\documentclass[nohyper,nobib,justified]{tufte-book}
\usepackage{amsmath}
\usepackage{amsthm}
\usepackage{nameref}
\usepackage{url}
\usepackage{xurl}
\usepackage[utf8]{inputenc}
\usepackage[backend=biber, natbib=true, style=numeric, refsegment=chapter]{biblatex}
\usepackage{amsfonts}
\usepackage{bm}
\usepackage{bbm}
\usepackage{amssymb}
\usepackage{graphicx}
\usepackage{subcaption}
\usepackage{accents}
\usepackage{mathtools}
\usepackage{booktabs}
\usepackage{units}
\usepackage{tikz,pgfplots,grffile}
\usepackage{xcolor}
\usepackage[most]{tcolorbox}
\usepackage[frozencache]{minted}
\makeatletter
\def\minted@error#1{\PackageWarning{minted}{#1}}
\makeatother
\usepackage{csquotes}
\tcbuselibrary{minted, listings}

\usepackage[ruled,vlined,algochapter]{algorithm2e}
\usepackage{algpseudocode}

\usepackage{xargs}
\usepackage{xspace}
\usepackage{colortbl}
\usepackage{xr}

\usepackage{hyperref}

\newcommand{\colorcode}[1]{{\begingroup\ttfamily\color{magenta}\nolinkurl{#1}\endgroup}}
\usepackage{enumitem}

\usepackage{xparse}
\NewDocumentCommand{\codeword}{v}{%
\texttt{\textcolor{magenta}{#1}}%
}
\makeatletter
\renewcommand{\fnum@listing}{Algorithm~\thelisting}
\makeatother
\newtcolorbox{definitionbox}{
  colback=blue!3!white,
  colframe=blue!50!black,
  boxrule=0.8pt,
  arc=4pt,
  left=8pt,
  right=8pt,
  top=6pt,
  bottom=6pt,
  fonttitle=\bfseries,
  coltitle=black,
  enhanced,
}
\DeclarePairedDelimiter{\rbr}{(}{)}
\newcommand{\defn}{\coloneqq}
\newcommand{\inv}[1]{#1^{-1}}
\newcommand{\pd}[3][]{%
    \frac{\partial^{#1}#2}{{\partial#3}^{#1}}
}
\newenvironment{hint}{\par\trivlist\item\emph{Hint:}\enspace\ignorespaces}{}
\usepackage{adjustbox}

\graphicspath{{./}}

\renewcommandx{\cite}[3][1={0pt},2={}]{\sidenote[][#1]{\fullcite[#2]{#3}}}

\usepackage{makeidx}
\makeindex

\setkeys{Gin}{width=\linewidth,totalheight=\textheight,keepaspectratio}

\usepackage{fancyvrb}
\fvset{fontsize=\normalsize}

\makeatletter
\renewcommand{\maketitlepage}{%
\begingroup%
\setlength{\parindent}{0pt}

{\fontsize{24}{24}\selectfont\textit{\@author}\par}

\vspace{1.75in}{\fontsize{36}{54}\selectfont\@title\par}

\vspace{0.5in}{\fontsize{14}{14}\selectfont\textsf{\smallcaps{\@date}}\par}

\vfill{\fontsize{14}{14}\selectfont\textit{\@publisher}\par}

\thispagestyle{empty}
\endgroup
}
\makeatother

\titlecontents{part}%
    [0pt]
    {\addvspace{0.5\baselineskip}}
    {\allcaps{Part~\thecontentslabel}\allcaps}
    {\allcaps{Part~\thecontentslabel}\allcaps}
    {}
    [\vspace*{0.5\baselineskip}]

\titlecontents{chapter}%
    [4em]
    {\addvspace{0.5\baselineskip}}
    {\contentslabel{2em}\textit}
    {\hspace{0em}\textit}
    {\qquad\thecontentspage}
    [\vspace*{0.25\baselineskip}]

\titlecontents{section}%
    [7em]
    {}
    {\contentslabel{3em}\textit}
    {\hspace{0em}\textit}
    {\qquad\thecontentspage}
    [\vspace*{0.1\baselineskip}]

\let\subsubsection\subsection
\let\subsection\section

\newcommand{\notessection}[1]{\section*{#1}}
\DeclareFixedFont{\ttb}{T1}{txtt}{bx}{n}{9} 
\DeclareFixedFont{\ttm}{T1}{txtt}{m}{n}{9}  

\usepackage{color}
\definecolor{deepblue}{rgb}{0,0,0.5}
\definecolor{deepred}{rgb}{0.6,0,0}
\definecolor{deepgreen}{rgb}{0,0.5,0}
\definecolor{backcolour}{rgb}{0.95,0.95,0.95}

\usepackage{listings}
\newcommand\pythonstyle{\lstset{
language=Python,
basicstyle=\ttm,
otherkeywords={self},             
keywordstyle=\ttb\color{deepblue},
emph={MyClass,__init__},          
emphstyle=\ttb\color{deepred},    
stringstyle=\color{deepgreen},
frame=tb,                         
backgroundcolor=\color{backcolour},
showstringspaces=false,            %
breakatwhitespace=false,
framexleftmargin=1em,
xleftmargin=1em,
}}

\newcommand\pythonstylenoborder{\lstset{
language=Python,
basicstyle=\ttm,
otherkeywords={self},             
keywordstyle=\ttb\color{deepblue},
emph={MyClass,__init__},          
emphstyle=\ttb\color{deepred},    
stringstyle=\color{deepgreen},                       
backgroundcolor=\color{backcolour},
showstringspaces=false,            %
breakatwhitespace=false,
framexleftmargin=1em,
xleftmargin=1em,
}}

\newcommand\gencodestyle{\lstset{
basicstyle=\ttm,
backgroundcolor=\color{backcolour},
showstringspaces=false,            %
breakatwhitespace=false,
framexleftmargin=1em,
xleftmargin=1em,
}}

\lstnewenvironment{python}[1][]
{
\pythonstyle
\lstset{#1}
}
{}

\lstnewenvironment{pythonnoborder}[1][]
{
\pythonstylenoborder
\lstset{#1}
}
{}

\lstnewenvironment{gencode}[1][]
{
\gencodestyle
\lstset{#1}
}
{}

\newcommand\pythoninline[1]{{\pythonstyle\lstinline!#1!}}

\theoremstyle{plain}
\newtheorem{theorem}{Theorem}[section]
\theoremstyle{definition}
\newtheorem{example}{Example}[section]
\newtheorem{definition}[theorem]{Definition}

\usepackage[framemethod=TikZ]{mdframed}
\usetikzlibrary{backgrounds,positioning,intersections, hobby, patterns, calc,cd,fit,quotes,math,decorations.pathmorphing, decorations.markings, shadows,shapes,automata, arrows}

\usepgfplotslibrary{patchplots,external,statistics}
\tikzstyle{boxCommentPlot}=[rectangle,very thick,draw=stanfordRed,rounded corners,align=center,font=\normalsize,inner sep=8pt]

\tikzcdset{arrow style=tikz, diagrams={>=To}}

\tikzset{
    invisible/.style={opacity=0},
    visible on/.style={alt={#1{}{invisible}}},
    alt/.code args={<#1>#2#3}{%
      \alt<#1>{\pgfkeysalso{#2}}{\pgfkeysalso{#3}}%
  }
}

\tikzstyle{block} = [draw, rectangle, minimum height=2.5em, minimum width=3.5em]
\tikzstyle{block1} = [draw, rectangle, minimum height=1.5em, minimum width=2.5em]
\tikzstyle{blockDyn} = [draw, rectangle, minimum height=2.5em, minimum width=3.5em, align=center, inner sep=10pt, thick, fill=white, copy shadow={draw=black,fill=black,opacity=1,shadow xshift=0.5ex,shadow yshift=-0.5ex}]
\tikzstyle{blockAlg} = [draw, rectangle, minimum height=1.5em, minimum width=2.5em, align=center, inner sep=10pt, thick]
\tikzstyle{sum} = [draw,circle]
\tikzstyle{arrow} = [thick]
\tikzstyle{input} = [coordinate]
\tikzstyle{output} = [coordinate]
\tikzstyle{pinstyle} = [pin edge={to-,thin,black}]

\DeclareMathOperator*{\argmax}{arg\,max}
\DeclareMathOperator*{\argmin}{arg\,min}

\newcommand{\mytilde}{\raise.17ex\hbox{$\scriptstyle\mathtt{\sim}$}} 

\renewcommand{\a}{\bm{a}}
\renewcommand{\v}{\bm{v}}
\newcommand{\e}{\bm{e}}
\newcommand{\x}{\bm{x}}
\newcommand{\bX}{\bm{X}}
\newcommand{\y}{\bm{y}}

\newcommand{\z}{\bm{z}}
\newcommand{\p}{\bm{p}}
\newcommand{\blam}{\bm{\lambda}}

\newcommand{\bu}{\bm{u}}
\newcommand{\bg}{\bm{g}}
\newcommand{\q}{\bm{q}}
\newcommand{\bv}{\bm{v}}
\newcommand{\bmu}{\bm{\mu}}

\newcommand{\m}{\bm{m}}
\newcommand{\w}{\bm{w}}
\newcommand{\bc}{\bm{c}}
\newcommand{\btheta}{\bm{\theta}}

\newcommand{\f}{f} 

\newcommand{\R}{\mathbb{R}}
\newcommand{\C}{\mathcal{C}}
\newcommand{\graph}{\mathcal{G}}
\newcommand{\edge}{\mathcal{E}}
\newcommand{\node}{\mathcal{V}}
\newcommand{\startnode}{q_{\mathrm{S}}}
\newcommand{\goalnode}{q_{\mathrm{G}}}

\newcommand{\X}{\mathcal{X}}
\newcommand{\U}{\mathcal{U}}
\newcommand{\dagname}{\text{DA\footnotesize{GGER}}}
\newcommand{\bmx}[2][1]{%
    \renewcommand*{\arraystretch}{#1}%
        \begin{bmatrix*}[c]%
            #2%
        \end{bmatrix*}%
    \renewcommand*{\arraystretch}{1.0}%
}
\newcommand{\tran}[1]{#1^\mathsf{T}}
\DeclarePairedDelimiter{\norm}{\lVert}{\rVert}
\newcommand{\stateNoise}{Q}
\newcommand{\measNoise}{R}
\newcommand{\figlabel}[1]{\addtocounter{figure}{-1}\refstepcounter{figure}\label{#1}}

\newcommand{\pubauthors}{Daniele Gammelli, Joseph Lorenzetti, Katie Luo,
Gioele Zardini and Marco Pavone}
\newcommand{\pubnotice}{This material will be published by Cambridge University
Press as Principles of Robot Autonomy by \pubauthors. This pre-publication
version is free to view and download for personal use only. Not for
re-distribution, re-sale or use in derivative works.
\textcopyright\ \pubauthors\ \the\year.}

\fancypagestyle{plain}{%
  \fancyhf{}%
  \setlength{\footskip}{3.6\baselineskip}%
  \fancyfoot[L]{\parbox[t]{\textwidth}{\scriptsize\setstretch{1}\raggedright\pubnotice}}%
}

\title{Principles of Robot\\ Autonomy}
\author{Daniele Gammelli, Joseph Lorenzetti, Katie Luo, Gioele Zardini, Marco Pavone}
\date{\today}

\usepackage{comment}
\usepackage{silence}
\usepackage{todonotes}
\setuptodonotes{inline}
\usepackage[noabbrev,capitalize]{cleveref}
\newcommand{\bel}{\text{bel}}
\newcommand{\belpred}{\overline{\bel}}

\newcommand{\controldim}{m}

\newcommand{\dynmodel}{f}
\newcommand{\dynJac}{F}
\renewcommand{\d}{d}
\newcommand{\definedas}{\coloneqq}

\newcommand{\expected}[2]{\mathbb{E}_{#1}\left[#2 \right]}
\DeclareMathOperator{\Exp}{Exp}

\newcommand{\given}{\mid}

\newcommand{\hamiltonian}{H}

\newcommand{\controlspace}{\mathcal{U}}

\newcommand{\lagrangian}{L}

\newcommand{\measmodel}{h}
\newcommand{\measJac}{H}
\newcommand{\maximize[1]}{\underset{#1}{\text{maximize}}\quad}
\newcommand{\minimize[1]}{\underset{#1}{\text{minimize}}\:\quad}

\newcommand{\Oc}{\mathcal{O}}

\newcommand{\outputdim}{p}

\newcommand{\particleset}{\mathcal{P}}

\newcommand{\pluseq}{\mathrel{+}=}
\newcommand{\reals}{\mathbb{R}}

\newcommand{\statespace}{\mathcal{X}}
\newcommand{\statedim}{n}

\newcommand{\mysidenote}[1]{\sidenote[][\baselineskip]{#1}}
\newcommand{\subjectto}{\text{subject to} \quad}

\newcommand{\tup}[1]{\left(#1\right)}

\renewcommand{\u}{\bm{u}}
\newcommand{\vCol}[1]{\left[{#1}\right]^\top}

\begin{document}

\maketitle

\newpage
\thispagestyle{empty}
\begin{fullwidth}
\setlength{\parindent}{0pt}
\setlength{\parskip}{\baselineskip}
\vspace*{\fill}

\noindent \pubnotice

\noindent Posted on arXiv with the written permission of Cambridge University Press.
The definitive version will be published by Cambridge University Press.

\noindent \emph{Third-party material.} Figures and other material credited to
external sources remain subject to the copyright of their respective owners;
permission for reuse must be sought directly from those rights holders.
\end{fullwidth}

\frontmatter
\chapter{Preface}
\notessection{What is robot autonomy?}
Depending on one's imagination, the term \emph{robot} often evokes one of two extremes.
For some, it conjures visions of fantastical do-it-all android butlers from various sci-fi futures, to this day brought to life only on the page or on the screen. 
These helpful servants are defined by their ability to deftly handle anything the world throws at them.
Endowed with human (or even super-human!) levels of artificial intelligence, our aspirational creations are unbounded in creativity and resourcefulness for problem-solving. 
For others, the word \emph{robot} refers instead to an assortment of mechatronic tools existing today, exemplified by the robotic arms and manipulators that have powered industry since the latter half of the twentieth century.
This more practically-grounded interpretation is characterized by precision and control, not only in the machines themselves but also in the carefully structured environments in which they operate.
Motion is optimized, and every element is arranged to occur exactly when and where it should.
The production line is designed to admit no surprises, and in the rare event that there are, human supervisors intervene to restore nominal operation.

In recent decades, however, a middle ground between these two extremes has begun to take shape. 
As illustrated in Figure~\ref{fig:robot_autonomy_examples}, contemporary systems---including the growing presence of self-driving cars, follow-me aerial drones, legged and humanoid robots, and free-flying space robots---demonstrate capabilities that lie between rigid industrial automation and fully general artificial agents.
In the course of their operation, such robots inevitably encounter novel situations and unanticipated combinations of tasks and constraints. 
These cannot be exhaustively specified in advance, nor can they rely on continuous human supervision to resolve every contingency. 
Instead, their functionality depends on an ability to act independently in the face of uncertainty and change.
In this sense, a defining feature of these systems is that they must exhibit some degree of \emph{autonomy}.

\emph{Robot autonomy} refers to a robot’s capacity to perceive its environment and to act in pursuit of its objectives without direct external guidance, particularly from human operators.
Central to this capability is the ability to make decisions based on an evolving understanding of the environment and its relation to the robot's goals.
This understanding is informed by a continual stream of sensory data, which is itself shaped by the robot's actions as it moves through and interacts with its surroundings.
This feedback loop connecting perception and action is what enables robots to operate in unstructured and uncertain scenarios, where the ability to react and replan is paramount.

\begin{figure}
    \centering
  \newcommand{\robotimg}[2]{%
    \includegraphics[width=\linewidth,height=0.666\linewidth,clip,trim=#1]{#2}%
  }
    
    \begin{subfigure}[t]{0.3\textwidth}
        \centering
      \robotimg{42 0 42 0}{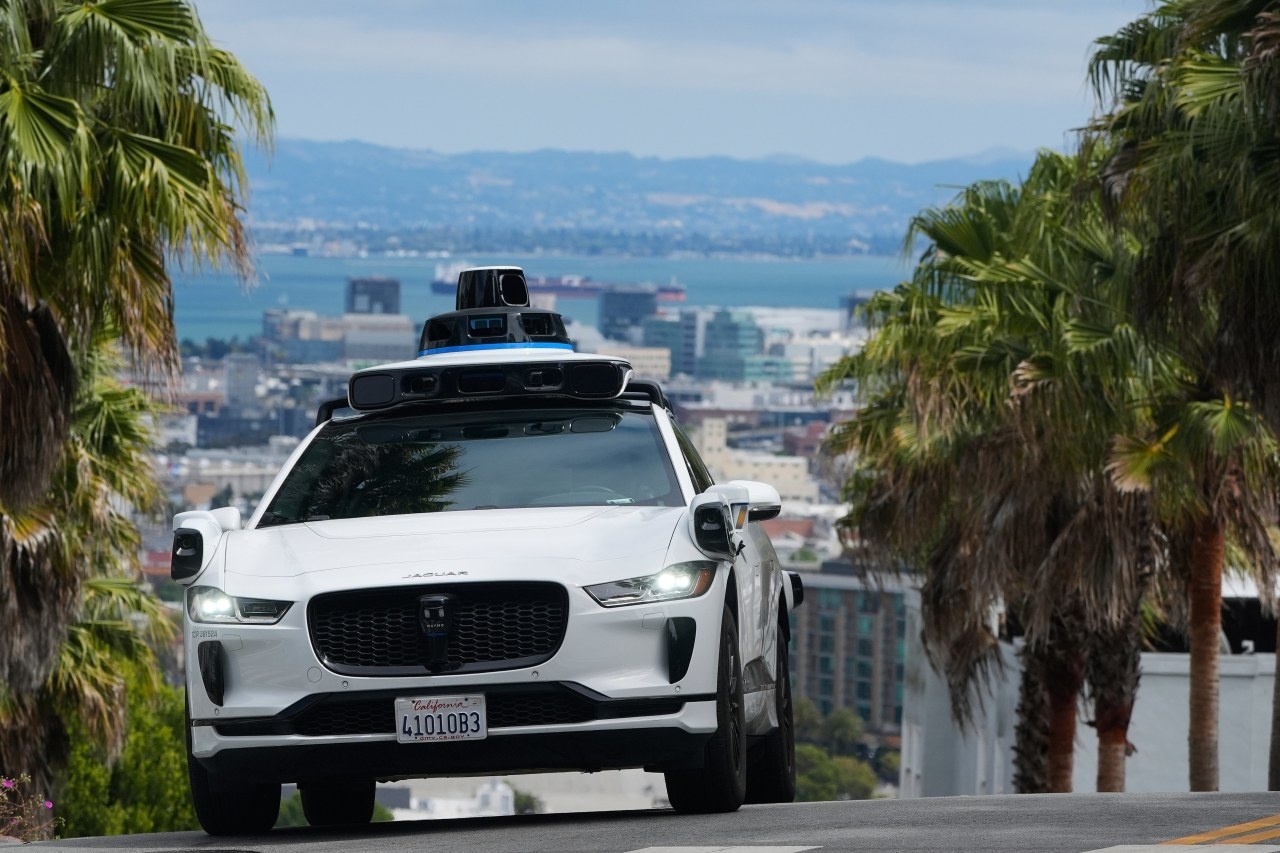}
        \caption{Waymo self-driving car.}
    \end{subfigure}
    \hfill
    \begin{subfigure}[t]{0.3\textwidth}
        \centering
      \robotimg{0 50 0 50}{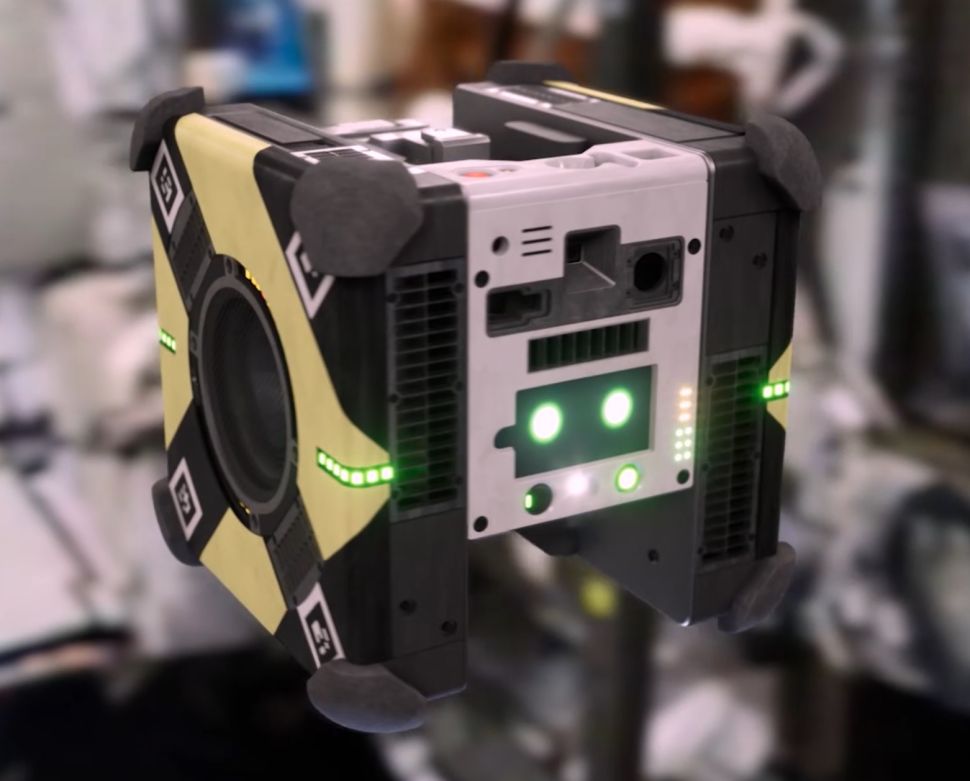}
        \caption{NASA's Astrobee robot.}
    \end{subfigure}
    \hfill
    \begin{subfigure}[t]{0.3\textwidth}
        \centering
      \robotimg{0 41 0 41}{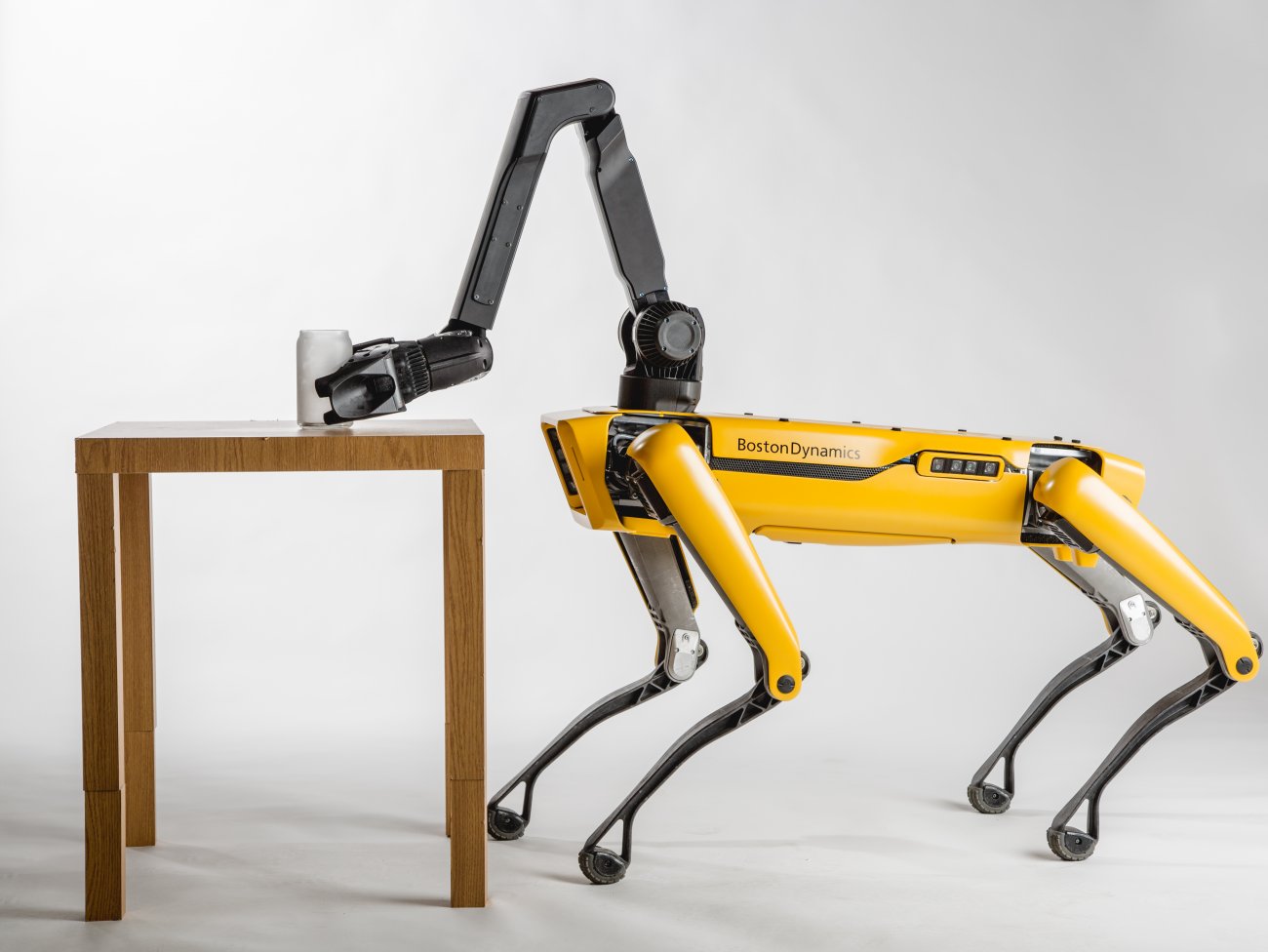}
        \caption{Boston Dynamics SpotMini.}
    \end{subfigure}

    \vspace{0.5em}

    \begin{subfigure}[t]{0.3\textwidth}
        \centering
      \robotimg{120 0 120 0}{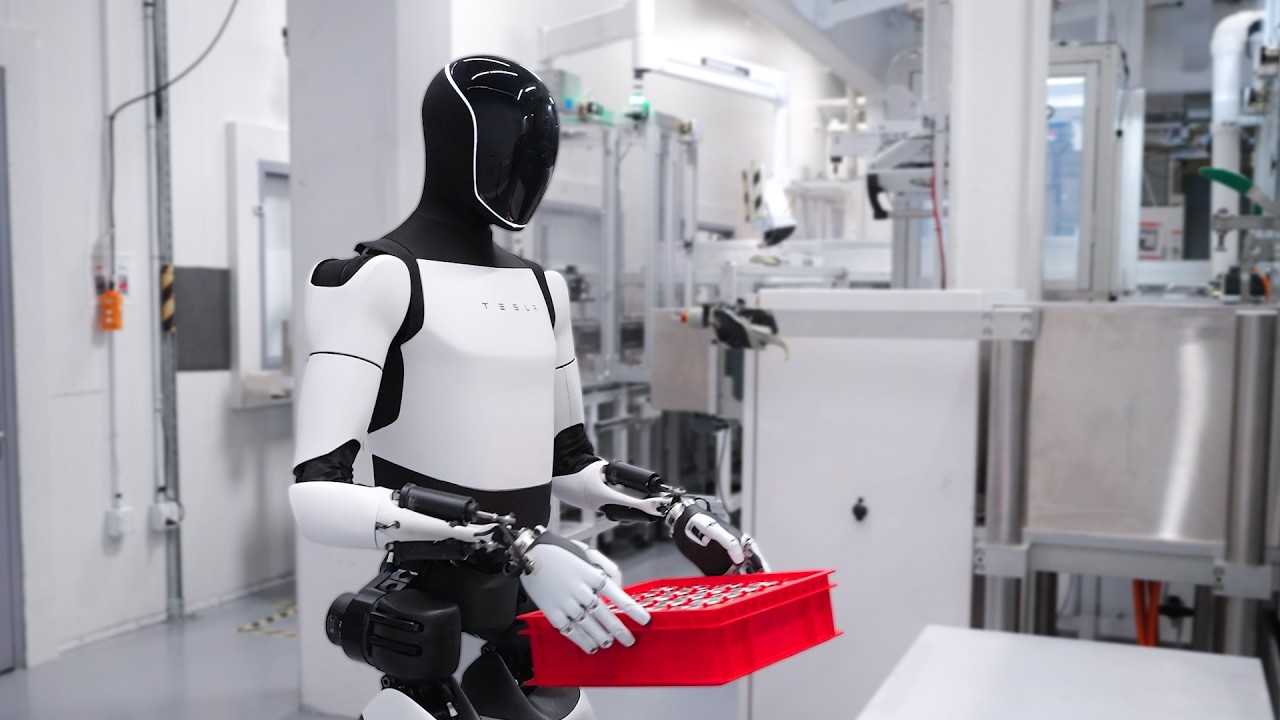}
        \caption{Tesla Optimus humanoid robot.}
    \end{subfigure}
    \hfill
    \begin{subfigure}[t]{0.3\textwidth}
        \centering
      \robotimg{0 0 0 0}{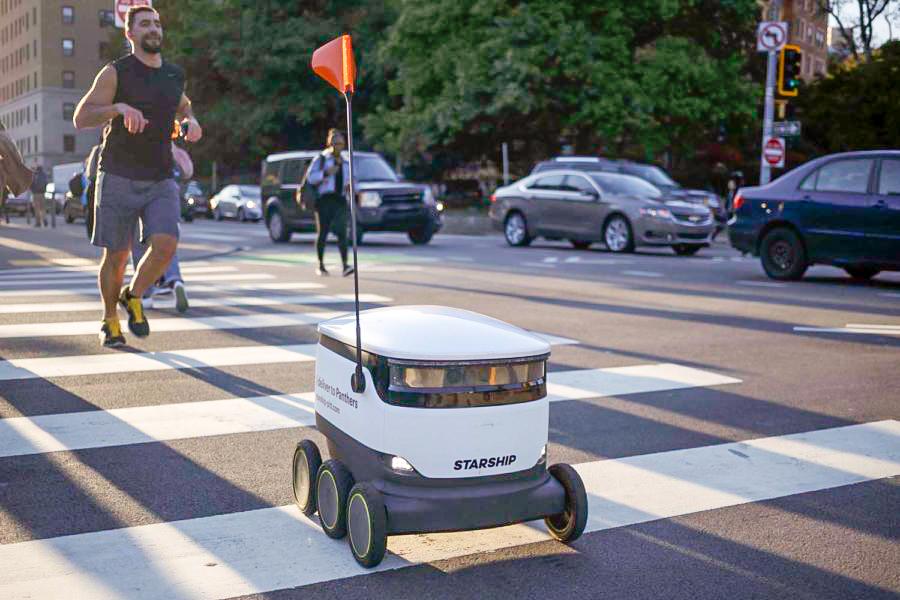}
        \caption{Starship delivery robot.}
    \end{subfigure}
    \hfill
    \begin{subfigure}[t]{0.3\textwidth}
        \centering
      \robotimg{0 2 0 2}{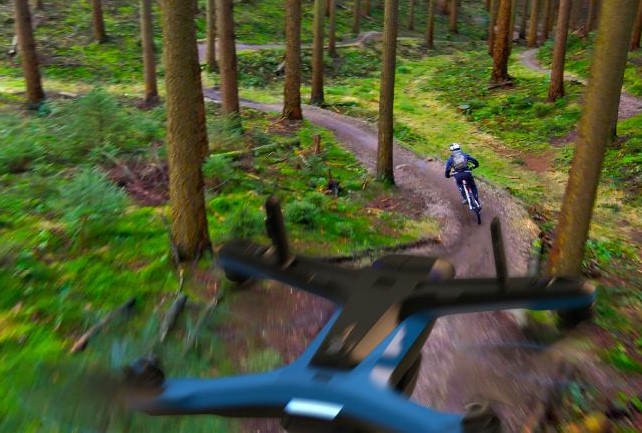}
        \caption{Skydio follow-me drone.}
    \end{subfigure}

    \caption{Autonomous robots deployed around the world (and in the skies above!) today.}
\end{figure}
\figlabel{fig:robot_autonomy_examples}

\notessection{Autonomy vs. Automation vs. Artificial Intelligence}
The notion of robot autonomy developed in this textbook overlaps with, but should not be conflated with, the broader and more established field of \emph{automation}, nor the rapidly evolving field of \emph{embodied artificial intelligence (AI)}. 
At a high level, what distinguishes autonomy as a subset of automation technologies is its emphasis on active decision making.
More generally, \emph{automation} refers to the reduction or elimination of human involvement in the execution of tasks through a combination of different technologies.
In many cases, it is advantageous to constrain the task domain or impose additional structure on the environment so as to eliminate the need for autonomous decision-making during operation.
This is not to say that automation technologies are incapable of handling variation; rather, such variations are typically anticipated during design, with predefined responses or safeguards incorporated in advance.
For instance, while the exact motor commands executed by a robotic welding system may not be explicitly prescribed—since the required levels of precision often necessitate closed-loop (feedback) control—the surrounding factory environment is carefully engineered in conjunction with the robot to ensure reliability and efficiency.
This co-design of robot and workspace is a hallmark of classical automation.
Beyond industrial manufacturing, well-established examples of automation include visual servoing, process control, and Computer Numerical Control (CNC) machine tools.
In contrast, autonomous robotic systems are generally designed under weaker assumptions about environmental structure.
Consequently, they must be endowed with the ability to understand their surroundings, reason, and implement novel courses of action.

On the other hand, the term \emph{artificial intelligence} refers to the broad goal of creating systems capable of perceiving, reasoning, planning, and problem-solving in ways that parallel, or exceed, human capabilities.
Within robotics, AI provides many of the computational tools that enable autonomy.
Classically, these have taken the form of algorithms that perform structured reasoning or search over well-defined spaces of possibilities.
More recently, however, the term AI has become closely associated with data-driven methods, particularly machine learning (ML), in which systems acquire capabilities from data rather than through explicit programming.
\emph{Embodied} AI, also referred to as \emph{physical} AI, extends this paradigm by grounding intelligence in a physical instantiation, or embodiment, that interacts with the world rather than existing solely in a virtual or abstract form.
This perspective has recently been accompanied by a shift toward \emph{end-to-end} system design, where the classical boundaries between perception, estimation, planning, and control are increasingly blurred or replaced by unified models.
Such models learn direct mappings from sensory inputs to actions, high-level decisions, or even predictions of future outcomes.
Prominent examples include \emph{vision-language-action} (VLA) models, which integrate multimodal input streams with action generation, and emerging \emph{world models}, which learn to predict the future evolution of the environment from rich sensory data such as images and video, often at internet scale.
As a result, modern approaches to robot autonomy span a spectrum, ranging from structured, model-based pipelines to fully learned, end-to-end systems.
In this textbook, we present both human-designed algorithms and machine-learned approaches to provide a unified and comprehensive view of the available methods---and because, in practice, it is often valuable to combine these approaches to various degrees depending on the application at hand.

\notessection{How is robot autonomy achieved?}

Robot autonomy is in its essence an interdisciplinary endeavor. 
While many areas of science and engineering often benefit from the exchange of ideas across fields, robot autonomy is fundamentally defined by the integration of techniques from multiple domains.
It is, by its very nature, a synthesis of computer vision, estimation theory, artificial intelligence, and control theory, just to name a few.
The science and practice drawn from each of these domains form critical components of the modern autonomy \emph{stack}, a term which in itself emphasizes the multifaceted, multi-component structure through which a robot perceives, reasons about, and interacts with its environment.
At a high level, robot autonomy can be understood as requiring three fundamental capabilities:
\begin{itemize}
    \item \emph{See:} A robot uses sensors, such as cameras, laser scanners, global positioning system measurements, and motor feedback, to collect raw data about its surroundings. 
    These signals are processed to extract semantic and geometric information of the robot's state and its environment.
    \item \emph{Think:} Building on these local perceptual signals, a robot first synthesizes sensor data over time into a coherent global estimate of its state in the environment, typically through filtering, localization, and mapping.
    On top of this estimate, it performs higher-level reasoning and decision-making, selecting behaviors that advance both immediate objectives and longer-horizon mission goals.
    \item \emph{Act:} A robot executes these higher-level decisions by translating them into physically realizable motions through trajectory generation and motion planning.
    These planned motions are then realized at the actuator level through control laws that are typically feedback-based and closed-loop, enabling robust execution despite disturbances, model mismatch, and uncertainty.
\end{itemize}

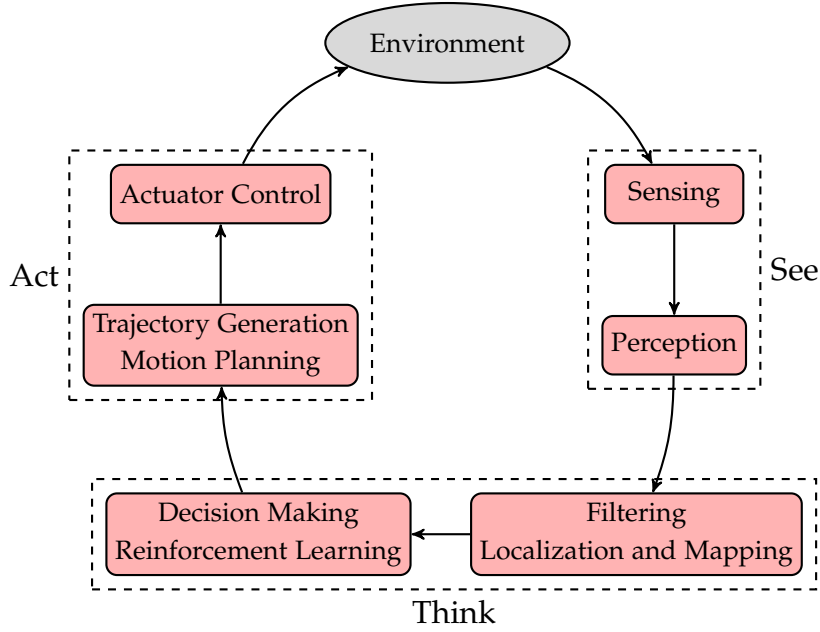
\begin{figure}[t]
    \begin{center}
    \begin{tikzpicture}[node distance=0cm,>=stealth',auto]
      \tikzstyle{part}=[draw=black,rounded corners,minimum height=2.2em,minimum width=5.2em,thick,fill=red!30]
      \tikzstyle{environment} = [shape=ellipse, draw=black, minimum height=3em, minimum width=3em, thick, fill=gray!30]
    
      \node[environment](env){Environment};
      
      \node[part, above of=env, xshift=3cm, yshift=-2cm](sensing){Sensing};
      \node[part, above of=env, xshift=3cm, yshift=-4cm, align=center](info){Perception};
      \node[draw, thick, dashed, inner sep=5pt, fit=(info) (sensing), label={[label distance=0.0cm]0:\Large See}] (see) {};
      
      \node[part, above of=env, xshift=2.5cm, yshift=-6.5cm, align=center](localize){Filtering  \\ Localization and Mapping};
      \node[part, above of=env, xshift=-2.5cm, yshift=-6.5cm, align=center](plan){Decision Making  \\ Reinforcement Learning};
      \node[draw, thick, dashed, inner sep=5pt, fit=(localize) (plan), label={[label distance=0.0cm]270:\Large Think}] (think) {};
      
      \node[part, above of=env, xshift=-3cm, yshift=-4cm, align=center](execute){Trajectory Generation \\ Motion Planning};
      \node[part, above of=env, xshift=-3cm, yshift=-2cm, align=center](actuate){Actuator Control};
      \node[draw, thick, dashed, inner sep=5pt, fit=(execute) (actuate), label={[label distance=0.0cm]180:\Large Act}] (act) {};

      \draw[->,thick] (env) to[bend left=20]
        node[pos=0.5,sloped,above=6pt,inner sep=1pt,fill=white,fill opacity=0.9,text opacity=1]
          {} (sensing);
      \draw[->,thick] (sensing) to[bend left=0]
        node[pos=0.5,sloped,above=6pt,inner sep=1pt,fill=white,fill opacity=0.9,text opacity=1]
          {} (info);
      \draw[->,thick] (info) to[bend left=10]
        node[pos=0.5,sloped,above=6pt,inner sep=1pt,fill=white,fill opacity=0.9,text opacity=1]
          {} (localize);
      \draw[->,thick] (localize) to[bend left=0]
        node[pos=0.5,sloped,above=6pt,inner sep=1pt,fill=white,fill opacity=0.9,text opacity=1]
          {} (plan);
      \draw[->,thick] (plan) to[bend left=10]
        node[pos=0.5,sloped,above=6pt,inner sep=1pt,fill=white,fill opacity=0.9,text opacity=1]
          {} (execute);
      \draw[->,thick] (execute) to[bend left=0]
        node[pos=0.5,sloped,above=6pt,inner sep=1pt,fill=white,fill opacity=0.9,text opacity=1]
          {} (actuate);
      \draw[->,thick] (actuate) to[bend left=20]
        node[pos=0.5,sloped,above=6pt,inner sep=1pt,fill=white,fill opacity=0.9,text opacity=1]
          {} (env);
    \end{tikzpicture}
    \end{center}
    \caption{The \emph{See-Think-Act} cycle. In the \emph{See} stage, raw sensor signals are processed into local semantic and geometric observations. In the \emph{Think} stage, these observations are integrated over time to form a global estimate of the robot's state and environment, which supports higher-level decision-making. In the \emph{Act} stage, decisions are translated into feasible trajectories and motion plans, and executed through actuator-level feedback control.}
\end{figure}
\figlabel{fig:see-think-act}

The \emph{See–Think–Act} paradigm provides a high-level blueprint for robot autonomy and is conceptually aligned with established human decision-making cycles such as the observe–orient–decide–act (OODA) loop and the plan–do–check–act (PDCA) cycle, used in domains ranging from business to military strategy. 
Most modern autonomy stacks adopt some form of \emph{See–Think–Act} as the foundation for their behavior, although the specific implementation can vary widely depending on the intended application. 
Indeed, designing this stack is a central responsibility of the roboticist.
In practice, the stages of this pipeline are often combined or treated implicitly. 
This is particularly evident in emerging end-to-end approaches, where the explicit boundaries between \emph{See}, \emph{Think}, and \emph{Act} are increasingly blurred. 
However, this shift does not eliminate the underlying functional roles these components play; rather, it reflects a different way of organizing them, where, ultimately, an autonomous robot must still possess each of these capabilities to some degree in order to operate effectively.

The steps of \emph{See-Think-Act} are typically presented as a loop, where the actions executed by a robot influence the observations it receives at the next cycle. 
Ideally, these observations are consistent with the robot’s expectations under its current plan. 
Closing the loop, however, is essential for providing corrective feedback, enabling the robot to adapt to previously unseen or unexpected changes in its environment. 
In this way, updates to situational understanding lead to revisions of the current plan, which in turn inform the selection of the next action, ultimately generating new observations that initiate the cycle once again.

In practice, however, this abstract loop is rarely implemented as a single, sequential process. 
Instead, it is realized as a network of asynchronously operating components that run at different rates while continually incorporating the most up-to-date information available.
Consider the example of a quadrotor drone. 
Object detection modules may process each incoming camera frame at around 30Hz and feed into object trackers operating at a similar rate, while more computationally intensive tasks, such as maintaining a 3D map of the environment, may update more slowly, on the order of 2Hz. 
The most recent outputs from these \emph{See} components can then inform a trajectory planner running at approximately 10Hz, which generates motion plans that account for both static and dynamic obstacles. 
At the lowest level, a flight controller tracks the planned trajectory and operates at a much higher frequency, often around 200Hz, using high-rate inertial measurements from accelerometers and gyroscopes that may reach 1000Hz.
Each component in this system consumes inputs either directly from sensors or from other components, and produces outputs that are used elsewhere in the stack or ultimately translated into actuation commands. 
Rather than viewing autonomy as a single closed loop, it is therefore more useful to think of the system as a graph: nodes correspond to functional components, and edges represent the flow of information between them. 
To make this abstraction concrete, and in the accompanying programming exercises, we will assume that communication between components is handled through the Robot Operating System (ROS), a widely used framework in both academic research and industry practice.

\notessection{Learning objectives}
This book has evolved from the course notes for the series of classes \emph{Principles of Robot Autonomy}, taught annually at Stanford University since Winter 2017. 
The course was initiated in response to the growing deployment of self-driving cars, drones, and mobile robots more broadly, signaling a transition of robot autonomy from a primarily academic pursuit to a collection of mature, field-tested tools and techniques on which practitioners can depend.
The aim of this book is therefore to equip the reader with a principled understanding of the theoretical, algorithmic, and practical aspects underlying modern robot autonomy.
This is arguably an ever-moving target, particularly on the implementation side, as software tools and system abstractions continue to evolve.
Nevertheless, because the performance of an autonomy stack depends critically on effective system integration, these practical considerations are essential and cannot be overlooked; accordingly, the \emph{Robot Autonomy Software} chapter provides an orienting overview of the software principles and ROS-based implementation paradigms that underpin modern robotic systems.

At the same time, many of the fundamental principles for endowing mobile autonomous robots with perception, planning, and decision-making capabilities are now well established.
The chapters that follow present the core techniques in modeling and control, motion planning and trajectory optimization, object detection and tracking, state estimation, simultaneous localization and mapping (SLAM), deep learning for perception and decision making, reinforcement learning, imitation learning, and more.
These methods are supported by mathematical tools drawn from optimization theory, geometry and coordinate transformations, filtering theory, machine learning, statistical learning theory, deep learning, and broader artificial intelligence.
By developing familiarity with these foundational tools, the reader will gain a coherent understanding of the broader autonomy stack, and will be well prepared to contribute new methods and ideas to the evolving field of robot autonomy.

\notessection{Structure of the book}
Before proceeding, let us outline the structure of the book and how the material is organized.
We begin with the \emph{Robot Autonomy Software} chapter, which provides an overview of the software principles and tools that underpin modern robotic systems. 
In particular, we introduce ROS and the computational abstractions used throughout the text. 
While not strictly part of the autonomy stack itself, this material serves as an essential foundation for understanding how the algorithms presented in later chapters are implemented in practice.
The book is then organized into four parts, which reflect key aspects of the \emph{See–Think–Act} paradigm illustrated in Figure~\ref{fig:see-think-act}. 

Part~I, \emph{Robot Motion Planning and Control}, focuses on the \emph{Act} portion of the autonomy stack, covering modeling, control, trajectory generation, and motion planning.
We begin here deliberately. 
By grounding the discussion in how robots ultimately produce motion, we establish a concrete understanding of the final outcome of autonomy, introduce core notation and system models, and provide a pedagogical foundation upon which higher-level reasoning can be built.

Part~II, \emph{Robot Perception}, corresponds to the \emph{See} stage. 
It develops the tools required for extracting meaningful information from raw sensor data, including camera modeling, geometric perception, and modern learning-based approaches for detection and recognition. 

Part~III, \emph{Robot Localization and Mapping}, bridges perception and reasoning by addressing state estimation, filtering, and SLAM, forming a critical component of the \emph{Think} stage. 

Finally, Part~IV, \emph{Robot Decision Making}, focuses on higher-level reasoning and planning under uncertainty, including sequential decision-making, dynamic programming, reinforcement learning, and imitation learning.

The book concludes with a \emph{Prospects} chapter, which reflects on the broader trajectory of the field. There, we revisit the \emph{See–Think–Act} paradigm and discuss emerging directions.

A central emphasis of this book is learning by doing. 
To that end, most chapters are accompanied by interactive Python implementations in Jupyter Notebooks that allow the reader to experiment directly with the concepts and algorithms presented in the text. 
These notebooks are designed to complement the theoretical material, providing hands-on intuition and practical experience. 
In addition, each chapter includes exercises of varying difficulty, as well as pointers to further reading that highlight extensions, open problems, and connections to current research.
Interactive notebooks and exercises are available online at:

\begin{center}
\texttt{https://github.com/StanfordASL/pora-exercises.git}
\end{center}

Ultimately, each chapter is structured to include three main components: core conceptual and mathematical content, accompanying notebooks and exercises for active exploration, and references for deeper study. 
Together, these elements are intended to support both a principled understanding and practical proficiency in robot autonomy.

We also release a regularly maintained website that includes additional teaching materials, updates, and resources related to the book.
The official website can be found at:
\begin{center}
\texttt{https://porabook.com}
\end{center}

\notessection{Acknowledgments}
These notes accompany and are based largely on the content of the courses \emph{AA174A / AA274A: Principles of Robot Autonomy I} and \emph{AA274B: Principles of Robot Autonomy II} at Stanford University. 
We would therefore like to acknowledge the students who have taken these courses and provided useful feedback since their initial offering in 2017. 
We also reserve special acknowledgements for the course assistants who were instrumental in developing and refining the course material, and in particular Benoit Landry and Edward Schmerling, who were instrumental in developing the first iteration of the courses. 
We would also like to acknowledge Yue Wang and Yan Wang for their contributions to the perception chapters, especially for developing the initial structure and organization, much of which informed and shaped the present version.
We are also grateful to the members of the Autonomous Systems Lab at Stanford University for many insightful discussions, careful proofreading, and helpful feedback on the material, all of which have helped shape and refine this book.

\tableofcontents

\chapter{Robot Autonomy Software}
\label{ch:software-note}
Modern autonomous robots are built as collections of interacting software components that perceive the environment, reason about goals, and execute actions in real time.
As discussed in the \emph{Preface}, autonomy is not defined by any single algorithm, but by the integration of many components across the \emph{See–Think–Act} cycle. 
Turning that conceptual paradigm into a functioning robotic system requires a practical software substrate through which sensing, estimation, planning, and control modules can communicate and operate together.
These components must operate concurrently, exchange information reliably, and adapt to changing conditions, all while meeting real-world constraints on timing, safety, and robustness. 
Coordinating such systems is as much a software challenge as it is an algorithmic or mechanical one.

In practice, most contemporary robotic systems are developed on top of a shared software infrastructure that helps manage this complexity. 
In this brief note, we provide a high-level overview of the \emph{Robot Operating System (ROS)}\cite{Joseph2018}\cite[\baselineskip]{QuigleyGerkeyEtAl2015}, which has emerged as the de facto standard software framework for robotics research, education, and many industrial applications. 
Despite its name, ROS is not an operating system in the traditional sense. 
Rather, it is a \emph{middleware} framework: a collection of tools, libraries, and design conventions that support modular, distributed robot software development.

ROS is best understood as a common language and ecosystem for robotics software rather than a fixed or complete solution. 
It provides abstractions for communication, data sharing, configuration, and system integration that allow developers to focus on higher-level autonomy and control algorithms. 
At the same time, the robotics software landscape is evolving rapidly, and ROS itself continues to change, with new versions, extensions, and alternatives appearing regularly. 
As such, this chapter is not intended as a comprehensive guide or tutorial, but rather as an orienting overview of the dominant software paradigms in modern robotics.
Readers are encouraged to view this material as a snapshot of current practice and to supplement it with the extensive and continually updated resources available online.

\notessection{Why Robot Software Is Different}

At first glance, robot programming may appear to be a straightforward application of classical software engineering. 
However, robots impose a unique set of constraints that distinguish them from most conventional software systems. 
A robot must simultaneously interface with numerous hardware devices, process streaming sensor data, make decisions under uncertainty, and generate real-time control signals, all while operating in a physical world that is only partially observable and constantly changing.

Unlike many traditional software applications, robot software must coordinate many heterogeneous components operating concurrently. 
A single robot may include cameras, lidars, inertial sensors, encoders, motors, and communication interfaces, each producing or consuming data at different rates. 
As a result, robot software must support:
\begin{itemize}
    \item \emph{Multitasking:} autonomous robots inherently require concurrent execution. 
    Multiple sensing, estimation, planning, and control processes must run in parallel, often at different frequencies, while exchanging data in a timely and reliable manner.
    
    \item \emph{Low-level device control:} robots must interface directly with a wide variety of hardware devices using communication protocols such as GPIO, USB, SPI, and others. 
    This requires being able to support multiple programming languages, such as C++ and Python.

    \item \emph{High-level object-oriented abstractions:} as autonomy stacks grow in size and complexity, software modularity becomes essential. 
    Encapsulation, inheritance, and code reuse enable developers to build complex robotic systems from reusable components.
    
    \item \emph{Community libraries and shared infrastructure:} given the breadth of robotics as a field, it is impractical for every developer to reimplement common algorithms from scratch.
    Access to third-party libraries and an active user community dramatically accelerates development and encourages standardization.
\end{itemize}

These requirements motivate the need for a software middleware specifically designed for robotics, one that bridges the gap between hardware-level control and high-level autonomy algorithms.

\notessection{A Brief History of ROS}

Before the development of ROS, robotics software was largely fragmented. 
Code written for one robot or laboratory was often difficult to reuse elsewhere, hindering collaboration and slowing progress across the field. 
In 2007, early versions of ROS emerged from the Stanford AI Robot (STAIR) project, motivated by the vision of a free and open-source robotics framework that would promote collaboration, reuse, and rapid experimentation.

\begin{marginfigure}
    \centering 
    \includegraphics[width=0.9\linewidth]{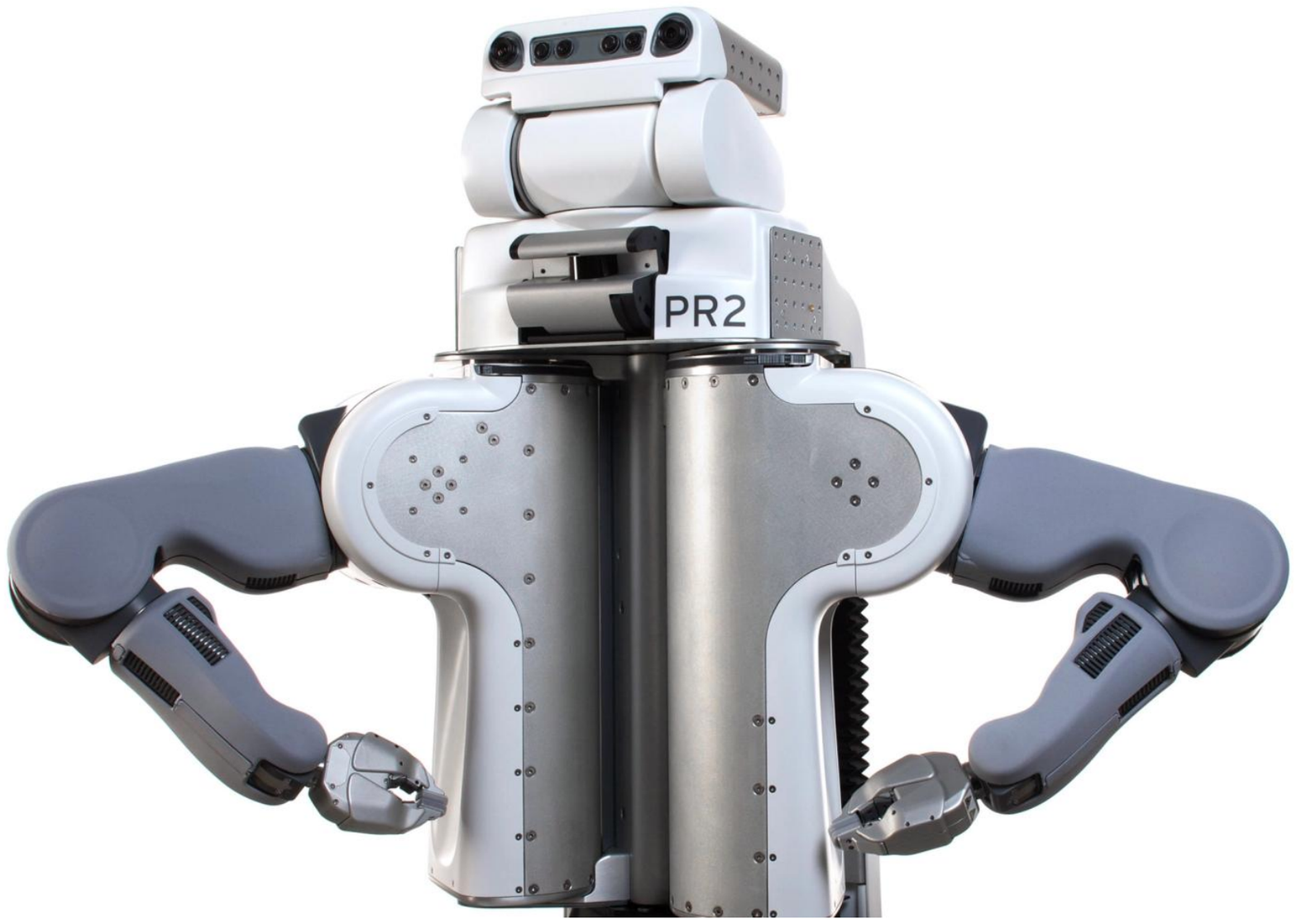}
    \caption{The PR2 robot, one of the first platforms to be widely supported by ROS.}
    \label{fig:pr2-robot}
\end{marginfigure}

This vision was further realized when the startup company Willow Garage, led by Scott Hassan, assumed stewardship of the project. 
Under this guidance, ROS evolved into a standardized development platform featuring a rich set of tools, libraries, and conventions.
The release of ROS~0.4 in 2009 coincided with the deployment of the PR2 mobile manipulation robot~(\cref{fig:pr2-robot}), several units of which were distributed to universities to catalyze collaborative development.
ROS~1.0 followed in 2010.

Oversight of ROS transitioned in 2012 to the Open Source Robotics Foundation (OSRF), supporting its development, distribution, and adoption. 
The first long-term support release, ROS Indigo Igloo, appeared in 2014. 
In 2017, OSRF was rebranded as Open Robotics, reflecting its broader mission of supporting open-source robotics software beyond ROS itself. 
Today, ROS is widely regarded as the closest thing to an industry standard for robot software, underpinning systems ranging from research prototypes to commercial autonomous platforms.

As robotics applications expanded beyond research settings into safety-critical and real-time domains, important limitations of ROS~1 became apparent.
In particular, ROS~1 was not designed with real-time guarantees, deterministic communication, or multi-robot deployments over unreliable networks as first-class concerns.
These challenges motivated the development of ROS~2, a major architectural redesign that builds on modern middleware standards such as the Data Distribution Service (DDS).

ROS~2 introduces native support for real-time communication, improved security, better handling of distributed and multi-robot systems, and more explicit control over quality-of-service parameters.
As such, ROS~2 reflects the evolving needs of the robotics community and is increasingly adopted in applications where reliability, scalability, and real-time performance are critical.

\notessection{Design Philosophy of ROS}

The architectural choices embodied in ROS reflect a deliberate set of design goals shaped by the practical realities of building large-scale robotic systems.
Rather than attempting to provide a monolithic, all-encompassing framework, ROS adopts a minimalist philosophy that emphasizes flexibility, reuse, and composability.
ROS' philosophy can be summarized in the following main principles:

\paragraph{Peer-to-peer.}
At its core, ROS is designed around a \emph{peer-to-peer} communication model.
Computation is distributed across multiple processes, potentially running on different machines, which discover and connect to one another dynamically at runtime.
This approach avoids centralized bottlenecks and allows robotic systems to scale naturally across heterogeneous computing resources, such as embedded controllers, edge devices, and cloud servers.

\paragraph{Multi-language.}
Another key design principle is \emph{language neutrality}.
Robotic systems often benefit from combining high-performance compiled code for time-critical components with higher-level scripting languages for rapid development and experimentation.
ROS supports this hybrid approach by defining communication interfaces independently of any specific programming language, allowing components written in different languages to interoperate seamlessly.

\paragraph{Tools-based.}
ROS is also explicitly \emph{tools-based}.
Instead of embedding all functionality within a single runtime environment, ROS provides a collection of command-line tools and graphical utilities for tasks such as launching systems, inspecting communication graphs, logging data, replaying experiments, and visualizing internal state.
This modular tooling approach simplifies debugging and experimentation, enabling developers to inspect and modify a running system without recompilation or redeployment.

\paragraph{Thin.}
Finally, ROS is intentionally \emph{thin}.
The middleware itself provides communication, configuration, and coordination mechanisms, but encourages core algorithms and device drivers to be implemented as standalone libraries with minimal dependence on ROS-specific infrastructure.
This separation improves testability, promotes code reuse outside of ROS, and reduces long-term maintenance costs as software systems evolve.

\paragraph{Free and open-source.}
ROS is released under the BSD license, which allows nearly unrestricted use, modification, and distribution, making it freely available for both academic and commercial use.
As such, ROS has fostered a vibrant ecosystem of users and contributors who share code, documentation, and best practices.

\medskip
Taken together, these principles position ROS not as a rigid framework, but as an enabling substrate upon which diverse robotic architectures can be constructed.

\notessection{Core ROS Concepts}
To reason effectively about ROS-based systems, it is helpful to understand a set of foundational abstractions that recur throughout the ecosystem.

\paragraph{Nodes.}
A \emph{node} is a single executable process that performs computation.
Nodes are intended to be lightweight and focused, each responsible for a well-defined function.
For example, a mobile robot might have separate nodes for processing camera images, fusing sensor data into a state estimate, generating motion plans, and sending commands to the motors.
A complete robotic system typically consists of many nodes executing concurrently.

\paragraph{Messages and Topics.}
Nodes communicate by exchanging \emph{messages}, which are data structures defined by user-specified types.
Messages are transmitted over named \emph{topics} using a publisher and subscriber model, shown schematically in Figure~\ref{fig:pubsub}. 
A node that produces data publishes messages to a topic, while any number of other nodes may subscribe to that topic to receive the data.
Publishers and subscribers are decoupled, meaning that they do not need to be aware of each other's existence or lifecycle, only the topic name and message type.
This communication pattern naturally supports streaming data flows, such as sensor measurements, state estimates, and control commands, and enables flexible one-to-many and many-to-many information sharing.
\begin{figure}[t] 
    \centering 
    \includegraphics[width=0.9\linewidth]{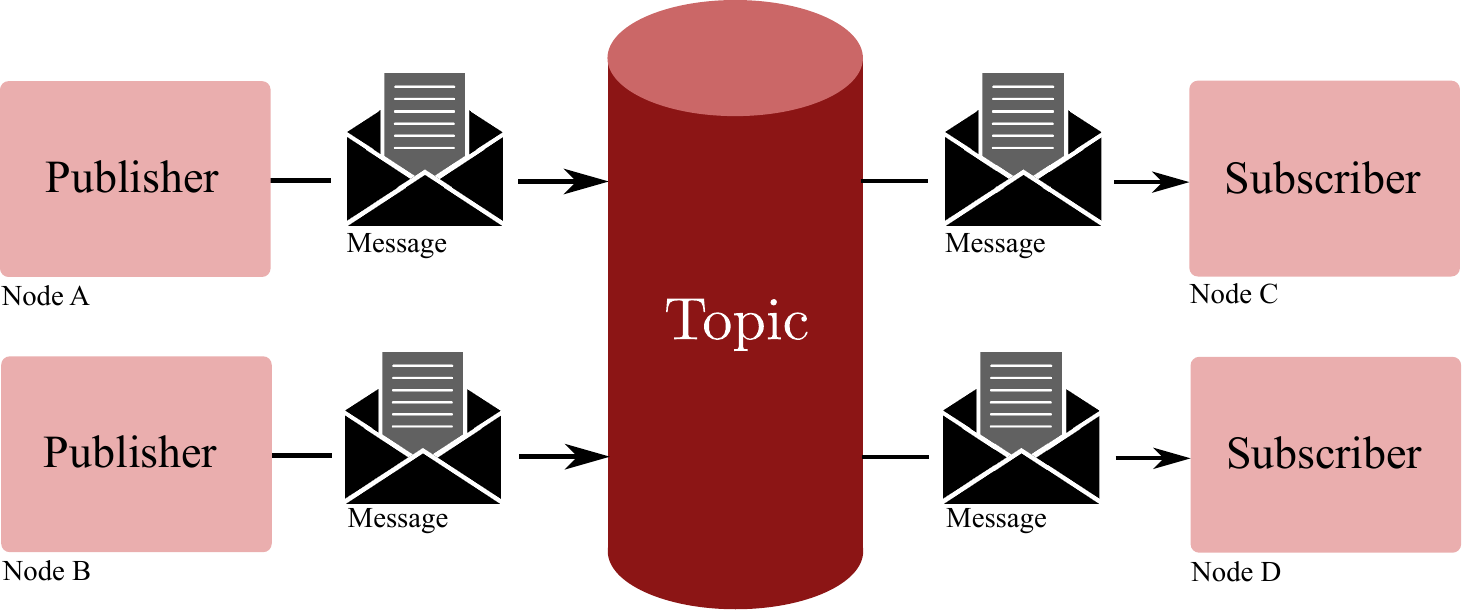}
    \caption{Publish/Subscribe Communication Model in ROS. 
    Topics are best used for unidirectional streaming communication, where a publisher continuously sends messages to one or more subscribers.}
\end{figure}
\figlabel{fig:pubsub}

\paragraph{Services.}
In addition to asynchronous message passing, ROS also supports synchronous request-response interactions through \emph{services}.
A service defines a pair of message types, one for the request and one for the response.
Nodes can advertise services that other nodes can call, allowing for more structured interactions such as parameter queries, configuration changes, or one-off computations.
For example, a node might request an updated map, or portion of a map, from a mapping node.
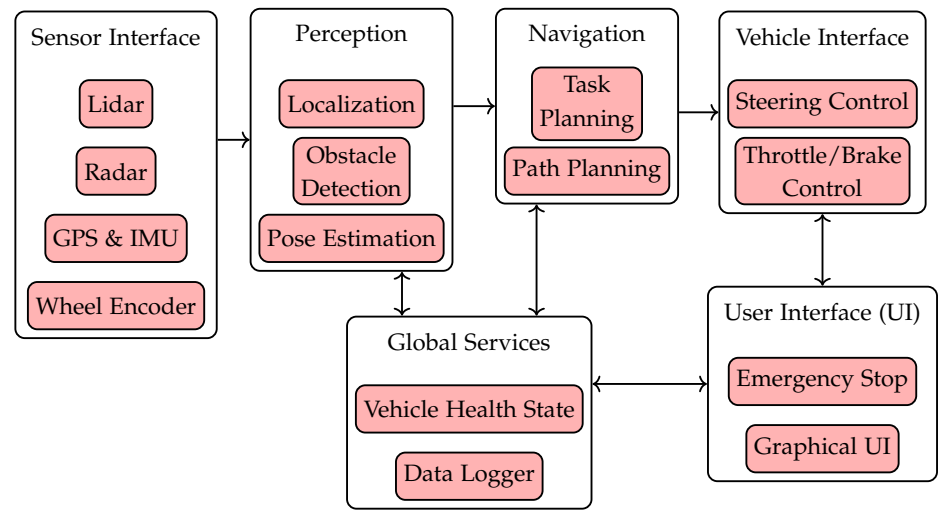
\begin{figure}[t]
    \begin{center}
    \resizebox{0.98\linewidth}{!}{
    \begin{tikzpicture}[node distance=2cm, ->, 
    typetag/.style={rectangle, draw=black!50, font=\scriptsize\ttfamily, anchor=west}]
            \tikzstyle{package} = [draw=black, rectangle, rounded corners, minimum height=3em, minimum width=3em, thick]
            \tikzstyle{node} = [draw=black, rectangle, rounded corners, minimum height=2em, minimum width=2em, thick, fill=red!30]
            \node[](sensor){Sensor Interface};
            \node[node, below of=sensor, xshift=0cm, yshift=1cm](lidar){Lidar};
            \node[node, below of=lidar, xshift=0cm, yshift=1cm](radar){Radar};
            \node[node, below of=radar, xshift=0cm, yshift=1cm](gps){GPS \& IMU};
            \node[node, below of=gps, xshift=0cm, yshift=1cm](encoder){Wheel Encoder};
            \node[package, fit={(sensor)(lidar)(radar)(gps)(encoder)}](sensor-group){};
            
            \node[right of=sensor, xshift=1.5cm](pcp){Perception};
            \node[node, below of=pcp, xshift=0cm, yshift=1cm](loc){Localization};
            \node[node, align=center, below of=loc, xshift=0cm, yshift=1cm](obs){Obstacle \\ Detection};
            \node[node, below of=obs, xshift=0cm, yshift=1cm](pose){Pose Estimation};
            \node[package, fit={(pcp)(loc)(obs)(pose)}](pcp-group){};
            
            \node[right of=pcp, xshift=1.5cm](nav){Navigation};
            \node[node, align=center, below of=nav, xshift=0cm, yshift=1cm](tlc){Task \\ Planning};
            \node[node, align=center, below of=tlc, xshift=0cm, yshift=1cm](pp){Path Planning};
            \node[package, fit={(nav)(tlc)(pp)}](nav-group){};
            
            \node[right of=nav, xshift=1.5cm](veh){Vehicle Interface};
            \node[node, align=center, below of=veh, xshift=0cm, yshift=1cm](steer){Steering Control};
            \node[node, align=center, below of=steer, xshift=0cm, yshift=1cm](throttle){Throttle/Brake \\ Control};
            \node[package, fit={(veh)(steer)(throttle)}](veh-group){};
            
            \node[below of=veh-group, xshift=0cm, yshift=-1cm](user){User Interface (UI)};
            \node[node, align=center, below of=user, xshift=0cm, yshift=1cm](estop){Emergency Stop};
            \node[node, align=center, below of=estop, xshift=0cm, yshift=1cm](gui){Graphical UI};
            \node[package, fit={(user)(estop)(gui)}](user-group){};
            
            \node[below of=pcp-group, xshift=1.75cm, yshift=-1cm](global){Global Services};
            \node[node, align=center, below of=global, xshift=0cm, yshift=1cm](health){Vehicle Health State};
            \node[node, align=center, below of=health, xshift=0cm, yshift=1cm](logger){Data Logger};
            \node[package, fit={(global)(health)(logger)}](global-group){};
            
            \draw[->, thick] (sensor-group.east|-pcp-group.west) to[] (pcp-group.west);
            \draw[->, thick] (pcp-group.east|-nav-group.west) to[] (nav-group.west);
            \draw[->, thick] (nav-group.east|-veh-group.west) to[] (veh-group.west);
            \draw[<->, thick] ([xshift=-1cm]pcp-group.south-|global-group.north) to[] ([xshift=-1cm]global-group.north);
            \draw[<->, thick] ([xshift=1cm]nav-group.south-|global-group.north) to[] ([xshift=1cm]global-group.north);
            \draw[<->, thick] (veh-group.south-|user-group.north) to[] (user-group.north);
            \draw[<->, thick] (global-group.east|-user-group.west) to[] (user-group.west);
    \end{tikzpicture}
    }
    \end{center}
    \caption{Example structure of a mobile robot software framework in ROS. 
    Each white box represents a \emph{package} that contains multiple \emph{nodes} (red boxes) responsible for specific functionality. 
    The arrows indicate communication pathways between different packages.}
    \end{figure}
\figlabel{fig:modularity}

\paragraph{Packages.}
A \emph{package} is a directory that contains related ROS nodes, libraries, datasets, configuration files, and any other resources needed to build and run a particular piece of functionality.
Packages are the primary unit of code organization and distribution in ROS, and can be shared and reused across different projects and robots.
An example of this modular structure is illustrated in Figure~\ref{fig:modularity}.
By separating functionality into well-defined components, ROS enables scalable development, testing, and maintenance of complex robotic systems.

\notessection{Software as the Glue of Autonomy}

While the algorithms presented in this book form the intellectual core of robot autonomy, their practical realization depends critically on software infrastructure. 
In real robotic systems, autonomy is not executed as a single loop but as a distributed graph of asynchronously updating components operating at different rates and exchanging information continuously.
Throughout this text, we focus on the core algorithms and their mathematical foundations, but it is important to recognize that software engineering is the glue that holds these components together in a functioning system.
\newpage
\printbibliography[segment=\therefsegment,heading=subbibliography,title={References}]

\mainmatter

\part{Robot Motion Planning and Control}

\chapter{Modeling Robot Dynamics}
\label{ch:model-dyn}
\newrefsegment
Robots can take on a wide variety of forms: they may have rigid or flexible structures, rely on different types of actuators for control, perceive their surroundings through diverse sensing modalities, or even exist as purely virtual agents.
Despite this variety of embodiments, nearly all robotic systems share a fundamental characteristic---they are \emph{dynamic} agents whose states evolve over time.
The most immediate example of robot dynamics is physical motion, encompassing changes in position, velocity, joint configurations, and sensor orientations.
A deep understanding of the dynamical properties of a robotic system is essential to its effective design and control.
For instance, building a bipedal robot capable of walking or running requires detailed modeling of its motion to ensure that the mechanical structure and actuators can withstand the forces and torques involved. 
Accurate dynamic models are also critical for designing control strategies that achieve stable, efficient, and responsive locomotion.

This chapter introduces several foundational topics in modeling robotic systems.
We will start with \cref{sec:ss_models} by introducing the concept of a \emph{state space model}, which provides a mathematical framework to describe the behavior of a robot's state over time. 
Next, in \cref{sec:kin_and_dyn}, we will detail how a robot's \emph{kinematics and dynamics} are used to derive these state space models, focusing on the principles that govern physical motion and constraints. 
We will then explore specific motion models for wheeled robots in \cref{sec:wheeled_robot_models}, including the \emph{unicycle} and \emph{differential drive} models, to illustrate practical applications of these concepts. 
Finally, in \cref{sec:simulating_dynamics}, we will discuss computational techniques for \emph{simulating} robot dynamics, emphasizing numerical integration methods such as the Euler and Runge-Kutta methods to approximate and analyze the time evolution of robotic systems.

\section{State Space Models}
\label{sec:ss_models}
A state space model is a mathematical framework for describing the behavior of a dynamical system. 
Every state space model consists of two key components: a \emph{state} and a \emph{model}.
\begin{definition}[State]
    The \emph{state} of a dynamical system at time~$t_0$ is a minimal set of variables~$\x(t_0)$ such that, given the control input~$\u(t)$ for all~$t \geq t_0$, the future evolution of the system’s state~$\x(t)$ for all~$t \geq t_0$ is uniquely determined, independently of the system’s behavior for~$t < t_0$.
\end{definition}
Formally, the state is a sufficient statistic of the system’s history: given the current state and the external inputs to the system, the system’s future behavior is fully determined, independently of how that state was reached.
The state of a robotic system can be finite or infinite-dimensional. 
For example, a simple mobile robot with a rigid body can be represented by a finite and low-dimensional state (e.g., the position and velocity of its center of mass), whereas a flexible robot might require an infinite-dimensional state to describe the continuous deformation of its body (e.g., a deflection function that describes the displacement at every point along the robot's length).

When modeling the dynamics of a robot, it is important to define the state in a way that aligns with the specific goals and requirements of the application.
In practice, the complexity of the state representation can often be reduced by omitting parts of the robot's dynamics that are irrelevant to the problem at hand.
For instance, in developing software to enable an autonomous car to navigate urban environments, it may be sufficient to model only its position, orientation, and velocity, while abstracting away the internal mechanics of the engine, tires, or suspension.
Conversely, if the car is being designed for high-performance racing, these specifics become critical to accurately capture and optimize its behavior.
Throughout this book, we focus on applications where the state is finite-dimensional and represent it as a vector~$\x \in \R^\statedim$, referred to as the \emph{state vector}.

The second key component of a state space model is the \emph{model} itself, which defines the rules and equations governing the evolution of the state over time by relating it to a set of \emph{inputs} and \emph{outputs}.

The \emph{inputs} to a model refer to external factors or control signals that influence the behavior of the system.\mysidenote{The term \emph{input} is used interchangeably with \emph{control} and \emph{action} in different domains of robotics.}
These can include forces, torques, voltages, commands, or any other external stimuli that drive the system’s dynamics.
As with the state, the dimensionality of the input may vary, but throughout this book we assume the input is represented by a finite-dimensional vector,~$\u\in \R^\controldim$.

The \emph{outputs} of a model are the observable variables or measurements derived from the system. 
These typically come from sensors or other measurement devices that capture data reflecting the system's state and behavior.
For example, a robot equipped with a global navigation satellite system (GNSS) sensor may directly measure its position but not its heading.
In some cases, the relationship between the state and the output is complex---for example, the connection between a robot’s inertial pose and a red-green-blue-depth (RGB-D) camera image.
The process of inferring the state from the outputs is referred to as \emph{state estimation}, and it will be explored in detail in Chapters \ref{ch:intro-to-localization} - \ref{ch:sensor-fusion}.
We assume the output is finite-dimensional and denote it by the vector~$\y\in\R^q$.
\begin{definition}[Model]
    A \emph{model} describes the evolution of a dynamical system through two types of equations: \emph{state equations} and \emph{observation equations}. 
    The state equations specify how the state evolves as a function of the current state and inputs, while the observation equations define how the state influences the measurable outputs.
\end{definition}

State equations are typically modeled as differential equations\mysidenote{Ordinary differential equations are most commonly used to model robotic systems. However, partial differential equations may be needed for more complex cases, such as systems with flexible or deformable structures, where dynamics vary over both space and time.} describing the changes in the state with respect to an independent variable, usually time:
\begin{equation}
    \label{eq:dynamics-ss}
    \dot{\x} = \dynmodel(\x(t),\u(t)),
\end{equation}
where~$\dot{\x} = \frac{\d \x(t)}{\d t}$ is the time derivative of the state vector~$\x(t)$, and~$f\colon \R^n\times \R^m\to \R^n$ is the \emph{dynamics} function of the system.

Observation equations take the form:
\begin{equation}
    \label{eq:rdt-ss}
    \y(t)=h(\x(t),\u(t)),
\end{equation}
where~$\y(t)$ denotes the output at time~$t$, and~$h\colon \R^n\times \R^m\to \R^q$ is a function that relates the state and inputs to the system's output.

Together, the definitions of the state~$\x$, input~$\u$, output~$\y$, and Equations~\eqref{eq:dynamics-ss} and~\eqref{eq:rdt-ss}, form the \emph{state space model}\mysidenote{In some contexts, it may be convenient to represent the evolution of a system using discrete-time difference equations. As we will discuss in~\cref{ch:openloop}, discrete-time models are particularly well-suited for systems that naturally evolve in discrete steps, or for approximating continuous-time systems within computational frameworks.}.

Throughout this book, we assume that time is the independent variable and simplify our notation by writing~$\theta$ in place of~$\theta(t)$ for time-dependent variables.
Accordingly, we use~$\dot{\theta}$ to represent time derivatives,~$\ddot{\theta} = \frac{\d^{2}\theta(t)}{\d t^2}$ for second derivatives, and~$\theta^{(m)} = \frac{\d^{m}\theta(t)}{\d t^m}$ for higher-order derivatives.

\smallskip
In summary, state space models provide a powerful formalism for modeling, analyzing, and controlling robotic systems. 
Mastering and effectively utilizing state space models is crucial for advancing robotic capabilities, enabling systems to autonomously navigate, interact, and adapt in dynamic and complex environments.

\subsubsection{Types of State Space Models}
State space models, as represented in \cref{eq:dynamics-ss} and \cref{eq:rdt-ss}, can be classified based on two key properties: linearity and time-invariance.

A model is said to be \emph{time-invariant} if the functions~$f$ and~$h$ do not explicitly depend on time~$t$; otherwise, it is \emph{time-varying}.
A model is said to be \emph{linear}\mysidenote{Often referred to as a \emph{linear system}.} if the functions~$f$ and~$h$ are linear functions of both the state~$\x$ and control~$\u$. 
More generally, a system is linear if it satisfies the \emph{superposition principle}, which states that the response to a linear combination of inputs is the corresponding linear combination of the individual responses.
Formally, if~$x_1(t)$ and~$x_2(t)$ are solutions corresponding to inputs~$u_1(t)$ and~$u_2(t)$, respectively, then for any scalars~$\alpha, \beta \in \mathbb{R}$, the trajectory~$\alpha x_1(t) + \beta x_2(t)$ is a solution corresponding to the input~$\alpha u_1(t) + \beta u_2(t)$.
If this property does not hold, the system is said to be \emph{nonlinear}.

\begin{example}[Linear time-invariant model]
Consider the system:
\begin{equation*}
    \begin{split}
    \dot{x}&=x+u,\\
    y&=x.
    \end{split}
\end{equation*}
This model is linear and time-invariant because the functions describing the system depend linearly on~$\x$ and~$\u$ and there is no explicit time dependence.
\end{example}

\begin{example}[Nonlinear time-varying model]
Consider the system:
\begin{equation*}
    \begin{split}
    \dot{x}&=t x+u,\\
    y&=x^2.
    \end{split}
\end{equation*}
This model is \emph{nonlinear} due to the quadratic output term~$x^2$, and it is also \emph{time-varying} because the state equation explicitly depends on time through the term~$tx$.
Note, however, that the state equation~$\dot{x} = t x + u$ is \emph{linear}, as it satisfies the superposition principle with respect to~$x$ and~$u$. 
Therefore, the nonlinearity in this system arises solely from the output equation, not from the state dynamics.
\end{example}

\paragraph{Linear models and their standard form.}
Linear models are particularly important due to their analytical tractability and widespread applicability. 
They are commonly expressed in a standard matrix form:
\begin{equation}
    \begin{split}
    \dot{\x}&=A(t)\x+B(t)\u,\\
    \y&=C(t)\x+D(t)\u,
    \end{split}
\end{equation}
where the matrices~$A(t),B(t),C(t),D(t)$ are of appropriate dimensions and define the linear relationships for the state dynamics and the output. 
We refer to the matrix~$A(t)$ as the \emph{dynamics} matrix,~$B(t)$ as the \emph{control} matrix,~$C(t)$ as the \emph{output} or \emph{sensor} matrix, and~$D(t)$ as the \emph{direct} or \emph{feed-forward} matrix\mysidenote{If the matrices~$A(t),B(t),C(t),$ and~$D(t)$ are time-invariant, we have a Linear Time-Invariant (LTI) system, which is a cornerstone of control theory.}.

Despite the inherent nonlinear nature of many real-world systems, linear models are often employed due to their simplicity and ease of manipulation. 
A common practice is to approximate nonlinear systems with linear models around specific operating points using a technique known as \emph{linearization}. 
As we will discuss in \cref{ch:closedloop}, this approach simplifies analysis and control design, making linear models a fundamental tool in engineering and control theory.

\subsubsection{Converting Higher-Order Models Into State Space Form}
The state space model, as described by \cref{eq:dynamics-ss}, is represented by first-order differential equations. 
However, many robotics applications involve higher-order differential equation models, such as those governing the dynamics of robotic arms or wheeled robots.
To analyze and control these systems in a unified framework, we can convert higher-order models into first-order state space form by introducing additional state variables.

Consider a linear~$n$-th order differential equation:
\begin{equation}
    \begin{split}
    \label{eq:n-th-diff}
    \theta^{(n)} + a_{n-1}\theta^{(n-1)} + \ldots + a_1\dot{\theta} + a_0\theta = u,
    \end{split}
\end{equation} 
where~$a_i \in \R$ are constants and~$\theta$ is the state variable.
To convert this differential equation into state space form, that is, a set of first-order differential equations, we can define an $n$-dimensional state vector:
\begin{equation*}
    \x=\begin{bmatrix} x_1\\ x_2\\ \vdots \\ x_n\end{bmatrix} \coloneqq
    \begin{bmatrix} \theta^{(n-1)} \\[+3pt] \theta^{(n-2)}\\ \vdots \\ \theta\end{bmatrix}.
\end{equation*}
Given that~$\dot{x}_1 = \theta^{(n)}$, $\dot{x}_2 = x_1$, $\dot{x}_3 = x_2$ and so on, we can express the higher-order differential equation as a system of first-order state space equations:
\begin{equation*}
    \begin{bmatrix} \dot{x}_1\\ \dot{x}_2 \\ \vdots \\ \dot{x}_n\end{bmatrix}=
    \begin{bmatrix} 
        -a_{n-1}x_1-\ldots-a_0x_n\\
        x_1\\
        \vdots\\
        x_{n-1}
    \end{bmatrix}+
    \begin{bmatrix}
        u\\
        0\\
        \vdots\\
        0
    \end{bmatrix}.
\end{equation*}
In this particular case, the model is linear, so it can be represented in matrix form:
\begin{equation*}
    \dot{\x}=
    \begin{bmatrix}
        -a_{n-1}&-a_{n-2}&\ldots &-a_1&-a_0\\
        1&0&\ldots&0&0\\
        0&1&&0&0\\
        \vdots &&\ddots&&\vdots\\
        0&0&&1&0
    \end{bmatrix}\x(t)+
    \begin{bmatrix}
        1\\
        0\\
        \vdots\\
        0
    \end{bmatrix}\u(t).
\end{equation*}

\smallskip
\noindent Let us consider a practical example of this technique: 
\begin{example}[Converting Newton's second law to state space form]
\label{ex:double-int}
Many robotic systems, such as robotic arms and wheeled robots, involve dynamics that can be described by Newton's second law. 
Understanding how to convert these higher-order dynamics into state space form is crucial for controlling and analyzing numerous robotics systems.

Newton's second law describes the acceleration response of a mass~$m$ resulting from a force~$F$:
\begin{equation*}
	F=m\ddot{s},
\end{equation*}
where~$s$ represents the one-dimensional positional displacement of the mass. 
Newton's second law, in this form, is a classic example of a \emph{double integrator} system\mysidenote{In control theory, a double integrator describes a system in which the output (here, the position~$s$) is obtained by integrating the input (acceleration) twice.}.

To convert this second-order differential equation into state space form, we first define the state vector:
\begin{equation*}
	\x \coloneqq \begin{bmatrix} s\\\dot{s}\end{bmatrix}.
\end{equation*}
Then, we represent the dynamics in state space form:
\begin{equation*}
	\dot{\x}=\begin{bmatrix} 0&1\\0&0\end{bmatrix}\x+\begin{bmatrix} 0\\ \frac{1}{m} \end{bmatrix} \u,
\end{equation*}
where~$u$ represents the applied force~$F$.
By defining the state vector~$\x$ and expressing the original second-order differential equation in this manner, we have effectively transformed it into a set of first-order differential equations.
\end{example}

\section{Kinematics and Dynamics}
\label{sec:kin_and_dyn}
Understanding a robot's physical motion is essential for enabling its autonomous operation. 
For instance, it is crucial to determine how an autonomous vehicle's actions, such as adjusting throttle and steering, affect its state and interaction with the environment. 
Similarly, it is important to understand how a robot manipulator's movements impact its ability to manipulate objects.
A crucial aspect of this understanding involves the concepts of \emph{kinematics} and \emph{dynamics}, which govern the robot's physical motion and the constraints it must adhere to.
\begin{definition}[Kinematics]
    \emph{Kinematics} is the study of the motion of physical systems, concerned with describing positions, velocities, and accelerations over time, without reference to the forces or torques that produce the motion.
\end{definition}
A robot's kinematics outline limitations on its motion that are determined by its physical state or geometry. 
These constraints arise from the physical structure of the system, such as joint limits, actuator placement, linkage geometry, and the ways in which different components are mechanically connected. 
They determine how the system can move, independent of any external forces.
For example, consider a wheeled robot. 
Static friction restricts the wheels from sliding laterally, meaning they cannot move in the direction parallel to the rotation axis. 
This kinematic constraint significantly limits the robot's ability to navigate and affects its overall maneuverability. 
By understanding these constraints, one can better grasp the feasible movements of the robot within its environment, which in turn informs how it can effectively interact with the world around it.

\begin{definition}[Dynamics]
    \emph{Dynamics} is the study of the motion of physical systems as determined by the forces and torques acting upon them. 
    It seeks to relate a system’s motion to the underlying physical causes of that motion, such as gravity, friction, or applied inputs.
\end{definition}
In the context of robotic or mechanical systems, dynamics are typically governed by Newton’s Second Law, which states that the acceleration of a body is proportional to the net force acting on it.
For example, the dynamics of an autonomous vehicle are described by the relationship between its acceleration and the external forces acting on it, including tire-road interaction, gravitational effects on slopes, and aerodynamic drag.

\smallskip
From the definitions above, kinematics describe constraints that arise from the robot's physical state or geometry, whereas dynamics explain how forces or inputs influence the robot's motion.
In this section, we first introduce the concept of \emph{generalized coordinates} for defining the physical configuration of a robot and discuss how the robot's kinematics and dynamics can be expressed in terms of these coordinates.
Then, we will demonstrate how to use the system's kinematics and dynamics to define a state space model for the robot's physical motion.
In the following chapters, we will explore how these models are applied to develop robust and high-performing algorithms for motion planning and control.

\subsubsection{Generalized Coordinates}
A robot’s physical state, also referred to as its \emph{configuration}, provides a complete specification of the position of every point on the robot at a given instant~\cite{Lozano1990}.
A configuration can be represented using a set of variables known as \emph{generalized coordinates}, denoted by~$\q(t) \in \mathbb{R}^{n_g}$.
Generalized coordinates form a set of parameters that uniquely describe the robot’s configuration relative to a reference frame.
Depending on the system, they may include joint angles, Cartesian positions, orientations, or other parameters defining the robot’s physical arrangement\mysidenote{The terms ``configuration'' and ``generalized coordinates'' are often used interchangeably. However, while the configuration refers to the robot’s physical arrangement in space, generalized coordinates are a specific mathematical representation of that arrangement. Multiple choices of generalized coordinates may represent the same configuration.}.

Importantly, the configuration, namely the generalized coordinates, typically constitutes only a part of the full system state~$\x$.
As discussed in previous sections, in a state-space model, the state is a complete set of variables sufficient to determine the system’s future evolution, given an external input.
This usually includes both the generalized coordinates~$\q$ and their time derivatives, called \emph{generalized velocities}, denoted as~$\dot{\q}$.
In some systems, the state may also include additional internal variables such as actuator dynamics, sensor states, or environmental parameters.

\begin{figure*}[tb] 
    \centering 
    \includegraphics[width=0.55\linewidth]{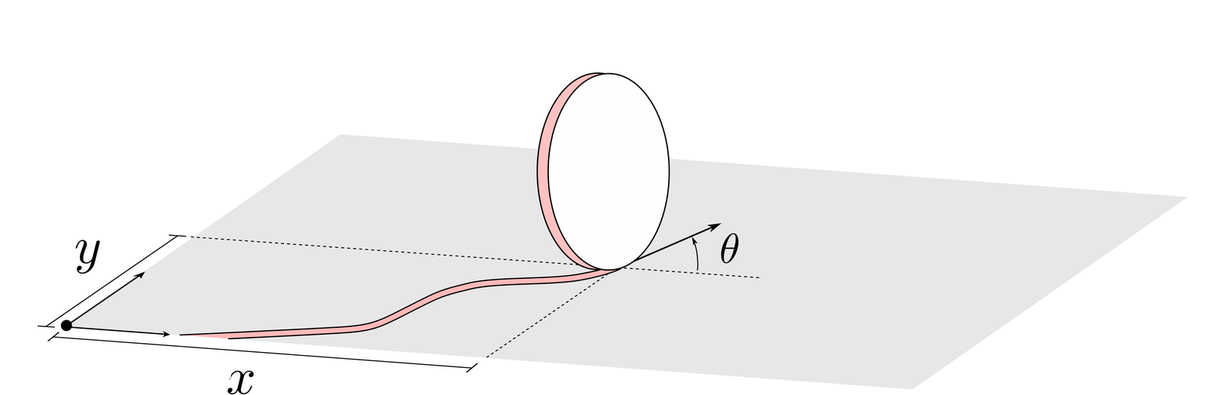}
    \caption{Generalized coordinates for a wheel rolling without slipping on a plane.} 
    \label{fig:non slip disk} 
\end{figure*}

\begin{example}[Rolling Wheel]
The configuration of a wheel rolling on a plane, as illustrated in \cref{fig:non slip disk}, can be represented by three parameters: the contact point position coordinates~$\tup{x, y}$ and the heading angle~$\theta$ relative to a fixed reference frame. 
This set of parameters,~$\q = \vCol{x,y,\theta}$, constitutes just one possible choice of generalized coordinates to define the wheel's configuration. 
Alternatively, the configuration could also be represented using the heading angle~$\theta$ along with a polar coordinate representation of the contact point position.
\end{example}

\subsubsection{Kinematic Constraints}
Once a specific set of generalized coordinates,~$\q$, is chosen to represent a robot's configuration, we can identify the relevant \emph{kinematic constraints} for the robot. 
These kinematic constraints establish relationships between the generalized coordinates and the generalized velocities, thereby describing the limitations on the robot's motion.
We refer the reader to~\citet{SicilianoEtAl2008Kinematics} for a comprehensive treatment of robotic kinematics and the formulation of the associated constraint equations.
\begin{definition}[Kinematic Constraints]
\emph{Kinematic constraints} are a set of constraints imposed on the generalized coordinates,~$\q$, and generalized velocities,~$\dot{\q}$. 
We express kinematic constraints mathematically as:
\begin{equation} 
\label{eq:kinconst}
    \tilde{a}_i(\q, \dot{\q}) = 0, \quad \quad i = 1, \dots, k < n_g,
\end{equation}
where~$k$ is the number of constraints and~$n_g$ is the number of generalized coordinates.
\end{definition}

In many robotics applications, kinematic constraints are linear with respect to the generalized velocities. 
These are known as \emph{Pfaffian constraints}, and can be mathematically expressed as:
\begin{equation}
\label{eq:pfaffian}
    \a_i^\top (\q)\dot{\q} = 0, \quad \quad i =1, \dots, k < n_g,
\end{equation}
where~$\a_i(\q) \in \R^{n_g}$.
Pfaffian constraints can also be compactly represented in matrix form as:
\begin{equation}
    A^\top (\q)\dot{\q}=\bm{0},
\end{equation}
where $A(\q) \in \R^{n_g \times k}$.

\begin{example}[Pendulum] 
\label{ex:pendulum}
\begin{marginfigure}
    \centering 
    \includegraphics[width=0.55\linewidth]{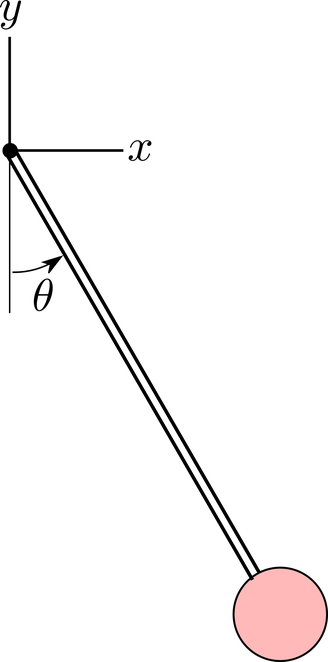}
    \caption{Generalized coordinates for a simple pendulum. } 
    \label{fig:pendulum} 
\end{marginfigure} 
\cref{fig:pendulum} shows a simple pendulum with a point mass and a rigid, massless, rod that rotates about a fixed pivot point.
We can choose to represent the configuration of the pendulum by the Cartesian coordinate position of the mass, assuming the pivot point is the reference frame origin.
The generalized coordinate vector for this choice is~$\q = \vCol{x, y}$, and the generalized velocity vector is~$\dot{\q} = \vCol{\dot{x}, \dot{y}}$.
The fact that the rod connecting the pivot point to the mass is rigid introduces a restriction on the motion of this system, which we represent by the kinematic constraint:
\begin{equation} 
\label{eq:pendholonomic}
\tilde{a}_1(\q, \dot{\q}) = x^2 + y^2 - L^2 = 0,    
\end{equation}
where~$L$ is the length of the rod. 
While this constraint is not in Pfaffian form, we can equivalently express it as a Pfaffian constraint by noting that:
\begin{equation*}
\tilde{a}_1(\q, \dot{\q}) = 0 \implies \frac{\partial \tilde{a}_1(\q, \dot{\q})}{\partial t} = 0.
\end{equation*}
For the pendulum kinematic constraint in \cref{eq:pendholonomic}, we have:
\begin{equation*}
\frac{\partial \tilde{a}_1(\q, \dot{\q})}{\partial t} = 2x\dot{x} + 2y\dot{y},
\end{equation*}
and therefore we can write the constraint in the Pfaffian form of \cref{eq:pfaffian} with:
\begin{equation} 
\label{eq:pendpfaffian}
\a_i^\top (\q) = \begin{bmatrix}2x & 2y\end{bmatrix}.
\end{equation}
The Pfaffian constraint in \cref{eq:pendpfaffian} implies that \cref{eq:pendholonomic} holds as long as the pendulum starts in a state~$\q(0)$ satisfying~$\tilde{a}_1(\q(0)) = 0$.

An alternative choice of generalized coordinates to represent the pendulum's configuration is to consider the angle~$\theta$ between the vertical and the pendulum's rod orientation,~$\q = [\theta]$. 
This choice fully specifies the configuration without requiring us to define any kinematic constraints, making it a more natural choice for this system. 
Note that since~$x = L\sin \theta$ and~$y = -L \cos \theta$ the kinematic constraint in \cref{eq:pendholonomic} is trivially satisfied for all~$\theta$.
\end{example}

\begin{example}[Rolling Wheel] 
\label{ex:noslipwheel}
Consider the wheel illustrated in \cref{fig:non slip disk}, which we can represent with the generalized coordinates $\q = \vCol{x, y, \theta}$. 
For this system, we can assume that the friction at the contact point between the wheel and the surface induces a no-slip condition. 
This no-slip condition is a constraint on the motion of the wheel that restricts the velocity component of the wheel in the lateral direction to always be zero.
Since the unit vector~$\e_v = \vCol{\cos\theta, \:\sin\theta}$ describes the heading of the wheel, the lateral direction is given by the perpendicular unit vector~$\e_{v,\perp} = \vCol{\sin\theta, \: -\cos\theta}$.
We can compute the lateral velocity from the dot product of the lateral direction unit vector and the velocity vector,~$\bv = \vCol{\dot{x}, \: \dot{y}}$, which gives the no-slip kinematic constraint:
\begin{equation} 
\label{eq:wheelnonholonomic}
a_1(\q,\dot{\q}) = \dot{x} \sin\theta - \dot{y} \cos \theta = 0.
\end{equation}
This constraint is linear in the generalized velocities,~$\tup{\dot{x}, \dot{y}}$, and therefore is a Pfaffian constraint.
\end{example}

\subsubsection{Holonomic and Nonholonomic Constraints}
Kinematic constraints often fall into two categories: \emph{holonomic} or \emph{nonholonomic}, depending on how they restrict the motion of the system. 
Holonomic constraints can be expressed solely as functions of the generalized coordinates, without involving generalized velocities. 
In contrast, nonholonomic constraints involve the generalized velocities and cannot be expressed solely in terms of the generalized coordinates.
\begin{definition} [Holonomic Constraints]
Kinematic constraints that can be expressed in the form:
\begin{equation}
    \tilde{a}_i(\q) = 0, \quad i = 1, \dots, k < n_g,
    \label{eq:holonomic}
\end{equation}
are called \emph{holonomic}.
\end{definition}
In robotics applications, holonomic constraints generally arise due to mechanical interconnections, such as rigid links and joints of a robotic arm.
We refer to a system that is only subject to holonomic constraints as a \emph{holonomic system}.
These constraints are a unique subclass of kinematic constraints that restrict the accessible configurations of the system. 
Specifically, for a system with~$n$ generalized coordinates under~$k$ holonomic constraints, the dimension of the space of accessible configurations is~$n - k$.
Holonomic constraints can always be equivalently expressed as Pfaffian constraints of the form \cref{eq:pfaffian}. 
This is because:
\begin{equation*}
\tilde{a}_i(\q) = 0 \implies \frac{\partial \tilde{a}_i(\q)}{\partial t} = 0,
\end{equation*}
and by differentiating the expression:
\begin{equation}
\frac{\partial \tilde{a}_i(\q)}{\partial t} = \frac{\partial \tilde{a}_i(\q)}{\partial \q}\dot{\q} = \a_i^\top (\q) \dot{\q},
\end{equation}
as we demonstrated in \cref{ex:pendulum}.
However, it is important to note that not all Pfaffian constraints are holonomic. 
For a Pfaffian constraint to be holonomic, it must be integrable to the form in \cref{eq:holonomic}.
Specifically, there must exist a scalar function~$\tilde{a}_i(\q)$ such that:
\begin{equation}
\a_i^\top (\q) \dot{\q} = \frac{\partial \tilde{a}_i(\q)}{\partial \q} \dot{\q} = 0.
\end{equation}
This implies that the Pfaffian constraint, when integrated, yields a constraint solely dependent on the generalized coordinates~$\q$, without any explicit dependence on their time derivatives~$\dot{\q}$.

\begin{example}[Pendulum]
Consider the pendulum from \cref{ex:pendulum}. 
The kinematic constraint in \cref{eq:pendholonomic} restricts the pendulum mass to lie on a circle of radius~$L$, which is a subset of all possible generalized coordinates. 
This constraint is holonomic since we can express it as a function of only the generalized coordinates.
\end{example}

\begin{example}[Rolling Wheel]
Consider the wheel from \cref{ex:noslipwheel}, where the kinematic constraint in \cref{eq:wheelnonholonomic} restricts the direction of motion.
In contrast to the pendulum, this constraint does not limit the wheel's ability to reach any configuration of generalized coordinates, namely the position and heading.
Mathematically, we cannot integrate the constraint in \cref{eq:wheelnonholonomic} to yield a constraint of the form~$\tilde{a}_i(\q) = 0$, and thus this constraint is not holonomic.
\end{example}

While holonomic constraints restrict the system's accessible configurations, kinematic constraints can also limit the motion between configurations. 
We refer to these constraints as \emph{nonholonomic} constraints.
A system that is subject to at least one nonholonomic constraint is referred to as a \emph{nonholonomic system}.
\begin{definition} [Nonholonomic Constraints]
Constraints that can be described in Pfaffian form,~$\a_i(\q)^\top \dot{\q} = 0$, but cannot be integrated to the form~$\tilde{a}_i(\q) = 0 $ are called \emph{nonholonomic}.
\end{definition}
That is, the Pfaffian expressions cannot be written as the total time derivative of any scalar function that depends only on the generalized coordinates.
In other words, there exists no scalar function~$\tilde{a}_i(\q)$ such that~$\frac{\d}{\d t} \tilde{a}_i(\q) = \a_i(\q)^\top \dot{\q}$.
Geometrically, nonholonomic constraints restrict the instantaneous generalized velocities to lie in the null space of~$A(\q)^\top$, where~$A(\q)$ is the matrix whose rows are the vectors~$\a_i(\q)^\top$ corresponding to each nonholonomic constraint.

\begin{example}[Rolling Wheel]
Consider the wheel example from \cref{ex:noslipwheel}, which has a nonholonomic constraint:
\begin{equation*}
a_1(\q)^\top \dot{\q} = \begin{bmatrix}
\sin \theta & -\cos \theta & 0
\end{bmatrix}\dot{\q}=0.
\end{equation*}
The null space of~$a_1(\q)^\top$ in this case is spanned by the vectors~$[\cos \theta, \: \sin \theta,\: 0]$ and~$[0, \: 0,\: 1]$, which suggests that all motion must be made up of a linear combination of these vectors. 
Intuitively, we expect this because~$[\cos \theta, \: \sin \theta,\: 0]$ is the rolling direction and~$[0, \: 0,\: 1]$ is the axis the wheel rotates about.
\end{example}

In summary, holonomic constraints restrict a system’s motion by confining its configurations to lower-dimensional manifolds—specifically, level sets defined by scalar equations of the form~$\tilde{a}_i(\q) = 0$. 
For a system with~$n$ generalized coordinates and~$k$ independent holonomic constraints, the configuration space is effectively reduced to a manifold of dimension~$n - k$. 
This reduction reflects a true loss of accessibility in the configuration space: the system can only evolve along a restricted subset of configurations determined by the initial conditions and the constraint equations.
In contrast, nonholonomic constraints act directly on the system’s instantaneous velocities, typically expressed in Pfaffian form as~$A(\q)^\top \dot{\q} = 0$, where~$A(\q)^\top$ is a full-rank matrix of dimension~$k \times n$. 
These constraints restrict the allowed directions of motion at each configuration by confining~$\dot{\q}$ to lie in an~$(n - k)$-dimensional subspace. 
However, since nonholonomic constraints are not integrable, they do not reduce the dimensionality of the configuration space itself. 
That is, although motion is constrained at each instant, the system may still be able to reach any configuration in the configuration space through admissible trajectories that respect the velocity constraints.
Thus, while holonomic constraints reduce the number of independent configuration variables and confine the system to a lower-dimensional subset of the configuration space, nonholonomic constraints preserve full accessibility of the configuration space but restrict how that space can be traversed.

\subsubsection{Kinematic Models}
Once we have chosen an appropriate set of generalized coordinates~$\q$ and have identified the relevant kinematic constraints, we can convert the kinematic constraints into a state space model of the form in \cref{eq:dynamics-ss}, which we refer to as a \emph{kinematic model}.
\begin{definition} [Kinematic Model]
\label{def:kinematic_model}
Given a generalized coordinate vector~$\q \in \R^{n_g}$, and~$k$ Pfaffian constraints\mysidenote{These Pfaffian constraints can come from a combination of holonomic and non-holonomic constraints.},~$A^\top (\q)\dot{\q}=\bm{0}$, a \emph{kinematic model} is a state space model of the form:
\begin{equation}
\dot{\q} = G(\q)\u,
\end{equation}
where~$\u \in \R^p$ is the input and where the column space of~$G(\q) \in \R^{n_g \times n_g-k}$ spans the null space of~$A^\top (\q)$. 
\end{definition}
Each input in~$\u$ corresponds to one degree of freedom of the system, and for any initial condition~$\q(0)$ and sequence of inputs~$\u(t)$ the solutions to the kinematic model are guaranteed to satisfy the Pfaffian constraints. 
We can prove that the trajectories of the kinematic model will satisfy the Pfaffian constraints by writing the model in the equivalent form:
\begin{equation*}
\dot{\q} = G(\q)\bu = \sum_{i=1}^{n-k} \bg_i(\q) u_i,
\end{equation*}
where~$\bg_i \in \R^{n_g}$ is the $i$-th column of~$G$ and~$u_i$ is the~$i$-th input.
In this form, we can more easily see that each input acts on the generalized velocity~$\dot{\q}$ through a particular mode that is defined by the vector~$\bg_i$.
Since we have specified in \cref{def:kinematic_model} that the column space of~$G$ spans the null space of~$A^\top(\q)$, we know that by definition:
\begin{equation*}
A^\top(\q)\bg_i(\q)u_i = 0,
\end{equation*}
for any input~$u_i \in \R$ and for all coordinates~$\q$.
Therefore, by definition each component of the input vector can only influence the generalized velocity in a way that satisfies the Pfaffian constraints.
Another way to see this mathematically is by substituting the kinematic model into the Pfaffian constraint equation:
\begin{equation*}
\begin{split}
A^\top (\q)\dot{\q} &= A^\top (\q) G(\q)\u, \\
&=A^\top (\q)\big(\sum_{i=1}^{n-k} \bg_i(\q) u_i\big), \\
&= \sum_{i=1}^{n-k} A^\top (\q) \bg_i(\q) u_i, \\
&= 0. \\
\end{split}
\end{equation*}

\begin{example}[Rolling Wheel]
Consider the rolling wheel example from \cref{ex:noslipwheel}, which has a single nonholonomic constraint:
\begin{equation*}
a_1(\q)^\top \dot{\q} = \begin{bmatrix}
\sin \theta & -\cos \theta & 0
\end{bmatrix}\dot{\q}=0,
\end{equation*}
where~$\q = \vCol{x,\:y, \:\theta}$. 
The null space of~$a_1(\q)^\top $ is spanned by the vectors~$\vCol{\cos \theta, \: \sin \theta,\: 0}$ and~$\vCol{0, \: 0,\: 1}$ and therefore the kinematic model is given by:
\begin{equation} 
\label{eq:wheelkinmodel}
\begin{bmatrix}
\dot{x} \\ \dot{y} \\ \dot{\theta}
\end{bmatrix} = \begin{bmatrix}
\cos \theta & 0 \\
\sin\theta & 0 \\
0 & 1
\end{bmatrix}\begin{bmatrix}
u_1 \\ u_2
\end{bmatrix}.
\end{equation}
In this case, the inputs~$u_1$ and~$u_2$ have an intuitive physical meaning:~$u_1$ is the speed at which the wheel is moving, and~$u_2$ is the wheel's angular rotation rate. 
\end{example}

\subsubsection{Dynamics Models}
Kinematic models describe the geometric constraints that limit a robot’s motion---for example, the fact that a robotic arm can only rotate about its joints.
However, kinematics alone does not explain how motion is generated or resisted. 
For that, we turn to dynamics, which describe how forces and torques influence motion by producing accelerations.

Newton's second law of motion is the foundation of robot dynamics, relating the net force acting on a body to its acceleration.
Applied to a single point mass, the law takes the form:
\begin{equation}
\label{eq:newtonsecond}
\bm{F}(\q, \dot{\q}) = m\ddot{\q},
\end{equation}
where~$m$ is the mass of the particle,~$\q = \vCol{x, y, z}$ is its position vector, and~$\bm{F}$ is the total force acting on the particle.
\begin{example}[Mass-spring-damper system]
A fundamental example in the study of dynamics is the one-dimensional mass-spring-damper system.
The system consists of a mass~$m$ attached to a spring and a damper, constrained to move along a line. 
The total force acting on the mass is typically composed of three terms:
\begin{enumerate}
    \item an external input force~$F_\text{external}$,
    \item a spring force~$-kx$ that resists displacement from the equilibrium position, and
    \item a damping force~$-c\dot{x}$ that resists velocity.
\end{enumerate}
The net force on the mass is:
\begin{equation*}
F = F_\text{external} - kx - c\dot{x},
\end{equation*}
where~$k > 0$ and~$c > 0$ are the spring and damping coefficients, respectively, and~$x$ is the displacement from equilibrium.
Substituting this expression into Newton’s second law yields the following second-order differential equation:
\begin{equation*}
m\ddot{x} + c\dot{x} + kx = F_\text{external}.
\end{equation*}
This equation models oscillatory motion with damping, and it arises in many robotics applications---for example, when analyzing joint compliance, actuator dynamics, or contact interactions.
\end{example}

While the mass-spring-damper system illustrates the dynamics of a single particle in one dimension, real-world robotic systems are often more complex. 
To capture their behavior, we extend Newton’s second law to systems of interconnected particles, typically modeled in robotics as \emph{rigid bodies}.
A rigid body is an idealized object in which the relative positions of all constituent particles remain fixed over time, regardless of external forces.
This assumption implies that the body does not deform and allows us to reduce a complex system of interacting particles to a simpler model governed by the motion of a finite set of parameters, for example, position and orientation of a frame fixed to the body.
As a result, rigid body dynamics provide a powerful and tractable framework for analyzing and simulating robotic systems\mysidenote{Interesting examples in robotics where the rigid body assumption may not hold include soft robots, robots with compliant end-effectors, or robots with lightweight flexible structures.}.

The motion of a rigid body in three-dimensional space can be described in terms of its translational and rotational dynamics.
Translational dynamics govern the motion of the body’s center of mass and are described by Newton’s second law, as expressed in \cref{eq:newtonsecond}, where the position variable refers specifically to the center of mass.
In three-dimensional space, a rigid body has three translational degrees of freedom, corresponding to movement along each of the Cartesian axes.
Rotational dynamics, on the other hand, describe how the body’s orientation evolves over time. 
A rigid body also has three degrees of freedom associated with its orientation in three-dimensional space, corresponding to rotation about each of its principal axes.
Unlike translation, orientation cannot be represented by a single vector, and several parameterizations are commonly used.
Notable examples include rotation matrices, which provide a full and unambiguous representation at the cost of redundancy; \emph{Euler angles}, which use a sequence of three rotations to represent orientation, and \emph{quaternions}, which offer a compact and singularity-free alternative well-suited for numerical applications.

The rotational dynamics\mysidenote{We refer the reader to~\citet{Shuster1981} for an in-depth treatment of rotational dynamics and attitude representations.} of a rigid body are governed by the time evolution of its angular momentum. 
Specifically, they are described by:
\begin{equation}
\label{eq:rotdynamics}
\bm{M} = \dot{\bm{H}},
\end{equation}
where~$\bm{M}$ denotes the total external moment (or torque) acting on the body, and~$\bm{H}$ is the angular momentum, typically computed about the center of mass.
This relationship is commonly referred to as \emph{Euler’s equation} for rotational dynamics and captures how applied torques influence changes in the body’s rotational motion.

An alternative to the Newton-Euler method---defined by Equations~\eqref{eq:newtonsecond}-\eqref{eq:rotdynamics}---for deriving the equations of motion for a rigid body is the \emph{Lagrange Method}. 
This approach is closely tied to the notion of generalized coordinates, generalized velocities, and kinematic constraints, and takes an energy-based approach. 
In the Lagrange method, the dynamics of a rigid body are derived from a scalar quantity called the \emph{Lagrangian}, defined as the difference between the kinetic and potential energies:
\begin{equation}
\label{eq:dyn_lagrangian}
L(\q, \dot{\q}) = T(\q, \dot{\q}) - V(\q),
\end{equation}
where~$T(\q, \dot{\q})$ and~$V(\q)$ denote the kinetic and potential energies of the system, respectively. 
The evolution of the system is governed by \emph{Lagrange’s equations}~\cite{SicilianoEtAl2008Lagrange}\cite{LynchEtAl2017Lagrange}, which incorporate both external influences and kinematic constraints. 
In the absence of constraints, the equations of motion are given by:
\begin{equation}
\label{eq:standard_lagranges_eq}
\begin{split}
\frac{\d}{\d t}\Big(\frac{\partial L}{\partial \dot{\q}_j} \Big) - \frac{\partial L}{\partial \q_j} = Q_j, \quad j = 1, \dots, n_g,
\end{split}
\end{equation}
where~$Q_j\in\R$ is a non-conservative \emph{generalized force} associated with the generalized coordinate~$\q_j$\mysidenote{Generalized forces are projections of physical forces and torques into the generalized coordinate space. Forces not derived from a potential---such as friction---are termed non-conservative. In contrast, forces like gravity are conservative.}, and~$n_g$ corresponds to the system’s degrees of freedom.
Equations~\eqref{eq:standard_lagranges_eq} describe how the generalized forces acting on the system relate to its position, velocity, and acceleration, providing a systematic way to derive the system’s dynamic model from its kinetic and potential energies.

In the presence of Pfaffian constraints\mysidenote{While Lagrange’s method can accommodate general constraints, we focus here on Pfaffian constraints for simplicity.}, Lagrange's equations take the form:
\begin{equation}
\label{eq:lagranges_eq}
\begin{split}
\frac{\d}{\d t}\Big(\frac{\partial L}{\partial \dot{\q}_j} \Big) - \frac{\partial L}{\partial \q_j} = Q_j + \sum_{i=1}^k \lambda_i a_{ij}(\q), \quad j = 1, \dots, n_g, \\
\a_i^\top (\q)\dot{\q} = 0, \quad \quad i =1, \dots, k,
\end{split}
\end{equation}
where~$a_{ij}$ is the~$j$-th component of the~$i$-th Pfaffian constraint vector~$\a_i(\q)$ and~$\lambda_i \in \R$ is a \emph{Lagrange multiplier}.
The first~$n_g$ equations describe the dynamics of the generalized coordinates under the influence of both external and constraint forces, while the remaining~$k$ equations represent the kinematic constraints themselves.
The complete system thus comprises~$n_g + k$ equations in~$n_g + k$ unknowns (the generalized coordinates and the Lagrange multipliers), and is commonly referred to as the \emph{standard non-holonomic form}.
If the system is holonomic and the generalized coordinates are chosen to be independent, the constraints are implicitly satisfied, and Lagrange’s equations reduce to the simpler, unconstrained form introduced in~\cref{eq:standard_lagranges_eq}.

\begin{example}[Pendulum]
Consider again the pendulum depicted in \cref{fig:pendulum}.
To model its dynamics, we analyze how gravity drives the motion of the mass.
Specifically, we will demonstrate four distinct approaches for deriving the equations of motion—using both Cartesian and polar coordinates, and applying both the Newton-Euler and Lagrange methods.
This comparison will highlight how the choice of generalized coordinates can influence the complexity of the derivation.

We begin by using Newton’s second law to derive the dynamics of the pendulum, focusing on the two forces acting on the mass: gravity and the force from the rod.
We assume that the rod’s force acts purely along its axis. 
Since the pendulum’s length is fixed, this force must counteract the component of gravity along the rod and generate the required centripetal acceleration.
The axial force exerted by the rod is given by:
\begin{equation*}
F_r = mg\cos\theta + \frac{mv^2}{L},
\end{equation*}
where~$m$ is the mass of the pendulum,~$g$ is gravitational acceleration,~$L$ is the length of the rod, and~$v$ is the speed of the mass.
The gravitational force is:
\begin{equation*}
F_g = mg,
\end{equation*}
acting along the negative~$y$-direction.
To compute the net force in Cartesian coordinates, we project both the rod’s force and the gravitational force onto the~$x$- and~$y$-axes:
\begin{equation*}
\begin{split}
F_x &= -\frac{mv^2}{L}\sin\theta - mg\sin\theta\cos\theta,\\
F_y &= \frac{mv^2}{L}\cos\theta - mg\sin^2\theta.
\end{split}
\end{equation*}
Applying Newton’s second law as defined in \cref{eq:newtonsecond} yields the equations of motion:
\begin{equation*}
\begin{split}
\ddot{x} &= -\frac{v^2}{L}\sin\theta - g\sin\theta\cos\theta,\\
\ddot{y} &= \frac{v^2}{L}\cos\theta - g\sin^2\theta.
\end{split}
\end{equation*}
To express these equations purely in terms of Cartesian coordinates, we substitute~$x = L\sin\theta$ and $y = -L\cos\theta$, leading to:
\begin{equation}
\label{eq:newton_pend}
\begin{split}
\ddot{x} &= \frac{1}{L^2}(gxy - xv^2),\\
\ddot{y} &= -\frac{1}{L^2}(gx^2 + yv^2),
\end{split}
\end{equation}
with~$v^2 = \dot{x}^2 + \dot{y}^2$.
This method requires careful force analysis, as the kinematic constraint (fixed-length rod) is handled implicitly through the projected components of the rod’s force.

As a second approach to deriving the equations of motion using Cartesian coordinates, we now apply the Lagrange method, which yields a slightly simpler formulation.
We begin by defining the kinetic and potential energies:
\begin{equation*}
T = \frac{1}{2}m(\dot{x}^2 + \dot{y}^2), \quad V = mgy,
\end{equation*}
and observe that there are no external non-conservative generalized forces\mysidenote{Gravity is a conservative force, so no generalized non-conservative forces appear.}.
As discussed in~\cref{ex:pendulum}, we recall the system's single Pfaffian constraint:
\begin{equation*}
x\dot{x} + y\dot{y} = 0.
\end{equation*}
Thus, using Lagrange’s equations introduced in~\cref{eq:lagranges_eq}, we obtain:
\begin{equation}
\begin{split}
m\ddot{x} &= \lambda x, \\
m\ddot{y} + mg &= \lambda y, \\
x\dot{x} + y\dot{y} &= 0.
\end{split}
\label{eq:lagrange_pendulum_cartesian}
\end{equation}
We can solve for the Lagrange multiplier~$\lambda$ by differentiating the constraint with respect to time:
\begin{equation*}
\frac{\d}{\d t}(x\dot{x} + y\dot{y}) = \dot{x}^2 + \dot{y}^2 + x\ddot{x} + y\ddot{y} = 0,
\end{equation*}
and by substituting the expressions for~$\ddot{x}$ and $\ddot{y}$ from the first two Lagrange's equations in~\eqref{eq:lagrange_pendulum_cartesian} to obtain:
\begin{equation*}
\dot{x}^2 + \dot{y}^2 + \frac{1}{m}x^2 \lambda + \frac{1}{m} y^2 \lambda - gy = 0.
\end{equation*}
Solving for~$\lambda$ yields:
\begin{equation*}
\lambda = \frac{m}{L^2}(gy - v^2),
\end{equation*}
where we used~$L^2 = x^2 + y^2$ and $v^2 = \dot{x}^2 + \dot{y}^2$.
Finally, substituting the expression for~$\lambda$ back into the equations of motion and simplifying, we find:
\begin{equation}
\label{eq:lagrange_cartesian_pend}
\begin{split}
\ddot{x} &= \frac{1}{L^2}(gxy - xv^2), \\
\ddot{y} &= -\frac{1}{L^2}(gx^2 + yv^2), \\
\end{split}
\end{equation}
which matches the result previously obtained using Newton’s method in \cref{eq:newton_pend}.

After applying both the Newton-Euler and Lagrange methods in Cartesian coordinates, we now replicate the derivation in polar coordinates.
To apply Euler's equation for rotational dynamics as given in~\cref{eq:rotdynamics}, we adopt a coordinate frame fixed at the pivot point.
The gravitational force acting on the mass generates a moment about the pivot:
\begin{equation*}
M = -mgL\sin\theta,
\end{equation*}
while the angular momentum of the system about the same point is:
\begin{equation*}
H = mL^2 \dot{\theta},
\end{equation*}
where~$mL^2$ is the \emph{moment of inertia}\mysidenote{In Euler’s equation, the moment of inertia plays a role analogous to mass in Newton’s second law.} about the pivot. 
Substituting into Euler’s equation yields the system dynamics:
\begin{equation}
\label{eq:euler_pend}
\ddot{\theta} = -\frac{g}{L}\sin\theta,
\end{equation}
which are considerably more compact than the corresponding equations derived in Cartesian coordinates.

The Lagrange method also becomes significantly simpler when using the polar coordinate~$\theta$, as there is no need to handle Pfaffian constraints explicitly. 
In this formulation, the kinetic and potential energies of the system are:
\begin{equation*}
T = \frac{1}{2}mL^2\dot{\theta}^2, \quad V = -mgL\cos\theta.
\end{equation*}
Applying \cref{eq:lagranges_eq}, and noting the absence of non-conservative generalized forces or constraints, we obtain the following equation of motion:
\begin{equation}
\label{eq:lagrange_polar_pend}
\begin{split}
\ddot{\theta} = -\frac{g}{L}\sin\theta,
\end{split}
\end{equation}
which, as expected, matches the result derived using Euler’s equation in \cref{eq:euler_pend}.
\end{example}

\subsection{Wheeled Robot Motion Models}
\label{sec:wheeled_robot_models}
Robots are developed in diverse forms, sizes, and configurations, each featuring distinct mobility solutions tailored to specific applications.
Among these, wheeled robots are particularly common because of their excellent mobility and simple design.
In this section, we demonstrate how the concepts from the preceding sections can be applied to two classic and widely used motion models for simple wheeled robots: the \emph{unicycle model} and the \emph{differential drive model}.

\subsubsection{Unicycle Model}
\label{subsubsec:uni}
\begin{marginfigure}
    \centering 
    \includegraphics[width=0.8\linewidth]{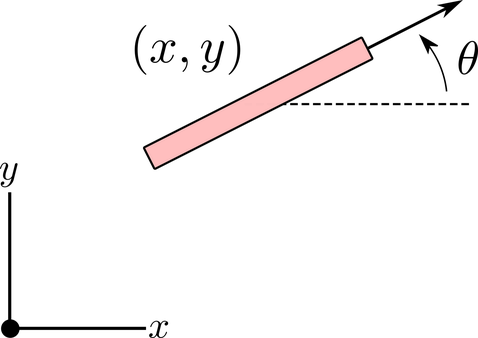}
    \caption{Generalized coordinates for a unicycle.}
    \label{fig:uni} 
\end{marginfigure} 
The unicycle model is one of the simplest kinematic models used for modeling robot motion. 
This model leverages the kinematics of the rolling wheel discussed in \cref{ex:noslipwheel}, essentially assuming the robot is constrained only by a no-slip constraint from a single wheel.
\cref{fig:uni} illustrates a simplified diagram of the generalized coordinates for the unicycle model.

The kinematic model is identical to the one presented in \cref{eq:wheelkinmodel}, namely:
\begin{equation}
    \label{eq:uni}
\begin{bmatrix}
\dot{x} \\ \dot{y} \\ \dot{\theta}
\end{bmatrix} = \begin{bmatrix}
\cos \theta & 0 \\
\sin\theta & 0 \\
0 & 1
\end{bmatrix}\begin{bmatrix}
v \\ \omega
\end{bmatrix},
\end{equation}
where~$v$ represents the forward speed and~$\omega$ denotes the rotational rate.

While the unicycle model may be a simplified representation of the robot's true kinematics, it remains valuable in many contexts where detailed dynamics are unnecessary.
Its main advantage lies in its simplicity, which often enables more computationally efficient algorithms.
In practice, such lower-fidelity models are often used in the early stages of a system’s design or decision-making process, and are later refined or supplemented with more accurate models when higher precision is required.

\subsubsection{Differential Drive Model}
\label{subsubsec:diff_drive}
\begin{marginfigure}
    \centering 
    \includegraphics[width=0.9\linewidth]{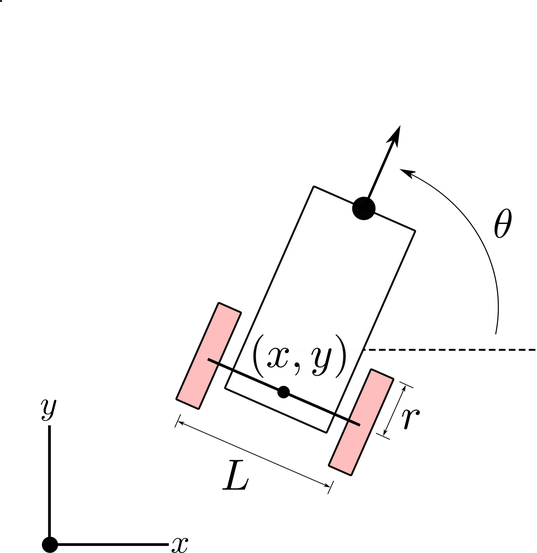}
    \caption{Generalized coordinates for a differential drive robot.} 
    \label{fig:dd} 
\end{marginfigure}
The differential drive model is a variation on the unicycle model from the previous section, with two wheels fixed on a shared rear axle and a passive front wheel that induces no additional kinematic constraints.
This model uses the same generalized coordinates as the unicycle model,~$\q = \vCol{x,\:y,\:\theta}$, but also requires the definition of certain geometric parameters: the width of the rear axle, denoted by $L$, and the radius of the wheels, denoted by~$r$, as illustrated in \cref{fig:dd}.

The differential drive model assumes the wheels roll without slipping, making the derivation of its kinematic constraints similar to that of a single rolling wheel, as discussed in \cref{ex:noslipwheel}. 
The heading vector of each wheel is given by~$\e_v = \vCol{\cos \theta, \: \sin \theta}$, and the lateral direction is~$\e_{v,\perp} = \vCol{\sin \theta, \: -\cos \theta}$. 
Using the lateral direction vector, we define the no-slip kinematic constraints for the wheels as:
\begin{equation*}
\dot{\p}_l^\top \e_{v,\perp} = 0, \quad \dot{\p}_r^\top \e_{v,\perp} = 0,
\end{equation*}
where~$\dot{\p}_l$ and~$\dot{\p}_r$ are the velocity vectors of the left and right wheels, respectively.
Next, we express the wheel velocity vectors~$\dot{p}_l$ and~$\dot{p}_r$ as functions of the generalized coordinates and velocities by leveraging the robot's geometry.
The positions of the left and right wheel centers, denoted as~$\p_l$ and~$\p_r$, respectively, can be computed from the generalized coordinates by:
\begin{equation*}
\begin{split}
    \p_l = \begin{bmatrix}
    x - \frac{L}{2}\sin \theta \\
    y + \frac{L}{2}\cos \theta
    \end{bmatrix}, \quad
    \p_r = \begin{bmatrix}
    x + \frac{L}{2}\sin \theta \\
    y - \frac{L}{2}\cos \theta
    \end{bmatrix}.
\end{split}
\end{equation*}
Taking the time derivative of these positions yields the velocity vectors:
\begin{equation*}
\begin{split}
    \dot{\p}_l = \begin{bmatrix}
    \dot{x} - \dot{\theta}\frac{L}{2}\cos \theta \\
    \dot{y} - \dot{\theta}\frac{L}{2}\sin \theta
    \end{bmatrix}, \quad
    \dot{\p}_r = \begin{bmatrix}
    \dot{x} + \dot{\theta}\frac{L}{2}\cos \theta \\
    \dot{y} + \dot{\theta}\frac{L}{2}\sin \theta
    \end{bmatrix}.
\end{split}
\end{equation*}
After some algebraic manipulation, we find that the no-slip kinematic constraints for each wheel are equivalent:
\begin{equation*}
    \begin{split}
\dot{\p}_l^\top \e_{v,\perp} &= \dot{\p}_r^\top \e_{v,\perp} = \dot{x}\sin \theta - \dot{y} \cos \theta = 0,
    \end{split}
\end{equation*}
indicating that the no-slip constraint for both wheels is redundant, and thus the constraint matches the single wheel constraint from \cref{ex:noslipwheel}. 
This is intuitive because the wheels are rigidly connected; hence, if one wheel cannot move laterally, neither can the other.
The kinematic model for the differential drive model is also identical to the single wheel model in \cref{eq:uni}, but the inputs can now be expressed in a more realistic form relative to the actual geometry of the robot.

In particular, instead of using the forward speed~$v$ and body rotation rate~$\omega$ as inputs, as in \cref{eq:uni}, the differential drive model uses the rotation rates of the left and right wheels,~$\omega_l$ and~$\omega_r$. 
We can derive a relationship between these sets of inputs by considering the geometry of the robot and the no-slip wheel assumption. 
First, denote the position~$\p = \vCol{x, \: y}$ in terms of the wheel center positions by~$\p = \frac{1}{2}(\p_l + \p_r)$, thus the velocity vector is~$\dot{\p} = \frac{1}{2}(\dot{\p}_l + \dot{\p}_r)$.
By the no-slip wheel assumption, the velocity~$v$ can be expressed as~$v = \e_v^\top \dot{p}$, leading to:
\begin{equation*}
\begin{split}
v &= \e_v^\top \dot{\p}, \\
&= \frac{1}{2}\e_v ^\top(\dot{\p}_l + \dot{\p}_r), \\
&= \frac{1}{2}(v_l + v_r), \\
&= \frac{r}{2}(\omega_l + \omega_r), \\
\end{split}
\end{equation*}
where~$r$ is the radius of the wheel and~$v_l$ and~$v_r$ are the speeds of the left and right wheels, respectively. 
Additionally, the no-slip condition on each wheel is given by~$v_l = \e_v^\top \dot{\p}_l$ and~$v_r = \e_v^\top \dot{\p}_r$, expanded as:
\begin{equation*}
\begin{split}
\dot{x}\cos{\theta} + \dot{y}\sin{\theta} - \dot{\theta} \frac{L}{2} &= v_l, \\
\dot{x}\cos{\theta} + \dot{y}\sin{\theta} + \dot{\theta} \frac{L}{2} &= v_r. \\
\end{split}
\end{equation*}
Since~$\dot{x}\cos{\theta} + \dot{y}\sin{\theta} = v$, we simplify these expressions to:
\begin{equation*}
\begin{split}
\frac{L}{2} \dot{\theta} &= v_r - v, \\
\frac{L}{2} \dot{\theta} &= v - v_l. \\
\end{split}
\end{equation*}
Combining these gives:
\begin{equation*}
\begin{split}
L\dot{\theta} &= v_r - v_l, \\
&= r(\omega_r - \omega_l), \\
\end{split}
\end{equation*}
establishing the relationship between the generalized velocity~$\dot{\theta}$ and the wheel rotational speeds.

In summary, the mapping between the inputs can be defined as:
\begin{equation*}
v = \frac{r}{2}(\omega_l + \omega_r), \quad \omega = \frac{r}{L}(\omega_r - \omega_l).
\end{equation*}
which allows us to define the differential drive model:
\begin{equation} 
\label{eq:dd}
\begin{bmatrix}
\dot{x} \\ \dot{y} \\ \dot{\theta} 
\end{bmatrix}
 = 
 \begin{bmatrix}
\frac{r}{2}\cos\theta & \frac{r}{2}\cos\theta \\ 
\frac{r}{2}\sin\theta & \frac{r}{2}\sin\theta \\ 
\frac{r}{L} & -\frac{r}{L}
\end{bmatrix}
\begin{bmatrix}
\omega_r \\ \omega_l
\end{bmatrix}.
\end{equation}
Despite the slight increase in complexity over the unicycle model, this model leverages the geometry of the robot to make the inputs more intuitive. 
This enhancement makes the differential drive model more suitable for certain motion planning and control tasks, as the robot's actuation typically originates from motors attached to the wheels' axles.

More generally, a kinematic state-space model should be interpreted only as a subsystem of a more comprehensive dynamical model.
In particular, kinematic models typically assume direct control over certain motion variables---such as velocity or angular rate---without accounting for how these quantities are generated or constrained by the physical system. 
For more realistic modeling, it is often necessary to extend the kinematic model to include additional integrators in front of the control inputs.

\begin{example}[Dynamic extension of the unicycle model]
\label{ex:dynamic_unicycle}
The unicycle model introduced in~\cref{eq:uni} assumes direct control over the forward velocity~$v$ and angular velocity~$\omega$, with the state defined by the variables~$\tup{x, y, \theta}$.
To reflect the fact that velocity~$v$ is itself the result of integrating an acceleration input~$a$, the model can be extended by treating~$v$ as an additional state, yielding the augmented state~$\tup{x, y, \theta, v}$ and input~$\tup{\omega, a}$. 
The dynamics become:
\begin{equation}
    \label{eq:uni_dynamic}
    \begin{bmatrix}
    \dot{x} \\ \dot{y} \\ \dot{\theta} \\ \dot{v}
    \end{bmatrix}
    =
    \begin{bmatrix}
    v\cos\theta \\ v\sin\theta \\ \omega \\ a
    \end{bmatrix}.
\end{equation}
This dynamic extension accounts for acceleration as a control input and enables the modeling of more realistic scenarios, such as those involving actuation limits.
\end{example}

\subsubsection{Bicycle/Simple Car Model}
\label{subsubsec:kinematic_bicycle}
The bicycle model is a simplified kinematic model commonly used to approximate the motion of vehicles with two front-steered wheels and two rear-driven wheels, such as cars or mobile robots with similar geometry. 
The model captures key steering dynamics while assuming no slip at the contact points of the wheels. 
It is called a “bicycle” model because the two front wheels and two rear wheels are collapsed into a single front and rear wheel aligned on a common axis, forming a virtual two-wheeled vehicle.
Compared to the unicycle and differential drive models introduced earlier, the bicycle model enforces more realistic kinematic constraints on how the system can turn. 
In particular, it captures the fact that the vehicle must steer to follow curved paths, and cannot rotate in place. 
As such, it provides a better approximation for many wheeled systems while still remaining relatively simple.

\Cref{fig:bicycle-model} shows the simplified geometry of the bicycle model. This model can be derived by enforcing nonholonomic constraints on the rolling direction of each wheel and assuming ideal no-slip contact, following the discussion in previous sections.
\Cref{fig:car-model} illustrates how the same kinematic model can be interpreted in the context of a four-wheeled vehicle. The key idea is that both front wheels steer with a common angle $\phi$, and the vehicle moves forward with velocity $v$, subject to the no-slip constraints. These assumptions lead to the following differential equations characterizing the car model:
\begin{marginfigure}
    \centering 
    \includegraphics[width=0.95\linewidth]{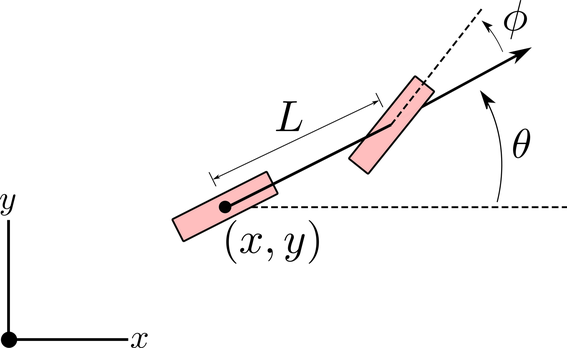}
    \caption{The bicycle model approximates the motion of a four-wheeled vehicle by collapsing each axle into a single wheel, aligned with the vehicle’s centerline. The state consists of the position, $\tup{x, y}$, of the rear axle center and the heading angle, $\theta$. The control inputs are the forward velocity, $v$, and the steering angle, $\phi$.}
    \label{fig:bicycle-model} 
\end{marginfigure} 
\begin{marginfigure}
    \centering 
    \includegraphics[width=0.95\linewidth]{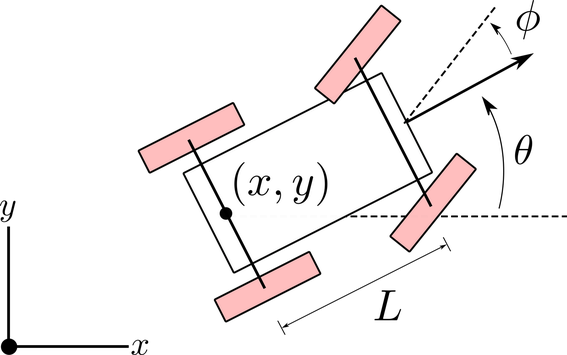}
    \caption{The same bicycle model applied to a car-like vehicle. The control and state definitions are identical to \cref{fig:bicycle-model}, but the visualization makes explicit the mapping between the simplified model and a four-wheeled car.}
    \label{fig:car-model} 
\end{marginfigure} 
\begin{equation} 
\label{eq:car-dynamics}
\begin{split}
    \dot{x} &= v\cos\theta,\\
    \dot{y} &= v\sin\theta,\\
    \dot{\theta} &= \frac{v}{L}\tan\phi, 
\end{split}
\end{equation}
where $\tup{x, y}$ is the position of the rear axle center, $\theta$ is the heading angle, $v$ is the forward speed, $\phi$ is the steering angle, and $L$ is the length of the wheelbase (the distance between the front and rear axles). 
Therefore, we define the state as $\x = \vCol{x, \: y, \: \theta}$ and the control input as $\u = \vCol{v, \:\phi}$.

\section{Simulating Robot Dynamics}
\label{sec:simulating_dynamics}
In \cref{sec:ss_models}, we introduced the concept of a \emph{state space model} to mathematically describe the evolution of a robot's state over time.
In \cref{sec:kin_and_dyn} we demonstrated how a robot's \emph{kinematics and dynamics} are used to derive a state space model that represents its physical motion.
In this section, we present several computational techniques for \emph{simulating} the changes in a robot's state over time.

The state space model in \cref{eq:dynamics-ss} is a general system of ordinary differential equations, which in most cases cannot be solved analytically.
Numerical simulation provides a practical approach to obtaining approximate solutions, allowing us to better understand a robot's dynamics and to test and validate algorithms for robot autonomy.
Typically, when we refer to \emph{simulating} a system, we mean approximately solving an \emph{initial value problem} (IVP) for a system of differential equations, as defined in \cref{eq:dynamics-ss}:
\begin{equation*}
\dot{\x} = h(\x(t), t), \quad \x(t_0) = \x_0,
\end{equation*} 
where~$h(\x(t), t) = f(\x(t), \u(t))$, and the input~$\u(t)$ may either be explicitly defined as a known function of time (e.g., a predefined control sequence), or computed at each time step based on the current state.

The objective of this initial value problem is to find the trajectory~$\x(t)$, starting from~$\x(t_0)$, that satisfies the differential equation\mysidenote{If~$h$ is Lipschitz continuous in~$x(t)$ and continuous in~$t$, the trajectory~$x(t)$ exists and is unique.}.
By the \emph{Fundamental Theorem of Calculus}\mysidenote{This expresses the inverse relationship between differentiation and integration: integrating the derivative~$\dot{\x} = h(\x(t), t)$ over time recovers the original function~$\x(t)$.}, the solution at time~$t$ can be written as:
\begin{equation*}
    \begin{split}
        \x(t)&=\x(t_0)+\int_{t_0}^t h(\x(\tau),\tau)\d\tau.
    \end{split}
\end{equation*}
In general, evaluating this integral analytically for arbitrary functions~$h(\x(t), t)$ is intractable.
Therefore, we typically resort to numerical integration methods that involve a \emph{discretization in time}:
\begin{equation}
    \begin{split}
        \x(t)&=\x(t_0)+\sum_{k=0}^{N-1}\int_{t_k}^{t_{k+1}}h(\x(\tau),\tau)\d\tau,
    \end{split}
\label{eq:discretized_integration}
\end{equation}
where~$t_0 < t_1 < \dots < t_N = t$ define a time grid, and each interval has width~$\Delta t_k = t_{k+1} - t_k$.
This decomposition breaks the continuous integration problem into a sum of smaller integrals over short intervals.
Within each interval~$[t_k, t_{k+1}]$, we can then approximate the integral using various numerical quadrature rules—such as the Euler method, the Midpoint method, or higher-order Runge-Kutta schemes---which trade off computational cost and accuracy.
In this section, we provide a concise introduction to some of the most widely used methods.

\begin{example}[Simple IVP]
To illustrate the different numerical integration methods, we will consider the initial value problem defined below and presented in Algorithm~\ref{alg:ivp_definition} as an example:
\begin{equation*}
\dot{x}(t) = x(t) \sin^2(t), \quad x(0) = 1.
\end{equation*}
Our goal is to approximate the trajectory \( x(t) \) over the interval \( [0, 10] \) using various integration methods.
\begin{listing}[!ht]
\begin{tcolorbox}[colback=gray!10, colframe=gray!50, title=Simple IVP (Running Example), boxrule=0.5mm, arc=0mm, label=alg:running_example]
\begin{minted}[escapeinside=||]{python}
import numpy as np

# Define the derivative function h(x, t)
def h(x, t):
    return x * np.sin(t) ** 2

# Define initial conditions and final time
x0 = 1.0 # Initial state x(t0)
t0 = 0.0 # Initial time t0
tf = 10.0 # Final time tf

|$\Delta$|t = 0.5 # Discretization step
t = np.arange(t0, tf + |$\Delta$|t, |$\Delta$|t) # Array of timestamps

# Compute analytical solution
x_true = x0*np.exp(((t-t0) - np.sin(t-t0)*np.cos(t+t0))/2)

def integrate(h, x0, t, method):
    x = np.zeros((t.size, x0.size))
    x[0] = x0
    for i in range(t.size - 1):
        |$\Delta$|t = t[i + 1] - t[i]
        x[i + 1] = method(h, x[i], t[i], |$\Delta$|t)
    return x

def method(h, x, t, |$\Delta$|t):
    # Implement here numerical integration method

# Test a specific numerical method
x_method = integrate(h, x0, t, method)
\end{minted}
\end{tcolorbox}
\caption{Definition of an illustrative initial value problem. 
The code for this example is available in the repository \colorcode{github.com/StanfordASL/pora-exercises} in the notebook \colorcode{ch01/simulation.ipynb}. 
In the following sections, we will explore different numerical integration methods and implement them in a custom \colorcode{method} function.}
\label{alg:ivp_definition}
\end{listing}

Concretely, we will explore different techniques to approximate the following analytical solution:
\begin{equation*}
    x(t) = x_0 \exp\left( \frac{t - t_0 - \sin(t-t_0)\cos(t+t_0)}{2} \right),
\end{equation*}
    
which, for our specific initial conditions \( x_0 = 1 \) and \( t_0 = 0 \), simplifies to:
\begin{equation*}
x(t) = \exp\left( \frac{t - \sin(t) \cos(t)}{2} \right).
\end{equation*}
\end{example}
    
\subsubsection{Euler Method}
One of the simplest techniques for approximating the integral within each time interval of the discretized problem, as shown in Equation \ref{eq:discretized_integration}, is the Euler method\mysidenote{Named after the Swiss mathematician Leonhard Euler and often referred to as the \emph{forward Euler} method.}.
This method approximates the integral over a short interval $[t_k, t_{k+1}]$ by evaluating the integrand at the beginning of the interval.

Given that the trajectory $\x(t)$ satisfies:
\begin{equation*}
\x(t_{k+1}) = \x(t_k) + \int_{t_k}^{t_{k+1}} h(\x(\tau), \tau) \d\tau,
\end{equation*}
the Euler method approximates this integral by assuming $h(\x(\tau), \tau)$ is constant within the interval, yielding:
\begin{equation*}
\x(t_{k+1}) \approx \x(t_k) + \Delta t \cdot h(\x(t_k), t_k),
\end{equation*}
where $\Delta t = t_{k+1} - t_k$ is the time step.

This approximation corresponds to a first-order Taylor series expansion:
\begin{equation}
\begin{split}
\x(t + \Delta t) &\approx \x(t) + \Delta t \, \dot{\x}(t), \\
&= \x(t) + \Delta t \cdot h(\x(t), t), \\
\end{split}
\end{equation}
which treats the rate of change $\dot{\x}(t)$ as constant across the interval.

Alternatively, the Euler method can be interpreted as a finite difference approximation of the time derivative:
\begin{equation*}
\dot{\x}(t) \approx \frac{\x(t + \Delta t) - \x(t)}{\Delta t}.
\end{equation*}
While computationally inexpensive, the Euler method has limited accuracy due to its reliance on information from the beginning of each interval.
The local truncation error\mysidenote{That is, the error introduced in a single time step.} is of order $\Oc(\Delta t^2)$, and errors can accumulate significantly over long trajectories unless small time steps are used.
 
As a concrete illustration, consider the running example introduced in \cref{alg:running_example}.
A simple implementation of Euler’s method is presented in Algorithm~\ref{alg:euler}.
\begin{listing}[ht!]
\begin{tcolorbox}[colback=gray!10, colframe=gray!50, title=Euler Method, boxrule=0.5mm, arc=0mm]
\begin{minted}[escapeinside=||]{python}
def euler(h, x, t, |$\Delta$|t):
    return x + |$\Delta$|t * h(x, t)

x_euler = integrate(h, x0, t, euler)
\end{minted}
\end{tcolorbox}
\caption{Python implementation of Euler's method.}
\label{alg:euler}
\end{listing}

\subsubsection{Midpoint Method}
The Midpoint method is a refinement of the Euler method that improves accuracy by evaluating the derivative at the midpoint of the time interval, rather than at its beginning.
Recall that Euler’s method approximates the next state using the derivative $\dot{\x}$ evaluated at time $t$:
\begin{equation*}
\x(t + \Delta t) \approx \x(t) + \Delta t \cdot h(\x(t), t).
\end{equation*}
In contrast, the Midpoint method approximates the integral by using the value of the derivative at $t + \frac{\Delta t}{2}$:
\begin{equation}
\label{eq:midpoint-a}
\x(t + \Delta t) \approx \x(t) + \Delta t \cdot h\left( \x\left(t + \frac{\Delta t}{2}\right), t + \frac{\Delta t}{2} \right).
\end{equation}
While this yields a more accurate estimate, it is not yet explicit, since the value $\x(t + \frac{\Delta t}{2})$ is not known in advance.

To resolve this, we approximate the midpoint using a single Euler step of size $\frac{\Delta t}{2}$:
\begin{equation}
\label{eq:midpoint-b}
    \begin{split}
    \x\left(t+\frac{\Delta t}{2}\right) &\approx \x(t) + \frac{\Delta t}{2} \cdot h(\x(t),t).
    \end{split}
\end{equation}
Substituting this estimate into \cref{eq:midpoint-a}, we arrive at the explicit form of the Midpoint method:
\begin{equation*}
    \x(t + \Delta t) \approx \x(t) + \Delta t \cdot h\left( \x(t) + \frac{\Delta t}{2} \cdot h(\x(t), t), \; t + \frac{\Delta t}{2} \right).
\end{equation*}
The Midpoint method improves the local truncation error with respect to the Euler method from $\Oc(\Delta t^2)$ to $\Oc(\Delta t^3)$, providing significantly better accuracy for small step sizes. The improvement comes at the cost of computing the derivative twice per step---once at the start of the interval and once at its midpoint.

As an illustration, a simple implementation for the running example introduced in \cref{alg:running_example} is presented in Algorithm~\ref{alg:midpoint}.
\begin{listing}[ht!]
\begin{tcolorbox}[colback=gray!10, colframe=gray!50, title=Midpoint Method, boxrule=0.5mm, arc=0mm]
\begin{minted}[escapeinside=||]{python}
def midpoint(h, x, t, |$\Delta$|t):
    t_mid = t + |$\Delta$|t / 2
    x_mid = x + (|$\Delta$|t / 2) * h(x, t)
    return x + |$\Delta$|t * h(x_mid, t_mid)

x_midpoint = integrate(h, x0, t, midpoint)
\end{minted}
\end{tcolorbox}
\caption{Python implementation of the Midpoint method.}
\label{alg:midpoint}
\end{listing}

\subsubsection{Runge-Kutta-4 Method}
Both the Euler and Midpoint methods approximate the change in $\x$ over a time step interval by evaluating the derivative $\dot{\x}$ at one or two specific points within the interval.
The Runge-Kutta family of methods generalizes this idea by using multiple evaluations of the derivative to achieve higher accuracy.
One of the most widely used methods in this family is the \emph{fourth-order Runge-Kutta method}\mysidenote{Often abbreviated as RK4.}, which computes four derivative estimates over the interval $[t, t + \Delta t]$:
\begin{equation*}
    \x(t+\Delta t)\approx \x(t)+\frac{\Delta t}{6}(k_1 + 2k_2 + 2k_3 + k_4),
\end{equation*}
where:
\begin{equation*}
    \begin{split}
        k_1&=h(\x(t),\:t), \\
        k_2&=h(\x(t)+\frac{\Delta t}{2}k_1, \:t+\frac{\Delta t}{2}), \\
        k_3&=h(\x(t)+\frac{\Delta t}{2}k_2, \:t+\frac{\Delta t}{2}), \\
        k_4&=h(\x(t)+\Delta tk_3, \:t+\Delta t).
    \end{split}
\end{equation*}

RK4 improves the local truncation error to $\Oc(\Delta t^5)$, offering significantly better accuracy than the Euler or Midpoint methods.
This comes at the cost of evaluating $h(\x, t)$ four times per step, but the method remains computationally efficient and stable for many practical applications.

Using the running example from \cref{alg:running_example}, an implementation of RK4 is shown in Algorithm~\ref{alg:rk4}.
\begin{listing}[ht!]
\begin{tcolorbox}[colback=gray!10, colframe=gray!50, title=RK4 Method, boxrule=0.5mm, arc=0mm]
\begin{minted}[escapeinside=||]{python}
def rk4(h, x, t, |$\Delta$|t):
    k1 = h(x, t)
    k2 = h(x + (|$\Delta$|t / 2) * k1, t + |$\Delta$|t / 2)
    k3 = h(x + (|$\Delta$|t / 2) * k2, t + |$\Delta$|t / 2)
    k4 = h(x + |$\Delta$|t * k3, t + |$\Delta$|t)
    return x + (|$\Delta$|t / 6) * (k1 + 2 * k2 + 2 * k3 + k4)

x_rk4 = integrate(h, x0, t, rk4)
\end{minted}
\end{tcolorbox}
\caption{Python implementation of the RK4 method.}
\label{alg:rk4}
\end{listing}

To compare the performance of these methods, we can visualize their outputs against the analytical solution:
\begin{figure}[h!] 
    \centering 
    \includegraphics[width=0.8\linewidth]{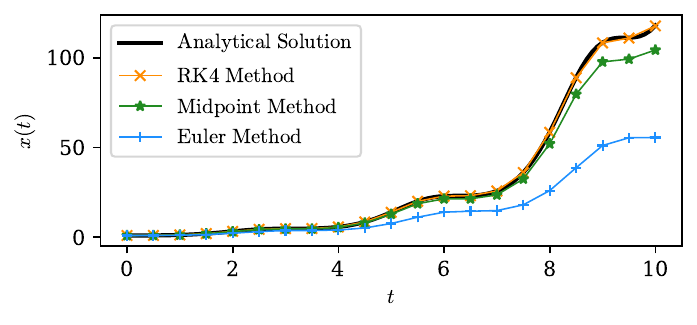}
    \caption{Visual comparison of various numerical integration methods and their approximations compared to the analytical solution for the initial value problem introduced in \cref{alg:running_example}, with $\Delta t = 0.5$.} 
    \label{fig:integration_comparison} 
\end{figure}

It is important to highlight that advanced simulation techniques extend far beyond these numerical integration methods, allowing for the high-fidelity simulation of robotic systems, including visualization in pixel space.
These techniques can incorporate detailed physical modeling, sensor data fusion, and learning-based approaches to create realistic and accurate simulations\mysidenote{One notable example is the use of Neural Radiance Fields (NeRFs) for generating photorealistic scenes.}.
As we will discuss in later chapters on perception, these advancements are crucial for tasks such as robot training, planning under uncertainty, and evaluating the autonomy stack in novel and previously unseen scenarios.

\section{Summary}
In this chapter, we introduced the fundamental principles underlying the modeling and simulation of robotic systems.
We began by introducing state space models, which provide a mathematical framework to describe the evolution of a robot's state over time.
Next, we discussed a robot's kinematics and dynamics, which characterize its motion and the constraints acting on it.
This included a discussion of generalized coordinates and kinematic constraints—both holonomic and nonholonomic—along with the formulation of kinematic models using Pfaffian constraints.
Throughout this chapter, we examined practical examples such as the rolling wheel, the pendulum, and wheeled robots like the unicycle and differential drive models. 
Finally, we introduced numerical integration techniques for simulating robot dynamics over time. We presented the Euler, Midpoint, and Runge-Kutta methods, highlighting their trade-offs and applications through concrete code examples.

\paragraph{To learn more.}
For readers interested in a deeper and more rigorous treatment of the concepts presented in this chapter,~\citet{SicilianoEtAl2007, SicilianoEtAl2008} offer comprehensive and widely adopted references. 
These texts cover the mathematical foundations of robot kinematics, dynamics, and control in greater depth, and provide additional examples, derivations, and exercises that complement and extend the material introduced here.

\section{Exercises}
The starter code for the exercises provided below is available online through GitHub. 
To get started, download the code by running in a terminal window:

\begin{tcolorbox}[colback=gray!10]
\begin{minted}{bash}
    git clone https://github.com/StanfordASL/pora-exercises.git
\end{minted}
\end{tcolorbox}

We denote Problems requiring hand-written solutions and coding in Python with \adjustbox{height=2ex, valign=c}{\includegraphics{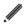}} and \adjustbox{height=2ex, valign=c}{\includegraphics{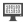}}, respectively.

\subsection*{\adjustbox{height=2ex, valign=c}{\includegraphics{figs/code.png}}\ Problem 1: Numerical Integration Methods}
In \cref{alg:running_example} we introduced a simple IVP and in \cref{fig:integration_comparison} we showed the differences between the Euler, midpoint, and fourth-order Runge-Kutta methods for solving it.
In this exercise we will explore the use of a more advanced numerical integration scheme provided by the SciPy Python library called \colorcode{odeint}.
Open the notebook \colorcode{ch01/exercises/simulation.ipynb} and practice by implementing the dynamics models for a damped pendulum and a bicycle, and then simulating them using \colorcode{odeint}.

\subsection*{\adjustbox{height=2ex, valign=c}{\includegraphics{figs/code.png}}\ Problem 2: Nonholonomic Wheeled Robot Dynamics}
The goal of this exercise is to familiarize yourself with some Python fundamentals that will be used throughout the book, such as NumPy and inheritance, as well as techniques for controlling nonholonomic wheeled robots. 

\begin{marginfigure}
    \centering 
    \includegraphics[width=0.9\linewidth]{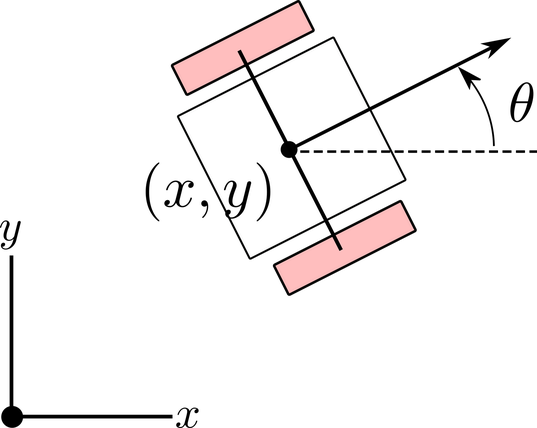}
    \caption{Generalized coordinates for a robot with unicycle kinematics.} 
    \label{fig:robot} 
\end{marginfigure}

Consider a simple robot with two wheels whose state is defined by the position of the center of the axle and the heading angle, shown in~\cref{fig:robot}. 
This robot's motion can be described by the simplest nonholonomic wheeled robot model, the unicycle model.

The \emph{kinematic} model we will use reflects the rolling without side-slip constraint, and is given below in \cref{robot_eq}. 
\begin{equation}
\begin{split}
\dot{x}(t) &= v(t) \cos(\theta(t)), \\
\dot{y}(t) &= v(t) \sin(\theta(t)), \\
\dot{\theta}(t) &= \omega(t).
\end{split}
\label{robot_eq}
\end{equation}
In this model, the robot state is~$\x = [x,y,\theta]^\top$,
where~$[x,y]^\top$ is the Cartesian location of the robot center and~$\theta$ is its heading with respect to the~$x$-axis. 
The robot control inputs are~$\u = [v,\omega]^\top$, where~$v$ is the velocity along the main axis of the robot and~$\omega$ is the angular velocity, subject to the control constraints: 
\[
|v(t)| \leq 0.75 \,\, \mathrm{m/s}, \qquad \text{and} \qquad |\omega(t)| \leq 1.0 \,\, \mathrm{rad/s}.
\]

In this problem, we will demonstrate the use of class inheritance in Python classes and the use of NumPy for vectorized operations. 
The notebook associated with this problem is \colorcode{ch01/exercises/nonholonomic\_wheeled\_robot\_dynamics.ipynb}.

We will be using a \colorcode{Dynamics} base class for two different dynamics models: the wheeled robot dynamics model in this exercise and a double integrator model in the next exercise.
The base class contains two unimplemented functions: \colorcode{step} and \colorcode{rollout}. 
The \colorcode{step} function will propagate the dynamics a single time step with disturbances, and the \colorcode{rollout} function will apply the \colorcode{step} function multiple times to retrieve a trajectory of states over multiple time steps. 
Because the feed-forward dynamics are subject to disturbances, the same control sequence will result in different trajectories. 
We will observe this by executing multiple rollouts of the dynamics using the same control sequence from the same initial state.

In the \colorcode{ch01/exercises/nonholonomic\_wheeled\_robot\_dynamics.ipynb} notebook, complete the \colorcode{RobotDynamics} class. 
Implement the function \colorcode{step} using discrete-time Euler integration with the kinematic equations described in \cref{robot_eq}. 
Then in the same class, implement the function \colorcode{rollout} with two \colorcode{for}-loops, calling the \colorcode{step} function. 
Run the cells that rollout the robot's dynamics and plot the control and state trajectories (this code has been written for you). 

\subsection*{\adjustbox{height=2ex, valign=c}{\includegraphics{figs/code.png}}\ Problem 3: Double Integrator Dynamics}
In this exercise, we consider the double integrator dynamics model:
\begin{equation}
\begin{split}
\dot{x}(t) &= v_x(t), \\
\dot{y}(t) &= v_y(t), \\
\dot{v}_x(t) &= a_x(t), \\
\dot{v}_y(t) &= a_y(t).
\end{split}
\label{doubleint_eq}
\end{equation}
In this model, the robot state is~$\x = [x, y, v_x, v_y]^\top$ and the robot control inputs are~$\u = [a_x, a_y]^\top$.

Notice that in the previous problem, we used a \colorcode{for}-loop to rollout several trajectories of the robot's dynamics. 
In this problem, we will use the same base dynamics class for a \colorcode{DoubleIntegratorDynamics} class, and use batching to reduce the number of \colorcode{for}-loops needed to perform multiple rollouts.
The notebook for this problem is \colorcode{ch01/exercises/double\_integrator\_dynamics.ipynb}.

\begin{enumerate}[label=\roman*.~]

\item To reduce the number of \colorcode{for}-loops needed to perform multiple rollouts, we will batch the dynamics equations applied in the function \colorcode{step}. 
Implement the function \colorcode{step} in the \colorcode{DoubleIntegratorDynamics} class. 

\item Fill in the code in function \colorcode{rollout} in the \colorcode{DoubleIntegratorDynamics} class using the \colorcode{step} function you just wrote. Note that you should only need one \colorcode{for}-loop!

\end{enumerate}
\newpage
\printbibliography[segment=\therefsegment,heading=subbibliography,title={References}]
\chapter{Open-Loop Control \& Trajectory Optimization}
\label{ch:openloop}
\newrefsegment
In Chapter 1, we introduced the state space model as a mathematical formulation for representing a robot’s dynamics. 
These models, typically expressed as systems of differential equations, provide a foundational framework that describes how the state of a robot evolves over time in response to control inputs. 
In particular, we saw how a state space model can be derived from the robot’s kinematics and dynamics, providing a compact yet expressive description of its physical motion.
In this chapter, we turn to a fundamental question: \emph{given a state space model of the robot, how can we determine the control inputs that will drive it to execute a desired behavior?}

\begin{figure}[t]
   \centering 
   \includegraphics[width=0.9\linewidth]{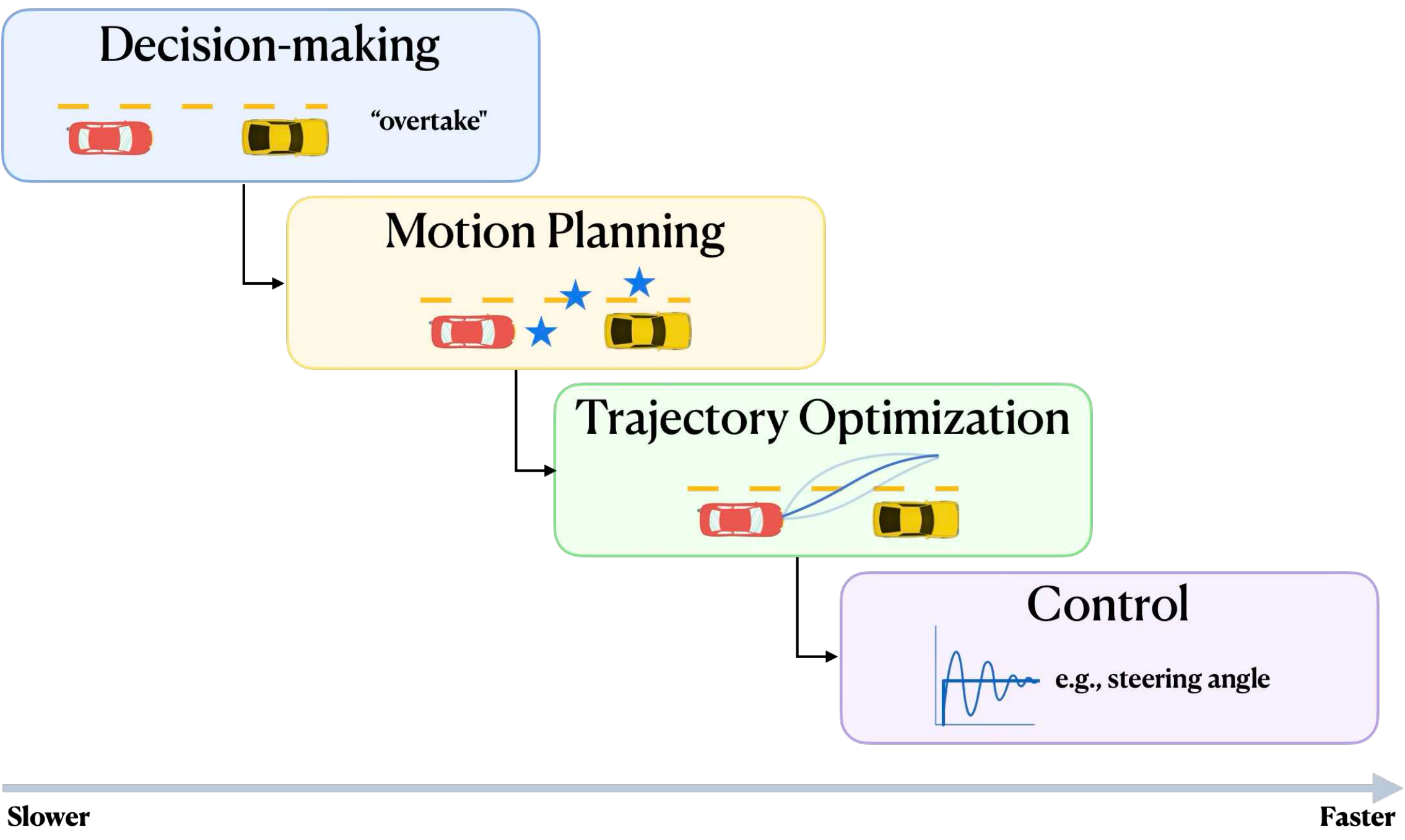}
   \caption{A hierarchical view of the relationship between decision-making, motion planning, trajectory optimization, and control.} 
   \label{fig:hierarchy} 
\end{figure}

As we will see throughout Chapters~\ref{ch:openloop}-\ref{ch:motion-planning}, robots must transform high-level goals into precise physical actions. 
This process is often described hierarchically, spanning \emph{decision-making}, \emph{motion planning}, \emph{trajectory optimization}, and \emph{control} (\cref{fig:hierarchy}).
Each layer plays a distinct role, yet they remain deeply interconnected.

At the higher level of this hierarchy, decision-making governs what tasks the robot should perform to fulfill its objectives.
It involves reasoning over goals, resources, and constraints, often under uncertainty. 
In a self-driving car, for example, this may correspond to deciding when to overtake, yield, or reroute.
Because it involves strategic considerations rather than immediate actuation, decision-making typically unfolds on the order of seconds to minutes.
Ultimately, decision-making defines the high-level objectives that guide subsequent layers of the hierarchy.

Once a high-level decision has been made, motion planning determines how to realize it in the robot’s physical environment. 
This is typically expressed in terms of the robot’s configuration space and involves finding a collision-free path that respects geometric and kinematic constraints.
For instance, motion planning may compute a path for a mobile robot to navigate a cluttered warehouse without collisions.
The timescale of motion planning is often on the order of hundreds of milliseconds to seconds.

Building on this path, trajectory optimization refines it into a time-parameterized trajectory that is dynamically feasible and optimized for performance criteria.
This entails solving continuous optimization problems that incorporate dynamics, actuator limits, and objectives such as energy efficiency, comfort, or safety margins.
The result is a trajectory specifying both the robot’s states and the control inputs needed to realize them over time, ensuring compatibility with the robot's actuation capabilities and dynamic constraints.
Trajectory optimization typically operates on shorter timescales, from tens to hundreds of milliseconds.

Finally, low-level control ensures that the robot faithfully executes the desired trajectory in the physical world, by converting the trajectory into actuation commands. 
Controllers must operate at high frequency, applying feedback to correct deviations caused by disturbances, modeling errors, or sensor noise. 
Whether adjusting wheel torques, joint forces, or thrust vectors, control is what closes the loop between higher-level plans and physical reality.

Together, these layers form the backbone of an autonomous system: decision-making provides strategic guidance, motion planning translates that guidance into feasible paths, trajectory optimization refines those paths into feasible and optimal trajectories, and control ensures that the robot can follow those trajectories in the real world.
In practice, the boundaries between these layers are often blurred.
Trajectory optimization, for instance, may be tightly integrated with planning or even embedded within control loops, while high-level decisions may be informed by the lower-level processes.
For the purposes of this book, we will adopt the hierarchical perspective outlined above, while acknowledging that real-world systems frequently combine or intertwine these processes.

\smallskip
In this chapter, we focus on trajectory optimization as a fundamental tool for computing trajectories that are both feasible and optimal.
While control and motion planning will be revisited in Chapter~\ref{ch:closedloop} and Chapter~\ref{ch:motion-planning}, respectively, our emphasis in this chapter is on the formulation and solution of the trajectory optimization problem.
We begin in~\cref{sec:optimal_control_problem} by introducing the trajectory optimization problem and casting it as a continuous optimization problem.
Building on this foundation, Sections~\ref{sec:indirect_methods} and~\ref{sec:direct_methods} present two major classes of solution strategies---indirect methods, which derive optimality conditions for the continuous problem using tools from the calculus of variations, and direct methods, which discretize and numerically solve the problem as a finite-dimensional nonlinear program.
Finally, in~\cref{sec:differentially_flat_systems}, we explore specialized techniques tailored to certain problem structures, known as differentially flat systems.

\section{The Optimal Control Problem}
\label{sec:optimal_control_problem}
Optimal control theory aims to determine control inputs that drive a dynamical system to satisfy its physical constraints while optimizing a performance criterion.
At a high level, formulating an optimal control problem requires three key components:
\begin{itemize}
    \item A \emph{mathematical model} of the system, typically expressed in state space form.
    \item A description of the \emph{physical constraints} the system must satisfy.
    \item A specification of the \emph{performance criterion} to be optimized.
\end{itemize}

\paragraph{Mathematical model.}

As discussed extensively in~\cref{ch:model-dyn}, the purpose of a mathematical model is to describe how the system's state evolves over time in response to control inputs.
Using the notation from~\cref{ch:model-dyn}, the system dynamics can be expressed as a set of ordinary differential equations:
\begin{equation}
    \dot{\x}(t) = \f\left(\x(t), \u(t)\right),
    \label{eq:system_dynamics}
\end{equation}
where $\x(t) \in \mathbb{R}^n$ is the state of the system at time $t$, $\u(t) \in \mathbb{R}^m$ is the control input, and $\f: \mathbb{R}^n \times \mathbb{R}^m \to \mathbb{R}^n$ defines the state evolution over time.
Throughout this book, we will often refer to $\x(t)$ and $\u(t)$ as the state and control \emph{sequences}, respectively, or equivalently as the state and control \emph{trajectories}.

\paragraph{Physical constraints.}
Once we define the system dynamics, the next step is to specify the physical constraints that the system must satisfy.
These constraints can take several forms, including:
\begin{itemize}
    \item \emph{Initial conditions}, which specify the state at the initial time $t_0$ as $\x(t_0) = \x_0$.
    \item \emph{Final conditions}, which specify the state at the final time $t_f$ as $\x(t_f) = \x_f$ or $\x(t_f) \in \mathcal{X}_f$, where $\mathcal{X}_f$ denotes a set of allowable terminal states.
    \item \emph{State constraints}, which require that the state remains within an allowable set $\mathcal{X}$ for all times $t \in [t_0, t_f]$, that is, $\x(t) \in \mathcal{X}$.
    \item \emph{Control constraints}, which enforce that the control input remains within an allowable set $\mathcal{U}$ for all times $t \in [t_0, t_f]$, that is, $\u(t) \in \mathcal{U}$.
\end{itemize}
Depending on whether the constraints are satisfied or not, we can define the concept of \emph{admissibility} for a control history and state trajectory.
\begin{definition}[Admissible State and Control Sequences]
    A state trajectory $\x(t)$ and a control sequence $\u(t)$ are admissible if they satisfy the state and control constraints at all times, that is:
    \begin{equation*}
    	\x(t) \in \mathcal{X} \quad \text{and} \quad \u(t) \in \mathcal{U}, \quad \forall t \in [t_0, t_f].
    \end{equation*}
\end{definition}
Admissibility is a key concept in optimal control, as it restricts the set of realizable trajectories\sidenote{In practice, this allows numerical methods to focus exclusively on admissible state and control sequences, rather than considering all possible solutions.}.

\paragraph{Performance criterion.}
The final component of an optimal control problem is the performance criterion to be optimized.
An optimal control is defined as one that minimizes (or maximizes) this performance criterion.
In some cases, the performance criterion may be implicitly defined by the problem statement (e.g., minimizing the time to reach a goal state), whereas in other cases, it must be explicitly designed (e.g., driving a car in a way that is comfortable for the passengers).

Throughout this book, we will focus on performance criteria that can be expressed as a cost functional\sidenote{A functional maps functions to real numbers; intuitively, we might say that a functional is a ``function over functions". Here, the cost functional maps a state trajectory and control sequence to a real number representing the overall cost.} of the form:
\begin{equation}
    J(\x(t), \u(t), t) = h(\x(t_f), t_f) + \int_{t_0}^{t_f} g(\x(t), \u(t), t) \, \d t,
    \label{eq:cost_functional}
\end{equation}
where $h: \mathbb{R}^n \times \mathbb{R} \to \mathbb{R}$ is the \emph{terminal} cost and $g: \mathbb{R}^n \times \mathbb{R}^m \times \mathbb{R} \to \mathbb{R}$ is the \emph{running} cost.
The terminal cost $h$ is evaluated at the final time $t_f$ and typically represents a cost associated with the state of the system at that time, such as a penalty for being far from a desired goal state.
The running cost $g$ is integrated over the time interval $[t_0, t_f]$ and represents the instantaneous cost incurred by the system at each time step, for example, the cost of energy consumption, or any other cost associated with the system's operation.
Depending on the problem, the final time $t_f$ may be finite and fixed, finite and free, or infinite\sidenote{In the case of an infinite final time, the terminal cost $h$ is typically ignored and set to zero.}.

\subsubsection{Problem Formulation}

As a result, an optimal control problem can be formulated as follows:
\begin{definitionbox}
\noindent Determine an admissible control sequence $\u^*(t)$ such that the system dynamics:
\begin{equation*}
\dot{\x}(t) = \f(\x(t), \u(t), t),
\end{equation*}
generate a corresponding admissible state trajectory $\x^*(t)$ that minimizes the performance criterion:
\begin{equation*}
J(\x(t), \u(t), t) = h(\x(t_f), t_f) + \int_{t_0}^{t_f} g(\x(t), \u(t), t) \, \d t,
\end{equation*}
where $\u^*(t)$ and $\x^*(t)$ are referred to as the \emph{optimal control sequence} and \emph{optimal state trajectory}, respectively.
\end{definitionbox}

This problem can be formally posed as the following optimization problem:
\begin{equation}
    \begin{aligned}
        \underset{\u(t)}{\text{minimize}} \quad & J(\x(t), \u(t), t), \\
        \text{subject to} \quad & \dot{\x}(t) = \f(\x(t), \u(t), t), \\ 
        & \x(t_0) = \x_0, \quad 
        \x(t_f) = \x_f,\\
        & \u(t) \in \mathcal{U}, \quad 
        \x(t) \in \mathcal{X}, \quad 
        t_0 < t < t_f.
    \end{aligned}
    \label{eq:optimal_control_problem}
\end{equation}
This general formulation serves as the starting point for the optimal control methods developed in the remainder of this book.
There are also several important attributes related to the solution of this problem that are worth highlighting:
\begin{enumerate}
	\item \emph{Existence:} A solution to an optimal control problem is not guaranteed to exist; there may be no control history that is both admissible and optimal.
	\item \emph{Uniqueness:} Even when a solution exists, it may not be unique. Multiple admissible control inputs can yield the same performance. While this can pose challenges for numerical algorithms, it also provides flexibility in selecting among equally good solutions depending on the application.
	\item \emph{Optimality:} The objective of optimal control is to find a control sequence that outperforms \emph{all} other admissible candidates. Thus, optimal control is interested in \emph{global} optimality, as opposed to \emph{local} optimality.
\end{enumerate}

\begin{example}[Autonomous Racing Optimal Control]
    \label{ex:auto_racing_ocp}
    Consider an autonomous racing scenario in which the goal is to complete a lap of a known course in the shortest possible time.
    We can formulate this as a finite-horizon optimal control problem, where the objective is to minimize the final time $t_f$ required to reach a designated goal position $(x_{\text{goal}}, y_{\text{goal}})$, subject to the constraint that the vehicle must remain on the track at all times.
    Let $\statespace_{\mathbb{course}}$ denote the set of admissible states that correspond to positions on the course.
    
    Suppose the vehicle is modeled using the simple kinematic car model from~\cref{eq:car-dynamics}, with state $\x = \vCol{x, \: y, \: \theta}$ representing position and heading, and control inputs $\u = \vCol{v, \:\phi}$ representing forward speed and steering angle, respectively. 
    The resulting optimal control problem is:
    \begin{equation*}
        \begin{aligned}
        \underset{v(t),\, \phi(t)}{\text{minimize}} \quad & t_f,\\
        \subjectto & \dot{x} = v \cos\theta, \quad \dot{y} = v \sin\theta, \quad \dot{\theta} = \frac{v}{L} \tan\phi,\\
        & \x \in \statespace_{\text{course}}, \quad \u \in \controlspace,\\
        & \x(t_0) = \x_0, \quad (x(t_f), y(t_f)) = (x_\text{goal}, y_\text{goal}).
        \end{aligned}
    \end{equation*}
    In this formulation, the cost depends only on the final time and not directly on the state or control at intermediate points.
    As a result, the time-optimal solution $(\x(t), \u(t))$ will lie on the boundaries of the admissible control and state sets.
    In practice, this means the vehicle will operate at full throttle and steer at the physical limits to cut the lap time, a strategy that achieves optimality mathematically—but may not make for a smooth or comfortable ride.
\end{example}

Throughout this book, we will see that the solution to the optimal control problem can take different forms, depending on whether and how it incorporates feedback from the current state of the system.
At the most fundamental level, we distinguish between \emph{open-loop} and \emph{closed-loop} control.
While closed-loop control will be the focus of~\cref{ch:closedloop}, this chapter addresses open-loop control.

\paragraph{Open-loop control.}
If the optimal control is computed purely as a function of time for a given initial state,
\begin{equation}
\u^*(t) = \ell(\x(t_0), t),
\end{equation}
it is said to be in \emph{open-loop} form.
In the context of trajectory optimization, restricting the focus to open-loop strategies is natural, as they balance computational efficiency---computing open-loop sequences is faster than computing closed-loop policies---with effectiveness, since robustness can be endowed through closed-loop tracking or by re-optimizing the trajectory in a receding-horizon fashion, as in Model Predictive Control\sidenote{Discussed further in~\cref{ch:closedloop}.}.

\medskip
\noindent Having introduced the optimal control problem, we now turn to methods for computing optimal open-loop solutions.
Fundamentally, Problem~\eqref{eq:optimal_control_problem} is an \emph{infinite-dimensional} optimization problem, where the optimization variables are functions of time.
In practice, solution methods must rely on discretization strategies, thereby approximating the infinite-dimensional problem with a finite-dimensional one.
Different discretization approaches give rise to distinct families of methods.

Broadly speaking, two main classes of methods exist: \emph{indirect methods} and \emph{direct methods}.
\emph{Indirect methods} follow an ``optimize-then-discretize'' paradigm: they first derive necessary conditions for optimality---typically in the form of boundary-value problems involving adjoint variables---and then apply numerical techniques to solve these conditions.
In contrast, \emph{direct methods} take a ``discretize-then-optimize'' approach: the state and control sequences are parameterized using finite-dimensional representations, and the resulting finite-dimensional optimization problem is solved numerically.

Beyond these two general classes, some systems admit further structural simplifications.
In particular, \emph{differentially flat systems} allow trajectories to be described in terms of a small set of variables that fully capture the system’s evolution\sidenote{These variables are commonly known as \emph{flat outputs}.}.
This property enables efficient trajectory generation and optimization, making these systems especially relevant in mobile robotics and aerospace applications.

The remainder of this chapter examines these three classes of methods in detail: indirect methods in~\cref{sec:indirect_methods}, direct methods in~\cref{sec:direct_methods}, and trajectory optimization for differentially flat systems in~\cref{sec:differentially_flat_systems}.

\subsection{Indirect Methods}
\label{sec:indirect_methods}
Indirect methods provide a principled framework for solving optimal control problems by drawing on ideas from calculus of variations (CoV)\sidenote{Calculus of variations extends the principles of classical calculus from functions to functionals. 
The central idea is to study how small perturbations---called \emph{variations}---of a candidate function influence the value of the functional.
By analyzing the first- and higher-order effects of these variations, one can derive necessary conditions for optimality.}.
For a comprehensive treatment of the calculus of variations and its applications in optimal control theory, we refer the reader to \citet{Kirk2004}.
At their core, indirect methods rely on the derivation of necessary optimality conditions (NOCs) that any solution must satisfy and then leverage numerical techniques to compute solutions consistent with these conditions.
In this way, the NOCs serve as the bridge between the continuous-time formulation of an optimal control problem and its numerical resolution.

Before turning to the derivation of such conditions for \emph{infinite}-dimensional optimization problems, let us first review key concepts from \emph{finite}-dimensional optimization, which will serve as a foundation for the discussion ahead.

\subsubsection{NOCs for Unconstrained Nonlinear Optimization Problems}
Consider the following \emph{finite}-dimensional optimization problem:
\begin{equation}
\minimize[\x \in \R^n] f(\x),
\label{eq:finite_dimensional_optimization_problem}
\end{equation}
where $f: \R^n \to \R$ is assumed continuously differentiable, i.e., $f \in C^1$.
We wish to identify the NOCs that any minimizer—local or global—must satisfy.

The key intuition is that at a local minimizer, no infinitesimal perturbation of the decision variable should decrease the objective. 
Formally, this requires analyzing how $f$ changes under small variations around a candidate minimizer $\x^*$.

\paragraph{First-order necessary condition.}
Let $\x^* \in \R^n$ be a local minimizer.
If $f \in C^1$, we can use gradients and Taylor series expansions to characterize the behavior of $f$ near $\x^*$.
For a small perturbation $\Delta x$, the cost variation is, up to first order:
\begin{equation*}
f(\x^* + \Delta x) - f(\x^*) \approx \nabla f(\x^*)^\top \Delta x.
\end{equation*}
If $\x^*$ is a local minimizer, then for sufficiently small $\Delta x$, the first-order term must be non-negative\sidenote{This is because, if we were to decrease the cost by perturbing $\x^*$ by $\Delta x$, then $\x^*$ would not be a local minimizer.}:
\begin{equation*}
\nabla f(\x^*)^\top \Delta x = \sum_{i=1}^n \frac{\partial f(\x^*)}{\partial x_i} \Delta x_i \geq 0.
\end{equation*}
In particular, by taking $\Delta \x$ to be positive and negative multiples of the coordinate unit vectors, that is, vectors having all components equal to zero except for one component equal to one, we obtain simultaneously:
\begin{equation*}
\frac{\partial f(\x^*)}{\partial x_i} \geq 0 \quad \text{and} \quad \frac{\partial f(\x^*)}{\partial x_i} \leq 0, \qquad i=1,\ldots,n,
\end{equation*}
which forces the condition:
\begin{equation*}
\frac{\partial f(\x^*)}{\partial x_i} = 0, \qquad i=1,\ldots,n,
\end{equation*}
or, more compactly:
\begin{equation*}
\nabla f(\x^*) = 0.
\end{equation*}
Thus, any local minimizer $\x^*$ must be a stationary point of $f$.

\paragraph{Second-order necessary condition.}
Assuming $f \in C^2$, consider again the Taylor expansion of $f$ around a local minimizer $\x^*$, this time up to second order:
\begin{equation*}
f(\x^* + \Delta x) - f(\x^*) \approx \nabla f(\x^*)^\top \Delta x + \frac{1}{2} \Delta x^\top \nabla^2 f(\x^*) \Delta x.
\end{equation*}
For $\x^*$ to be a local minimizer, the second-order variation must be nonnegative for all sufficiently small $\Delta \x$, that is:
\begin{equation*}
\nabla f(\x^*)^\top \Delta x + \frac{1}{2} \Delta x^\top \nabla^2 f(\x^*) \Delta x \geq 0.
\end{equation*}
Using the first-order condition $\nabla f(\x^*)=0$, the linear term vanishes, leaving:
\begin{equation*}
\Delta x^\top \nabla^2 f(\x^*) \Delta x \geq 0.
\end{equation*}
Thus, the Hessian $\nabla^2 f(\x^*)$ must be positive semidefinite at any local minimizer.

\begin{theorem}[Necessary Conditions for Unconstrained Local Minimizers]
Let $\x^*$ be a local minimizer of $f: \R^n \to \R$. If $f \in C^1$ in an open set containing $\x^*$, then:
\begin{equation}
\nabla f(\x^*) = 0 \quad \text{(first-order NOC)}.
\end{equation}
If, in addition, $f \in C^2$, then:
\begin{equation}
\nabla^2 f(\x^*) \succeq 0 \quad \text{(second-order NOC)}.
\end{equation}
\end{theorem}

\subsubsection{NOCs for Constrained Nonlinear Optimization Problems}
Having introduced the NOCs for unconstrained problems, this section extends the discussion to optimization problems subject to constraints.
The definition of optimality conditions in the constrained setting requires the introduction of auxiliary variables, known as \emph{Lagrange multipliers}. 
These variables are associated with the constraints and facilitate the characterization of optimal solutions while providing insights into the sensitivity of the optimal cost with respect to perturbations in the constraints.
In this section, we limit our discussion on the theory of Lagrange multipliers to the case of equality constrained optimization.
For a comprehensive treatment of optimality conditions in finite-dimensional optimization, the reader is referred to \citet{Bertsekas2016}.

Consider the following constrained optimization problem:
\begin{equation}
    \begin{aligned}
    \minimize[\x \in \R^n] & f(\x), \\
    \subjectto & h_i(\x) = 0, \quad i = 1, \ldots, m,
    \end{aligned}
\label{eq:finite_dimensional_optimization_problem_constrained}
\end{equation}
where $f: \R^n \to \R$ and $h_i: \R^n \to \R$ are continuously differentiable.

For compactness, define the constraint function $h: \R^n \to \R^m$ as:
\begin{equation}
h(\x) = (h_1(\x), \ldots, h_m(\x))^\top,
\end{equation}
so that the constraints can be written simply as $h(\x) = 0$.

The \emph{Lagrange multiplier theorem} for equality-constrained optimization states that, if $\x^*$ is a local minimizer, then there exist scalars $\lambda_1^*, \ldots, \lambda_m^*$, called \emph{Lagrange multipliers}, such that:
\begin{equation}
    \nabla f(\x^*) + \sum_{i=1}^m \lambda_i^* \nabla h_i(\x^*) = 0.
\label{eq:lagrange_multiplier_condition}
\end{equation}
To interpret this condition, observe that the cost gradient $\nabla f(\x^*)$ must be orthogonal to the subspace of \emph{first-order feasible variations}:
\begin{equation*}
V(\x^*) \definedas \{\Delta \x \, | \, \nabla h_i(\x^*)^\top \Delta \x = 0, \quad i = 1, \ldots, m\}.
\end{equation*}
This subspace consists of all variations $\Delta \x$ that preserve feasibility to first order \sidenote{That is, variations for which $\x = \x^* + \Delta \x$ satisfies $h(\x) = 0$ to first order.}.
Thus, condition~\eqref{eq:lagrange_multiplier_condition} ensures that the first-order cost variation $\nabla f(\x^*)^\top \Delta \x = 0$ for all $\Delta \x \in V(\x^*)$.
This statement is analogous to the $\nabla f(\x^*) = 0$ condition of unconstrained optimization.

Formally, the necessary conditions for equality constrained optimality are summarized as follows:
\begin{theorem}[Lagrange Multiplier Theorem --- Necessary Conditions for Equality Constrained Local Minimizers]
    Let $\x^*$ be a local minimizer of $f: \R^n \to \R$ subject to the equality constraints $h_i(\x) = 0$, $i = 1, \ldots, m$, and assume the constraint gradients $\nabla h_1(\x^*), \ldots, \nabla h_m(\x^*)$ are linearly independent.
    Then there exists a unique vector $[\lambda_1^*, \ldots, \lambda_m^*]^\top$, called the Lagrange multiplier vector, such that:
    \begin{equation}
        \nabla f(\x^*) + \sum_{i=1}^m \lambda_i^* \nabla h_i(\x^*) = 0.
    \end{equation}
\end{theorem}

It is often convenient to express these conditions using the \emph{Lagrangian function} $\lagrangian : \R^{n+m} \to \R$ defined as:
\begin{equation} 
    \label{eq:lagrangian}
    \lagrangian(\x,\blam) \definedas f(\x) + \sum_{i=1}^m \lambda_i h_i(\x).
\end{equation}
The first-order NOCs for a local minimum $\x^*$ then take the compact form:
\begin{equation} 
\label{eq:finite-noc}
    \begin{split}
    \nabla_{\x} \lagrangian(\x^*,\blam^*) &= 0, \\
    \nabla_{\blam} \lagrangian(\x^*,\blam^*) &= 0,
    \end{split}
\end{equation}
where $\nabla_{\x} \lagrangian$ and $\nabla_{\blam} \lagrangian$ denote the gradients with respect to $\x$ and $\blam$, respectively, and where the system in~\eqref{eq:finite-noc} consists of $n + m$ equations in $n + m$ unknowns---namely, the $n$ components of $\x^*$ and the $m$ components of $\blam^*$.

\medskip
In practice, optimality conditions serve as a powerful tool to \emph{filter} candidate solutions for global or local minima and often form the foundation of numerical optimization algorithms.
For instance, in the unconstrained case of Problem~\eqref{eq:finite_dimensional_optimization_problem}, one might (i) find all stationary points by solving $\nabla f(\x) = 0$, and (ii) apply the second-order test by checking $\nabla^2 f(\x) \succeq 0$ at each candidate.
This same philosophy extends naturally to infinite-dimensional optimal control problems, where any candidate solution must satisfy the corresponding NOCs.
However, as we move to infinite-dimensional problems, the nature of the NOCs changes significantly: rather than yielding algebraic equations as in the finite-dimensional case, the NOCs for optimal control take the form of differential equations.

\subsubsection{Pontryagin's Minimum Principle}
\label{sec:pontryagins_minimum_principle}
Extending the concept of necessary optimality conditions to infinite-dimensional problems leads to Pontryagin’s Minimum Principle (PMP), a cornerstone of optimal control theory.
Specifically, the PMP generalizes the finite-dimensional NOCs to the infinite-dimensional setting.

Consider the problem of finding an admissible control $\u^*(t) \in \mathcal{U}$ that drives the system:
\begin{equation}
\dot \x(t) = \dynmodel(\x(t), \u(t), t), 
\label{eq:pmp_control_system}
\end{equation}
along a trajectory that minimizes the cost functional:
\begin{equation*}
J(\x(t), \u(t), t) = h(\x(t_f), t_f) + \int_{t_0}^{t_f} g(\x(t), \u(t), t) \, \d t.
\end{equation*}

To derive the NOCs, we define the \emph{Hamiltonian}, the analog of the Lagrangian in finite-dimensional optimization:
\begin{equation} 
    \label{eq:hamiltonian}
    \hamiltonian(\x(t), \u(t), \p(t), t) \definedas g(\x(t), \u(t), t) + \p(t)^\top\dynmodel(\x(t), \u(t), t),
\end{equation}
where $\p(t) \in \R^n$ is the \emph{costate}\sidenote{The term \emph{costate} highlights that there is one costate associated with each state variable, analogous to Lagrange multipliers in finite-dimensional optimization.} vector.

Similarly to the finite-dimensional case, where necessary conditions for optimality are derived by considering the cost increment $\Delta f = f(\x + \Delta \x) - f(\x)$ in response to a perturbation $\Delta \x$, here we analyze the increment $\Delta J$ under \emph{variations} around a candidate function.

\begin{theorem}[Pontryagin's Minimum Principle; for a comprehensive treatment, we refer the reader to Chapter 5 in \citet{Kirk2004}]
\label{thm:pontryagins_minimum_principle}
Let $\u^*(t)$ be an optimal control with associated state trajectory $\x(t)$ for the system in~\eqref{eq:pmp_control_system} over $[t_0, t_f]$.
Then there exists a costate vector $\p^*(t)$ such that, for all $t \in [t_0, t_f]$, the following conditions hold:
\begin{equation} 
\label{eq:pontryagins_minimum_principle}
    \begin{split}
    \dot{\x}^*(t) &= \frac{\partial{\hamiltonian}}{\partial{\p}}(\x^*(t), \u^*(t), \p^*(t), t), \\
    \dot{\p}^*(t) &= -\frac{\partial{\hamiltonian}}{\partial{\x}}(\x^*(t), \u^*(t), \p^*(t), t), \\
    \u^*(t) &= \argmin_{\u \in \mathcal{U}} \hamiltonian(\x^*(t), \u, \p^*(t), t),
    \end{split}
\end{equation}
along with the boundary conditions:
\begin{equation}
\label{eq:pontryagins_minimum_principle_boundary_conditions}
\begin{aligned}
    & \left[\frac{\partial h}{\partial \x}\left(\x^*(t_f), t_f\right) - \p^*(t_f)\right]^\top \delta \x_f \\
    & + \left[\hamiltonian(\x^*(t_f), \u^*(t_f), \p^*(t_f), t_f) + \frac{\partial h}{\partial t}\left(\x^*(t_f), t_f\right)\right] \delta t_f = 0,
\end{aligned}
\end{equation}
where $\delta \x_f$ and $\delta t_f$ denote the variations of the final state and time, respectively \sidenote{As we will discuss in the remainder of this section, the boundary conditions in~\cref{eq:pontryagins_minimum_principle_boundary_conditions} depend on whether the final state and time are fixed (i.e., $\delta \x_f = 0$ or $\delta t_f = 0$) or free (i.e., $\delta \x_f$ or $\delta t_f$ are arbitrary).}.
\end{theorem}
Equations~\eqref{eq:pontryagins_minimum_principle} constitute the necessary conditions for optimality.
They form a system of $2n$ first-order differential equations---$n$ for the state and $n$ for the costate---together with $m$ algebraic equations defining the control input.
Solving these equations produces $2n$ constants of integration.
Half of these constants are determined by the initial conditions $\x^*(t_0) = \x_0$.
The remaining $n$ (or $n+1$, if the final time is free) are specified by the boundary conditions in~\cref{eq:pontryagins_minimum_principle_boundary_conditions}.
This results in a two-point boundary value problem, which may be solved analytically in special cases, or numerically using methods such as shooting or collocation\cite{Hertling1970}.

In practice, once the initial state is fixed, the boundary conditions are obtained by substituting the appropriate assumptions into~\cref{eq:pontryagins_minimum_principle_boundary_conditions}.
Common cases include:

\paragraph{Fixed final time and fixed final state.} If both $t_f$ and $\x(t_f)$ are fixed, then $\delta t_f = 0$ and $\delta \x_f = 0$, leaving the sole boundary condition:
\begin{equation*}
\x^*(t_f) = \x_f.
\end{equation*}

\paragraph{Fixed final time and free final state.} 
If $t_f$ is fixed but $\x(t_f)$ is free, then $\delta t_f = 0$ while $\delta \x_f$ is arbitrary.
Hence, the boundary condition is:
\begin{equation*}
\frac{\partial h}{\partial \x}\left(\x^*(t_f), t_f\right) - \p^*(t_f) = 0.
\end{equation*}

\paragraph{Free final time and fixed final state.} 
If $\x(t_f)$ is fixed but $t_f$ is free, then $\delta \x_f = 0$ while $\delta t_f$ is arbitrary.
Thus, the boundary condition is:
\begin{equation*}
\hamiltonian(\x^*(t_f), \: \u^*(t_f), \: \p^*(t_f), \: t_f) +\frac{\partial{h}}{\partial{t}}(\x^*(t_f), \: t_f) = 0.  
\end{equation*}

\paragraph{Free final time and free final state.} 
If both $\x(t_f)$ and $t_f$ are free, then $\delta \x_f$ and $\delta t_f$ are arbitrary, and both coefficients in~\cref{eq:pontryagins_minimum_principle_boundary_conditions} must be set to zero.
That is:
\begin{equation*}
\begin{split}
&\frac{\partial h }{\partial \x} (\x^* (t_f), \: t_f) - \p^*(t_f) = 0, \quad \text{($n$ equations)}\\
&\hamiltonian(\x^*(t_f), \: \u^*(t_f), \: \p^*(t_f), \: t_f) +\frac{\partial{h}}{\partial{t}}(\x^*(t_f), \: t_f) = 0,  \quad \text{($1$ equation)}.  
\end{split}
\end{equation*}
While these four cases cover many problems of practical interest, more general boundary conditions can be found in~\citet{Kirk2004}.

\subsubsection{Solving a Two-Point Boundary Value Problem}
\label{sec:solving_two_point_boundary_value_problem}
Finding solutions that satisfy the necessary optimality conditions in \cref{eq:pontryagins_minimum_principle} is a nontrivial task, as these must simultaneously satisfy a system of $2n$ differential equations together with boundary conditions imposed at both $t_0$ and $t_f$.
This type of problem, where conditions are specified at two distinct points in time, is known as a \emph{two-point boundary value problem} (TPBVP).

Over the years, a number of numerical procedures have been developed for solving TPBVPs. 
Two broad classes of approaches are commonly used:
\begin{itemize}
    \item \textbf{Shooting methods}, which reformulate the TPBVP as an initial value problem by guessing the unknown boundary conditions (e.g., the initial costate), simulating the system forward, and then iteratively adjusting the guess until the terminal boundary conditions are satisfied. 
    Although conceptually straightforward, shooting methods may suffer from numerical instability, especially for long time horizons.  

    \item \textbf{Collocation methods}, whereby the solution is approximated by a parametric function with unknown parameters at a set of discrete points (called collocation points).
    These methods turn the TPBVP into a large system of nonlinear algebraic equations that can be solved using computational techniques. 
    Collocation methods are robust and widely used in practice because they avoid the instability issues of shooting methods.  
\end{itemize}
Modern scientific computing environments provide high-level implementations of these ideas.  
For example, the \colorcode{scikits.bvp\_solver} package in Python or the function \colorcode{bvp4c} in MATLAB implement numerical algorithms for solving TPBVPs with relatively little effort from the user.  

Most solvers assume that the system of necessary conditions in \cref{eq:pontryagins_minimum_principle}, along with its boundary conditions, can be expressed in the standard form: 
\begin{equation} 
\label{eq:standard-tpbvp}
\dot{\z} = g(\z, \: t), \quad l(\z(t_0), \: \z(t_f)) = 0,
\end{equation}
where $\z(t)$ collects the unknown functions (such as states and costates), $g$ encodes their dynamics, and $l$ encodes the two-point boundary constraints. 

To illustrate how TPBVPs can be solved in practice, consider the toy dynamics:
\begin{equation*}
\dot{\z}(t) =
\begin{bmatrix}
\dot{z}_1(t) \\
\dot{z}_2(t)
\end{bmatrix} = 
\begin{bmatrix}
z_2(t) \\
-\,z_1(t)
\end{bmatrix},
\end{equation*}
with boundary conditions $z_1(t_0) = 0$ and $z_1(t_f) = -2$. The boundary conditions can equivalently be expressed in standard form as:
\begin{equation*}
l(\z(t_0), \z(t_f)) =
\begin{bmatrix}
z_1(t_0) \\
z_1(t_f) + 2
\end{bmatrix} = 0.
\end{equation*}

In Python, the system dynamics $g(\z,t)$ and boundary conditions $l(\z(t_0), \z(t_f))$ can be passed directly to \colorcode{solve\_bvp} as shown in Algorithm~\ref{alg:solvebvp_toy}.

Many optimal control problems can, in fact, be cast into the standard TPBVP form in \cref{eq:standard-tpbvp} and solved directly with off-the-shelf BVP solvers such as \colorcode{solve\_bvp}, sometimes after simple reformulations. 
Common cases include problems with conditions at special points, such as free-end problems, switching points, interface points, or discontinuities.
\cref{ex:ocp} illustrates these ideas in the context of a free-final-time problem.

\begin{listing}[ht!]
\begin{tcolorbox}[colback=gray!10, colframe=gray!50, 
    title=Solving a TPBVP with \texttt{solve\_bvp}, boxrule=0.5mm, arc=0mm]
\begin{minted}[escapeinside=||]{python}
from scipy.integrate import solve_bvp
import numpy as np

# Dynamics: |$\dot{z} = g(z,t)$|
def g(t, z):
    return np.vstack((z[1], -z[0]))

# Boundary conditions: |$l(z(t_0), z(t_f))=0$|
def l(z0, zf):
    return np.array([z0[0], zf[0] + 2])

# Time mesh and initial guess for |$z(t)$|
t_mesh = np.linspace(0, 4, 5)
z_guess = np.zeros((2, t_mesh.size))

# Solve TPBVP
sol = solve_bvp(g, l, t_mesh, z_guess)
z_sol = sol.sol(np.linspace(0, 4, 100))
\end{minted}
\end{tcolorbox}
\caption{Example usage of \colorcode{solve\_bvp} for a TPBVP in standard form.
The code for this example is available in the repository \colorcode{github.com/StanfordASL/pora-exercises} in the notebook \colorcode{ch02/tpbvp.ipynb}.}
\label{alg:solvebvp_toy}
\end{listing}

\begin{example}[Free Final Time Optimal Control Problem; see Example 6.1 in \citet{How2008}] 
    \label{ex:ocp}
    \theoremstyle{definition}
    Consider the double integrator system:
    \begin{equation*}
        \ddot{x} = u,
    \end{equation*}
    where $x \in \R$ is the state and $u \in \R$ is the control input.
    The control objective is to find a trajectory that minimizes the cost:
    \begin{equation*}
    J(x, u) = \frac{1}{2}\alpha t_f^2 + \int_{0}^{t_f} \frac{1}{2}\beta u^2(t) \d t,
    \end{equation*}
    and satisfies the boundary conditions:
    \begin{equation*}
    x(0) = 10,\quad \dot{x}(0) = 0,\quad x(t_f) = 0,\quad \dot{x}(t_f) = 0.
    \end{equation*}
    This is a free final time problem with fixed boundary conditions on the state.
    The cost penalizes both the duration of the maneuver (through the $\alpha t_f^2$ term) and the control effort (through the integral of $u^2$), with the trade-off governed by the weights $\alpha$ and $\beta$.
    
    We can equivalently express the double integrator dynamics as a first-order system of differential equations by setting $x_1 = x$ and $x_2 = \dot{x}$:
    \begin{equation*}
    \dot{x}_1 = x_2, \quad \dot{x}_2 = u,
    \end{equation*}
    so that the state vector is $\x = \vCol{x_1, \: x_2}$ and the boundary conditions become:
    \begin{equation*}
    x_1(0) = 10, \quad x_2(0) = 0,\quad x_1(t_f) = 0,\quad x_2(t_f) = 0.
    \end{equation*}
    The Hamiltonian is given by:
    \begin{equation*}
    \hamiltonian = \frac{1}{2}\beta u^2 + p_1x_2 + p_2u,
    \end{equation*}
    where $p_1$ and $p_2$ are the costate variables. 
    Next, we construct the NOCs from \cref{eq:pontryagins_minimum_principle} by taking the partial derivatives of $\hamiltonian$ with respect to $p$, $x$, and $u$:
    \begin{equation*}
    \begin{split}
    \dot{x}_1^* &= x_2^*, \\
    \dot{x}_2^* &= u^*, \\
    \dot{p}_1^* &= 0, \\
    \dot{p}_2^* &=-p_1^*, \\
    0 & = \beta u^* + p_2^*.
    \end{split}
    \end{equation*}
    Thus, from the last condition, the optimal control satisfies:
    \begin{equation*}
        u^* = -\frac{1}{\beta}p_2^*.
    \end{equation*}
    Since this is a free-final-time problem with fixed terminal state, the boundary conditions for the NOCs are given by:
    \begin{equation*}
    \begin{split}
    &x_1^*(0) = 10, \\
    &x_2^*(0) = 0, \\
    &x_1^*(t_f) = 0, \\
    &x_2^*(t_f) = 0, \\
    &\frac{1}{2}\beta u^*(t_f)^2 + p_1^*(t_f)x_2^*(t_f)+p_2^*(t_f)u^*(t_f) + \alpha t_f = 0.
    \end{split}
    \end{equation*}
    However, the necessary conditions obtained above do not immediately match the “standard” form required by numerical TPBVP solvers in \cref{eq:standard-tpbvp}, which assumes fixed final time.
    To cast the problem into standard form, one can apply the \emph{time-scaling} strategy.
    First, the time horizon is rescaled to the fixed interval $[0,1]$ by using the scaled time variable $\tau = t/t_f$.
    Next, the derivatives must be adjusted according to the new time variable.
    By the chain rule, differentiation with respect to $\tau$ introduces a scaling factor, that is, $\frac{\partial}{\partial \tau} \coloneqq \frac{\partial}{\partial t} \frac{\partial (\tau t_f)}{\partial \tau} = \frac{\partial}{\partial t} t_f$.
    Finally, the final time $t_f$ is replaced by an auxiliary state variable $r$ with trivial dynamics $\dot{r} = 0$.
    This results in a TPBVP with fixed final time, namely equal to $1$, and an additional state variable $r$ that encodes the original final time $t_f$.
    
    For this example, the time-scaled cost becomes:
    \begin{equation*}  
        J(x, u, r) = \frac{1}{2}\alpha r^2 + \int_{0}^{1} \frac{1}{2}\beta r u^2(\tau) \d \tau,
    \end{equation*}
    with corresponding Hamiltonian:
    \begin{equation*}  
        \hamiltonian = \frac{1}{2}\beta r u^2 + p_1 r x_2 + p_2 r u.
    \end{equation*}
    As a result, the necessary conditions for optimality become:
    \begin{equation*}
    \begin{split}
        \dot x_1^* &= r^* x_2^*, \\
        \dot x_2^* &= r^* u^*, \\
        \dot p_1^* &= 0, \\
        \dot p_2^* &=-r^* p_1^*, \\
        \dot r^* &= 0, \\
        0 &= \beta r^* u^* + r^* p_2^*,
    \end{split}
    \end{equation*}
    with boundary conditions:
    \begin{equation*}
    \begin{split}
    &x_1^*(0) = 10, \\
    &x_2^*(0) = 0, \\
    &x_1^*(1) = 0, \\
    &x_2^*(1) = 0, \\
    &\alpha r^* + \frac{1}{2}\beta r^* u^*(1)^2 + r^* p_1^*(1)x_2^*(1) + r^* p_2^*(1)u^*(1) = 0.
    \end{split}
    \end{equation*}

    After this reformulation, the problem adheres to the standard form and can be solved numerically.
\end{example}

For a systematic treatment of how nonstandard boundary value problems can be reformulated into standard form suitable for general-purpose solvers, see \citet{AscherRussell1981}.
For a practical implementation of TPBVP solvers for free-final time optimal control problems, we refer the reader to the notebook \colorcode{ch02/free\_final\_time\_optimal\_control.ipynb} in the repository
\newline 
\noindent\colorcode{github.com/StanfordASL/pora-exercises}.

\subsection{Direct Methods}
\label{sec:direct_methods}
So far, we introduced indirect methods, which involve deriving the necessary optimality conditions of the continuous-time optimal control problem and then discretizing them to numerically solve the resulting two-point boundary value problem.
While indirect methods provide deep theoretical insights, they can be challenging to apply in practice due to the difficulty of solving boundary value problems, particularly for complex nonlinear dynamics or large-scale systems.

Direct methods take the opposite approach. 
Rather than deriving the optimality conditions analytically, the problem is discretized first.
This reduces the continuous-time optimal control problem to a finite-dimensional nonlinear optimization problem, which can then be solved using general-purpose numerical optimization algorithms.

The process of converting the continuous-time optimal control problem into a discretized form amenable to numerical optimization is known as \emph{transcription}.
While there exist many different transcription methods, a simple and widely used approach is the \emph{forward Euler discretization}.
This method selects a discretization $0 = t_0 < t_1 < \ldots < t_N = t_f$ of the time interval $[0, t_f]$ and approximates the state and control sequences assuming a zero-order hold on both the states and control inputs, meaning that both the state and the control input are constant over each time interval $[t_i, t_{i+1})$.
The system dynamics are then integrated forward using Euler integration:
\begin{equation}
\x_{i+1} \approx \x_i + h_i \f(\x_i, \u_i), \quad h_i = t_{i+1}-t_i.
\end{equation}

Direct methods are typically grouped into two main families:
\begin{itemize}
    \item \emph{State and control parameterization methods (also known as direct collocation methods)}: here, both the control inputs and the state trajectories are discretized, and the dynamics are introduced explicitly as algebraic constraints linking the state and control variables at each discretization point (e.g., trapezoidal and Hermite-Simpson collocation, Gauss-Lobatto methods, etc.).
    \item \emph{Control parameterization methods (also known as direct shooting methods)}: in this approach, only the control inputs are discretized, and the state trajectories are obtained by numerically integrating the system dynamics forward in time.
    As a result, the optimization variables are solely the discretized controls, while the states are implicitly defined by the integration of the dynamics (e.g., using single or multiple shooting techniques).
\end{itemize}

In what follows, we illustrate the fundamental concepts of both families of methods through a concrete example.
\begin{example}[Zermelo's Problem - Continuous-time problem]
\label{ex:zermelos_problem}
Consider the problem of steering a boat from a point $(0,0)$ to a point $(M, \ell)$ in a river with a current. The boat can be controlled by adjusting its direction and its speed is constant. The dynamics of the boat are described by the following differential equations:
    \begin{equation*}
        \begin{aligned}
            \dot{x}(t) &= v \cos(u(t)) + \text{flow}(y(t)), \quad t \in [0, t_f], \\
            \dot{y}(t) &= v \sin(u(t)), \quad t \in [0, t_f], 
        \end{aligned}
    \end{equation*}
where $(x(t), y(t))$ is the position of the boat with $x$ defining the coordinate along the river and $y$ the coordinate across the river, $u(t)$ is the control input (the direction of the boat), and $v$ is the constant speed of the boat. 
    For simplicity, assume that the river flow is described by an arbitrary function acting in the $x$-direction, with its intensity depending on the position $y(t)$, namely $\text{flow}(y(t))$.

    The objective is to minimize the control effort over time, which can be formulated as an optimal control problem:
    \begin{align*}
        &\underset{u(t)}{\text{minimize}} && \int_0^{t_f} u(t)^2 \, \d t, \\
        &\text{subject to} && \dot{x}(t) = v \cos(u(t)) + \text{flow}(y(t)), \quad t \in [0, t_f], \\
        &&& \dot{y}(t) = v \sin(u(t)), \quad t \in [0, t_f], \\
        &&& \left(x(0), y(0)\right) = (0, 0), \\
        &&& \left(x(t_f), y(t_f)\right) = (M, \ell), \\
        &&& |u(t)| \leq u_{\text{max}}, \quad t \in [0, t_f].
    \end{align*}
\end{example}

\subsubsection{Direct Collocation Methods}
\label{sec:transcription_methods}
Consider the Problem introduced in~\cref{ex:zermelos_problem}.
Applying a state and control parametrization, that is a collocation method, leads to the following finite-dimensional nonlinear program, equivalently described in Algorithm~\ref{alg:zermelo_direct_math_comments}:
\begin{align*}
    &\underset{(x, y, u)}{\text{minimize}} && \sum_{i=0}^{N-1} h_i u_i^2, \\
    &\text{subject to} && x_{i+1} = x_i + h_i \left(v \cos(u_i) + \text{flow}(y_i)\right), \quad i = 0, \ldots, N-1, \\
    &&& y_{i+1} = y_i + h_i v \sin(u_i), \quad i = 0, \ldots, N-1, \\
    &&& (x_0, y_0) = (0, 0), \\
    &&& (x_N, y_N) = (M, \ell), \\
    &&& |u_i| \leq u_{\text{max}}, \quad i = 0, \ldots, N-1.
\end{align*}
In this formulation, both the state and control trajectories are discretized and treated as decision variables.
The system dynamics are not enforced through numerical integration, but rather as algebraic equality constraints linking consecutive discretization points.

\begin{listing}[p]
\begin{tcolorbox}[colback=gray!10, colframe=gray!50,
    title=Collocation Formulation of Zermelo's Problem, boxrule=0.5mm, arc=0mm]
\begin{minted}[escapeinside=||]{python}
# Decision variables: |$(x_i, y_i, u_i), \; i=0,\dots,N$|
get_x = lambda z: z[:N + 1]
get_y = lambda z: z[N + 1:-N]
get_u = lambda z: z[-N:]
get_z = lambda x, y, u: np.concatenate([x, y, u])

# Cost: |$\sum_{i=0}^{N-1} h_i u_i^2$|
cost = lambda z: np.sum(h * np.square(get_u(z)))

def constraints(z):
    x, y, u = get_x(z), get_y(z), get_u(z)
    constraints = []
    for i in range(N):
        # |$x_{i+1} = x_i + h_i \left(v \cos(u_i) + \text{flow}(y_i)\right)$|
        constraints.append(x[i+1] - x[i] - h*(v*np.cos(u[i]) 
                           + flow(y[i])))
        # |$y_{i+1} = y_i + h_i v \sin(u_i)$|
        constraints.append(y[i+1] - y[i] - h*v*np.sin(u[i]))
    # Boundary conditions: |$(x_0, y_0) = (0,0), \; (x_N, y_N) = (M, \ell)$|
    constraints.extend([x[0], y[0], x[N] - M, y[N] - l])
    return np.array(constraints)

# State bounds
x_lower = np.zeros(N + 1)
x_upper = M * np.ones(N + 1)
y_lower = np.zeros(N + 1)
y_upper = l * np.ones(N + 1)

# Control bounds: |$|u_i| \leq u_{\text{max}}$|;
u_lower = -u_max * np.ones(N) # control constraint
u_upper = u_max * np.ones(N)  # control constraint

bounds = Bounds(
	get_z(x_lower, y_lower, u_lower), 
	get_z(x_upper, y_upper, u_upper))

# Solve the NLP
result = minimize(cost, z0, bounds=bounds,
                    constraints={'type': 'eq', 'fun': constraints})
\end{minted}
\end{tcolorbox}
\caption{Direct collocation approach to Zermelo's problem using forward Euler discretization.
The code for this example is available in the repository \colorcode{github.com/StanfordASL/pora-exercises} in the notebook \colorcode{ch02/zermelos\_problem.ipynb}.} 
\label{alg:zermelo_direct_math_comments}
\end{listing}

\subsubsection{Direct Shooting Methods}
Many of the concepts introduced for state and control parametrization carry over to control parametrization methods, with a key distinction in how the dynamics are handled.
In control parametrization (shooting) methods, the optimization variables consist only of the control inputs at each discretization point.
The state trajectory is not explicitly optimized but is instead computed recursively by forward simulation of the system dynamics.
In other words, a candidate sequence of controls uniquely determines the corresponding states, which are then used to evaluate the cost and any state constraints.

Concretely, let us revisit Zermelo’s problem from Example~\ref{ex:zermelos_problem}, this time using a shooting method transcription.
In this formulation, the control inputs $\{u_i\}_{i=0}^{N-1}$ are treated as the optimization variables, while the states $\{(x_i, y_i)\}_{i=0}^{N-1}$ are computed recursively from the dynamics.

The resulting finite-dimensional optimization problem is:
\begin{align*}
    &\underset{u}{\text{minimize}} && \sum_{i=0}^{N-1} h_i u_i^2, \\
    &\text{subject to} && (x_N, y_N) = (M, \ell), \\
    &&& |u_i| \leq u_{\text{max}}, \quad i = 0, \ldots, N-1, \\
    &&& \text{where, recursively,} \\
    &&& x_{i+1} = x_{i} + h_i \left(v \cos(u_{i}) + \text{flow}(y_{i})\right), \quad i = 0, \ldots, N-1, \\
    &&& y_{i+1} = y_{i} + h_i v \sin(u_{i}), \quad i = 0, \ldots, N-1.
\end{align*}
Here, the dynamics are no longer constraints in the optimization problem, but rather equations that implicitly determine the state evolution given a candidate control sequence.
Algorithm~\ref{alg:zermelo_shooting} provides a Python implementation of this shooting method approach to Zermelo's problem.

\begin{listing}[ht!]
\begin{tcolorbox}[colback=gray!10, colframe=gray!50,
    title=Shooting Formulation for Zermelo's Problem, boxrule=0.5mm, arc=0mm]
\begin{minted}[escapeinside=||]{python}
# Decision variables: |$(u_i), \; i=0,\dots,N-1$|; Cost: |$\sum_{i=0}^{N-1} h_i u_i^2$|
cost = lambda u: np.sum(h * np.square(u))

# States computed recursively from |$x_0=0, y_0=0$|
dynamics = lambda x, y, u: (
        x + h * (v * np.cos(u) + flow(y)),
    y + h * v * np.sin(u)
)

def inequality_constraints(u):
    x, y = 0, 0 # initial condition (x(0), y(0)) = (0, 0)
    constraints = []
    for ui in u:
        x, y = dynamics(x, y, ui)
        # |$(x_i, y_i) >= (0,0)$| (box constraint with below)
        constraints.extend([x, y])
        # |$(x_i, y_i) <= [M, \ell]$|
        constraints.extend([M - x, l - y])
    # |$(x_N, y_N) >= [M, \ell]$| (enforcing equality with the above)
    constraints.extend([x - M, y - l])
    return constraints

bounds = Bounds(-u_max * np.ones(N), 
                u_max * np.ones(N)) # |$\lvert u_i\rvert \le u_{\text{max}}$|

# Solve NLP
result = minimize(cost, u0, bounds=bounds, 
                  constraints={'type': 'ineq',
                  	     'fun': inequality_constraints})
\end{minted}
\end{tcolorbox}
\caption{Direct shooting approach to Zermelo’s problem using forward Euler discretization.
The code for this example is available in the repository \colorcode{github.com/StanfordASL/pora-exercises} in the notebook \colorcode{ch02/zermelos\_problem.ipynb}.} 
\label{alg:zermelo_shooting}
\end{listing}

\medskip
Both approaches come with their own advantages and limitations.
Control parameterization methods generally result in smaller optimization problems, making the method computationally attractive. 
The dynamics are enforced exactly through integration (up to the accuracy of the chosen numerical integrator), which is especially useful when the dynamics are complex or only accessible via a black-box simulator.
However, state constraints may be difficult to enforce, as the states are not explicit optimization variables but rather implicitly defined through the integration of the dynamics. 
This can lead to numerical instability or infeasibility when state constraints are critical. 
Moreover, errors from numerical integration may accumulate, potentially reducing the accuracy of the solution.

On the other hand, state and control parametrization methods treat both states and controls as optimization variables.
This allows state constraints to be imposed directly, improving numerical stability and robustness, often leading to better-conditioned optimization problems when constraints play a central role.
However, the resulting optimization problem generally grows significantly in size, since all states and controls at every discretization point are treated as decision variables.
This higher dimensionality increases computational cost and can make the solver more sensitive to initial guesses.

In practice, both methods are widely used, and the choice between them often depends on the problem structure, the availability of simulators or system models, and the importance of accurately handling state constraints.

\subsection{Differentially Flat Systems}
\label{sec:differentially_flat_systems}
Computing open-loop control sequences by directly solving optimal control problems can often be computationally intensive.
In many applications, it is useful to trade off strict optimality\sidenote{That is, the theoretical best performance according to a given cost functional.} for computational tractability by seeking ``good’’ trajectories that are simpler to compute, even if slightly sub-optimal.

For a special class of systems known as \emph{differentially flat systems}, generating such feasible trajectories is considerably simpler.
A system is differentially flat if there exists a set of outputs, called \emph{flat outputs}, such that all system states and inputs can be expressed as algebraic functions of these outputs and a finite number of their derivatives.
This property allows trajectory generation to be performed in the space of the flat outputs, eliminating the need to solve differential equations as part of the optimization process and greatly reducing computational complexity.

Differentially flat models arise in several common robotics applications, including simple car models, quadrotors, and many other wheeled or aerial vehicles.
Their relative simplicity and expressiveness make them particularly attractive for trajectory planning and open-loop control synthesis.

\begin{example}[Differentially Flat Autonomous Vehicle Control] 
\label{ex:carflatness}
Recall the motion planning task from \cref{ex:auto_racing_ocp} where the objective was to compute an open-loop control sequence to drive a vehicle through a course to a goal position in minimum time.
If we relax the requirement of optimality and instead aim simply to find a feasible trajectory that follows the course, we can exploit the differential flatness of the kinematic car model.
Specifically, consider the kinematic car model from \cref{eq:car-dynamics}:
\begin{equation*}
\begin{split}
    \dot{x} &= v\cos\theta,\\
    \dot{y} &= v\sin\theta,\\
    \dot{\theta} &= \frac{v}{L}\tan\phi,
\end{split}
\end{equation*}
where $(x,y)$ is the vehicle position, $\theta$ is the heading, $v$ is the speed, $\phi$ is the steering angle, and $L$ is the wheelbase.

This system is differentially flat with flat outputs $(x(t), y(t))$.
Therefore, it is sufficient to specify any differentiable trajectory for $x(t)$ and $y(t)$ that respects the course constraints.
From these trajectories, the remaining state and control variables---which are the quantities needed for practical implementation---can be computed analytically.
The heading is obtained from the velocity direction as:
\begin{equation*}
\theta = \tan^{-1}\left(\frac{\dot{y}}{\dot{x}}\right).
\end{equation*}
and the speed along the trajectory can be computed using either component of the velocity:
\begin{equation*}
v = \frac{\dot{x}}{\cos\theta}, \quad \text{or}  \quad v = \frac{\dot{y}}{\sin\theta}.
\end{equation*}
Finally, the steering angle is determined from the heading dynamics:
\begin{equation*}
\phi = \tan ^{-1}\left(\frac{L\dot{\theta}}{v}\right).
\end{equation*}
In this way, a feasible trajectory for the vehicle can be generated entirely by specifying smooth flat output trajectories, from which all states and, importantly, the control inputs can be derived directly.
\end{example}

We formalize this concept through the notion of differential flatness\sidenote{\citet{Murray2009} is a good resource for a comprehensive treatment on differential flatness.}.

\begin{definition}[Differential Flatness]
A nonlinear system with state $\x \in \R^\statedim$ and control $\u \in \R^\controldim$:
\begin{equation} 
\label{eq:diffflatsys}
\dot{\x}(t) = \dynmodel(\x(t),\u(t)),
\end{equation}
is \emph{differentially flat} if there exists a function $\alpha$ such that:
\begin{equation}
\z = \alpha (\x,\u,\dot{\u},\dots,\u^{(a)}),
\end{equation}
where $\u^{(i)}$ denotes the $i$-th time derivative of $\u$, and such that the system trajectories can be expressed as functions of the flat output $\z \in \R^m$ and a finite number of its derivatives:
\begin{equation} 
\label{eq:ztoxu}
\begin{split}
\x &= \beta (\z,\dot{\z},\dots,\z^{(b)}) \\
\u &= \gamma (\z,\dot{\z},\dots,\z^{(c)}).
\end{split}
\end{equation}
\end{definition}

In other words, a system is said to be differentially flat if there exists a set of outputs $\z$ (with the same dimension as the input vector $\u$) that completely determine both the states and the inputs, without requiring integration of the system dynamics.
For trajectory optimization, this property is particularly advantageous: since the evolution of a flat system is fully characterized by its flat outputs, trajectories can be computed directly in the output space and then mapped to the corresponding inputs, thereby avoiding expensive integration of the dynamics.

In the following sections, we explore different techniques to exploit differential flatness for open-loop trajectory design, including how to parameterize trajectories in the flat output space, handle initial and terminal state constraints, and enforce control constraints.

\subsubsection{Trajectory Parameterization}
Our primary limitation when planning a trajectory in the flat output space is that it must be \emph{differentiable}.
A common approach is to parameterize each component of the flat output $\z$ using $N$ smooth basis functions:
\begin{equation} 
\label{eq:flat}
z_j(t) = \sum_{i=1}^{N} \alpha_i^{[j]} \psi_i(t),
\end{equation}
where $z_j$ is the $j$-th element of $\z$, $\alpha_i^{[j]} \in \R$ are parameters that define the trajectory, and $\psi_i(t)$ are smooth basis functions.

Polynomial basis functions are a natural choice, e.g., $\psi_1(t) = 1$, $\psi_2(t) = t$, $\psi_3(t) = t^2$, etc.
A key advantage of this parameterization is that $z_j(t)$ is linear in the variables $\alpha_i^{[j]}$, which facilitates translating constraints on $\z$ and its derivatives directly into constraints on the coefficients $\alpha_i^{[j]}$.

\subsubsection{Equality Constraints}
\label{subsubsec:flatequalityconst}
A key component of any open-loop motion planning problem is the enforcement of boundary conditions.
Typically, this means ensuring that the system begins at a prescribed initial state $\x(0) = \x_0$ and often that it reaches a desired terminal state, $\x(t_f) = \x_f$, at some final time $t_f$.
When planning in the flat output space, these state conditions must be expressed as constraints on the flat output $\z(t)$ and its derivatives.
Recalling the mapping in \cref{eq:ztoxu}, this leads to:
\begin{equation} 
\label{eq:flatbc}
\begin{split}
\x_0 &= \beta (\z(0),\dot{\z}(0),\dots,\z^{(q)}(0)), \\
\x_f &= \beta (\z(t_f),\dot{\z}(t_f),\dots,\z^{(q)}(t_f)). \\
\end{split}
\end{equation}

In practice, this means that boundary conditions on $z_j(0), \dot{z}_j(0), \dots , z_j^{(q)}(0)$ and $z_j(t_f), \dot{z}_j(t_f), \dots , z_j^{(q)}(t_f)$ must be enforced.
When using a smooth basis function parameterization of the form in \cref{eq:flat}, these conditions translate directly into algebraic constraints on the coefficients $\alpha_i^{[j]}$.
By differentiating \cref{eq:flat} $q$ times, we obtain:
\begin{equation}
\begin{split}
\dot{z}_j(t) &= \sum_{i=1}^{N} \alpha_i^{[j]} \dot{\psi_i}(t), \\
&\vdots \\
z_j^{(q)}(t) &= \sum_{i=1}^{N} \alpha_i^{[j]} \psi_i^{(q)}(t). \\
\end{split}
\end{equation}
which allows us to express the boundary conditions as a system of linear equations:
\begin{equation} 
\label{eq:diffflatlinear}
\begin{bmatrix}
    \psi_1(0) & \psi_2(0) & \dots & \psi_N(0) \\
    \dot{\psi_1}(0) & \dot{\psi_2}(0) & \dots & \dot{\psi_N}(0) \\
    \vdots & \vdots & & \vdots \\
    \psi_1^{(q)}(0) & \psi_2^{(q)}(0) & \dots & \psi_N^{(q)}(0) \\
    \psi_1(t_f) & \psi_2(t_f) & \dots & \psi_N(t_f) \\
    \dot{\psi_1}(t_f) & \dot{\psi_2}(t_f) & \dots & \dot{\psi_N}(t_f) \\
    \vdots & \vdots & & \vdots \\
    \psi_1^{(q)}(t_f) & \psi_2^{(q)}(t_f) & \dots & \psi_N^{(q)}(t_f) \\
\end{bmatrix}
\begin{bmatrix}
    \alpha_1^{[j]} \\
    \alpha_2^{[j]} \\
    \vdots \\
    \alpha_N^{[j]} \\
\end{bmatrix} =
\begin{bmatrix}
    z_j(0) \\
    \dot{z}_j(0) \\
    \vdots \\
    z^{(q)}_j(0) \\
    z_j(t_f) \\
    \dot{z}_j(t_f) \\
    \vdots \\
    z_j^{(q)}(t_f)
\end{bmatrix}.
\end{equation}
Assuming the matrix formed by the basis functions has a sufficient number of columns and that it is full column rank, we can solve for (possibly non-unique) $\alpha_i^{[j]}$ that solve the trajectory generation problem.

More generally, any equality constraint on the flat outputs or their derivatives—beyond just initial and terminal states—can be written in this linear form. 
For instance, waypoints can be added as additional equality constraints. 
However, if too many constraints are imposed, the system may become overdetermined, leaving no feasible solution. 
In such cases, one must increase the richness of the basis functions, for example by using higher-order polynomials or additional functions, which improves flexibility but also increases computational complexity.

\subsubsection{Inequality Constraints via Time Scaling}
Having addressed equality constraints in \cref{subsubsec:flatequalityconst}, we now turn to inequality constraints.
These commonly arise in motion planning and control to enforce actuator limits or safety bounds on the state.
For example, the simple car model from \cref{ex:carflatness} may have a speed constraint of the form:
\begin{equation*}
\vert v(t)\vert \leq v_\text{max}.
\end{equation*}

A useful technique for handling such constraints in the flat-output space is \emph{time scaling}.
The idea is to first plan a trajectory that satisfies the equality constraints (e.g., by solving \cref{eq:diffflatlinear}), and then adjust its temporal evolution—speeding up or slowing down along the path—to enforce the inequality constraints.

Formally, let $\x(t)$ be a trajectory satisfying the equality constraints. We can separate the \emph{geometric path}\sidenote{The geometric path of a trajectory is the sequence of states $\x$ of the trajectory, but not associated with a particular time} from its timing by introducing a path parameter $s(t)$:
\begin{equation*}
\x(t) = \x(s(t)),
\end{equation*}
with $s(0) = s_0$, $s(t_f) = s_f$, and $\dot{s}(t) > 0$\sidenote{The condition $\dot{s}(t)>0$ ensures invertibility, so each $t$ corresponds to a unique $s$.}. 
The geometric path $\x(s)$ captures the sequence of states, while the choice of $s(t)$ determines how quickly the system traverses that path. 
Varying $s(t)$ is referred to as \emph{time scaling}.

\begin{example}[Time Scaling for a Simple System]
\label{ex:simple_time_scale}
Consider a scalar system with state $x \in \R$ and a straight-line path connecting an initial and terminal state, $x_0$ and $x_f$:
\begin{equation*}
    x(s) = x_0 + s(x_f - x_0), \quad s \in [0,1].
\end{equation*}
Choosing a cubic polynomial for $s(t)$,
\begin{equation*}
    s(t) = \frac{3}{T^2}t^2 - \frac{2}{T^3}t^3, \quad t \in [0,T],
\end{equation*}
which satisfies $s(0) = 0$, $s(T) = 1$, and $\dot s(t) > 0$, yields the temporal trajectory:
\begin{equation*}
    x(t) = x_0 + \left(\frac{3}{T^2}t^2 - \frac{2}{T^3}t^3 \right)(x_f - x_0).
\end{equation*}
Here, $T$ controls the duration of the trajectory. 
If we impose a velocity bound:
\begin{equation*}
   \lvert \dot{x}\rvert  \leq \dot{x}_{\text{max}},
\end{equation*}
then:
\begin{equation*}
\begin{split}
    \dot{x} &= 6\left(\frac{t}{T^2} - \frac{t^2}{T^3} \right)(x_f - x_0), \\
   \ddot{x} &= 6\left(\frac{1}{T^2} - \frac{2t}{T^3} \right)(x_f - x_0), \\
\end{split}
\end{equation*}
with the maximum velocity attained at $t = \tfrac{T}{2}$.
We can then convert this into a constraint on $T$ to ensure the inequality constraint is satisfied:
\begin{equation*}
T \geq \frac{3(x_f - x_0)}{2\dot{x}_{\text{max}}}.
\end{equation*}
\end{example}
\cref{ex:simple_time_scale} illustrates the key idea: inequality constraints can often be transformed into conditions on the timing law $s(t)$, without altering the geometric path itself.

For general state-space systems, time scaling is often more complex.
Given a feasible trajectory $(\x(t), \u(t))$ of the system dynamics in \cref{eq:diffflatsys}, we can rewrite it as a geometric path $(\x(s), \u(s))$ using a path parameter $s(t)$:
\begin{equation}
\frac{\d\x(s)}{\d s} \frac{\d s(t)}{\d t} = \dynmodel(\x(s), \u(s)).
\end{equation}
For time scaling, we replace $s(t)$ with a new path parameter $\tilde{s}(t)$ over a possibly different interval $t \in [0,\tilde{t}_f]$, with $\tilde{s}(0)=s_0$ and $\tilde{s}(\tilde{t}_f)=s_f$\sidenote{The geometric path is still defined on the interval $[s_0, s_f]$, which must remain the same for any new time scaling law.}.
The new scaling must still satisfy the dynamics:
\begin{equation}
\frac{\d\x(\tilde{s})}{\d\tilde{s}} \frac{\d\tilde{s}(t)}{\d t} = \dynmodel(\x(\tilde{s}), \u(\tilde{s})).
\end{equation}
Since the geometric path is fixed---as it was previously defined---the terms $\frac{\d\x(\tilde{s})}{\d\tilde{s}}$ and $\x(\tilde{s})$ are also fixed.
Therefore, time scaling with a new path parameter $\tilde{s}(t)$, is only admissible if an appropriate $\tilde{\u}(\tilde{s})$ can be found.
Fortunately, for many systems---including those commonly studied in motion planning---this is possible with the right choice of $\tilde{s}(t)$.

\begin{example}[Time Scaling for the Simple Car Model]
\label{ex:time_scale_simple_car}
\theoremstyle{definition} 
Consider again the simple car model from \cref{eq:car-dynamics}:
\begin{equation*}
\begin{split}
	\dot{x} &= v\cos\theta,\\
    \dot{y} &= v\sin\theta,\\
    \dot{\theta} &= \frac{v}{L}\tan\phi,
\end{split}
\end{equation*} 
and suppose we have identified a candidate trajectory $\x_c(t)$ with control $\u_c(t)$ by leveraging the differential flatness of the model through \cref{eq:diffflatlinear} and mapping the flat outputs $\z_c(t)$ into the state and control space.

For this model, a natural choice for the path parameter $s$ is the arc-length, defined as:
\begin{equation*}
    s(t) = \int_0^t  v(\tau) \d \tau.
\end{equation*}
such that $\dot{s}(t) = v(t) > 0$.
With this choice, the geometric path $\x_c(s)$ is defined over $s \in [0, L_{\text{path}}]$, where $L_{\text{path}}$ is the total length of the path.
Rewriting the dynamics in terms of an arbitrary time scaling $\tilde{s}(t)$ gives:
\begin{equation*}
\begin{split}
\frac{\d x_c(\tilde{s})}{\d\tilde{s}}\dot{\tilde{s}} &= v(\tilde{s})\cos\theta_c(\tilde{s}),\\
\frac{\d y_c(\tilde{s})}{\d\tilde{s}}\dot{\tilde{s}}&= v(\tilde{s})\sin\theta_c(\tilde{s}),\\
\frac{\d\theta_c(\tilde{s})}{\d\tilde{s}}\dot{\tilde{s}} &= \frac{v(\tilde{s})}{L}\tan\phi(\tilde{s}),
\end{split}
\end{equation*}
which must hold for any admissible time scaling $\tilde{s}(t)$\sidenote{The trivial choice $\tilde{s}(t) = s(t)$ reproduces the original candidate trajectory with control inputs $\u_c(t)$.}.

By adopting the arc-length parameterization, we have $\dot{\tilde{s}} = v(\tilde{s})$, so these equations reduce to:
\begin{equation*}
\begin{split}
\frac{\d x_c(\tilde{s})}{\d\tilde{s}} &= \cos\theta_c(\tilde{s}),\\
\frac{\d y_c(\tilde{s})}{\d\tilde{s}}&= \sin\theta_c(\tilde{s}),\\
\frac{\d\theta_c(\tilde{s})}{\d\tilde{s}} &= \frac{1}{L}\tan\phi(\tilde{s}).
\end{split}
\end{equation*}
The first two equations are automatically satisfied for any choice of $\tilde{s} \in [s_0, s_f]$, since the original candidate trajectory satisfies the dynamics. 
On the other hand, the third equation is satisfied provided we reuse the same steering input, $\phi(\tilde{s}) = \phi_c(\tilde{s})$.
Therefore, the dynamics remain consistent for any choice of time scaling $\tilde{s}(t)$: the geometric path is preserved, while the temporal evolution along that path is left free.
This observation is powerful as we may freely adjust the speed input $v(t)$, subject only to $\dot{\tilde{s}}(t) > 0$, without altering the geometry of the trajectory.
In practice, this allows us to easily enforce inequality constraints on the speed $\lvert v(t)\rvert \leq v_\text{max}$.
\end{example}

\cref{ex:time_scale_simple_car} shows a relatively straightforward application of time scaling to a model derived from kinematic constraints.
This idea extends naturally to a wide class of kinematic models of the form:
\begin{equation} 
\label{eq:kinmodel}
    \dot{\x}(t) = G(\x(t))  \u(t).
\end{equation}
Applying the chain rule, we obtain:
\begin{equation*}
    \frac{\d\x(s)}{\d s} \dot{s} = G(\x(s(t)))  \u(t),
\end{equation*}
which can be rewritten as:
\begin{equation} 
\label{eq:kinematicgeometricmodel}
    \frac{\d \x(s)}{\d s} = G(\x(s)) \u_g(s),
\end{equation}
where $\u_g(s) = \tfrac{\u(t)}{\dot{s}(t)}$ is the \emph{geometric control}\sidenote{Since $s(t)$ must be strictly increasing, we require $\dot{s}(t) > 0$.}.
Equation \eqref{eq:kinematicgeometricmodel} shows that the geometric path $\x(s)$ is fully determined by the geometric control $\u_g(s)$, independent of the time parametrization.
Therefore, once the geometric control and geometric path are defined, we can temporally scale the trajectory $\x(t)$ using the path parameter $s(t)$ without changing the geometric path.
The corresponding control inputs are recovered via $\u(t) = \dot{s}(t)\u_g(s)$.

In summary, for models of the form \eqref{eq:kinmodel}, we can perform time scaling by:
\begin{enumerate}
\item Selecting a path parameter $s$ (e.g., arc-length), computing $s(t)$ for the original trajectory $\x(t)$, and determining the interval $[s_0, s_f]$.
\item Re-parameterizing the control $\u(t)$ in terms of $s$. 
\item Computing the geometric control $\u_g(s) = \u(s(t))/\dot{s}(t)$ for $s \in [s_0, s_f]$.
\item Defining a new path parameter function $\tilde{s}(t)$ over the interval $[0, \tilde{t}_f]$ with $\dot{\tilde{s}}(t) > 0$, $\tilde{s}(0) = s_0$, and $\tilde{s}(\tilde{t}_f) = s_f$.
\item Computing the new control inputs as $\tilde{\u}(t) = \u_g(\tilde{s}(t)) \dot{\tilde{s}}(t)$ for all $t \in [0,\tilde{t}_f]$.
\end{enumerate}

\begin{example}[Time Scaling for the Unicycle Model] \label{ex:timescaleuni}
\theoremstyle{definition}
Consider the kinematic unicycle model:
\begin{equation}
\begin{split}
    \dot{x} &= v\cos\theta,\\
    \dot{y} &= v\sin\theta,\\
    \dot{\theta} &= \omega, 
\end{split}
\end{equation}
where $\tup{x, y}$ denote the position, $\theta$ the heading, $v$ the forward velocity, and $\omega$ the rotation rate. 
We define the state as $\x = \vCol{x, \: y, \: \theta}$ and the control as $\u = \vCol{v, \:\omega}$.

A natural path parameter for this system is again the arc-length:
\begin{equation*}
    s(t) = \int_0^t  v(\tau) \d\tau,
\end{equation*}
such that $\dot{s}(t) = v(t) > 0$.
If the trajectory is defined over $t \in [0,T]$ with total length $L_{\text{path}}$, then $s(0) = 0$ and $s(T) = L_{\text{path}}$.
The corresponding geometric controls are:
\begin{equation*}
\begin{split}
v_g(s) &= \frac{v(s)}{\dot{s}(t)} = 1, \\
\omega_g(s) &= \frac{\omega(s)}{\dot{s}(t)} = \frac{\omega(s)}{v(s)},
\end{split}
\end{equation*}
where the fact that $v_g(s)=1$ follows directly from $\dot{s}(t) = v(s(t))$.
Introducing a new timing law $\tilde{s}(t)$ generates a new velocity profile $\tilde{v}(\tilde{s}) = \dot{\tilde{s}}(t)$ along the path, which can use to solve for the new $\tilde{\omega}$ inputs by:
\begin{equation*}
\begin{split}
\tilde{\omega}(\tilde{s}) &= \omega_g(\tilde{s}) \dot{\tilde{s}}(t) = \frac{\omega(\tilde{s})}{v(\tilde{s})} \tilde{v}(\tilde{s}).
\end{split}
\end{equation*}
In practice, it is often simpler to directly prescribe a velocity profile $\tilde{v}(\tilde{s})$ along the path and compute the corresponding angular velocity $\tilde{\omega}(\tilde{s}) = \frac{\omega(\tilde{s})}{v(\tilde{s})} \tilde{v}(\tilde{s})$. 
Finally, to determine the new controls as functions of time, we note that:
\begin{equation*}
    \tau(s) = \int_0^s \frac{1}{\tilde{v}(s')}\d s',
\end{equation*}
defines a function $\tau(s)$ that maps each point $s \in [0, L_{\text{path}}]$ to a new time.
\end{example}

\begin{example}[Planar Quadrotor Control]
In this example we consider the control of a planar quadrotor system. 
The quadrotor is modeled with six state variables: horizontal position $x$, vertical position $y$, orientation angle $\phi$, and their respective velocities. 
The control inputs are the thrusts $T_1$ and $T_2$ from the two rotors.
The objective is to minimize the energy consumption, represented by the integral of the squared thrusts over time:
\begin{equation*}
\minimize[] \int_0^{t_f} T_1(t)^2 + T_2(t)^2 \d t.
\end{equation*}
The system dynamics are given by the following differential equations:
\begin{equation*}
    \begin{bmatrix}
    \dot{x} \\
    \dot{v}_x \\
    \dot{y} \\
    \dot{v}_y \\
    \dot{\phi} \\
    \dot{\omega}
    \end{bmatrix} = 
    \begin{bmatrix}
    v_x \\
    -\frac{(T_1+T_2)}{m}\sin\phi \\
    v_y \\
    \frac{(T_1+T_2)}{m}\cos\phi - g \\
    \omega \\
    \frac{(T_2-T_1)\ell}{I_{zz}}
    \end{bmatrix},
\end{equation*}
where $m$ is the mass, $g$ is the gravitational acceleration, $\ell$ is the distance from the center of mass to each rotor, and $I_{zz}$ is the moment of inertia about the $z$-axis.

This system is differentially flat, with flat outputs $(x, y)$. 
For a practical implementation of differential flatness for trajectory generation applied to this system, refer to the notebook \colorcode{ch02/differentially\_flat\_planar\_quadrotor.ipynb} in the repository \colorcode{github.com/StanfordASL/pora-exercises}.
\end{example}

\section{Summary}
\label{sec:ol_ctrl_summary}
In this chapter, we explored how trajectory optimization provides a fundamental framework for computing open-loop motions and establishing a foundation for autonomous decision-making in robotic systems.

We began by formalizing the optimal control problem---a mathematical formalization for the task of driving a system’s state evolution through admissible control inputs while optimizing a performance criterion. 
This formulation involves three key components: the system’s mathematical model, the physical constraints, and the performance criteria.

We then introduced two major families of methods for solving optimal control problems and compute optimal open-loop control sequences: direct methods and indirect methods.
Indirect methods adopt an “optimize-then-discretize” approach, deriving analytical necessary conditions for optimality and solving the resulting two-point boundary value problem numerically. 
In contrast, direct methods follow a “discretize-then-optimize” strategy, transcribing the continuous problem into a finite-dimensional nonlinear program that can be solved with standard optimization solvers.

Lastly, we discussed differentially flat systems, a special class of systems for which trajectory generation is significantly simplified. 
For these systems, planning can be performed in a lower-dimensional ``flat output" space using techniques like polynomial parameterization. 
We showed how initial, terminal, and waypoint constraints can be translated into linear algebraic equations, and how inequality constraints on state and control can be managed through time scaling.

\paragraph{To learn more.}
For comprehensive treatments of optimal control, we point the reader to several excellent references, including \citet{Murray2009}, \citet{Kirk2004}, \citet{Rao2010}, and \citet{Kelly2017}.
\citet{Kirk2004} offers a foundational perspective on indirect methods, including detailed derivations of the calculus of variations and Pontryagin's Minimum Principle.
For a thorough exploration of direct methods and modern optimization-based approaches, \citet{Rao2010} and \citet{Kelly2017} provide in-depth coverage of transcription techniques, such as collocation and shooting, and their formulation as nonlinear programming problems. 
Finally, \citet{Murray2009} gives an extensive overview of differential flatness, illustrating how this property can be exploited for efficient trajectory generation.

\section{Exercises}
The starter code for the exercises provided below is available online through GitHub. 
To get started, download the code by running in a terminal window:

\begin{tcolorbox}[colback=gray!10]
\begin{minted}{bash}
    git clone https://github.com/StanfordASL/pora-exercises.git
\end{minted}
\end{tcolorbox}

We denote Problems requiring hand-written solutions and coding in Python with \adjustbox{height=2ex, valign=c}{\includegraphics{figs/write.png}} and \adjustbox{height=2ex, valign=c}{\includegraphics{figs/code.png}}, respectively.

\subsection*{\adjustbox{height=2ex, valign=c}{\includegraphics{figs/write.png}}\ Problem 1: Extremal Curves}
[This exercise is inspired by \citet{Kirk2004}, Chapter 4, Problem 4.9]

\noindent Given the functional:
\begin{equation*}
    J(x) = \int_0^1 \rbr*{ \frac{1}{2}\dot{x}(t)^2 + 5x(t)\dot{x}(t) + x(t)^2 + 5x(t) } \d t,
\end{equation*}
find an extremal curve $x^* : [0,1] \to \R$ that satisfies $x^*(0) = 1$ and $x^*(1) = 3$.

\subsection*{\adjustbox{height=2ex, valign=c}{\includegraphics{figs/write.png}}\ Problem 2: Minimum Control Effort}
Consider the dynamics:
\begin{equation*}
    \dot{x}(t) = -2x(t) + u(t),
\end{equation*}
with the initial constraint $x(0) = 2$, terminal constraint $x(1) = 0$, and cost functional:
\begin{equation*}
    J(u) = \int_0^1 u(t)^2\,\d t.
\end{equation*}
Write down the Hamiltonian and use the necessary optimality conditions to derive an optimal control $u^*(t)$ and corresponding state trajectory $x^*(t)$.

\subsection*{\adjustbox{height=2ex, valign=c}{\includegraphics{figs/write.png}}\ Problem 3: Zermelo's Ship}
Zermelo's ship must travel through a region of strong currents. The position of the ship is denoted by $(x(t),y(t)) \in \R^2$. The ship travels at a constant speed $v > 0$, yet its heading $\theta(t)$ can be controlled. The current moves in the positive $x$-direction with speed $w(y(t))$. The equations of motion for the ship are:
\begin{equation*}
\begin{aligned}
    \dot{x}(t) &= v\cos\theta(t) + w(y(t)), \\
    \dot{y}(t) &= v\sin\theta(t).
\end{aligned}
\end{equation*}
We want to control the heading $\theta(t)$ such that the ship travels from a given initial position $(x(t_0), y(t_0)) = (x_0,y_0)$ to the origin $(0,0)$ in minimum time.
\begin{enumerate}
    \item Suppose $w(y(t)) = \frac{v}{h}y(t)$, where $h > 0$ is a known constant. Show that an optimal control law $\theta^*(t)$ must satisfy a linear tangent law of the form:
    \begin{equation*}
        \tan\theta^*(t) = \alpha - \frac{v}{h}t,
    \end{equation*}
    for some constant $\alpha \in \R$.
    
    \item Suppose $w(y(t)) \equiv \beta$ for some constant $\beta > 0$. Derive an expression for the optimal transfer time $t_1^* - t_0$.
\end{enumerate}

\subsection*{\adjustbox{height=2ex, valign=c}{\includegraphics{figs/write.png}}\ Problem 4: Singular Arc for Dubins' Car}
The kinematics of Dubins' car are described by:
\begin{equation*}
\begin{aligned}
    \dot{x} &= v\cos\theta, \\
    \dot{y} &= v\sin\theta, \\
    \dot{\theta} &= u,
\end{aligned}
\end{equation*}
where $(x,y) \in \R^2$ is the car's position, $\theta \in \R$ is the car's heading, $v > 0$ is the car's constant known speed, and $u$ is the controlled turn rate. The turn rate is bounded, meaning $u \in [-\bar{\omega},\bar{\omega}]$, where $\bar{\omega} > 0$ is a known constant. 

The car starts at $(x,y) = (0,0)$ with a heading of $\theta = 0$ at $t = 0$. We want the car to drive to $(x,y) = (0,c)$ in the least amount of time possible, where $c > 0$ is a given constant.

\begin{enumerate}
    \item Use Pontryagin's maximum principle to express the optimal control input $u^*(t)$ as a function of the optimal co-state $p^*(t) \defn (p_x^*(t), p_y^*(t), p_\theta^*(t)) \in \R^3$.
    \begin{hint}
        You should discover that the maximum condition for $u^*(t)$ is not informative whenever $p^*_\theta(t) \equiv \bar{p}_\theta$ for a particular fixed value $\bar{p}_\theta \in \R$. When such a lack of information persists over a non-trivial time interval, i.e., any time interval $[t_1,t_2]$ with $t_2 > t_1 \geq 0$, this is known as a \emph{singular arc}. To compute $u^*(t)$ in this case, use the fact that $p^*_\theta(t) \equiv \bar{p}_\theta$ is constant in time along this singular arc.
    \end{hint}

    \item Use boundary conditions to argue why $p^*(t)$ might end in a singular arc. Suppose we know $p^*(t)$ begins on a non-singular arc, then switches once to and ends on a singular arc. For this particular case, argue why $u^*(0) = \bar{\omega}$ and describe the optimal state trajectory $(x^*(t), y^*(t), \theta^*(t))$ and control trajectory $u^*(t)$ in words without explicitly deriving them.
\end{enumerate}

\subsection*{\adjustbox{height=2ex, valign=c}{\includegraphics{figs/write.png}}\ \adjustbox{height=2ex, valign=c}{\includegraphics{figs/code.png}}\ Problem 5: Single Shooting for a Unicycle}
Consider the kinematic model of a unicycle:
\begin{equation*}
\begin{aligned}
    \dot{x}         &= v \cos(\theta), \\
    \dot{y}         &= v \sin(\theta), \\
    \dot{\theta}    &= \omega,
\end{aligned}
\end{equation*}
where $(x, y)$ is the planar position of the vehicle, $\theta$ is its heading angle, $v$ is its forward velocity, and $\omega$ is its angular velocity. Overall, the state and control input for this system are $x \defn (x, y, \theta) \in \R^3$ and $u \defn (v, \omega) \in \R^2$, respectively. We have overloaded $x$ to denote both horizontal position $x \in \R$ and the full state vector $x \in \R^3$.

Our task is to drive the vehicle from the starting configuration $x(0) = (0, 0, \pi/2)$ to the target configuration $x(T) = (5, 5, \pi/2)$ in minimum time with as little control effort as possible. To this end, we consider the objective:
\begin{equation*}
J(x,u) = \int_0^T \rbr*{ \alpha + v(t)^2 + \omega(t)^2 } \,\d t,
\end{equation*}
where $\alpha > 0$ is a chosen constant weighting factor and $T$ is the free final time.
\begin{enumerate}
    \item Derive the Hamiltonian and necessary optimality conditions, specifically
    \begin{enumerate}
        \item the ODE for the state and co-state,
        \item the optimal control as a function of the state and co-state, and
        \item the boundary conditions, including the additional condition for free final time $T$.
    \end{enumerate}
\end{enumerate}\vspace{-1.5em}
\begin{hint}
    Since the control set is unbounded, use the weak maximum condition.
\end{hint}

In practice, you might use a boundary value problem (BVP) solver from an existing computing library, such as \href{https://docs.scipy.org/doc/scipy/reference/generated/scipy.integrate.solve_bvp.html}{\colorcode{scipy.integrate.solve\_bvp}}, but in this problem we will use a bit of nonlinear optimization theory and JAX to write our own!
\begin{enumerate}[resume]
    \item In the file \colorcode{ch02/exercises/unicycle\_optimal\_control.ipynb}, complete the implementations of \colorcode{unicycle\_dynamics}, \colorcode{hamiltonian}, \colorcode{optimal\_control}, and \colorcode{noc\_ode}. Use $\alpha = 0.25$.
\end{enumerate}

In the single shooting method, we need to initialize estimates of the initial co-state $p(0)$ and final time $T$. We then integrate the state and co-state dynamics forward in time from $t = 0$ to $t = T$, at which point we check whether the terminal boundary conditions are satisfied.
\begin{enumerate}[resume]
    \item Use the ODE integration from \colorcode{noc\_trajectories} to complete \\\noindent\colorcode{boundary\_residual}, which should compute a measure of how far off each of your terminal boundary conditions is from satisfaction, given guesses for the initial co-state $p(0)$ and final time $T$.

    \item Finally, in \colorcode{newton\_step} and \colorcode{single\_shooting}, implement the Newton-Raphson root-finding method for \colorcode{boundary\_residual}. 
    Now, if you provide an appropriate guess for the initial costate and final time, you can solve the problem in \colorcode{unicycle\_optimal\_control.ipynb} and see a plot of the optimal solution. 
    You may find that whether or not your BVP solver converges to a solution is highly dependent on the quality of your initial guess---indeed, initialization is a major challenge when applying indirect methods for optimal control!
    \begin{hint}
        For finding roots of a function $f : \R^n \to \R^n$, each iteration of the Newton-Raphson method entails improving a current best guess $x^{(k)}$ at iteration $k$ using the update rule:
        \begin{equation*}
        x^{(k+1)} = x^{(k)} - \inv{\pd{f}{x}(x^{(k)})}f(x^{(k)}).
        \end{equation*}
    \end{hint}
\end{enumerate}
\newpage
\printbibliography[segment=\therefsegment,heading=subbibliography,title={References}]
\chapter{Closed-Loop Control \& Trajectory Tracking}
\label{ch:closedloop}
\newrefsegment
In \cref{ch:openloop}, we introduced trajectory optimization as a foundational tool for \emph{open-loop} optimal control, where the objective is to compute a time-parameterized control sequence that minimizes a given cost function subject to system dynamics and constraints.
Within a robot’s autonomy stack, open-loop control is often a key component, generating control sequences for tasks such as manipulation, locomotion, and navigation.
However, because open-loop strategies depend only on time and not on the current system state, they are inherently vulnerable to disturbances, modeling errors, and other execution-time uncertainties.

In this chapter, we turn our attention to \emph{closed-loop} optimal control.
While the core elements of the problem remain the same---namely, the system dynamics, state and input constraints, and a performance criterion---the object of optimization is fundamentally different.
Specifically, rather than seeking an open-loop sequence $\u(t)$, closed-loop control aims to determine an optimal control law of the form $\u(t) = \pi(\x(t), t)$, which explicitly depends on the current state in addition to the current time.
Such closed-loop control laws are often referred to as \emph{feedback controllers} or \emph{control policies}.

\begin{figure}[t]
    \centering 
    \includegraphics[width=0.8\linewidth]{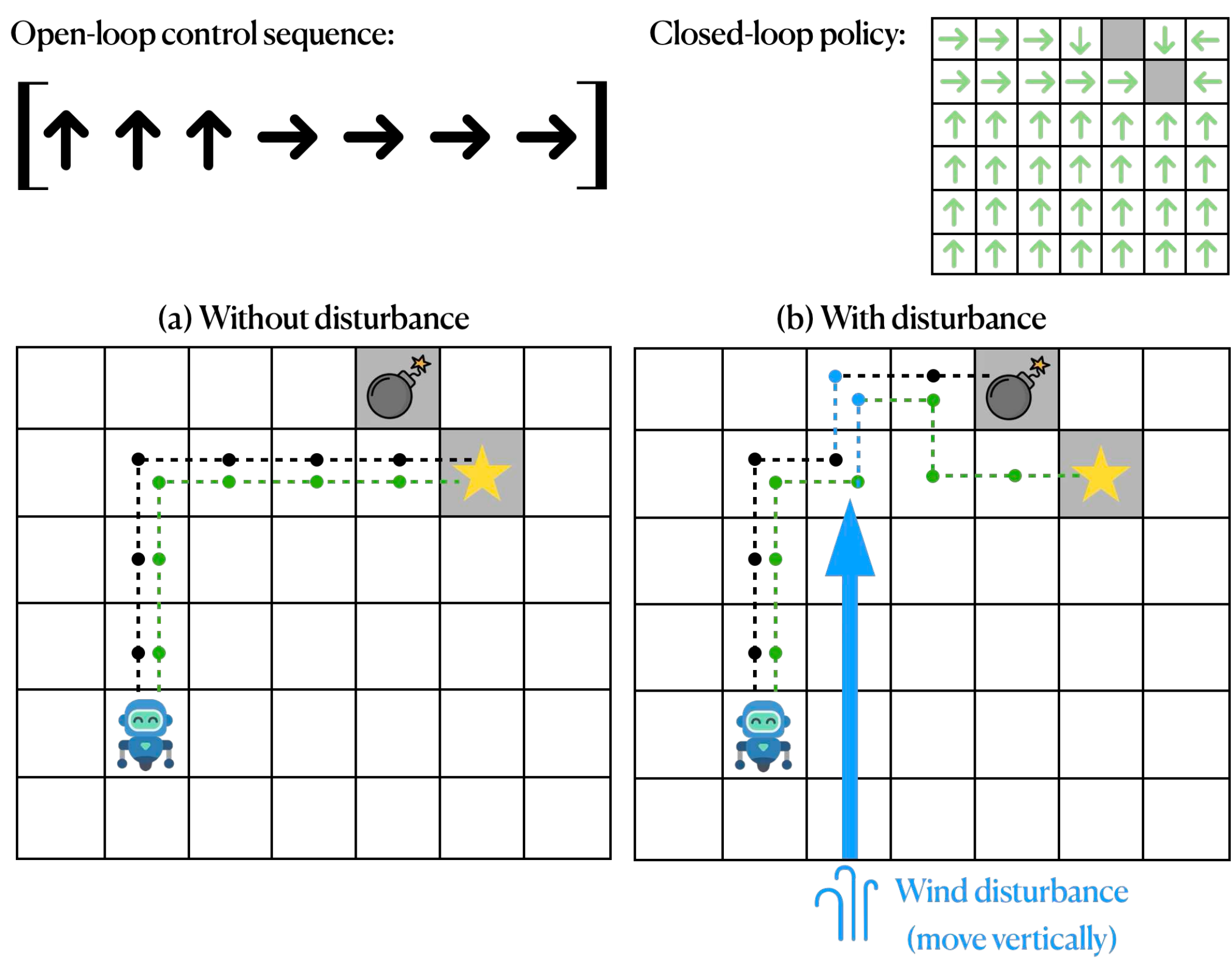}
    \caption{Comparison between open-loop and closed-loop control in a 2D gridworld navigation task.
    Without disturbances (left figure), we can see both open and closed-loop policies lead to same path, ending at the goal (star).
    With a wind disturbance (right figure), the fixed series of actions from the open-loop control sequence fails to take the robot to the goal after it is pushed upward by the wind.
    In contrast, the closed-loop policy specifies an action for each state, allowing the system to react to the disturbance and successfully reach the goal by adjusting its behavior based on the current state.} 
    \label{fig:ol_vs_cl} 
\end{figure}

Obtaining solutions to an optimal control problem in the form of a closed-loop policy is particularly powerful, as it provides a rule prescribing optimal behavior from \emph{any} state the system might encounter.
Under idealized assumptions---that is, exact system models, perfect knowledge of the state, and absence of disturbances---the optimal open-loop sequence and the optimal closed-loop policy would produce identical behavior.
In practice, however, such conditions are rarely met, with real systems inevitably facing model uncertainties, external disturbances, and measurement noise.
In these settings, closed-loop policies offer a distinct advantage, as they continuously adapt the control input to the actual system state, thereby providing levels of robustness that are difficult to achieve through open-loop strategies (\cref{fig:ol_vs_cl}).

\smallskip
Formally, we define a closed-loop control law as follows:
\begin{definition}[Closed-loop control law]
A closed-loop control law is a function $\pi : \mathbb{R}^n \times \mathbb{R} \to \mathbb{R}^m$ that maps the current state\sidenote{When the full system state is not directly measurable, closed-loop control laws can be defined based on the current measured system outputs, i.e., $\u(t) = \pi(\y(t), t)$, where $\y(t)$ denotes the measured system output at time $t$. In this case, the closed-loop policy is referred to as an \emph{output feedback policy}.} $\x(t) \in \mathbb{R}^n$ and time $t \in \mathbb{R}$ to a control input $\u(t) \in \mathbb{R}^m$:
\begin{equation}
\label{eq:closed_loop_control_law}
\u(t) = \pi(\x(t), t ).
\end{equation}
\end{definition}

Despite their theoretical appeal and practical advantages, the main drawback of closed-loop control approaches is computational, as solving for an optimal closed-loop policy is generally more expensive than computing an open-loop input sequence.
In practice, a useful compromise is achieved by following a \emph{two-step design}, which aims to combine the strengths of both open- and closed-loop paradigms. 
In this scheme, one first computes a nominal reference trajectory by solving an open-loop optimal control problem, using tools such as those introduced in \cref{ch:openloop}.
Then, a closed-loop controller is designed to \emph{track} this reference trajectory during execution.
Conceptually, the resulting controller blends a \emph{feedforward} term---the nominal trajectory---with a \emph{feedback} term that reacts to deviations from it. 
As we will discuss in more depth throughout this chapter, these ideas are at the core of what is commonly referred to as \emph{trajectory tracking control}.

Formally, the two-step design can be expressed as:
\begin{equation}
    \label{eq:tracking_controller}
    \u(t) = \bar{\u}(t) + \pi(\x(t), \bar{\x}(t), t),
\end{equation}
where $\bar{\u}(t)$ and $\bar{\x}(t)$ denote the nominal input and state trajectories, and $\pi$ is a \emph{trajectory-tracking control law} ensuring that the actual trajectory remains close to the planned one despite model mismatch or disturbances.

Building on these concepts, this chapter explores key strategies for closed-loop optimal control and their application to trajectory tracking.
We begin in~\cref{subsec:pid_control} with the classical paradigm of \emph{feedback control}, illustrated through the widely used \emph{proportional–integral–derivative (PID)} controllers and their application to tracking, with a focus on differentially flat systems.
Next, in~\cref{subsec:optimal_control_linear}, we turn to linear optimal control, focusing on the \emph{linear quadratic regulator (LQR)}, its key extensions, and its application to tracking control.
We then discuss nonlinear optimal control problems in~\cref{subsec:nonlinear_closed_loop}, first introducing techniques that compute optimal solutions in closed-loop form---such as the \emph{Hamilton-Jacobi-Bellman (HJB)} equation and \emph{Dynamic Programming (DP)}---and then showing how, for tracking purposes, linear methods can be extended to nonlinear systems through \emph{linearization}.
Building on this, we present how LQR ideas generalize to nonlinear optimal control through algorithms such as \emph{iterative LQR (iLQR)} and \emph{differential dynamic programming (DDP)}, that \emph{simultaneously} generate an open-loop trajectory and closed-loop tracking controller in two-step design form.
Finally, in~\cref{subsec:mpc}, we introduce \emph{model predictive control (MPC)}, a powerful framework that brings together ideas of open-loop and closed-loop optimal control through the framework of receding horizon optimization, and discuss how it can be applied to tracking.

\subsection{Classical Feedback Control}
\label{subsec:pid_control}
The central goal of classical feedback control is to regulate a system’s output so that it tracks a desired reference signal.
For example, in the context of an autonomous vehicle, one might want to control the vehicle's speed to match a target velocity or maintain a specific distance from another vehicle.

The classical paradigm of feedback control is built on a simple yet powerful principle: continuously measure the system output, compare it against the desired reference, and compute corrective actions based on the resulting discrepancy, known as the \emph{error signal}.
These corrective inputs are then applied to the system, ensuring the output continually adjusts toward the desired behavior.
For example, accelerate if the vehicle is below the target speed, or decelerate it if it is above.

Classical feedback controllers are generally designed to satisfy a set of fundamental desiderata:
\begin{itemize}
    \item \textit{Stability:} the controller should ensure that the system remains \textit{stable}. 
    While there are several formal notions of stability, the essential requirement is that, loosely speaking, the system ``is bounded in its behavior''.
    \item \textit{Tracking:} the controller should minimize the error between the system output and the desired reference, ensuring the system follows the commanded trajectory as closely as possible.
    \item \textit{Disturbance rejection:} the controller should attenuate the effects of external disturbances and measurement noise, keeping the system performance relatively unaffected by unexpected inputs.
    \item \textit{Robustness:} the controller should perform well despite uncertainties in the system model or parameters, ensuring reliable operation under varying conditions.
\end{itemize}

This feedback principle underlies many widely used strategies in control engineering.
Among the simplest and most widely adopted strategies arising from this principle is the proportional–integral–derivative (PID) controller\sidenote{PID control is most naturally formulated for single-input, single-output (SISO) linear systems, where the error signal is scalar. Extensions to multi-input multi-output or nonlinear systems are possible, but they often rely on heuristic tuning and provide weaker guarantees.}.
Despite its conceptual simplicity, PID has endured the test of time and remains a dominant tool in control engineering, from regulating temperature in industrial furnaces to stabilizing the flight of drones.

\subsubsection{Structure of PID Control}
\label{subsubsec:pid_structure}
A PID controller computes the control input as a weighted combination of three terms derived from the tracking error $e(t) = r(t) - y(t)$, where $r(t)$ denotes the desired reference and $y(t)$ the measured output:
\begin{equation}
u(t) = k_p \, e(t) + k_i \int_0^t e(\tau) \d\tau + k_d \frac{\d}{\d t} e(t).
\end{equation}
The three terms serve complementary purposes:
\begin{itemize}
    \item Proportional (P): reacts immediately to deviations, producing corrective action proportional to the current error.
    \item Integral (I): accumulates past error, driving steady-state error to zero.
    \item Derivative (D): predicts future trends by responding to the rate of change, improving transient behavior and damping oscillations.
\end{itemize}

\begin{figure}[t]
    \begin{center}
        \begin{tikzpicture}[node distance=2cm, ->]
            \tikzstyle{block} = [draw=black, rectangle, minimum height=2em, minimum width=3em, thick]
            \node[draw, circle, inner sep=1pt](sum) {$+$};
            \node[right=1.5cm of sum, block](i){$k_i \int_{0}^{t}e(\tau)\d \tau$};
            \node[above=1cm of i, block](p){$k_p e(t)$};
            \node[below=1cm of i, block](d){$k_d \frac{\d e(t)}{\d t}$};
            \node[draw, circle, inner sep=1pt, right=1.5cm of i](sum2) {$+$};
            \node[right=1.5cm of sum2, block](plant) {System Dynamics};
            \draw[->](sum) to node[above, pos=0.3]{$e(t)$} (i) {};
            \draw[->](i) to (sum2) {};
            \draw[->](sum2) to node[above, pos=0.5]{$u(t)$} (plant){};
            \draw[->] ($(plant.east)+(0,0)$) -- ($(plant.east)+(1,0)$) node[above, pos=0.5]{$y(t)$};
            \draw[->] (p) -| (sum2.north) {};
            \draw[->] (d) -| (sum2.south) {};
            \draw[->] ($(sum.east)+(1,0)$) |- (p.west) {};
            \draw[->] ($(sum.east)+(1,0)$) |- (d.west) {};
            \draw[->] ($(plant.east)+(0.5,0)$) -- ($(plant.east)+(0.5,-2.5)$) -| (sum.south) node[left, pos=0.9] {$-$};
            \draw[->] ($(sum.west)+(-1,0)$) -- node[above, pos=0.5] {$r(t)$}(sum.west) {};
        \end{tikzpicture}
    \end{center}
    \caption{Block diagram for a PID controller in a feedback loop. The error signal $e(t)$ is computed as the difference between the reference $r(t)$ and the output $y(t)$. The control input $u(t)$ is generated by combining the proportional, integral, and derivative terms of the error.}
    \label{fig:pid}
    \end{figure}
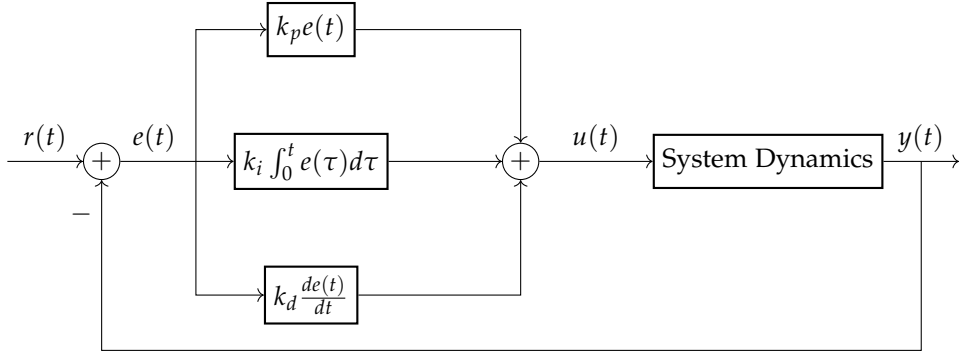

By adjusting the gains $(k_p, k_i, k_d)$, we can shape how aggressively the controller responds to disturbances, how quickly it eliminates offsets, and how smoothly it approaches the reference trajectory\sidenote{Although tuning PID controllers is not trivial, heuristic rules such as the Ziegler–Nichols method provide systematic starting points for gain selection, but they often require further refinement in practice.}.
The structure of a PID controller is depicted in the block diagram in~\cref{fig:pid}. 
PID control is ubiquitous in industry precisely because of this simplicity: the controller can be deployed with only limited implementation effort and modest computational resources, which makes it attractive for embedded systems.

\begin{example}[PD control of a double-integrator system]
    \label{ex:pd_double_int}
    To illustrate the mechanics of PID-type control, consider a double-integrator system with dynamics:
    \begin{equation*}
        \dot{\x}(t)=\begin{bmatrix} 0 & 1 \\ 0 & 0 \end{bmatrix}\x(t)+\begin{bmatrix} 0 \\ 1 \end{bmatrix}u(t),
    \end{equation*}
    where $\x = \vCol{x_1, x_2}$.
    Suppose the control objective is to regulate $x_1$ to the origin. 
    A proportional–derivative controller of the form:
    \begin{equation*}
        u(t) = - k_p e(t) - k_d \frac{\d e(t)}{\d t}, \quad e(t) = x_1(t),
    \end{equation*}
    yields closed-loop dynamics:
    \begin{equation*}
    \dot{\x}(t)=\begin{bmatrix} 0 & 1 \\ -k_p & -k_d \end{bmatrix}\x(t),
    \end{equation*}
    where we have used the fact that $\dot{e} = \dot{x}_1 = x_2$.
    The eigenvalues of this system are:
    \begin{equation*}
    \lambda = -\frac{k_d}{2} \pm \frac{1}{2}\sqrt{k_d^2 - 4k_p},
    \end{equation*}
    and, as a result, stability requires selecting $k_p$ and $k_d$ such that the real parts of these eigenvalues are negative; that is, $k_p > 0$ and $k_d > 0$\sidenote{Note that a proportional controller alone, with $k_d = 0$, will not make the system stable since at best the eigenvalues would be purely complex, and thus the system response would be an undamped oscillation.}.
    By appropriately tuning $k_p$ and $k_d$, we can trade off speed of convergence against overshoot, oscillations, and sensitivity to measurement noise.
    For example, one might set the proportional gain $k_p$ to a very large value in order to drive rapid convergence to the origin.
    In practice, however, excessively large gains tend to amplify measurement noise and can cause the controller to behave poorly.
    For example, with large proportional gain $k_p$, the controller may introduce oscillations, while a large derivative gain $k_d$ increases damping but may also slow the system’s response.
\end{example}

\medskip
Beyond PID and PD control, classical control theory provides a rich suite of tools for analysis and design. 
These methods are often developed in the frequency domain\sidenote{Frequency-domain analysis uses the Laplace transform to represent system dynamics as algebraic relations---referred to as \emph{transfer functions}---rather than differential equations.}, where tools such as \emph{Bode plots} and the \emph{Nyquist stability criterion} offer powerful ways to assess and shape system behavior.
For an in-depth treatment of these topics, we refer the reader to \citet{AstromMurray2009}.

\subsubsection{Application of PID Control to Tracking Problems}
Although PID controllers are most commonly associated with regulation tasks, as discussed in~\cref{ex:pd_double_int}, the same structure introduced in~\cref{subsubsec:pid_structure} can also be employed for trajectory tracking, where the objective is to follow a time-varying nominal trajectory.
In this setting, the reference signal $r(t)$ evolves over time, and the controller seeks to minimize the instantaneous tracking error $e(t) = r(t) - y(t)$ throughout the motion.
While this direct application of the PID framework can perform reasonably well for simple scenarios, such as slowly varying or smooth trajectories, it remains largely heuristic, requiring ad hoc design and tuning of the gains for each specific task and system.
As a result, PID-based tracking tends to be fragile and poorly generalizable, with performance degrading even under minor variations to the problem setup.
There is, however, an important class of systems where PID-type control can be systematically and effectively applied to tracking: differentially flat systems.

As introduced in~\cref{sec:differentially_flat_systems}, differential flatness provides a powerful framework for computing open-loop control sequences.
Recall that a system with state $\x \in \R^n$ and input $\u \in \R^m$:
$$
\dot{\x} = f(\x, \u),
$$
is said to be differentially flat if there exists a function $\alpha$ such that:
$$
\z = \alpha(\x, \u, \dot{\u}, \dots, \u^{(a)}),
$$
where $\u^{(i)}$ denotes the $i$-th derivative of $\u$, and such that both the system states and inputs can be expressed as algebraic functions of the flat outputs $\z \in \R^m$ and a finite number of their derivatives:
$$
\x = \beta(\z, \dot{\z}, \dots, \z^{(b)}), \quad
\u = \gamma(\z, \dot{\z}, \dots, \z^{(c)}).
$$
This structural property implies that the full system trajectory is fully determined once the flat outputs $\z(t)$ are known.
In practice, we can therefore design a trajectory directly in the flat-output space---using, for example, spline interpolation or polynomial parameterizations---and reconstruct the corresponding state and control trajectories algebraically, without the need to integrate the system dynamics.

While in~\cref{ch:openloop} we focused on leveraging differential flatness for open-loop trajectory generation, it is important to highlight how these systems also lend themselves naturally to trajectory tracking.
In particular, differential flatness enables the trajectory tracking problem for a nonlinear system to be reduced to a \emph{linear} tracking problem in the flat-output space, where classic control methods---such as PID---can be applied.

\smallskip
Differentially flat systems possess a particularly useful feature, as they can be \emph{feedback linearized} to yield a linear system in the flat-output space.
Specifically, given a differentially flat system with flat output $\z = (z_1, \dots, z_m)$ there exist a vector of integers $\bm{r} = (r_1, r_2, \dots, r_m)$ such that:
\begin{equation}
    \begin{split}
    \x & = \beta(z_1, \dot{z}_1, \dots, z_1^{(r_1)}, \dots, z_m, \dot{z}_m, \dots, z_m^{(r_m)}), \\
    \u & = \gamma(z_1, \dot{z}_1, \dots, z_1^{(r_1+1)}, \dots, z_m, \dot{z}_m, \dots, z_m^{(r_m+1)}),
    \end{split}
\end{equation}
and such that the system dynamics can be equivalently expressed as a linear system of the form:
\begin{equation}
    \begin{split}
    z_i^{(r_1+1)} & = w_1, \\
    z_2^{(r_2+1)} & = w_2, \\
    \vdots \\
    z_m^{(r_m+1)} & = w_m,
    \label{eq:flat_feedback_linearized}
    \end{split}
\end{equation}
where $\bm{w} = (w_1, \dots, w_m)$ is a \emph{virtual} control input that can be algebraically related to the original system inputs $\u$\sidenote{For a detailed treatment of feedback linearization for differentially flat systems, we refer the reader to \citet{Levine2009} and \citet{Murray2009}.}.
The linear system in~\cref{eq:flat_feedback_linearized} can effectively be controlled using standard linear control techniques, such as PID control.

In particular, given a reference flat output trajectory $\z_d = (z_{d,1}, \dots, z_{d,m})$---computed, for example, by any open-loop method---and its corresponding virtual input trajectory $\bm{w}_d = (w_{d,1}, \dots, w_{d,m})$, we can define the \emph{component-wise} tracking error:
$$
e_i \coloneqq z_i - z_{d, i}, \quad i = 1, \ldots, m,
$$
which implies the following error dynamics:
$$
e_i^{(r_i+1)} = w_i - w_{d,i}.
$$
To ensure convergence of the tracking error to zero, we can choose the following control law:
$$
w_i = w_{d,i} - \sum_{j=0}^{r_i} k_{i,j} e_i^{(j)},
$$
which, applied to the system in~\cref{eq:flat_feedback_linearized}, yields the following closed-loop dynamics:
$$
z_{i}^{(r_i+1)} = w_{d,i} - \sum_{j=0}^{r_i} k_{i,j} e_i^{(j)}.
$$
Since $w_{d,i} = z_{d,i}^{(r_i+1)}$ by construction, the resulting tracking error dynamics take the form:
$$
e_i^{(r_i+1)} + \sum_{j=0}^{r_i} k_{i,j} e_i^{(j)} = 0,
$$
where the gains $k_{i,j} > 0$ are selected to enforce stability.

This procedure effectively reduces the nonlinear trajectory tracking problem to a set of decoupled linear tracking problems, one for each flat output, which can be solved using standard linear control techniques.
Below, we illustrate this procedure with a concrete example.

\begin{example}[PD control for a dynamically extended unicycle]
\label{ex:pd-control-extended-unicycle}
    Consider the dynamically extended unicycle model introduced in~\cref{ex:dynamic_unicycle}:
    \begin{equation*}
    \begin{split}
    \dot{x} &= v \cos\theta, \\
    \dot{y} &= v \sin\theta, \\
    \dot{v} &= a, \\
    \dot{\theta} &= \omega,
    \end{split}
    \end{equation*}
    with state $\x = [x, y, v, \theta]^\top$ and control input $\u = [a, \omega]^\top$.
    This system is differentially flat with flat outputs $(x, y)$ and its dynamics can be expressed as:
    \begin{equation}
    \begin{bmatrix}
    \ddot{x} \\[3pt] \ddot{y}
    \end{bmatrix}
    = 
    \underbrace{
    \begin{bmatrix}
    \cos \theta & -v \sin \theta \\
    \sin \theta &  v \cos \theta
    \end{bmatrix}
    }_{J(\theta,v)}
    \underbrace{
    \begin{bmatrix}
    a \\[3pt] \omega
    \end{bmatrix}
    }_{\u}
    \coloneqq
    \begin{bmatrix}
    w_1 \\[3pt] w_2
    \end{bmatrix},
    \label{eq:unicycle_flat_dynamics}
    \end{equation}
    where $\bm{w} = [w_1, w_2]^\top$ is a virtual control input representing the flat-output accelerations, effectively transforming the system into a system of the form in~\cref{eq:flat_feedback_linearized}.

    As a result, given a reference trajectory $(x_d, y_d)$, we can design a feedback law for $\bm{w}$ that ensures tracking of the desired trajectory.
    Specifically, we can select the following PD-type virtual control:
    \begin{equation}
    \label{eq:pd-unicycle-control}
    \begin{split}
    w_1 &= \ddot{x}_d + k_{px}(x_d - x) + k_{dx}(\dot{x}_d - \dot{x}), \\
    w_2 &= \ddot{y}_d + k_{py}(y_d - y) + k_{dy}(\dot{y}_d - \dot{y}),
    \end{split}
    \end{equation}
    with positive control gains $k_{px}, k_{dx}, k_{py}, k_{dy} > 0$.
    Under this law, the Cartesian tracking error obeys a second-order linear differential equation:
    $$
    \ddot{e}_x + k_{dx}\dot{e}_x + k_{px} e_x = 0, \qquad
    \ddot{e}_y + k_{dy}\dot{e}_y + k_{py} e_y = 0,
    $$
    ensuring that the tracking error $(e_x, e_y) = (x_d - x, y_d - y)$ converges exponentially to zero.

    Finally, assuming the Jacobian matrix $J(\theta,v)$ is invertible, the corresponding control inputs $(a, \omega)$ can be obtained algebraically by:
    $$
    \begin{bmatrix} a \\[3pt] \omega \end{bmatrix}
    = J^{-1}(\theta,v)
    \begin{bmatrix} w_1 \\[3pt] w_2 \end{bmatrix}.
    $$
    Thus, a conceptually simple PD control law in the flat-output space translates into a nonlinear state feedback law for the original system.
\end{example}

\subsection{Optimal Closed-loop Control for Linear Systems}
\label{subsec:optimal_control_linear}
Despite their widespread use and practical appeal, classical feedback controllers lack systematic methods to mathematically quantify and optimize performance. 
As introduced in~\cref{ch:openloop}, modern control theory addresses these issues by formulating controller design as an explicit optimization problem.

In this section, we focus on the linear–quadratic setting\sidenote{That is, where the system dynamics are linear and the cost function is quadratic.}, which represents one of the most elegant and widely applicable frameworks for closed-loop optimal control.
Specifically, we focus on the linear quadratic regulator problem by first presenting its infinite-horizon, continuous-time formulation.
Then, we discuss several important extensions of this framework---such as finite-horizon, discrete-time, and time-varying formulations---and illustrate how it can be applied to linear tracking problems.

\subsubsection{The Linear Quadratic Regulator}
\label{subsubsec:lqr}
The linear quadratic regulator (LQR) provides a principled solution to the \emph{regulation} problem, where the primary objective is to drive the state of a linear system to the origin while optimally balancing state deviations against control effort\sidenote{While this may seem like a restrictive setting at first, we will see in the remainder of this chapter that this formulation extends naturally to a wide range of practically relevant problems.}.
Formally, consider the linear time-invariant (LTI) system:
\begin{equation}
    \dot{\x}(t) = A \x(t) + B \u(t),
    \label{eq:linear_dynamics}
\end{equation}
where $A \in \R^{\statedim \times \statedim}$ and $B \in \R^{\statedim \times \controldim}$ are constant matrices.
The objective is to determine a control policy $\u(t)$ that minimizes the quadratic cost functional:
\begin{equation}
    J(\u) = \int_0^\infty \x(t)^\top Q \x(t) + \u(t)^\top R \u(t) \d t,
\label{eq:lqr_cost}
\end{equation}
where $Q \in \R^{\statedim \times \statedim}$ is a symmetric positive semidefinite matrix that penalizes deviations of the state from the origin and $R \in \R^{\controldim \times \controldim}$ is a symmetric positive definite matrix that penalizes control effort.
Thus, in its \emph{infinite-horizon} form, the LQR problem is formalized as the following optimal control problem:
\begin{equation}
\begin{split}
    \minimize[\u] &\int_{0}^{\infty} \x(t)^\top Q\x(t)+ \u(t)^\top R \u(t) \d t,\\
    \subjectto &\dot{\x}(t)=A\x(t)+B\u(t).
\end{split}
\end{equation}
Assuming that the pair $(A,B)$ is \emph{stabilizable} and $(Q, A)$ is \emph{detectable}\sidenote{These assumptions guarantee the existence of a unique stabilizing solution to the infinite-horizon LQR problem. For formal definitions and proofs, we refer the reader to \citet{Murray2009} and \citet{Bertsekas2000}.}, the LQR problem admits a unique optimal solution in the form of a stationary linear state-feedback controller:
\begin{equation}
    \u(t) = -K \x(t),
    \label{eq:linear_feedback}
\end{equation}
where the optimal gain matrix $K \in \R^{\controldim \times \statedim}$ is given by:
\begin{equation}
    K = R^{-1} B^\top P.
    \label{eq:lqr_gain}
\end{equation}
Here, $P \in \R^{\statedim \times \statedim}$ is the unique positive semidefinite solution to the \emph{continuous-time algebraic Riccati equation}:
\begin{equation}
    A^\top P + PA - PBR^{-1}B^\top P + Q = 0.
    \label{eq:algebraic_riccati}    
\end{equation}
The Riccati equation can be derived either through the calculus of variations---via the Pontryagin Maximum Principle introduced in~\cref{ch:openloop}---or through dynamic programming, using the Hamilton–Jacobi–Bellman equation\sidenote{Both derivations lead to the Riccati equation and the same optimal linear feedback law. We refer the reader to \citet{Bertsekas2000} for an in-depth treatment of both approaches.}.
Substituting the optimal closed-loop policy from~\cref{eq:linear_feedback} into the system dynamics from~\cref{eq:linear_dynamics} yields the closed-loop system:
\begin{equation}
    \dot{\x}(t) = (A - BK)\x(t),
\end{equation}
which is guaranteed to be asymptotically stable, i.e., the eigenvalues of the matrix $A - BK$ have strictly negative real parts.
Thus, the LQR problem provides a direct recipe for the regulation problem: solve the Riccati equation\sidenote{In practice, solutions to the algebraic Riccati equation are obtained numerically using standardized software packages. Popular implementations include \colorcode{scipy.linalg.solve\_continuous\_are} in Python or \colorcode{icare} in MATLAB.}, compute the optimal gain matrix $K$, and apply the feedback law $\u(t) = -K\x(t)$.

As a concrete example, Algorithm~\ref{alg:lqr_double_integrator} demonstrates how to compute the optimal infinite-horizon LQR controller for a simple double integrator system with dynamics $\ddot{x} = u$.

\begin{listing}[ht!]
\begin{tcolorbox}[colback=gray!10, colframe=gray!50, 
    title=Double Integrator Infinite-horizon LQR, boxrule=0.5mm, arc=0mm]
\begin{minted}[escapeinside=||]{python}
import numpy as np
from scipy.linalg import solve_continuous_are

# System dynamics matrices for simple double integrator |$\ddot{x}$| = u
A = np.array([[0, 1], [0, 0]])
B = np.array([[0], [1]])

# Define cost function matrices
Q = np.array([[1, 0], [0, 1]])
R = np.array([[1]])

# Solve the continuous algebraic Riccati equation
P = solve_continuous_are(A, B, Q, R)

# Compute optimal feedback gain matrix
K = -np.linalg.inv(R) |$@$| B.T |$@$| P

# Verify eigenvalues are negative (closed-loop system is stable)
eig_val, eig_vec = np.linalg.eig(A + B |$@$| K)
print(eig_val)
\end{minted}
\end{tcolorbox}
\caption{Computing the optimal infinite-horizon LQR controller for a simple double integrator with dynamics $\ddot{x} = u$ in Python.
The code for this example is available in the repository \colorcode{github.com/StanfordASL/pora-exercises} in the notebook \colorcode{ch03/lqr.ipynb}.}
\label{alg:lqr_double_integrator}
\end{listing}

\paragraph{Designing the cost matrices.}
A central question in LQR design is how to select the cost matrices $Q$ and $R$, as these determine the trade-off between state regulation and control effort, thereby shaping the overall closed-loop performance.
For a valid solution to exist, the matrices must satisfy $Q=Q^\top \succeq 0$ and $R=R^\top \succ 0$.
To simplify the discussion, we further assume $Q=Q^\top \succ 0$.
Together with the stabilizability and detectability assumptions stated above, this guarantees the existence of a unique stabilizing solution to the algebraic Riccati equation.
In practice, the specific choice of $Q$ and $R$ depends on the designer’s understanding of the system dynamics and performance objectives.
A particularly simple and commonly adopted approach is to use diagonal weight matrices, where each diagonal element directly penalizes a corresponding state or control variable:
$$
Q = \begin{bmatrix}
    q_1 & 0 & \cdots & 0 \\
    0 & q_2 & \cdots & 0 \\
    \vdots & \vdots & \ddots & \vdots \\
    0 & 0 & \cdots & q_n 
\end{bmatrix}, \quad
R = \begin{bmatrix}
    r_1 & 0 & \cdots & 0 \\
    0 & r_2 & \cdots & 0 \\
    \vdots & \vdots & \ddots & \vdots \\
    0 & 0 & \cdots & r_m 
\end{bmatrix},
$$
where $q_i > 0$ and $r_j > 0$ for all $i = 1, \ldots, n$ and $j = 1, \ldots, m$.
With this choice, each diagonal entry directly specifies how strongly the corresponding (squared) state or input contributes to the overall cost.
In general, states (or equivalently, control inputs) that are particularly important to regulate are assigned larger weights, ensuring that deviations in those directions are penalized more heavily. 
Conversely, less critical states or inputs are assigned smaller weights, reflecting their reduced influence on the system's overall performance.

\paragraph{Finite-horizon formulation.}
The LQR formulation can be naturally extended to the \emph{finite-horizon} setting, where performance is evaluated only over a fixed time interval $[0, t_f]$.
In this case, the optimal control problem can be formulated as:
\begin{equation}
    \begin{split}
    \minimize[\u] &\x(t_f)^\top Q_f\x(t_f) +\int_{0}^{t_f} \x(t)^\top Q\x(t)+ \u(t)^\top R \u(t) \d t,\\
    \subjectto &\dot{\x}(t)=A\x(t)+B\u(t),
\end{split}
\label{eq:finite_horizon_lqr}
\end{equation}
where $Q_f \in \R^{\statedim \times \statedim}$ and $Q \in \R^{\statedim \times \statedim}$ are symmetric positive semidefinite matrices, and $R \in \R^{\controldim \times \controldim}$ is symmetric positive definite.
The matrix $Q_f$ specifies the terminal cost, penalizing deviations of the state from the origin at the final time $t_f$.

As in the infinite-horizon case, the optimal control law retains the linear form $\u(t) = -K(t)\x(t)$, but the gain matrix becomes \emph{time-varying}.
Specifically, the optimal gain matrix $K(t)$ is given by:
\begin{equation}
    K^*(t) = R^{-1} B^\top P(t),
    \label{eq:finite_horizon_gain}
\end{equation}
where $P(t)$ is the positive semidefinite matrix that solves the \emph{continuous-time differential Riccati equation}:
\begin{equation*}
    \dot{P}(t)= -A^\top P(t)-P(t)A + P(t)BR^{-1}B^\top P(t) - Q,
\end{equation*}
with terminal condition $P(t_f) = Q_f$.

As the horizon length $t_f$ tends to infinity, and under the standard assumptions introduced for the infinite-horizon case---namely, that $(A,B)$ is stabilizable, $(Q, A)$ is detectable, $Q=Q^\top \succeq 0$, and $R=R^\top \succ 0$---the solution $P(t)$ of the differential Riccati equation converges to the steady-state solution $P$ of the algebraic Riccati equation from~\cref{eq:algebraic_riccati}.
Correspondingly, the time-varying gain matrix $K(t)$ converges to the constant gain matrix $K$ from~\cref{eq:lqr_gain}.
In practice, for sufficiently long horizons, it is common to use the infinite-horizon gain directly, thereby avoiding the need to compute or store the entire gain schedule $K(t)$.

\paragraph{Discrete-time LQR.}
Up to this point, we have focused on continuous-time formulations.
However, both the finite-horizon and infinite-horizon problems can be posed just as naturally in discrete time, where the system evolves according to the difference equation:
\begin{equation}
    \x_{t+1} = A\x_t + B\u_t,
\end{equation}
and where, as discussed in~\cref{ch:openloop}, the time horizon $t_f$ is discretized into $N \in \mathbb{N}$ intervals $t = 0, 1, \ldots, N-1$ of length $\Delta t = t_f / N$.
In this setting, the structure of the optimal solution mirrors the continuous-time case, where the optimal state-feedback law is again linear in the state, $\u_t = -K_t \x_t$, with gains obtained by solving a Riccati equation---now in its discrete form.

\smallskip
Specifically, the \emph{finite-horizon} problem in discrete time is formulated as:
\begin{equation}
    \begin{split}
    \minimize[\u] &\x_N^\top Q_f \x_N + \sum_{t=0}^{N-1} \x_t^\top Q \x_t + \u_t^\top R \u_t,\\
    \subjectto &\x_{t+1} = A\x_t + B\u_t, \quad t = 0, \ldots, N-1,
\end{split}
\label{eq:discrete_finite_horizon_lqr}
\end{equation}
where $Q_f=Q_f^\top \succeq 0$, $Q=Q^\top \succeq 0$, and $R=R^\top \succ 0$.
The optimal solution is obtained through a backward Riccati recursion, which proceeds from the terminal condition $P_{N} = Q_f$ and iterates backward in time for $t = N-1, N-2, \ldots, 0$ according to:
\begin{equation}
    P_t = Q + A^\top P_{t+1} A - A^\top P_{t+1} B \bigl(R + B^\top P_{t+1} B \bigr)^{-1} B^\top P_{t+1} A,
\label{eq:discrete_riccati}
\end{equation}
with the corresponding feedback gain:
\begin{equation}
    K_t = \bigl(R + B^\top P_{t+1} B \bigr)^{-1} B^\top P_{t+1} A.
\end{equation}
This recursion can be viewed as the discrete-time analogue of integrating the continuous-time differential Riccati equation backward from $t_f$ to $0$. 
At each step, $P_t$ is updated one stage earlier, effectively propagating future cost information back through time to determine the optimal closed-loop policy.

\smallskip
In the \emph{infinite-horizon} setting, the problem is formulated as:
\begin{equation}
    \begin{split}
    \minimize[\u] &\sum_{t=0}^{\infty} \x_t^\top Q \x_t + \u_t^\top R \u_t,\\
    \subjectto &\x_{t+1} = A\x_t + B\u_t,
\end{split}
\end{equation}
where $Q=Q^\top\succeq 0$ and $R=R^\top\succ 0$.
Analogous to the continuous-time case, and under the standard stabilizability and detectability conditions, the backward Riccati recursion converges---as the horizon length $t_f$ tends to infinity---to a steady-state matrix $P$, which satisfies the \emph{discrete-time algebraic Riccati equation}:
\begin{equation}
\label{eq:dare}
    P = Q + A^\top P A - A^\top P B (R + B^\top P B)^{-1} B^\top P A.
\end{equation}
The corresponding optimal feedback gain is then time-invariant and given by:
\begin{equation}
    K = (R + B^\top P B)^{-1} B^\top P A.
\end{equation}

\smallskip
Thus, while the overall structure of the LQR solution carries over seamlessly to discrete time, the computational machinery changes, where instead of solving a continuous-time differential or algebraic Riccati equation, one either propagates a backward recursion over a finite horizon (finite case) or solves a fixed-point equation (infinite case). 
This recursive viewpoint is also closely aligned with the principles of dynamic programming, which is a foundational concept in optimal control and reinforcement learning and will be discussed in more detail in~\cref{ch:sequential-decision-making}.

\medskip
Below, we summarize key extensions of the LQR problem using its discrete-time formulation.

\paragraph{Time-varying LQR with cross-quadratic costs.}
We now consider the finite-horizon, discrete-time LQR problem with both time-varying dynamics and cross-quadratic cost terms:
\begin{equation}
    \begin{split}
    \minimize[\u] &\frac{1}{2}\x_N^\top Q_f\x_N + \sum_{t=0}^{N-1} \left(\frac{1}{2}\x_t^\top Q_t\x_t + \frac{1}{2}\u_t^\top R_t\u_t + \x_t^\top S_t \u_t\right),\\
    \subjectto &\x_{t+1}=A_t\x_t+B_t\u_t, \quad t=0,\ldots,N-1,
    \end{split}
\end{equation}
where $Q_f=Q_f^\top \succeq 0$, and where $Q_t=Q_t^\top \succeq 0$, $R_t=R_t^\top \succ 0$, and $S_t$ are time-varying cost matrices.
The system matrices $A_t$ and $B_t$ are also explicitly time-dependent, capturing nonstationary dynamics.
This formulation extends the standard LQR problem by incorporating the cross-term matrices $S_t$, which penalize state–control interactions. 
Such terms naturally arise when the cost of applying a control action depends on the current system state.
For example, in a vehicle control scenario, the cost of applying a braking force may vary with the vehicle's speed.

The optimal solution retains the linear feedback form $\u_t = -K_t \x_t$, with:
\begin{equation}
    K_t = (R_t + B_t^\top P_{t+1} B_t)^{-1}(B_t^\top P_{t+1} A_t + S_t^\top),
\end{equation}
and where the matrices $P_t$ are obtained from the backward Riccati recursion:
\begin{equation}
    P_t = Q_t + A_t^\top P_{t+1} A_t - (A_t^\top P_{t+1} B_t + S_t)(R_t + B_t^\top P_{t+1} B_t)^{-1}(B_t^\top P_{t+1} A_t + S_t^\top),
\end{equation}
with terminal condition $P_{t_f} = Q_f$.

\paragraph{LQR with affine dynamics and quadratic and linear costs.}
We now consider a more general discrete-time LQR formulation that incorporates affine dynamics together with quadratic, linear, and constant terms in the cost. 
Let $Q_{f}=Q_f^\top \succeq 0$, $Q_t=Q_t^\top \succeq 0$, and $R_t=R_t^\top \succ 0$. 
The cost function is given by:
\begin{equation}
    \begin{split}
    J(\u) & = h(\x_{N}) + \sum_{t=0}^{N-1} g(\x_t, \u_t), \\
    h(\x_N) & = \frac{1}{2}\x_N^\top Q_f \x_N + \bm{q}_f^\top \x_N + \alpha_f, \\
    g(\x_t, \u_t) & = \frac{1}{2}\x_t^\top Q_t \x_t + \frac{1}{2}\u_t^\top R_t \u_t + \x_t^\top S_t \u_t + \\ 
     & + \bm{q}_t^\top \x_t + \bm{r}_t^\top \u_t + \alpha_t, \quad t = 0, \ldots, N-1,
    \end{split}
    \label{eq:lqr_general}
\end{equation}
subject to the affine dynamics:
\begin{equation}
    \x_{t+1} = A_t \x_t + B_t \u_t + \bm{c}_t.
\end{equation}
The optimal control policy can be derived using dynamic programming, yielding a time-varying affine closed-loop control law\sidenote{In this section, we cite the result without proof and refer the reader to \citet{Bertsekas2000} for a detailed derivation of the optimal solution for various LQR extensions.}.
To streamline notation, let us define the following intermediate quantities:
\begin{equation}
    \begin{split}
    \eta_t &\coloneqq \alpha_t + \beta_{t+1} + p_{t+1}^\top \bm{c}_t + \frac{1}{2}\bm{c}_t^\top P_{t+1} \bm{c}_t, \\
    h_{x,t} &\coloneqq \bm{q}_t + A_t^\top \left(p_{t+1} + P_{t+1} \bm{c}_t\right), \\
    h_{u,t} &\coloneqq \bm{r}_t + B_t^\top \left(p_{t+1} + P_{t+1} \bm{c}_t\right), \\
    H_{xx,t} &\coloneqq Q_t + A_t^\top P_{t+1} A_t, \\
    H_{xu,t} &\coloneqq S_t + A_t^\top P_{t+1} B_t, \\
    H_{uu,t} &\coloneqq R_t + B_t^\top P_{t+1} B_t.
    \end{split}
    \label{eq:lqr_general_matrices}
\end{equation}
The optimal law remains affine in the state:
\begin{equation}
\u_t = -K_t\x_t - k_t,
\end{equation}
with parameters given recursively by:
\begin{equation}
\begin{aligned}
    P_{N} &= Q_f, \\
    p_{N} &= \bm{q}_f, \\
    \beta_{N} &= \alpha_f, \\
    K_t^* &\coloneqq H_{uu,t}^{-1} H_{xu,t}^\top, \\
    k_t &\coloneqq H_{uu,t}^{-1} h_{u,t}, \\
    P_{t} &\coloneqq H_{xx,t} - H_{xu, t} K_t^*, \\
    p_{t} &\coloneqq h_{x,t} - H_{xu,t} k_t, \\
    \beta_{t} &\coloneqq \eta_t - \frac{1}{2} h_{u,t}^\top k_t.
\end{aligned}
\end{equation}
This affine–quadratic formulation generalizes the classical LQR by accommodating affine dynamics and nonhomogenous cost terms.
As we will see in the remainder of this chapter, this formulation plays a central role in extending LQR techniques to nonlinear systems, where such affine and cross-linear structures naturally arise through linearization and quadratic approximation of the dynamics and cost functions.

\medskip
In summary, the LQR framework represents a powerful tool for closed-loop optimal control of linear systems with quadratic costs. 
As we will see next, the principles underlying LQR can be extended to tackle more complex scenarios, including trajectory tracking and nonlinear dynamics.

\subsubsection{Linear Tracking Problems}
\label{subsubsec:linear_tracking}
As discussed in the previous section, the LQR framework provides a principled solution to the regulation problem, in which the goal is to drive the system state to the origin.
In many practical settings, however, tasks may extend well beyond regulation.
Robotic manipulators must follow preplanned motions, autonomous vehicles must pass through waypoints, and industrial systems often operate around time-varying setpoints.
In such scenarios, the control objective is one of trajectory tracking rather than regulation.
Thus, restricting control design to regulation about the origin is often insufficient for many real-world applications.

The tracking problem can be naturally formulated within the two-step design paradigm introduced earlier in~\cref{eq:tracking_controller}.
In the first step, an \emph{open-loop} optimal control problem is solved to obtain a nominal state–control trajectory:
$$
\left(\bar{\x}(t), \bar{\u}(t)\right), \quad t \in [0, t_f],
$$ 
that satisfies the system dynamics and optimizes a chosen performance criterion\sidenote{These nominal trajectories are typically computed using trajectory optimization methods, as discussed in~\cref{ch:openloop}.}.
In the second step, a \emph{closed-loop} controller is designed to ensure that the actual system trajectory remains close to this nominal trajectory, compensating for disturbances, modeling errors, and measurement noise.

Formally, consider the linear system given by:
\begin{equation}
\dot{\x}(t) = A\x(t) + B\u(t),
\label{eq:lti_dynamics}
\end{equation}
and let $(\bar{\x}(t), \bar{\u}(t))$ denote a nominal trajectory we wish to track, which satisfies the same dynamics, that is:
\begin{equation}
\dot{\bar{\x}}(t) = A\bar{\x}(t) + B\bar{\u}(t).
\label{eq:lti_nominal}
\end{equation}
We define the \emph{deviation} (or \emph{error}) variables as:
\begin{equation}
\delta\x(t) = \x(t) - \bar{\x}(t), \qquad
\delta\u(t) = \u(t) - \bar{\u}(t),
\label{eq:error_variables}
\end{equation}
which represent the difference between the actual state and control inputs and their nominal counterparts.
By using the definition of $\delta \x(t)$ in~\cref{eq:error_variables} and substituting the dynamics from~\cref{eq:lti_dynamics,eq:lti_nominal}, we can derive the dynamics of the deviation variables:
\begin{align*}
\delta\dot{\x}(t) 
&= \dot{\x}(t) - \dot{\bar{\x}}(t) \\
&= [A\x(t) + B\u(t)] - [A\bar{\x}(t) + B\bar{\u}(t)] \\
&= A(\x(t) - \bar{\x}(t)) + B(\u(t) - \bar{\u}(t)) \\
&= A\delta\x(t) + B\delta\u(t),
\end{align*}
which have the same linear structure as the original system, only now expressed in terms of the deviation variables.
This observation is central, as it allows the direct application of LQR techniques to the tracking problem.

\paragraph{LQR formulation for tracking.}
Tracking performance can be naturally expressed through a quadratic cost on the deviation variables:
\begin{equation}
    J(\delta \u(t)) = \delta \x(t_f)^\top Q_f \delta \x(t_f) + \int_{0}^{t_f} \delta \x(t)^\top Q \delta \x(t) + \delta \u(t)^\top R \delta \u(t) \, \d t,
\end{equation}
where $Q \succeq 0$, $R \succ 0$, and $Q_f \succeq 0$ are weighting matrices that penalize deviations of the state and control from their nominal values.
The tracking problem can thus be expressed as the finite-horizon optimal control problem:
\begin{equation}
\begin{split}
    \minimize[\delta \u] & \delta \x(t_f)^\top Q_f \delta \x(t_f) + \int_{0}^{t_f} \delta \x(t)^\top Q \delta \x(t) + \delta \u(t)^\top R \delta \u(t) \, \d t, \\
    \subjectto & \delta \dot{\x}(t) = A \delta \x(t) + B \delta \u(t),
\end{split}
\end{equation}
which is precisely the finite-horizon LQR problem introduced in~\cref{subsubsec:lqr}, but now formulated in terms of the deviation variables.
Therefore, all the results derived for finite-horizon LQR apply directly to this tracking formulation, and the optimal control law takes the form:

\begin{equation}
    \delta \u(t) = -K(t) \delta \x(t),
\end{equation}
where $K(t)$ is the time-varying optimal feedback gain obtained by solving the differential Riccati equation from~\cref{eq:finite_horizon_gain}.

Finally, substituting the definitions of the deviation variables from~\cref{eq:error_variables} yields the optimal control law in terms of the original state and control variables:
\begin{equation}
    \u(t) = \bar \u(t) - K(t) \bigl(\x(t) - \bar \x(t)\bigr).
\end{equation}
This expression is a clear instantiation of the two-step design paradigm, where the control input consists of a feedforward term $\bar \u(t)$, derived from the nominal trajectory, and a feedback term $-K(t) (\x(t) - \bar \x(t))$ that actively compensates for any deviations of the actual state from the nominal trajectory.

\subsection{Optimal Closed-loop Control for Nonlinear Systems}
\label{subsec:nonlinear_closed_loop}
The previous section focused on linear systems, for which the LQR framework provides elegant closed-form solutions.
However, many real-world systems are inherently nonlinear, making the extension of optimal closed-loop control to nonlinear dynamics crucial for numerous practical applications.

In this section, we first briefly contextualize dynamic programming and the HJB equation---two foundational methods for deriving globally optimal closed-loop control laws in~\cref{subsubsec:dp_hjb}, referring the reader to~\cref{ch:reinforcement-learning} and~\citet{Bertsekas2000} for a more in-depth discussion.
We then turn our attention to approaches that aim to leverage the structure and insights from linear optimal control to solve nonlinear optimal control problems.
For trajectory-tracking, we begin by showing how the notion of \emph{linearization} enables the application of LQR tracking techniques to nonlinear systems in~\cref{subsubsec:nonlinear_lqr}.
Finally, we extend ideas from linearization and LQR to develop algorithms that \emph{simultaneously} generate an open-loop trajectory and closed-loop tracking controller in two-step design form.
Specifically, we present iLQR and DDP in~\cref{subsubsec:ilqr_ddp}.

\subsubsection{Dynamic Programming and the Hamilton–Jacobi–Bellman Equation}
\label{subsubsec:dp_hjb}
When deriving optimal closed-loop control policies for nonlinear systems, one typically distinguishes between two complementary formulations: discrete-time and continuous-time.

In the discrete-time setting, the primary tool is dynamic programming, originally developed by Richard Bellman in the 1950s. 
Dynamic programming provides a systematic framework for solving optimal control problems with additive cost functions and nonlinear dynamics, and it underlies numerous classical and modern optimal control algorithms, including many learning-based approaches\sidenote{For instance, reinforcement learning algorithms discussed in~\cref{ch:reinforcement-learning}.}.
Intuitively, dynamic programming decomposes the optimization problem into a sequence of smaller subproblems that can be solved recursively by exploiting the \emph{principle of optimality}, which informally states that for a sequence of optimal decisions, the \emph{tail} of the optimal sequence is also optimal for a \emph{tail subproblem}.
This recursive structure leads to the celebrated \emph{Bellman equation}, which dramatically simplifies the search for optimal policies by transforming the global optimization problem into a tractable sequence of smaller optimization problems.

In the continuous-time setting, the analogous formulation is given by the HJB equation. 
The HJB equation is a nonlinear partial differential equation whose solution enables the derivation of an optimal control law.
Conceptually, it captures the infinitesimal version of the dynamic programming principle, providing a continuous-time characterization of optimality.

Both the discrete-time dynamic programming paradigm and the continuous-time HJB equation offer powerful, theoretically grounded approaches for deriving globally optimal closed-loop policies for nonlinear systems. 
However, they also suffer from severe computational challenges, such as the \emph{curse of dimensionality}, which limit their direct applicability to high-dimensional systems. 

Practical methods inspired by dynamic programming principles will be explored further in~\cref{ch:reinforcement-learning}, while for an in-depth discussion of the HJB framework, we refer the reader to~\citet{Bertsekas2000}.

\subsubsection{Linear Methods for Nonlinear Tracking Control}
\label{subsubsec:nonlinear_lqr}
A common strategy for controlling nonlinear systems is to approximate their dynamics locally by a linear system and then apply linear control techniques, such as LQR tracking from~\cref{subsubsec:linear_tracking}.
The process of approximating a nonlinear system by a linear one is called \emph{linearization} and it leverages the fact that any smooth nonlinear function can be approximated locally by its first-order Taylor expansion.
Below, we first describe the linearization process and then show how to apply LQR tracking to the resulting linearized system.

\paragraph{Linearization of nonlinear systems.}
Consider the nonlinear system dynamics:
\begin{equation}
    \dot{\x}(t) = f(\x(t), \u(t)),
    \label{eq:nonlinear_dynamics_actual}
\end{equation}
where $f: \R^{\statedim} \times \R^{\controldim} \to \R^{\statedim}$ is a smooth nonlinear function that governs the system's evolution.
Let $\left(\bar \x(t), \bar \u(t)\right)$ denote a nominal state and control trajectory that satisfies the dynamics, that is:
\begin{equation}
    \dot{\bar \x}(t) = f(\bar \x(t), \bar \u(t)).
    \label{eq:nonlinear_dynamics_reference}
\end{equation}
For the purposes of trajectory-tracking, we are interested in controlling the system so that it remains close to this nominal trajectory, i.e., such that the state deviations $\delta \x(t) = \x(t) - \bar \x(t)$ are small.
As in the linear case (see~\cref{subsubsec:linear_tracking}), let us first derive the dynamics of these deviation variables and then design a controller to stabilize them around zero.

Using~\cref{eq:nonlinear_dynamics_actual,eq:nonlinear_dynamics_reference}, the deviation dynamics can be written as:
\begin{equation}
    \delta \dot{\x}(t) = \dot{\x}(t) - \dot{\bar \x}(t) = f(\x(t), \u(t)) - f(\bar \x(t), \bar \u(t)),
    \label{eq:delta_dynamics}
\end{equation}
where, to approximate the right-hand side, we can linearize the nonlinear function $f$ around the nominal trajectory.
Specifically, assuming that the deviations  $\delta\x(t)$ and $\delta\u(t) = \u(t) - \bar{\u}(t)$ are small, a first-order Taylor expansion of $f$ yields:
\begin{equation}
    \dynmodel(\x(t),\u(t)) \approx f(\bar{\x}(t),\bar{\u}(t)) + \underbrace{\frac{\partial f}{\partial \x}(\bar{\x}(t),\bar{\u}(t))}_{\coloneqq A(t)}\delta \x(t)+\underbrace{\frac{\partial f}{\partial \u}(\bar{\x}(t),\bar{\u}(t))}_{\coloneqq B(t)}\delta \u(t),
\label{eq:taylor_expansion_linearization}
\end{equation}
where $A(t) \in \R^{\statedim \times \statedim}$ and $B(t) \in \R^{\statedim \times \controldim}$ are the Jacobians of $f$ with respect to the state and control input, evaluated along the nominal trajectory, that is, at $(\bar \x(t), \bar \u(t))$.

Substituting the approximation from~\cref{eq:taylor_expansion_linearization} into~\cref{eq:delta_dynamics}, we obtain:
\begin{equation}
\begin{split}
    \delta\dot{\x}(t)
    &\approx
    \bigl[f(\bar{\x}(t),\bar{\u}(t))
    + A(t)\,\delta\x(t)
    + B(t)\,\delta\u(t)\bigr]
    - f(\bar{\x}(t),\bar{\u}(t)),
\end{split}
\end{equation}
which yields the linearized dynamics for the deviation variables:
\begin{equation}
    \delta \dot{\x}(t) \approx A(t) \delta \x(t) + B(t) \delta \u(t).
    \label{eq:linearized_dynamics}
\end{equation}

\paragraph{Applying LQR tracking to the linearized system.}
Once in this form, standard linear control techniques, such as the LQR tracking framework, can be directly applied to design a feedback controller that stabilizes the deviation dynamics around zero.
For instance, applying the finite-horizon LQR tracking framework from~\cref{subsubsec:linear_tracking} yields the optimal feedback law:
$$
\delta \u(t) = -K(t)\, \delta \x(t).
$$
which, expressed in the original variables, gives the closed-loop controller:
$$
\u(t) = \bar{\u}(t) - K(t) \bigl( \x(t) - \bar{\x}(t) \bigr),
$$
where $K(t)$ is the time-varying feedback gain computed by solving the differential Riccati equation associated with the linearized system.

\paragraph{Linearizing around an equilibrium point.}
A particularly important special case arises when the nominal trajectory corresponds to a constant equilibrium $(\bar \x, \bar \u)$.
A state $\bar \x$ is referred to as an equilibrium if there exists a control input $\bar \u$ (called the equilibrium input) such that $f(\bar \x, \bar \u) = 0$.
In other words, if the system is initialized at $\bar \x$ and the constant input $\bar \u$ is applied for all $t \geq t_0$, the state will remain fixed at $\bar \x$ indefinitely.
Linearizing about $(\bar \x, \bar \u)$ produces a system of the form in~\cref{eq:linearized_dynamics}, where $A$ and $B$ would be constant matrices (not time-varying) evaluated at the equilibrium.

\paragraph{Local validity of linearization.}
It is important to highlight that linearization provides only a local approximation, and the linearized model accurately captures the system dynamics only in a neighborhood of the nominal trajectory. 
If the state deviates significantly from $\bar{\x}(t)$, the linear approximation may no longer hold, and the resulting controller may perform poorly or even destabilize the system.

\medskip
\noindent Below, we illustrate the linearization process with a concrete example.
\begin{example}[Inverted pendulum]
    \label{ex:inverted_pendulum}
    \begin{marginfigure}
        \centering 
        \includegraphics[width=0.85\linewidth]{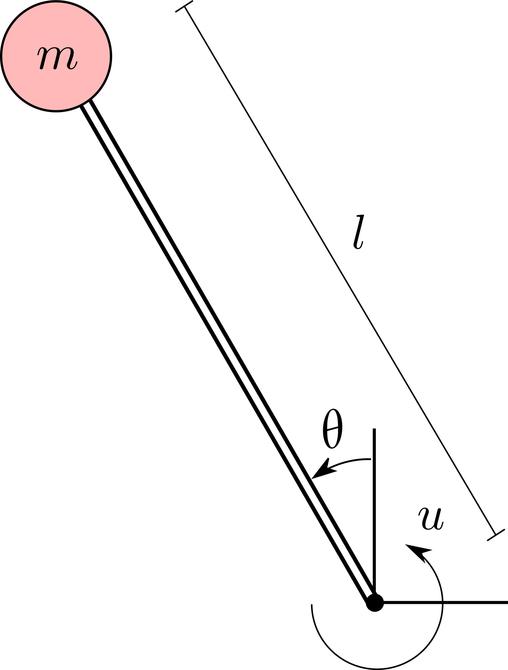}
        \caption{An inverted pendulum consisting of a point mass $m$ attached to a rigid rod of length $l$. 
        The motion is described by the angle $\theta$ from the upright vertical, and $u$ denotes the control torque applied about the pivot.} 
        \label{fig:inverted_pendulum} 
    \end{marginfigure} 
    Consider the inverted pendulum shown in \cref{fig:inverted_pendulum}.
    Its nonlinear dynamics are given by:
    \begin{equation*}
        ml^2\ddot{\theta} = mgl\sin(\theta) + u,
    \end{equation*}
    where $m$ denotes the mass, $l$ is the length of the rod, $g$ is the acceleration due to gravity, $\theta$ is the pendulum angle, and $u$ is the control torque.
    Introducing the state vector $\x \coloneqq \vCol{\theta \ \dot{\theta}}$, the system can be written in state-space form as:
    \begin{equation*}
        \dot{\x} = f(\x,u) = \begin{bmatrix} \dot{\theta}\\ 
        \frac{g}{l} \sin(\theta)+\frac{1}{ml^2} u
        \end{bmatrix}.
    \end{equation*}
    The upright stationary position $\bar x=\vCol{0,0}$ is an equilibrium point for this system with equilibrium control input $\bar u = 0$.
    Linearizing about this equilibrium---one where $\delta \x = \x - \bar{\x} = \x$ and $\delta u = u - \bar{u} = u$---yields the linear model:
    \begin{equation*}
        \delta \dot{\x} = \begin{bmatrix} 0&1\\ \frac{g}{l} &0\end{bmatrix}\delta \x + \begin{bmatrix} 0\\ \frac{1}{ml^2}\end{bmatrix}\delta u,
    \end{equation*}
    whose state matrix has one eigenvalue with positive real part, and is therefore unstable.

    To stabilize the pendulum, we can design a proportional–derivative controller of the form:
    \begin{equation*}
    \delta u(t)=-k_p \theta(t) - k_d\dot{\theta}(t),
    \end{equation*}
    which would yield the closed-loop dynamics:
    \begin{equation*}
        \delta \dot{\x} = \begin{bmatrix} 0&1\\ \frac{g}{l} - \frac{1}{ml^2}k_p & -\frac{1}{ml^2}k_d
        \end{bmatrix}\delta \x.
    \end{equation*}
    By selecting suitable gains $k_p > 0$ and $k_d > 0$, we can ensure that both eigenvalues of the closed-loop system have negative real parts, ensuring stability around the upright position---analogously to the double-integrator PD design in \cref{ex:pd_double_int}.
    Of course, PD control is only one option, and other methods---such as the LQR---can also be applied to the linearized model.
\end{example}

Algorithm~\ref{alg:inv_pend_jax} shows a Python implementation of the linearization process for the inverted pendulum example using JAX, a library for high-performance numerical computing and automatic differentiation.
\begin{listing}[ht!]
\begin{tcolorbox}[colback=gray!10, colframe=gray!50, 
    title=Inverted Pendulum Dynamics Linearization, boxrule=0.5mm, arc=0mm]
\begin{minted}[escapeinside=||]{python}
import jax
import jax.numpy as jnp

def inverted_pendulum_dynamics(x, u, g=9.81, m=1, l=1):
    """
    Evaluate the inverted pendulum dynamics.
    """
    |$\theta$|, d|$\theta$|_dt = x
    dx_dt = jnp.array([d|$\theta$|_dt, (g/l)*jnp.sin(|$\theta$|) + (1/m*l**2)*u])
    return dx_dt

# Linearize around the stationary upright position with zero 
# control (i.e. the pendulum is perfectly balanced)
f_jac = jax.jacobian(inverted_pendulum_dynamics, argnums=(0, 1))
x = jnp.array([0., 0.])
u = 0.                
A, B = f_jac(x, u) # Evaluate Jacobian at equilibrium point
\end{minted}
\end{tcolorbox}
\caption{Linearizing the inverted pendulum dynamics from \cref{ex:inverted_pendulum} in Python using the JAX library.
The code for this example is available in the repository \colorcode{github.com/StanfordASL/pora-exercises} in the notebook \colorcode{ch03/jax\_linearization.ipynb}.}
\label{alg:inv_pend_jax}
\end{listing}

\subsubsection{Iterative LQR (iLQR) and Differential Dynamic Programming (DDP)}
\label{subsubsec:ilqr_ddp}
In the previous sections, we saw how the LQR framework---along with its various extensions---provides a principled foundation for regulation and tracking, and how the same ideas can be applied to both linear and nonlinear systems.
Specifically, we highlighted how LQR can be naturally embedded within a two-step design paradigm, serving as a practical alternative to directly solving the full nonlinear optimal control problem by combining elements of open-loop trajectory generation and closed-loop tracking control.

In this section, we adopt a different yet complementary perspective, and explore how ideas from linearization and LQR can be extended to develop algorithms that \emph{simultaneously} generate an open-loop trajectory \emph{and} a closed-loop tracking controller in two-step design form.
This approach leads to two closely related algorithms: the \emph{iterative linear quadratic regulator (iLQR)} and \emph{differential dynamic programming (DDP)}.
As we will see, the key insight underlying both methods is that the structure which makes LQR tractable---namely, quadratic cost and linear dynamics---can be embedded within an iterative optimization scheme to efficiently handle nonlinear dynamics and non-quadratic costs.

\paragraph{Iterative LQR.}
Recall from the LQR tracking problem in~\cref{subsubsec:linear_tracking} that the controller acts to drive deviation variables $(\delta \x, \delta \u)$ to zero, thereby ensuring the system follows a specified reference.
At its core, iLQR uses the same machinery, but with a different objective: instead of driving the deviation variables to zero, the optimal deviations are used to \emph{modify the nominal trajectory itself}.
By repeating this process, iLQR gradually refines the trajectory until no further improvement can be made, yielding a locally optimal solution.

Formally, consider the discrete-time finite-horizon nonlinear optimal control problem:
\begin{equation}
    \begin{split}
    \minimize[\u] & h(\x_N) + \sum_{t=0}^{N-1} g(\x_t, \u_t), \\
    \subjectto & \x_{t+1} = f(\x_t, \u_t), \quad t = 0, \ldots, N-1, \\
    \end{split}
\end{equation}
Given a feasible nominal trajectory $(\bar x_0, \bar u_0, \ldots, \bar x_N, \bar u_N)$, let us linearize the dynamics and quadratize the cost around the nominal trajectory as:
\begin{equation}
\begin{aligned}
    \delta \x_{t+1} \approx & \underbrace{\frac{\partial f}{\partial x}\left(\bar{x}_t, \bar{u}_t\right)}_{\coloneqq A_t} \delta \x_t + \underbrace{\frac{\partial f}{\partial u}\left(\bar{x}_t, \bar{u}_t\right)}_{\coloneqq B_t} \delta \u_t+\underbrace{0}_{\coloneqq c_t}, \\
    h(x_N) \approx & \underbrace{h\left(\bar{x}_N\right)}_{\coloneqq\alpha_f}+\underbrace{\nabla h\left(\bar{x}_N\right)}_{\coloneqq q_f} \delta \x_N + \frac{1}{2} \delta \x_N \underbrace{\nabla^2 h\left(\bar{x}_N\right)}_{\coloneqq Q_f} \delta \x_N, \\
    g_t(x_t, u_t) \approx & \underbrace{g_t\left(\bar{x}_t, \bar{u}_t\right)}_{\coloneqq \alpha_t}+\underbrace{\nabla_x g\left(\bar{x}_t, \bar{u}_t\right)}_{\coloneqq q_t} \delta \x_t+\underbrace{\nabla_u g\left(\bar{x}_t, \bar{u}_t\right)}_{\coloneqq r_t} \delta \u_t \\
    & +\frac{1}{2} \delta \x_t^{\top} \underbrace{\nabla_{x x}^2 g_t\left(\bar{x}_t, \bar{u}_t\right)}_{\coloneqq Q_t} \delta \x_t + \frac{1}{2} \delta \u_t^{\top} \underbrace{\nabla_{u u}^2 g_t\left(\bar{x}_t, \bar{u}_t\right)}_{\coloneqq R_t} \delta \u_t \\
    & + \delta \x_t^{\top} \underbrace{\nabla_{x u}^2 g_t\left(\bar{x}_t, \bar{u}_t\right)}_{\coloneqq S_t} \delta \u_t,
\end{aligned}
\end{equation}
where $Q,R,S$ are Hessians and $q,r$ the gradients of the cost with respect to state and input, evaluated along $(\bar \x, \bar \u)$.
This reduces the problem to the LQR formulation introduced in~\cref{eq:lqr_general}, which can be solved via the Riccati equations to obtain the optimal deviations $(\delta \x^*, \delta \u^*)$.

\smallskip
\noindent iLQR alternates between two complementary steps:
\begin{itemize}
    \item \emph{Backward pass:} linearize the dynamics and quadratize the cost around $(\bar \x, \bar \u)$.
    Solve the resulting LQR problem to obtain the affine control law for the deviation variables:
    \begin{equation}
        \delta \u_t^* = -K_t \delta \x_t - k_t.
    \end{equation}
    
    \item \emph{Forward pass:} starting from the initial condition $x_0$, propagate the nonlinear system forward using:
    \begin{equation}
        \u_t = \bar \u_t - k_t - K_t(\x_t - \bar \x_t),
    \end{equation}
    and update the nominal trajectory to $(\bar \x, \bar \u) = (\x, \u)$.
\end{itemize}
These steps are repeated until the trajectory converges, typically measured by negligible improvement in cost or small changes in the control sequence.

\paragraph{Practical considerations.}
While iLQR provides a powerful framework for trajectory optimization, its performance depends critically on several implementation details:
\begin{itemize}
    \item \textit{Local optimality.} iLQR produces solutions that are only \emph{locally optimal}.
    The resulting open-loop trajectory $(\bar \x, \bar \u)$ minimizes the cost only in a neighborhood of the initial guess, and the associated feedback law stabilizes the system only locally.
    Consequently, a good initialization of the nominal trajectory is often critical to success.

    \item \textit{Second-order terms.} The quadratic expansion of the cost introduces second-order terms $H_{xx, t}$ and $H_{uu, t}$, which in general may not be positive semidefinite and positive definite, respectively.
    To ensure well-posed Riccati recursions, these terms are often regularized to be invertible, for example by adding a multiple of the identity matrix $(H_{xx,t} + \mu I$, $H_{uu, t} + \mu I)$, with $\mu > 0$, or by projecting them to the nearest valid matrices.

    \item \textit{Termination criteria.} Since iLQR is iterative, a stopping rule must be defined.
    In practice, iterations are terminated either when the change in the control trajectory is sufficiently small, or when the improvement in cost between successive iterations falls below a chosen threshold.

    \item \textit{Forward pass robustness.} During the forward pass, care must be taken to ensure that the updated trajectory does not deviate too far from the one used in the linearization. 
    Common strategies include penalizing large deviations more heavily or performing a line search on the step size used to update the controls and states.
\end{itemize}
These considerations are critical for achieving robust and efficient performance in practice. 
A comprehensive collection of tips and detailed mathematical treatment of iLQR can be found in~\citet{Tassa2011}.

\paragraph{Differential dynamic programming.}
Closely related to iLQR is DDP, an algorithm rooted in both LQR and dynamic programming\sidenote{Introduced in~\cref{subsubsec:dp_hjb} and which will be further discussed in~\cref{ch:sequential-decision-making}.}.
Like iLQR, DDP alternates between a backward pass (computing locally optimal gains) and a forward rollout (updating the trajectory).
The key difference lies in the treatment of the backward pass.
In iLQR, the dynamics are linearized to first order, so the resulting formulation involves second-order terms only from the cost function.
In contrast, DDP also expands the dynamics to second order, so that curvature information from the nonlinear dynamics explicitly enters the Riccati recursion.

Formally, compared to the solution introduced in~\cref{eq:lqr_general_matrices}, DDP augments the matrices $H_{xx, t}$, $H_{uu, t}$, and $H_{xu, t}$ with additional second-order terms, as follows:
\begin{equation}
    \begin{aligned}
    H_{xx, t} &\coloneqq Q_t + A_t^\top P_{t+1} A_t + \sum_{i=1}^n p_{t+1, i}\, \nabla^2_{xx} f_i(\bar \x_t, \bar \u_t), \\[2pt]
    H_{uu, t} &\coloneqq R_t + B_t^\top P_{t+1} B_t + \sum_{i=1}^n p_{t+1, i}\, \nabla^2_{uu} f_i(\bar \x_t, \bar \u_t), \\[2pt]
    H_{xu, t} &\coloneqq S_t + A_t^\top P_{t+1} B_t + \sum_{i=1}^n p_{t+1, i}\, \nabla^2_{xu} f_i(\bar \x_t, \bar \u_t).
    \end{aligned}
\end{equation}
In practice, this richer approximation often improves convergence behavior and solution accuracy, especially in strongly nonlinear systems.
However, it also makes DDP more computationally expensive than iLQR, since it requires evaluating and storing second-order derivatives of the dynamics at every iteration.

\subsection{Model Predictive Control (MPC)}
\label{subsec:mpc}
We now turn to Model Predictive Control (MPC), also known as \emph{Receding Horizon Control} (RHC), which has become one of the most influential methodologies in modern control, both in theory and in practice.
To appreciate its role, it is helpful to revisit the distinction between open-loop and closed-loop control.

In an open-loop formulation, the control problem is posed as finding a trajectory $\u(t)$ that optimizes a given performance criterion subject to system constraints.
While conceptually straightforward and computationally cheaper, this approach does not account for disturbances or deviations from the nominal trajectory that may arise during execution.

By contrast, closed-loop control directly maps the measured state to a control action through a feedback policy.
This design enables the control input to adapt to the current system state.
However, computing optimal feedback policies can quickly become intractable, even for systems of moderate size\sidenote{Earlier, we introduced PID controllers and LQR as computationally efficient examples of closed-loop control. These methods, however, are restricted to relatively simple settings such as linear systems with quadratic costs. More general cases often quickly become computationally prohibitive.}.

\medskip
At a high level, MPC can be seen as a principled unification of these two paradigms, using repeated open-loop optimization to effectively achieve closed-loop behavior.
At each time step, MPC solves a finite-horizon open-loop optimal control problem, yielding an input sequence $u_0, u_1, \ldots, u_{N-1}$. 
Only the first control action $u_0$ is applied, after which the horizon shifts forward, the state is re-measured, and the optimization is solved again.
This receding-horizon strategy effectively embeds feedback into open-loop optimization, combining the robustness of closed-loop adaptation with the computational tractability of open-loop planning.

Naturally, MPC requires additional design elements to ensure reliable operation. 
Two key concerns are \emph{persistent feasibility} (the guarantee that a valid solution can be found at each step) and \emph{stability} (ensuring the closed-loop system behaves well over time). 
These are typically addressed through terminal constraints, tailored terminal costs, or other problem-specific design choices, which we will return to later in this section.

\medskip
Historically, MPC emerged in the 1970s within the process control community, particularly in chemical engineering. 
This application domain was ideal for two reasons. 
First, safety-critical constraints (e.g., temperature limits) could be enforced explicitly within the optimization problem. 
Second, chemical processes evolve on relatively slow timescales, giving ample time to solve optimization problems online---even with the limited computational power of the era. 
With today’s hardware, these limitations have largely disappeared, and MPC is now deployed in real-time domains such as robotics, where numerical optimization problems may be solved at tens or hundreds of Hertz.

Today, MPC is considered one of the cornerstones of modern control.
Alongside PID control and LQR, it is among the most widely applied methodologies, valued for its ability to explicitly handle constraints while maintaining robustness.
The material in this section provides the foundations necessary to understand and implement MPC.
For readers interested in deeper coverage of both theory and practice, we recommend the textbooks by \citet{BorrelliEtAl2017} and \citet{RawlingsEtAl2017}.

\subsubsection{The Receding Horizon Framework}
\begin{figure}[t!]
    \centering 
    \includegraphics[width=0.9\linewidth]{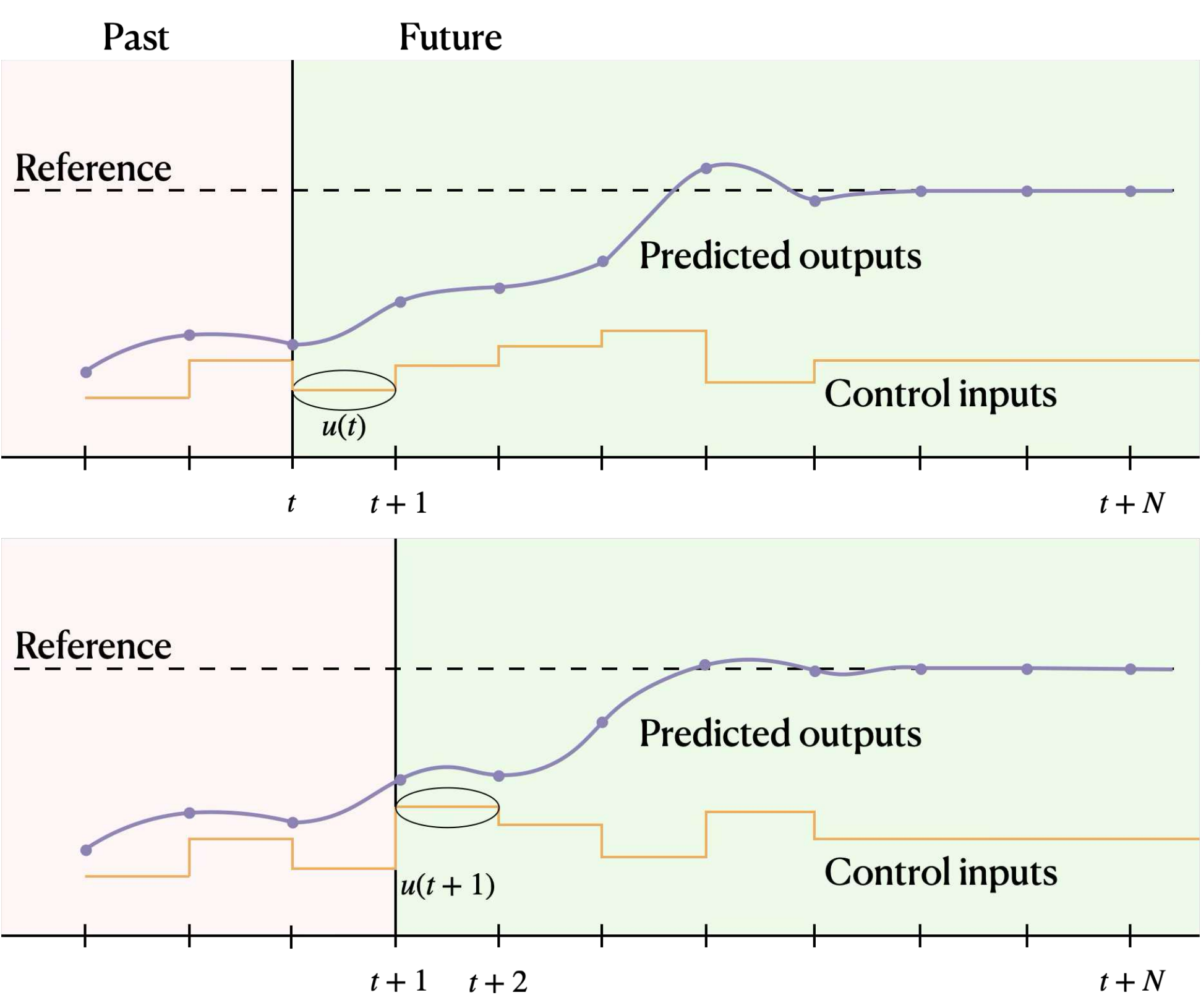}
    \caption{The receding horizon principle. At each time step, a finite-horizon optimal control problem is solved based on the current state measurement. Only the first control input is applied, and the process repeats at the next time step.}
    \label{fig:rhc} 
\end{figure}
The underlying setting for MPC is the discrete-time infinite-horizon optimal control problem. 
That is, in principle, we would like to design a feedback policy that minimizes a cost accumulated over an infinite time horizon, that is, with final time $t_f = +\infty$. 
Solving for closed-loop policies in this setting is, however, typically intractable.

MPC addresses this challenge through a suboptimal---but computationally tractable---approximation. 
Essentially, instead of solving the infinite-horizon problem directly, it repeatedly solves a finite-horizon optimal control problem. 
The idea is simple and illustrated in~\cref{fig:rhc}.
At the current time $t$, the system state is measured, and an optimal control sequence is computed over a horizon of length $N$. 
As in any open-loop formulation, the optimization relies on the system model to predict how the state will evolve under candidate input sequences, selecting the one that minimizes the cost over the finite horizon. 
This process yields a sequence of control inputs:
$$
u_t, u_{t+1}, \ldots, u_{t+N-1},
$$
along with the corresponding predicted state trajectory. 
Crucially, only the \emph{first} control input $u_t$ is applied to the system. 
Once time advances to $t+1$, the remaining inputs $u_{t+1}, \ldots, u_{t+N-1}$ are discarded, and a new optimization problem is solved based on the updated state measurement. 
This process is then repeated recursively.

This scheme is often referred to as the receding horizon framework, where the horizon “moves forward” with time, and at each step a new optimization problem is solved based on the latest state information.
While it may appear wasteful to discard the unused portion of the control trajectory, this is precisely what allows MPC to incorporate updated measurements and disturbances, ensuring feedback is embedded into the process.

If the finite-horizon problems are designed and tuned carefully---through appropriate horizon lengths, terminal costs, or constraints---the closed-loop behavior of MPC can closely approximate that of the true infinite-horizon optimal controller. 
In practice, MPC often achieves performance nearly indistinguishable from the infinite-horizon optimal solution, while remaining computationally feasible.

\subsubsection{Basic Formulation}
To formalize the ideas introduced above, consider the discrete-time linear time-invariant (LTI) system:
\begin{equation}
    \x_{t+1} = A\x_t + B\u_t,
\label{eq:mpc_system}
\end{equation}
where $\x_t \in \R^n$ and $\u_t \in \R^m$ denote the state and input vectors, respectively.
Both are subject to constraints of the form:
\begin{equation}
    \x_t \in \X,
    \quad
    \u_t \in \U,
    \quad \forall t \geq 0,
\label{eq:mpc_constraints_form}
\end{equation}
where $\X \subseteq \R^n$ and $\U \subseteq \R^m$ are convex polyhedral sets describing admissible states and inputs.

At time $t$, given the measured state $\x_t$, MPC computes an input sequence over a prediction horizon of length $N$:
$$
\u_{t|t}, \, \u_{t+1|t}, \, \dots, \, \u_{t+N-1|t},
$$
where the notation $\u_{t+k|t}$ indicates “the input applied at time $t+k$ as predicted at time $t$”.
Correspondingly, we denote the predicted state trajectory as:
$$
\x_{t|t}, \, \x_{t+1|t}, \, \dots, \, \x_{t+N|t},
\qquad \text{with } \x_{t|t} = \x_t.
$$

\paragraph{Finite-horizon problem.}
The control inputs are obtained by solving the finite-horizon optimal control problem:
\begin{equation}
\begin{aligned}
    J_t^*(\x_t) = \minimize[\U_{t \to t+N|t}] &
    h(\x_{t+N|t}) + \sum_{k=0}^{N-1} g(\x_{t+k|t}, \u_{t+k|t}), \\
    \subjectto & \x_{t+k+1|t} = A\x_{t+k|t} + B\u_{t+k|t}, \quad k=0,\dots,N-1, \\
    & \x_{t+k|t} \in \X, \ \u_{t+k|t} \in \U, \quad k=0,\dots,N-1, \\
    & \x_{t+N|t} \in \X_f, \quad \x_{t|t} = \x_t,
\end{aligned}
\label{eq:mpc_ocp}
\end{equation}
where $\U_{t \to t+N|t}$ denotes the sequence of decision variables ${\u_{t|t},\dots,\u_{t+N-1|t}}$, $\X_f$ is a terminal set, and the functions $g(\cdot,\cdot)$ and $h(\cdot)$ represent stage and terminal costs.

\paragraph{Control law.}
Only the first element of the optimal input sequence is applied to the system:
\begin{equation}
    \u_t = \u_{t|t}^*(\x_t),
\label{eq:mpc_feedback_law}
\end{equation}
where the dependency on $\x_t$ arises from the fact that the optimization problem in~\cref{eq:mpc_ocp} has the condition $\x_{t|t} = \x_t$.
At the next step, $t+1$, the state is re-measured, and Problem~\eqref{eq:mpc_ocp} is solved again with updated information.
This iterative procedure defines the receding horizon control law, embedding feedback into the overall process.

\paragraph{Closed-loop dynamics.}
The resulting closed-loop system can be written compactly as:
\begin{equation}
    \x_{t+1} = A\x_t + B \pi_t(\x_t) = f_{\mathrm{cl}}(\x_t),
    \qquad t \geq 0,
\label{eq:mpc_closed_loop_dyn}
\end{equation}
where $\pi_t(\x_t) \coloneqq \u_{t|t}^*(\x_t)$ is the MPC feedback law at time $t$.
Because the dynamics, cost, and constraints are time-invariant, the problem can be equivalently written by fixing the initial time $t=0$.
Formally, let $\x_0 = \x_t$ and $\U_0 = {\u_0,\dots,\u_{N-1}}$, we can rewrite the problem in~\cref{eq:mpc_ocp} as:
\begin{equation}
\begin{aligned}
    J_0^*(\x_t) = \minimize[\U_0] &
    h(\x_N) + \sum_{k=0}^{N-1} g(\x_k,\u_k), \\
    \subjectto & \x_{k+1} = A\x_k + B\u_k, \quad k=0,\dots,N-1, \\
    & \x_k \in \X, \ \u_k \in \U, \quad k=0,\dots,N-1, \\
    & \x_N \in \X_f, \quad \x_0 = \x_t.
\end{aligned}
\label{eq:mpc_time_invariant_ocp}
\end{equation}
This formulation is often more convenient for analysis, since it avoids carrying the explicit time index through the condition $\x_0 = \x_t$.

\paragraph{Typical cost functions.}
In practice, MPC problems are often defined by quadratic stage and terminal costs of the form:
\begin{equation}
h(\x_N) = \x_N^\top Q_f \x_N,
\qquad
g(\x_k,\u_k) = \x_k^\top Q \x_k + \u_k^\top R \u_k,
\label{eq:mpc_costs}
\end{equation}
with $Q_f \succeq 0$, $Q \succeq 0$, and $R \succ 0$.
This leads to the overall cost function:
\begin{equation}
J_0(\x_0) = \x_N^\top Q_f \x_N + \sum_{k=0}^{N-1} \big( \x_k^\top Q \x_k + \u_k^\top R \u_k \big).
\label{eq:mpc_cost_function}
\end{equation}
Quadratic costs are not only natural but also lead to convex optimization problems when paired with polyhedral constraints.
Alternative norms, such as $\ell_1$ or $\ell_\infty$, are sometimes used to promote robustness or sparsity in the control actions\sidenote{For instance, $\ell_1$ penalties are commonly used in predictive control formulations for systems with actuator limitations or to induce sparse actuation.}.

\paragraph{Implementation challenges.}
While MPC provides a powerful framework, it also raises two critical implementation challenges:
\begin{enumerate}
    \item \emph{Persistent feasibility.} Even if the optimization problem is feasible at the current time step, there is no guarantee that feasibility will be preserved in the future, as the chosen inputs may drive the system into a region of the state space from which no admissible solution exists.
    This situation arises when the controller fails to “look ahead” sufficiently to maintain long-term feasibility.
    For example, consider an autonomous car driving towards an obstacle at high speed, and the MPC controller chooses to brake too late because of a short-sighted horizon. As a result, the car may end up in a state where it cannot stop in time to avoid a collision, leading to infeasibility in subsequent control steps.
    \item \emph{Closed-loop stability.} Feasibility alone does not guarantee that the closed-loop trajectories converge to the origin. 
    In fact, it is possible for the MPC law to produce admissible control inputs that indefinitely satisfy the constraints but fail to stabilize the system.
\end{enumerate}
These issues highlight a fundamental tension: MPC is based on solving short-horizon problems, yet we ultimately require effective long-term behavior. 
How can a strategy that looks only $N$ steps ahead ensure both feasibility and stability over a potentially infinite horizon?

A central insight is that these properties can be enforced through appropriate design of the terminal cost and terminal constraint set---denoted as $h(\cdot)$ and $\X_f$ in~\eqref{eq:mpc_ocp}, respectively.
Roughly speaking, the terminal cost serves as a surrogate for the infinite-horizon tail of the problem, while the terminal set ensures that the system remains within a region where the behavior is well-understood and manageable.
Together, these design choices represent a flexible tool to reconcile the short-sighted nature of MPC with the long-term guarantees we ultimately care about.

While a full treatment of these issues is beyond the scope of this book, we refer the interested reader to~\citet{BorrelliEtAl2017} for comprehensive discussions and rigorous treatments of these topics.

\subsubsection{MPC for Tracking: Delta Input $(\delta \u)$ Formulation}
\label{subsec:mpc_tracking}
A particularly common use case for MPC is reference tracking, where the objective is to have the system state follow a reference trajectory $\bm{r}_0, \bm{r}_1, \ldots $ over time.
Here, $\bm{r}_t \in \R^p$ may denote the full state reference (in which case $p=n$) or a partial output reference (with $p < n$).
In this section, we introduce the $\delta\u$ formulation\cite{BorrelliEtAl2017} which is particularly advantageous in trajectory tracking and regulation tasks around nonzero operating points.

Formally, consider the following LTI system:
\begin{equation}
\begin{split}
    \x_{t+1} &= A\x_t + B\u_t, \\
    \u_t &= \u_{t-1} + \delta\u_t, \\
    \y_t &= C\x_t,
\end{split}
\label{eq:du_mpc_system}
\end{equation}
where $\delta \u_t \in \R^m$ represents the change in control input from the previous time step, $\y_t \in \R^p$ denotes the system output at time $t$, and $C \in \R^{p \times n}$ is a linear transformation mapping the state to the output.
Assuming the pair $(A,B)$ is controllable, it is often desirable to parameterize the optimization problem in terms of the \emph{control increments} rather than the absolute control inputs. 
Instead of optimizing the inputs $\u_t, \u_{t+1}, \ldots, \u_{t+N-1}$ directly, the controller optimizes their increments:
\begin{equation}
    \delta\u_t = \u_t - \u_{t-1},
    \label{eq:delta_u_def}
\end{equation}
and reconstructs the actual inputs recursively as:
\begin{equation}
    \u_t = \u_{t-1} + \delta\u_t,
    \label{eq:u_recursion}
\end{equation}
with the idea that, once the system has reached a steady state, the control inputs will remain constant (that is, $\delta \u_t = 0$).

As a result, the MPC problem can be readily reformulated as:
\begin{equation}
\begin{aligned}
    J_0^*(\x_t) = \minimize[\delta \u_0, \ldots, \delta \u_{N-1}] &
    \sum_{k=0}^{N-1} y_k^\top Q \bar{x}_k + \delta u_k^\top R \delta u_k, \\
    \subjectto & \x_{k+1} = A\x_k + B\u_k, \quad k=0,\dots,N-1, \\
    & y_k= C \x_k, \quad k=0,\dots,N-1, \\
    & \x_k \in \X, \quad \u_k \in \U, \quad k=0,\dots,N-1, \\
    & \x_N \in \X_f, \\
    & \u_k = \u_{k-1} + \delta\u_k, \quad k=0,\dots,N-1, \\
    & \x_0 = \x(t), \quad \u_{-1} = \u(t-1),
\end{aligned}
\label{eq:du_formulation}
\end{equation}
where $\x(t)$ and $\u(t-1)$ denote the current state and the last applied control input, respectively.

Ultimately, the $\delta\u$ formulation does not change the fundamental structure of MPC but reparameterizes the optimization problem in a way that is better conditioned for tracking tasks.

\subsubsection{Receding Horizon as a General Principle for Real-World Control}
\label{subsec:receding_horizon_general}
In practice, virtually all implementations of optimal control---whether framed as MPC, closed-loop form, or as a two-step design---operate according to the receding horizon framework. 
Even when a nominal trajectory is first computed offline and subsequently tracked by a feedback controller, the controller is typically reinitialized and re-optimized as new state information becomes available. 
In other words, the nominal trajectory serves only as a temporary reference, continuously adjusted as the system evolves and new measurements are obtained. 
This viewpoint helps reconcile the apparent distinction between MPC and trajectory-tracking control: both ultimately rely on repeated, finite-horizon optimization combined with feedback, differing mainly in how frequently and to what extent the optimization is repeated. 
Recognizing this continuity is important, as it clarifies that the so-called “receding horizon” behavior is not unique to MPC, but rather a practical necessity in virtually all control architectures that seek robustness to uncertainty and disturbances.

\section{Summary}
\label{sec:cl-ctrl-summary}
In this chapter, we introduced strategies for closed-loop control and trajectory tracking. 
Specifically, we discussed how the goal of closed-loop control is to compute a control policy that continuously adapts to the system’s evolving state. 
While generally more computationally demanding than open-loop methods, closed-loop control provides substantial advantages in terms of robustness to disturbances, model inaccuracies, and external perturbations.

We also introduced a tractable compromise through the concept of a \emph{two-step design}, in which an open-loop trajectory is first computed and then tracked using a feedback controller. 
This formulation allowed us to naturally transition to trajectory-tracking control, bridging open-loop optimization and closed-loop stabilization.

Building on these concepts, the chapter presented several key methods for closed-loop control and trajectory tracking. 
We began with classical feedback control, illustrated by the PID controller and its application to trajectory tracking, particularly for differentially flat systems.
We then introduced linear optimal control methods, focusing on the linear quadratic regulator, which provides an elegant closed-form solution for linear systems via the Riccati equation and extends naturally to tracking problems.

Next, we turned to nonlinear optimal control. 
We first contextualized techniques for obtaining globally optimal closed-loop solutions---namely dynamic programming and the HJB equation---and then showed how the notion of linearization enables the use of LQR for local stabilization and trajectory tracking in nonlinear systems. 
Building on this, we discussed how algorithms such as iLQR and DDP extend these principles to simultaneously generate an open-loop trajectory and a closed-loop tracking controller in two-step design form.

Finally, we presented model predictive control as a powerful framework that unifies open-loop optimization with closed-loop feedback through a receding-horizon strategy, offering a versatile method for handling constraints in real-time applications.

\paragraph{To learn more.}
For a deeper exploration of the topics covered in this chapter, several key resources are available.
An in-depth treatment of feedback control and PID controllers can be found in \citet{AstromMurray2009}.
For a rigorous treatment of optimal control from a dynamic programming perspective, which provides the theoretical underpinnings for the Riccati equation, \citet{Bertsekas2000} is an essential reference.
Finally, for comprehensive coverage of model predictive control, from fundamental principles to advanced theory and applications, we refer the reader to the textbooks by \citet{RawlingsEtAl2017} and \citet{BorrelliEtAl2017}.

\section{Exercises}
The starter code for the exercises provided below is available online through GitHub. 
To get started, download the code by running in a terminal window:

\begin{tcolorbox}[colback=gray!10]
\begin{minted}{bash}
    git clone https://github.com/StanfordASL/pora-exercises.git
\end{minted}
\end{tcolorbox}

We denote Problems requiring hand-written solutions and coding in Python with \adjustbox{height=2ex, valign=c}{\includegraphics{figs/write.png}} and \adjustbox{height=2ex, valign=c}{\includegraphics{figs/code.png}}, respectively.

\subsection*{\adjustbox{height=2ex, valign=c}{\includegraphics{figs/code.png}}\ Problem 1: Inverted Pendulum PD Control}
In the notebook \colorcode{ch03/exercises/pid\_control.ipynb}, implement the PD controller for the inverted pendulum dynamics.
Then, play around with different values of the gains $k_p$ and $k_d$ and use JAX and NumPy to compute the Jacobian of the closed-loop dynamics.
Verify the eigenvalues of the linearized matrix are stable.
Finally, run the provided code to simulate the nonlinear closed-loop dynamics.

\subsection*{\adjustbox{height=2ex, valign=c}{\includegraphics{figs/write.png}}\ \adjustbox{height=2ex, valign=c}{\includegraphics{figs/code.png}}\ Problem 2: Cart-pole LQR Control}
In this problem, we consider a cart-pole formulation in which we will design a controller to balance an inverted pendulum by linearizing the dynamics around a single stationary point.
Therefore, we will consider a trajectory initialized in a neighborhood about the final, upright state and the optimal control is a closed-form solution derived through Riccati recursion. This system has two degrees of freedom corresponding to the horizontal position $x$ of the cart, and the angle $\theta$ of the pendulum (where $\theta = 0$ occurs when the pendulum is hanging straight downwards). We can apply a force $u \in \R$ to push the cart horizontally, where $u > 0$ corresponds to a force in the positive $x$-direction. With the state $s \defn (x,\theta,\dot{x},\dot{\theta}) \in \R^4$, we can write the continuous-time dynamics of the cart-pole system as:
\begin{equation*}
    \dot{s} = f(s,u) = \bmx{
        \dot{x} \\
        \dot{\theta} \\
        \frac{
            m_p(\ell\dot{\theta}^2 + g\cos\theta)\sin\theta + u
        }{
            m_c + m_p\sin^2\theta
        } \\[0.5em]
        -\frac{
            (m_c+m_p)g\sin\theta + m_p\ell\dot{\theta}^2\sin\theta\cos\theta + u\cos\theta
        }{
            \ell(m_c + m_p\sin^2\theta)
        }
    },
\end{equation*}
where $m_p$ is the mass of the pendulum, $m_c$ is the mass of the cart, $\ell$ is the length of the pendulum, and $g$ is the acceleration due to gravity. We can discretize the continuous-time dynamics using Euler integration with a fixed time step $\Delta{t}$ to get the approximate discrete-time dynamics:
\begin{equation*}
    s_{k+1} \approx s_k + \Delta{t} f(s_k,u_k),
\end{equation*}
where~$s_k$ and~$u_k$ are the state and control input, respectively, at time $t = k\Delta{t}$.
The code for this exercise is located in \\\noindent\colorcode{ch03/exercises/cartpole\_lqr\_control.ipynb}.

\begin{enumerate}[]
    \item Consider the upright state $\bar{s} \defn (0,\pi,0,0)$ with $\bar{u} \defn 0$, and define $\tilde{s}_k \defn s_k - \bar{s}$. Linearizing the approximate discrete-time dynamics $s_{k+1} \approx s_k + \Delta{t} f(s_k,u_k)$ about $(\bar{s},\bar{u})$ yields an approximate LTI system of the form:
    \begin{equation*}
        \tilde{s}_{k+1} \approx A\tilde{s}_k + Bu_k.
    \end{equation*}
    Express~$A$ and~$B$ in terms of $m_p$, $m_c$, $\ell$, $g$, and $\Delta{t}$. You may use the fact that:
    \begin{equation*}
        \frac{\partial f}{\partial s}(\bar{s}, \bar{u}) = \bmx[1.25]{
            0 & 0 & 1 & 0 \\
            0 & 0 & 0 & 1 \\
            0 & \frac{m_p g}{m_c} & 0 & 0 \\
            0 & \frac{(m_c+m_p)g}{m_c\ell} & 0 & 0
        }, \quad
        \frac{\partial f}{\partial u}(\bar{s}, \bar{u}) = \bmx[1.25]{
            0 \\ 0 \\ \frac{1}{m_c} \\ \frac{1}{m_c\ell}
        }.
    \end{equation*}
\end{enumerate}

We will design a stabilizing LQR controller for this discrete-time LTI system to solve:
\begin{equation*}\begin{aligned}
    \minimize[u]
        ~& \sum_{k=0}^\infty \rbr*{
            \frac{1}{2}\tran{\tilde{s}_k}Q\tilde{s}_k
            + \frac{1}{2}\tran{u_k}Ru_k
        },
    \\
    \subjectto
        ~& \tilde{s}_{k+1} = A\tilde{s}_k + Bu_k,\
            \forall k \in \mathbb{N}_{\geq 0},
\end{aligned}
\end{equation*}
for fixed $Q,R \succ 0$. Recall that after $N$ iterations of the discrete-time Riccati recursion:
\begin{equation*}
\begin{aligned}
    K_k &= -\inv{(R + \tran{B}P_{k+1}B)}\tran{B}P_{k+1}A, \\
    P_k &= Q + \tran{A}P_{k+1}(A + B K_k),
\end{aligned}
\end{equation*}
the cost-to-go matrices $\{P_k\}_{k=0}^N$ and the time-varying feedback gains $\{K_k\}_{k=0}^{N-1}$ describe the optimal LQR controller for a finite-horizon version of the problem above. If $(A,B)$ is stabilizable, then these iterates asymptotically converge to some $P_\infty \succ 0$ and $K_\infty$. In fact, $J^*(s_0) \defn \tran{(s_0-\bar{s})}P_\infty{(s_0-\bar{s})}$ is the \emph{infinite-horizon} optimal cost-to-go for any initialization $s_0$, and $u_k = K_\infty \tilde{s}_k$ is the optimal feedback policy, which happens to be linear and \emph{time-invariant}\footnote{
    The infinite-horizon LQR problem also converges for fixed $Q \succeq 0$ and $R \succ 0$, as long as $(A,B)$ is stabilizable and $(A,Q)$ is detectable.
}.
\begin{enumerate}[resume]
    \item Write code to approximate $P_\infty$ and $K_\infty$ for the linearized, discretized cart-pole system by initializing $P_\infty = 0$ and then applying the Riccati recursion until convergence with respect to the maximum element-wise norm condition $\norm{P_k - P_{k-1}}_\mathrm{max} < 10^{-4}$. Use $m_p = 2~\mathrm{kg}$, $m_c = 10~\mathrm{kg}$, $\ell = 1~\mathrm{m}$, $g = 9.81~\mathrm{m/s}^2$, $\Delta{t} = 0.1~\mathrm{s}$, $Q = I_4$, and $R = I_1$. Report the value of $K_\infty$ with two decimal places for each entry.

    \item Use the provided code to simulate the continuous-time, nonlinear cart-pole system with the linear feedback controller $u = K_\infty \tilde{s}$. Initialize the system at $s = (0,3\pi/4,0,0)$, and use a controller sampling rate of $10~\mathrm{Hz}$. Verify from the simulation output that the behavior stabilizes as expected.


    \item We will now use an LQR controller to track a time-varying trajectory. Specifically, we will aim to balance the pendulum upright (that is, $\bar{\theta}(t) \equiv \pi$) while oscillating the position of the cart to track a desired reference $\bar{x}(t) = a\sin(2\pi t / T)$, where $a > 0$ and $T > 0$ are known constants.
    \begin{enumerate}[label=(\alph*)]
        \item Normally, as discussed in this chapter, you would have to re-linearize the system around the desired trajectory at each time step. Why is this not the case for this particular problem (that is, why can you just reuse $A$ and $B$)?

        \item Repeat part~(c) for this case with $a = 10$ and $T = 10$, except this time initialize the system upright at $s(0) = (0,\pi,0,0)$. For each state plot, overlay the corresponding entry from the reference trajectory $\bar{s}(t)$.

        \item You may notice that this controller does not have good tracking performance. You could try increasing the state penalty matrix $Q$ to, e.g., $Q = 10 I_4$. However, this should only improve tracking for $x(t)$ and $\dot{x}(t)$, while $\theta(t)$ and $\dot{\theta}(t)$ still oscillate around $\pi$ and $0$, respectively. What physical characteristic of the desired trajectory (or lack thereof) causes this to happen?
    \end{enumerate}
\end{enumerate}

\subsection*{\adjustbox{height=2ex, valign=c}{\includegraphics{figs/write.png}}\ \adjustbox{height=2ex, valign=c}{\includegraphics{figs/code.png}}\ Problem 3: Cart-pole Swing Up (iLQR)}
In this problem, we will implement a controller to solve the cart-pole ``swing up'' problem and will assume that we do not have actuation constraints. 
In the swing-up problem, the pendulum begins hanging downwards and is then brought to the upright position. Unlike in the cart-pole balancing problem, it is no longer sufficient to linearize around a single stationary point. Therefore, we will formulate the optimal control problem into a convex sub-problem and iteratively solve for optimal perturbations from a reference trajectory. Since we do not consider the case with limited actuation capabilities, the solution to each convex sub-problem can be solved with Riccati recursion, i.e., we will iteratively develop our solution using iterative LQR (iLQR) control. In this case, iLQR control provides an elegant and easy to implement closed-loop policy. However, through the standard LQR formulation at each iteration, we generally cannot reason, at least directly, over constraints on the state or control space: in the case of limited actuation, we require the more flexible and sophisticated framework provided by Sequential Convex Programming. The code for this exercise is located in \colorcode{ch03/exercises/cartpole\_iterative\_lqr.ipynb}.

Recall that the cart-pole is a continuous-time system with dynamics of the form $\dot{s} = f(s,u)$. To compute the iLQR control law, we will consider the Euler discretized dynamics:
\begin{equation}
    s_{k+1} \approx f(s_k,u_k) \defn s_k + \Delta{t}f(s_k, u_k),
\end{equation}
with time step $\Delta{t} > 0$. The provided code will then simulate this control law on the original continuous-time system.

\begin{enumerate}
    \item For a given operating point $(\bar{s}_k,\bar{u}_k)$, suppose we define the Jacobians:
    \begin{equation}
        A_k \defn \frac{\partial f}{\partial s}(\bar{s}_k,\bar{u}_k),
        \quad
        B_k \defn \frac{\partial f}{\partial u}(\bar{s}_k,\bar{u}_k).
    \end{equation}
    Use JAX in the function \colorcode{linearize} to write a single line of code that computes $A_k$ and $B_k$, given $f$, $\bar{s}_k$ and $\bar{u}_k$.
\end{enumerate}

For our iLQR controller, we will use the quadratic cost function:
\begin{equation}
    J(s,u) \defn 
    \frac{1}{2}\tran{(s_N - s_\mathrm{goal})} Q_N (s_N - s_\mathrm{goal})
    + \frac{1}{2}\sum_{k=0}^{N-1} \rbr*{\tran{(s_k-s_\mathrm{goal})} Q (s_k-s_\mathrm{goal}) + \tran{u_k} R u_k},
\end{equation}
where $s_\mathrm{goal}$ is the goal state (i.e., the upright position). The entries of $Q_N \succ 0$ are chosen to be large so that the terminal cost acts as a soft terminal ``constraint''.

\begin{enumerate}[resume]
    \item  Rewrite the cost function in terms of the deviations $\tilde{s}_N \defn s_N - \bar{s}_N$, $\tilde{s}_k \defn s_k - \bar{s}_k$, and $\tilde{u}_k \defn u_k - \bar{u}_k$. This will result in a quadratic cost function with linear terms of the form $\tran{q_N}\tilde{s}_N$, $\tran{q_k}\tilde{s}_k$, and $\tran{r_k}\tilde{u}_k$.
    Identify the vectors $q_N$, $q_k$, and $r_k$.

    \item Complete the iLQR controller code in the function \\\noindent\colorcode{discreteIterativeLQR.solve}. Specifically, your code must update the controller gain and offset terms $\{Y_k\}_{k=0}^{N-1}$ and $\{y_k\}_{k=0}^{N-1}$, respectively\footnote{
        In \cref{subsubsec:ilqr_ddp}, we labeled these terms as $K_t$ and $k_t$, but here we try to avoid confusion with the discrete index $k$.
    }, the nominal trajectory $(\bar{s},\bar{u})$, and the deviations $(\tilde{s}, \tilde{u})$.

    \item Implement the function \colorcode{compute\_control} for the \colorcode{discreteIterativeLQR} class to apply either the open-loop iLQR control input or the closed-loop iLQR policy, depending on the value of the Boolean flag \colorcode{closed\_loop}. Run your code for both cases to simulate the system and generate plots of the state and control input over time. You should notice that the iLQR control sequence does not accomplish the task if applied open-loop.
\end{enumerate}

\subsection*{\adjustbox{height=2ex, valign=c}{\includegraphics{figs/code.png}}\ Problem 4: Linear MPC}
In \cref{subsec:mpc}, we introduced some of the fundamental aspects of model predictive control.
For this exercise, you will implement a simple MPC scheme for a linear system to drive the system to the origin while minimizing the cost function:
\begin{equation*}
J(\x_0) = \sum_{t=0}^{\infty} \big( \x_t^\top Q \x_t + \u_t^\top R \u_t \big),
\end{equation*}
and satisfying the inequality constraints on the state and control, $\x_t \leq \x_u$, $\x_t \geq \x_l$, and $\norm{\u_t}_\infty \leq r_u$.
To solve this problem with a finite-horizon MPC scheme, we will use the cost function from \cref{eq:mpc_cost_function} which adds a terminal cost and we will also add a terminal constraint $\norm{\x_N}_\infty \leq r_N$ to the problem.
In the notebook \colorcode{ch03/exercises/linear\_mpc.ipynb}:
\begin{enumerate}
\item Implement the function \colorcode{solve\_mpc} to solve a finite horizon MPC problem with the quadratic stage and terminal costs, inequality constraints on the state and control, and the terminal constraint discussed above.
\item Run the provided code to simulate the closed-loop system with the terminal cost $Q_f = Q$ and with no terminal constraint (that is, set $r_N = \infty$).
Analyze the resulting plots, explain why there is a difference between the trajectory computed in each MPC iteration's optimization problem and the final closed-loop trajectory.
\item Next, run the provided code that specifies a different terminal cost matrix $Q_f$ by the solution to the discrete-time algebraic Riccati equation (DARE) from \cref{eq:dare} in the section on the linear quadratic regulator.
What do you notice about the difference between the MPC trajectory computed at each time step and the closed-loop system behavior?
Why is this different than when using $Q_f = Q$?
Notice that the total cost of the closed-loop trajectory is lower when using the DARE terminal cost matrix, why is that?
\item As we discussed in \cref{subsec:mpc}, persistent feasibility is a challenge with MPC, where we want to guarantee that if we find a feasible solution at one time step then we will be able to find a feasible solution at future time steps.
One simple solution to this problem is to apply a terminal constraint $\x_N = 0$, which forces the optimizer to find a trajectory to the origin by the end of the finite horizon, thereby eliminating the issue with being short-sighted.
However, this will cause the controller to be sub-optimal and can require a longer horizon to find a feasible solution.
Run the provided code and play around with the controller's horizon, how does the closed-loop trajectory's cost change as the horizon changes?
\end{enumerate}

\subsection*{\adjustbox{height=2ex, valign=c}{\includegraphics{figs/code.png}}\ Problem 5: Extended Unicycle Trajectory Tracking Control}
In \cref{ex:pd-control-extended-unicycle}, we introduced a closed-loop trajectory tracking problem for a dynamically extended unicycle model where we apply a PD controller.
In this example, we showed that since the system is differentially flat we could write its dynamics as the second order linear system in \cref{eq:unicycle_flat_dynamics}.
From these dynamics, we can develop a PD controller for the virtual inputs $w_1$ and $w_2$ given by \cref{eq:pd-unicycle-control}, and then can map these virtual controls back into the actual controls $(a, \omega)$ algebraically using:
\begin{equation*}
\begin{bmatrix} a \\[3pt] \omega \end{bmatrix}
    = J^{-1}(\theta,v)
    \begin{bmatrix} w_1 \\[3pt] w_2 \end{bmatrix}.
\end{equation*}

In this exercise, you will compute a desired reference trajectory for the extended unicycle and then will design a PD controller to track the reference in the presence of external disturbances.
The code implementation aspects of the exercise can be completed in the notebook \\\noindent\colorcode{ch03/exercises/unicycle\_trajectory\_tracking.ipynb}.

\begin{enumerate}
\item First, you will design the reference trajectory for the unicycle to follow.
While there are several ways to accomplish this, for this exercise you will use a polynomial basis expansion of the form:
\begin{equation*}
x_d(t) = \sum_{i=1}^n x_i \psi_i(t), \quad y_d(t) = \sum_{i=1}^n y_i \psi_i(t),
\end{equation*}
where $\psi_i$ for $i = 1, \dots, n$ are the basis functions, and $x_i$ and $y_i$ are coefficients that can be selected.
Use the basis functions $\psi_1(t) = 1$, $\psi_2(t) = t$, $\psi_3(t) = t^2$, $\psi_4(t) = t^3$.
\begin{enumerate}
\item Write a set of linear equations in the coefficients $x_i$ and $y_i$ for $i = 1, \dots, 4$ to express the following initial and final conditions:
\begin{equation*}
\begin{split}
x(0) = x_0, \quad y(0) = y_0, \quad v(0) = v_0, \quad \theta(0) = \theta_0, \\
x(t_f) = x_f, \quad y(t_f) = y_f, \quad v(t_f) = v_f, \quad \theta(t_f) = \theta_f. \\
\end{split}
\end{equation*}
\item Implement the function \colorcode{compute\_traj\_coeffs} to compute the coefficients $x_i$ and $y_i$.
\item Then, implement the function \colorcode{compute\_traj} to use the coefficients and the basis functions to compute the full reference trajectory $x_d(t)$ and $y_d(t)$.
Additionally, \colorcode{compute\_traj} should also compute $v_d(t)$\sidenote{Hint: from the dynamics model we can see $v = \sqrt{\dot{x}^2 + \dot{y}^2}$.}, $\theta_d(t)$, $\dot{x}_d(t)$, $\dot{y}_d(t)$, $\ddot{x}_d(t)$, and $\ddot{y}_d(t)$.
\item Now, implement the function \colorcode{compute\_controls} to compute $a(t)$ and $\omega(t)$ for a given state trajectory.
Run the provided code to compute a trajectory and control sequence for the initial and final conditions:
\begin{equation*}
\begin{split}
x(0) = 0, \quad y(0) = 0, \quad v(0) = \frac{1}{2}, \quad \theta(0) = -\frac{\pi}{2}, \\
x(t_f) = 5, \quad y(t_f) = 5, \quad v(t_f) = \frac{1}{2}, \quad \theta(t_f) = -\frac{\pi}{2}, \\
\end{split}
\end{equation*}
with $t_f = 25$.
Why do we have to choose $v(t_f) > 0$?
What would happen if we let $v(t) = 0$ at some point along the trajectory?
\item Run the provided code to simulate the unicycle system with the controls $a(t)$ and $\omega(t)$ computed previously in open-loop while subject to disturbances.
Notice how the system does not reach the target state in the presence of noise.
\end{enumerate}

\item Now that we can compute a reference trajectory for the unicycle to follow, we want to compute a closed-loop controller to track the trajectory robustly in the presence of disturbances.
Specifically, you will use the PD controller from \cref{eq:pd-unicycle-control} to compute the virtual controls $w_1$ and $w_2$.
\begin{enumerate}
\item Write down a system of equations for computing the control inputs $a$ and $\omega$ in terms of the virtual controls $w_1 = \ddot{x}$ and $w_2 = \ddot{y}$ and the unicycle state $\x = [x, y, v, \theta]^\top$.
Use this to implement the function \\\noindent\colorcode{TrajectoryTracker.compute\_control}.
\item Run the provided code to run the closed-loop tracking controller.
How does the performance compare to the open-loop?
Experiment with different starting and ending conditions, and different amounts of noise.
\end{enumerate}
\end{enumerate}
\newpage
\printbibliography[segment=\therefsegment,heading=subbibliography,title={References}]
\chapter{Motion Planning}
\label{ch:motion-planning}
\newrefsegment
As introduced in Chapters~\ref{ch:openloop} and \ref{ch:closedloop}, autonomous systems transform high-level goals into concrete physical actions through a hierarchy of processes, ranging from strategic decision-making to real-time control.
While open-loop trajectory optimization offers a principled framework for generating dynamically feasible state and control sequences, it often struggles to handle complex, state-dependent constraints, such as collision avoidance.
Moreover, trajectory optimization methods typically yield only \emph{locally optimal} solutions, whose quality can strongly depend on initialization and may fail to capture the global structure of the problem.

To address these limitations, this chapter focuses on \emph{motion planning}, which takes a more \emph{global} perspective by explicitly reasoning about the full set of environmental constraints, obstacles, and trajectory feasibility.
To maintain computational tractability, motion planning often relies on simplified dynamical models, producing coarse or geometric paths that ensure collision-free navigation.
In practice, motion planning and trajectory optimization often complement each other, with motion planning establishing global feasibility by finding an initial collision-free path, and trajectory optimization refining this path into a smooth, dynamically consistent, and optimized trajectory.

The study of the motion planning problem, formally defined in the 1970s, has a rich history.
Early work in the 1980s focused on exact, combinatorial algorithms designed to capture the geometry of configuration spaces with mathematical rigor.
While theoretically elegant, these methods suffered from severe computational bottlenecks in high-dimensional problems.
The 1990s brought a paradigm shift with the introduction of sampling-based approaches, which offered scalable solutions for complex motion planning problems.
The 2000s marked the widespread deployment of planning algorithms on real-time systems, from autonomous vehicles to robotic manipulators. 
Today, research in motion planning is vibrant and expanding, focusing on integrating differential and logical constraints, planning under uncertainty, leveraging parallel computation, and incorporating learning-based methods.

The relevance of motion planning is reflected in its breadth of applications.
To name a few, in autonomous driving, planning algorithms compute safe and efficient maneuvers for vehicles navigating dense traffic. 
In humanoid robotics, planners coordinate whole-body motions for tasks such as walking, climbing, or manipulation in cluttered environments.
In surgical robotics, motion planning is used to design precise, collision-free tool trajectories inside the human body. 
Even outside classical robotics, motion planning techniques have found applications in fields such as protein folding, where high-dimensional configuration spaces must be navigated to determine protein folding pathways.

Over the past decades, a diverse range of algorithmic approaches has been developed to address motion planning problems. 
These approaches are commonly grouped into four main categories: grid-based planning, combinatorial planning, sampling-based planning, and potential field methods.
At a high level, \emph{grid-based planners} discretize the robot's environment into a grid and use graph search algorithms to compute a feasible path through the grid cells.
\emph{Combinatorial planners} construct explicit representations of the configuration space that capture the connectivity of the free space required for planning.
\emph{Sampling-based planners} leverage random sampling and collision detection to incrementally explore the configuration space, offering scalability to high-dimensional problems where explicit representations become intractable.
Finally, \emph{potential field methods} construct artificial potential functions that attract the robot toward its goal while pushing it away from obstacles.

In this chapter, we begin by formally defining the motion planning problem in~\cref{subsec:mot-plan-prob-formulation}, introducing the core concepts and notation. 
We then explore the four major algorithmic paradigms for motion planning discussed above, namely grid-based planning in~\cref{subsec:grid-based-planning}, combinatorial planning in~\cref{sec:combinatorial-motion-planning}, sampling-based planning in~\cref{sec:sampling-based-motion-planning}, and potential field methods in~\cref{sec:potential-field-methods}.

\subsection{Problem Formulation}
\label{subsec:mot-plan-prob-formulation}
We begin by formally defining the motion planning problem in its most basic formulation.
Let $\mathcal{W} \subseteq \mathbb{R}^2$ denote the robot's \emph{workspace}, i.e., the physical environment in which the robot operates.
Within this workspace lies a known obstacle region $\mathcal{O} \subset \mathcal{W}$, characterized by a polygonal (piecewise-linear) boundary.
The robot itself is modeled as a rigid polygon that must navigate through the workspace without intersecting any obstacles\sidenote{For a 3D workspace, the only differences are that $\mathcal{W} \subseteq \mathbb{R}^3$, and the robot and the obstacle region are represented as polyhedra.}\sidenote{The basic formulation presented here extends well beyond rigid polygons and polyhedra to include robots with complex geometries. For an in-depth treatment of these extensions, we refer the reader to \citet{LaValle2006}.}.

With this setup, the fundamental motion planning problem can be informally stated as follows: \emph{given an initial placement of the robot, compute how to gradually move it into a desired goal placement without colliding with any obstacles.}
As an illustration, consider the workspace in~\cref{fig:2dworkspace}, where the task is to move an L-shaped robot from its initial position to a target position while avoiding polygonal obstacles.
In this case, the output of a motion planning algorithm will be a path through the set of all intermediate transformations of the robot, from start to goal.

Although the motion planning problem is naturally described in the robot's workspace, it really lives in another space: the set of all rigid-body transformations that describe the robot’s possible placements.
As discussed in~\cref{ch:model-dyn}, this set is referred to as the \emph{configuration space} or $\C$-space. 
In this formulation, obstacles in the workspace induce forbidden regions in $\C$, and the motion planning problem reduces to finding a continuous path in $\C$ that avoids collisions.

Let us briefly review the notion of configuration space through an example.
\begin{example}[L-shaped Robot] 
	\label{ex:mot-plan}
	\theoremstyle{definition}
	Consider the L-shaped robot in~\cref{fig:2dworkspace}. The task is to move it from an initial placement to a goal placement in a two-dimensional world with polygonal obstacles.
	Suppose the robot is described by a state:
	$$
	\x = \begin{bmatrix}x & y & \theta & \dot{x} & \dot{y} & \dot{\theta}\end{bmatrix}^\top,
	$$ 
	where $(x,y)$ denote position, $\theta$ denotes orientation, and $(\dot{x}, \dot{y}, \dot{\theta})$ denote velocities.
	For the simplest version of motion planning, we restrict the attention to the robot's \emph{configuration}:
	$$
	\q = \begin{bmatrix}x & y & \theta\end{bmatrix}^\top,
	$$
	which fully describes its degrees of freedom.
	In other words, every combination of $(x,y,\theta)$ corresponds to a unique placement of the robot in the workspace, and is called a \emph{configuration}.
	This abstraction simplifies the problem: instead of computing a trajectory for the full state, we seek a sequence of collision-free configurations, as shown in the right-side graphic of~\cref{fig:2dworkspace}.
	The resulting geometric path in configuration space can then be provided to trajectory optimization methods for refinement---allowing them to incorporate system dynamics---and subsequently to closed-loop control methods (e.g., tracking controllers introduced in~\cref{ch:closedloop}) for execution.
	\begin{figure}[tb] 
		\centering 
		\includegraphics[width=0.95\linewidth]{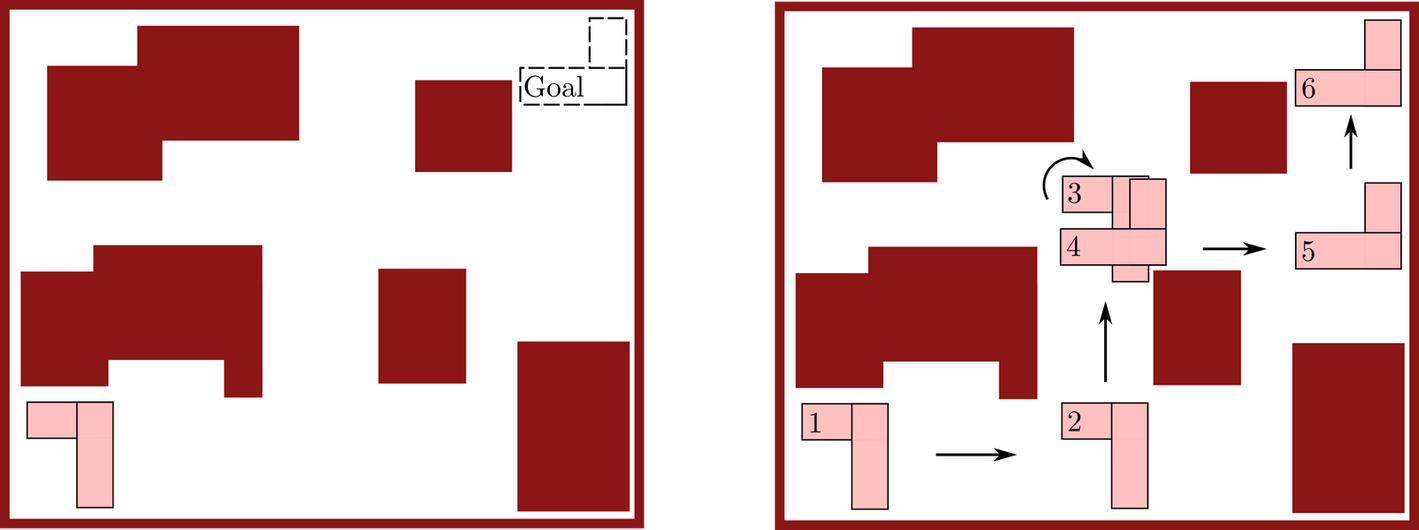}
		\caption{Motion planning in a two-dimensional workspace with obstacles.
		The task is to move the L-shaped robot from start (bottom-left) to goal (top-right) without collisions.}
		\label{fig:2dworkspace} 
	\end{figure}
	\begin{figure}[tb]
		\centering
		\includegraphics[width=0.80\textwidth]{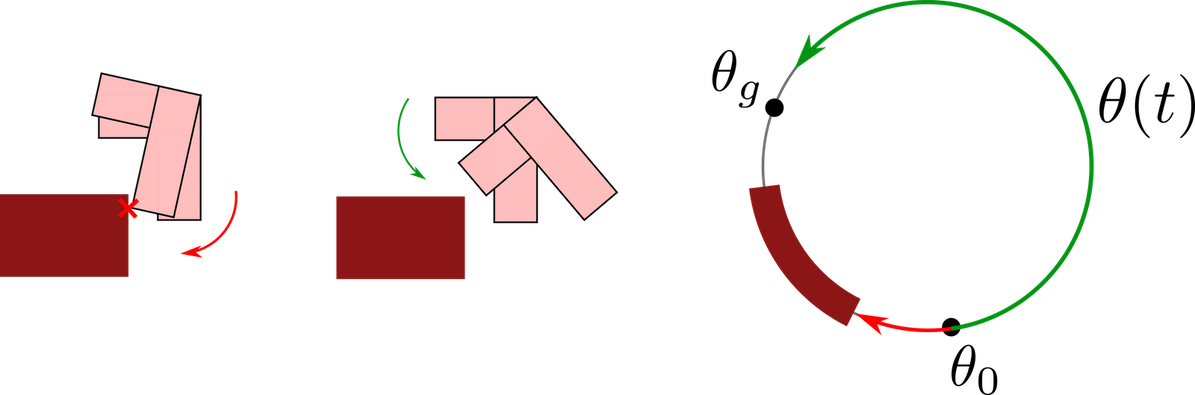}
		\caption{Configuration space defined on $\R^2 \times \mathcal{S}^1$.
		Rotating clockwise from $\theta_0$ to $\theta_g$ leads to collision, while rotating counter-clockwise yields a feasible path.}
		\label{fig:rotational-dof-fig}
	\end{figure}
	In this example, the configuration space is $\C = \R^2 \times \mathcal{S}^1 \subset \R^3$, where $\R^2$ represents the robot's position in the plane and $\mathcal{S}^1$ is the one-dimensional unit circle manifold representing the robot's orientation.
	The presence of the manifold $\mathcal{S}^1$ reflects the periodicity of the orientation variable $\theta$, where $\theta$ and $\theta \pm 2\pi k$ are equivalent for all integers $k$.
	This periodicity has important practical implications for planning, as the robot can reach the same orientation either by rotating clockwise or counter-clockwise\sidenote{For instance, a heading change of $\pi/2$ radians can be achieved by turning left by $\pi/2$ or turning right by $3\pi/2$.}. 
	In the scenario depicted in~\cref{fig:rotational-dof-fig}, suppose the robot has an initial heading $\theta_0$ and a goal heading $\theta_g$.
	If one were to consider only clockwise rotations, there would be no feasible path to the goal without colliding with the obstacle.
	However, once periodicity is properly accounted for by modeling orientation as $\mathcal{S}^1$, the robot can simply rotate counter-clockwise to reach the goal without collision.
	Therefore, in this example, the motion planning problem reduces to finding a continuous path in $\C = \R^2 \times \mathcal{S}^1$ that avoids the forbidden regions induced by the obstacles in the workspace.

	Crucially, the concept of configuration space generalizes to robots with more complex geometries and higher degrees of freedom (e.g., robotic arms with multiple joints).
\end{example}

\paragraph{Free space in configuration space.}
A fundamental concept in motion planning is the \emph{free space}, denoted by $\C_{\mathrm{free}}$.
Intuitively, $\C_{\mathrm{free}}$ is the set of all robot configurations in which the robot does not collide with any obstacles in the workspace. 
Formally, let $R(q) \subseteq \mathcal{W}$ denote the set of points in the workspace occupied by the robot when placed at configuration $q$. 
The free space is defined as:
\begin{equation}
	\C_{\mathrm{free}} = \{ q \in \C \,|\, R(q) \cap \mathcal{O} = \emptyset \}.
\end{equation}
Similarly, the \emph{obstacle region} in configuration space, denoted by $\C_{\mathrm{obs}}$, is defined as the complement of the free space:
\begin{equation}
	\C_{\mathrm{obs}} = \{ q \in \C \mid R(q) \cap \mathcal{O} \neq \emptyset \}.
\end{equation}
\begin{figure}[tb]
	\centering
	\includegraphics[width=0.75\linewidth]{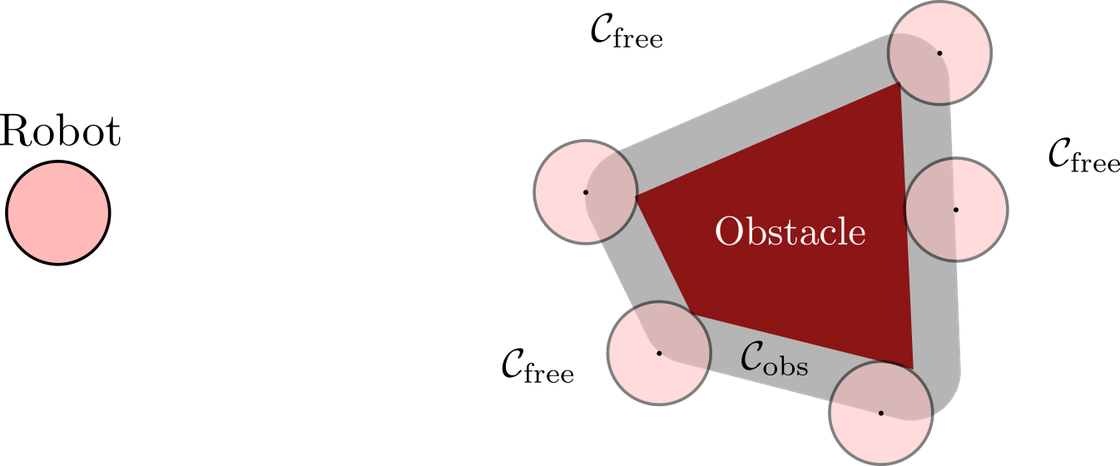}
	\caption{Free ($\C_{\mathrm{free}}$) and forbidden ($\C_{\mathrm{obs}}$) spaces of the configuration space for a circular robot in the presence of a polygonal obstacle. 
	The forbidden region accounts for the robot’s physical dimensions.}
	\label{fig:cspace-obstacle}
\end{figure}
To illustrate these definitions, consider the simple case of a circular robot navigating among polygonal obstacles, as shown in~\cref{fig:cspace-obstacle}. 
Although the robot itself is a disk of nonzero radius, collision checking in configuration space reduces to verifying that a \emph{point-like} representation of the robot (its configuration) does not intersect the obstacle regions in configuration space.
In this case, this equivalence is achieved by inflating the workspace obstacles by the robot’s radius.
The red region in the figure represents the obstacle in the physical workspace, $\mathcal{O}$, whereas the grey region corresponds to the inflated obstacle in configuration space, $\C_{\mathrm{obs}}$.

Once $\C_{\mathrm{free}}$ (and, correspondingly, $\C_{\mathrm{obs}}$) has been computed, the robot’s physical dimensions no longer need to be explicitly considered, and the robot can be treated as a point moving through the configuration space, as illustrated in the previous example. 
With this abstraction, the geometric complexity of the robot and the environments are absorbed into the structure of the configuration space itself.

\begin{figure}[tb]
	\centering
	\includegraphics[width=0.75\linewidth]{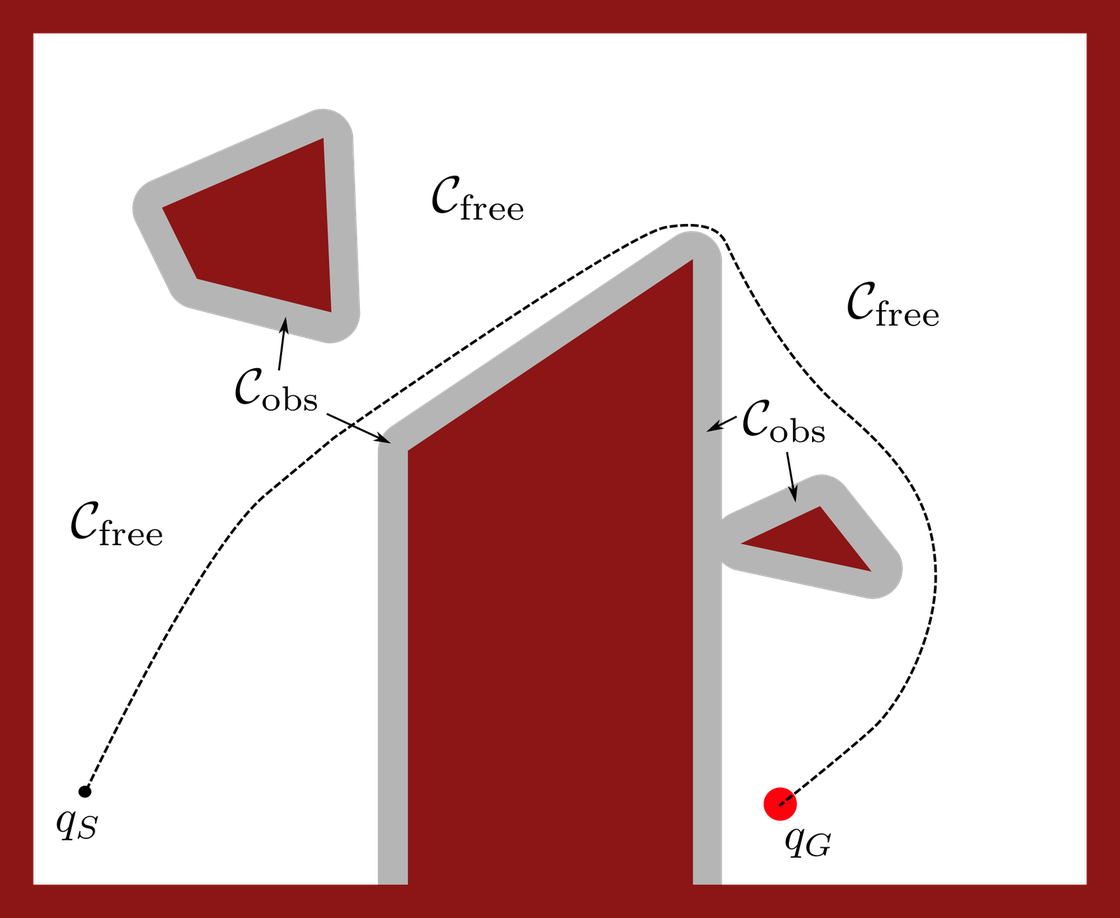}
	\caption{Example of a path planning problem in configuration space. 
	The free space $\C_{\mathrm{free}}$ excludes the forbidden configurations $\C_{\mathrm{obs}}$ induced by the polygonal obstacles. 
	Motion planning reduces to finding a continuous path from the start to the goal within $\C_{\mathrm{free}}$.
	Note that for a simple point robot in 2D, the $\C$-space and the workspace would coincide.}
	\label{fig:cspace-path}
\end{figure}

Within this abstracted setting, the motion planning problem reduces to finding a continuous path $\tau: [0,1] \to \C_{\mathrm{free}}$ such that $\tau(0) = \startnode$ and $\tau(1) = \goalnode$, where $\startnode$ and $\goalnode$ denote the start and goal configurations, respectively. 
An example is shown in~\cref{fig:cspace-path}, where the robot must navigate around forbidden regions in $\C$-space to reach the goal. 
By working in configuration space, the problem of collision-free navigation becomes purely geometric, focused on finding a path that avoids the forbidden regions\sidenote{As we will see in~\cref{subsubsec:kinodynamic-planning}, this purely geometric interpretation is true for static environments and without kinodynamic constraints.}.

\medskip
\noindent In the remainder of this chapter, we explore various algorithmic strategies that aim to solve either discretized (\cref{subsec:grid-based-planning}) or continuous (Sections~\ref{sec:combinatorial-motion-planning}-\ref{sec:potential-field-methods}) versions of the motion planning problem.

\subsection{Grid-based Motion Planning}
\label{subsec:grid-based-planning}
A natural way to simplify the motion planning problem is to discretize the robot’s continuous configuration space into a grid. 
Instead of reasoning about infinitely many possible configurations, we approximate the environment by dividing it into a finite set of cells. 
Each cell is classified as either free (i.e., collision-free, meaning the robot can occupy it) or forbidden (i.e., meaning it would lead to a collision with an obstacle). 
The robot is allowed to move between adjacent free cells, and the planning task reduces to finding a sequence of connected free cells that leads from the start to the goal.

This discretization naturally converts motion planning into a \emph{graph search problem}.
Specifically, each free cell in the grid is represented as a vertex in a graph, and an edge is added between two vertices whenever the corresponding cells are adjacent and both free. 
Formally, we construct a graph $\graph = (\node,\edge)$, where each vertex $v \in \node$ corresponds to a free cell, and each edge $(v,u) \in \edge$ corresponds to a valid robot move between adjacent cells. 
Planning then amounts to finding a path in the graph from the start vertex to the goal vertex.

This perspective is powerful because it allows motion planning to draw on decades of results from graph theory and computer science. 
Algorithms originally developed for solving shortest-path problems---such as breadth-first search or Dijkstra’s algorithm---can be directly applied to motion planning. 
In other words, the seemingly geometric problem of ``navigating through a space with obstacles'' is reduced to the purely combinatorial problem of ``finding a path through a graph''.
This approach, however, relies on an explicit characterization of $\C_{\mathrm{free}}$, which is necessary to construct the grid and classify each cell as either free or occupied.
In practice, obtaining such an exact characterization is often infeasible—particularly in high-dimensional configuration spaces—thereby motivating alternative strategies, such as sampling-based planning, discussed later in this chapter.

\paragraph{Connectivity and neighborhood structure.}
The way adjacency is defined depends on the robot model and the chosen grid structure:
\begin{itemize}
	\item In a 4-connected grid, each cell has up to four neighbors (up, down, left, right).
	\item In an 8-connected grid, diagonal moves are also allowed, giving each cell up to eight neighbors.
	\item More generally, one can consider $k$-connected neighborhoods (e.g., 16-connected) or hexagonal tilings, trading off simplicity, path quality, and computational cost.
\end{itemize}
The choice of connectivity has practical consequences. 
A coarse neighborhood may restrict the robot’s ability to approximate smooth trajectories, while a richer neighborhood increases branching factors and search complexity. 
Regardless of the specific choice, the key point is that the grid structure induces a well-defined graph on which search algorithms can operate.

\subsubsection{Label Correcting Algorithms}
\label{subsubsec:label-correcting-algorithms}
Once the motion planning problem has been cast as a graph search, the next step is to design an algorithm that systematically explores the graph to find a path from start to goal. 
A broad and powerful family of methods for this task is known as \emph{label-correcting algorithms}.

The term “label” refers to a numerical value associated with each vertex in the graph, which represents the cost of the best-known path from the start vertex to that vertex.
At a high level, label correcting algorithms proceed by iteratively refining these cost labels.
Initially, only the start vertex has a cost label of zero, while all others are initialized to infinity (or an undefined state). 
The algorithm proceeds by exploring the graph and updating, or \emph{correcting}, these labels whenever a better (lower-cost) path to a vertex is discovered.

\medskip
Formally, let $\startnode$ denote the start vertex and $\goalnode$ the goal vertex\sidenote{We use the notation $q$ for graph vertices to emphasize the connection between graph search and path planning in configuration space, where $q$ typically denotes a configuration.}.
For each vertex $q$, let $C(q)$ represent the label, i.e., the cost of the best path found so far from $\startnode$ to $q$.
This value is also known as the \emph{cost of arrival}.

Label-correcting algorithms operate by maintaining a \emph{frontier} (also referred to as the \emph{alive set} or a \emph{priority queue}) of vertices whose neighbors may still admit cost improvements.
At each iteration, a vertex $q$ is extracted from the frontier and \emph{expanded}.
\emph{Expansion} refers to the process of examining all outgoing edges from $q$ to its neighboring vertices $q'$.
For every neighbor\sidenote{In graph theory, two vertices are denoted as neighbors if they are connected by an edge.} $q^\prime$ of $q$, the algorithm attempts to improve (or \emph{relax}) the current cost estimate associated with $q'$.
Specifically, given an edge $(q, q')$ with cost $c(q, q')$, the relaxation step verifies whether the path through $q$ offers a lower cost of arrival to $q'$ than any previously known path---namely, whether:
$$ 
C(q) + c(q,q^\prime) < C(q^\prime).
$$
If this condition holds, the label of $q'$ (i.e., $C(q^\prime)$) is updated and $q^\prime$ is reinserted into the frontier for further exploration.
This process continues until no label can be further improved.

\paragraph{General structure of label-correcting algorithms.}
All algorithms in this family share the following structure:
\begin{itemize}
	\item \emph{Initialization:} Set $C(\startnode) = 0$ and $C(q) = \infty$ for all other vertices $q$. Initialize the frontier with the start vertex $Q = \{\startnode\}$.
	\item \emph{Main Loop:} Until the frontier $Q$ is not empty, repeat:
		\begin{enumerate}
			\item Extract a vertex $q$ from the frontier $Q$ according to a specific \emph{selection rule}\sidenote{The choice of selection rule is the main distinguishing factor between different label-correcting algorithms.}.
			\item For each neighbor $q^\prime$ of $q$, perform the relaxation step, i.e.:
			\begin{enumerate}
				\item Check if $C(q) + c(q,q^\prime) < C(q^\prime)$.
				\item If so, update $C(q^\prime) \leftarrow C(q) + c(q,q^\prime)$ and add $q^\prime$ to the frontier $Q$ if it is not already present.
			\end{enumerate}
		\end{enumerate}
	\item \emph{Termination:} The algorithm terminates when the frontier is empty. At this point, $C(q)$ contains the cost of the shortest path from $\startnode$ to each reachable vertex $q$.
\end{itemize}
The pseudocode for the general structure of a label-correcting algorithm is provided in Algorithm~\ref{alg:label-correcting}.
\begin{algorithm}[t]
	\caption{Structure of a Label-Correcting Algorithm}
	\label{alg:label-correcting}
	\DontPrintSemicolon
	\KwData{$\startnode$, $\goalnode$, directed graph $\graph=(\node,\edge)$, edge costs $c(\cdot,\cdot)$, selection rule $\mathsf{Select}$}
	\KwResult{Shortest path from $\startnode$ to $\goalnode$ (if reachable)}
	\For{$q \in \node \setminus \{\startnode\}$}{
		$C(q) \gets \infty$\\
	}
	$C(\startnode) \gets 0$ \\
	$Q \gets \{\startnode\}$ \tcc*[r]{Initialize priority queue}
	\While{$Q$ is not empty}{
		$q \gets \mathsf{Select}(Q)$; $Q.\mathrm{remove}(q)$ \\
		\For{$q' \in \{\,u \mid (q,u)\in \edge\,\}$}{
			$\tilde{C} \gets C(q) + c(q,q')$ \\
			\If{$\tilde{C} < C(q')$}{
				$C(q') \gets \tilde{C}$; \quad $\mathrm{parent}(q') \gets q$ \\
				\If{$q' \neq \goalnode$ \textbf{and} $q' \notin Q$}{
					$Q.\mathrm{add}(q')$
				}
			}
		}
	}
	\eIf{$C(\goalnode) = \infty$}{
		\Return{N.A.} \tcc*[r]{Goal unreachable}
	}{
		\tcc*[r]{Reconstruct path by backtracking parents}
		$\mathrm{path} \gets [\,]$; $q \gets \goalnode$ \\
		\While{$q \neq \startnode$}{$\mathrm{path}$.$\mathrm{prepend}(q)$; $q \gets \mathrm{parent}(q)$}
		\Return{$\mathrm{path}$}
	}
\end{algorithm}

\paragraph{The role of the selection rule.}
The selection rule for choosing the next vertex to expand is one of the key factors that differentiates various label-correcting algorithms, ultimately defining their efficiency and performance.
Several classical search methods can be seen as special cases of the label-correcting framework, for example:
\begin{itemize}
	\item \emph{Depth-First Search (DFS):} in DFS~(\cref{fig:depth-first-search}), the frontier is managed as a \emph{stack} (last-in, first-out). 
	This means that the algorithm always continues along the most recently discovered path before backtracking. 
	DFS is appealing for its simplicity and low memory footprint, since the number of active nodes is proportional to the depth of the search. 
	However, DFS is exposed to the risk of getting trapped in deep but unproductive branches of the search tree, leading to poor performance in finding the shortest path.
	\begin{marginfigure}
		\begin{center}
			\begin{tikzpicture}[node distance=0.5cm, ->]
				\tikzstyle{visted} = [draw=black, circle, rounded corners, minimum height=1.5em, minimum width=1.5em, fill=red!30]
				\tikzstyle{node} = [draw=black, dashed, circle, rounded corners, minimum height=1.5em, minimum width=1.5em, fill=gray!10]
				\node[visted](s0){\tiny $0$};
				\node[visted, below of=s0, xshift=-1.33cm, yshift=-0.1cm](s1){\tiny $1$};
				\node[node, below of=s0, xshift=1.33cm, yshift=-0.1cm](s2){\tiny};
				\node[visted, below of=s1, xshift=-.66cm, yshift=-0.2cm](s3){\tiny $2$};
				\node[visted, below of=s1, xshift=.66cm, yshift=-0.2cm](s4){\tiny $5$};
				\node[node, below of=s2, xshift=-.66cm, yshift=-0.2cm](s5){\tiny};
				\node[node, below of=s2, xshift=.66cm, yshift=-0.2cm](s6){\tiny};
				\node[visted, below of=s3, xshift=-0.33cm, yshift=-0.25cm](s7){\tiny $3$};
				\node[visted, below of=s3, xshift=0.33cm, yshift=-0.25cm](s8){\tiny $4$};
				\node[visted, below of=s4, xshift=-0.33cm, yshift=-0.25cm](s9){\tiny $6$};
				\node[node, below of=s4, xshift=0.33cm, yshift=-0.25cm](s10){\tiny};
				\node[node, below of=s5, xshift=-0.33cm, yshift=-0.25cm](s11){\tiny};
				\node[node, below of=s5, xshift=0.33cm, yshift=-0.25cm](s12){\tiny};
				\node[node, below of=s6, xshift=-0.33cm, yshift=-0.25cm](s13){\tiny};
				\node[node, below of=s6, xshift=0.33cm, yshift=-0.25cm](s14){\tiny};
				\draw[->] (s0) to[] (s1);
				\draw[->] (s0) to[] (s2);
				\draw[->] (s1) to[] (s3);
				\draw[->] (s1) to[] (s4);
				\draw[->] (s2) to[] (s5);
				\draw[->] (s2) to[] (s6);
				\draw[->] (s3) to[] (s7);
				\draw[->] (s3) to[] (s8);
				\draw[->] (s4) to[] (s9);
				\draw[->] (s4) to[] (s10);
				\draw[->] (s5) to[] (s11);
				\draw[->] (s5) to[] (s12);
				\draw[->] (s6) to[] (s13);
				\draw[->] (s6) to[] (s14);
			\end{tikzpicture}
		\end{center}
		\caption{Depth-first search}
		\label{fig:depth-first-search}
	\end{marginfigure}
	\item \emph{Breadth-First Search (BFS):} in BFS~(\cref{fig:breadth-first-search}), the frontier is a \emph{queue} (first-in, first-out). 
	The algorithm expands all nodes at a given “depth” before moving on to the next, effectively exploring the graph in concentric layers around the start node.
	Compared to DFS, BFS requires significantly more memory, as it must store all frontier nodes at a given depth.
	\begin{marginfigure}
		\begin{center}
			\begin{tikzpicture}[node distance=0.5cm, ->]
				\tikzstyle{visted} = [draw=black, circle, rounded corners, minimum height=1.5em, minimum width=1.5em, fill=red!30]
				\tikzstyle{node} = [draw=black, dashed, circle, rounded corners, minimum height=1.5em, minimum width=1.5em, fill=gray!10]
				\node[visted](s0){\tiny $0$};
				\node[visted, below of=s0, xshift=-1.33cm, yshift=-0.1cm](s1){\tiny $1$};
				\node[visted, below of=s0, xshift=1.33cm, yshift=-0.1cm](s2){\tiny $2$};
				\node[visted, below of=s1, xshift=-.66cm, yshift=-0.2cm](s3){\tiny $3$};
				\node[visted, below of=s1, xshift=.66cm, yshift=-0.2cm](s4){\tiny $4$};
				\node[visted, below of=s2, xshift=-.66cm, yshift=-0.2cm](s5){\tiny $5$};
				\node[node, below of=s2, xshift=.66cm, yshift=-0.2cm](s6){\tiny};
				\node[node, below of=s3, xshift=-0.33cm, yshift=-0.25cm](s7){\tiny};
				\node[node, below of=s3, xshift=0.33cm, yshift=-0.25cm](s8){\tiny};
				\node[node, below of=s4, xshift=-0.33cm, yshift=-0.25cm](s9){\tiny};
				\node[node, below of=s4, xshift=0.33cm, yshift=-0.25cm](s10){\tiny};
				\node[node, below of=s5, xshift=-0.33cm, yshift=-0.25cm](s11){\tiny};
				\node[node, below of=s5, xshift=0.33cm, yshift=-0.25cm](s12){\tiny};
				\node[node, below of=s6, xshift=-0.33cm, yshift=-0.25cm](s13){\tiny};
				\node[node, below of=s6, xshift=0.33cm, yshift=-0.25cm](s14){\tiny};
				\draw[->] (s0) to[] (s1);
				\draw[->] (s0) to[] (s2);
				\draw[->] (s1) to[] (s3);
				\draw[->] (s1) to[] (s4);
				\draw[->] (s2) to[] (s5);
				\draw[->] (s2) to[] (s6);
				\draw[->] (s3) to[] (s7);
				\draw[->] (s3) to[] (s8);
				\draw[->] (s4) to[] (s9);
				\draw[->] (s4) to[] (s10);
				\draw[->] (s5) to[] (s11);
				\draw[->] (s5) to[] (s12);
				\draw[->] (s6) to[] (s13);
				\draw[->] (s6) to[] (s14);
			\end{tikzpicture}
		\end{center}
		\caption{Breadth-first search}
		\label{fig:breadth-first-search}
	\end{marginfigure}
	\item \emph{Dijkstra's Algorithm (also known as Best-First Search):} in Dijkstra’s algorithm (Algorithm~\ref{alg:dijkstra}), the frontier is a priority queue ordered by the tentative label $C(q)$, i.e., the best cost of arrival discovered so far. 
	At each iteration, the vertex with the smallest label is selected for expansion.
	Formally, we express this greedy selection of the next vertex as:
	\begin{equation}
		\mathsf{Select}(Q) = \arg\min_{q \in Q} C(q).
	\end{equation}
	A key property of Dijkstra’s algorithm is that once a vertex is extracted from the priority queue, its label is guaranteed to be final and optimal.
	In other words, each node will be expanded at most once, leading to a more efficient exploration of the graph.
	Because of this, Dijkstra’s algorithm provides the foundation for many widely used planning methods such as $A^*$ and $D^*$.
\end{itemize}

\begin{listing}[ht!]
\begin{tcolorbox}[colback=gray!10, colframe=gray!50, 
    title=Dijkstra's Algorithm, boxrule=0.5mm, arc=0mm]
\begin{minted}[escapeinside=||]{python}
def dijkstra(start_node, edges):
    visited = set()
    Q = [(0, start_node)]
    while Q:
        current_cost, current_node = heapq.heappop(Q)
        if current_node in visited:
            continue
        visited.add(current_node)
        for neighbor, edge_cost in edges[current_node]:
            if neighbor in visited:
                continue
            new_cost = current_cost + edge_cost
            if new_cost < neighbor.cost:
                neighbor.cost = new_cost
                neighbor.parent = current_node.idx
                heapq.heappush(Q, (new_cost, neighbor))
\end{minted}
\end{tcolorbox}
\caption{Python implementation of Dijkstra's algorithm.
See this code used in the context of motion planning in the repository \colorcode{github.com/StanfordASL/pora-exercises} in the notebook \colorcode{ch04/prm\_star.ipynb}.}
\label{alg:dijkstra}
\end{listing}

\begin{figure}[htbp]
	\begin{center}
	\begin{tikzpicture}[>=stealth]
	\def\nodesize{8.5mm} 
	\tikzset{
	  visited/.style  ={draw=black, circle, minimum size=\nodesize, inner sep=0pt, fill=red!30},
	  unvisited/.style={draw=black, dashed, circle, minimum size=\nodesize, inner sep=0pt, fill=gray!10},
	  lbl/.style      ={font=\small},
	  capt/.style     ={font=\scriptsize}
	}
	
	\begin{scope}[xshift=-3.6cm]
	  \node[visited]   (A) at (0,0)                  {$A{\scriptstyle (1)}$};
	  \node[unvisited] (B) at (-1.15,-1.20)          {$B$};
	  \node[visited]   (C) at ( 1.15,-1.20)          {$C{\scriptstyle (2)}$};
	  \node[visited]   (D) at (0,-2.55)              {$D{\scriptstyle (3)}$};
	  \node[visited]   (E) at (0,-3.95)              {$E{\scriptstyle (4)}$};
	
	  \draw[->] (A) -- node[lbl, above left]  {$1$} (B);
	  \draw[->] (A) -- node[lbl, above right] {$5$} (C);
	  \draw[->] (B) -- node[lbl, left]        {$1$} (D);
	  \draw[->] (C) -- node[lbl, right]       {$1$} (D);
	  \draw[->] (D) -- node[lbl, right]       {$1$} (E);
	
	  \node[capt] at ([yshift=6pt]A.north) {(a) DFS (stack)};
	\end{scope}
	
	\begin{scope}
	  \node[visited] (A2)  at (0,0)         {$A{\scriptstyle (1)}$};
	  \node[visited] (B2)  at (-1.15,-1.20) {$B{\scriptstyle (2)}$};
	  \node[visited] (C2)  at ( 1.15,-1.20) {$C{\scriptstyle (3)}$};
	  \node[visited] (D2)  at (0,-2.55)     {$D{\scriptstyle (4)}$};
	  \node[visited] (E2)  at (0,-3.95)     {$E{\scriptstyle (5)}$};
	
	  \draw[->] (A2) -- node[lbl, above left]  {$1$} (B2);
	  \draw[->] (A2) -- node[lbl, above right] {$5$} (C2);
	  \draw[->] (B2) -- node[lbl, left]        {$1$} (D2);
	  \draw[->] (C2) -- node[lbl, right]       {$1$} (D2);
	  \draw[->] (D2) -- node[lbl, right]       {$1$} (E2);
	
	  \node[capt] at ([yshift=6pt]A2.north) {(b) BFS (queue)};
	\end{scope}
	
	\begin{scope}[xshift=3.6cm]
	  \node[visited]   (A3) at (0,0)         {$A{\scriptstyle (1)}$};
	  \node[visited]   (B3) at (-1.15,-1.20) {$B{\scriptstyle (2)}$};
	  \node[unvisited] (C3) at ( 1.15,-1.20) {$C$};
	  \node[visited]   (D3) at (0,-2.55)     {$D{\scriptstyle (3)}$};
	  \node[visited]   (E3) at (0,-3.95)     {$E{\scriptstyle (4)}$};
	
	  \draw[->] (A3) -- node[lbl, above left]  {$1$} (B3);
	  \draw[->] (A3) -- node[lbl, above right] {$5$} (C3);
	  \draw[->] (B3) -- node[lbl, left]        {$1$} (D3);
	  \draw[->] (C3) -- node[lbl, right]       {$1$} (D3);
	  \draw[->] (D3) -- node[lbl, right]       {$1$} (E3);
	
	  \node[capt] at ([yshift=6pt]A3.north) {(c) Dijkstra (priority by $C$)};
	\end{scope}
	\end{tikzpicture}
	\end{center}
	
	\caption{Effect of the selection rule on exploration order. 
	Numbers inside nodes indicate the order of expansion (removal from the frontier). 
	DFS follows the most recently discovered branch and may yield a suboptimal-cost route; 
	BFS explores all nodes at the current depth before proceeding deeper;
	Dijkstra expands in order of lowest cost from the start node.}
	\label{fig:selection-rule-comparison}
\end{figure}
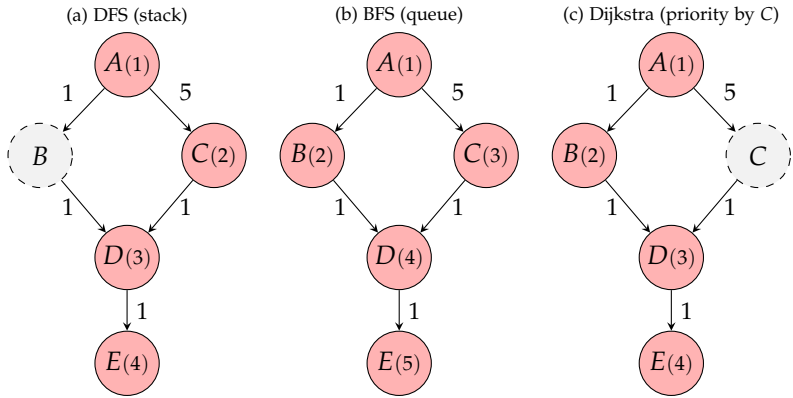

\noindent To make the differences between these search strategies more concrete, consider the simple graph in~\cref{fig:selection-rule-comparison}.
In this example, the start vertex is $A$ and the goal vertex is $E$.
The numbers in parentheses indicate the order of expansion for each of the three search strategies, and the numbers next to each edge indicate the cost of traversing that edge.
By observing the order of expansion, we can see how each strategy explores the graph differently, with DFS going deep into one branch, BFS exploring all nodes at the current depth before moving deeper, and Dijkstra's algorithm expanding nodes based on the lowest cumulative cost.

\paragraph{Beyond Dijkstra: toward more informed search for motion planning.}
Dijkstra’s algorithm is one of the most widely used methods for graph search, owing to its simplicity and the guarantee that it always finds an optimal path, if one exists\cite{Bertsekas2000}.
In particular, Dijkstra is a \emph{correct} algorithm, in that it always finds a least-cost path from the start node to the goal node, provided such a path exists.
\begin{theorem}[Correctness of a graph search algorithm; \citet{Bertsekas2000}]
If a feasible path exists from $\startnode$ to $\goalnode$, then the algorithm terminates in finite time with $C(\goalnode)$ equal to the optimal cost of traversal, $C^*(\goalnode)$.
\end{theorem}
While this property makes Dijkstra appealing, its exploration strategy may result in significant wasted effort.
By always expanding the frontier node with the smallest accumulated cost of arrival, Dijkstra effectively explores the search space in “cost contours” radiating outward from the start. 
This strategy guarantees optimality for the path that is returned, but may expand many vertices that are irrelevant for reaching the goal, a drawback that becomes especially pronounced in the large and structured graphs typical of motion planning (see~\cref{fig:dijkstra-inefficiency}).

These limitations motivated the development of more informed search strategies that incorporate additional guidance toward the goal.
The most influential among these are the \emph{A* algorithm}, which augments Dijkstra with heuristic estimates of the remaining cost to the goal, and its dynamic extension, the \emph{D* algorithm}, which adapts the search as new information about the environment becomes available.

\begin{marginfigure}
	\usetikzlibrary{patterns}
	\begin{center}
	\begin{tikzpicture}[x=0.35cm,y=0.35cm,>=stealth]
		\tikzstyle{visited}  = [draw=black, circle, rounded corners, minimum width=0.9em, minimum height=0.9em, fill=red!30]
		\tikzstyle{free}     = [draw=black, dashed, circle, rounded corners, minimum width=0.9em, minimum height=0.9em, fill=gray!10]
		\tikzstyle{frontier} = [circle, draw=blue!80, fill=white, minimum size=0.9em, line width=0.5pt, pattern=crosshatch, pattern color=blue!80]
		\tikzstyle{goal}     = [circle, fill=green!70!black, draw=green!60!black, inner sep=1pt]
		\tikzstyle{start}    = [star, star points=5, fill=black, draw=black, inner sep=0.6pt]
		\fill[gray!50] (8,4) rectangle (13,6);   
		\fill[gray!50] (11,1) rectangle (13,4);  
		
		\draw[white] (0,0) rectangle (16,12);

		\foreach \i in {0,...,16}{
			\foreach \j in {0,...,12}{
				\pgfmathtruncatemacro{\inTop}{(\i>=9) && (\i<=15) && (\j>=4) && (\j<=7)}
				\pgfmathtruncatemacro{\inLeg}{(\i>=13) && (\i<=15) && (\j>=1) && (\j<=4)}
				\ifnum\inTop=0\relax
				\ifnum\inLeg=0\relax
					\pgfmathsetmacro{\dist}{veclen(\i-1,\j-1)}
					\pgfmathparse{\dist <= 6.2 ? 1 : 0}
					\ifnum\pgfmathresult=1\relax
						\node[visited] at (\i,\j) {};
					\else
						\pgfmathparse{\dist <= 7.0 ? 1 : 0}
						\ifnum\pgfmathresult=1\relax
							\node[frontier] at (\i,\j) {};
						\fi
					\fi
				\fi
				\fi
			}
		}
		
		\node[start] at (0,0) {};
		\node[anchor=west] at (-1.8,0) {\footnotesize $\startnode$};
		\node[goal]  at (12,9) {};
		\node[anchor=west] at (12.5,9) {\footnotesize $\goalnode$};
		\node[frontier] at (0,8) {};
		\node[frontier] at (2,8) {};
		\node[frontier] at (8,0) {};
		\node[frontier] at (8,2) {};
	\end{tikzpicture}
	\end{center}
	\caption{Dijkstra's expansion in grid-based motion planning. 
	The algorithm explores from $\startnode$ by selecting from the frontier (crosshatch-filled nodes) the node with the lowest cost of arrival (the solid-filled nodes have been previously visited).
	By doing so, the exploration process may include regions that do not help reach $\goalnode$.}
	\label{fig:dijkstra-inefficiency}
\end{marginfigure}

\paragraph{The A* algorithm.}
The $A^*$ algorithm improves upon Dijkstra’s selection rule by augmenting the cost of arrival with a heuristic estimate of the cost-to-go, i.e., the cost from the current node to the goal.
Formally, instead of expanding the vertex that minimizes $C(q)$, $A^*$ expands the vertex that minimizes:
\begin{equation}
f(q) = C(q) + h(q),
\end{equation}
where $h(q)$ is a heuristic function estimating the optimal remaining cost from $q$ to the goal.
Consequently, the relaxation step for each neighbor $q^\prime$ of $q$ is also strengthened to update $f(q^\prime)$ rather than just the cost of arrival $C(q^\prime)$:
\begin{equation}
	f(q^\prime) \leftarrow \min\big(f(q^\prime), C(q) + c(q,q^\prime) + h(q^\prime)\big).
\end{equation}
As long as $h(\cdot)$ is \emph{admissible} (i.e., it never overestimates the true cost-to-go), $A^*$ is guaranteed to return an optimal path. 
Intuitively, underestimating the cost-to-go ensures that nodes are not prematurely discarded, which could otherwise lead to suboptimal solutions.
In practice, heuristics such as the Euclidean or Manhattan distance to the goal often reduce the number of vertices explored, since the search is biased toward the target rather than expanding uniformly in all directions. 
This makes $A^*$ one of the most widely used graph search algorithms for the purposes of motion planning.

\paragraph{The D* algorithm.}
In many robotic applications, the environment is only partially known in advance, and new information (such as previously unseen obstacles) may be discovered during execution. 
Recomputing an $A^*$ search from scratch each time the map changes can be computationally challenging. 
The $D^*$ algorithm (\emph{Dynamic A*}) addresses this challenge by incrementally repairing the solution when changes are detected. 
Rather than discarding the existing search tree, $D^*$ efficiently updates cost labels and frontier priorities, reusing past computations whenever possible. 
This makes it particularly suitable for autonomous navigation in dynamic or uncertain environments, where the robot must adapt its plan online as new information is gathered. 
For an in-depth treatment of $D^*$, we refer the reader to \citet{Stentz1995}.

\paragraph{Pros and cons of grid-based planning.}
Grid-based planning methods offer several appealing advantages that explain their long-standing popularity in robotics. 
Perhaps the most important benefit is their \emph{simplicity}: the underlying idea of discretizing the configuration space into cells and treating planning as a graph search problem is straightforward to implement and reason about. 
Once the grid is constructed, classical search algorithms such as DFS, BFS, or Dijkstra can be directly applied. 
This also makes grid-based methods relatively \emph{fast} in certain settings, particularly when the resolution of the grid is well-matched to the scale of the environment and the complexity of the obstacles.

Despite these advantages, grid-based methods come with important limitations. 
A first challenge is that they are inherently \emph{resolution dependent}. 
If the grid resolution is too coarse, narrow passages or fine obstacle boundaries may be missed, and the planner may fail to find a feasible solution even if one exists in the continuous space. 
On the other hand, using a very fine grid increases the computational burden significantly, as the number of grid cells grows rapidly with finer resolution. 
Thus, achieving the right tradeoff between resolution and tractability is nontrivial.

A second drawback is that grid-based planning scales poorly with robot complexity. 
While it is effective for simple 2D robots moving in two-dimensional workspaces (i.e., where configuration space and workspace coincide), the size of the grid grows \emph{exponentially} with the number of degrees of freedom (DOFs) of the robot. 
Moreover, grid-based planning requires an exact characterization of the free space \(\C_{\mathrm{free}}\), which is often difficult to compute in high-dimensional configuration spaces.
These limitations motivate the development of sampling-based methods, which are discussed in~\cref{sec:sampling-based-motion-planning}.
As a result, grid-based methods are primarily used for robots with few DOFs or for simplified planning problems where the dimensionality of the configuration space is deliberately reduced.

In summary, grid-based planning remains a widely used and important approach, particularly for low-dimensional problems. 
However, its reliance on discretization and its exponential scaling with dimension limit its applicability to more complex robotic systems.

\subsection{Combinatorial Motion Planning}
\label{sec:combinatorial-motion-planning}
In this section, we return to the \emph{continuous} formulation of the motion planning problem and explore combinatorial approaches. 
The key idea is to construct an exact representation of the connectivity of the free space $\C_{\mathrm{free}}$ without resorting to approximations.
Instead of discretizing the space arbitrarily, combinatorial methods compute a \emph{roadmap} that captures the essential topological structure of $\C_{\mathrm{free}}$ to enable planning in continuous spaces.

Due to this property, combinatorial motion planning algorithms are referred to as \emph{exact}, as they find paths through the continuous configuration space without resorting to approximations.
Combinatorial planners are also \emph{complete}, meaning they are guaranteed to find the optimal path if one exists, or correctly report failure otherwise.
This is in contrast to grid-based planners, which are only \emph{resolution complete}, guaranteeing a solution only if one exists at the chosen discretization resolution.

However, like grid-based methods, combinatorial planning becomes computationally challenging in high-dimensional configuration spaces, since computing the exact geometry of $\C_{\mathrm{free}}$, and its decomposition into a roadmap, is often prohibitively expensive.
As a result, such approaches are best suited for robots with a small number of degrees of freedom or for low-dimensional planning problems.

\paragraph{The roadmap.}
A roadmap is a graph $\graph = (\node,\edge)$ embedded in the configuration space, where each vertex $q \in \node$ corresponds to a configuration in $\C_{\mathrm{free}}$, and each edge $(q,q^\prime) \in \edge$ represents a continuous, collision-free path between the corresponding configurations. 
Let $\mathcal{S} \subset \C_{\mathrm{free}}$ denote the set of all configurations represented by the vertices in $\graph$. 
For $\graph$ to be a valid roadmap, it must satisfy two key conditions that ensure it accurately represents the structure of $\C_{\mathrm{free}}$:
\begin{enumerate}
    \item \emph{Accessibility:} from any configuration $q \in \C_{\mathrm{free}}$, it must be simple to compute a continuous, collision-free path to any configuration $s \in \mathcal{S}$.
    Typically, $s$ is chosen as the nearest vertex to $q$ (assuming $\C$ is a metric space).
    This condition ensures that every configuration in the free space can be connected to the roadmap without leaving $\C_{\mathrm{free}}$.

    \item \emph{Connectivity-preserving:} using the accessibility condition, it must be possible to connect any two configurations $\startnode, \goalnode \in \C_{\mathrm{free}}$ to some $s_1, s_2$ in the roadmap, respectively.
    The connectivity-preserving property requires that if there exists a continuous, collision-free path between $\startnode$ and $\goalnode$ in $\C_{\mathrm{free}}$, then there must also exist a corresponding path between $s_1$ and $s_2$ in the roadmap $\graph$. 
    In other words, no feasible path in $\C_{\mathrm{free}}$ is lost because the roadmap fails to capture the underlying connectivity of the free space.  
    This property is essential for the completeness of combinatorial planning algorithms.
\end{enumerate}
Once these conditions are satisfied, the roadmap provides an exact representation of the planning problem, where vertices correspond to representative configurations in accessible regions of $\C_{\mathrm{free}}$, and edges encode the connectivity between them.  
Motion planning then reduces to a graph search problem, where the start and goal configurations are connected to the roadmap and a path is searched for in the resulting graph.

While this structure may appear similar to the grid-based approach discussed earlier, the key distinction lies in how the space is represented.
A roadmap is constructed to capture the true geometry and connectivity of $\C_{\mathrm{free}}$, rather than imposing a fixed discretization of $\C$ onto a uniform grid.
In other words, vertices in a roadmap can correspond to \emph{any} configuration within the free space, allowing feasible paths to always be preserved in the graph representation, whereas grid-based methods may fail to represent certain valid paths due to the coarseness of the underlying discretization.
This two-step procedure---roadmap construction followed by graph search---provides a powerful and general framework for motion planning\sidenote{For an in-depth treatment of the concepts of roadmaps, and a formal definition of the requirements for completeness and optimality, see \citet{LaValle2006}.}.

\subsubsection{Cell Decomposition}
As discussed above, combinatorial methods must construct a finite data structure that exactly encodes the planning problem.
One way to achieve this is through \emph{cell decomposition} methods, which partition the free space $\C_{\mathrm{free}}$ into a finite collection of regions, called \emph{cells}, that can be used to construct a roadmap.

A useful way to think about cell decompositions is through three key properties that make them suitable for motion planning:
\begin{enumerate}
	\item \emph{Trivial connectivity within cells:} computing a path from one configuration to another inside a cell must be easy. 
	For instance, if every cell is convex, then any two points in the cell can be connected by a straight-line segment that remains in $\C_{\mathrm{free}}$.
	\item \emph{Adjacency extraction:} it must be straightforward to determine which cells are adjacent to one another, so that a roadmap can be built by connecting neighboring cells.
	\item \emph{Efficient query location:} for a given initial and goal configuration $(\startnode, \goalnode)$, it should be efficient to determine which cells contain them.
\end{enumerate}
When these conditions are satisfied, the motion planning problem reduces to a graph search problem, where vertices correspond to representative configurations in each cell, and edges connect neighboring cells.

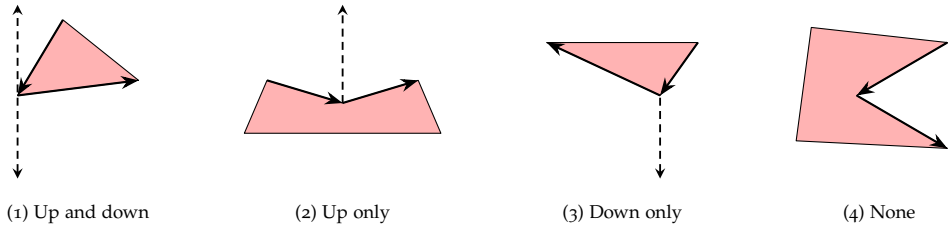
\begin{figure}[htbp]
	\centering
	\begin{tikzpicture}[x=1cm,y=1cm,>=stealth]
	  \tikzset{
		poly/.style={fill=red!30, draw=black, line join=round},
		edgearr/.style={-Stealth, line width=0.8pt},
		guideBoth/.style={<->, densely dashed, line width=0.7pt},
		guideUp/.style={->, densely dashed, line width=0.7pt},
		guideDown/.style={<-, densely dashed, line width=0.7pt},
		caseLbl/.style={font=\scriptsize}
	  }
	
	  \begin{scope}[shift={(0,0)}]
		\filldraw[poly] (-0.2,1.0) -- (0.8,0.2) -- (-0.8,0) -- cycle;
		\draw[edgearr] (-0.2,1.0) -- (-0.8,0);
		\draw[edgearr] (-0.8,0) -- (0.8,0.2);
		\draw[guideBoth] (-0.8,-1.1) -- (-0.8,1.2);
		\node[caseLbl, below=2pt] at (0,-1.25) {(1) Up and down};
	  \end{scope}
	
	  \begin{scope}[shift={(3.5,0)}]
		\filldraw[poly] (-1.0,0.2) -- (0.0,-0.1) -- (1.0,0.2) -- (1.3,-0.5) -- (-1.3,-0.5) -- cycle;
		\draw[edgearr] (-1.0,0.2) -- (0.0,-0.1);
		\draw[edgearr] (0.0,-0.1) -- (1.0,0.2);
		\draw[guideUp] (0,-0.1) -- (0,1.2);
		\node[caseLbl, below=2pt] at (0,-1.25) {(2) Up only};
	  \end{scope}
	
	  \begin{scope}[shift={(7.2,0)}]
		\filldraw[poly] (-1.0,0.7) -- (1.0,0.7) -- (0.5,0.0) -- cycle;
		\draw[edgearr] (1.0,0.7) --(0.5,0.0);
		\draw[edgearr] (0.5,0.0) -- (-1.0,0.7);
		\draw[guideDown] (0.5,-1.1) -- (0.5,0.0);
		\node[caseLbl, below=2pt] at (0,-1.25) {(3) Down only};
	  \end{scope}
	
	  \begin{scope}[shift={(10.6,0)}]
		\filldraw[poly] (-0.9,0.9) -- (0.9,0.7) -- (-0.3,0.0) -- (0.9,-0.7) -- (-1.1, -0.6) -- cycle;
		\draw[edgearr] (0.9,0.7) -- (-0.3,0.0);
		\draw[edgearr] (-0.3,0.0) -- (0.9,-0.7);
		\node[caseLbl, below=2pt] at (0,-1.25) {(4) None};
	  \end{scope}
	\end{tikzpicture}
	\caption{Four general cases of vertical ray extension from a vertex of a polygonal obstacle.}
	\label{fig:vertical-decomposition-cases}
\end{figure}

\paragraph{Vertical cell decomposition.}
A widely used technique for 2D environments is the \emph{vertical decomposition}, also called \emph{trapezoidal decomposition}. 
Suppose the obstacles are polygonal, and let $P$ denote the set of vertices defining $\C_{\mathrm{obs}}$. 
At each vertex $p \in P$, rays are extended vertically upward and downward through $\C_{\mathrm{free}}$ until they intersect either another obstacle or the workspace boundary. 
Depending on the local geometry, four distinct cases arise, corresponding to whether a vertical extension is possible upward, downward, in both directions, or in neither, as illustrated in \cref{fig:vertical-decomposition-cases}.

The result of this process, illustrated in~\cref{fig:cell-decompsition}, is a decomposition of $\C_{\mathrm{free}}$ into a collection of 2-cells (open trapezoids and degenerate triangles) and 1-cells (open vertical line segments forming the boundaries between trapezoids). 

Once the decomposition is available, a roadmap $\graph = (\node,\edge)$ can be constructed. 
For each 2-cell $C_i$, a representative sample point $q_i \in C_i$ is chosen---commonly the centroid, though the exact choice is not critical. 
Each 1-cell is also assigned a sample point.
These points are visualized as black dots in~\cref{fig:cell-decompsition}. 
The roadmap graph is then defined as follows: every cell corresponds to a vertex, and for each 2-cell, edges are added to connect its sample point with the sample points of adjacent 1-cells lying on its boundary.

By construction, both the accessibility and the connectivity conditions are satisfied, i.e., every sample point is accessible via a straight-line path within its cell, and any two adjacent cells are connected by an edge in the roadmap.
Thus, the roadmap provides an exact representation of the planning problem. 
Once the roadmap is constructed, the explicit cell decomposition is no longer needed, and planning reduces to connecting $\startnode$ and $\goalnode$ to the roadmap and performing a graph search.

\begin{example}[2D Cell Decomposition]
	\label{ex:celldecomp}
	Consider the two-dimensional configuration space in \cref{fig:cell-decompsition}. Using vertical decomposition, the free space is partitioned into trapezoids and triangles separated by vertical segments. A roadmap is obtained by placing representative vertices (shown as black dots) inside the cells and along their shared boundaries, and connecting them according to adjacency. To solve a motion planning query, start and goal configurations are connected to their respective cells, and a path is found via a standard graph search algorithm.
	\begin{figure}[ht]
	\centering
	\includegraphics[width=0.74\textwidth]{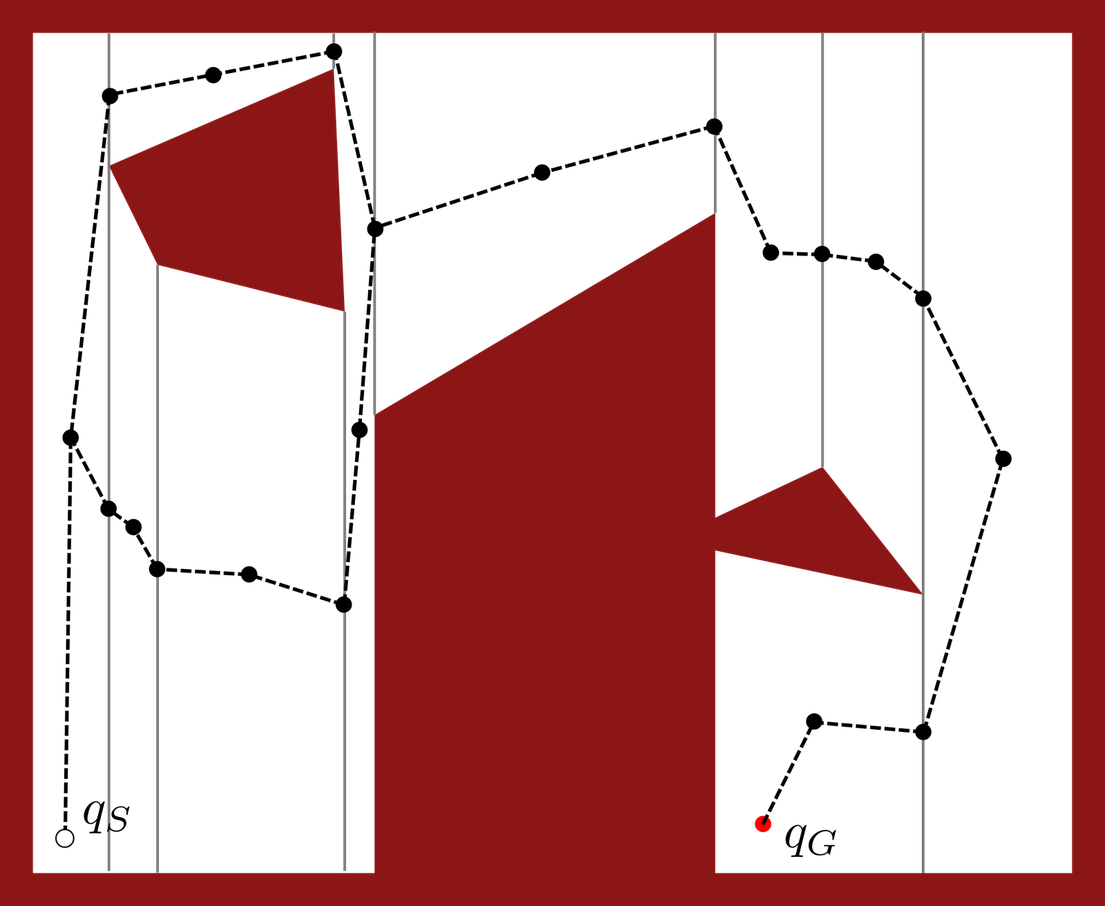}
	\caption{Example of a vertical cell decomposition in two dimensions. The free space $\C_{\mathrm{free}}$ is decomposed into trapezoids, with roadmap vertices (solid dots) placed inside cells and along boundaries. A motion planning query reduces to graph search on this roadmap.}
	\label{fig:cell-decompsition}
	\end{figure}
\end{example}

\subsubsection{Other Roadmap Construction Methods}
Beyond cell decomposition, several other methods have been developed to construct roadmaps, often exploiting geometric properties of the free space.

For instance, \emph{maximum clearance roadmaps} attempt to maintain as much distance as possible from obstacles, effectively following the “skeleton” of the free space. 
A well-known example is the \emph{generalized Voronoi diagram}, where the roadmap consists of points in $\C_{\mathrm{free}}$ equidistant to at least two obstacles. 
Such roadmaps have the advantage of producing paths that maximize safety margins, which is particularly useful for robots operating in tight or uncertain environments.
Another approach is the \emph{shortest path roadmap}, which constructs the roadmap by connecting pairs of points in $\C_{\mathrm{free}}$ with the shortest possible paths that avoid obstacles.
For an in-depth discussion of these and other roadmap construction techniques, we refer the reader to \citet{LaValle2006}.
Each of these roadmap constructions provides a different balance between ease of computation, path quality, and robustness. 
The choice of method depends on the geometry of the robot and environment, as well as the requirements of the task at hand.

\subsection{Sampling-based Motion Planning}
\label{sec:sampling-based-motion-planning}
The limitations of combinatorial and grid-based methods, particularly their computational complexity in high-dimensional spaces, have motivated the development of \emph{sampling-based motion planning} algorithms.
At a high level, these algorithms explicitly avoid the need for an explicit representation of $\C_{\mathrm{free}}$ and $\C_{\mathrm{obs}}$, instead relying on \emph{random sampling} to capture the structure of the configuration space.
To do so, sampling-based methods combine random sampling in $\C$-space with black-box \emph{collision detection} algorithms that can determine whether a configuration or path segment is collision-free (i.e., lies in $\C_{\mathrm{free}}$).
By incrementally connecting such collision–free samples, sampling-based methods are able to build a roadmap or tree structure that captures the feasible connectivity of the free space \emph{without ever explicitly representing it}, in practice ensuring significant computational speed ups.
This paradigm offers several advantages:
\begin{itemize}
	\item \emph{Conceptual simplicity}: algorithms are relatively easy to understand and implement.
	\item \emph{Generality}: the same framework applies to different robots and environments.
	\item \emph{Extensibility}: the methodology can be extended beyond purely geometric settings to handle kinodynamic planning, differential constraints, and uncertainty.
\end{itemize}
At the same time, sampling–based methods have inherent limitations. 
Unlike exact combinatorial approaches, they generally offer weaker guarantees regarding the existence or quality of the resulting solution.
Moreover, it may be challenging to know a priori how many samples are needed to find a solution, or to ensure that the planner will find a solution if one exists.
On the theoretical side, most methods offer \emph{probabilistic} guarantees rather than deterministic ones: they are \emph{probabilistically complete} (i.e., if a feasible path exists, the probability of finding one approaches 1 as the number of samples $n$ tends to infinity) and, for certain variants, \emph{asymptotically optimal} (i.e., if $C_n$ is the cost of the best path found after $n$ samples and $C^\star$ is the optimal cost, then $C_n$ tends to $C^\star$ with probability 1). 
Both properties are asymptotic and probabilistic in nature, as they provide no finite-sample guarantee but only that the probability of failure (or of suboptimality) vanishes as $n$ grows.
These properties will be discussed in more detail in Section~\ref{subsubsec:theoretical-guarantees}.

Traditionally, sampling-based motion planning algorithms fall into two main approaches: \emph{probabilistic roadmaps} (PRMs) and \emph{rapidly-exploring random trees} (RRTs).
PRMs are \emph{multi-query planners}, as their precomputed structures can be reused to answer multiple planning problems in the same environment. 
By contrast, RRTs are \emph{single-query planners}, designed to find a solution for a specific start–goal pair $(\startnode, \goalnode)$ in a given free configuration space $\C_{\mathrm{free}}$.
Despite this distinction, both families of algorithms share two fundamental components. 
The first is a collision detection routine, $\texttt{CollisionFree}(q)$, which determines whether a configuration $q$ lies within the free space $\C_{\mathrm{free}}$. 
The second is a local planner, $\texttt{LocalPlanner}(q, q^\prime)$, which produces a short path segment—often a straight line in configuration space or a dynamically feasible motion—and verifies its validity by checking for collisions.

Below, we provide an overview of PRMs and RRTs, along with their theoretical properties and practical considerations.

\subsubsection{Probabilistic Roadmaps (PRMs)}
\label{subsubsec:prm}
PRMs are among the most influential sampling–based methods for motion planning.
At a high level, PRMs are a multi-query planner that constructs a roadmap in the configuration space by randomly sampling configurations and connecting them to form a graph.
The PRM algorithm consists of two main phases: a \emph{construction phase} and a \emph{query phase}.
During the construction phase, the algorithm samples a set of configurations $\{q_1, q_2, \ldots, q_n\}$ from $\C_{\mathrm{free}}$ using a uniform or biased sampling strategy.
Each sampled configuration is then used to construct a roadmap that encodes the connectivity of the free space.
As a result, once constructed, the roadmap can be queried to solve multiple planning problems in the same environment by simply attaching start and goal configurations to the roadmap and running a graph search.

\noindent Formally, the PRM algorithm can be summarized as follows:
\begin{enumerate}
	\item \emph{Construction Phase}:
	\begin{enumerate} 
		\item Sample $n$ configurations $\{q_1, q_2, \ldots, q_n\}$ from $\C$ and discard those that lie in $\C_{\mathrm{obs}}$ using the collision detection method $\texttt{CollisionFree}(q)$.
		\item Draw an edge between each pair of configurations $q_i$ and $q_j$ according to a chosen connection rule---for example, if they are within a certain distance threshold $r$, i.e., $||q_i - q_j|| \leq r$, \emph{and} if the local planner $\texttt{LocalPlanner}(q_i, q_j)$ finds a collision-free path between them, e.g., a straight line segment in $\C$-space.
	\end{enumerate}
	\item \emph{Query Phase}: given a query $(\startnode, \goalnode) \in \C_{\mathrm{free}}$, connect $\startnode$ and $\goalnode$ to the nearest nodes in the roadmap using the local planner, and then use a graph search algorithm (e.g., Dijkstra's or $A^*$) to find a path from $\startnode$ to $\goalnode$ through the roadmap.
\end{enumerate}

\begin{figure}[t]
	\centering
	 \includegraphics[width=0.65\textwidth]{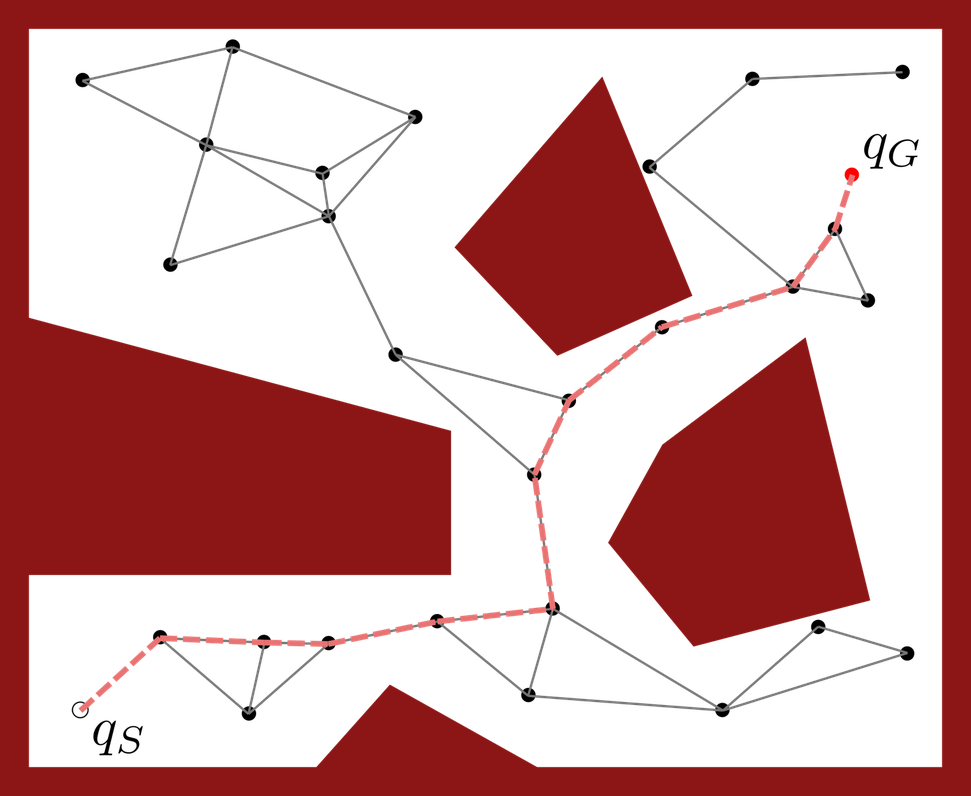}
	\caption{Roadmap generated by the PRM algorithm. 
    Solid dots denote randomly sampled configurations, and the lines connecting these dots represent collision-free connections established between configurations within a predefined connection radius $r$. 
    The initial configuration $\startnode$ and the goal configuration $\goalnode$ are connected through the roadmap, with the resulting solution path highlighted by the dashed red line.
    This path is obtained by applying a graph search algorithm to the constructed roadmap.}
	\label{fig:prm-graph}
  \end{figure}

\cref{fig:prm-graph} illustrates an example of the PRM algorithm in a simple 2D environment.

\paragraph{Design considerations.}
The effectiveness of PRM depends on several key design decisions that shape both its computational performance and its ability to capture the connectivity of the free space. 
A first consideration concerns the sampling strategy. 
The most straightforward approach is to generate configurations uniformly at random in the configuration space and retain only those that lie in $\C_{\mathrm{free}}$. 
While this method provides unbiased coverage, it may perform poorly in environments dominated by narrow passages, since these regions occupy little volume and are therefore rarely sampled. 
To address this limitation, various biased sampling strategies have been proposed, which aim to increase the likelihood of sampling configurations in challenging areas of the configuration space.
More generally, heuristic biasing strategies can direct samples toward regions that are likely to be relevant for specific planning tasks, trading uniformity for effectiveness in complex environments.

Another important design choice is the method used to connect sampled configurations.
Two rules are most common: the fixed-radius rule, where each vertex attempts to connect to all samples within a ball of radius $r$, and the $k$-nearest neighbor rule, where each vertex connects to its $k$ closest samples. 
As will be discussed in the remainder of this chapter, both connection rules are supported by theoretical results that tie the values of $r$ and $k$ to the number of samples and the dimensionality of the space (see, e.g., \citet{KaramanFrazzoli2011}). 
These results ensure that, as the number of samples grows, the roadmap becomes sufficiently connected to capture the topology of $\C_{\mathrm{free}}$ without introducing an excessive number of samples

\paragraph{Characteristics of PRMs.}
The probabilistic roadmap framework is particularly well suited to environments that remain static across multiple planning queries. 
Since the construction of the roadmap can be computationally demanding—dominated by nearest-neighbor queries and collision checking—it is most effective when the investment in preprocessing can be amortized over many queries posed in the same workspace. 
In these contexts, the roadmap serves as a reusable data structure that compactly encodes the connectivity of the free space, enabling queries to be answered quickly with standard graph search algorithms.

For these reasons, PRM has become a method of choice in domains where multiple planning queries must be solved in high-dimensional but largely static environments. 
Its strength lies not in producing a single solution quickly, but in building a reusable structure that captures the topology of the free space and can then be reused to efficiently solve many queries.

Overall, PRM-like motion planning algorithms are widely recognized to find ``good'' paths in practice, even in high-dimensional configuration spaces.
However, this may require a large number of expensive collision checks, thus incurring significant computational costs.

\subsubsection{Rapidly-Exploring Random Trees (RRTs)}
\label{subsubsec:rrt}
RRT is a foundational sampling–based algorithm designed for \emph{single-query} motion planning. 
In contrast to the precomputation strategy of PRMs, RRT incrementally grows a tree\sidenote{A tree is a special type of graph that is connected and acyclic. In other words, there is exactly one path between any two vertices in a tree.}---denoted $\mathcal{T} = (\node, \edge)$---rooted at the initial configuration $\startnode$, seeking to connect it to a specified goal configuration $\goalnode$.
As a result, RRT is particularly well suited for scenarios in which only a single query needs to be solved, such as when the environment dynamically changes.

RRT proceeds iteratively, where each iteration consists of the following steps:
\begin{enumerate}
	\item Sample a random configuration $q_{\mathrm{rand}}$ from the configuration space $\C$.
	\item Find the nearest vertex $q_{\mathrm{near}}$ in the tree $\mathcal{T}$ to the sampled configuration $q_{\mathrm{rand}}$ under an appropriate distance metric $d(\cdot,\cdot)$ (e.g., Euclidean distance).
	\item Generate a new configuration $q_{\mathrm{new}}$ by moving from $q_{\mathrm{near}}$ toward $q_{\mathrm{rand}}$, ensuring that the motion from $q_{\mathrm{near}}$ to $q_{\mathrm{new}}$ is collision-free.
	\item Update the tree $\mathcal{T}$ by adding $q_{\mathrm{new}}$ as a new vertex and connecting it to $q_{\mathrm{near}}$ with an edge $(q_{\mathrm{near}}, q_{\mathrm{new}})$.
\end{enumerate}
The algorithm continues expanding the tree until a vertex is added that is sufficiently close to the goal configuration $\goalnode$\sidenote{This proximity can be verified, for example, by attempting to connect the newly added vertex to the goal and checking whether the connection is collision-free. If successful, the algorithm terminates.}.
At that point, a path from $\startnode$ to $\goalnode$ can be extracted by backtracking through the tree.
If after a budget of $N$ iterations no feasible connection to the goal is found, the algorithm terminates with failure.

\begin{figure}[t]
	\centering
	\includegraphics[width=0.65\textwidth]{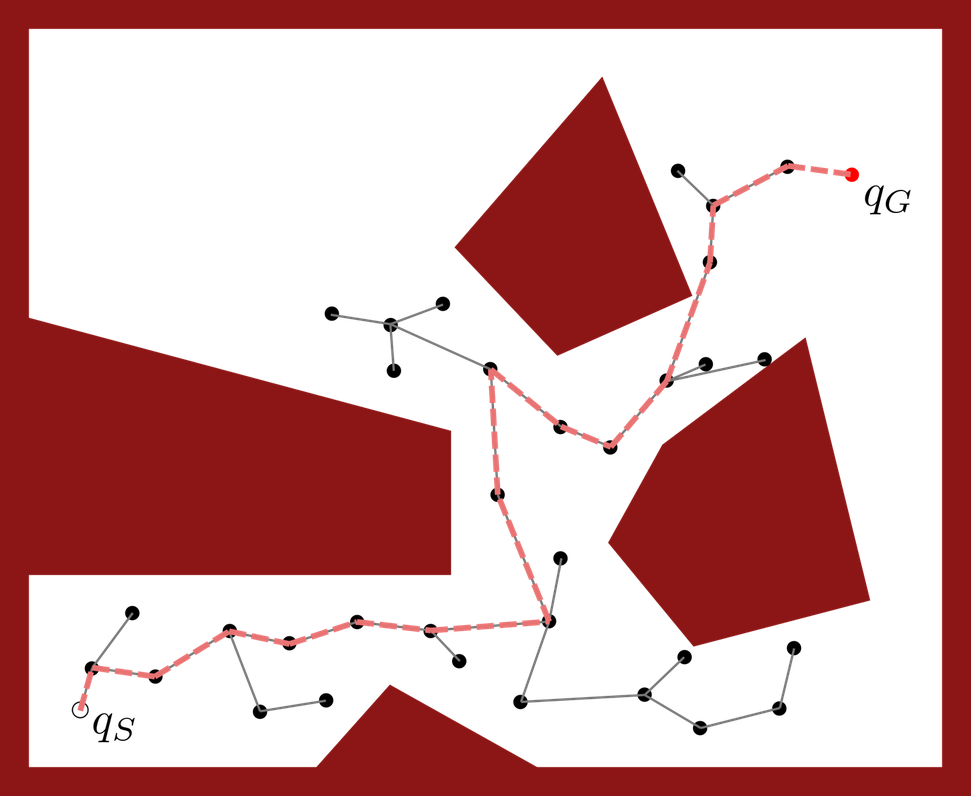}
	\caption{Exploration tree generated by the RRT algorithm. 
	Solid dots indicate tree vertices added over iterations. 
	At each step the tree is extended from the nearest vertex toward a random sample.}
	\label{fig:rrt-graph}
  \end{figure}

\cref{fig:rrt-graph} illustrates an example of the RRT algorithm in a simple 2D environment.

\begin{marginfigure}
	\includegraphics[width=\textwidth]{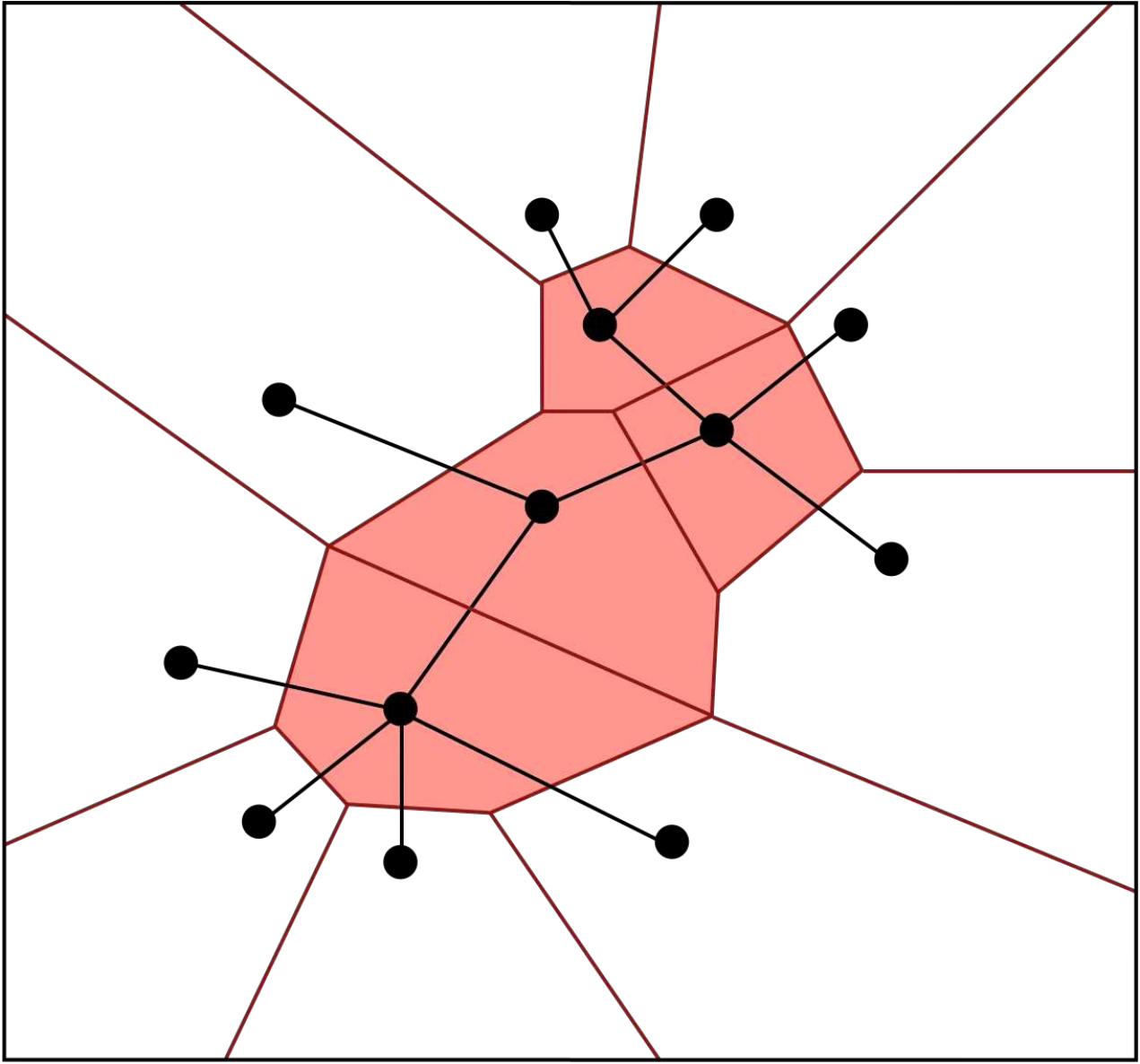}
	\caption{Voronoi bias in RRT.}
	\label{fig:voronoi-bias}
\end{marginfigure}

\paragraph{Voronoi bias.}
A key insight into RRT’s efficiency comes from interpreting its sampling dynamics through the lens of Voronoi diagrams (\cref{fig:voronoi-bias}).
Each vertex in the tree $\mathcal{T}$ defines a Voronoi cell in configuration space, consisting of all points closer to that vertex than to any other. 
Since random samples are uniformly distributed, the probability of expanding a given vertex is proportional to the volume of its Voronoi cell. 
Vertices on the frontier of the tree tend to own large cells, and therefore are more likely to be selected for expansion. 
This \emph{Voronoi bias} implicitly drives the tree outward, rapidly exploring uncovered regions without requiring any explicit mechanism for frontier selection.

Overall, RRT provides a powerful framework for single-query motion planning. 
Specifically, RRT-like motion planning algorithms are widely used in practice due to their ability to find feasible paths quickly in high-dimensional spaces.
However, the paths produced by RRT are often suboptimal.
Its extensions, particularly RRT$^*$ and its variants, combine RRT's exploratory power with provable guarantees on optimality, making them widely used in practical applications.
Importantly, RRT (as well as PRM) does not require an explicit characterization of $\C_{\mathrm{free}}$, which is key to its scalability to higher-dimensional $\C$ spaces.

\subsubsection{Theoretical Guarantees}
\label{subsubsec:theoretical-guarantees}
Both PRM and RRT algorithms come with theoretical guarantees that underpin their effectiveness in motion planning tasks.
Specifically, we consider two key properties which formalize the sense in which planners like PRM, RRT, and their related extensions PRM$^*$ and RRT$^*$ succeed in the limit of large sampling budgets: \emph{probabilistic completeness} and \emph{asymptotic optimality}.

\paragraph{Probabilistic completeness.}
A planner is said to be probabilistically complete if, whenever a feasible path exists, the \emph{probability} that the algorithm fails to find one approaches zero as the number of samples $n$ tends to infinity.
Both PRM and RRT satisfy probabilistic completeness under reasonable assumptions on the configuration space and the sampling process\sidenote{Intuitively, the sampling distribution must assign nonzero probability to every open subset of the free space.}.

While an in-depth treatment of probabilistic completeness is beyond the scope of this chapter, we refer the reader to \textcite{KavrakiSvestkaEtAl1996} and \textcite{LaValle1998}, which establish these guarantees for PRM, RRT, and related algorithms.

\paragraph{Asymptotic optimality.}
A planner is asymptotically optimal if, as the number of samples $n$ tends to infinity, the \emph{probability} that cost of the best path found by the algorithm converges to the optimal cost approaches one.
Achieving asymptotic optimality requires careful control of the connectivity of the underlying random graph.
If connections are too sparse, the graph may fail to capture near-optimal paths.
If connections are too dense, computational costs may increase significantly.
Thus, guarantees of asymptotic optimality rely on choosing a connection radius that decreases at an appropriate rate relative to the number of samples to balance these competing effects.

\textcite{KaramanFrazzoli2011} provided the first rigorous study of asymptotic optimality for sampling-based planners.
In particular, they proved the following result for PRM$^*$:
\begin{theorem}
	If the connection radius $r(n)$ satisfies:
	$$
	r(n) \geq \gamma \left( \frac{\log n}{n} \right)^{1/d},
	$$
	where $d$ is the dimension of the configuration space and $\gamma$ is a constant that depends only on $d$ and $\C_\mathrm{free}$, then the cost of the best path returned by $PRM^*$ after $n$ samples converges to the optimal cost \emph{with probability one} as $n\to\infty$.
\end{theorem}

\textcite{KaramanFrazzoli2011} also introduced an asymptotically optimal variant of RRT, called RRT$^*$, which incorporates a rewiring step to improve path quality over time.

\begin{example}[PRM* in 2D Workspace]
Explore \colorcode{ch04/prm\_star.ipynb} in the repository \colorcode{github.com/StanfordASL/pora-exercises} for an example implementation of the PRM* algorithm that is asymptotically optimal as the number of samples increases.
In this example, we consider a simple 2D workspace with some obstacles, and try to plan the shortest path from a start to goal position.
Play around with the number of nodes in the PRM to see the optimality and execution time trade-off.
\end{example}

\subsubsection{Fast Marching Tree Algorithm (FMT$^*$)}
\label{subsubsec:fmt}
The \emph{Fast Marching Tree} (FMT$^*$) algorithm, introduced by~\citet{JansonSchmerlingEtAl2015}, is a more recent addition to the family of sampling–based motion planners. 
Its key contribution is to achieve asymptotic optimality---like PRM$^*$ and RRT$^*$---while performing dramatically fewer collision checks, often the computational bottleneck in motion planning.

\paragraph{High-level description.}
FMT$^*$ operates on a fixed set of $n$ collision–free samples drawn from $\C_{\mathrm{free}}$, together with the initial condition $\startnode$, i.e., $V = \{\startnode\} \cup \{q_1, \dots, q_n\} \subset \C_{\mathrm{free}}$.
Rather than attempting to construct a global roadmap (as in PRM$^*$) or expand a tree through random sampling (as in RRT$^*$), FMT$^*$ performs graph construction and graph search \emph{concurrently}.
At each iteration, it expands the current tree outward in \emph{cost–to–arrive space}, always advancing from the lowest–cost node in the tree and connecting it to nearby unvisited samples via the best available one–step connection.
This approach is reminiscent of Dijkstra’s algorithm and of the Fast Marching Method for solving Eikonal equations~\cite{Sethian1996}, in that expanded nodes in the tree never need to be revisited.

At any point during the algorithm, the sample set $V$ is partitioned into three disjoint subsets: (i) $V_{\mathrm{open}}$, nodes currently part of the tree and eligible for expansion; (ii) $V_{\mathrm{unvisited}}$, samples not yet connected to the tree; and (iii) $V_{\mathrm{closed}}$, nodes already expanded and thus excluded from further connections.
Initially, $V_{\mathrm{open}} = \{\startnode\}, V_{\mathrm{unvisited}} = V \setminus \{\startnode\}, \text{and} \, V_{\mathrm{closed}} = \emptyset$.
At each iteration, the algorithm extracts the lowest–cost node $q^\prime$ in $V_{\mathrm{open}}$ and considers all unvisited neighbors $q \in V_{\mathrm{unvisited}}$ within a radius $r_n$ from $q^\prime$.
For each such $q$, it computes the best parent $q_p \in V_{\mathrm{open}}$ minimizing the cost:
\begin{equation}
q_p^\ast = \argmin_{q_p \in \mathcal{N}_{V_{\mathrm{open}}}(q)} \Big[ C(q_p) + c(q_p, q) \Big],
\end{equation}
where $C(q_p)$ is the cost–to–arrive at $q_p$, $c(q_p, q)$ is the local path cost, and $\mathcal{N}_{V_{\mathrm{open}}}(q)$ denotes the set of neighbors of $q$ among the samples in $V_{\mathrm{open}}$.
Crucially, only this best candidate edge $(q_p^\ast, q)$ is collision–checked.
If the edge is feasible, $q$ is added to the tree and moved from $V_{\mathrm{unvisited}}$ to $V_{\mathrm{open}}$.
After all neighbors of $q^\prime$ have been processed, $q^\prime$ is moved from $V_{\mathrm{open}}$ to $V_{\mathrm{closed}}$.
The process repeats until either a node in the goal region is added to the tree or $V_{\mathrm{open}}$ becomes empty.

FMT$^*$’s efficiency stems from its \emph{lazy} evaluation strategy, such that only one edge per candidate node is collision–checked, instead of all possible ones as in PRM$^*$.
While this laziness may occasionally discard a better edge, such events occur with vanishing probability as the number of samples increases. 
Indeed, FMT$^*$ is asymptotically optimal and, as $n$ tends to infinity, the cost of the best path returned converges to the optimal cost \emph{with probability one}.
The efficiency gain is substantial, and compared to PRM$^*$, the ratio of collision checks performed by FMT$^*$ converges to zero, making it particularly well suited to high–dimensional problems where collision checking dominates computation.

In this sense, FMT$^*$ occupies a natural middle ground between PRM$^*$ and RRT$^*$. 
Like PRM$^*$, it begins from a fixed set of pre–sampled configurations, but instead of forming a global roadmap, it incrementally connects nodes into a tree rooted at the start. 
Like RRT$^*$, it constructs a single tree, but its expansion enables a more effective exploration and often improved solution quality.


\subsubsection{Kinodynamic Planning}
\label{subsubsec:kinodynamic-planning}
The geometric motion planning algorithms introduced so far assume no differential constraints on the robot's motion.
This assumption simplifies the planning task, as it allows the problem to be decomposed into two steps: first, computing a collision-free path that ignores dynamics, and then smoothing or reparameterizing this path so that it can be followed by the robot, as discussed in~\cref{ch:closedloop}.
However, purely geometric plans are not always straightforward to refine into dynamically feasible, optimized trajectories, often requiring downstream trajectory optimizers or controllers to perform aggressive corrections.

The \emph{kinodynamic motion planning problem} extends geometric motion planning by explicitly accounting for a robot’s dynamics. 
Formally, let $\X \subseteq \mathbb{R}^n$ and $\U \subseteq \mathbb{R}^m$ denote the state and control spaces of a robotic system, respectively, and consider dynamics of the form:
\begin{equation}
	\dot{\x}(t) \;=\; f(\x(t),\u(t)), \qquad \x(t) \in \X,\; \u(t) \in \U.
\end{equation}
As in geometric planning, the robot’s configuration $q$ can be derived from the full dynamic state $\x$, which encodes both position and the information needed for collision checking.
The kinodynamic planning problem then seeks to drive the robot from an initial state $\x_{\mathrm{init}}$ to a goal region $\X_{\mathrm{goal}}$ while satisfying both collision avoidance and the system’s dynamical constraints.

By embedding kinematic and dynamic constraints at the planning stage, the planner produces nominal motions that are dynamically feasible by construction. 
This, in turn, provides better initializations for trajectory optimization and reduces the burden on the tracking controller, ultimately improving the overall system performance at execution time.

\begin{figure}[t]
	\centering
	\includegraphics[width=0.7\textwidth]{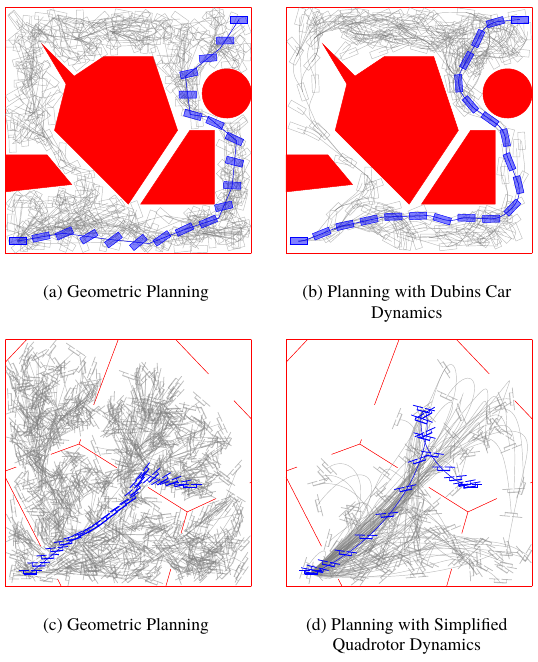}
	\caption{Comparison of geometric and kinodynamic planning. 
	Obstacles are the solid polygons/circles (top figures) and solid straight line segments (bottom figures), trees of motion considered by the sampling-based planner are shown throughout the configuration space, and solution trajectories are highlighted.
	In each row, the robots start and end in the same configuration, yet the addition of differential constraints in the right-hand figures greatly restricts their possible motions, making the planning problem significantly more challenging.
	In the top row, enforcing the simple car dynamics prevents the vehicle from sliding laterally (b), resulting in a more complex trajectory compared to the geometric case (a).
	In the bottom row, incorporating momentum through simplified quadrotor dynamics yields a smoother but more circuitous trajectory (d) compared to the geometric case (c), reflecting the constraints imposed by the vehicle’s inertia.
	[Figure from \citet{SchmerlingPavone2019}].}
	\label{fig:kinodynamic-vs-geometric}
\end{figure}

This formulation highlights the fundamental distinction between geometric and kinodynamic planning.
In geometric planning, feasible paths are arbitrary collision-free curves in configuration space.
In contrast, kinodynamic planning restricts admissible trajectories to those consistent with the system’s equations of motion.
For example, a wheeled robot cannot move sideways, and a quadrotor cannot instantaneously stop due to inertia.
These differential constraints drastically alter the search space, as illustrated in \cref{fig:kinodynamic-vs-geometric}, where adding dynamics constraints significantly restricts the feasible set of trajectories.

\paragraph{Forward-propagation-based algorithms.}
A practical way to incorporate dynamics into sampling-based planners is through \emph{forward propagation}.
Rather than directly connecting sampled states, the planner samples a control input $\u \in \U$ and a duration $t$, then integrates the system dynamics forward in time to produce a new state.
For example, the kinodynamic extension of RRT proceeds by: (i) sampling a random state and finding its nearest neighbor $q_{\mathrm{near}}$ in the tree, (ii) sampling a random control $\u \in \U$ and propagation time $t$, and (iii) simulating the system forward from $q_{\mathrm{near}}$ to obtain a new state. 
If the resulting trajectory segment is collision-free, it is added to the tree; otherwise, it is discarded.
Notably, under mild regularity assumptions on $f(\x(t),\u(t))$, variants of RRT with forward propagation have been shown to be probabilistically complete.

Kinodynamic planning thus unifies obstacle avoidance with differential constraints in a single framework, which is particularly important for robotics, where dynamical considerations highly restrict the set of feasible paths, for example in high-speed ground vehicles, aerial robots, or systems with significant momentum.

\subsubsection{Should Probabilistic Planners be Probabilistic?}
\label{subsubsec:deterministic-sampling}
A natural question is whether the success of sampling-based algorithms depends fundamentally on the randomness of the sampling process.
In particular, would the theoretical guarantees and practical performance of planners such as PRM$^*$ or FMT$^*$ still hold if the algorithms were \emph{de-randomized}, i.e., run on deterministic samples?

This question is important for several reasons. 
Deterministic sampling sequences could substantially ease certification in safety-critical systems, since they remove probabilistic uncertainty. 
They would also allow planners to exploit offline computation more effectively, and simplify certain operations such as nearest-neighbor search.

Recent results show that the answer is affirmative, with carefully chosen low-dispersion deterministic sampling sequences, one can retain asymptotic optimality and completeness guarantees~\cite{JansonIchterEtAl2018}.
Thus, randomness is not strictly necessary for either correctness or performance. 
Probabilistic sampling provides convenience and generality, but deterministic low-dispersion sequences can yield equally strong guarantees while offering advantages in predictability, efficiency, and ease of certification.

\subsection{Potential Field Methods}
\label{sec:potential-field-methods}
\begin{figure}[tb]
	\centering
	\includegraphics[width=0.8\textwidth]{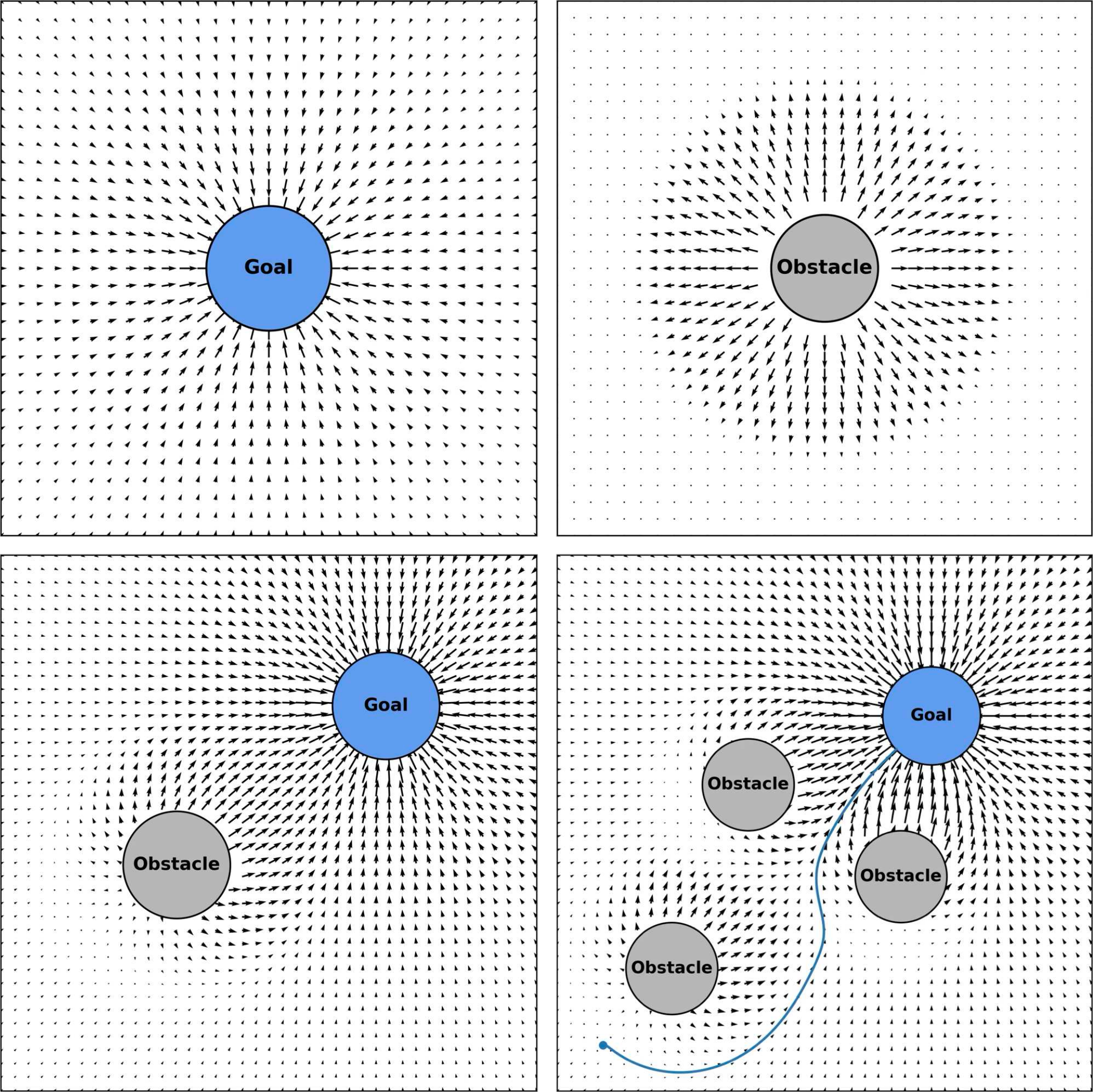}
	\caption{A potential field in a 2D configuration space. 
	The attractive potential (top left) creates a basin of attraction at the goal. 
	The repulsive potential (top right) creates barriers around obstacles. 
	The combined potential field (bottom left) defines a surface whose negative gradient guides the robot from the start to the goal, resulting in a smooth trajectory (bottom right).}
	\label{fig:potential-field-methods}
\end{figure}

The planning methods described in the previous sections focus on capturing the global connectivity of the robot’s free space in a compact graph representation, which is then searched for a feasible path.
In contrast, the approach introduced in this section is based on a fundamentally different idea.
Potential field methods~\cite{Khatib1986} frame the motion planning problem using a physics-inspired analogy, where the robot is treated as a particle moving under the influence of an \emph{artificial potential field} defined over the configuration space.
This field is constructed to guide the robot toward its goal while simultaneously pushing it away from obstacles, resulting in a smooth, continuous path, as illustrated in \cref{fig:potential-field-methods}.

The core of this approach is the design of an artificial potential function $U(q): \C \to \R$, which is typically composed of two components: an \emph{attractive potential} $U_{\mathrm{att}}(q)$ that pulls the robot toward the goal configuration $q_{\mathrm{goal}}$, and a \emph{repulsive potential} $U_{\mathrm{rep}}(q)$ that pushes it away from obstacles:
\begin{equation}
	U(q) = U_{\mathrm{att}}(q) + U_{\mathrm{rep}}(q).
\end{equation}
The motion of the robot is then determined by following the negative gradient of this total potential $-\nabla U(q)$, which represents the direction of steepest descent in the potential field.
Specifically, the artificial force acting on the robot at configuration $q$ is given by:
\begin{equation}
	\begin{split}
	F(q) = -\nabla U(q) & = -\nabla U_{\mathrm{att}}(q) - \nabla U_{\mathrm{rep}}(q) \\
	& = -F_{\mathrm{att}}(q) - F_{\mathrm{rep}}(q),
	\end{split}
\end{equation}
where $F_{\mathrm{att}}(q)$ and $F_{\mathrm{rep}}(q)$ are the \emph{attractive} and \emph{repulsive forces}, respectively.
A path is generated by starting at the initial configuration $\startnode$ and iteratively taking small steps in the direction of this force, effectively performing gradient descent on the potential surface until reaching the goal configuration $\goalnode$.

Below, we describe common choices for the attractive and repulsive potential functions and refer the reader to \citet{Latombe1991} for a more comprehensive treatment of potential field methods.

\paragraph{Attractive potential.}
The attractive potential is designed to create a basin of attraction around the goal configuration---ideally, a unique global minimum---encouraging the robot to move toward it.
A common choice for the attractive potential is a parabolic function of the form:
\begin{equation}
	U_{\mathrm{att}}(q) = \frac{1}{2} k_{\mathrm{att}} \, \rho^2(q, \goalnode),
\end{equation}
where $k_{\mathrm{att}}$ is a positive scaling factor, and $\rho(q, \goalnode)$ denotes a distance metric (e.g., Euclidean distance) between the current configuration $q$ and the goal configuration $\goalnode$.
The function $U_{\mathrm{att}}(q)$ is thus positive or null, and attains its minimum value at $\goalnode$, where $U_{\mathrm{att}}(\goalnode) = 0$.

The resulting attractive force (pointing toward the goal) is given by:
\begin{equation}
	\begin{split}
		F_{\mathrm{att}}(q) & \coloneqq -\nabla U_{\mathrm{att}}(q) \\
		& = -k_{\mathrm{att}} \, \rho(q, \goalnode) \nabla \rho(q, \goalnode) \\
		& = -k_{\mathrm{att}} (q - \goalnode), \\
		& \text{where } \rho(q, \goalnode) = ||q - \goalnode|| \text{ and } \nabla \rho(q, \goalnode) = \frac{q - \goalnode}{||q - \goalnode||},
	\end{split}
\end{equation}
which is a linear function of the distance to the goal.

\paragraph{Repulsive potential.}
The repulsive potential is designed to create a potential barrier around obstacles, preventing the robot from colliding with them.
Moreover, it should have a limited range of influence, affecting the robot only when it is within a certain distance from an obstacle.
One way to achieve these properties is to define the repulsive potential as:
\begin{equation}
	U_{\mathrm{rep}}(q) = 
	\begin{cases}
		\frac{1}{2} k_{\mathrm{rep}} \left( \frac{1}{\rho(q, \C_{\mathrm{obs}})} - \frac{1}{\rho_0} \right)^2 & \text{if } \rho(q, \C_{\mathrm{obs}}) \leq \rho_0, \\
		0 & \text{if } \rho(q, \C_{\mathrm{obs}}) > \rho_0,
	\end{cases}
\end{equation}
where $k_{\mathrm{rep}}$ is a positive scaling factor, $\rho(q, \C_{\mathrm{obs}})$ denotes the distance from the configuration $q$ to the nearest obstacle in the configuration space, and $\rho_0$ is a threshold distance beyond which the repulsive potential has no effect, also referred to as the \emph{distance of influence}.
Formally, the distance to the nearest obstacle is defined as:
\begin{equation}
	\rho(q, \C_{\mathrm{obs}}) = \min_{q^\prime \in \C_{\mathrm{obs}}} \rho(q, q^\prime),
\end{equation}
where $\C_{\mathrm{obs}}$ is the set of configurations that result in collisions with obstacles.

If $\C_{\mathrm{obs}}$ is a convex region, $\rho(q, \C_{\mathrm{obs}})$ is differentiable everywhere in $\C_{\mathrm{free}}$, and the resulting repulsive force is given by:
\begin{equation}
	\begin{split}
	F_{\mathrm{rep}}(q) & \coloneqq -\nabla U_{\mathrm{rep}}(q) \\
	& =
	\begin{cases}
		k_{\mathrm{rep}} \left( \frac{1}{\rho(q, \C_{\mathrm{obs}})} - \frac{1}{\rho_0} \right) \frac{1}{\rho^2(q, \C_{\mathrm{obs}})} \nabla \rho(q, \C_{\mathrm{obs}}) & \text{if } \rho(q, \C_{\mathrm{obs}}) \leq \rho_0, \\
		0 & \text{if } \rho(q, \C_{\mathrm{obs}}) > \rho_0.
	\end{cases}
	\end{split}
\end{equation}
Specifically, let $q_{\mathrm{obs}}^\ast$ be the point on the boundary of the obstacle closest to $q$, i.e., $q_{\mathrm{obs}}^\ast = \arg\min_{q^\prime \in \C_{\mathrm{obs}}} \rho(q, q^\prime)$.
Then, the gradient $\nabla \rho(q, \C_{\mathrm{obs}})$ is a unit vector pointing away from $\C_{\mathrm{obs}}$ and supported by the line segment connecting $q_{\mathrm{obs}}^\ast$ to $q$.

\paragraph{Advantages and disadvantages.}
The primary advantage of potential field methods is their simplicity and computational efficiency.
These methods were originally developed as an online collision avoidance strategy for mobile robots\cite{Khatib1986}, applicable in dynamic environments where obstacles may not be known in advance.
Instead of performing a complex search over a graph or constructing a roadmap, path generation simply involves evaluating the potential function and its gradient at the robot's current configuration.

However, since potential field planners essentially act as gradient descent algorithms on the potential function, they are susceptible to local minima.
Specifically, the planner can become trapped in regions where the attractive and repulsive forces balance out, resulting in a net force of zero at a configuration that is not the goal.
This issue commonly occurs in environments with concave obstacles (e.g., a U-shaped trap) or in narrow corridors where the repulsive forces from opposing walls can cancel the attractive pull toward the goal.
Because of this, standard potential field planners are not \emph{complete} and may fail to find a path even when one exists.

Several techniques have been proposed to address the local minima problem.
One approach is to introduce a random ``jiggle'' (e.g., a small random perturbation to the robot's configuration) to escape the basin of a local minimum.
A more theoretically grounded solution involves constructing special potential fields called \emph{navigation functions}, which are provably free of local minima except for the goal \citep{Rimon1990}.
However, constructing such functions is computationally expensive and generally only feasible for simple environments.
Another related approach, the \emph{wavefront planner} or \emph{brushfire algorithm}, discretizes the configuration space (similar to grid-based methods) and propagates a potential wave outward from the goal, effectively creating a potential field on a grid that is guaranteed to be free of local minima.

In summary, potential field methods provide a fast and reactive framework for local motion planning and obstacle avoidance.
While their simplicity is appealing, their unreliability for global planning due to the local minima problem limits their use as a standalone, complete motion planner.

\section{Summary}
\label{sec:motion_planning_summary}
In this chapter, we introduced motion planning as the critical link that translates high-level task specifications into feasible, collision-free paths.
We formulated the core problem as finding a sequence of actions that drives a robot from a start configuration to a goal configuration while avoiding obstacles.
A central theme was the abstraction of motion into the robot’s configuration space ($\C$-space), where planning reduces to finding a continuous path in the free space, $\C_{\mathrm{free}}$.

We began by discussing \emph{grid-based methods}, which discretize the configuration space and cast the problem as a shortest-path search on a graph.
This allowed us to introduce a general family of label-correcting algorithms, including Dijkstra’s algorithm for finding shortest paths, and its informed extension, A*, which uses heuristics to guide the search efficiently toward the goal.
We then turned to \emph{combinatorial planning}, an exact approach that constructs a roadmap by decomposing the continuous free space into simple cells, guaranteeing completeness but facing computational challenges in high-dimensional spaces.
Next, we discussed \emph{sampling-based methods}, which avoid the explicit construction of the free space by relying on random sampling and collision detection to incrementally build $\C_{\mathrm{free}}$.
We presented the PRM algorithm, a multi-query planner that builds a reusable graph of the free space, and the single-query RRT algorithm, which incrementally grows a tree from the start configuration.
We summarized their key theoretical properties, including probabilistic completeness and asymptotic optimality, which motivated advanced planners like PRM*, RRT* and the FMT* algorithm.
We further extended these concepts to \emph{kinodynamic planning}, which incorporates system dynamics, and discussed how deterministic low-dispersion sampling can provide similar theoretical guarantees as probabilistic methods while offering benefits in predictability and certification.
Finally, we introduced \emph{potential field methods}, which generate paths by defining artificial potential functions that attract the robot to the goal while repelling it from obstacles, enabling fast and reactive local planning at the cost of potential local minima.

\paragraph{To learn more.}
For a deeper exploration of the topics covered in this chapter, several key resources are available.
For a comprehensive and foundational treatment of motion planning, from configuration spaces to combinatorial and sampling-based algorithms, the textbook by \citet{LaValle2006} is an essential reference.
An in-depth presentation of grid-based planning methods and shortest path algorithms is provided in \citet{Bertsekas2000}.
A detailed treatment of the D* algorithm for planning in dynamic environments can be found in \citet{Stentz1995}.
The seminal work on asymptotic optimality for sampling-based planners, which introduced PRM* and RRT*, is presented in \citet{KaramanFrazzoli2011}.
For an in-depth discussion of the FMT* algorithm, we refer the reader to \citet{JansonSchmerlingEtAl2015}.
Kinodynamic planning is thoroughly discussed in \citet{SchmerlingPavone2019}.
The extension of performance guarantees to deterministic sampling patterns is explored in \citet{JansonIchterEtAl2018}.
The original formulation of potential field methods for robot motion planning was introduced by \citet{Khatib1986}, and a comprehensive discussion can be found in \citet{Latombe1991}.
Finally, for a more modern review of motion planning techniques and their applications in robotics, we recommend the survey by \citet{Hauser2020}.

\section{Exercises}
The starter code for the exercises provided below is available online through GitHub. 
To get started, download the code by running in a terminal window:

\begin{tcolorbox}[colback=gray!10]
\begin{minted}{bash}
    git clone https://github.com/StanfordASL/pora-exercises.git
\end{minted}
\end{tcolorbox}

We denote Problems requiring hand-written solutions and coding in Python with \adjustbox{height=2ex, valign=c}{\includegraphics{figs/write.png}} and \adjustbox{height=2ex, valign=c}{\includegraphics{figs/code.png}}, respectively.

\subsection*{\adjustbox{height=2ex, valign=c}{\includegraphics{figs/code.png}}\ Problem 1: A* motion planning}
In this exercise, you will implement the A* grid-based motion planning algorithm for some simple two-dimensional environments.
In the files \\\noindent\colorcode{ch04/exercises/a\_star.ipynb} and \colorcode{ch04/exercises/a\_star.py}, you will implement the key parts of the A* algorithm (see \cref{alg:astar}), run the algorithm on some randomly generated path planning problems, and then explore a way to smooth the resulting discrete paths, which could be useful for practical robot motion planning tasks.

\begin{algorithm}[ht!]
\caption{A* Motion Planning}
\label{alg:astar}
\DontPrintSemicolon
\KwData{Start node $\startnode$, goal node $\goalnode$}
\KwResult{Shortest path from $\startnode$ to $\goalnode$ (if reachable)}
	$\mathcal{O} \gets \{\startnode\}$ \tcc*[r]{Initialize open set}
	$\mathcal{C} \gets \{\}$ \tcc*[r]{Initialize closed set}
	$C(\startnode) \gets 0$ \tcc*[r]{Initialize cost-to-arrive}
	$\tilde{C}(\startnode) \gets \mathsf{distance}(\startnode, \goalnode)$ \tcc*[r]{Initialize estimated cost}
	\While{$\mathcal{O}$ is not empty}{
		$q \gets \mathsf{lowest\_est\_cost\_through}(\mathcal{O})$\\
		\If{$q = \goalnode$}{
			\Return{$\mathrm{reconstruct\_path}()$}
		}
		$\mathcal{O}.\mathrm{remove}(q)$ \\
		$\mathcal{C}.\mathrm{add}(q)$ \\
		\For{$q' \in \mathrm{free\_neighbors}(q)$}{
			\If{$q' \in \mathcal{C}$}{
				\textbf{continue}
			}
			\If{$q' \not\in \mathcal{O}$}{
				$\mathcal{O}.\mathrm{add}(q')$ \\
			}
			\ElseIf{$C(q) +  \mathsf{distance}(q,q') > C(q')$}{
				\textbf{continue}
			}
			$\mathrm{parent}(q') \gets q$ \\
			$C(q') \gets C(q) +  \mathsf{distance}(q,q')$ \\
			$\tilde{C}(q') \gets C(q') + \mathsf{distance}(q',\goalnode)$ \\
		}
	}
	\Return{$\mathrm{Failure}$}
\end{algorithm}

In this implementation of A*, we will represent the free space by a graph, which is traversed by sampling and collision-checking states from a deterministic grid. 
This implementation can be categorized as informed, deterministic sampling-based planning (``informed'' due to the A* heuristic).

\begin{enumerate}
 \item Implement the remaining functions in \colorcode{a\_star.py} within the \colorcode{Astar} class. 
These functions represent many of the key functional blocks at play in motion planning algorithms:
     \begin{itemize}
         \item \colorcode{is\_free} which checks whether a state is collision-free and valid.
         \item \colorcode{distance} which computes the travel distance between two points.
         \item \colorcode{get\_neighbors} which finds the free neighbor states of a given state.
         \item \colorcode{solve} which runs the A* motion planning algorithm.
     \end{itemize}
\end{enumerate}

\textbf{Note:} Notice that we collision-check states but do not collision-check edges. This saves us some computation (collision-checking is often one of the most expensive operations in motion planning).
Also, in this case the obstacles are aligned with the grid, so paths will remain collision-free. However, outside such special circumstances we should add edge collision-checking and/or inflate obstacles to guarantee collision-avoidance.

\begin{enumerate}
\setcounter{enumi}{1}
\item Planning a path on a grid is often not very desirable for a real robot that would have to track the trajectory.
In this exercise, we will smooth the paths from A* by fitting a cubic spline to the path nodes. 
Implement this within the \colorcode{compute\_smooth\_plan} function of \colorcode{a\_star.ipynb}. 
\end{enumerate}

\textbf{Note:} There are many ways to ensure smoothed solutions are collision-free (for example, collision-checking smoothed paths and running a dichotomic search on the smoothing parameters to find a tight fit against obstacles, or inflating obstacles in the original planning to give additional room for smoothing). 
This strategy can be used on geometric sampling-based planning methods as well.

\subsection*{\adjustbox{height=2ex, valign=c}{\includegraphics{figs/code.png}}\ Problem 2: Rapidly-exploring random trees}
In this exercise, you will implement the RRT sample-based motion planning algorithm to plan paths in simple 2D environments.
In the files \\\noindent\colorcode{ch04/exercises/rrt.ipynb} and \colorcode{ch04/exercises/rrt.py}, you will implement the key parts of the RRT algorithm and define a \colorcode{GeometricRRT} planner that leverages simple straight line connections between nodes.

For this implementation of RRT, we consider a ``Geometric'' RRT problem where nodes are connected with simple straight lines.

\begin{algorithm}[ht!]
\caption{RRT Motion Planning}
\label{alg:rrt}
\DontPrintSemicolon
\KwData{Start node $\startnode$, goal node $\goalnode$, max iterations $N_{\max}$, goal sampling bias $\alpha$}
\KwResult{Path from $\startnode$ to $\goalnode$ (if reachable)}
	$\mathcal{T} \gets \{\startnode\}$ \tcc*[r]{Initialize tree}
	\For{$k \leftarrow 1$ \KwTo $N_{\max}$}{
		\If{$\mathrm{rand()} < \alpha$}{
			$q' \leftarrow \goalnode$
		}
		\Else{
			$q' \leftarrow \mathrm{random\_sample()}$ \\
		}
		$q \leftarrow \mathrm{nearest\_neighbor}(q', \mathcal{T})$ \\
		$q' \leftarrow \mathrm{steer\_towards}(q, q')$ \\
		\If{$\mathrm{is\_free\_motion}(q, q')$}{
			$\mathrm{parent}(q') \leftarrow q$ \\
			$\mathcal{T}.\mathrm{add}(q')$ \\
			\If{$q' = \goalnode$}{
				\Return{$\mathrm{reconstruct\_path}()$}
			}
		}
	}
	\Return{$\mathrm{Failure}$}
\end{algorithm}

\begin{enumerate}
\item Implement the remaining functions in \colorcode{rrt.py} within the \colorcode{RRT} and \colorcode{GeometricRRT} classes:
     \begin{itemize}
         \item \colorcode{RRT.solve} which runs the RRT algorithm in \cref{alg:rrt}.
         \item \colorcode{GeometricRRT.nearest\_neighbor} which computes the nearest neighbor in the current tree to a given point using Euclidean distance.
         \item \colorcode{GeometricRRT.steer\_towards} to compute a new state from a target state following a straight line path.
     \end{itemize}
\item Implement the function \colorcode{RRT.shortcut\_path} to try to find a shorter path from the existing tree by removing nodes from the path that aren't strictly needed.
Run the code in \colorcode{rrt.ipynb} to check your work.
\item Run the provided code (\colorcode{RRT.solve\_optimal}) to compare the standard RRT algorithm you implemented against the RRT* algorithm.
\end{enumerate}
\newpage
\printbibliography[segment=\therefsegment,heading=subbibliography,title={References}]

\part{Robot Perception}
\chapter{Introduction to Robot Sensors}
\label{ch:robot-sensors}
\newrefsegment
The three main pillars of robotic autonomy are perception, planning, and control, which correspond to the see, think, and act stages of autonomy. 
The \emph{perception} component consists of the numerous challenges associated with a robot sensing and understanding its environment, and a key element of perception is the sensors the robot uses to extract meaningful information about the world.
In the next few chapters, we focus on the robot perception problem, and in particular we introduce common sensors utilized in robotics applications, discuss their key performance characteristics, and describe strategies for extracting useful information from the sensor measurements\cite{DudekJenkin2008,SiegwartNourbakhshEtAl2011}.

Robots operate in diverse environments which often require diverse sets of sensors for effective perception. 
For example, a self-driving car may utilize cameras, lidar, and radar for detecting objects in the environment. 
It also requires sensors for characterizing the physical state of the vehicle itself, such as inertial measurement units (IMU), GNSS positioning sensors\sidenote{Global Navigation Satellite System}, and more. 
In this chapter, we will begin by introducing the different types of sensors in Section~\ref{sec:ch6_sensorclass}. Next, we will discuss the performance characteristics of sensors in Section~\ref{sec:ch6_sensorperformance}, and then we will discuss common errors and uncertainty quantification for them in Section~\ref{sec:ch6_sensorerrors}. Finally, we will survey some of the most common sensors used in mobile robotics applications in Section~\ref{sec:ch06_commonsensors}.

\section{Sensor Classifications}
\label{sec:ch6_sensorclass}
We use the terms \emph{proprioceptive} and \emph{exteroceptive} to distinguish between sensors that measure the environment and sensors which measure quantities related to the robot itself.
\begin{definition}[Proprioceptive]
Proprioceptive sensors measure values internal to the robot.
For example, a proprioceptive sensor might measure motor speed, wheel load, robot arm joint angles, or battery voltage.
\end{definition}

\begin{definition}[Exteroceptive]
Exteroceptive sensors acquire information from the robot's environment.
For example, exteroceptive sensors measure distances to objects, light intensity, and sound amplitude. 
\end{definition}
Generally speaking, exteroceptive sensor measurements are more likely to require interpretation by the robot in order to extract meaningful environmental features.
In addition to characterizing sensors based on what they measure, we also characterize sensors as \emph{passive} or \emph{active} based on how they operate. 
\begin{definition}[Passive sensor]
Passive sensors, such as thermometers and cameras, measure ambient environmental energy entering the sensor.
\end{definition}
\begin{definition}[Active sensor]
Active sensors, such as ultrasonic sensors, lidar and radar, emit energy into the environment and measure the reaction.
\end{definition}
Classifying a sensor as active or passive is important because each exhibits unique characteristics and challenges.
For example, passive sensors are heavily influenced by environmental conditions, such as a camera's reliance on good ambient lighting to take quality images.

\section{Sensor Performance}
\label{sec:ch6_sensorperformance}
Different types of sensors exhibit varying performance attributes.
While some sensors maintain exceptional accuracy in controlled laboratory settings, their performance may suffer in natural real-world environments.
Conversely, other sensors offer narrow, high-precision data across a variety of settings.
We quantify and compare sensor performance characteristics by defining metrics related to \emph{design specifications} and \emph{in situ}\sidenote{In situ metrics quantify how well a sensor performs in the real environment.} performance. 

\subsubsection{Design Specification Metrics} 
A number of performance characteristics are specifically considered when designing a sensor, and which are also used to quantify its overall nominal performance capabilities.
\begin{enumerate}
    \item \emph{Dynamic range} quantifies the ratio between the lower and upper limits of the sensor inputs under normal operation. 
    We usually express this metric in decibels (dB), and compute it as:
    \begin{equation*}
        \text{DR} = 10 \log_{10}(r) \:\:[\text{dB}],
    \end{equation*}
    where $r$ is the ratio between the upper and lower limits. 
    In addition to the dynamic range ratio, the actual range is also an important sensor metric. 
    For example, an optical rangefinder has a minimum operating range and gives spurious data when measurements are taken with the object closer than that minimum.
    \item \emph{Resolution} is the minimum difference between two values that can be detected by a sensor. 
    The lower limit of the dynamic range of a sensor is usually equal to its resolution\sidenote{This is not necessarily the case for digital sensors}. 
    \item \emph{Linearity} characterizes whether or not the sensor's output depends linearly on the input.
    \item \emph{Bandwidth} or \emph{frequency} is used to measure the speed with which a sensor can provide a stream of readings. 
    We usually express this metric in units of Hertz (Hz), which is measurements per second. 
    High bandwidth sensors are desirable so that downstream information can be updated at a high rate. 
    For example, mobile robots may have to limit their maximum speed based on the bandwidth of their obstacle detection sensors.
\end{enumerate}

\subsubsection{In Situ Performance Metrics} 
Metrics related to the design specifications can be reasonably quantified in a laboratory environment and then extrapolated to predict performance during real-world deployment. 
However, several important sensor metrics cannot be adequately characterized in lab settings since they are influenced by complex interactions between the environment. 
\begin{enumerate}
    \item \emph{Sensitivity} defines the ratio of change in the output from the sensor to a change in the input. 
    High sensitivity is often undesirable because any noise to the input can be amplified, but low sensitivity might degrade the ability to extract useful information from the sensor's measurements.
    \emph{Cross-sensitivity} defines the sensitivity to environmental parameters that are unrelated to the sensor's target quantity. 
    For example, a flux-gate compass can demonstrate high sensitivity to magnetic north and is therefore useful for mobile robot navigation. 
    However, the compass also has high sensitivity to ferrous building materials, so much so that its cross-sensitivity often makes the sensor useless in some indoor environments. 
    High cross-sensitivity of a sensor is generally undesirable, especially when it cannot be modeled.
    \item \emph{Error} of a sensor is defined as the difference between the sensor's output measurements and the true values being measured, within some specific operating context. 
    Given a true value, $v$, and a measured value, $m$, we define the error as $e\definedas m-v$.
    \item \emph{Accuracy} is defined as the degree of conformity between the sensor's measurement and the true value, and is often expressed as a proportion of the true value, for example we may state that a sensor has 97.5\% accuracy. 
    Therefore, small error corresponds to high accuracy and large error corresponds to low accuracy. 
    For a measurement, $m$, and true value, $v$, we define the accuracy as $a\definedas 1-|m-v|/v$. 
    Characterizing sensor accuracy is challenging since obtaining the true value, $v$, can be difficult or impossible.
    \item \emph{Precision} defines the reproducibility of the sensor results. 
    For example, a sensor has high precision if multiple measurements of the same environmental quantity are similar. 
    It is important to note that precision is not the same as accuracy\sidenote{A very precise sensor can still be highly inaccurate}.
\end{enumerate}

\section{Sensor Errors and Uncertainty Modeling}
\label{sec:ch6_sensorerrors}
When discussing in situ performance metrics such as accuracy and precision, it is important to be able to reason about the sources of sensor errors. 
In particular, it is important to distinguish between two main types of error, \emph{systematic errors} and \emph{random errors}.
\begin{enumerate}
    \item \emph{Systematic errors} are caused by factors or processes that can in theory be modeled because they are deterministic and therefore reproducible and predictable. 
    Calibration errors are a common source of systematic errors in sensors.
    \item \emph{Random errors} cannot be predicted using a sophisticated model since they are stochastic and unpredictable. 
    Hue instability in a color camera, spurious rangefinding errors, and black level noise in a camera are all examples of random errors.
\end{enumerate}

To reliably employ a sensor in practice, it is beneficial to characterize its systematic and random errors to allow for corrections that improve its accuracy and provide information about its precision. 
We refer to the process of quantifying sensor errors and identifying their origins as \emph{error analysis}.
This analysis often entails identifying all sources of systematic errors, modeling random errors\sidenote{For example, using Gaussian distributions.}, and assessing the cumulative effect of errors on the sensor's output.

However, conducting a comprehensive error analysis can be difficult due to several factors.
A significant challenge arises due to a \emph{blurring} between systematic and random errors that is the result of changes to the operating environment. 
For instance, exteroceptive sensors on a mobile robot face varying measurement conditions as the robot navigates, with the sensor's performance potentially influenced by the robot's own movement.
Therefore, while we can classify sensor errors as systematic or random in controlled environments, accurately characterizing these errors becomes substantially more complex in real-world settings.

If we could perfectly model and understand all systematic errors in sensor measurements we could theoretically correct for them.
However, in practice, this is often not feasible.
We therefore characterize uncertainty due to random errors by using \emph{probability distributions}.
Given the practical challenge of identifying all sources of random error, we commonly make assumptions when modeling the error distribution.
We commonly assume that random errors have a zero-mean, and that the distribution is symmetric and \emph{unimodal}\sidenote{A very common distribution that fits these properties is the Gaussian distribution.}.
These assumptions can make mathematical analysis easier, but they also have limitations.
For example, some assumptions, such as the unimodality of the distribution, may not hold true in real-world applications.

\begin{example}[Sensor uncertainty assumptions]
Consider a sonar sensor, an active sensor that uses acoustic pulses to measure distance.
Suppose the sonar's accuracy is high, with random errors mainly stemming from noise from internal timing circuits. 
We could reasonably assume this noise is unimodal and possibly Gaussian.
However, in scenarios where the sonar encounters materials causing coherent reflections, distance over-estimations become likely and could result in a bias towards positive errors.
A comprehensive distribution that also captures this effect should be bimodal and asymmetric.
\end{example}

\section{Common Sensors in Mobile Robotics}
\label{sec:ch06_commonsensors}
In mobile robotics applications, we encounter both proprioceptive and exteroceptive sensors working together to enable autonomous operation. 
Proprioceptive sensors provide the robot with information about its internal state—encoders measure joint positions and wheel rotations, IMUs track orientation and acceleration, and heading sensors determine the robot's direction. 
These sensors form the foundation for understanding the robot's own configuration and motion. 
Exteroceptive sensors, on the other hand, gather information about the surrounding environment—active ranging sensors measure distances to obstacles, beacons provide absolute position references, and vision sensors capture rich visual information about the scene. 
The following sections examine key examples of each type, with particular emphasis on those most commonly used in mobile robotics.

\subsubsection{Proprioceptive Sensors}

\paragraph{Encoders.} 
Encoders are proprioceptive electro-mechanical sensors that convert mechanical motion into a series of digital signals that can be interpreted to measure relative or absolute position measurements\sidenote{Thanks to their extensive use across many domains, significant advancements have been made in developing affordable encoders that provide high resolution.}.
One common application of encoders in robotics is for sensing the rotation angle and speed of wheels or motors.
This is important for being able to design good control laws for wheel speed control and motor-driven joints.

One common type of encoder is the \emph{optical encoder}.
Optical encoders work by directing light through slits in a rotating metal or glass disc onto a photodiode, creating sine or square wave pulses corresponding to the disc's rotation.
We can then integrate the number of wave peaks to determine how much the disc has rotated.
The encoder's resolution, expressed in cycles per revolution (CPR), determines its minimum angular resolution.
In terms of bandwidth, it is critical that the encoder is sufficiently fast to handle the expected shaft rotation rates\sidenote{Encoder bandwidth is generally not a concern in mobile robot applications.}. 
Quadrature encoders are also common in robotics applications to additionally sense the direction of rotation. 
As with most proprioceptive sensors, encoders typically operate in a very predictable and controlled environment and we can account for their systematic errors and cross-sensitivities. 
In practice, we often assume perfect accuracy of optical encoders since their errors are typically dwarfed by errors in downstream components.

\paragraph{Inertial measurement unit (IMU).}
Inertial measurement units (IMU) are devices that use gyroscopes and accelerometers to estimate relative position, orientation, velocity, and acceleration with respect to an inertial reference frame. 
An accelerometer measures net acceleration due to external forces, including gravity, while a gyroscope measures angular velocity.
Modern IMUs, such as those in mobile phones, typically use Micro Electro-Mechanical Systems (MEMS) technology for both components.

\begin{figure}[tbh]
\begin{center}
    \begin{tikzpicture}[node distance=2cm, ->]
        \tikzstyle{node} = [draw=black, ellipse, rounded corners, minimum height=3em, minimum width=3em, thick, fill=red!30, align=center]
        \tikzstyle{action} = [draw=black, rectangle, minimum height=2em, minimum width=2em, thick, fill=gray!30, align=center]
        \node[node](s0){\small Rate \\ \small gyroscope};
        	\node[action, below of=s0, xshift=0cm](s0i0){\small Integrate \\ \small (to orientation)};
        	\node[node, right of=s0, xshift=1cm](s1){\small Accelerometer};
        	\node[action, below of=s1, xshift=0cm, yshift=0cm](s1i1){\small Transform to \\ \small local frame};
        	\node[action, right of=s1i1, xshift=0.5cm, yshift=0cm](s1i2){\small Gravity \\ \small correction};
        	\node[action, right of=s1i2, xshift=0.5cm, yshift=0cm](s1i3){\small Integrate \\ \small (to velocity)};
        \node[node, above of=s1i3, yshift=0cm](s2){\small Initial \\ \small velocity};
        	\node[action, right of=s1i3, xshift=0.5cm, yshift=0cm](s1i4){\small Integrate \\ \small (to position)};
        	\node[node, above of=s1i4, yshift=0cm](s3){\small Initial \\ \small position};
        	\node[node, below of=s1i2, yshift=0cm, fill=white!30](s4){\small Acceleration};
        	\node[node, below of=s1i3, yshift=0cm, fill=white!30](s5){\small Velocity};
        	\node[node, below of=s1i4, yshift=0cm, fill=white!30](s6){\small Position};
		\draw[->, thick] (s0) to[] (s0i0);
		\draw[->, thick] (s0i0) to[] (s1i1);
		\draw[->, thick] (s1) to[] (s1i1);
		\draw[->, thick] (s1i1) to[] (s1i2);
		\draw[->, thick] (s1i2) to[] (s1i3);
		\draw[->, thick] (s1i3) to[] (s1i4);
		\draw[->, thick] (s2) to[] (s1i3);
		\draw[->, thick] (s3) to[] (s1i4);
		\draw[->, thick] (s1i2) to[] (s4);
		\draw[->, thick] (s1i3) to[] (s5);
		\draw[->, thick] (s1i4) to[] (s6);
    \end{tikzpicture}
\end{center}
\caption{Inertial measurement unit (IMU) block diagram.}
\label{fig:IMU}
\end{figure}
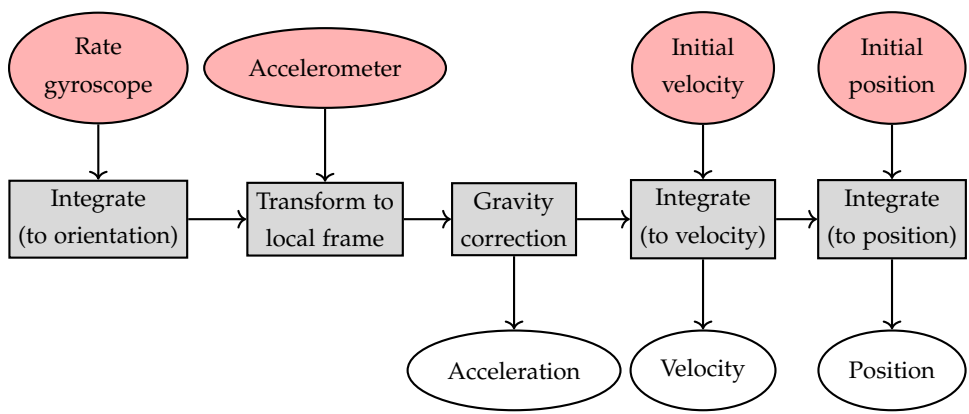

We show the general working principle of an IMU in \cref{fig:IMU}.
First, we integrate gyroscope data to estimate the vehicle orientation while three accelerometers estimate the instantaneous acceleration along each axis. 
We then transform the acceleration into the local navigation frame using the current estimate of the vehicle orientation relative to gravity and subtract the gravity vector from the measurement. 
Next, we integrate the resulting acceleration to obtain the velocity and integrate again to compute the position, provided that we know both the initial velocity and position. 

One of the fundamental issues with IMUs is the phenomenon called \emph{drift}, which describes the slow accumulation of errors over time. 
Drift in any one component will also affect the downstream components. 
For example, drift in the gyroscope leads to errors in the estimation of the vehicle orientation relative to gravity, which results in incorrect cancellation of the gravity vector. 
Additionally, errors in acceleration measurements will cause the integrated velocity to drift in time, which will in turn also cause position estimate drift. 
We can account for drift by using periodic references to some external measurement, such as GNSS position measurements, cameras, or other sensors.

\paragraph{Heading sensors.}
Heading sensors determine the robot's orientation in space and can be proprioceptive (gyroscopes) or exteroceptive (compasses).
Compasses measure the Earth's magnetic field to provide an estimate of direction.
Digital compasses using the Hall effect are inexpensive but often suffer from poor resolution and accuracy, while flux-gate compasses have improved resolution and accuracy but are more expensive.
Both types are vulnerable to magnetic field disturbances, making them less suitable for indoor applications.
Gyroscopes preserve their orientation with respect to a fixed inertial reference frame and can be mechanical or optical.
While high-quality gyroscopes can achieve excellent accuracy (angular drift of about 0.1 degrees in 6 hours), they are expensive and still suffer from drift over time.

\subsubsection{Exteroceptive Sensors}

\paragraph{Active ranging.}
Active ranging sensors provide direct distance measurements to objects in the vicinity of the sensor. 
These sensors are important in robotics for localization and environment reconstruction. 
There are two main types of active ranging sensors: time-of-flight active ranging sensors and geometric active ranging sensors\sidenote{Examples of time-of-flight sensors include ultrasonic, laser rangefinder, and time-of-flight cameras, and examples of geometric sensors include optical triangulation and structured light sensors.}.

\begin{enumerate}
    \item \emph{Time-of-flight Active Ranging:} 
Time-of-flight active ranging sensors make use of the propagation speed of sounds or electromagnetic waves. 
In particular, the travel distance is given by: 
\begin{equation*}
    d = \frac{c t}{2},
\end{equation*}
where $d$ is the distance to the target, $c$ is the speed of wave propagation, and $t$ is the measured round-trip time of flight.
Note that the time of flight is significantly smaller when using electromagnetic signals, on the order of nanoseconds for distances on the order of meters, which can make these types of sensors more challenging to develop in an affordable and robust way.
The quality of time-of-flight range sensors depends on several factors including uncertainties in determining the exact time of arrival of the reflected signal, inaccuracies in the time of flight measurement, the dispersal cone of the transmitted beam\sidenote{Mainly with ultrasonic range sensors.}, interaction with the target\sidenote{For example, surface absorption, specular reflections.}, and the speed of the mobile robot and dynamic targets.
    \item \emph{Geometric Active Ranging:} Geometric active ranging sensors use geometric properties in the measurements to establish distance readings. 
Generally, these sensors project a known pattern of light and then we can use geometric properties to analyze the reflection and estimate range via triangulation.
Optical triangulation sensors (1D) transmit a collimated beam toward the target and use a lens to collect reflected light and project it onto a position-sensitive device or linear camera. 
Structured light sensors (2D or 3D) project a known light pattern such as a point, line, or texture, onto the environment. 
The reflection is captured by a receiver and then, together with known geometric values, we can estimate range via triangulation. 
\end{enumerate}

\paragraph{Beacons.}
Beacons are signaling devices with precisely known positions\sidenote{Stars and lighthouses are classic examples.} that enable position determination through relative measurements. 
The GNSS positioning system is an advanced example of beacons that works by processing synchronized signals from at least four satellites to estimate three position coordinates and a clock correction variable.
Indoor positioning systems often use camera-based motion capture or ultra-wideband beacons for similar functionality in GPS-denied environments.

\paragraph{Vision sensors.}
Vision sensors have become crucial for perception in robotics due to their ability to capture an enormous amount of information about the environment\sidenote{The human eye provides millions of bits of information per second.}.
Unlike the sensors discussed above which provide specific measurements like distance or orientation, cameras capture rich visual data that can be processed to extract various types of information including object detection, depth estimation, motion tracking, and scene understanding.
The main challenges associated with vision-based sensing are related to processing digital images to extract salient information like object depth, motion and object detection, color tracking, feature detection, scene recognition, and more.
We generally refer to the analysis and processing of images as \emph{computer vision} and \emph{image processing}.
The next chapter will explore camera models and calibration in detail, followed by techniques for extracting 3D information from visual data.

\section{Summary}
This chapter introduced the fundamental concepts of robot sensors, including their classifications, performance metrics, and common types used in mobile robotics. 
Understanding sensor characteristics—including their limitations and error sources—is essential for effective robot perception. 
While we've surveyed various sensor types, the remainder of this part will focus specifically on vision-based sensing, beginning with camera models and calibration in the next chapter. Later chapters will introduce how these sensors can be used to extract useful information about the environment, such as object detections.

\paragraph{To learn more.}
To dive deeper into robot sensors and perception, \textit{Introduction to Autonomous Mobile Robots} by \citet{SiegwartNourbakhshEtAl2011} provides a comprehensive overview of various sensor types and their applications in robotics. 
Additionally, readers interested in robotic sensor details are encouraged to read \citet{DudekJenkin2008}.
\newpage
\printbibliography[segment=\therefsegment,heading=subbibliography,title={References}]
\chapter{Camera Models and Calibration}
\label{ch:cameras}
\newrefsegment
While the previous chapter surveyed various sensor types used in robotics, vision sensors warrant special attention due to their unique ability to capture rich, high-dimensional information about the environment.
A single camera image can contain information about object identity, pose, color, texture, and spatial relationships—information that would require multiple specialized sensors to obtain otherwise.
However, extracting useful information from camera images requires understanding how three-dimensional scenes are projected onto two-dimensional image planes, and how we can calibrate cameras to enable accurate measurements.

This chapter explores the mathematical foundations of camera models, from the basic pinhole model to modern RGB-D sensors, and presents practical methods for camera calibration that enable quantitative vision-based perception. 
In Section~\ref{sec:ch07_imageformulation} we introduce the principles of image formation and the mathematical models that describe how cameras capture images.
Next, in Section~\ref{sec:ch07_cameramodels}, we derive the perspective projection equations and discuss practical considerations such as lens distortion.
Then, we will discuss an increasingly popular type of camera, RGB-D cameras, in Section~\ref{sec:ch07_rgbdcameras}, which provide both color and depth information.
Finally, in Section~\ref{sec:ch07_perspectiveprojection} and Section~\ref{sec:ch07_cameracalibration}, we derive the mathematical model for perspective projection and present methods for camera calibration, which is essential for using these models in practice.

\section{Digital Cameras and Image Formation}
\label{sec:ch07_imageformulation}
Modern cameras consist of a sensor that captures light and converts the resulting signal into a digital image. 
Light falling on an imaging sensor is usually picked up by an active sensing area, integrated for the duration of the exposure\sidenote{The duration of exposure is usually expressed as the shutter speed, such as $1/125$, $1/60$, or $1/30$ of a second.}, and then passed to a set of sense amplifiers. 

\subsubsection{Image Sensors}
The two main kinds of sensors used in digital cameras today are charge-coupled devices (CCD) and complementary metal-oxide-semiconductor (CMOS) sensors.
A CCD chip is an array of light-sensitive picture elements called pixels, and can contain between 20,000 and several million pixels total. 
We can think of each pixel as a light-sensitive discharging capacitor that is $5$ to $25\mu \text{m}$ in size.
While complementary metal oxide semiconductor (CMOS) chips also consist of an array of pixels, they are quite different from CCD chips. 
In particular, along the side of each pixel are several transistors specific to that pixel. 
CCD sensors have typically outperformed CMOS for quality-sensitive applications such as digital single-lens-reflex cameras, while CMOS sensors are better for low-power applications. 
However, today, CMOS sensors are standard in most digital cameras.

\subsubsection{Image Formation}
Rays of light reflected by an object tend to be scattered in many directions and may consist of different wavelengths. 
Averaged over time, the emitted wavelengths and directions for a specific object can be precisely described using object-specific probability distribution functions. 
In particular, the light reflection properties of a given object are the result of how light is reflected, scattered, or absorbed based on the object's surface properties and the wavelength of the light. 
For example, an object might look blue because blue wavelengths of light are primarily scattered off the surface while other wavelengths are absorbed. 

\begin{figure}[ht]
\centering
\includegraphics[width=0.95\textwidth]{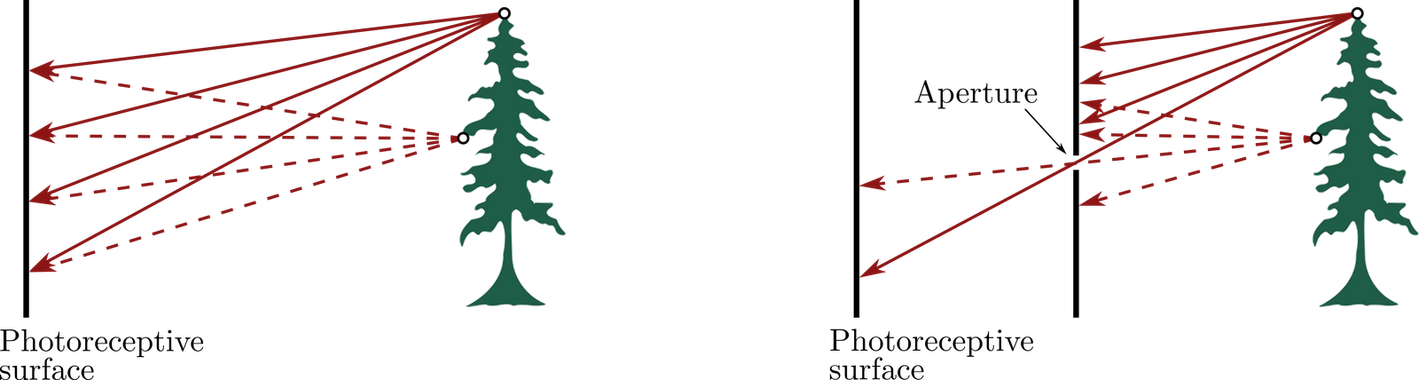}
\caption{Light rays on a photoreceptive surface, referred to as the image plane. 
On the left, numerous rays being reflected and scattered by the object leads to blurry images whereas, on the right, a barrier has been added so that the scattered light rays can be distinguished.}
\label{fig:LightRays}
\end{figure}

Cameras capture images by sensing reflected light rays on a photoreceptive surface such as a CCD or a CMOS sensor. 
Since light reflecting off an object is generally scattered in many directions, exposing a planar photoreceptive surface to these reflected rays would result in many rays being captured at each pixel, which would lead to blurry images.
A solution to this issue is to add a barrier in front of the photoreceptive surface that only lets some of the rays pass through an aperture, as we show in \cref{fig:LightRays}. 
The earliest approach to filtering light rays in this way was to have a small hole in the barrier surface. 
We refer to cameras with this type of filter as \emph{pinhole} cameras.

\section{Camera Models}
\label{sec:ch07_cameramodels}
\subsubsection{Pinhole Camera Model} 
\begin{figure}[ht]
\centering
\includegraphics[width=0.85\textwidth]{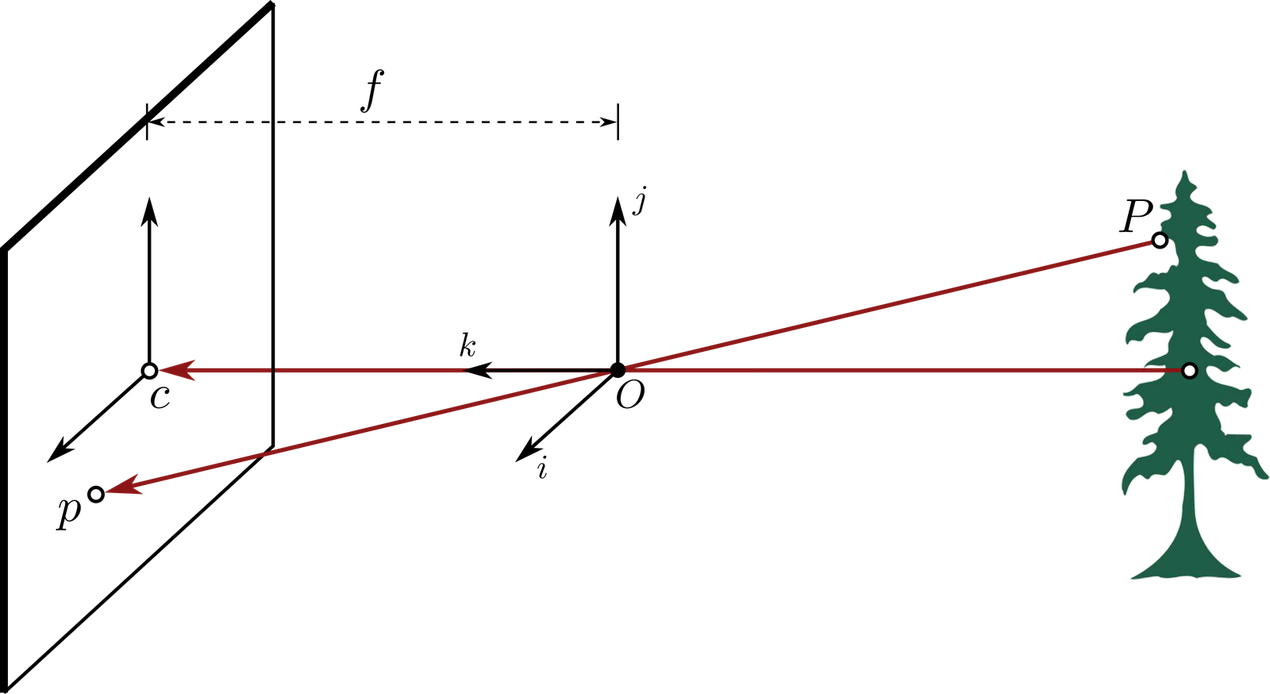}
\caption{Pinhole camera model. 
Due to the geometry of the pinhole camera system, the object's image is inverted on the image plane. 
In this figure, $O$ is the camera center, $c$ is the image center, and $p$ is the principal point.}
\label{fig:PinholeMath}
\end{figure}

A pinhole camera has no lens but rather a single small aperture. 
Light from the scene passes through this pinhole aperture and projects an inverted image onto the image plane, as we show in \cref{fig:PinholeMath}. 
While modern cameras do not operate in this way, we can use the principles of the pinhole camera to derive useful mathematical models. 

We start by defining several useful references to help develop the mathematical pinhole camera model. 
First, the \emph{camera reference frame} is centered at a point, $O$, that is at a focal length, $f$, in front of the image plane, as we show in \cref{fig:PinholeMath}. 
We define this reference frame, with directions $\tup{i,j,k}$, such that the $k$ axis is coincident with the \emph{optical axis} that points toward the image plane. 
We denote the coordinates of a point in the camera frame by $P = \tup{X,Y,Z}$. 
When a ray of light is emitted from a point, $P$, and passes through the pinhole at point $O$, it gets captured on the image plane at a point $p$. 
Since these points are all collinear, we can deduce the following relationships between the coordinates $P = \tup{X,Y,Z}$ and $p = \tup{x,y,z}$:
\begin{equation*}
x = \lambda X, \quad y = \lambda Y, \quad z = \lambda Z,
\end{equation*}
for some $\lambda \in \R$. 
This leads to the relationship:
\begin{equation*}
\lambda = \frac{x}{X}= \frac{y}{Y}= \frac{z}{Z}.
\end{equation*}
From the geometry of the camera, we can see that $z = f$ where $f$ is the focal length, such that we can rewrite these expressions as:
\begin{equation} 
\label{eq:pinhole}
    x=f\frac{X}{Z}, \quad  y = f\frac{Y}{Z}.
\end{equation}
Therefore, we can compute the position of the pixel on the image plane that captures a ray of light from the point $P$.

\subsubsection{Thin Lens Model}
One of the main issues with having a fixed pinhole aperture is that there is a trade-off associated with the aperture's size. 
A large aperture allows a greater number of light rays to pass through, which leads to image blurring. 
A small aperture lets through fewer light rays, but the resulting image is darker. 
As a solution, lenses focus light by refraction and can be used to replace the aperture, avoiding the need for these trade-offs. 

\begin{figure}[ht]
\centering
\includegraphics[width=0.8\textwidth]{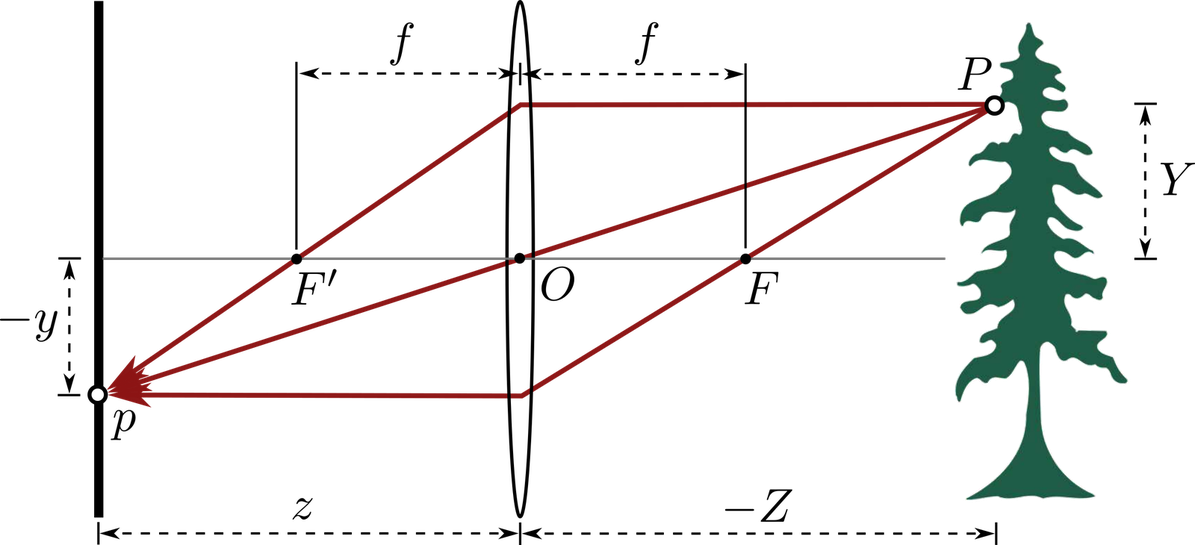}
\caption{The thin lens model.}
\label{fig:Lens}
\end{figure}

We can develop a mathematical model for lenses similar to the pinhole model by using properties from Snell's law. 
\cref{fig:Lens} shows a diagram of the most basic lens model, which is the \emph{thin lens model}\sidenote{The \emph{thin lens model} assumes no optical distortion due to the curvature of the lens.}. 
Snell's law states that rays passing through the center of the lens are not refracted, and those that are parallel to the optical axis are focused on the focal point, labeled $F'$. 
In addition, all rays passing through $P$ are focused by the thin lens on the point $p$. 
We develop a mathematical model similar to \cref{eq:pinhole} from the geometry of similar triangles:
\begin{equation}
    \frac{y}{Y}=\frac{z}{Z}, \quad \frac{y}{Y}=\frac{z-f}{f}  = \frac{z}{f} -1,
\end{equation}
where again the point $P$ has coordinates $\tup{X,Y,Z}$, its corresponding point, $p$, on the image plane has coordinates $\tup{x,y,z}$, and $f$ is the focal length.
Combining these two equations yields the \emph{thin lens equation}:
\begin{equation} 
\label{eq:thinlens}
    \frac{1}{z}+\frac{1}{Z}=\frac{1}{f}.
\end{equation}
Note that in this model, and for a particular focal length, $f$, a point, $P$, is only in sharp focus if the image plane is located a distance $z$ from the lens. 
In practice, an acceptable focus is possible within some range of distances referred to as depth of field or depth of focus. 
Additionally, if $Z$ approaches infinity, light would focus a distance of $f$ away from the lens. 
Therefore, this model is essentially the same as a pinhole model if the lens is focused at a distance of infinity.

\subsubsection{Radial Distortion}
\begin{figure}[ht]
\includegraphics[width=0.75\textwidth]{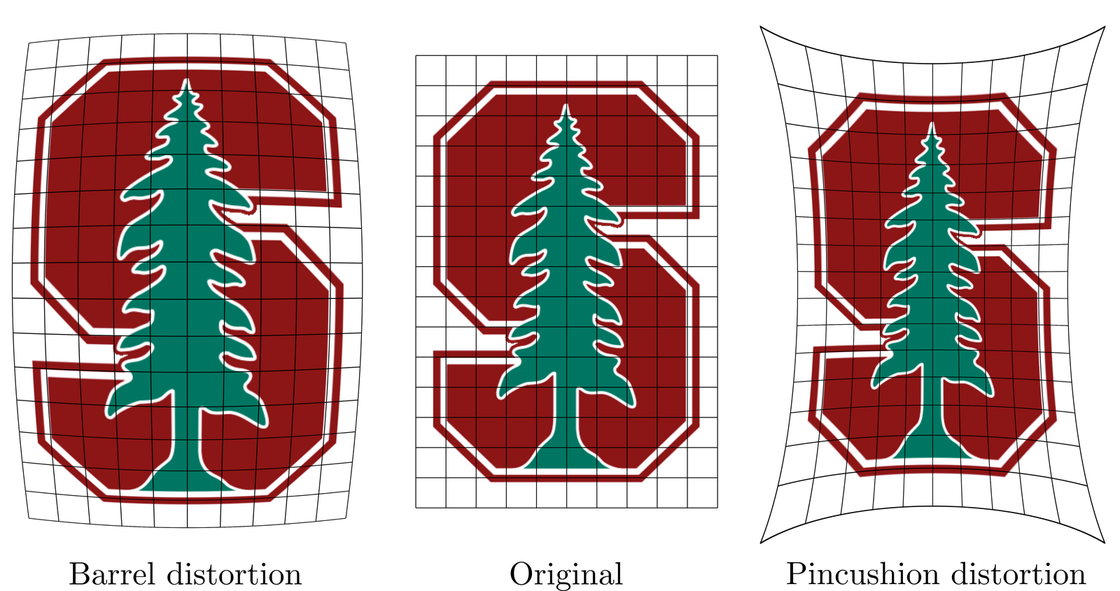}
\centering
\caption{Different kinds of radial distortions that are seen in real lenses, which may affect the accuracy of the pinhole camera model.}
\label{fig:distortion}
\end{figure}

The pinhole camera model provides a nominal camera model for which it is relatively straightforward to develop a mathematical model of the perspective projection. 
However, in practice, this model is not a perfect representation of the imaging process. 
One effect that is not captured by the pinhole model is \emph{radial distortion}, which is an effect seen in real lenses where either barrel distortion or pincushion distortion will affect the real pixel coordinates. 
We show images of barrel and pincushion distortion in \cref{fig:distortion}.

There are methods we can use to correct for image distortion. 
A simple and efficient way is to model the relationship between the ideal pixel coordinates, $\tup{u,v}$, and the distorted pixel coordinates, $\tup{u_d,v_d}$, as:
\begin{equation}
\begin{bmatrix}
    u_d \\
    v_d
\end{bmatrix}
=
(1+kr^2)
\begin{bmatrix}
    u-u_{cd} \\
    v-v_{cd}
\end{bmatrix}
+
\begin{bmatrix}
    u_{cd} \\
    v_{cd}
\end{bmatrix},
\end{equation}
where $k \in \mathbb{R}$ is the radial distortion factor, $\tup{u_{cd}, v_{cd}}$ are the pixel coordinates of the image center, and $r^2=(u-u_{cd})^2+(v-v_{cd})^2$ is the square of the distance between the ideal pixel location and the center of distortion.
Note that $k$ differs across cameras and needs to be predetermined through calibration.

\section{RGB-D Cameras}
\label{sec:ch07_rgbdcameras}
While traditional cameras capture color information through RGB channels, RGB-D cameras additionally provide depth information for each pixel, creating a 2.5D representation of the scene.
These sensors have become increasingly important in robotics due to their ability to provide both appearance and geometric information in a single, compact package.

\subsubsection{Depth Sensing Technologies}
RGB-D cameras employ various technologies to capture depth information alongside color:

\paragraph{Structured light.}
These sensors project a known infrared pattern onto the scene and use triangulation to compute depth from the pattern's deformation.
The original Microsoft Kinect exemplifies this approach, projecting a speckle pattern that is captured by an IR camera offset from the projector.
Structured light sensors provide high accuracy at close range (0.5 to 4 m) but struggle in bright ambient light conditions.

\paragraph{Time-of-flight (ToF).}
ToF cameras emit modulated infrared light and measure the phase shift of the reflected signal to determine distance.
These sensors offer good accuracy across their operating range and work well in varying lighting conditions, though they typically have lower resolution than structured light sensors and can suffer from multipath interference.

\paragraph{Stereo infrared.}
Some RGB-D cameras, such as the Intel RealSense D-series, use stereo vision with infrared cameras and an optional IR pattern projector.
This approach combines the robustness of stereo vision with active illumination to handle textureless surfaces.

\subsubsection{Applications in Robotics and Practical Considerations}
RGB-D cameras have enabled significant advances in several robotics applications, particularly in structured indoor environments.
For indoor navigation and mapping, RGB-D sensors are ideal when GPS is unavailable and lighting can be controlled.
They provide dense 3D point clouds at frame rate, enabling real-time obstacle avoidance and map building, with popular SLAM systems like RGB-D SLAM and ElasticFusion leveraging these sensors for creating detailed 3D maps.
In object manipulation tasks, RGB-D cameras excel by providing both the object's appearance and its precise 3D geometry, allowing robots to identify objects using color and texture while planning grasps based on accurate depth information.
The relatively high accuracy at close range (typically 0.5 to 4 m) makes them well-suited for tabletop manipulation and pick-and-place operations.

However, several practical considerations limit their use.
Most consumer RGB-D sensors operate effectively only between 0.5 and 5 meters, with accuracy degrading quadratically with distance—millimeter accuracy at one meter may degrade to several centimeters at maximum range.
Material properties significantly affect performance: transparent and highly reflective surfaces cannot be measured, while dark materials may return weak signals leading to missing depth values.
RGB-D cameras also require careful calibration between the color and depth sensors, which are typically offset from each other, and factory calibration may not suffice for high-precision applications.
Additionally, multiple RGB-D cameras can interfere with each other's IR projectors, and ambient infrared light (such as sunlight) makes outdoor use challenging.
Despite these limitations, RGB-D cameras offer an attractive balance of cost, size, and capability for many indoor robotics applications, providing rich 3D perception without the computational complexity of stereo matching or the cost of laser scanners.

\section{Perspective Projection and Coordinate Transformations}
\label{sec:ch07_perspectiveprojection}
The pinhole camera model can be used to mathematically define relationships between points in the scene and points on the image plane.
Our objective is to derive a mathematical model that maps a point $P_W$ expressed in world frame coordinates to a point $p$ on the image plane in pixel coordinates.
We accomplish this by combining two transformations: from world frame to camera frame coordinates ($P_W \to P_C$), and from camera coordinates to image coordinates ($P_C \to p$).

\subsubsection{Camera Frame to Image Coordinates ($P_C \to p$)}
The first step we consider is how to map a point in the scene expressed in camera frame coordinates, $P_C$, to the corresponding point on the image plane, $p$, using the pinhole camera model. 
In \cref{sec:ch07_cameramodels}, we presented the pinhole camera equations:
\begin{equation} 
\label{eq:PC2xy}
    x=f\frac{X_C}{Z_C}, \quad  y = f\frac{Y_C}{Z_C},
\end{equation}
where $P_C = \tup{X_C, Y_C, Z_C}$, $p = \tup{x, y}$, and $f$ is the focal length of the pinhole camera\sidenote{We generally do not include the $z$ term of $p$ simply because $z=f$ is a fixed value.}. 

\begin{figure}[ht]
\includegraphics[width=0.55\textwidth]{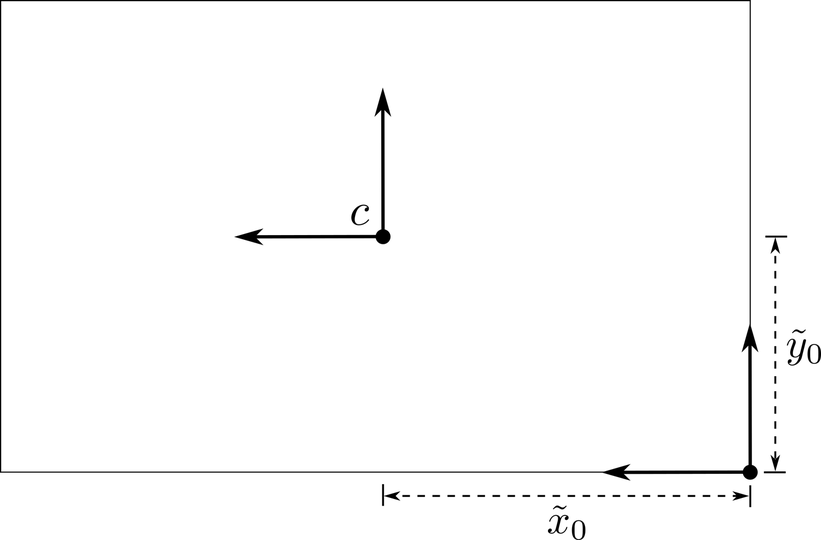}
\centering
\caption{We define a new reference frame with coordinates denoted by $\tilde{(\cdot)}$ with its origin in the lower corner of the image plane. 
The image center coordinates in this new frame are $(\tilde{x}_0, \tilde{y}_0)$.}
\label{fig:camera_coordinates}
\end{figure}

Note that the quantities $x$ and $y$ are coordinates in the \emph{camera frame}, but it is often desirable to express the point $p$ in terms of \emph{pixel coordinates}. 
Pixel coordinates are generally defined with respect to a reference frame in the lower corner of the image plane to avoid negative coordinates. 
We show this new reference frame in \cref{fig:camera_coordinates}, where we define the image center, $c$, with coordinates $\tup{\tilde{x}_0, \tilde{y}_0}$, where $\tilde{(\cdot)}$ is the notation we use to denote a coordinate with respect to this new reference frame.
In this new reference frame, we map the point $P_C$ to the coordinates $\tup{\tilde{x}, \tilde{y}}$ by:
\begin{equation} 
\label{eq:PC2xtyt}
    \tilde{x} = f\frac{X_C}{Z_C} + \tilde{x}_0, \quad  \tilde{y} = f\frac{Y_C}{Z_C} + \tilde{y}_0.
\end{equation}
Finally, given the number of pixels per unit distance, we can map these new coordinates to pixel coordinates. 
In particular, we map the point $P_C$ to pixel coordinates $\tup{u,v}$ by:
\begin{equation} 
\label{eq:PC2uv}
   u = \alpha \frac{X_C}{Z_C} + u_0, \quad  v = \beta \frac{Y_C}{Z_C} + v_0,
\end{equation}
where $\alpha = k_xf$, $u_0 = k_x \tilde{x}_0$, $\beta = k_y f$, $v_0 = k_y \tilde{y}_0$, and $k_x$ and $k_y$ are the number of pixels per unit distance in image coordinates.

Note that the transformation from the point $P_C$ in camera frame coordinates to $p$ in pixel coordinates given by \cref{eq:PC2uv} is not linear. 
However, we can represent this transformation as a linear mapping\sidenote{Expressing the perspective projection as a linear map will simplify the mathematics later on.} through an additional change of coordinates. 
In particular, we will express the points $P_C$ and $p$ in \emph{homogeneous coordinates}.

For a two-dimensional point $\tup{x_1,x_2}$ or a three-dimensional point $\tup{x_1,x_2,x_3}$ in Euclidean space, we represent the point in homogeneous coordinates by the transformation:
\begin{equation}
\tup{x_1,\:x_2} \to \tup{\alpha x_1, \:\alpha x_2, \:\alpha}, \quad \text{and} \quad \tup{x_1,\:x_2,\:x_3} \to \tup{\alpha x_1,\:\alpha x_2, \:\alpha x_3, \: \alpha},
\end{equation}
for any $\alpha \neq 0$. 
These new coordinates are called homogeneous coordinates because we can choose the scaling factor, $\alpha$, arbitrarily as long as $\alpha \neq 0$. 
We transform a set of homogeneous coordinates back by: 
\begin{equation}
\tup{y_1, \: y_2, \:y_3} \to \tup{\frac{y_1}{y_3}, \: \frac{y_2}{y_3}}, \quad \text{and} \quad \tup{y_1,\:y_2,\:y_3,\:y_4} \to \tup{\frac{y_1}{y_4}, \: \frac{y_2}{y_4}, \: \frac{y_3}{y_4}}.
\end{equation}
We will denote when a point is described in homogeneous coordinates using the superscript $h$.
For example, we express the point $P_C = \tup{X_C, Y_C, Z_C}$ in camera frame coordinates with $\alpha = 1$ in homogeneous coordinates by:
\begin{equation*}
    P_C^h = \tup{X_C, Y_C, Z_C, 1},
\end{equation*}
and we can express the pixel coordinate $p = \tup{u,v}$ in homogeneous coordinates by:
\begin{equation*}
    p^h = \tup{Z_C u, \: Z_C v, \:Z_C} = \tup{\alpha X_C + u_0 Z_C, \: \beta Y_C + v_0 Z_C, \: Z_C},
\end{equation*}
by choosing $\alpha = Z_C$ and substituting the expressions from \cref{eq:PC2uv}. 
With the expression of these points in homogeneous coordinates, we can see that their relationship is transformed from the nonlinear relationship in \cref{eq:PC2uv} to the \emph{linear} relationship:
\begin{equation}
\begin{bmatrix}
\alpha & 0 & u_{0} & 0 \\
0 & \beta & v_{0} & 0  \\
0 & 0 & 1 & 0
\end{bmatrix}
\begin{bmatrix}
X_{c} \\
Y_{c} \\
Z_{c} \\
1
\end{bmatrix}
=
\begin{bmatrix}
\alpha X_{c} + u_{0}Z_{c} \\
\beta Y_{c} + v_{0}Z_{c} \\
Z_{c} \\
\end{bmatrix}.
\end{equation}

Often, in practice, we also add a skewness parameter, $\gamma$\sidenote{The skewness parameter generally ends up being close to zero.}, and we can write this linear relationship in the more compact form:
\begin{equation} 
\label{eq:PC2uvhomo}
    \begin{bmatrix}
        K & 0_{3 \times 1}
    \end{bmatrix} P_C^h = p^h, \quad K \definedas \begin{bmatrix}
\alpha & \gamma & u_{0} \\
0 & \beta & v_{0}  \\
0 & 0 & 1
\end{bmatrix}.
\end{equation}
We refer the matrix $K$ in \cref{eq:PC2uvhomo} as the \emph{camera matrix} or \emph{matrix of intrinsic parameters} because it contains the five parameters that define the fundamental characteristics of the camera from the perspective of the pinhole camera model. 
While these parameters may be specified by the camera manufacturer, we often estimate them in practice by performing a camera calibration.

\subsubsection{World Frame to Camera Frame ($P_W \to P_C$)}
\begin{figure}[ht]
\centering
\includegraphics[width=0.65\textwidth]{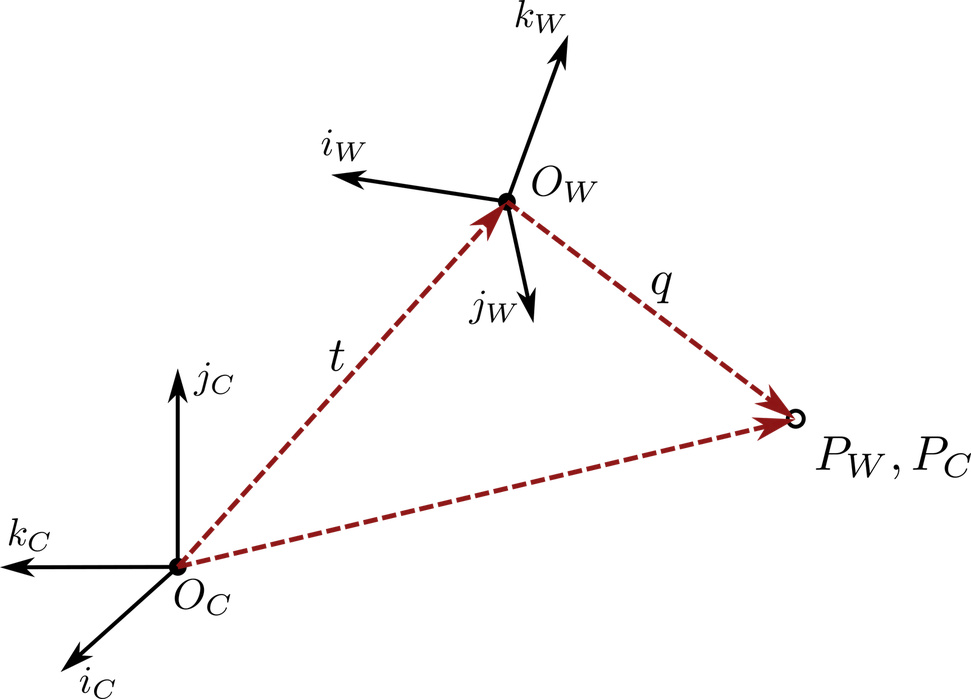}
\caption{A depiction of the point $P$ expressed either in camera coordinates, $P_C$, or in world frame coordinates, $P_W$. 
We denote the world frame origin by $O_W$ and the camera frame origin by $O_C$.}
\label{fig:Pc2Pw}
\end{figure}

We can express a point, $P$, in the scene (see \cref{fig:PinholeMath}) either in terms of camera frame coordinates, $P_C$, or world frame coordinates, $P_W$. 
While we discussed the use of the pinhole model to map $P_C$ coordinates to pixel coordinates, $p$, in the previous section, in this section we discuss the mapping between the camera and world frame coordinates of the point $P$, as we show in \cref{fig:Pc2Pw}.

From \cref{fig:Pc2Pw}, we can write $P_C$ as:
\begin{equation}
P_{C} = t + q,
\end{equation}
where $t$ is the vector from $O_C$ to $O_W$, expressed in camera frame coordinates, and $q$ is the vector from $O_W$ to $P$, expressed in camera frame coordinates. 
However, the vector $q$ is the same vector as $P_W$, just expressed with respect to a different coordinate frame. 
The coordinates are related by a rotation:
\begin{equation}
q = RP_W,
\end{equation}
where $R$ is the rotation matrix relating the camera frame to world frame defined as:
\begin{equation} 
\label{eq:rotmatrix}
R \definedas
\begin{bmatrix}
i_w \cdot i & j_w \cdot i & k_w \cdot i \\
i_w \cdot j & j_w \cdot j & k_w \cdot j \\
i_w \cdot k & j_w \cdot k & k_w \cdot k
\end{bmatrix},
\end{equation}
where $i$, $j$, and $k$ are the unit vectors that define the camera frame and $i_w$, $j_w$, and $k_w$ are the unit vectors that define the world frame. 
To summarize, we can map the point $P_W$ to camera frame coordinates $P_C$ by:
\begin{equation}
P_C = t + R P_W,
\end{equation}
where $t$ is the vector in camera frame coordinates from $O_C$ to $O_W$ and $R$ is the rotation matrix defined in \cref{eq:rotmatrix}.
Similar to the previous section, we can equivalently express this transformation for the case where the points $P_W$ and $P_C$ are expressed in homogeneous coordinates:
\begin{equation} 
\label{eq:Pw2Pchomo}
\begin{split}
\begin{bmatrix}
    P_C \\ 1
    \end{bmatrix} = \begin{bmatrix}
R & t \\
0_{1\times3} & 1
\end{bmatrix}
\begin{bmatrix}
P_W \\ 1
\end{bmatrix}.
\end{split}
\end{equation}

\subsubsection{Complete Projection Pipeline ($P_W \to p$)}
The objective of the perspective projection task is to find a way to mathematically relate the position of a point in world frame coordinates, denoted $P_W$, to the corresponding pixel coordinates, $p$, on the image plane. 
With the relationship from \cref{eq:Pw2Pchomo} that we developed for mapping $P_W$ to the camera frame coordinates, $P_C$, and the relationship in \cref{eq:PC2uvhomo} for mapping $P_C$ to pixel coordinates, $p$, we can now define the direct mapping from $P_W$ to $p$. 
In particular, combining the two transformations together yields:
\begin{equation*}
p^h = \begin{bmatrix}
K \quad 0_{3\times 1}
\end{bmatrix}
\begin{bmatrix}
R && t \\
0_{1 \times 3} && 1
\end{bmatrix}
P_W^h,
\end{equation*}
which we can simplify to:
\begin{equation} 
\label{eq:Pw2uvhomo}
p^h = K\begin{bmatrix}
    R & t
\end{bmatrix} P_W^h.
\end{equation}
In \cref{eq:Pw2uvhomo}, $P_W^h$ is the homogeneous coordinate representation of $P_W$ and $p^h$ is the homogeneous coordinate representation of $p$. 
Recall that the matrix $K \in \mathbb{R}^{3\times3}$ is the matrix of intrinsic camera parameters, and the matrix $[R \:\:\: t] \in \mathbb{R}^{3\times 4}$ contains \emph{extrinsic} parameters\sidenote{Extrinsic parameters describe the camera's position and orientation relative to the points in the scene.}. 
Note that the total number of degrees of freedom is 11, where 5 are from the intrinsic parameters that define $K$, 3 are from the rotation matrix, $R$, and 3 are from the position vector, $t$.

\section{Camera Calibration}
\label{sec:ch07_cameracalibration}
Before we can use camera models in practice, we need to determine the camera's intrinsic and extrinsic parameters.
Camera calibration is the process of estimating these parameters, which is essential for quantitative computer vision applications.
We present two main approaches: the Direct Linear Method for basic calibration, and auto-calibration for cases where calibration targets are unavailable.

\subsubsection{Direct Linear Method}
\label{subsubsec:direct-linear-calibration}
One approach is the \emph{direct linear calibration} method, which requires a set of known correspondences, $p_i \xleftrightarrow[]{} P_{W,i}$ for $i = 1,\dots,n$.

\paragraph{Direct linear calibration: Step 1.}
\label{subsubsec:dlc_step1}
For direct linear calibration, the first step is to write each corresponding pair of points, $p_i = \tup{u_i,v_i}$ and $P_{W,i} = \tup{X_{W,i}, Y_{W,i}, Z_{W,i}}$, in homogeneous coordinates and then use the expression in \cref{eq:Pw2uvhomo} to write:
\begin{equation} 
\label{eq:correspondence}
    p_i^h = M P_{W,i}^h, \quad i = 1,\ldots, n,
\end{equation}
where we refer to $M = K[R \:\:\: t]$ as the \emph{projection matrix}. 
Next, we use the $n$ correspondences to estimate the projection matrix, $M$, and then later we can extract the intrinsic and extrinsic parameters from $M$. 
A useful first step to determine $M$ is to rewrite it in terms of its rows:
\begin{equation}
    M = \begin{bmatrix}
        m_1 \\ m_2 \\ m_3
    \end{bmatrix},
\end{equation}
where $m_i \in \R^{1 \times 4}$ is the $i$-th row of $M$. 
By considering the rows of $M$ individually, we can write the relationship in \cref{eq:correspondence} as:
\begin{equation*}
    \begin{bmatrix}
        \alpha u_i \\ \alpha v_i \\ \alpha
    \end{bmatrix} = \begin{bmatrix}
         m_1 \cdot P_{W,i}^h \\ m_2 \cdot P_{W,i}^h \\ m_3 \cdot P_{W,i}^h \\
    \end{bmatrix}, \quad i = 1,\dots,n
\end{equation*}
which by mapping the homogeneous coordinates, $p_i^h$, back to the original coordinates, $p_i$, yields the $2n$ expressions:
\begin{equation*}
\begin{split}
    u_i &= \frac{m_1 \cdot P_{W,i}^h}{m_3 \cdot P_{W,i}^h}, \quad i = 1,\dots,n, \\
    v_i &= \frac{m_2 \cdot P_{W,i}^h}{m_3 \cdot P_{W,i}^h}, \quad i = 1,\dots,n,
\end{split}
\end{equation*}
or equivalently, by some algebraic manipulation, yields the expressions:
\begin{equation}
\begin{split}
    u_i(m_3 \cdot P_{W,i}^h) - (m_1 \cdot P_{W,i}^h) &= 0, \quad i = 1,\dots,n \\
    v_i(m_3 \cdot P_{W,i}^h) - (m_2 \cdot P_{W,i}^h) &= 0, \quad i = 1,\dots,n.
\end{split}
\end{equation}
We can now combine these $2n$ equations together in one large matrix equation:
\begin{equation}
\label{eq:Pmeq}
    \tilde{P}m = 0, \quad m \definedas \begin{bmatrix}
        m_1^\top  \\ m_2^\top  \\ m_3^\top 
    \end{bmatrix},
\end{equation}
where $m \in \R^{12 \times 1}$ is a vector consisting of the stacked rows of $M$ and $\tilde{P} \in \R^{2n \times 12}$ is a matrix of known coefficients determined by the quantities $u_i$, $v_i$, and $P^h_{W,i}$. 
For a more concrete representation of how we define $\tilde{P}$, the first couple rows are given by:
\begin{equation} 
\tilde{P} =
     \begin{bmatrix}
  -(P_{W,1}^h)^\top  & 0_{1\times 4} & u_{1} (P_{W,1}^h)^\top  \\
  0_{1 \times 4} & -(P_{W,1}^h)^\top  & v_{1} (P_{W,1}^h)^\top   \\
  -(P_{W,2}^h)^\top  & 0_{1\times 4} & u_{2} (P_{W,2}^h)^\top  \\
  \vdots & \vdots & \vdots
 \end{bmatrix}.
\end{equation}
Note that we must have at least six correspondences, $n \geq 6$, to ensure that $m$ is uniquely defined. 
With this sufficient number of correspondences, we could ideally directly solve \cref{eq:Pmeq}. 
However, in practice, a more robust procedure is to build $\tilde{P}$ with more than 6 points, which gives an overdetermined set of equations that may not have a solution\sidenote{This is particularly true in real-world applications where noise corrupts the data.}.
Therefore, to compute $m$, we formulate an optimization problem:
\begin{equation} 
\label{eq:mopt}
\begin{split}
	\minimize[m] & \lVert \tilde{P} m \rVert^2, \\
    \subjectto & \lVert m \rVert^2 = 1,
\end{split}
\end{equation}
where the constraint $\lVert m \rVert^2 = 1$ is required to ensure that the optimization problem cannot be solved by trivially choosing $m_i=0$ for each $i = 1,\dots,12$. 
We call this optimization problem a \emph{constrained least-squares} problem.

\begin{example}[Constrained least-squares optimization] 
\label{ex:constlsq}
\theoremstyle{definition}
The constrained least squares problem:
\begin{equation*}
\begin{split}
	\minimize[x] & \lVert A x \rVert^2, \\
    \subjectto & \lVert x \rVert^2 = 1,
\end{split}
\end{equation*}
with $x\in \R^n$ and $A \in \R^{m \times n}$ and $m > n$ is a finite-dimensional optimization problem. 
Consider the corresponding Lagrangian:
\begin{equation*}
L = x^\top  A^\top Ax + \lambda (1 - x^\top x),
\end{equation*}
and the necessary optimality conditions:
\begin{equation*}
\begin{split}
\nabla_x L &= 2(A^\top A - \lambda I)x = 0, \\
\nabla_\lambda L &= 1 - x^\top x = 0. \\
\end{split}
\end{equation*}
We can write the first necessary optimality condition as $A^\top A x = \lambda x$, and therefore any $x$ that satisfies this condition must be an eigenvector of the matrix $A^\top A$. 
Additionally, while all the eigenvectors satisfy this condition, the optimum is the eigenvector associated with the smallest eigenvalue.
We can efficiently compute this eigenvector by using a singular value decomposition of $A = U\Sigma V^\top $ and then choosing $x$ to be the column of $V$ associated with the smallest singular value, since $A^\top A = V\Sigma^2 V^\top $.
\end{example}

\paragraph{Direct linear calibration: Step 2.}
Once we have solved the optimization problem in \cref{eq:mopt} to compute the vector $m$, the projection matrix, $M$, is completely defined. 
The next step in the camera calibration process is to extract the intrinsic and extrinsic camera parameters from the matrix $M$. 
For this step, we will express the matrix $M$ in terms of its columns:
\begin{equation*}
M = \begin{bmatrix}
    c_1 & c_2 & c_3 & c_4
\end{bmatrix},
\end{equation*}
where $c_i$ is the $i$-th column of $M$. 
We can factorize $M$ as:
\begin{equation}
    M = K\begin{bmatrix}
        R & t
    \end{bmatrix},
\end{equation}
by taking the first three columns of $M$ and performing a \emph{RQ factorization}:
\begin{equation}
\begin{bmatrix}
    c_1 & c_2 & c_3
\end{bmatrix} = KR,
\end{equation}
where $R$ is an orthogonal matrix and $K$ is an upper triangular matrix. 
Once $K$ is known, we can compute the vector $t$ by $t = K^{-1} c_4$.

\subsubsection{A Flexible Camera Calibration Method} 
\label{subsubsec:zhang}
The projection matrix, $M$, is defined for a specific set of extrinsic parameters $R$ and $t$. 
In practice, however, it might be desirable for us to estimate the camera's intrinsic parameters from $N$ different images from different perspectives, and therefore with $N$ different projection matrices due to the varying extrinsic parameters. 
In this case, we can apply an alternative procedure\cite{Zhang2000} to the direct linear calibration method to extract the intrinsic parameters, $K$. 

We begin by assuming that the known points, $P_W$, for each individual image lie on a plane. 
For example, the calibration scene might consist of a pattern, such as a checkerboard pattern, on a planar surface. 
In this case, we can assume that the world frame origin lies on the plane such that $Z_W = 0$ for all points on the plane. 
Since $Z_W = 0$, we can simplify the relationship between $p^h$ and $P^h_W$ given by \cref{eq:Pw2uvhomo} to:
\begin{equation}
p^h = H \tilde{P}_W^h,
\end{equation}
with:
\begin{equation}
    H = K\begin{bmatrix}
    r_1 & r_2 & t
\end{bmatrix}, \quad \tilde{P}_W^h = \begin{bmatrix}
    X_W & Y_W & 1
\end{bmatrix}^\top ,
\end{equation}
where $H$ is called the \emph{homography matrix}\sidenote{The homography matrix maps points between two-dimensional planes while a projection matrix maps three-dimensional points to points on a two-dimensional plane.}, $\tilde{P}_W^h$ is the simplified position of the point $P$ in world frame, written in homogeneous coordinates, and $r_i$ is the $i$-th column of the rotation matrix, $R$. 
Note that we can still estimate the homography matrix, $H$, using the same procedure discussed earlier.

Next, we identify a set of constraints on the intrinsic parameter matrix, $K$, by writing the homography, $H$, as:
\begin{equation*}
    H = \begin{bmatrix}
        Kr_1 & Kr_2 & Kt
    \end{bmatrix} = \begin{bmatrix}
        \tilde{c}_1 & \tilde{c}_2 & \tilde{c}_3
        \end{bmatrix},
\end{equation*}
and noting that since $r_1$ and $r_2$ are orthonormal we have:
\begin{equation} 
\label{eq:zhangconst}
    \tilde{c}_1^\top  B \tilde{c}_2 = 0, \quad \tilde{c}_1^\top  B \tilde{c}_1 = \tilde{c}_2^\top  B \tilde{c}_2,
\end{equation}
where $B = K^{-\top}K^{-1} \in \R^{3 \times 3}$ is a symmetric matrix.
We can therefore solve for the intrinsic camera parameters, $K$, by using the constraints in \cref{eq:zhangconst} to solve for the symmetric matrix $B$ and then backing out the parameters that define $K$.
To compute the matrix $B$ from the constraints in \cref{eq:zhangconst}, we can employ several useful tricks. 
The main trick is to notice that even though $B$ consists of nine parameters, it is symmetric, and, therefore, we only need six parameters to specify it fully.
Therefore, we reparameterize the matrix $B \in \R^{3\times 3}$ as a vector $b \in \R^6$ as:
\begin{equation}
    b = \begin{bmatrix}
        B_{11} & B_{12} & B_{22} & B_{13} & B_{23} & B_{33}
    \end{bmatrix}^\top .
\end{equation}
This reparameterization is useful because it allows us to rewrite the expression $\tilde{c}_i^\top B\tilde{c}_j$ as:
\begin{equation}
    \tilde{c}_i^\top B\tilde{c}_j = v_{ij}^\top  b,
\end{equation}
where:
\begin{equation*}
\begin{split}
   v_{ij} = \begin{bmatrix}
        \tilde{c}_{i1}\tilde{c}_{j1}, & \tilde{c}_{i1}\tilde{c}_{j2}+\tilde{c}_{i2}\tilde{c}_{j1}, & \tilde{c}_{i2}\tilde{c}_{j2}, & \tilde{c}_{i3}\tilde{c}_{j1} + \tilde{c}_{i1}\tilde{c}_{j3}, & \tilde{c}_{i3}\tilde{c}_{j2} + \tilde{c}_{i2}\tilde{c}_{j3}, & \tilde{c}_{i3}\tilde{c}_{j3}
    \end{bmatrix}^\top ,
\end{split}
\end{equation*}
and where $\tilde{c}_{ik}$ is the $k$-th element of the column vector $\tilde{c}_i$ and $\tilde{c}_{jk}$ is the $k$-th element of the column vector $\tilde{c}_j$. 
With this reparameterization, we can rewrite the constraints in \cref{eq:zhangconst} as:
\begin{equation*}
\begin{split}
\tilde{c}_1^\top  B \tilde{c}_2 = 0 &\implies v_{12}^\top b = 0, \\
\tilde{c}_1^\top  B \tilde{c}_1 = \tilde{c}_2^\top  B \tilde{c}_2 &\implies (v_{11} - v_{22})^\top  b = 0,
\end{split}
\end{equation*}
or by combining them:
\begin{equation} 
\label{eq:bconst}
    \begin{bmatrix}
        v_{12}^\top  \\ (v_{11} - v_{22})^\top 
    \end{bmatrix}b = 0,
\end{equation}
which is linear with respect to the unknowns vector $b$. 
Importantly, while the homographies, $H$, are different for each image due to the different extrinsic parameters, the intrinsic camera parameters represented by the vector $b$ are the same.
Therefore, with $N$ images from the same camera, even with potentially different perspectives, we can stack the constraints in \cref{eq:bconst} to give:
\begin{equation} 
\label{eq:allbconst}
    Vb = 0,
\end{equation}
where $V \in \R^{2N \times 6}$. 
In the case where we include the skewness parameter, $\gamma$, in $K$, there must be $N \geq 3$ images in order to specify $B$ uniquely. 
Similarly to the approach for computing the projection matrix in the previous section, we can compute the vector $b$ as the solution to the constrained least squares problem:
\begin{equation} 
\label{eq:bopt}
\begin{split}
	\minimize[b] & \lVert Vb \rVert^2, \\
    \subjectto & \lVert b \rVert^2 = 1.
\end{split}
\end{equation}
Once we have computed $b$, we can solve for the intrinsic camera parameters, $K$, by leveraging the definition of $B = K^{-T}K^{-1}$. 
In particular, we compute the intrinsic parameters by:
\begin{equation} 
\label{eq:B2K}
\begin{split}
    v_0 &= \frac{B_{12}B_{13} - B_{11}B_{23}}{B_{11}B_{22} - B_{12}^2}, \\
    \lambda &= B_{33} - \frac{B_{13}^2 + v_0(B_{12}B_{13} - B_{11}B_{23})}{B_{11}}, \\
    \alpha &= \sqrt{\frac{\lambda}{B_{11}}}, \\
    \beta &= \sqrt{\frac{\lambda B_{11}}{B_{11}B_{22} - B_{12}^2}}, \\
    \gamma &= \frac{-B_{12}\alpha^2\beta}{\lambda}, \\
    u_0 &= \frac{\gamma v_0}{\beta} - \frac{B_{13}\alpha^2}{\lambda},
\end{split}
\end{equation}
where we can think of $\lambda$ as a scaling parameter that accounts for the fact that there are five unknown camera intrinsic parameters but six degrees of freedom in $B$.

Once we have extracted the camera intrinsic parameters, $K$, from this procedure, given any new homography, $H$, we can compute the extrinsic parameters by:
\begin{equation}
\begin{split}
    r_1 &= \frac{K^{-1}\tilde{c}_1}{\lVert K^{-1}\tilde{c}_1 \rVert}, \\
    r_2 &= \frac{K^{-1}\tilde{c}_2}{\lVert K^{-1}\tilde{c}_2 \rVert}, \\
    r_3 &= r_1 \times r_2, \\
    t &= \frac{K^{-1}\tilde{c}_3}{\lVert K^{-1}\tilde{c}_1 \rVert}.
\end{split}
\end{equation}
As one final step, we note that the matrix $R$ defined with column vectors $r_1$, $r_2$, and $r_3$ will not generally satisfy the orthonormality property of a rotation matrix. 
We can correct this issue by again using optimization methods to compute a valid rotation matrix that best corresponds to these column vectors:
\begin{equation} 
\label{eq:Ropt}
\begin{split}
	\minimize[R] \:\:& \lVert R - Q \rVert^2, \\
    \subjectto & R^\top R = I,
\end{split}
\end{equation}
where:
\begin{equation*}
    Q = \begin{bmatrix}
        r_1 & r_2 & r_3
    \end{bmatrix}.
\end{equation*}
We solve this problem by choosing $R = UV^\top$, where $U$ and $V$ are defined by the singular value decomposition of $Q = U\Sigma V^\top $.

\subsubsection{Camera Auto-Calibration}
\label{subsec:auto-calibration}
The direct linear transformation from \cref{subsubsec:direct-linear-calibration} and Zhang's flexible calibration method from \cref{subsubsec:zhang} require point correspondences to calculate the intrinsic and extrinsic parameters.
Auto-calibration offers an alternative approach that does not make this assumption by utilizing multiple views of a static scene to determine the camera's intrinsic and extrinsic parameters.
This approach leverages the fact that a camera's intrinsic parameters remain constant across different views of the same scene.
By identifying correspondences between points in multiple images, we can estimate the intrinsic matrix, $K$, and extrinsic matrices, $R$ and $t$, by bundle adjustment based on the static scene geometry.
In total, the camera auto-calibration process consists of five key steps.

First, we use an algorithm such as scale-invariant feature transform (SIFT) to identify key points in the scene and their correspondence points across multiple images from different views.
The SIFT algorithm is a computer vision method to detect, describe, and match local features in images.
It is robust in detecting and describing local features in images, making it effective for finding correspondences under varying conditions.
Details of the SIFT algorithm will be discussed in the following chapters.

Second, we use the correspondences between a pair of images to compute the \emph{fundamental matrix}.
The fundamental matrix, $F \in \R^{3\times 3}$, relates corresponding points between a pair of images of the same scene\sidenote{The fundamental matrix describes the \emph{epipolar geometry} between two image views.}.
For a pair of images $(I, I')$ of the same scene, the fundamental matrix, $F$, satisfies:
\begin{equation}
\label{eq:fund}
    p^\top F p' = 0,
\end{equation}
where $p$ and $p'$ are corresponding points in images $I$ and $I'$, respectively.
We will discuss the fundamental matrix more in the context of stereo vision in \cref{ch:stereo-vision}.

To compute the fundamental matrix for the image pair, we first construct a point correspondence matrix, $W$, where each row represents a correspondence between points in the two images.
Let $(u_i, v_i)$ and $(u_i', v_i')$ be the coordinates of one set of matched points, $p_i$ and $p_i'$. 
We form each row of the matrix $W$ using one set of corresponding coordinates as:
\begin{equation}
    W = \begin{bmatrix}
            u_1 u_1' & u_1 v_1' & u_1 & v_1 u_1' & v_1 v_1' & v_1 & u_1'    & v_1'    & 1      \\
            \vdots   & \vdots   & \vdots & \vdots   & \vdots   & \vdots & \vdots & \vdots & \vdots \\
            u_n u_n' & u_n v_n' & u_n   & v_n u_n' & v_n v_n' & v_n   & u_n'    & v_n'    & 1
    \end{bmatrix}.
\end{equation}
From this point correspondence matrix, we then compute the fundamental matrix by solving the linear system $Wf = 0$ using a singular value decomposition (SVD), where $f$ is the vectorized form of the fundamental matrix. 
We describe the specifics of this procedure in more detail in \cref{subsubsec:epi}.
	
Third, we compute the \emph{essential matrix}, which is similar to the fundamental matrix in that it relates corresponding points in two images based on scene geometry.
For two images, $I$ and $I'$, we define the essential matrix by the rotation matrix, $R$, and translation vector, $t$, relating the coordinate frames of the two images by:
\begin{equation}
\label{eq:essential2}
    E = [t]_\times R,
\end{equation}
where $[t]_\times$ is the matrix representation of the cross product.		
Assuming the intrinsic matrix, $K$, remains the same between the images, we can also define the essential matrix, $E$, with respect to the fundamental matrix as:
\begin{equation}
\label{eq:essential}
    E = K^\top F K.
\end{equation}
Therefore, we can first compute the essential matrix from the previously computed fundamental matrix and the intrinsic parameter matrix, $K$, and then we can compute the camera extrinsic rotation and translation parameters, $R$ and $t$, by the singular value decomposition:
\begin{equation}
\label{eq:svd}
    E = U \Sigma V^\top, 
\end{equation}
where $\Sigma = \text{diag}(1, 1, 0)$.
From this decomposition, there are two possible solutions for the camera extrinsic parameters:
\begin{equation}
\begin{aligned}
    R_1 &= UWV^\top, \quad & R_2 &= UW^\top V^\top, \\
    t_1 &= U[:, 2], \quad & t_2 &= -U[:, 2],
\end{aligned}
\end{equation}
where:
\begin{equation}
    W = \begin{bmatrix}
            0 & -1 & 0 \\
            1 &  0 & 0 \\
            0 &  0 & 1
    \end{bmatrix}.
\end{equation}
We can identify the correct solution by using each to compute the three-dimensional points from the two-dimensional matched correspondence points, as we describe in the next step.
The extrinsic parameters, $R$ and $t$, we should use are the ones that ensure that most of the three-dimensional points lie in front of both cameras, meaning they will have positive depth values.

The fourth step is to perform triangulation to compute the 3D points in the scene from the 2D image correspondence points.
We define the projection matrix, $P$, for an image as:
\begin{equation}
    P = K\begin{bmatrix}
    R & t
\end{bmatrix},
\end{equation}
where again $K$ is the camera intrinsic matrix, $R$ is the rotation matrix, and $t$ is the translation vector.
This projection matrix maps a 3D point in homogeneous coordinates into the 2D camera frame coordinates by:
\begin{equation}
    \begin{bmatrix}
        u \\
        v \\
        1
    \end{bmatrix} = P
    \begin{bmatrix}
        X \\
        Y \\
        Z \\
        1
    \end{bmatrix}.
\end{equation}
Therefore, for matched points across two images, we can compute the 3D coordinates by solving the linear system:
\begin{equation}
\label{eq:triangulation}
    \begin{bmatrix}
        u P_{3}^\top - P_{1}^\top \\
        v P_{3}^\top - P_{2}^\top \\
        u' P_{3}'^\top - P_{1}'^\top \\
        v' P_{3}'^\top - P_{2}'^\top
    \end{bmatrix}
    \begin{bmatrix}
        X \\
        Y \\
        Z \\
        1
    \end{bmatrix} = 0, 
\end{equation}
where $(u,v)$ are the coordinates and $P_{1}$, $P_{2}$, and $P_{3}$ are the first, second, and third rows of the projection matrix, $P$, for image $I$, respectively, and $(u',v')$ are the coordinates and $P_{1}'$, $P_{2}'$, and $P_{3}'$ are the rows of the projection $P'$ for image $I'$.
We can solve \cref{eq:triangulation} using a least squares method.

So far, we have used correspondences in pairs of images from the same camera of the same scene to estimate the image extrinsics, $R$ and $t$, and computed estimates of the 3D scene points by triangulation.
These computations require knowledge of the camera intrinsic matrix, $K$, which is the quantity we are trying to estimate.
We can leverage the previous steps to compute $K$ by using an iterative optimization-based procedure, where we begin with an estimate\sidenote{For example, we could start with an estimate by referencing the camera manufacturer's data, or using some other simpler method.} of $K$ and refine it until convergence.
In particular, we refine the estimate of the camera intrinsics and extrinsics by solving the optimization:
\begin{equation}
\label{eq:auto-calibration-cost}
    \sum_{i=1}^N \sum_{j=1}^M \| p_{ij} - P_i(K, R_i, t_i) X_j \|^2, 
\end{equation}
where $p_{ij}$ is an observed 2D point in image $i$ that corresponds to the $j$-th 3D point, $X_j$ is the $j$-th estimated 3D point, $P_i(K, R_i, t_i)$ is the projection matrix for image $i$, $M$ is the total number of triangulated 3D points, and $N$ is the total number of images.
Note that each projection matrix is a function of the intrinsic parameters, which are constant across all images, as well as the extrinsic parameters for the image.
We can optimize this cost function by applying a nonlinear method, such as Levenberg-Marquardt, to compute a new set of parameters.
We then repeat the steps listed above, computing new extrinsics, triangulation points, and optimizing, until convergence.

\subsection{RGB-D Camera Calibration}
RGB-D cameras present additional calibration challenges beyond traditional cameras.
In addition to calibrating the RGB camera using standard methods, we must also:
\begin{itemize}
    \item Calibrate the depth sensor's intrinsic parameters
    \item Determine the extrinsic transformation between RGB and depth sensors  
    \item Correct for systematic depth measurement errors
    \item Account for the different fields of view and resolutions of the two sensors
\end{itemize}
Many manufacturers provide factory calibration, but applications requiring high accuracy often necessitate custom calibration procedures using specialized targets that are visible in both RGB and infrared.

\section{Summary}
This chapter presented the mathematical foundations of camera models and calibration techniques essential for vision-based robotics.
Starting from the basic pinhole model, we developed the perspective projection equations that relate 3D world points to 2D image coordinates.
We explored practical considerations including lens models and distortion, and extended our discussion to modern RGB-D sensors that provide both color and depth information.
The calibration methods presented enable us to estimate the camera parameters necessary for quantitative vision applications.
With calibrated cameras, we can now proceed to extract three-dimensional information from images, which we explore in the next chapter through stereo vision and structure from motion techniques.

\paragraph{To learn more.}
For a deeper dive into camera models and calibration techniques, readers are encouraged to refer to the \textit{Foundations of Computer Vision} textbook by \citet{torralba2024foundations} for a more in-depth discussion on image systems and camera models. For practical calibration algorithms and implementations, the OpenCV library\cite{Bradski2000} provides extensive resources and code examples. Interested readers can consult the works directly from \citet{Tsai1987}, \citet{lowe1999object}, and \citet{triggs2000bundle} for further insights into the algorithms discussed.

\section{Exercises}
The starter code for the exercises provided below is available online through GitHub. 
To get started, download the code by running in a terminal window:

\begin{tcolorbox}[colback=gray!10]
\begin{minted}{bash}
    git clone https://github.com/StanfordASL/pora-exercises.git
\end{minted}
\end{tcolorbox}

We denote Problems requiring hand-written solutions and coding in Python with \adjustbox{height=2ex, valign=c}{\includegraphics{figs/write.png}} and \adjustbox{height=2ex, valign=c}{\includegraphics{figs/code.png}}, respectively.

\subsection*{\adjustbox{height=2ex, valign=c}{\includegraphics{figs/code.png}}\ Problem 1: Camera Calibration: Extrinsics}
In this exercise, you will implement the key parts of a method to compute the camera extrinsic parameters $R$ and $t$.
Specifically, you will use the steps in \cref{subsubsec:dlc_step1} to compute the homography matrix $H$ given a calibration image.
Then, given the camera intrinsic matrix $K$, you will use the method at the end of \cref{subsubsec:zhang} to compute the extrinsics, $R$ and $t$, for the image.
For this exercise, you will use a chessboard with known dimensions to help with the camera calibration process.
This is a convenient choice because it provides a high contrast and the grid corner features are easy to detect.

In the file \colorcode{ch06/exercises/camera\_extrinsics.ipynb}:
\begin{enumerate}
\item First, implement the function \colorcode{generate\_chessboard\_3D\_world\_coordinates} to compute a grid of 3D world coordinates that correspond to the grid corners of the chessboard.
Note that we can choose to set all of the z-coordinates to be $0$ since we know the chessboard is a 2D plane.
Then, implement the function \colorcode{generate\_chessboard\_2D\_pixel\_coordinates} to compute the 2D pixel coordinates of the grid corners from an image using the function \colorcode{findChessboardCorners} from the open-source computer vision package \colorcode{cv2}.
\item Second, implement the function \colorcode{compute\_homography} using the method from \cref{subsubsec:dlc_step1}.
Note that generally in \cref{subsubsec:dlc_step1} we are computing the projection matrix $M \in \R^{3 \times 4}$, but in this problem we have a slightly simplified problem of computing the homography matrix $H \in \R^{3 \times 3}$ which just maps points between two 2D planes (since we know the chessboard grid points lie on a common plane).
In other words, when using the approach from \cref{subsubsec:dlc_step1} rather than dealing with rows $m_i \in \R^{1\times4}$ you should be considering rows of the homography matrix $h_i \in \R^{1\times3}$ and be using the vectors $P^h_{W} = [X_{W}, Y_W, 1]^\top$.
Once the homography matrix is available, implement the function \colorcode{compute\_extrinsics} using the method at the end of \cref{subsubsec:zhang} to compute the extrinsics $R$ and $t$ given the intrinsic matrix $K$.
Additionally, implement the function \colorcode{transform\_world\_to\_pixel} to transform a point in the world frame into the pixel coordinates from the camera intrinsics and extrinsics.
Run the provided code to see the result of your computations on some example chessboard images.
\end{enumerate}

\subsection*{\adjustbox{height=2ex, valign=c}{\includegraphics{figs/code.png}}\ Problem 2: Camera Calibration: Intrinsics}
In Problem 1, we provided the camera intrinsic matrix, $K$. In this exercise\sidenote{Note you will need to first complete Problem 1 since this will leverage some of that problem's code.}, you will use the flexible calibration method described in \cref{subsubsec:zhang} to compute the intrinsic matrix yourself.
Specifically, explore the notebook \newline\noindent\colorcode{ch06/exercises/camera\_intrinsics.ipynb}.
You will need to implement the function \colorcode{compute\_intrinsics} to compute the matrix $K$ from a list of homography matrices from different calibration images.
\newpage
\printbibliography[segment=\therefsegment,heading=subbibliography,title={References}]
\chapter{Stereo Vision and Structure From Motion}
\label{ch:stereo-vision}
\newrefsegment
In \cref{ch:cameras}, we introduced the mathematical relationship between the position of a point, $P$, in a scene, expressed in world frame coordinates, $P_W$, and the corresponding point, $p$, in pixel coordinates that gets projected onto the image plane of a camera. 
This relationship is based on the pinhole camera model, and requires knowledge about the camera's intrinsic and extrinsic parameters.
We also presented methods for camera calibration, which allow us to determine these parameters.
Given a calibrated camera with known parameters, a fundamental problem in robotic perception is how to leverage images to recover three-dimensional information about the structure of the environment\sidenote{While we could use other sensors to recover three-dimensional scene information, such as ultrasonic sensors or laser rangefinders, cameras capture a broad range of information that goes beyond depth sensing and are attractive based on their cost and size.}.
The camera projection model alone does not provide us with enough information to fully determine the 3D position of a point from a single image, specifically because we cannot determine the point's \emph{depth}\sidenote{Unless you are willing to make strong assumptions, for example that you know the physical dimensions of the objects in the environment.}.

In this chapter, we introduce \emph{stereo vision} in Section~\ref{sec:ch08_stereovision} and \emph{structure from motion} in Section~\ref{sec:ch08_sfm}, two approaches for extracting 3D information from camera images.
Both leverage multiple images of a scene to determine three-dimensional structure: stereo vision uses images from different viewpoints captured simultaneously by two or more cameras, while structure from motion uses images captured sequentially from a single moving camera.
These techniques form the visual foundation for many robotic applications including navigation, mapping, and manipulation.

\section{Stereo Vision}
\label{sec:ch08_stereovision}
Stereopsis\sidenote{From \emph{stereo}, meaning solidity, and \emph{opsis}, meaning vision or sight.} is the process in visual perception leading to the sensation of depth from two slightly different projections of the world onto the retinas of the two eyes.
The difference in the two retinal images is called horizontal \emph{disparity}, retinal disparity, or binocular disparity, and arises from our eyes' different positions in the head. 
This disparity enables our brain to fuse the two retinal images into a single percept with depth information. 
For example, if you hold your finger vertically in front of you and alternate closing each eye, you will see that the finger jumps from left to right—this lateral displacement is the disparity between your eyes.

Computational stereopsis, or \emph{stereo vision}, is the process of obtaining depth information from images captured by two or more cameras observing the same scene from different perspectives. 
This process consists of two major steps: fusion and reconstruction. 
Fusion involves solving the correspondence problem—identifying which pixels in each image correspond to the same 3D point. 
Reconstruction uses these correspondences to triangulate the 3D position of scene points, including their depth.

\subsubsection{Epipolar Geometry}
\label{subsubsec:epi}
The first step in stereo vision is to establish correspondences between images\sidenote{We generally assume that the perspectives differ only slightly, such that features appear similar across images.}. 
This task can be challenging, as incorrect matches lead to large reconstruction errors. 
Epipolar geometry provides powerful constraints that simplify correspondence search and improve matching accuracy.

\paragraph{Epipolar constraints.}
\begin{figure}[ht]
  \begin{center}
	\includegraphics[width=.8\textwidth]{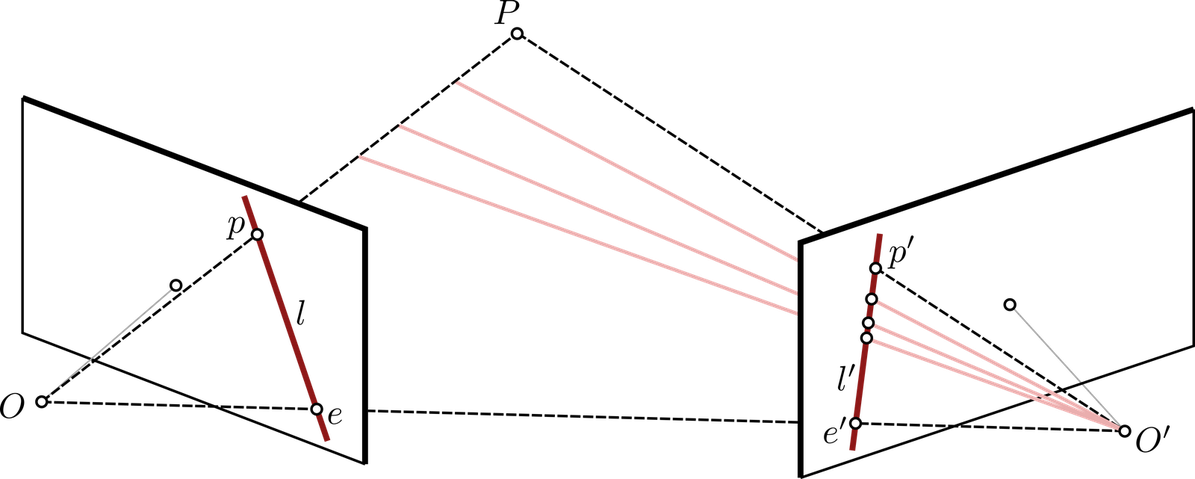}
  \end{center}
  \caption{The point $P$ in the scene, the optical centers $O$ and $O'$ of the two cameras, and the two image projections $p$ and $p'$ all lie in the same epipolar plane. 
  The lines $l$ and $l'$ are the epipolar lines. 
  If point $p$ is observed in one image, its correspondence must lie on the epipolar line $l'$ in the other image.}
  \label{fig:epi}
\end{figure}

Consider the image projections $p$ and $p'$ of a scene point $P$ observed by two cameras with optical centers $O$ and $O'$, as shown in \cref{fig:epi}.
These five points all lie in the \emph{epipolar plane}, defined by the two rays $OP$ and $O'P$. 
The intersection of this plane with each image plane forms the \emph{epipolar lines} $l$ and $l'$, which pass through the \emph{epipoles} $e$ and $e'$—the projections of each camera center onto the other camera's image plane.
This geometric relationship provides a powerful constraint: if $p$ and $p'$ are projections of the same point $P$, then $p$ must lie on epipolar line $l$ and $p'$ must lie on epipolar line $l'$. 
This \emph{epipolar constraint} reduces the correspondence search from a two-dimensional problem to a one-dimensional search along epipolar lines.
Mathematically, we express this constraint using the coplanarity of the vectors:
\begin{equation}
\overline{Op} \cdot [\overline{OO'} \times \overline{O'p'}] = 0.
\end{equation}

\paragraph{Fundamental and essential matrices.}
When the world reference frame coincides with the first camera's frame (origin at $O$), we can express the epipolar constraint as:
\begin{equation} 
\label{eq:epiconst}
    p^\top F p' = 0,
\end{equation}
where $F \in \R^{3 \times 3}$ is the \emph{fundamental matrix}.
The fundamental matrix encodes the epipolar geometry between two views and has seven degrees of freedom (it is defined up to scale and has rank 2).
It depends only on the cameras' intrinsic parameters and their relative pose:
\begin{equation}
    F = K^{-\top}EK'^{-1},
\end{equation}
where $K$ and $K'$ are the intrinsic parameter matrices for the two cameras, and $E$ is the \emph{essential matrix}:
\begin{equation}
    E = [t]_\times R = \begin{bmatrix}
    0 & -t_3 & t_2 \\
    t_3 & 0 & -t_1 \\
    -t_2 & t_1 & 0
    \end{bmatrix}R,
\end{equation}
with $R$ and $t = \vCol{t_1, t_2, t_3}$ being the rotation and translation that transform points from the second camera frame to the first.

The fundamental matrix also defines the epipolar lines: $l = Fp'$ and $l' = F^\top p$. 
The epipoles satisfy $F^\top e = 0$ and $Fe' = 0$, confirming that $F$ is singular (rank 2).
To estimate $F$ from image correspondences, we use the fact that each correspondence $(p_i, p'_i)$ provides one linear constraint.
With $p = \vCol{u, v, 1}$ and $p' = \vCol{u', v', 1}$ in homogeneous coordinates, we can rewrite the epipolar constraint as:
\begin{equation}
\begin{bmatrix}
uu' & uv' & u & vu' & vv' & v & u' & v' & 1
\end{bmatrix}f = 0,
\end{equation}
where $f$ is the vectorized form of $F$.
Given $n \geq 8$ correspondences, we stack these constraints into a matrix equation $Wf = 0$ and solve:
\begin{equation} 
\label{eq:fopt}
\begin{split}
\minimize[f] & \lVert Wf \rVert^2,\\
\subjectto & \lVert f \rVert^2 = 1.
\end{split}
\end{equation}
The solution is the eigenvector corresponding to the smallest eigenvalue of $W^\top W$.
Since the resulting matrix may not have rank 2, we enforce the singularity constraint by computing the SVD of $\tilde{F}$ and setting the smallest singular value to zero.

\paragraph{Image rectification.}
Epipolar rectification transforms stereo image pairs such that epipolar lines become horizontal and aligned across images.
This transformation simplifies correspondence search to a one-dimensional problem along image rows, significantly reducing computational cost.
The rectified configuration is equivalent to having two cameras with parallel optical axes and aligned image rows, separated by a baseline distance.
\begin{figure}[ht]
  \begin{center}
	\includegraphics[width=0.7\textwidth]{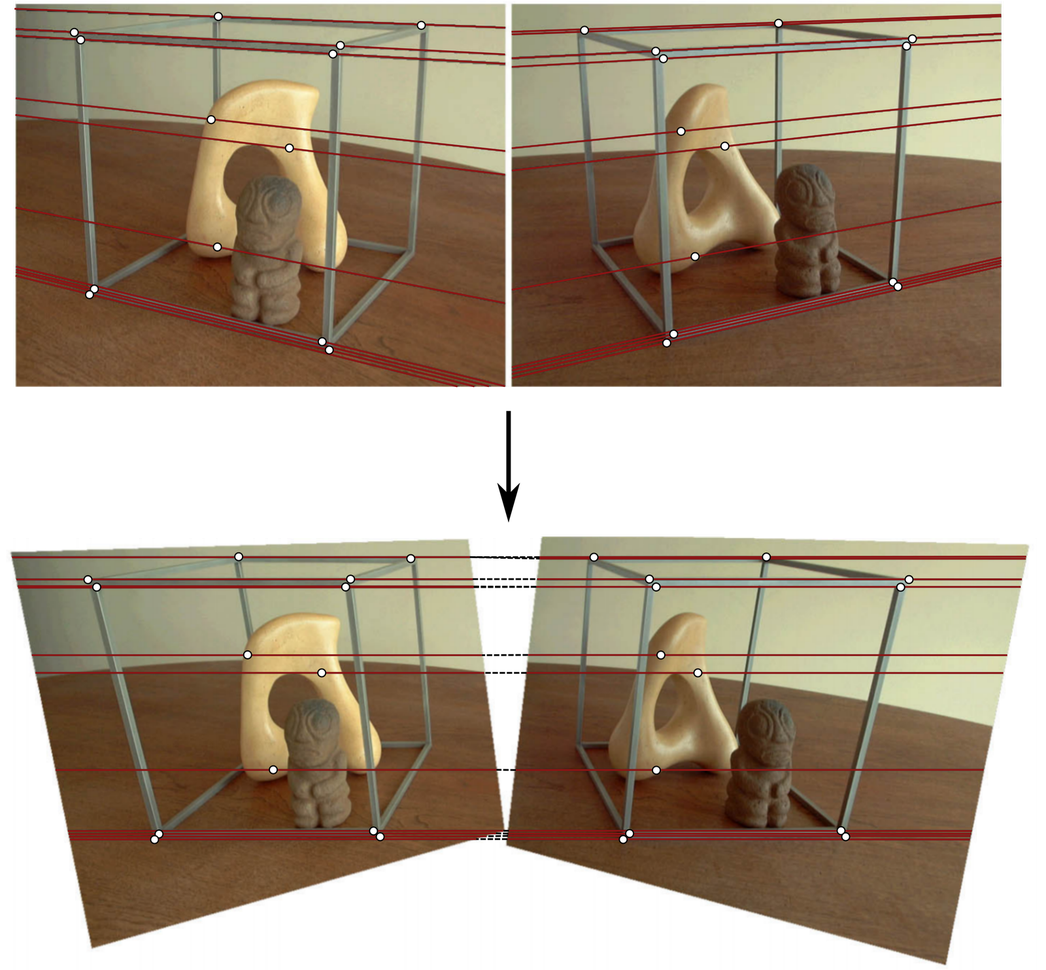}
  \end{center}
  \caption{Epipolar rectification transforms images so that corresponding points lie on the same horizontal scan line. Original epipolar lines (left) become horizontal and aligned after rectification (right).}
  \label{fig:rect}
\end{figure}
After rectification, as shown in \cref{fig:rect}, correspondence search is constrained to horizontal scan lines, making stereo matching algorithms more efficient and robust.

\subsection{Correspondence and Reconstruction}

With the geometric constraints established through epipolar geometry and simplified through image rectification, we now turn to the practical challenges of stereo vision.
The correspondence problem—determining which pixels in each image represent the same scene point—remains the most critical and challenging step, as errors here propagate directly to the reconstructed 3D structure.
Once reliable correspondences are found, triangulation transforms these matched points into 3D coordinates, with the accuracy depending fundamentally on the system geometry and image measurements.

\paragraph{The correspondence problem.}
Even with epipolar constraints and rectification, finding correct correspondences remains challenging.
Occlusions occur when points visible in one view are hidden in another, while repetitive patterns in textured regions create ambiguous matches where multiple locations appear identical.
Conversely, uniform regions lack sufficient texture for reliable feature matching, and perspective distortions cause features to appear different across viewpoints despite representing the same scene point.
Modern stereo matching algorithms address these challenges through robust feature descriptors, correlation-based matching over local windows, or learned representations that capture scene semantics beyond low-level appearance.

\begin{figure}[ht]
\centering
\includegraphics[width=0.9\textwidth]{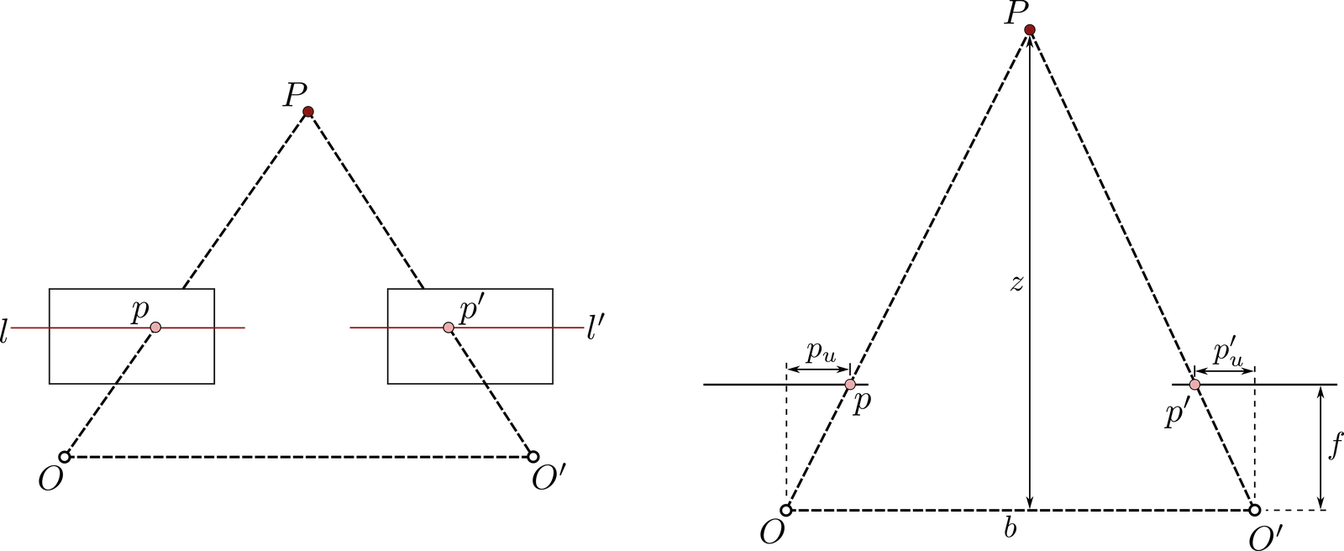}
\caption{Triangulation with rectified stereo cameras. The depth $z$ is inversely proportional to the disparity $d = p_u - p'_u$.}
\label{fig:recttri}
\end{figure}

\paragraph{Triangulation and disparity.}
Once correspondences are established, we reconstruct 3D points through triangulation.
For rectified stereo pairs with parallel optical axes separated by baseline $b$, the geometry simplifies considerably, as shown in \cref{fig:recttri}.
From similar triangles, we derive the fundamental stereo equation:
\begin{equation}
    z = \frac{bf}{d},
\end{equation}
where $z$ is depth, $b$ is baseline, $f$ is focal length, and $d \coloneqq p_u - p'_u$ is the \emph{disparity}—the difference in horizontal coordinates between corresponding points.

This inverse relationship between depth and disparity has important implications for stereo system design.
Near objects produce large disparities and thus accurate depth estimates, while distant objects yield small disparities with correspondingly less accurate depth measurements.
Since depth resolution decreases quadratically with distance, baseline selection becomes critical: larger baselines improve depth accuracy for distant objects but increase occlusions where one camera cannot see points visible to the other.
This fundamental trade-off must be considered when designing stereo systems for specific applications.

\begin{figure}[ht]
\centering
\includegraphics[width=0.95\textwidth]{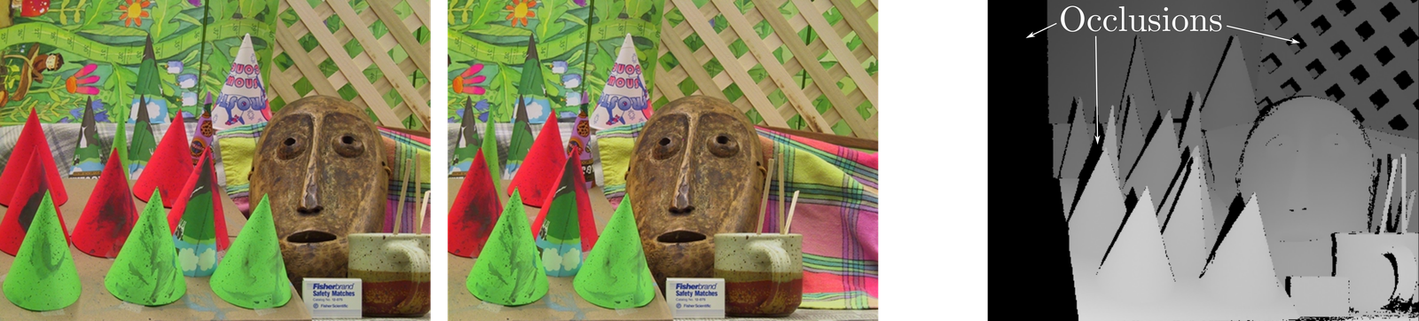}
\caption{Disparity map from stereo images. Brighter values indicate larger disparities (closer objects). Black regions are occlusions where correspondence cannot be established.}
\label{fig:disparity}
\end{figure}

\paragraph{Disparity maps.}
A disparity map encodes the disparity value for each pixel, providing a dense depth representation of the scene.
\cref{fig:disparity} shows an example where brighter regions indicate larger disparities (closer objects) and black regions represent occluded areas where no correspondence exists.
Dense disparity estimation extends the correspondence problem from sparse feature matching to every pixel, requiring additional regularization to handle ambiguous regions while preserving depth discontinuities at object boundaries.

\section{Structure From Motion}
\label{sec:ch08_sfm}
The \emph{structure from motion (SFM)} method uses a similar principle as stereo vision, but uses a single camera to capture multiple images from different perspectives while moving within the scene. 
In this case, the intrinsic camera parameter matrix, $K$, will be constant across images, but the extrinsic parameters consisting of the rotation matrix, $R$, and relative position vector, $t$, will be different for each image.
\begin{figure}[ht]
  \begin{center}
	\includegraphics[width=0.75\textwidth]{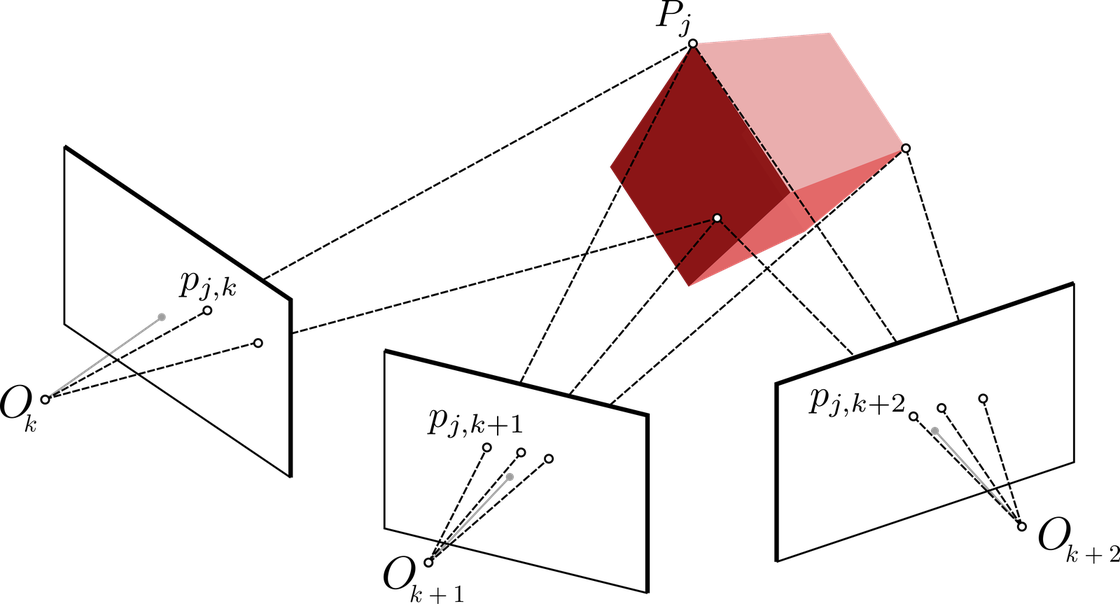}  \end{center}
  \caption{A depiction of the structure from motion (SFM) method. 
  A single camera is used to take multiple images from different perspectives, which provides enough information to reconstruct the 3D scene.}
  \label{fig:sfm}
\end{figure}
Consider a case where we take $m$ images of $n$ fixed 3D points from different perspectives. 
This would lead to $m$ projection matrices, $\mathcal{P}_k = K[R_k \: t_k]$, and $n$ 3D points, $P_j$, that we would need to determine by leveraging the projection relationships:
\begin{equation*}
    p_{j,k}^h = \mathcal{P}_k P^h_j, \quad j = 1,\dots,n, \quad k=1,\dots,m.
\end{equation*}
Notice that there is quite a bit of similarity between this problem and the camera auto-calibration problem discussed in \cref{ch:cameras}, except here we assume we already know the camera intrinsic parameters.

The fundamental challenge in structure from motion is to simultaneously recover both the camera poses and the 3D point positions from only the 2D image observations.
This requires solving a large optimization problem with many unknowns, which we approach either incrementally by adding one image at a time, or globally by considering all images simultaneously.
Both approaches rely on establishing feature correspondences across images and minimizing reprojection errors through bundle adjustment.
Structure from motion methods also have some unique limitations.
Most notably, there exists an inherent ambiguity in the absolute scale of the scene that cannot be determined from images alone\sidenote{For example, a bigger object at a longer distance and a smaller object at a closer distance can yield identical projections.}.
Additionally, errors accumulate over long sequences, leading to drift that must be corrected through loop closure when the camera revisits a previous location.
Despite these challenges, structure from motion has proven invaluable for robotic applications where carrying multiple cameras is impractical.

\subsubsection{Two-View Geometry}
The structure from motion pipeline begins with reconstructing the scene from two views, establishing the initial geometry that subsequent images will extend.
Given correspondences between two images taken from positions $O_1$ and $O_2$, we first estimate the fundamental matrix using the same techniques described for stereo vision.
With known intrinsics, we recover the essential matrix $E = K^\top F K$ and decompose it to obtain the relative camera pose.

The decomposition of the essential matrix yields four possible solutions for the rotation and translation $(R, t)$.
We identify the correct solution by verifying that reconstructed points have positive depth in both camera views.
Once the relative pose is established, we triangulate the matched features to obtain initial 3D points.
For each correspondence $(p_{j,1}, p_{j,2})$, the 3D point $P_j$ satisfies:
\begin{equation}
    \begin{bmatrix}
        (p_{j,1} \times) \mathcal{P}_1 \\
        (p_{j,2} \times) \mathcal{P}_2
    \end{bmatrix} P_j = 0,
\end{equation}
where $\mathcal{P}_1$ and $\mathcal{P}_2$ are the projection matrices for cameras at $O_1$ and $O_2$, and $(p \times)$ denotes the skew-symmetric matrix formed from the homogeneous coordinates of $p$.
This linear system is solved using least squares, typically with singular value decomposition for numerical stability.

\subsubsection{Sequential Reconstruction}
Sequential structure from motion extends the initial two-view reconstruction by incrementally incorporating new images, as illustrated in \cref{fig:sfm}.
Each new image captured from position $O_k$ undergoes pose estimation through perspective-n-point (PnP) solving, which finds the camera position $(R_k, t_k)$ that minimizes reprojection error for correspondences between image points $p_{j,k}$ and existing 3D points $P_j$.
Once registered, the image contributes new scene points through triangulation with previously registered views from positions $O_{k-1}, O_{k-2}, \ldots$.

The critical challenge is maintaining consistency as errors accumulate.
Bundle adjustment jointly optimizes camera poses and 3D points to minimize total reprojection error:
\begin{equation}
\minimize[\{R_k, t_k\}, \{P_j\}] \sum_{k} \sum_{j \in \mathcal{V}_k} \lVert p_{j,k} - \pi(K, R_k, t_k, P_j) \rVert^2,
\end{equation}
where $p_{j,k}$ represents the observed projection of point $P_j$ in the image taken from camera position $O_k$, and $\mathcal{V}_k$ denotes the set of points visible from viewpoint $k$.
However, global optimization after each image is computationally prohibitive.
Instead, we perform local bundle adjustment over a sliding window of recent images (typically spanning positions $O_{k-w}$ through $O_k$), maintaining local accuracy while deferring global consistency to a final optimization step.
This windowed approach achieves near-global accuracy at a fraction of the computational cost, making the method practical for long image sequences.
Image selection order also affects reconstruction quality—we typically choose the next viewpoint that maximizes correspondences with the current reconstruction, balancing accurate pose estimation with effective triangulation of new points.

\paragraph{Global structure from motion.}
An alternative to sequential reconstruction is global structure from motion, which estimates all camera poses simultaneously before triangulating points.
This approach first constructs an epipolar graph connecting all image pairs with sufficient matches, then solves for all rotations and translations globally through rotation and translation averaging.
While potentially more accurate than incremental methods, global approaches require solving large optimization problems and may be less robust to outliers in practice.
The choice between sequential and global methods often depends on the specific application requirements and computational resources available.

\subsubsection{Visual Odometry}
One particularly important application of the structure from motion concept is \emph{visual odometry}, which estimates robot motion in real-time using visual input. 
Visual odometry prioritizes speed and local accuracy over global consistency, maintaining only a sliding window of recent frames and performing limited bundle adjustment.
This approach operates at camera frame rates by trading global optimality for computational efficiency.
Visual odometry has proven invaluable for robot navigation, particularly in environments where wheel odometry is unreliable or unavailable.
Mars rovers, for instance, rely on visual odometry to navigate terrain where wheel slip would cause traditional odometry to fail dramatically\sidenote{The Mars Exploration Rovers Spirit and Opportunity used stereo visual odometry to traverse over 45 kilometers combined, far exceeding their planned 600-meter missions.}.
The technique also enables navigation for flying robots and underwater vehicles where wheel odometry is impossible.

Modern visual odometry systems often combine multiple approaches for robustness.
Feature-based methods track distinctive image features across frames, providing robustness to illumination changes but potentially failing in textureless environments.
Direct methods minimize photometric error using raw pixel intensities, exploiting all image information but requiring good initialization and small inter-frame motions.
Hybrid approaches leverage the strengths of both, using features for robustness and direct alignment for accuracy.

\subsubsection{Loop Closure and Drift Mitigation}
Over extended sequences, small errors in structure from motion accumulate into significant drift.
Loop closure detection identifies when the camera revisits a previous location, providing constraints to correct this accumulated error globally.
This involves recognizing previously seen places despite changes in viewpoint and lighting, typically using visual vocabularies or learned features, followed by geometric verification and global optimization incorporating the loop constraints.
Successfully detecting and closing loops transforms structure from motion from a local reconstruction technique into a method capable of mapping large environments—a capability essential for autonomous navigation.

\section{Summary}
This chapter presented two fundamental approaches for extracting 3D information from camera images.
Stereo vision leverages simultaneous views from multiple cameras, using epipolar geometry to constrain correspondence search and enable real-time depth estimation through triangulation.
Structure from motion reconstructs scene geometry from sequential images captured by a moving camera, trading real-time performance for the flexibility of a single-camera system.
Both techniques face similar challenges—establishing correspondences between images despite occlusions, repetitive textures, and perspective distortions.
The reconstruction accuracy depends fundamentally on baseline configuration: stereo systems use fixed baselines while structure from motion adapts its effective baseline through camera motion.
However, structure from motion suffers from scale ambiguity that must be resolved through additional sensors or known scene dimensions.
These complementary approaches have enabled numerous robotic applications, from Mars rover navigation using visual odometry to autonomous vehicle perception using stereo depth.
Modern systems increasingly combine both techniques with other sensors—visual-inertial odometry fuses structure from motion with IMU data to recover metric scale, while stereo visual odometry leverages multiple cameras with motion estimation for robust navigation.

\paragraph{To learn more.}
For a comprehensive treatment of stereo vision and structure from motion, including advanced algorithms and practical implementations, readers are encouraged to consult the textbooks \textit{Introduction to Autonomous Mobile Robots} by \citet{SiegwartNourbakhshEtAl2011} and \textit{Computer Vision: A Modern Approach} by \citet{ForsythPonce2011}. Additionally, for a deeper understanding of epipolar geometry and multi-view reconstruction, readers can refer to the works \citet{Fusiello2000} and \citet{LoopZhang1999}.
\newpage
\printbibliography[segment=\therefsegment,heading=subbibliography,title={References}]
\chapter{Classical Methods for Perception}
\label{ch:classical_perception}
\newrefsegment
The previous chapters focused on using camera models to identify the relationship between points in a 3D scene and their projections onto the camera image, as well as how to leverage those models to reconstruct 3D scene structure from images. 
In this chapter, we introduce methods for extracting various types of information from images through both low-level image processing and higher-level feature extraction techniques.

We begin with image processing fundamentals, including filtering, feature detection, and description in Section~\ref{sec:ch09_imgprocessingfundamentals}. We then discuss geometric feature extraction methods in Section~\ref{sec:ch09_geometricfeaturesextraction} for identifying structure in sensor data. Finally, in Section~\ref{sec:ch09_featurebaseddetection}, we cover feature-based object detection approaches and discuss how classical perception methods remain relevant in modern robotics applications.

\subsection{Image Processing Fundamentals}
\label{sec:ch09_imgprocessingfundamentals}

At its core, image processing is a form of signal processing where the input signal is an image, such as a photo or a video, and the output is either an image or a set of parameters associated with the image. Extracting visual content from raw images is important for mobile robots to be able to intelligently interpret their surroundings\sidenote{Information extracted through image processing can have a significant impact on a robot's ability to perform fundamental tasks including localization, mapping, and decision making.}. While a large number of image processing techniques exist, in this chapter, we focus on some of the more fundamental methods that are relevant for robotics\cite{SiegwartNourbakhshEtAl2011}. 

\subsubsection{Image Filtering}
\begin{figure}[ht]
  \centering
  \includegraphics[width=.75\textwidth]{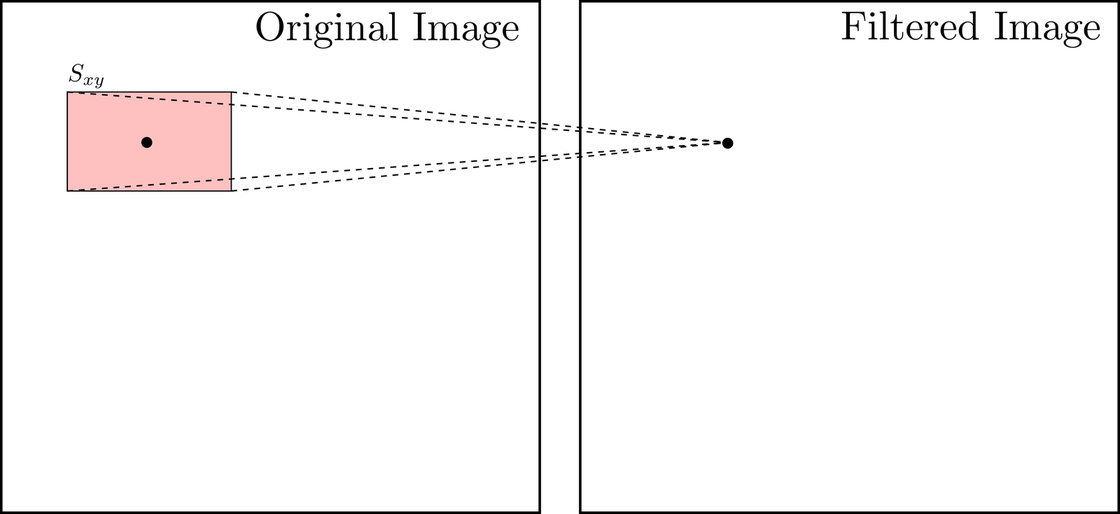}
    \caption{Illustration of the concept of spatial filtering. 
    The spatial filter operates on a neighborhood, $S_{xy}$, of each point in the original image to produce a new pixel in the filtered image.}
    \label{fig:spatial_filter_concept_fig}
\end{figure}

Image filtering is one of the principal tasks in image processing. 
The term \emph{filter} comes from frequency domain signal processing and refers to the process of accepting or rejecting certain frequency components of a signal\sidenote{For example, eliminating high-frequency noise is a classic filtering problem.}.
Perhaps the most common type of image filtering is \emph{spatial filtering}.
The basic principle of spatial filtering is that a particular pixel is modified in the filtered image based on the pixels in the immediate spatial neighborhood, as we show in \cref{fig:spatial_filter_concept_fig}.

Mathematically, we describe an image as a function, $I(x,y)$, that maps a pixel at coordinate $(x,y)$ in the domain $[a,b]\times[c,d]$ to either a scalar for grayscale images or a three-dimensional vector corresponding to red, green, and blue values for color images.
A spatial filter for an image, $I(x,y)$, consists of a neighborhood of pixels around a particular point, $(x,y)$, under examination, which we denote as $S_{xy}$\sidenote{This region is typically rectangular.}, and a predefined operation, $F$, that is performed on the image pixels encompassed by the neighborhood $S_{xy}$.
We define a new image, $I'(x,y)$, by applying the spatial filter operation $F$ to all pixels, $(x,y)$, in the original image, $I$.

In general, filters can leverage linear or nonlinear operations, but many of the most fundamental filters are linear and we can express them mathematically as:
\begin{equation}
\label{eq:correlation}
    I'(x,y) = F \circ I = \sum_{i=-N}^N \sum_{j=-M}^M F(i,j)I(x+i,y+j),
\end{equation}
where $N$ and $M$ are integers that define the width and height of a rectangular neighborhood, $S_{xy}$. 
Based on the size of this neighborhood, we say that this filter is of size $(2N+1) \times (2M+1)$. 
We generally refer to the filter operation $F$ as a \emph{mask} or \emph{kernel}. 
Broadly speaking, we refer to filters expressed by \cref{eq:correlation} as \emph{correlation filters}.

\emph{Convolution} filters are another class of linear filters that we commonly use.
Convolution filters are similar to correlation filters, but use reverse image indices\sidenote{In fact, correlation and convolution filters are identical when the filter mask is symmetric in both the horizontal and vertical directions.}.
In particular, we express convolution filters mathematically by:
\begin{equation}
\label{eq:convolution}
    I'(x,y) = F \ast I = \sum_{i=-N}^N \sum_{j=-M}^M F(i,j)I(x-i,y-j).
\end{equation}
Convolution filters are associative, meaning that for two different filter masks, $F$ and $G$, it is true that $F*(G*I) = (F*G)*I$. 
This associative property is useful for tasks such as smoothing an image before applying a differentiation filter. 
Suppose the mask $F$ implements a derivative filter and $G$ implements a smoothing filter, then sequentially applying these filters would result in $F*(G*I)$. 
However, because of the associative property, we can convolve the masks together first such that only the single filter $(F*G)*I$ needs to be applied to the image.

\begin{figure}[ht]
  \centering
  \includegraphics[width=.65\textwidth]{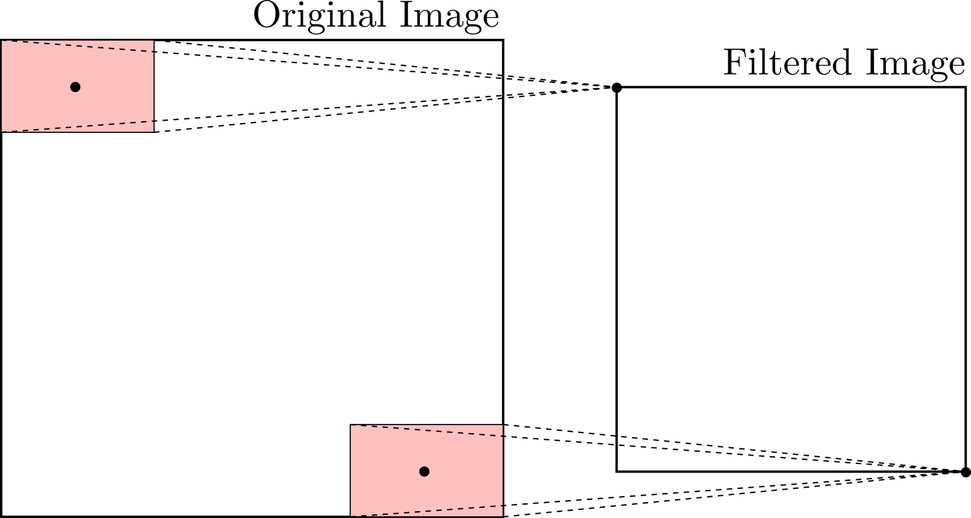}
    \caption{Due to the width and height of the mask, the filtered image may be smaller than the original. 
    This can be fixed with several techniques, such as padding.}
    \label{fig:nopaddingexample}
\end{figure}

Note that in both correlation and convolution filters, the boundaries of the image need some special care because of the width and height of the mask. 
For example, in \cref{fig:nopaddingexample} we show how the filtered image is smaller than the original due to the width and height of the mask. 
Some possible options to handle this include padding the image, cropping it, extending it, or wrapping it. 
However, as images are generally quite large relative to the mask size, the exact approach likely won't vary the final result significantly.

\begin{example}[Practical tricks for image filtering] 
\label{ex:padding}
\theoremstyle{definition}
When implementing correlation and convolution filters, we can leverage special tricks to simplify the process. 
In this example, we introduce two simplification tricks: a change in indexing and zero-padding. 

First, to accommodate varying sizes of filters, including even and odd sized filters, we can change the indexing such that the coordinate of interest is associated with the top-left element in the window rather than the center. 
For a correlation filter, this would correspond to:
\begin{equation}
\label{eq:correlation_newindex}
    I'(x,y) = F \circ I = \sum_{i=0}^{K-1} \sum_{j=0}^{L-1} F(i,j)I(x+i, y+j),
\end{equation}
where $K$ and $L$ are integers that define the width and height of the filter, and the pixel $(x,y)$ is at row $x$ and column $y$. 
Note that this formulation results in an output image, $I'$, that is shifted up and to the left. 
To see this shift, consider the top-left pixel at $x=0$ and $y=0$ in the new image, $I'$. 
We generate this new pixel value by applying the filter, $F$, over the pixels in the original image at rows $\{0,\dots,K-1\}$ and columns $\{0,\dots,L-1\}$, which is not centered at $(0,0)$ in the original image, $I$.
In practice, this shifting is not an issue as long as we always index with respect to the top-left corner. 
We show an example of top-left indexing in \cref{fig:topleftfilter}.
\begin{figure}[ht]
  \centering
  \includegraphics[width=.65\textwidth]{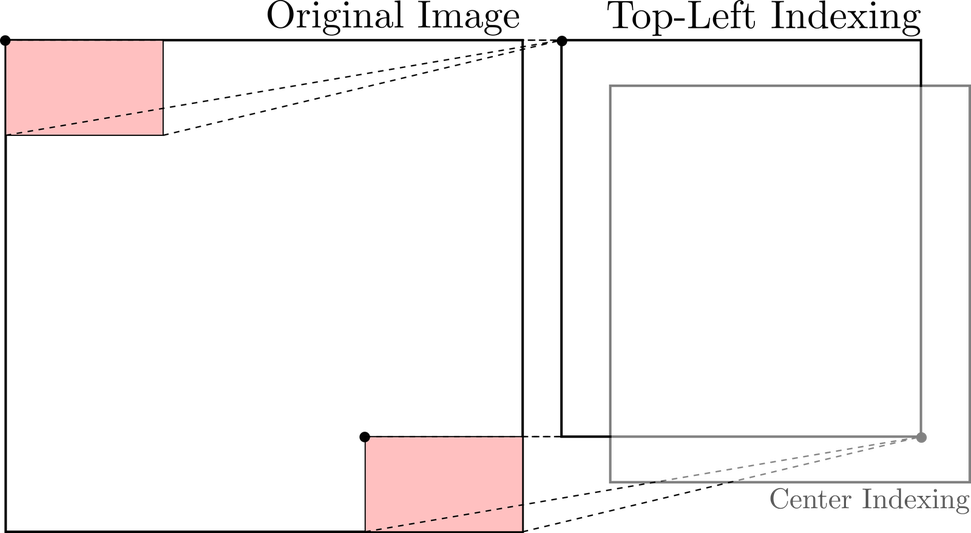}
    \caption{Top-left indexing is typically easier to implement than center indexing. 
    Notice that when top-left indexing, it appears as if the filtered image has shifted with respect to when we use center indexing.}
    \label{fig:topleftfilter}
\end{figure}

Zero-padding\sidenote{Also commonly referred to as \emph{same padding}.} is another simple trick that we can use to ensure that the output filtered image, $I'$, has the same dimension as the input image, $I$. 
In this approach, we pad the left and right boundaries of the image by $\lfloor K/2 \rfloor$ columns of zeros, and pad the top and bottom boundaries by $\lfloor L/2 \rfloor$ rows of zeros, where $\lfloor \cdot \rfloor$ denotes the \emph{floor} operation. 
For example, the image:
\begin{equation*}
I = \begin{bmatrix}
    1 & 2 & 3 \\
    4 & 5 & 6 \\
    7 & 8 & 9 \\
    \end{bmatrix},
\end{equation*}
becomes:
\begin{equation*}
I_\text{padded} = \begin{bmatrix}
    0 & 0 & 0 & 0 & 0 \\
    0 & 1 & 2 & 3 & 0 \\
    0 & 4 & 5 & 6 & 0 \\
    0 & 7 & 8 & 9 & 0 \\
    0 & 0 & 0 & 0 & 0 \\
    \end{bmatrix},
\end{equation*}
for filters $F \in \R^{3\times 3}$, $F \in \R^{2 \times 2}$, $F \in \R^{2 \times 3}$ and $F \in \R^{3 \times 2}$. 
When using this padding rule with a correlation filter from \cref{eq:correlation_newindex} and a filter, $F$, with $K = 2,3$ and $L= 2,3$, we can define the new image, $I'$, for values $x \in \{1,2,3\}$ and $y \in \{1,2,3\}$, resulting in $I'$ being the same dimension as the original image, $I$. 
We show an example use of padding combined with top-left indexing graphically in \cref{fig:paddingfilter}.
\begin{figure}[ht]
  \centering
  \includegraphics[width=.7\textwidth]{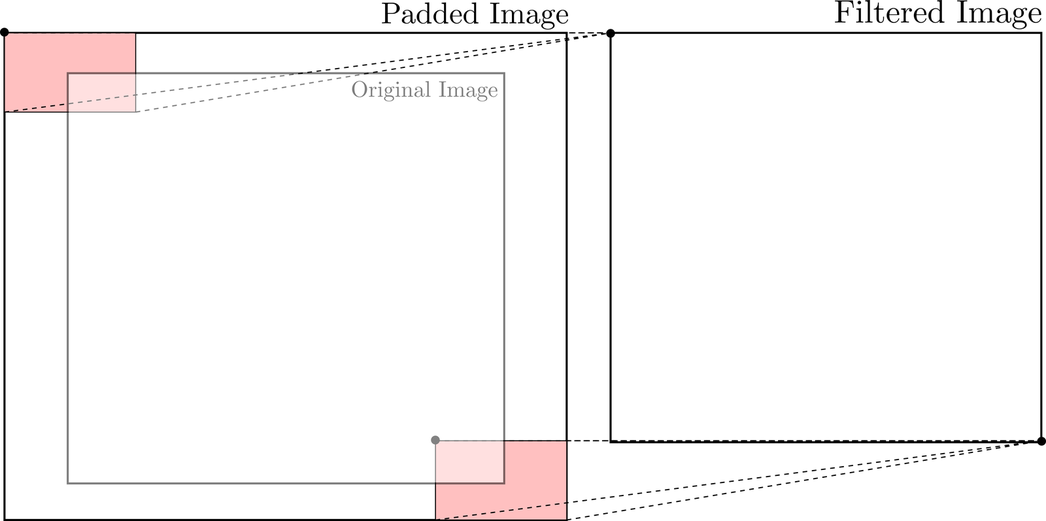}
    \caption{Image padding is a commonly used technique to ensure that the size of the filtered image is the same size as the original.}
    \label{fig:paddingfilter}
\end{figure}
\end{example}

\paragraph{Moving average filter.}
The moving average filter returns the average of the pixels in the mask, which achieves a smoothing effect\sidenote{Smoothing removes sharp features in the image.}. 
For example, we can define a moving average filter with a normalized\sidenote{The normalization is used to maintain the overall brightness of the image.} $3 \times 3$ mask with $F$ from \cref{eq:correlation} defined as:
\begin{equation*}
    F = \frac{1}{9}\begin{bmatrix}
    1 & 1 & 1 \\
    1 & 1 & 1 \\
    1 & 1 & 1 \\
    \end{bmatrix}.
\end{equation*}
Due to the symmetry of the mask, the correlation filter from \cref{eq:correlation} and convolution filter from \cref{eq:convolution} will be identical. 

\begin{figure}[ht]
  \centering
  \includegraphics[width=.8\textwidth]{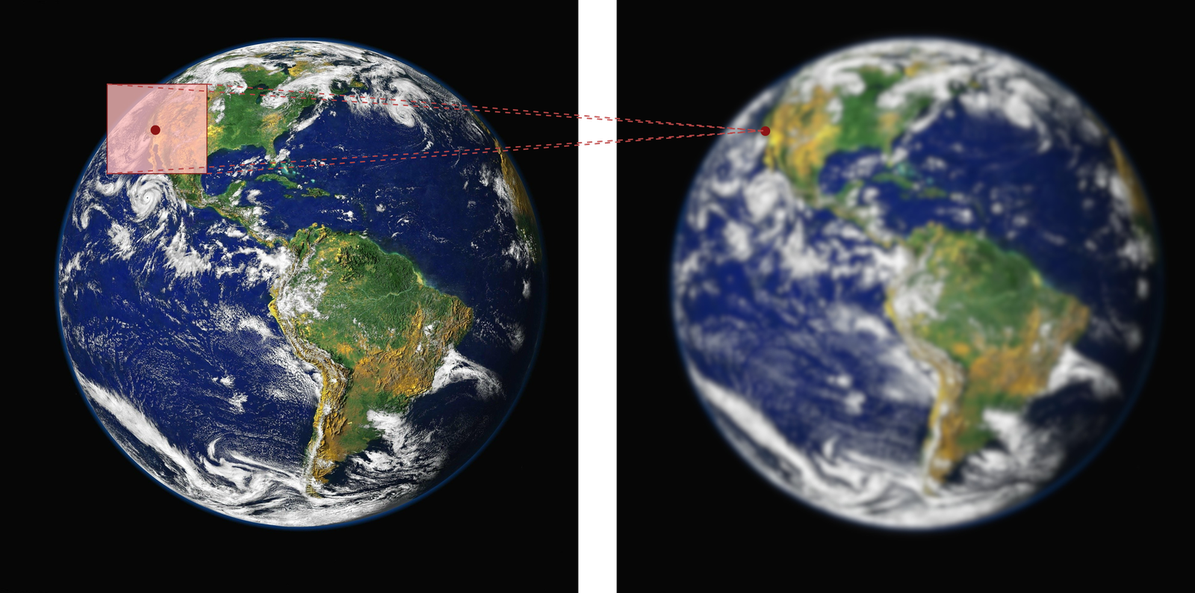}
    \caption{Example of a Gaussian smoothing filter, which produces a smoothing (blurring) effect on the filtered image.}
    \label{fig:gaussianfilter}
\end{figure}

\paragraph{Gaussian smoothing filter.}
\label{par:gaussian_smoothing_filter}
Gaussian smoothing filters are similar to the moving average filter, but instead of weighting all of the pixels evenly they are weighted by the Gaussian function:
\begin{equation*}
G_\sigma(x,y) = \frac{1}{2\pi\sigma^2} \exp \bigg(-\frac{x^2 + y^2}{2\sigma^2} \bigg).
\end{equation*}
We use this function to obtain the mask operation, $F$, by sampling the function about the center pixel.
For example, for the center pixel with $i=j=0$ in \cref{eq:correlation}, we sample $G_\sigma(0,0)$. 
For a normalized $3\times3$ mask with $\sigma= 0.85$, this filter is approximately defined by:
\begin{equation*}
F = \frac{1}{16}
\begin{bmatrix}
1 & 2 & 1\\
2 & 4 & 2\\
1 & 2 & 1
\end{bmatrix}.
\end{equation*}
Like the moving average filter, this filter mask is symmetric and therefore yields identical results with respect to the correlation or convolution filters. 
The advantage of the Gaussian filter is that it provides more weight to the neighboring pixels that are closer. 
We show an example of this filter in \cref{fig:gaussianfilter}.

\paragraph{Separable masks.}
We call a mask \emph{separable} if it can be broken down into the convolution of two kernels, $F = F_1 \ast F_2$. 
If a mask is separable into smaller masks, then it is often cheaper to apply $F_1$ followed by $F_2$, rather than by $F$ directly. 
One special case of this is when we can represent the mask as an outer product of two vectors, meaning it is equivalent to the 2D convolution of those two vectors. 
If a separable mask has shape $M\times M$ and the input image has size $w\times h$, then the computational complexity of directly performing the convolution is $O(M^2wh)$. 
By separating the masks, the computational cost is $O(2Mwh)$, which is linear in $M$ rather than quadratic. 
As an example, consider the moving average filter mask from before: 
\begin{equation*}
F = \frac{1}{9}
\begin{bmatrix}
1 & 1 & 1\\
1 & 1 & 1\\
1 & 1 & 1
\end{bmatrix} = \frac{1}{9}
\begin{bmatrix}
1 \\
1 \\
1
\end{bmatrix}
\begin{bmatrix}
1 & 1 & 1\\
\end{bmatrix}. 
\end{equation*}
As another example, we note that the Gaussian smoothing filter mask is also separable. 
To see why this is, note that we can decompose the Gaussian weighting function as:
\begin{equation*}
\begin{split}
G_\sigma(x,y) &= \frac{1}{2\pi\sigma^2} \exp \bigg(-\frac{x^2 + y^2}{2\sigma^2} \bigg)\\
    &= \frac{1}{\sqrt{2\pi}\sigma}\exp \bigg(-\frac{x^2}{2\sigma^2}\bigg)\frac{1}{\sqrt{2\pi}\sigma} \exp \bigg(-\frac{y^2}{2\sigma^2}\bigg)\\
    &= g_\sigma(x) \cdot g_\sigma(y).
\end{split}
\end{equation*}

\paragraph{Image differentiation filters.}
We can identify some image features, such as edges, by looking at the spatial derivatives in the pixel intensity values in both the vertical and horizontal directions. 
Since we represent images as functions defined over a discrete domain, the traditional method for differentiating continuous functions is not applicable. 
Instead, we can compute differences between pixels using techniques like the central difference method:
\begin{equation} 
\label{eq:cendiff}
\begin{split}
 \frac{\partial I} {\partial x} &= \frac{I(x+1,y) - I(x-1,y)}{2},\\
\frac{\partial I} {\partial y} &= \frac{I(x,y+1) - I(x,y-1)}{2}.  
\end{split}
\end{equation}
where $\partial I/\partial x$ is the derivative in the horizontal direction and $\partial I/\partial y$ is the derivative in the vertical direction. 
We can also define the derivatives using just one side instead of a central difference, for example $\frac{\partial I}{\partial x} = I(x+1,y) - I(x,y)$.

We can also differentiate an image using convolution filters. 
In particular, one common approach is to use a convolution filter of the form \cref{eq:convolution} defined with a mask, $F$, called a \emph{Sobel mask}\sidenote{Also referred to as simply a \emph{Sobel operator}.}. 
We denote this mask as $S_x$ for the $x$ direction and $S_y$ for the $y$ direction:
\begin{equation}
S_x = \begin{bmatrix}
1 & 0 & -1\\
2 & 0 & -2\\
1 & 0 & -1
\end{bmatrix}, \quad S_y =
\begin{bmatrix}
1 & 2 & 1\\
0 & 0 & 0\\
-1 & -2 & -1
\end{bmatrix}.
\end{equation}
Sobel masks are similar to the central difference method but use more neighboring pixels when calculating the derivative\sidenote{Specifically, they also consider the rows above and below to compute the difference.}. 
Note that Sobel masks are \emph{separable}.

\paragraph{Similarity measures.}
We can also use filtering to find similar features in different images, which can be useful for solving the correspondence problem in stereo vision or structure-from-motion techniques. 
In particular, we can compute the similarity between the pixel $(x,y)$ in image $I_1$ and pixel $(x', y')$ in image $I_2$ by:
\begin{equation} 
\label{eq:similarity}
\begin{split}
SAD &= \sum_{i=-N}^N \sum_{j=-M}^M \lvert I_1(x+i,y+j)-I_2(x^\prime+i,y^\prime+j)\rvert, \\
SSD &= \sum_{i=-N}^N \sum_{j=-M}^M [I_1(x+i,y+j)-I_2(x^\prime+i,y^\prime+j)]^2,
\end{split}
\end{equation}
where SAD is an acronym for \emph{sum of absolute differences}, SSD is an acronym for \emph{sum of squared differences}, and $N$ and $M$ define the size of the window around the pixels that we consider.

\subsubsection{Image Feature Detection}
A local feature\sidenote{Also sometimes referred to as interest points, interest regions, or keypoints.} in an image is a pattern that differs from its immediate neighborhood in terms of intensity, color, or texture. 
We can generally categorize local features in several ways, for example by whether or not they provide semantic content. 
For example, features that may provide semantic content include edges or other geometric shapes, such as lanes of a road or blobs corresponding to blood cells in medical images. 
Features that do not provide semantic content may also be useful, for example in feature tracking, camera calibration, 3D reconstruction, image mosaicing, and panorama stitching. 
In these cases, it may be more important that the feature be able to be located accurately and robustly over time. 
A third category of features are those that may not have semantic interpretations individually, but may have meaning as a collection.
For instance, we could recognize a scene by counting the number of feature matches between the observed scene and a query image. 
In this case, only the number of matches is relevant and not the location or type of feature. 
Applications where these types of features are important include texture analysis, scene classification, video mining, and
image retrieval.

We discuss several feature detection strategies below. 
While many strategies exist for different types of features, our focus will be on two common features that are often useful in robotics: edges and corners.

\paragraph{Edge detection.}
An \emph{edge} in an image is a region where there is a significant change in intensity values along one direction, and negligible change along the orthogonal direction. 
In one dimension an edge corresponds to a point where there is a sharp change in intensity, which mathematically can be thought of as a point of a function having a large first derivative and a small second derivative. 
Many edge detectors rely on this concept by differentiating images and looking for spikes in the derivative.
We can evaluate an edge detector based on several criteria for robustness and performance, including accuracy, localization, and single response. 
Good accuracy implies few false positives or negatives\sidenote{In this case, a false positive is a detection of an edge that isn't real, and a false negative is a missed edge.}, good localization implies that the detected edge should be exactly where the true edge is in the image, and a single response implies that only one edge is detected for each real edge. 
Noise and discretization effects can make edge detection challenging in practice.

Most edge detection methods rely on two key steps: smoothing and differentiation. 
We perform differentiation in both the vertical and horizontal directions to find locations in the image with high intensity gradients. 
However, differentiation alone is vulnerable to false positives due to image noise, which is why many algorithms will first smooth the image. 

\begin{example}[Edge detection in 1D]
\theoremstyle{definition}
In \cref{fig:noisy}, we show an example of how noise can corrupt image differentiation. 
\begin{figure}[ht]
  \centering
  \includegraphics[width=0.5\textwidth]{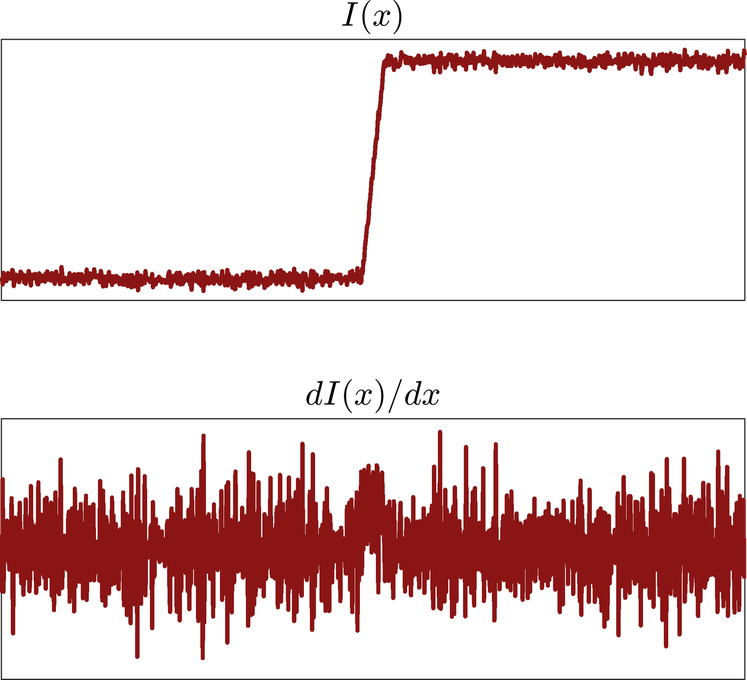}
    \caption{Differentiation of a signal with noise can be particularly challenging. 
    We can address this by first smoothing the signal.}
    \label{fig:noisy}
\end{figure}
Notice that in this case it is impossible to identify the jump in the signal due to the noise levels.
Smoothing filters, such as the Gaussian smoothing filter discussed earlier, can help remedy this problem. 
In particular, suppose the original signal in \cref{fig:noisy} is defined by $I(x)$. 
We can compute a smoothed version by applying a smoothing convolution filter:
\begin{equation*}
s(x) = g_\sigma(x) \ast I(x),
\end{equation*}
where $g_\sigma(x)$ represents a Gaussian smoothing filter, and then by applying the differentiation filter:
\begin{equation*}
s'(x)=\frac{\d}{\d x}\ast s(x).
\end{equation*}
We show this process in \cref{fig:gauss}.
\begin{figure}[ht!]
  \centering
  \includegraphics[width=0.55\textwidth]{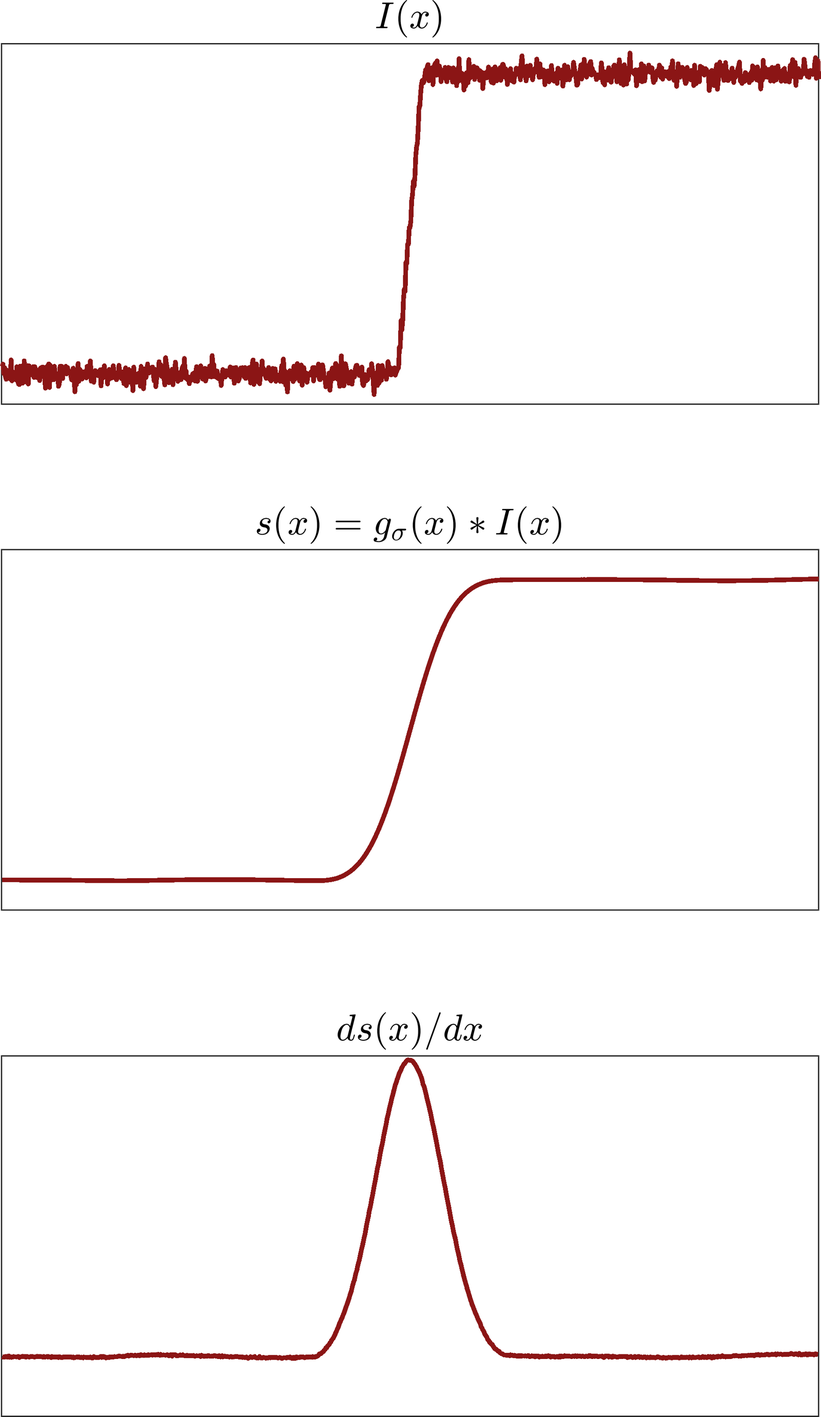}
    \caption{Edge detection through convolution with a Gaussian smoothing filter, followed by a differentiation filter.}
    \label{fig:gauss}
\end{figure}
Note that since these filters are convolutions, we can leverage the associativity property to combine them into a single filter:
\begin{equation*}
s'=(\frac{\d}{\d x} * g_\sigma) * I.
\end{equation*}
\end{example}

\begin{example}[Edge detection in 2D]
\theoremstyle{definition}
Edge detection in a two-dimensional image is quite similar to the example previously discussed for one dimension. 
Let the smoothing filter be the Gaussian smoothing filter from before, and consider a differentiation filter such as the Sobel filter. 
We can write the gradient of the smoothed image in both the $x$ and $y$ directions as:
\begin{equation*}
\nabla S= \begin{bmatrix}
\frac{\partial}{\partial x} * G_\sigma * I \\ \frac{\partial}{\partial y} * G_\sigma * I \end{bmatrix}= \begin{bmatrix}
G_{\sigma,x} * I\\G_{\sigma,y} * I
\end{bmatrix}=\begin{bmatrix}
S_x\\S_y
\end{bmatrix},
\end{equation*}
where $I$ is the original image and we use the associativity property of the smoothing and differentiation convolution filters to define the combined filters $G_{\sigma,x}$ and $G_{\sigma,y}$. 
We can then compute the magnitude of the gradient by:
\begin{equation*}
\lvert\nabla S\rvert =\sqrt{S_x^2+S_y^2},
\end{equation*}
which we can use to compare against a predefined threshold value for edge detection. 
To guarantee that we define thin edges, it is also possible to filter out points with gradient magnitude above the threshold that are not local maxima.
We show an example of this process in \cref{fig:sobel}.
\begin{figure}[ht]
  \centering
  \includegraphics[width=.9\textwidth]{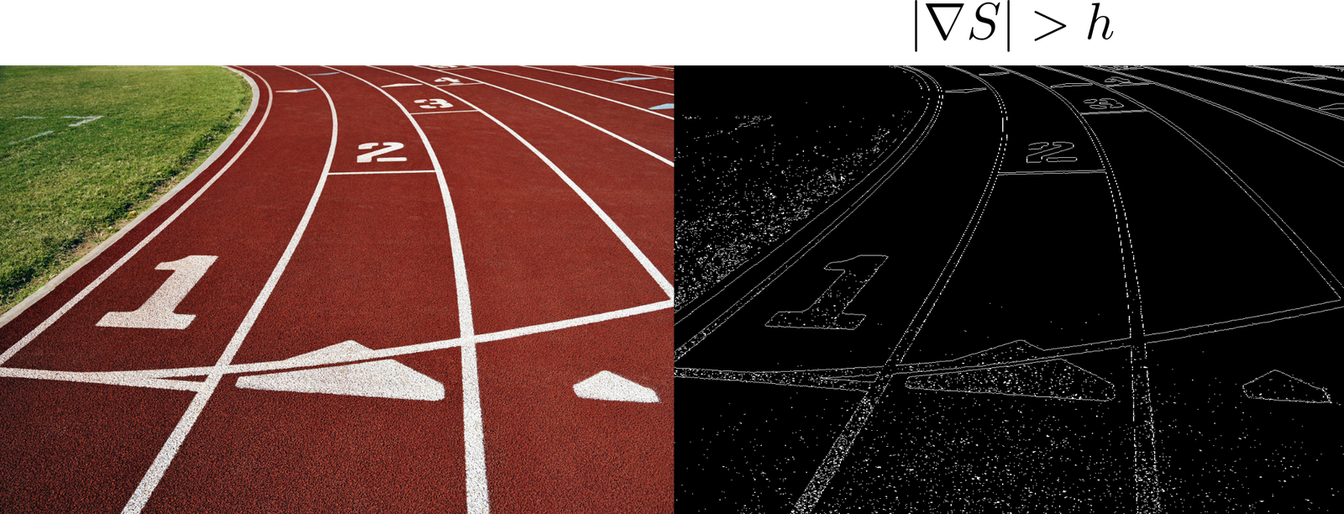}
    \caption{Edge detection using the Sobel edge detector.}
    \label{fig:sobel}
\end{figure}
\end{example}

\paragraph{Corner detection.}
A \emph{corner} in an image is defined as an intersection of two or more edges, and also sometimes as a point where there is a large intensity variation in every direction. 
Important properties of corner detectors include repeatability and distinctiveness. 
The repeatability of a corner detector quantifies how well we can find the same features in multiple images even under geometric and photometric transformations. 
Distinctiveness refers to whether the information carried by the patch surrounding the feature is distinctive, which we can use to reliably produce correspondences. 
Both of these properties are particularly important in applications such as panorama stitching and 3D reconstruction.

We can generally think of corner detection in a similar way to edge detection, except that instead of looking for change along one direction there should be changes in all directions. 
One well-known corner detector is known as the Harris detector\cite{Harris1988}, which has the useful property that the detection is invariant to rotations and linear intensity changes, such as geometric and photometric invariance. 
However, the Harris detector is not invariant to scale changes or geometric affine changes, which has led to the development of scale-invariant detectors such as the Harris-Laplacian detector or the scale-invariant feature transform (SIFT) detector.

\subsubsection{Image Descriptors}
Image \emph{descriptors} describe features so that they can be compared across images, or used for object detection and matching. 
Similar to image detectors, it is desirable for image descriptors to be repeatable\sidenote{For example, invariant with respect to pose, scale, and illumination.} and distinct. 
Perhaps the simplest example of a descriptor is an $n\times m$ window of pixel intensities centered at the feature, which we can normalize to be illumination invariant. 
However, such a descriptor is not invariant to pose or scale and is not distinctive, and therefore is generally not useful in practice.

\subsection{Geometric Feature Extraction}
\label{sec:ch09_geometricfeaturesextraction}

It is common in robotic localization and mapping to represent the environment using simple geometric primitives\sidenote{Common geometric primitives include lines, circles, corners, and planes.} that we can efficiently extract from sensor data. 
In this section, we present some techniques for line extraction from range data\sidenote{Range data can generally come from a variety of sources, including laser rangefinders, radar, or even computer vision.}. 
Lines are one of the most fundamental geometric primitives that we would want to extract from data, and techniques for extracting other primitives are conceptually similar.

There are two main challenges with extracting lines from range data. 
The first is \emph{segmentation}, which is the task of identifying which data points belong to which line, and inherently also identifying how many lines there are. 
The second is \emph{fitting}, which is the task of estimating the parameters that define a line given a set of points. 
For simplicity, in this chapter, we consider line extraction problems based on two-dimensional range data.

\subsubsection{Line Segmentation}
The line segmentation problem is to determine how many lines exist in a given set of data and which data points correspond to each line. 
We will discuss three popular algorithms for line segmentation: the \emph{split-and-merge} algorithm, the \emph{random sample consensus (RANSAC)} algorithm, and the \emph{Hough-transform} algorithm.

\begin{figure*}[t]
\centering
	\includegraphics[width=0.85\textwidth]{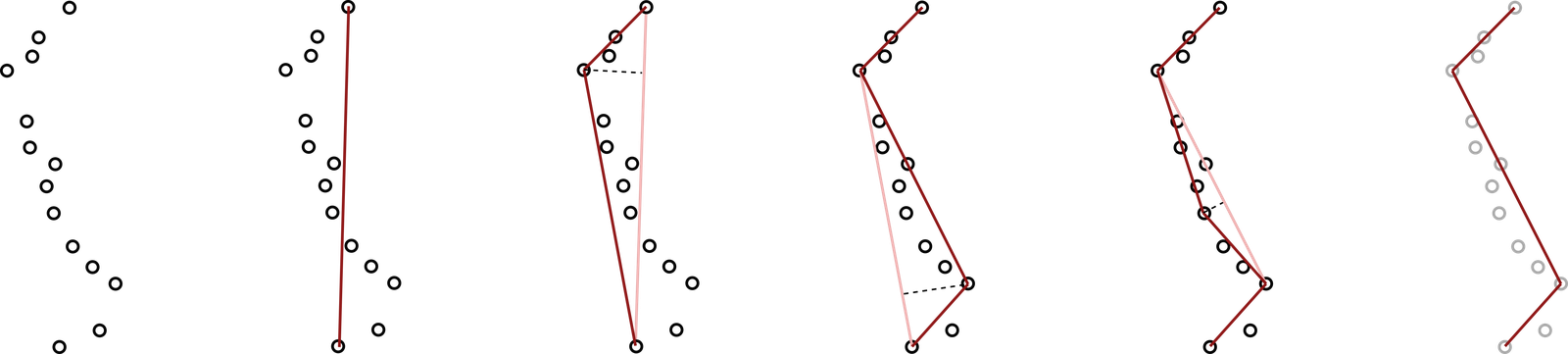}
	\caption{Iterative-end-point-fit variation of the split-and-merge algorithm for extracting lines from data.}
	\label{fig:splitmerge}
\end{figure*}

\paragraph{Split-and-merge.} 
The split-and-merge algorithm is a popular line extraction algorithm that is fast but not very robust to outliers. 
The split-and-merge algorithm repeatedly fits lines to sets of points and then splits the set of points into two sets if any point lies more than a specified distance, $d$, from the line. 
By repeating this process until no more splits occur, we are guaranteed that all points will lie less than the distance, $d$, to a line. 
After this splitting process is complete, a second step merges any of the newly formed lines that are collinear. 
We present this algorithm in more detail in \cref{alg:splitmerge}.
\begin{algorithm}[ht]
\caption{Split-and-Merge} 
\label{alg:splitmerge}
	\KwData{Set, $S$, of $N$ points, distance threshold, $d > 0$}
	\KwResult{A list, $L$, of sets of points, each resembling a line}
	$L \xleftarrow{} [S]$ \\
	$i \xleftarrow{} 1$ \\
	\While{$i \leq \text{length}(L)$}{
	    Fit a line $(\alpha,r)$ to the set $L[i]$ \\
	    Detect the point $P \in L[i]$ with maximum distance, $D$, to the line $(\alpha, r)$ \\
	    \eIf{$D < d$}{
	        $i \xleftarrow{} i + 1$
	    }
	    {
	    Split $L[i]$ at $P$ into new sets, $S_1$ and $S_2$ \\
	    $L[i] \xleftarrow{} S_1$ \\
	    $L[i+1] \xleftarrow{} S_2$ \\
	    }
	}
	Merge collinear sets in $L$
\end{algorithm}
A popular variant of the split-and-merge algorithm is known as the iterative-end-point-fit algorithm. 
This algorithm is the split-and-merge algorithm in \cref{alg:splitmerge} where the line is constructed by simply connecting the first and the last points of the set. 
We show this approach graphically in \cref{fig:splitmerge}.

\paragraph{Random sample consensus (RANSAC).}
Random Sample Consensus (RANSAC)\cite{FischlerBolles1981} is an algorithm to estimate the parameters of a model from a set of data that may contain outliers\sidenote{This problem is sometimes referred to as \emph{robust} model parameter estimation.}. 
Outliers are data points that do not fit the model and may be the result of high noise in the data, incorrect measurements, or simply points which come from objects that are unrelated to the current model. 
For example, a laser scan of an indoor environment may contain distinct lines from the surrounding walls but also points from other static and dynamic objects such as chairs or humans. 
In this case, if the goal is to extract lines to represent the walls, then any data point corresponding to other objects would be an outlier. 
In general, we can apply RANSAC to many parameter estimation problems, and typical applications in robotics include line extraction from 2D range data, plane extraction from 3D point clouds, and structure-from-motion\sidenote{Where the goal in structure-from-motion problems is to identify image correspondences which satisfy a rigid body transformation.}. 
We focus on using RANSAC for line extraction from two-dimensional data below.

\begin{figure}[ht]
  \centering
  \includegraphics[width=0.85\textwidth]{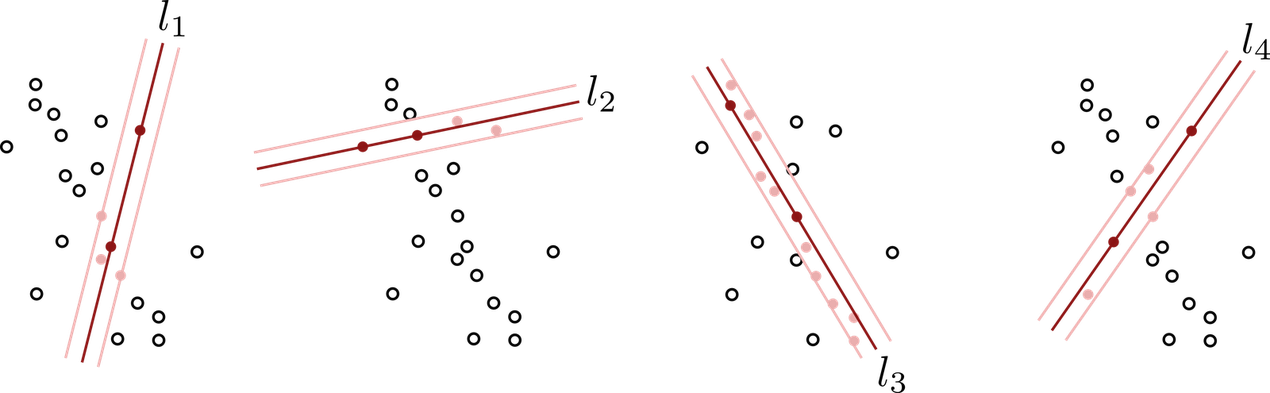}
\caption{Example of the RANSAC algorithm, showing four iterations of the algorithm. 
If the algorithm was terminated after these four iterations, line $l_3$ would be returned since it contains the maximum number of points.}
\label{fig:ransac-working}
\end{figure}

RANSAC is an iterative method and is non-deterministic\sidenote{In other words, it is stochastic or random. 
Running the algorithm twice on the same inputs will not necessarily produce the same results.}. 
Given a dataset, $S$, of $N$ points, we start by randomly selecting a sample of two points from $S$. 
Next, we construct a line from the two sampled points and compute the distance of all other points to this line. 
We then define the set of \emph{inliers}, which is comprised of all points whose distance to the line is within a predefined threshold, $d$. 
By repeating this process $k$ times, we generate $k$ inlier sets and their associated lines and return the inlier set with the most points. 
We detail this procedure in \cref{alg:ransac} and illustrate the process in \cref{fig:ransac-working}.
\begin{algorithm}[ht]\caption{Random Sample Consensus (RANSAC) for Line Extraction} \label{alg:ransac}
	\KwData{Set, $S$, of $N$ points, distance threshold, $d$}
	\KwResult{Set with maximum number of inliers and corresponding line}
	\While{$i \leq k$}{
	    Randomly select two points from $S$. \\
	    Fit line, $l_i$, through the two points. \\
	    Compute distance of all other points to $l_i$. \\
	    Construct set of points, $\tilde{S}_i$, with distance less than $d$ to $l_i$. \\
	    Store line, $l_i$, and set of points, $\tilde{S}_i$.\\
	    $i \xleftarrow{} i + 1$
	}
	Choose set $\tilde{S}_i$ with maximum number of points.
\end{algorithm}

Due to the probabilistic nature of the algorithm, as the number of iterations, $k$, increases the probability of finding a good solution increases. 
This approach is used over a brute force search of all possible combinations of two points since the total number of combinations is $N(N-1)/2$, which can be extremely large. 
In fact, we can perform a simple statistical analysis of RANSAC.
Let $p$ be the \emph{desired} probability of finding a set of points free of outliers and let $w$ be the probability of selecting an inlier from the dataset, $S$, of $N$ points, which we can express as:
\begin{equation*}
    w \coloneqq \frac{\text{\# inliers}}{N}.
\end{equation*}
Assuming we draw point samples independently from $S$, the probability of drawing two inliers is $w^2$, and $1-w^2$ is the probability that at least one is an outlier. 
Therefore, with $k$ iterations, the probability that RANSAC never selects two points that are both inliers is $(1-w^2)^k$. 
We can therefore find the minimum number of iterations, $\bar{k}$, needed to find an outlier-free set with probability $p$ by solving:
\begin{equation*}
    1-p = (1-w^2)^k,
\end{equation*}
for $k$. 
In other words, we can compute $\bar{k}$ as:
\begin{equation*}
    \bar{k} = \frac{\log (1-p)}{\log (1-w^2)}.
    \label{eq:magic-k}
\end{equation*}
While the value of $w$ may not be known exactly\sidenote{There are advanced versions of RANSAC that can estimate $w$ in an adaptive online fashion.}, we can still use this expression to get a good estimate of the number of iterations, $k$, that we need for good results. 
It is important to note that this probabilistic approach often leads to a much smaller number of iterations than a brute force search through all combinations. 
We can attribute this to the fact that $\bar{k}$ is only a function of $w$ and not the total number of samples, $N$, in the dataset. 

Overall, the main advantage of RANSAC is that it is a generic extraction method and can be used with many types of features given a feature model. 
It is also simple to implement and is robust to data outliers. 
The main disadvantages are that the algorithm needs to run multiple times to extract multiple features, and there are no guarantees that the solutions will be optimal.

\begin{figure}[ht]
  \centering
  \includegraphics[width=0.75\textwidth]{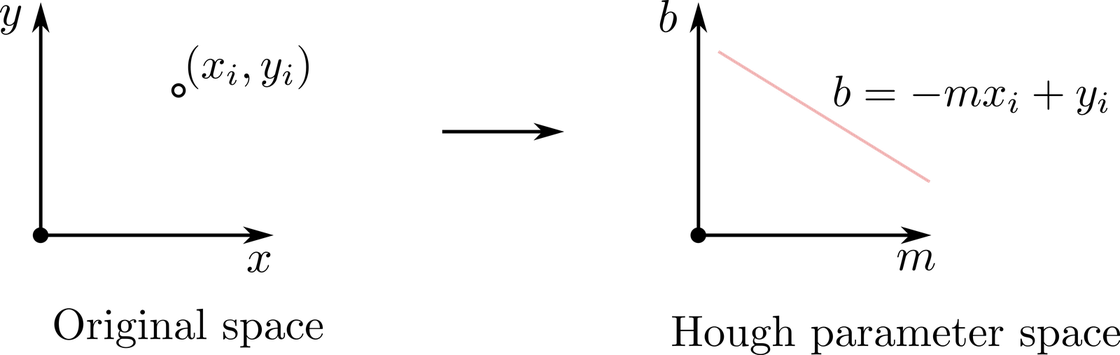}
\caption{Each point, $(x_i, y_i)$, in the original space maps to a \emph{line} in the Hough space which describes all possible parameters $m$ and $b$ that would generate a line passing through the point $(x_i, y_i)$.}
\label{fig:hough1}
\end{figure}

\begin{figure}[ht]
  \centering
  \includegraphics[width=0.75\textwidth]{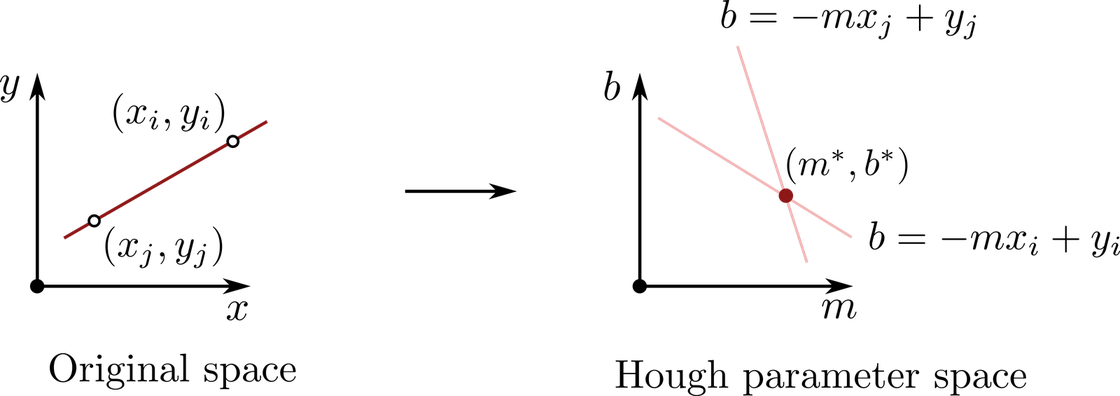}
\caption{All points on a line in the original space yield lines in the Hough space that intersect at a common point.}
\label{fig:hough2}
\end{figure}

\paragraph{Hough transform.}
In the Hough transform algorithm, each point, $(x_i,y_i)$, of the dataset, $S$, votes for a \emph{set} of possible line parameters, $(m,b)$, where $m$ is the slope and $b$ is the intercept point. 
For any given point, $(x_i,y_i)$, the candidate set of line parameters, $(m,b)$, that could pass through this point must satisfy $y_i = mx_i + b$, which we can also write as:
\begin{equation*}
\quad b=-mx_i + y_i.
\end{equation*}
Therefore, each point, $(x,y)$, in the original space maps to a \emph{line}, $(m,b)$, in the Hough space, as we show in \cref{fig:hough1}. 
The Hough transform algorithm exploits this fact by noting that two points on the same line in the original space will yield two \emph{intersecting} lines in Hough space. 
In particular, the point where they intersect in the Hough space corresponds to the parameters $m^*$ and $b^*$ that defines the line passing between the points in the original space, as we show in \cref{fig:hough2}.

We can apply this concept to the line segmentation problem by searching in the Hough space for intersections among the lines that correspond to each point, $(x,y)$, in the set, $S$. 
In practice, we do this by discretizing the Hough space with a grid and simply counting for each grid cell the number of lines corresponding to $(x_i,y_i)$ points from $S$ that pass through it. 
We choose local maxima among the cells as lines that ``fit'' the data set, $S$.

\begin{figure}
\centering
	\includegraphics[width=0.75\textwidth]{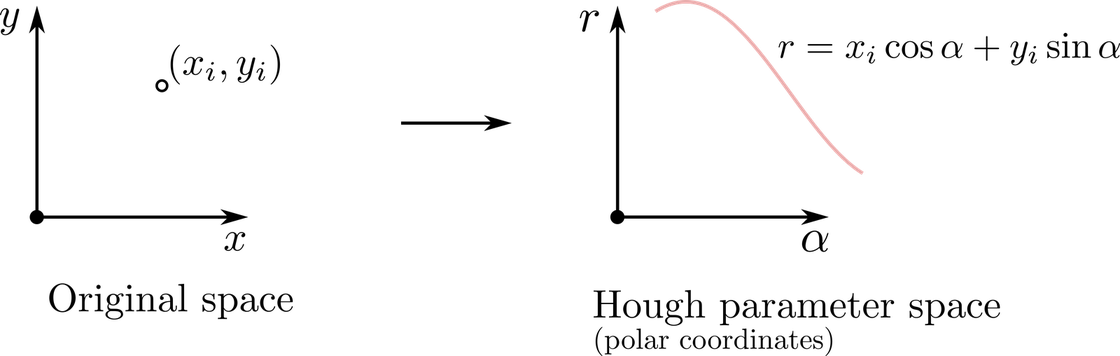}
	\caption{Representation of a point, $(x_i, y_i)$, in the Hough space when using a polar coordinate representation of a line with parameters $\alpha$ and $r$.}
	\label{fig:hough3}
\end{figure}

\begin{figure}
\centering
	\includegraphics[width=0.85\textwidth]{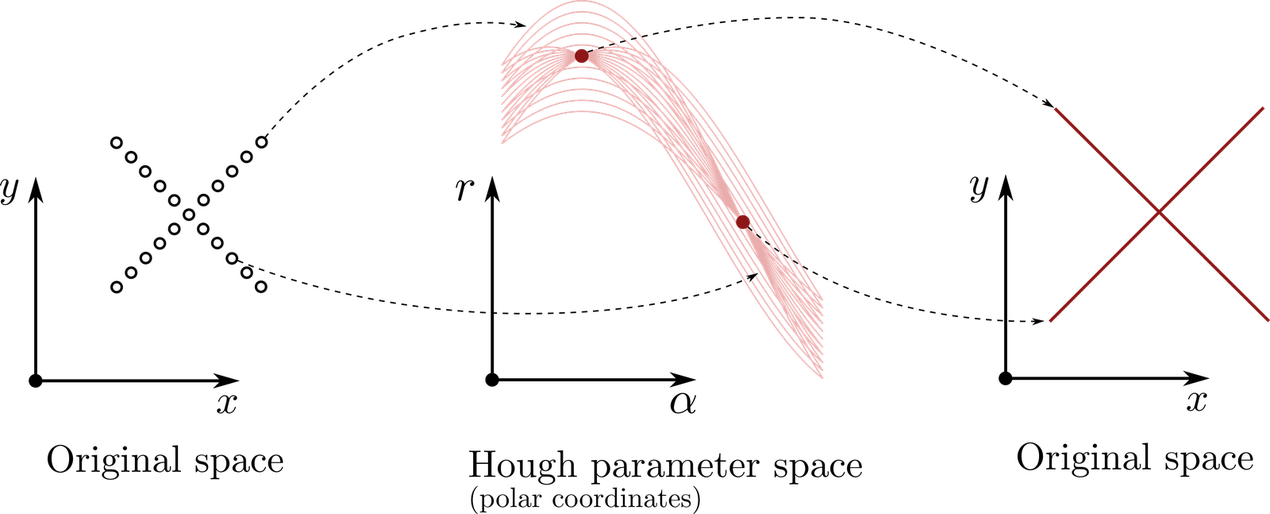}
	\caption{Example of the Hough transformation using a polar coordinate representation of lines.}
	\label{fig:hough4}
\end{figure}

However, performing a discretization of the Hough space requires a trade-off between range and resolution, in particular because the slope, $m$, can range from $-\infty$ to $\infty$. 
Alternatively, we can use a polar coordinate representation of the Hough space which defines a line as:
\begin{equation*}
x \cos \alpha + y \sin \alpha = r,
\end{equation*}
where $(\alpha,r)$ are the new line parameters. 
With this representation, we map a point, $(x_i,y_i)$, from the original space to the polar Hough space, $(\alpha,r)$, as a sinusoidal curve, as we show in \cref{fig:hough3}. 
We provide an example of the Hough transform using the polar representation in \cref{fig:hough4}.

\subsubsection{Point Cloud Registration}
\label{subsec:pointcloud-reg}

In robotics, another important sensor modality consists of point clouds, which we can obtain from lidar or RGB-D sensors. One important consideration is to align two point clouds, generally to localize a sensor in its surroundings or to merge data from multiple viewpoints into a unified representation. We refer to this alignment problem formally as \emph{point cloud registration}, which is the task of finding the geometric transformation that best aligns one point cloud to another.

The point cloud registration problem can be formulated as follows. Given a source point cloud $P \coloneqq \{p_1, \ldots, p_N\}$ and a reference point cloud $Q \coloneqq \{q_1, \ldots, q_M\}$, where each $p_i, q_j \in \R^3$, our goal is to find the rigid transformation consisting of a rotation matrix $R \in SO(3)$ and translation vector $t \in \R^3$ that best aligns $P$ to $Q$. We can express this mathematically as minimizing the error metric:
\begin{equation}
E(R, t) = \sum_{i=1}^{N} \| Rp_i + t - q_i^* \|^2,
\end{equation}
where $q_i^*$ denotes the closest point in $Q$ to the transformed point $Rp_i + t$. This formulation leads to a challenging optimization problem because both the transformation parameters and the point correspondences are unknown.

\begin{figure}[ht]
  \centering
  \includegraphics[width=0.75\textwidth]{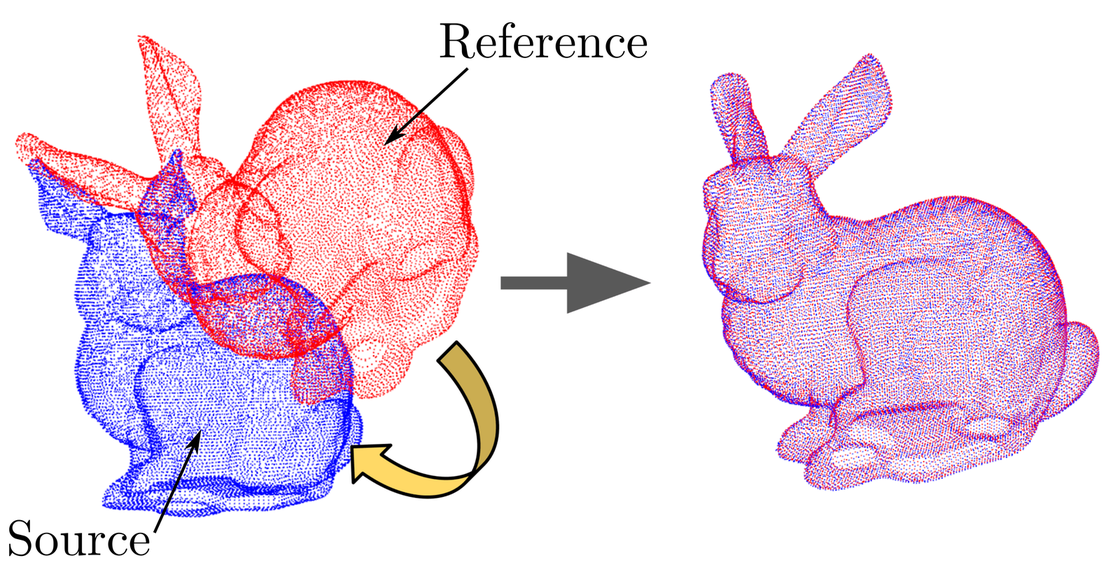}
    \caption{Illustration of the ICP algorithm iteratively aligning a source point cloud to a reference point cloud. The algorithm alternates between finding correspondences and estimating the transformation until convergence.}
    \label{fig:icp}
\end{figure}

\paragraph{Iterative closest point.}
The Iterative Closest Point (ICP) algorithm\cite{zhang1994iterative} is a widely used method for solving the point cloud registration problem. The algorithm alternates between two steps: finding point correspondences and estimating the optimal transformation given those correspondences. 
We present the complete ICP algorithm in \cref{alg:icp} and illustrate the iterative alignment process in \cref{fig:icp}.
\begin{algorithm}[ht]
\caption{Iterative Closest Point (ICP)} 
\label{alg:icp}
	\KwData{Source point cloud $P$, reference point cloud $Q$, initial transformation $(R_0, t_0)$ (optional), convergence threshold $\epsilon$}
	\KwResult{Refined transformation $(R, t)$}
	Initialize $(R, t) \xleftarrow{} (R_0, t_0)$ or $(I, 0)$ if no initial guess provided \\
	$E_{\text{prev}} \xleftarrow{} \infty$ \\
	\Repeat{$|E - E_{\text{prev}}| < \epsilon$}{
	    \tcp{Step 1: Find correspondences}
	    \For{each point $p_i \in P$}{
	        Find closest point $q_i^* \in Q$ to $Rp_i + t$ \\
	    }
	    \tcp{Step 2: Estimate transformation}
	    Compute optimal $(R_{\text{new}}, t_{\text{new}})$ that minimizes $\sum_{i=1}^{N} \| R_{\text{new}}p_i + t_{\text{new}} - q_i^* \|^2$ \\
	    \tcp{Step 3: Apply transformation}
	    $(R, t) \xleftarrow{} (R_{\text{new}}, t_{\text{new}})$ \\
	    $E_{\text{prev}} \xleftarrow{} E$ \\
	    $E \xleftarrow{} \sum_{i=1}^{N} \| Rp_i + t - q_i^* \|^2$ \\
	}
\end{algorithm}

The correspondence step in ICP requires finding the nearest neighbor in $Q$ for each transformed point in $P$. Note that while a naive implementation would require $O(NM)$ distance computations per iteration, in practice, we can accelerate this using spatial data structures such as KD-trees, which reduce the average complexity to $O(N \log M)$. However, it is important to note that the nearest neighbor matching can produce incorrect correspondences, particularly when the point clouds are far from alignment or when they have limited overlap.
The transformation estimation step computes the optimal rotation and translation given the current correspondences. This has a closed-form solution that we can obtain using singular value decomposition (SVD). First, we compute the centroids of the corresponding point sets:
\begin{equation}
\bar{p} = \frac{1}{N}\sum_{i=1}^{N} p_i, \quad \bar{q} = \frac{1}{N}\sum_{i=1}^{N} q_i^*.
\end{equation}
Next, we construct the cross-covariance matrix:
\begin{equation}
H = \sum_{i=1}^{N} (p_i - \bar{p})(q_i^* - \bar{q})^T.
\end{equation}
Computing the SVD of $H = U\Sigma V^T$, the optimal rotation is $R = VU^T$, and the optimal translation is $t = \bar{q} - R\bar{p}$\sidenote{We must check that $\det(R) = 1$ to ensure a proper rotation. If $\det(R) = -1$, we negate the column of $V$ corresponding to the smallest singular value.}.

The ICP algorithm converges when the change in error between iterations falls below a threshold $\epsilon$, or when the change in transformation parameters is sufficiently small. The algorithm is guaranteed to monotonically decrease the alignment error and converge to a local minimum, though the quality of the final alignment is highly sensitive to the initial transformation estimate. In practice, several strategies can improve ICP's robustness and performance, including outlier rejection methods that discard point pairs with distances exceeding a threshold to prevent corrupted correspondences from degrading the solution. As ICP remains a local optimization method, it benefits significantly from good initialization, which we can obtain from odometry, inertial sensors, or coarse global registration methods.

\subsection{Feature-Based Object Detection}
\label{sec:ch09_featurebaseddetection}

Another high-level information extraction task that is common in robotics is \emph{object recognition}. 
Object recognition is the task of classifying or naming discrete objects in the world, usually based on images or video. 
This is a particularly challenging task because real-world scenes are commonly made up of many varying types of objects which can appear at different poses and can occlude each other. 
Additionally, objects within a specific class can have a large amount of variability, for example breeds of dogs or car models. 
In this section, we introduce common methods for feature-based object detection, namely \emph{template matching} and \emph{bag of visual words}.

\subsubsection{Template Matching}
Template matching\cite{PerveenKumarEtAl2013} is a machine vision technique for identifying parts of an image that match a given image pattern\sidenote[][4\baselineskip]{Advanced template matching algorithms enable finding pattern occurrences regardless of their orientation and local brightness.}. 
This approach has seen success in a variety of applications, including manufacturing quality control, mobile robotics, and more. 
The two primary components needed for template matching are the source image, $I$, and a template image, $T$.

Given a source and template image, one approach to template matching is to leverage the linear spatial correlation filters discussed earlier in this chapter. 
In particular, a naive approach would be to use the normalized template image as a filter mask in a correlation filter. 
By applying this filter mask to every pixel in the source image, the resulting output would quantify the similarity of that region of the source image to the template. 
This type of approach is sometimes referred to as a \emph{cross-correlation}.
Another approach based on linear spatial filters would be to leverage the similarity filters that compute the sum of absolute differences (SAD) metric for each pixel in the source image. 
Regions of the source image similar to the template would correspond to low SAD scores.
The disadvantages of these approaches are that they do not handle rotations or scale changes, which are quite common in real-world applications.

One solution to the scaling issue in correlation filter based template matching is to simply re-scale the source image multiple times and perform template matching on each. 
We can use this concept, referred to as using \emph{image pyramids}\cite{Szeliski2010}, to accelerate object search by first using a coarser resolution image to localize the object, and then using finer resolution images for actual detection. 
We can build image pyramids in several ways. 
One naive approach is to simply eliminate some rows and columns of the image. 
Another approach is to first use a Gaussian smoothing filter to remove high frequency content from the image and then subsample the image. 
We refer to the sequence of images resulting from this approach as a \emph{Gaussian pyramid}.

\subsubsection{Bag of Visual Words}
The key idea behind the bag of visual words\sidenote{The model originated in natural language processing, where we consider texts such as documents, paragraphs, and sentences as collections, or ``bags'', of words.} approach is that we can simplify object representations by considering them as a collection of their subparts\sidenote{For example, a bike is an object with wheels, a frame, and handlebars.}, and we refer to the subparts as \emph{visual words}. 
In this approach, we search a source image for \emph{visual words}, and we create a distribution of visual words that we find in the image in the form of a histogram. 
We can then perform object detection by comparing this distribution to a set of training images. 
For example, suppose the source image contains a human face and the recognized features included eyes and a nose. 
Then, by comparing the distribution to training images, we would likely determine that the training images that also have eyes and a nose are also images of faces.

\subsubsection{Classical Perception Approaches Today}

While modern deep learning methods have revolutionized many computer vision tasks, classical perception techniques remain essential in robotics, particularly for specific instance detection and registration tasks. The fundamental methods we have discussed—feature detection, description, and matching—continue to form the backbone of many practical robotic systems. We will explore modern deep learning approaches to perception in detail in a subsequent chapter, but it is important to understand where classical methods continue to excel.

One area where classical methods remain dominant is in \emph{instance detection}, where the goal is to identify and localize a specific object rather than classify object categories. For example, detecting a particular coffee cup on a cluttered desk, recognizing a specific tool in a manufacturing environment, or localizing a known landmark for robot navigation all benefit from classical approaches. Unlike category-level classification (e.g., ``this is a cup''), instance detection requires identifying \emph{which specific} cup among many possible cups.

The typical pipeline for instance detection leverages the classical methods we have covered:
\begin{enumerate}
\item \emph{Feature Detection and Description}: Extract distinctive keypoints from both the query image (the specific object to find) and the target image (the scene to search). Scale-invariant detectors like SIFT or its variants remain popular because they are robust to changes in viewpoint, scale, and illumination.

\item \emph{Feature Matching}: Match features between the query and target images using descriptor similarity. This step identifies potential correspondences between the known object and regions in the scene.

\item \emph{Geometric Verification}: Use robust estimation techniques like RANSAC to fit a geometric transformation (such as a homography or rigid body transformation) to the matched features. This step filters out incorrect matches and verifies that the pattern of features is geometrically consistent with the object model.

\item \emph{Pose Estimation}: Once verified correspondences are obtained, estimate the 3D pose of the object relative to the camera, enabling the robot to interact with or manipulate the object.
\end{enumerate}

This classical pipeline offers several advantages that keep it relevant today. First, it requires only a small number of example images or even a 3D model of the specific object, whereas deep learning approaches typically require large labeled datasets. Second, it provides explicit geometric reasoning, yielding not just detection but precise pose estimation needed for robotic manipulation. Third, it is interpretable—engineers can inspect features, matches, and geometric fits to understand and debug system behavior.

Classical methods also remain important in \emph{registration} problems, where the goal is to align sensor data from different viewpoints or modalities. Applications include point cloud alignment for 3D reconstruction, image stitching for panoramas, and multi-sensor fusion, which we will discuss in later chapters. The Iterative Closest Point (ICP) algorithm and feature-based registration using RANSAC continue to be workhorses in these domains. However, modern systems increasingly adopt hybrid approaches that combine classical and learning-based methods. For instance, learned feature detectors and descriptors (such as SuperPoint\cite{detone2018superpoint}) can replace hand-crafted features like SIFT while maintaining the geometric reasoning framework. Similarly, learned feature matching networks can improve correspondence quality before geometric verification. These hybrid pipelines leverage the strengths of both paradigms: the data efficiency and geometric rigor of classical methods with the representational power of learned features.

In summary, while deep learning has transformed object classification and semantic understanding, classical perception methods remain indispensable for tasks requiring precise instance detection, geometric reasoning, and data-efficient operation. Understanding these fundamental techniques is essential for roboticists working on manipulation, localization, and any application where knowing exactly which object is where matters more than simply recognizing what category it belongs to.

\section{Summary}
\label{sec:classic-pcp-summary}
In this chapter, we introduced methods for extracting various types of information from images and sensor data through a pipeline of low-level image processing and higher-level feature extraction techniques.
We began with image processing fundamentals, covering filtering operations such as Gaussian smoothing for noise reduction and Sobel operators for differentiation and edge detection. We discussed the mathematical principles of correlation and convolution, along with practical implementation tricks like zero-padding. This foundation extended to feature detection, where we explored strategies for identifying semantically useful patterns like edges and corners, and the role of descriptors for representing these features.

Building on these low-level techniques, the chapter then presented methods for geometric feature extraction, focusing on the challenge of identifying structure like lines in range data. We detailed and compared three core segmentation algorithms: the fast but less robust Split-and-Merge, the robust but stochastic RANSAC, and the model-based Hough Transform. We also introduced the Iterative Closest Point (ICP) algorithm as a fundamental tool for point cloud registration.
Finally, we transitioned to feature-based object detection, covering classical approaches such as template matching and the Bag of Visual Words model. We concluded by discussing the enduring relevance of these classical perception methods in modern robotics, particularly for precise instance detection and geometric verification tasks, and noted their role in hybrid systems that combine classical pipelines with learned components.

\paragraph{To learn more.}
For a deeper exploration of the topics covered in this chapter, several key resources are available.
A comprehensive introduction to the computer vision algorithms that underpin robotic perception can be found in \citet{Szeliski2010} and \cite{Moravec1977}.
The original papers on key algorithms provide invaluable insight: \citet{FischlerBolles1981} for RANSAC, \citet{Harris1988} for the corner detector, and \citet{zhang1994iterative} for ICP.
For a deeper dive into feature descriptors, explore SIFT \citet{Lowe2004} and its modern learned counterparts like SuperPoint \citep{detone2018superpoint}.
Finally, for a broader perspective on how these classical techniques are employed in state-of-the-art systems, consult robotics and computer vision texts such as \citet{SiegwartNourbakhshEtAl2011}.

\section{Exercises}
The starter code for the exercises provided below is available online through GitHub. 
To get started, download the code by running in a terminal window:

\begin{tcolorbox}[colback=gray!10]
\begin{minted}{bash}
    git clone https://github.com/StanfordASL/pora-exercises.git
\end{minted}
\end{tcolorbox}

We denote Problems requiring hand-written solutions and coding in Python with \adjustbox{height=2ex, valign=c}{\includegraphics{figs/write.png}} and \adjustbox{height=2ex, valign=c}{\includegraphics{figs/code.png}}, respectively.

\subsection*{\adjustbox{height=2ex, valign=c}{\includegraphics{figs/write.png}}\adjustbox{height=2ex, valign=c}{\includegraphics{figs/code.png}}\ Problem 1: Correlation and Gaussian Smoothing Filters}
In this exercise, you will explore developing a correlation and Gaussian smoothing filter using the top-left indexing approach defined by \cref{eq:correlation_newindex}.
Specifically, complete the following:
\begin{enumerate}
\item First, consider an image and its zero-padded version:
\begin{equation*}
I = \begin{bmatrix}
    7 & 4 & 1 \\
    8 & 5 & 2 \\
    9 & 6 & 3 \\
    \end{bmatrix}, \quad
I_\text{padded} = \begin{bmatrix}
    0 & 0 & 0 & 0 & 0 \\
    0 & 7 & 4 & 1 & 0 \\
    0 & 8 & 5 & 2 & 0 \\
    0 & 9 & 6 & 3 & 0 \\
    0 & 0 & 0 & 0 & 0 \\
    \end{bmatrix}.
\end{equation*}
Compute by hand the resulting image $I'$ from applying the following correlation filters:
\begin{enumerate}
\item \begin{equation*}
F = \begin{bmatrix}
    0 & 0 & 0 \\
    0 & 1 & 0 \\
    0 & 0 & 0 \\
    \end{bmatrix}.
\end{equation*}
\item \begin{equation*}
F = \begin{bmatrix}
    1 & 0 & 0 \\
    0 & 0 & 0 \\
    0 & 0 & 0 \\
    \end{bmatrix}.
\end{equation*}
\item \begin{equation*}
F = \begin{bmatrix}
    1 & 1 & 1 \\
    0 & 0 & 0 \\
    -1 & -1 & -1 \\
    \end{bmatrix}.
\end{equation*}
What is this filter doing to the image?
Why might this be useful in computer vision?
How would this be different than the functionality of the filter:
\begin{equation*}
F' = \begin{bmatrix}
    -1 & 0 & 1 \\
    -1 & 0 & 1 \\
    -1 & 0 & 1 \\
    \end{bmatrix}.
\end{equation*}
\item \begin{equation*}
F = \frac{1}{16}\begin{bmatrix}
    1 & 2 & 1 \\
    2 & 4 & 2 \\
    1 & 2 & 1 \\
    \end{bmatrix}.
\end{equation*}
What is this filter doing to the image?
Why might this be useful in computer vision?
How would this be different than the functionality of the filter:
\begin{equation*}
F' = \frac{1}{9}\begin{bmatrix}
    1 & 1 & 1 \\
    1 & 1 & 1 \\
    1 & 1 & 1 \\
    \end{bmatrix}.
\end{equation*}
\end{enumerate}
\item In the file \colorcode{ch08/exercises/correlation\_filter.ipynb}, you will now implement the correlation filter.
Specifically, implement the function \colorcode{correlate\_image} using \cref{eq:correlation_newindex}.
Run the provided code to see the result of your implementation for a horizontal and vertical edge detector filter applied to a test image.	
\item Also in the file \colorcode{ch08/exercises/correlation\_filter.ipynb}, implement the Gaussian smoothing filter from \cref{par:gaussian_smoothing_filter} in the function \colorcode{create\_gaussian\_filter}.
Run the provided code to see the result of your implementation with $\sigma = 0.5$ and $\sigma = 2$.
Explain the general impact of varying $\sigma$ on the resulting image.	
\end{enumerate}

\subsection*{\adjustbox{height=2ex, valign=c}{\includegraphics{figs/code.png}}\ Problem 2: Iterative Closest Point (ICP) for Point Cloud Registration}
In this exercise, you will use the ICP algorithm defined in \cref{alg:icp} to register two different point clouds in the same reference frame.
In the notebook \colorcode{ch08/exercises/icp.ipynb}, complete the following:
\begin{enumerate}
\item Run the provided code to load two point clouds: one that represents the target ``full'' point cloud, and another that is just a partial point cloud of the same object.
The goal will be to determine the transformation that will align the partial point cloud to the target point cloud.
\item The first step before running ICP is to get an initial transformation estimate.
To accomplish this we will use the RANSAC algorithm.
Explore and run the provided code to run RANSAC to get this initial estimate.
\item Now, you will implement a basic version of the ICP algorithm.
Implement the functions \colorcode{nearest\_neighbor} and \colorcode{icp}.
Run the provided code, there should now be a rough alignment between the partial and target point clouds.
Note that this basic version of ICP is not necessarily robust to noise and outliers.
\item Use the \colorcode{open3D} library's functions to run a more robust version of the ICP algorithm.
Run the provided code to see how well the point clouds align with this implementation.
\end{enumerate}

\newpage
\printbibliography[segment=\therefsegment,heading=subbibliography,title={References}]
\chapter{Deep Learning Architectures for Perception}
\label{ch:vision-networks}
\newrefsegment
Modern computer vision has shifted from hand-crafted features to end-to-end learning with deep neural networks. Rather than manually designing feature extractors like SIFT or HOG descriptors, contemporary approaches learn hierarchical representations directly from data, achieving strong performance across diverse visual tasks. This transformation has been particularly impactful for robotics, where robust visual perception is essential for autonomous operation in complex, unstructured environments.
Consider a mobile robot navigating indoors: it must process RGB images to recognize objects while interpreting 3D LiDAR point clouds to understand spatial layout and plan collision-free paths. An autonomous vehicle must fuse information from multiple cameras and LiDAR scanners, processing both 2D image grids and irregular 3D point clouds in real-time. These scenarios require neural network architectures that effectively process different data modalities. \textit{Convolutional Neural Networks (CNNs)} leverage spatial locality for processing camera images. \textit{Transformers} use self-attention to capture long-range dependencies and enable transfer learning. For 3D data, \textit{point-based networks} like PointNet process unordered point clouds directly, while \textit{voxel-based architectures} discretize 3D space into regular grids. Each architecture provides learned feature representations essential for downstream robotic tasks.

In this chapter, we explore fundamental neural network architectures for robotic perception. We begin with CNNs in Section~\ref{sec:ch11_cnn} and Transformers in Section~\ref{sec:ch11_transformers} for image processing, then discuss point-based approaches in Section~\ref{sec:ch11_pointbased} and voxel-based approaches in Section~\ref{sec:ch11_voxelbased} for 3D sensor data.
These architectural foundations serve as building blocks for the detection, segmentation, and scene understanding methods in subsequent chapters.

\section{Convolutional Neural Networks}
\label{sec:ch11_cnn}

Convolutional neural networks (CNNs) are a type of deep learning architecture that is very common in the fields of computer vision and image processing.
The architecture of a CNN contains a special structure that leverages convolution filters similar to those we discussed in \cref{ch:classical_perception}.
However, in the context of machine learning, the convolution filters embedded in the structure of a CNN are optimized for the desired task and do not require human specification, giving higher performance and reducing the amount of required manual engineering.
CNN architectures comprise several main components: convolutional layers, nonlinear activations, pooling layers, and fully-connected layers.
In the following sections, we discuss these components in more detail.

\subsubsection{Convolution Layers}
\begin{figure}[ht]
\centering
\includegraphics[width=0.4\textwidth]{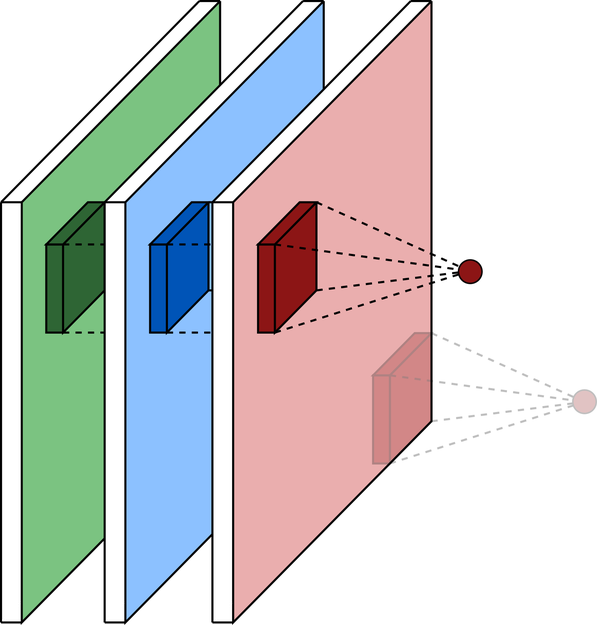}
\caption{A convolution filter being applied to a 3-channel RGB image.}
\label{fig:convfilter}
\end{figure}
One of the main structural concepts that is unique to the architecture of a CNN is the use of convolution layers. 
Convolution layers exploit the underlying \emph{spatial locality} structure in images by using sliding, learned filters which are often much smaller than the image itself. 
Mathematically, these filters perform operations in a manner similar to other linear filters used in image processing, such as Gaussian smoothing filters. For a 2D convolution operation, we can express the computation at each output location as:
\begin{equation}
    Y_{i,j} = \sum_{u=0}^{m-1} \sum_{v=0}^{n-1} X_{i+u,j+v} \cdot W_{u,v} + b,
    \label{fig:convolution_equation}
\end{equation}
where $(i,j)$ represents the spatial position in the output feature map, $X$ is the input image, $W$ is the filter with dimensions $m \times n$, and $b$ is the bias term. This operation slides the filter across the input, computing a weighted sum at each position.
For multi-channel inputs like RGB images with $C$ channels, the convolution extends to:
\begin{equation*}
    Y_{i,j} = \sum_{c=0}^{C-1} \sum_{u=0}^{m-1} \sum_{v=0}^{n-1} X_{i+u,j+v,c} \cdot W_{u,v,c} + b,
\end{equation*}
where the filter now has dimension $m \times n \times C$ and aggregates information across all input channels to produce a single output value at each spatial location.
For example, in \cref{fig:convfilter}, we show how a filter is applied over an image with three color channels (red, green, and blue), so $C = 3$.
In this case, the filter has dimension $m \times n \times 3$, which is vectorized to a weight vector, $w$, with $3mn$ elements. 
The \emph{stride} of the filter describes how many positions it shifts by when sliding over the input. 
The output of the filter is then passed through a nonlinear activation\sidenote{Typically a ReLU function.}.

\begin{figure}[ht] 
\centering
\includegraphics[width=0.5\textwidth]{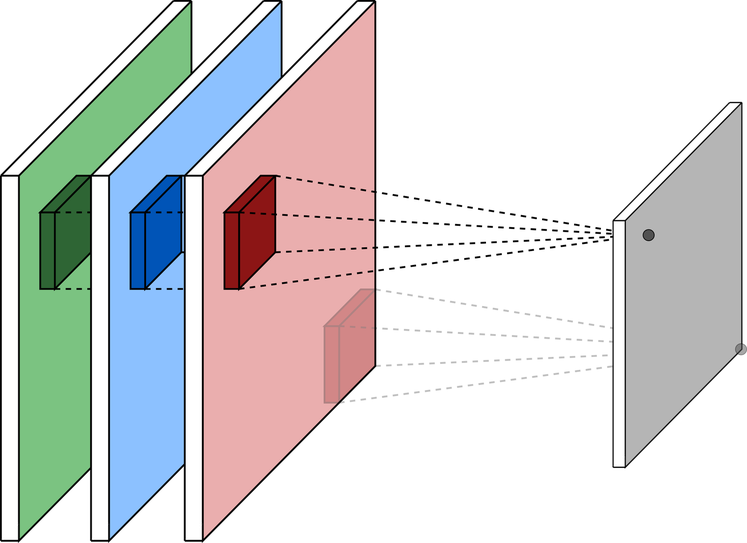}
\caption{The outputs of a convolution filter and activation function applied across an image make up a new image, called an \emph{activation} map.}
\label{fig:cnn-activations}
\end{figure}

\begin{figure}[ht] 
\centering
\includegraphics[width=0.65\textwidth]{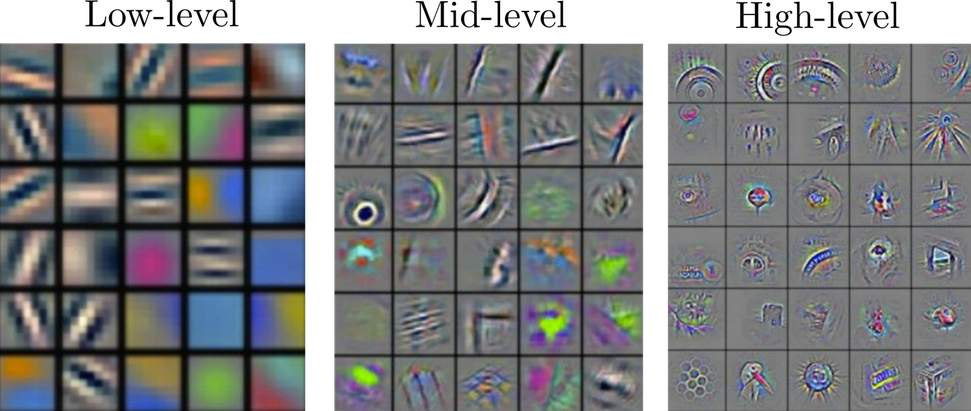}
\caption{Low-level, mid-level, and high-level feature visualizations in a convolutional neural network from Zeiler and Fergus (2014)\nocite{ZeilerFergus2014}.}
\label{fig:convfeatures}
\end{figure}

Once we have applied the filter to the entire image, the collection of outputs from the nonlinear activation function creates a new filtered image, which we typically refer to as an \emph{activation map}, as shown in \cref{fig:cnn-activations}.
In practice, a number of different filters are usually learned in each convolution layer, which produces a corresponding number of activation maps as the output\sidenote{Besides the number of filters applied to the input, the width and height of the filter, the amount of padding on the input, and the stride of the filter are other hyperparameters.}. 
This is crucial such that each filter can focus on learning one specific relevant feature. 
We show examples of different filters that might be learned in different convolution layers of a CNN in \cref{fig:convfeatures}\cite{ZeilerFergus2014}. 
Notice that the low-level features which are learned in earlier convolution layers look a lot like edge detectors, which are more basic and fundamental features, while later convolution layers have filters that look more like actual objects.

In general, using convolution layers to exploit the spatial locality of images provides several benefits.
First, \emph{parameter sharing} applies the same filter parameters at all spatial locations, keeping the total number of learned parameters much smaller than fully-connected layers would require.
Second, \emph{sparse interactions} from having filters smaller than the image enable better detection of small, meaningful features and improve computational efficiency through fewer operations.
Third, convolutional layers are \emph{equivariant to translation}, meaning that convolving a shifted image produces the same result as shifting the convolution output of the original image\sidenote{However, convolution is not equivariant to changes in scale or rotation.}, allowing feature detection regardless of position.
Finally, convolutional layers can naturally handle images of varying sizes when needed.

\subsubsection{Important CNN Components}
\begin{marginfigure} 
\centering
\includegraphics[width=0.9\textwidth]{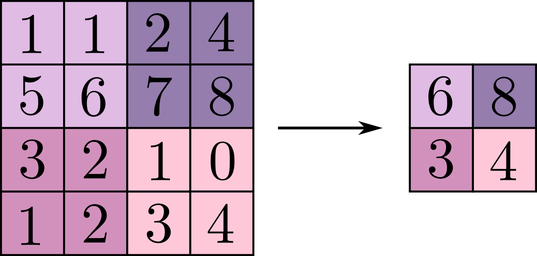}
\caption{Max pooling example with $2 \times 2$ filter and stride of $2$.}
\label{fig:pooling}
\end{marginfigure}

In addition to convolution layers, there are several other important components that make up the architecture of a CNN.
\paragraph{Pooling layers.}
Pooling is the second major structural component in CNNs. 
Pooling layers typically come after convolution layers and their nonlinear activation functions. 
The primary function of a pooling layer is to replace the output of the convolution layer's activation map at particular locations with a \emph{summary statistic} from other spatially local outputs. 
This helps make the network more robust against small translations in the input, helps improve computational efficiency by reducing the size of the input\sidenote[][-4\baselineskip]{This occurs because it lowers the resolution.}, and is useful in enabling the input images to vary in size\sidenote[][-3\baselineskip]{The size of the pooling can be modified to keep the size of the pooling layer output constant.}. 
The most common type of pooling is \emph{max pooling}, but other types also exist, such as \emph{mean pooling}. A typical max-pooling operation is shown in \cref{fig:pooling}.

Computationally, both max and mean pooling layers operate with the same filtering idea as in the convolution layers. 
Specifically, a filter of width, $m$, and height, $n$, slides around the layer's input with a particular stride. 
The difference between the two comes from the mathematical operation performed by the filter, which as their names suggest are either a maximum element or the mean over the filter. 
If the output of the convolution layer has $N$ activation maps, the output of the pooling layer will also have $N$ images, since the pooling filter is only applied across the spatial dimensions.

\paragraph{Fully connected layers.}
Downstream of the convolution and pooling layers are fully connected layers. 
These layers make up what is essentially just a standard neural network, which is appended to the end of the network. 
The function of these layers is to take the output of the convolution and pooling layers, which we can think of as a highly condensed representation of the image, and perform a classification or regression. 
Generally, the total number of fully connected layers at the end of the CNN will only make up a fraction of the total number of layers.

\paragraph{CNN performance.}
We can say that a CNN learns how to process images \emph{end-to-end} because it essentially learns how to perform two steps simultaneously: feature extraction and classification or regression\sidenote{In other words, it learns the entire process from image input to the desired output.}. 
In contrast, classical approaches to image processing use hand-engineered feature extractors. 
Since 2012, the performance of end-to-end learning approaches to image processing have dominated and continue to improve\sidenote{In some specific applications, hand-engineered features may still be better. 
For example, we might use engineering insight to identify a structure to the problem that a CNN could not easily learn.}. 
This continuous improvement has generally been realized with the use of deeper networks with more parameters, and also by combining CNN architectures with other techniques such as Transformers.

\subsubsection{Notable CNN Architectures}

Several landmark CNN architectures have significantly advanced computer vision and demonstrated the power of deep learning.

\paragraph{AlexNet (2012).} AlexNet was the first deep CNN to achieve breakthrough performance on ImageNet, popularizing the use of ReLU activations and dropout regularization. It demonstrated that deeper networks could dramatically outperform traditional hand-engineered methods, marking a turning point in computer vision.

\paragraph{ResNet (2015).} ResNet introduced residual connections that allow information to skip layers, enabling the training of much deeper networks—up to 152 layers—without suffering from vanishing gradients. ResNet showed that network depth itself could be a key factor in improving performance, establishing residual connections as a fundamental architectural component.

\paragraph{YOLO (You Only Look Once).} YOLO pioneered real-time object detection by treating detection as a single regression problem rather than a multi-stage classification task. YOLO demonstrated how CNN architectures can be adapted for various computer vision tasks beyond image classification, proving that speed and accuracy need not be mutually exclusive.

These architectures have not only achieved state-of-the-art results in their respective domains but have also influenced countless subsequent designs and established important principles for CNN development.

\section{Transformers}
\label{sec:ch11_transformers}

Transformers are deep learning architectures that have been widely applied across various domains, including natural language processing, computer vision, robotics, and more. Compared to CNNs, Transformers enforce fewer structural constraints on the input data. As long as we can organize the input into a set or sequence of tokens, it can be processed by a Transformer-based model. Since the internal structure of the Transformer is purely learned from data rather than hand-engineered, it requires less domain-specific knowledge, which makes it easier to generalize across different data modalities. Transformers are also computationally efficient and scalable due to their ability to be parallelized, allowing for extremely large models with hundreds of billions of parameters to be trained.

\subsubsection{Transformer Architecture Fundamentals}

The core innovation of the Transformer architecture is the self-attention mechanism, which allows the model to learn relationships between different elements in the input sequence or set. Unlike CNNs that have built-in spatial inductive biases through convolution operations, Transformers learn all spatial and semantic relationships directly from data. This flexibility comes at a cost: Transformers typically require larger datasets to achieve similar performance as architectures with stronger inductive biases, but they can also achieve superior performance when sufficient data is available.

\paragraph{Tokens and embeddings.}

Transformers take inputs in the form of a set or sequence of tokens. A \emph{token} is a numeric representation of the raw input data, expressed as a vector. The process of converting raw inputs into tokens is called \emph{tokenization}, and the specific method depends on the data modality. For language tasks, a token might represent a word, subword, or character, generated through a learned dictionary lookup. For computer vision tasks, a token typically represents a \emph{patch}—a square subset of the input image. For example, we can convert an image patch of size $P \times P$ with $C$ color channels into a token vector of size $CP^2$ by flattening the patch into a one-dimensional vector.

We then transform each token into a token \emph{embedding} vector, which is a high-dimensional latent space representation. The embedding process is typically a learned linear operation that maps tokens into a space where semantically or structurally similar tokens have similar representations. We also add a \emph{positional embedding} to each token embedding to encode information about the token's position in the sequence or spatial location. This positional information is crucial because the self-attention mechanism itself is permutation-invariant and does not inherently encode order or position. For example, if a sentence places \emph{dog} before \emph{cat}, the model needs that ordering to capture the sentence's correct meaning. Similarly, in vision tasks, knowing which patch came from the top-left versus bottom-right of an image provides essential spatial context\sidenote{Positional embeddings can be learned parameters or fixed sinusoidal functions. Learned embeddings are more flexible but require more data, while fixed embeddings can generalize to sequence lengths not seen during training.}.

Once we have transformed the raw input into embedding vectors, we aggregate them as rows of an input matrix, $X^0 \in \R^{N \times D}$, where $N$ is the \emph{context size}\sidenote{In general, larger context sizes will give better performance because we can capture more unique information. However, this comes at the cost of increased computational requirements, which scale quadratically with context size due to the attention mechanism.} and $D$ is the dimensionality of the embedding space.

\paragraph{Self-attention mechanism.}

The self-attention mechanism is the key component that allows Transformers to learn relationships between all elements in the input. Each embedding vector in $X^0$ initially represents only a single token and does not contain contextual information from other tokens. For example, when the phrase \emph{toy car} appears in a sentence, the embedding vector for \emph{car} does not yet encode the modifier \emph{toy}. The goal of self-attention is to allow the model to modify each embedding vector based on its relevance to all other tokens in the sequence.

Mathematically, \emph{scaled dot-product} self-attention is defined as:
\begin{equation}
    O = \mathrm{Attention}(Q, K, V) = \mathrm{softmax}\left(\frac{QK^\top}{\sqrt{d_k}}\right)V,
    \label{eq:scaled_dot_product_eq}
\end{equation}
where $Q \in \R^{m \times d_k}$ is the \emph{query} matrix, $K \in \R^{n \times d_k}$ is the \emph{key} matrix, $V \in \R^{n \times d_v}$ is the \emph{value} matrix, and $O \in \R^{m \times d_v}$ is the \emph{output} matrix. The intuition behind this formulation is that the term $QK^\top$ computes the dot product between all pairs of queries and keys, where a larger dot product indicates that a particular pair are similar or relevant to each other. We then apply the softmax function to each row of the resulting matrix to normalize the rows so that their elements sum to one\sidenote{The division by $\sqrt{d_k}$, where $d_k$ is the dimensionality of the keys and queries, helps stabilize gradients during training by preventing the dot products from becoming too large.}. Finally, multiplication by the value matrix $V$ produces an output where each row is a weighted sum of all value vectors, with weights determined by the normalized attention scores.

In the Transformer architecture, we typically define these matrices as $Q = XW_Q$, $K = XW_K$, and $V = XW_V$, where $X$ is the input matrix with each row corresponding to one embedding vector, and $W_Q$, $W_K$, and $W_V$ are matrices of learnable parameters. While we present the mathematical form using matrix notation for implementation clarity, it is often easier to reason about the transformation of a single embedding vector. Returning to our toy car example, the attention mechanism would modify the \emph{car} embedding vector into a new vector that captures relevant information from other words in the sentence. The value matrix $V$ provides potential modifications corresponding to each word, and the attention scores determine which modifications to apply. In this case, the relevance weighting would likely show a strong match between \emph{car} and \emph{toy}, resulting in an updated embedding that represents \emph{toy car} rather than a generic car.

A single attention mechanism is limited in the types of relationships it can capture, constrained by the finite parameter matrices $W_Q$, $W_K$, and $W_V$. To increase the model's expressive capacity, Transformers use \emph{multi-head attention}, which runs multiple attention mechanisms in parallel, each with unique parameter matrices $W_Q^i$, $W_K^i$, and $W_V^i$ for $i = 1, \ldots, h$, where $h$ is the number of attention heads. The outputs from all heads are concatenated and multiplied by another learned matrix $W_O$ to produce the final output. This allows the model to attend to different types of relationships simultaneously—for example, one head might learn syntactic relationships while another learns semantic relationships.

\paragraph{Transformer layer components.}

\begin{figure}[ht]
\centering
\includegraphics[width=0.5\textwidth]{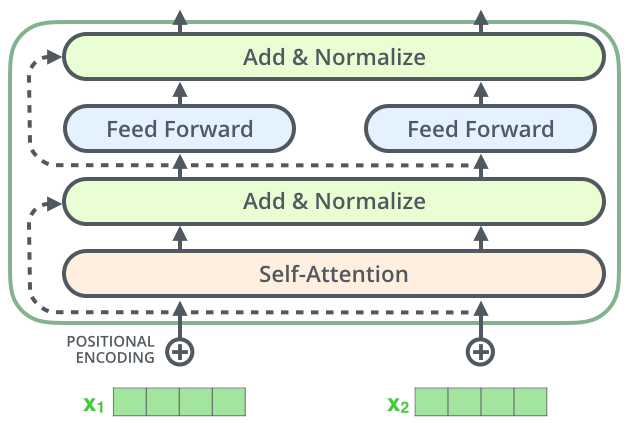}
\caption{A single Transformer layer showing the multi-head attention block and feed-forward network, each with residual connections and layer normalization from \textit{the Illustrated Transformer} (2018)\nocite{alammar2018illustrated}.}
\label{fig:transformer-layer}
\end{figure}

Each Transformer layer consists of two main components arranged sequentially: a multi-head self-attention layer followed by a feed-forward network, with both components wrapped in residual connections and layer normalization operations, as shown in \cref{fig:transformer-layer}.

After the multi-head self-attention layer processes the input, we add the attention output back to the input in what is called a \emph{residual connection}:
\begin{equation*}
\bar{X} = X + \mathrm{MultiHeadAttention}(X).
\end{equation*}
This residual connection allows gradients to flow more easily during training and helps the network learn identity mappings when beneficial. We then apply layer normalization to stabilize training:
\begin{equation*}
\tilde{X} = \mathrm{LayerNorm}(\bar{X}).
\end{equation*}

Next, we pass the normalized output through a position-wise feed-forward network, typically implemented as a \emph{multi-layer perceptron} (MLP). This MLP consists of two linear transformations with a non-linear activation function\sidenote{The ReLU or GELU activation functions are commonly used.} in between:
\begin{equation*}
\mathrm{FFN}(x) = W_2 \cdot \mathrm{ReLU}(W_1 x + b_1) + b_2,
\end{equation*}
where the MLP is applied independently to each position (each row of the input matrix). The feed-forward network typically expands the dimensionality in the first layer and then projects back to the original dimension in the second layer, allowing the network to learn complex non-linear transformations of the attention outputs. This component is followed by another residual connection and layer normalization:
\begin{equation*}
X' = \mathrm{LayerNorm}(\tilde{X} + \mathrm{FFN}(\tilde{X})).
\end{equation*}

A complete Transformer architecture consists of multiple such layers stacked in sequence, where the output of one layer becomes the input to the next. The depth of the network (number of stacked layers) is a key hyperparameter that significantly impacts model capacity and performance\sidenote{Modern Transformers can have dozens or even hundreds of layers. For example, GPT-3 has 96 layers, while some vision models use 32 or more layers.}. Note that the dimensions of the inputs and outputs of each Transformer layer are typically the same ($N \times D$), allowing for flexible stacking of arbitrary depth.

\paragraph{Final linear and unembedding layer.}

After passing through all Transformer layers, we must convert the final embedding representations back into a format suitable for the specific task. The \emph{unembedding layer} transforms the learned embedding vectors into task-specific outputs. For classification tasks, we typically take a single embedding vector—either a special classification token or an aggregated representation of all tokens—and pass it through a linear layer followed by a softmax function:
\begin{equation*}
    o = \mathrm{softmax}(xW + b),
\end{equation*}
where $x$ is the selected embedding vector, and $W$ and $b$ are learned parameters. The output $o$ is a probability distribution over the possible classes, where the vector size matches the number of classes and all elements sum to one.

For other tasks, the unembedding layer may take different forms. In sequence-to-sequence tasks like machine translation, we apply a linear transformation to each position's embedding to predict the next token in the output sequence. In dense prediction tasks like image segmentation, we may upsample the embeddings back to the original input resolution and apply per-position classification. The specific design of the final layers depends entirely on the task requirements, while the core Transformer layers remain largely the same across different applications.

\subsubsection{Vision Transformers (ViTs)}

Vision Transformers\cite{DosovitskiyEtAl2021} represent the most important Transformer architecture for robotics practitioners, serving as the foundation for modern perception pipelines in embodied AI systems. Unlike earlier vision models that required careful hand-engineering of features and architectural components, ViTs enable end-to-end learning from raw images to task-specific outputs. They have become ubiquitous in robotic applications, from object recognition and scene understanding in autonomous navigation systems to visual representations for manipulation policies and multi-modal reasoning in household robots. The ability to pre-train ViTs on large-scale image datasets and then fine-tune them for specific robotic tasks with limited data has made them particularly valuable for real-world deployments where collecting task-specific training data is expensive or impractical.

\paragraph{Architecture adaptation for images.}

Vision Transformers adapt the general Transformer architecture to process images by treating image patches as tokens. Given an input image of size $H \times W$ with $C$ color channels, ViT divides the image into a grid of non-overlapping patches of size $P \times P$, resulting in $N = \frac{HW}{P^2}$ patches. Each patch is flattened into a vector of dimension $C \cdot P^2$ and then linearly projected to the embedding dimension $D$ through a learned embedding matrix. Common choices for patch size include $P = 16$ or $P = 32$, which balance the trade-off between computational cost (smaller patches create more tokens) and the ability to capture fine-grained details\sidenote{For a standard $224 \times 224$ image with $P = 16$, this creates $14 \times 14 = 196$ patch tokens.}.

In addition to the patch embeddings, ViT introduces a special learnable \emph{class token}, denoted $x_{\text{cls}}$, which is prepended to the sequence of patch embeddings. This class token has no correspondence to any image patch but serves as a global representation that the network can use to aggregate information from all patches for classification tasks. The class token's embedding after passing through all Transformer layers is used as the input to the final classification head.
Vision Transformers use learnable 2D positional embeddings that encode each patch's spatial location in the original image grid. Unlike 1D positional embeddings used in language models, these embeddings must capture 2D spatial structure. The standard approach uses a separate learned embedding for each position in the patch grid, which is added to the corresponding patch embedding. Alternative approaches include using sinusoidal positional encodings extended to 2D or learning relative positional biases. The complete input to the first Transformer layer is thus:
\begin{equation*}
X^0 = [x_{\text{cls}}, x_1^p + e_1^{\text{pos}}, x_2^p + e_2^{\text{pos}}, \ldots, x_N^p + e_N^{\text{pos}}],
\end{equation*}
where $x_i^p$ are the linearly embedded patches and $e_i^{\text{pos}}$ are the positional embeddings.

\paragraph{Training considerations.}

A key characteristic of Vision Transformers is their data requirements compared to CNNs. Because ViTs lack the built-in inductive biases of convolution operations—such as translation equivariance and local connectivity—they require substantially larger training datasets to achieve comparable performance with similar model sizes. When trained on smaller datasets like ImageNet-1K (1.3 million images), ViTs typically underperform similarly-sized CNNs. However, when pre-trained on larger datasets such as ImageNet-21K (14 million images) or even larger proprietary datasets, ViTs can match or exceed CNN performance, and their performance continues to improve with dataset scale in ways that CNNs do not.

This observation has led to a standard training paradigm for ViTs: pre-training on large-scale datasets followed by fine-tuning on smaller, task-specific datasets. Pre-training can be done with supervised learning on labeled image datasets or through self-supervised methods that learn representations from unlabeled images. Self-supervised pre-training methods such as masked image modeling—where random patches are masked and the model learns to predict them—have proven particularly effective for ViTs and enable training on massive unlabeled image corpora. After pre-training, the model can be fine-tuned on downstream tasks with relatively small amounts of labeled data, often requiring only the final classification layers to be retrained while the Transformer layers remain largely fixed or are fine-tuned with small learning rates.

For robotics applications, this pre-training and fine-tuning paradigm is especially valuable. Robotic systems often operate in specialized environments with limited task-specific data, but can leverage visual representations learned from generic internet-scale image datasets. A ViT pre-trained on diverse visual data captures general-purpose features like edge detection, object recognition, and spatial relationships that transfer effectively to robotic tasks like grasping, navigation, and scene understanding.

\paragraph{Practical considerations.}

The choice between Vision Transformers and CNNs for robotic applications involves several practical trade-offs. ViTs excel when large pre-trained models are available and when the task benefits from modeling long-range dependencies across the entire image. Their flexibility in handling variable input sizes and their ability to scale to very large model sizes make them attractive for applications where computational resources are available and high accuracy is critical. However, CNNs remain competitive or superior in scenarios with limited data, real-time constraints on resource-constrained hardware, or tasks where strong spatial inductive biases are beneficial.

Computationally, ViTs have quadratic complexity with respect to the number of patches due to the self-attention mechanism, making them more expensive than CNNs for high-resolution images. Recent variants like Swin Transformer address this by using local attention windows and hierarchical structures, combining some of the efficiency benefits of CNNs with the flexibility of Transformers. For robotics practitioners, the decision often comes down to whether pre-trained models are available for the specific visual domain, the computational budget of the deployment platform, and whether the task requires the global reasoning capabilities that attention mechanisms provide. In many modern robotic systems, hybrid approaches that combine CNN backbones for efficient feature extraction with Transformer layers for high-level reasoning have proven effective, leveraging the complementary strengths of both architectures.

\section{Point Cloud Processing and Point-Based Networks}
\label{sec:ch11_pointbased}

Robotic systems operating in 3D environments frequently rely on LiDAR sensors, depth cameras, and other sensing modalities that produce point cloud data. Point clouds represent 3D data as collections of points in space, where each point is defined by its position $\mathbf{p}$ and may include additional attributes such as color, intensity, or surface normals\sidenote{Common representations include Cartesian coordinates $(x, y, z)$, spherical coordinates $(r, \theta, \phi)$, or cylindrical coordinates $(r, \theta, z)$ depending on the sensor type.}. In autonomous driving, LiDAR sensors capture millions of points per second to create detailed 3D maps enabling real-time obstacle detection and navigation. Unlike images with regular 2D grid structure, point clouds possess unique characteristics requiring specialized neural network architectures.

\subsubsection{Point Cloud Characteristics}
Point clouds differ fundamentally from images in three key ways that shape neural network design.
Point clouds are \emph{unordered sets} with no inherent sequential or spatial ordering. A point cloud with $N$ points can be represented as $\{\mathbf{p}_1, \mathbf{p}_2, \ldots, \mathbf{p}_N\}$ where each point $\mathbf{p}_i \in \R^d$ (typically $d = 3$). Crucially, any permutation of these points represents the same geometric object, requiring neural networks to be \emph{permutation invariant}.
\begin{figure}[ht]
\centering
\includegraphics[width=0.75\textwidth]{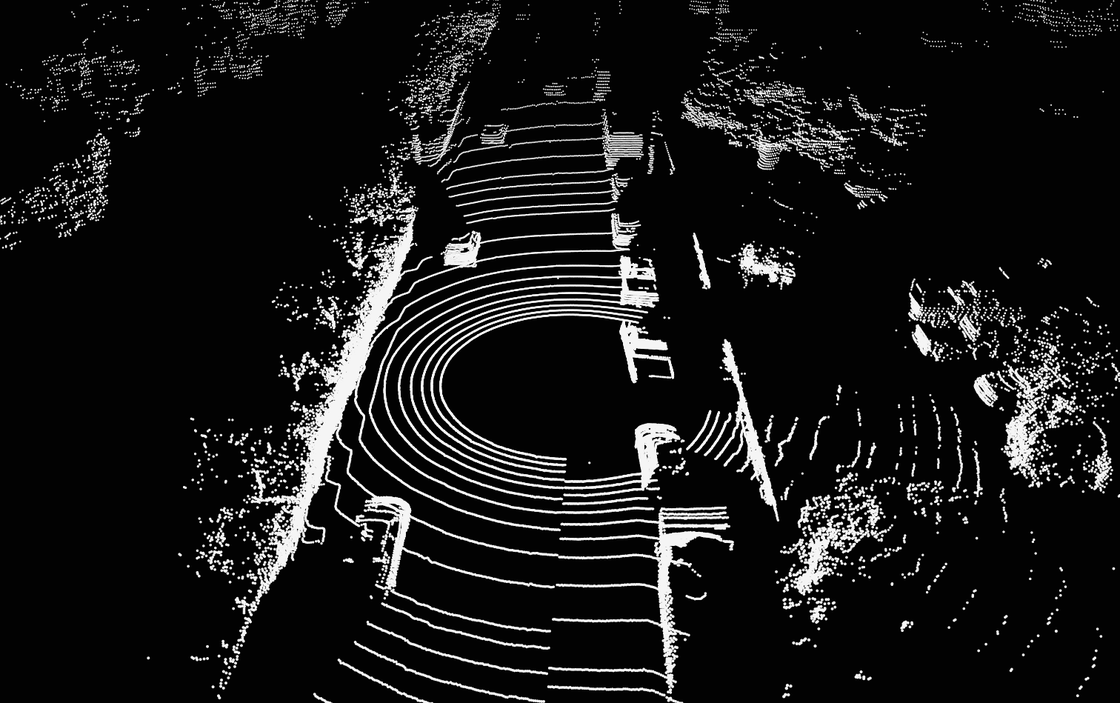}
\caption{LiDAR point cloud data from an autonomous vehicle showing cars and road infrastructure.}
\label{fig:lidar-pointcloud}
\end{figure}

\begin{figure}[ht]
\centering
\includegraphics[width=0.75\textwidth]{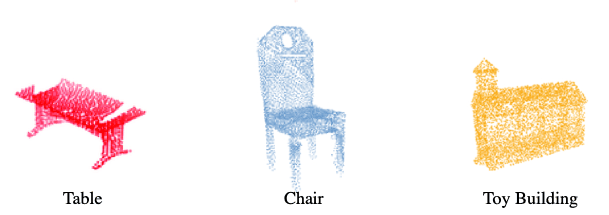}
\caption{Point clouds of different objects showing variable cardinality: a simple table (fewer points), a detailed chair (more points), and a toy building facade (most points).}
\label{fig:variable-cardinality}
\end{figure}

Point clouds exhibit \emph{variable cardinality}, containing vastly different numbers of points across samples. While images have fixed resolution (e.g., $224 \times 224$ pixels), point clouds might contain anywhere from 1,000 to 100,000 points depending on scanning resolution, sensor distance, and object complexity, posing challenges for batch processing and network design. \cref{fig:lidar-pointcloud} and \cref{fig:variable-cardinality} illustrate this variability.
Point clouds are also \emph{sparse and irregular}, with points distributed non-uniformly throughout 3D space at varying local densities determined by scanning angle, distance, and surface properties. Most 3D space contains no points, and local neighborhood structure around each point varies significantly, contrasting sharply with the dense, regular grid of image pixels.

\subsubsection{PointNet Architecture}

PointNet\cite{QiEtAl2017a} approaches point cloud processing through several key architectural innovations that ensure permutation invariance while extracting meaningful geometric features. The architecture consists of point-wise feature extraction, symmetric aggregation functions, and spatial transformation components that work together to process unordered point sets effectively.

\begin{figure}[ht]
\centering
\includegraphics[width=0.85\textwidth]{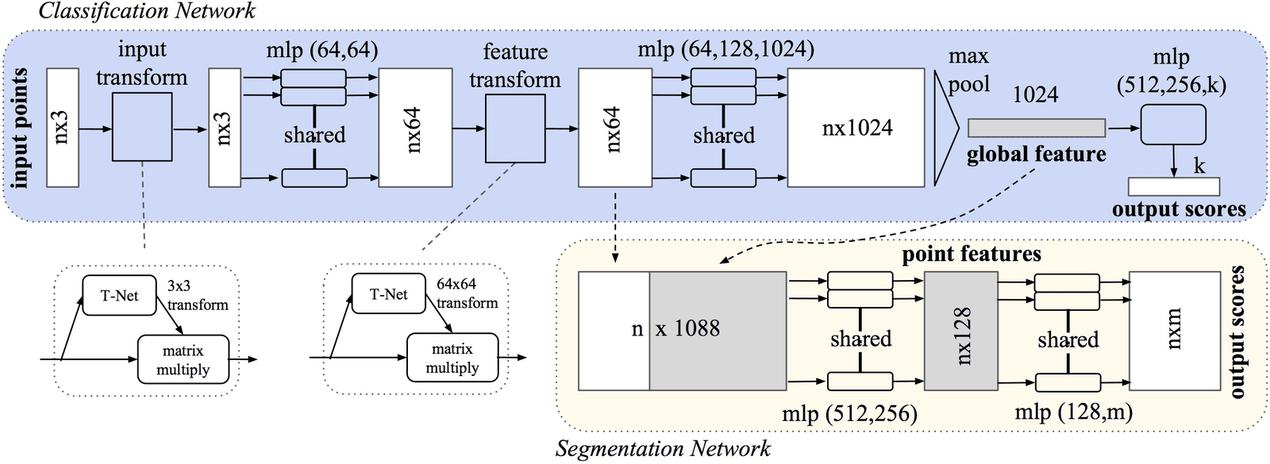}
\caption{PointNet architecture showing point-wise MLPs, transformation networks (T-Net), and symmetric aggregation for classification and segmentation tasks, from Qi et al. (2017)\nocite{QiEtAl2017a}.}
\label{fig:pointnet-architecture}
\end{figure}

\paragraph{Point-wise multi-layer perceptrons.}
PointNet applies multi-layer perceptrons (MLPs) independently to each point in the cloud. Given $N$ points where each point $p_i \in \R^3$ represents spatial coordinates, PointNet computes:
\begin{equation*}
    h_i = \text{MLP}(p_i),
\end{equation*}
where $h_i \in \R^k$ is the learned feature representation. This point-wise processing maintains permutation invariance because the same transformation applies to each point regardless of input order. The MLP consists of fully connected layers with ReLU activations, progressively increasing dimensionality from 3D coordinates to higher-dimensional spaces (e.g., 64, 128, 1024 dimensions). The MLP parameters are shared across all points, similar to parameter sharing in CNN filters, but without spatial locality constraints. The overall architecture is shown in \cref{fig:pointnet-architecture}.

\paragraph{Symmetric aggregation functions.}
After extracting point-wise features, PointNet must aggregate these features into a single global representation while preserving permutation invariance. This is achieved through symmetric functions that produce the same output regardless of input ordering. The most commonly used symmetric function in PointNet is the element-wise maximum:
\begin{equation*}
    g = \maximize[i=1,\ldots,N] h_i,
\end{equation*}
where the max operation is applied element-wise across all feature vectors $h_i$. 
This is provably permutation invariant because for any permutation $\sigma$:
\begin{equation*}
\max_i h_{\sigma(i)} = \max_i h_i.
\end{equation*}
While alternative symmetric functions like summation or mean could be used, max pooling has the advantage of being selective, allowing the network to focus on the most discriminative features across all points. However, this global aggregation approach means that PointNet captures only global features and may miss important local geometric structures, which motivates the hierarchical extensions in PointNet++, which we will discuss later in this section.

\paragraph{Transformation networks (T-Nets).}
To achieve invariance to geometric transformations such as rotation and translation, PointNet incorporates transformation networks (T-Net) that learn to align point clouds to a canonical orientation. The T-Net is itself a mini-PointNet that predicts a transformation matrix $T \in \R^{k \times k}$:
\begin{equation*}
    T = \text{T-Net}(\{p_1, p_2, \ldots, p_N\}).
\end{equation*}
This transformation matrix is then applied to either the input coordinates (input transform) or intermediate features (feature transform). For the input transform, $T \in \R^{3 \times 3}$ aligns the spatial coordinates, while for the feature transform, $T \in \R^{64 \times 64}$ normalizes the feature space. To ensure the stability of optimization, a regularization term is added to the loss function that encourages the transformation matrix to be close to orthogonal:
\begin{equation*}
    L_{\text{reg}} = ||I - TT^T||_F^2,
\end{equation*}
where $||\cdot||_F$ denotes the Frobenius norm. This regularization prevents the transformation from becoming degenerate and helps maintain the geometric properties of the point cloud.

\subsubsection{Notable Point-Based Architectures}

Several landmark point-based architectures have extended PointNet's core ideas to address its limitations and improve performance on complex 3D understanding tasks.

\paragraph{PointNet++.}
While PointNet effectively captures global features, it struggles to learn local geometric patterns due to its reliance on global max pooling. PointNet++\cite{QiEtAl2017b} addresses this limitation by introducing a hierarchical architecture that learns features at multiple scales, similar to how CNNs build hierarchical representations through multiple convolutional layers.
PointNet++ introduces set abstraction layers that recursively apply PointNet to local regions. Given a point cloud with $N$ points, each set abstraction layer samples $N'$ representative points (where $N' < N$), groups nearby points around each representative point, and applies a PointNet to extract local features. This process creates a hierarchical pyramid of features, where early layers capture fine-grained local details and later layers capture broader geometric patterns.

Set abstraction in PointNet++ consists of three key operations: sampling, grouping, and feature extraction. \emph{Sampling} uses farthest point sampling (FPS) to select representative points that provide good coverage of the entire point cloud. Given a set of points, FPS iteratively selects the point that is farthest from all previously selected points, ensuring diverse spatial coverage. \emph{Grouping} then defines local regions around each selected point using either ball query (all points within radius $r$) or $k$-nearest neighbors. This creates local point sets of varying sizes that capture the local geometry around each representative point. \emph{Feature extraction} applies PointNet to each local region to learn features that capture local geometric patterns while maintaining permutation invariance within each region.
For tasks requiring point-wise predictions like semantic segmentation, PointNet++ includes feature propagation layers that upsample features from coarser to finer resolutions. These layers use inverse distance weighted interpolation to propagate features from subsampled points back to the original point cloud:
\begin{equation*}
    f^{(j)}(x) = \frac{\sum_{i=1}^k w_i(x) f_i^{(j-1)}}{\sum_{i=1}^k w_i(x)}, \quad w_i(x) = \frac{1}{d(x, x_i)^p},
\end{equation*}
where $f^{(j)}$ represents features at layer $j$, $d(x,x_i)$ is the distance between points, and $p$ is typically set to 2. Skip connections between corresponding abstraction and propagation layers help preserve fine-grained details, similar to U-Net architectures in image segmentation.

\paragraph{Dynamic graph CNN (DGCNN).}
An alternative approach to processing point clouds treats them as graph structures, where points serve as nodes and edges are defined based on spatial proximity or learned relationships. Dynamic Graph Convolutional Neural Networks (DGCNN)\cite{WangEtAl2019} exemplify this approach by constructing graphs dynamically in feature space rather than just coordinate space. DGCNN applies edge convolution operations that aggregate information from neighboring points:
\begin{equation*}
    x_i' = \maximize[j:(i,j) \in \mathcal{E}] h_\theta(x_i, x_j - x_i),
\end{equation*}
where $x_i$ and $x_j$ are feature vectors of connected points, $h_\theta$ is a learnable function (typically an MLP), and the edge set $\mathcal{E}$ is dynamically updated based on feature similarity after each layer. This dynamic graph construction allows the network to capture both geometric and semantic relationships that evolve as features are learned. The edge convolution operation differs from standard graph convolutions by explicitly modeling the edge information $(x_j - x_i)$, which captures the relative geometric relationships between neighboring points. 
This approach has shown success in tasks like point cloud classification and part segmentation.

\subsubsection{Applications and Limitations}
After processing through these point-based architectures—PointNet's point-wise MLPs and symmetric aggregation, PointNet++'s hierarchical set abstraction layers, or DGCNN's dynamic graph convolutions—the networks produce rich point-wise feature representations that encode both local geometric patterns and global shape information. These learned features serve as inputs for downstream tasks including 3D object detection, semantic segmentation of points into categories like road, building, or vegetation, and instance segmentation for identifying individual objects. We will explore training methods for these detection and segmentation tasks in subsequent chapters.

Despite their effectiveness, point-based methods face computational challenges when processing large-scale point clouds. Real-world applications like autonomous driving can generate point clouds with millions of points per frame, making the $O(N^2)$ complexity of neighborhood search in DGCNN or the recursive sampling in PointNet++ computationally prohibitive. Memory requirements also scale poorly, as each point must be processed individually, leading to irregular memory access patterns that are inefficient on modern GPU architectures. In the following section, we discuss voxel-based and pillar-based approaches that leverage regular grid structures for efficient 3D convolutions. By discretizing 3D space into regular voxels or vertical pillars, these methods can apply standard convolutional operations while maintaining spatial locality and enabling efficient parallel processing. This structured representation trades some geometric precision for computational efficiency and scalability, making it particularly suitable for real-time applications in autonomous driving and robotics where processing speed is critical.

\section{Voxel-Based 3D Processing}
\label{sec:ch11_voxelbased}

In the previous section, we discussed that the computational limitations of point-based methods have motivated the development of grid-based approaches that discretize 3D space into regular structures, enabling the application of efficient convolutional operations. Rather than processing individual points with irregular neighborhoods, voxel-based and pillar-based methods transform point clouds into structured representations where standard CNNs can be applied. This paradigm shift trades some geometric precision for substantial computational advantages, making real-time processing of large-scale point clouds feasible for applications like autonomous driving. By leveraging the regularity of grid structures, these methods can utilize optimized convolution implementations and parallel processing capabilities of modern hardware.

\subsubsection{Grid-Based Representations}

Grid-based methods transform irregular point clouds into structured representations by discretizing 3D space into regular units. Two primary approaches have emerged: voxel-based representations that divide space into cubic voxels, and pillar-based\sidenote[][-1cm]{Frequently used in autonomous vehicle or navigation domain, where the scene can be viewed as a 2D ``map'' instead of a true 3D scene, for computational efficiency.} representations that use vertical columns extending through the entire height of the scene.

\begin{figure}[ht]
    \centering
    \includegraphics[width=0.65\textwidth]{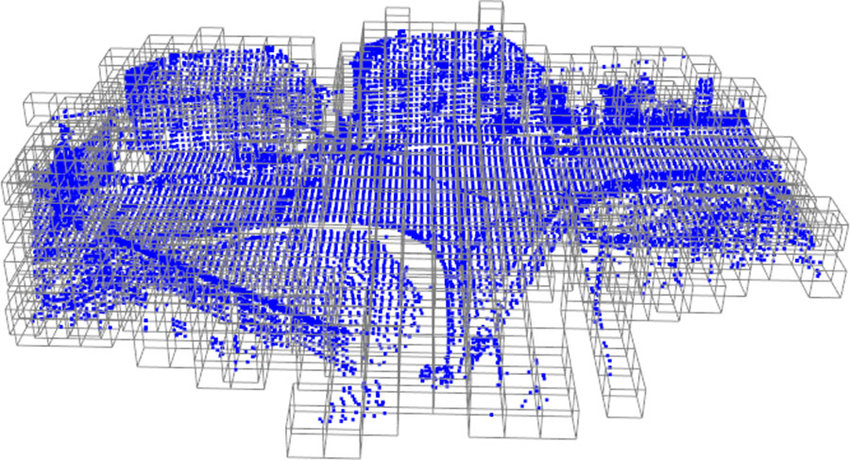}
    \caption{Comparison of point cloud representations: original point cloud (dots) overlaid on the voxel-based discretization into 3D cubic cells from Kang et al. (2018)\nocite{kang2018voxel}.}
    \label{fig:voxel-pillar-comparison}
\end{figure}

The \emph{voxel-based representation} (\cref{fig:voxel-pillar-comparison}) creates a full 3D regular grid structure of size $L \times W \times H$ by partitioning space into cubic cells of size $v_l \times v_w \times v_h$\sidenote{In practice, these dimensions are often set to be equal, creating cubic voxels.}, where each voxel can contain zero or more points from the original point cloud. This approach preserves complete spatial relationships in all three dimensions, enabling rich 3D feature learning through volumetric convolutions. Points within each voxel are aggregated into a single feature representation, typically through operations like mean pooling, max pooling, or learned aggregation functions.

In contrast, the \emph{pillar-based representation} adopts a 2.5D approach, looking at the scene from a bird's-eye view, and treats vertical columns (``pillars'') as the fundamental processing unit. The space is transformed into a 2D grid structure of size $L \times W$. Each pillar extends vertically through the entire height range of the point cloud, effectively collapsing the height dimension during initial processing. This approach is particularly useful in self-driving settings, where the scene processed is often very large and the reduction to 2D significantly improves computational efficiency. Points within each pillar are aggregated while preserving some height information through encoding strategies, but the primary spatial reasoning occurs in the horizontal, bird's-eye view plane.

The choice between these representations involves significant trade-offs in computational complexity and spatial information preservation. Voxel-based methods provide richer spatial context by maintaining full 3D neighborhood relationships with memory scaling as $O(L \times W \times H)$, enabling detection of complex 3D geometric patterns but requiring computationally expensive 3D convolutions. Pillar-based approaches reduce complexity by projecting the problem into 2D with memory scaling as $O(L \times W)$, enabling the use of mature 2D CNN architectures and optimized implementations, but potentially losing important vertical structure information crucial for multi-level feature detection. Both representations face spatial resolution trade-offs, where finer grids capture more geometric detail at exponentially higher computational cost, and must address sparsity challenges where most grid cells remain empty, motivating the development of sparse convolution techniques.

\subsubsection{3D Convolution Fundamentals}

Once point clouds are discretized into regular grid structures, we can apply convolutional operations to learn hierarchical feature representations. This section covers the fundamental operations that enable efficient processing of voxelized 3D data.

\paragraph{3D convolution operations.}

\begin{figure}[ht]
\centering
\includegraphics[width=0.4\textwidth]{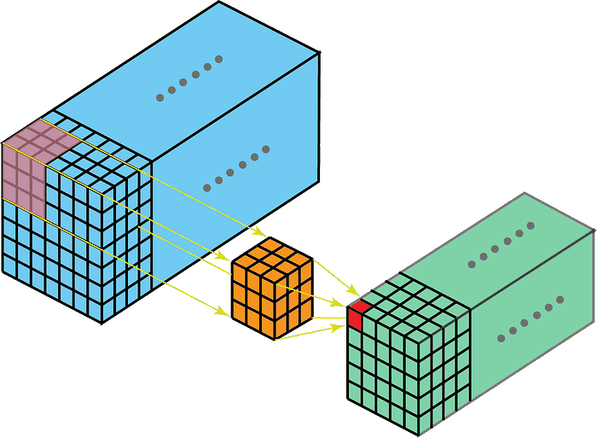}
\caption{A 3D convolution filter sliding across a volumetric input, showing how the $(k_x, k_y, k_z)$ filter operates in all three spatial dimensions.}
\label{fig:3d-convolution}
\end{figure}

3D Convolutional Neural Networks extend the successful principles of 2D CNNs to volumetric data by operating directly on 3D grids of voxels. While 2D convolutions slide filters across height and width dimensions of images, 3D convolutions add depth as a third spatial dimension, enabling the network to capture spatial relationships to understand the 3D scene.
Specifically, a 3D convolution applies a filter of size $(k_x, k_y, k_z)$ across all three spatial dimensions of the input volume. Mathematically, for an input volume $X$ and filter $W$, the 3D convolution operation can be expressed as:
\begin{equation}
    Y_{i,j,k} = \sum_{u=0}^{k_x-1} \sum_{v=0}^{k_y-1} \sum_{w=0}^{k_z-1} X_{i+u,j+v,k+w} \cdot W_{u,v,w} + b,
    \label{eq:conv_3d_equation}
\end{equation}
where $(i,j,k)$ represents the spatial position in the output volume and $b$ is the bias term. Note that in practice, both the input $X$ and output $Y$ typically have an additional channel dimension for multi-channel feature maps, and the weight $W$ is a 4D tensor that includes both spatial dimensions and input/output channel dimensions, while the bias $b$ is a vector with one element per output channel. The geometric interpretation is shown in \cref{fig:3d-convolution}.

Common kernel sizes include $3 \times 3 \times 3$ for capturing local 3D patterns and $1 \times 1 \times 1$ for channel-wise feature mixing without spatial aggregation\sidenote{Larger kernels like $5 \times 5 \times 5$ can capture broader spatial context but significantly increase computational cost due to the cubic scaling of operations.}. The key advantage of 3D convolutions over approaches that process 2D slices independently is their ability to learn features that span multiple depths, such as the full 3D shape of objects or volumetric textures. The receptive field in 3D grows cubically with network depth, allowing deeper layers to capture increasingly global context, though this rapid growth must be balanced against increased computational cost.

\paragraph{Sparse convolutions.}

Real-world point clouds exhibit extreme sparsity when discretized into voxel grids. In many applications such as autonomous driving and indoor scene processing, the majority of 3D space consists of empty air or unoccupied regions. Standard dense 3D convolutions waste significant computation on empty space, making them impractical for large-scale applications. Sparse convolutions address this by computing only on occupied voxels and their neighborhoods, using efficient data structures to maintain compact representations of non-empty regions\sidenote{Popular implementations include \texttt{spconv} for PyTorch/TensorFlow and Minkowski Engine for general sparse tensor operations.}.

\begin{example}[Memory savings in sparse convolutions]
A typical autonomous-vehicle LiDAR scene discretized at 10cm resolution over a $100m \times 100m \times 10m$ volume would require:
\begin{equation*}
    \text{Grid size} = \frac{100m}{0.1m} \times \frac{100m}{0.1m} \times \frac{10m}{0.1m} = 1000 \times 1000 \times 100,
\end{equation*}
\begin{equation*}
    \text{Total voxels} = 1000 \times 1000 \times 100 = 100 \text{ million voxel features per layer},
\end{equation*}
In contrast, sparse representations store only the occupied voxels, making memory usage proportional to the number of non-empty voxels rather than total grid size. The sparser the scene, the greater the memory savings.
\end{example}

\subsubsection{Notable Voxel-Based Architectures}

Several landmark architectures have demonstrated the effectiveness of grid-based representations for 3D perception tasks, particularly in autonomous driving applications where real-time performance is critical.

\paragraph{VoxelNet.}
VoxelNet\cite{ZhouTuzel2018} was one of the pioneering architectures for processing voxel-based representations. The architecture addresses the challenge of processing irregular point clouds by first voxelizing them into a 3D voxel grid, then leveraging 3D convolutions to extract features and detect objects.
\begin{figure}[ht]
\centering
\includegraphics[width=0.85\textwidth]{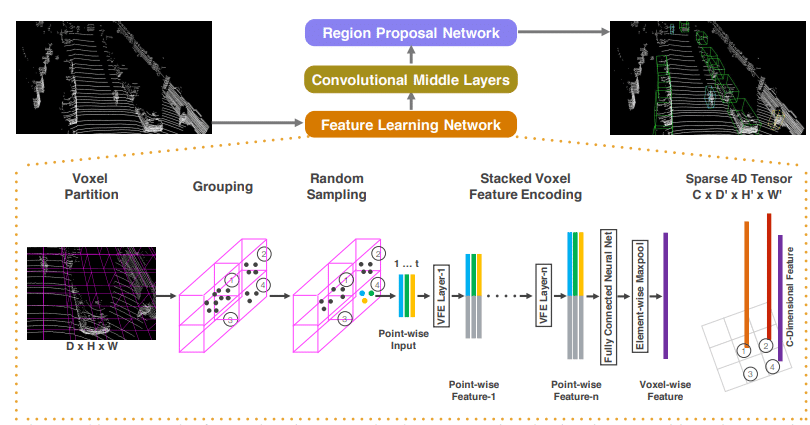}
\caption{VoxelNet architecture from Zhou and Tuzel (2018)\nocite{ZhouTuzel2018}, showing the complete pipeline from point cloud voxelization through VFE layers, 3D convolutional middle layers, to the Region Proposal Network for 3D object detection.}
\label{fig:voxelnet-architecture}
\end{figure}
A core component of VoxelNet is its Voxel Feature Encoding (VFE) layers, which process the variable number of points within each voxel to produce fixed-size feature representations. The full architecture is shown in \cref{fig:voxelnet-architecture}. Given a voxel containing points $\{p_1, p_2, \ldots, p_n\}$, where each point $p_i = (x_i, y_i, z_i, r_i)$ includes spatial coordinates and optional reflectance intensity, the VFE layers apply point-wise multi-layer perceptrons to each point independently:
\begin{equation*}
    f_i = \text{MLP}(p_i).
\end{equation*}

To capture contextual information within each voxel, VoxelNet augments each point with the centroid of all points in the same voxel. For a voxel containing $n$ points, the centroid is computed as $\bar{p} = \frac{1}{n}\sum_{i=1}^n p_i$, and each point is then represented as the concatenation $[p_i, p_i - \bar{p}]$, providing both absolute and relative spatial information. The VFE layers then aggregate features across all points in the voxel using element-wise max pooling, ensuring permutation invariance:
\begin{equation*}
    v = \maximize[i=1,\ldots,n] f_i,
\end{equation*}
where $v$ is the final voxel-level feature representation. This aggregation step converts the variable-sized point sets within each voxel into fixed-size feature vectors suitable for subsequent 3D convolution operations.

After voxel feature encoding, VoxelNet applies a series of 3D convolutional middle layers to build hierarchical representations of the scene. These layers follow standard 3D CNN design principles, progressively increasing receptive field size while extracting increasingly abstract features. The sparse nature of voxel occupancy makes this stage well-suited for sparse convolution implementations to improve computational efficiency. The 3D convolutional layers aggregate information across neighboring voxels to capture larger geometric structures, build multi-scale representations through progressive downsampling, and prepare features for the final object detection stage.

\paragraph{PointPillars.}
While VoxelNet processes full 3D voxels, PointPillars\cite{LangEtAl2019} takes a different approach by using vertical pillars that extend through the entire height of the scene. The key innovation lies in the Pillar Feature Network (PFN), which encodes points within each pillar and then converts the resulting pillar features into a 2D ``pseudo-image'' representation. This transformation allows PointPillars to leverage mature 2D CNN architectures for subsequent processing, rather than computationally expensive 3D convolutions.
\begin{figure}[ht]
    \centering
    \includegraphics[width=0.85\textwidth]{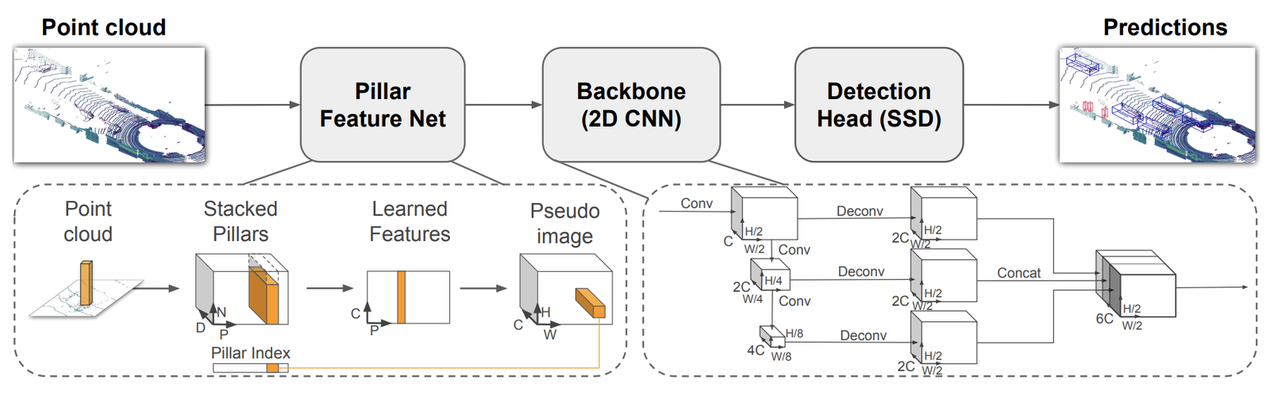}
    \caption{PointPillars architecture showing pillar feature encoding and 2D CNN backbone for efficient real-time 3D object detection, from Lang et al. (2019)\nocite{LangEtAl2019}.}
    \label{fig:pointpillar-architecture}
\end{figure}
The pillar-based approach offers significant computational advantages by reducing the problem from 3D to 2.5D, enabling the use of optimized 2D convolution operations and existing hardware accelerations designed for image processing. This design choice makes PointPillars particularly suitable for real-time applications where computational efficiency is crucial, achieving inference speeds suitable for autonomous driving while maintaining competitive detection accuracy, as shown in \cref{fig:pointpillar-architecture}.

\paragraph{Other variants.}
Building on the success of VoxelNet and PointPillars, several variants have been developed to further improve performance and efficiency. SECOND (Sparsely Embedded Convolutional Detection)\cite{YanEtAl2018} combines voxel-based processing with sparse convolution techniques for improved efficiency, significantly reducing computational requirements while maintaining accuracy. Voxel R-CNN extends the voxel-based approach with refined detection stages, while VoxelNet introduces architectural improvements that further push the boundaries of voxel-based 3D detection performance.
Similar to point-based architectures, after processing through these voxel-based architectures—VoxelNet's VFE layers and 3D convolutions, or PointPillars' pillar encoding and 2D CNNs—the networks produce rich feature representations that encode geometric patterns and spatial relationships across the scene. These learned features serve as input to Region Proposal Networks (RPNs) that generate 3D bounding box proposals for object detection. The key contribution of these approaches lies in demonstrating that the entire pipeline—from raw point cloud processing to 3D object detection—can be trained end-to-end, allowing the networks to learn optimal feature representations specifically for the detection task rather than relying on hand-crafted features. We will explore the details of training object detection networks and designing appropriate loss functions in subsequent chapters on object detection and segmentation.

\section{Summary}
\label{sec:vision-summary}
In this chapter, we explored fundamental neural network architectures that form the backbone of modern robotic perception, which has shifted from hand-crafted features to end-to-end learning from data.
We began with Convolutional Neural Networks (CNNs), detailing their core components—convolutional layers, pooling, and fully-connected layers—that leverage spatial locality and translation equivariance for processing image data. We discussed landmark architectures like AlexNet, ResNet, and YOLO that demonstrated the power of deep, hierarchical feature learning.
We then introduced the Transformer architecture, whose self-attention mechanism captures long-range dependencies without built-in spatial biases. We covered its key elements, including tokenization, positional embeddings, and multi-head attention, and focused on its adaptation to vision through Vision Transformers (ViTs), which process images as sequences of patches and excel when pre-trained on large datasets.
Finally, we addressed the challenge of 3D sensor data by examining point-based networks like PointNet, which use permutation-invariant operations on point sets, and its hierarchical extension, PointNet++. We then discussed voxel-based methods, which discretize point clouds into regular grids to enable efficient 3D convolutions, as seen in VoxelNet, and pillar-based approaches like PointPillars that project data into a 2D representation for computational efficiency.

\paragraph{To learn more.}
For a deeper exploration of the topics covered in this chapter, several key resources are available.
A comprehensive foundation in deep learning concepts relevant to all architectures discussed can be found in \citet{GoodfellowBengioCourville2016}.
The seminal paper on the Transformer architecture is presented by \citet{VaswaniEtAl2017}, while its application to vision is detailed in \citet{DosovitskiyEtAl2021}.
For in-depth studies on 3D perception, the original papers on PointNet \citep{QiEtAl2017a} and VoxelNet \citep{ZhouTuzel2018} are essential reading.
Finally, for a broader perspective on computer vision algorithms that contextualize these learning-based approaches, we refer the reader to \citet{Szeliski2010}.

\section{Exercises}
The starter code for the exercises provided below is available online through GitHub. 
To get started, download the code by running in a terminal window:

\begin{tcolorbox}[colback=gray!10]
\begin{minted}{bash}
    git clone https://github.com/StanfordASL/pora-exercises.git
\end{minted}
\end{tcolorbox}

We denote Problems requiring hand-written solutions and coding in Python with \adjustbox{height=2ex, valign=c}{\includegraphics{figs/write.png}} and \adjustbox{height=2ex, valign=c}{\includegraphics{figs/code.png}}, respectively.

\subsection*{\adjustbox{height=2ex, valign=c}{\includegraphics{figs/code.png}}\ Problem 1: Convolutional Neural Network (CNN)}
In this exercise you will implement a basic convolutional neural network and use it to classify images from the CIFAR-10 dataset.
The CIFAR-10 dataset consists of a large number of small RGB images of objects belonging to ten different classes.
In the notebook \colorcode{ch09/exercises/cnn.ipynb}, complete the following:
\begin{enumerate}
\item Run the provided code to load the CIFAR-10 dataset.
Take a look at some of the sample images, what is the dimension of each image?
\item Complete the implementation of the \colorcode{SimpleCNN} class to define the model architecture.
Specifically, your model should have two convolution layers with ReLU activation and max pooling.
Use the provided values to define the parameters of each of the features, such as the convolution kernel size and number of output channels.
Following the convolution layers, your model should have two fully connected layers separated by a ReLU activation.
Use the provided value to define the dimension of the hidden layer, and you should be able to determine the appropriate size of the first fully connected layer input based on the last convolution layer output size.
Additionally, implement the remaining code in the training loop to train your model using the provided \colorcode{criterion} and \colorcode{optimizer}.
Run the provided code to train your model and evaluate the model's performance on a test dataset.
\item Run the provided code to display the confusion matrix from the test dataset results.
What is this showing you? 
Are there any surprising results or does this match your intuition?
\item How many parameters does your model have in total?
\end{enumerate}
\newpage
\printbibliography[segment=\therefsegment,heading=subbibliography,title={References}]
\chapter{Object Detection and Recognition}
\label{ch:object-detection}
\newrefsegment
For a robot to safely navigate and interact with the world around it, it needs visual understanding capabilities beyond simple image classification—not only identifying what objects are present in an image, but also determining where they are located. 
Consider an autonomous vehicle navigating a busy intersection that must detect multiple pedestrians, vehicles, and cyclists while simultaneously understanding which pixels belong to the drivable road surface versus sidewalks or building facades. 
A household robot organizing a cluttered kitchen must not only detect individual objects like cups and plates, but also understand their precise boundaries to enable careful grasping and placement. 
These scenarios require three visual understanding tasks essential for robotics. 
\textit{Object detection} identifies what objects are present and where they are located using bounding boxes. \textit{Semantic segmentation} classifies every pixel into scene categories like ``road'' or ``vegetation''. \textit{Instance segmentation} combines both capabilities, identifying individual object instances and their precise boundaries. 
Furthermore, many robotics applications require reasoning about full 3D structures, motivating the extension of detection and segmentation to 3D sensor data, using the point cloud and voxel processing architectures from the previous chapter.

In this chapter, we will explore methodological developments that tackle these robotics perception tasks. The progression from expensive two-stage detectors to real-time one-stage approaches addresses the need for low-latency decisions in dynamic environments, while efficient 3D processing methods address computational challenges of real-time LiDAR processing. We will demonstrate how the CNN, PointNet, and voxel-based architectures from the previous chapter can be extended to enable robust visual understanding for autonomous robotic systems. In Section~\ref{sec:ch12_2dobjectdetection}, we will cover the foundations of 2D object detection, including the evolution from two-stage to one-stage detectors. In Section~\ref{sec:ch12_3dobjectdetection}, we will discuss how to extend these detection paradigms to 3D sensor data. Finally, in Section~\ref{sec:ch12_segmentation}, we will explore semantic and instance segmentation methods for both 2D and 3D data.

\section{2D Object Detection Foundations}
\label{sec:ch12_2dobjectdetection}
Object detection extends beyond image classification by not only identifying what objects are present in an image, but also determining where they are located. This dual requirement—classification and localization—fundamentally shapes the architectural design of detection systems. Unlike classification networks that output a single prediction per image, detection networks must handle varying numbers of objects at different scales and positions, requiring specialized architectures that can efficiently process these challenges.

\begin{definition}[Object detection]
Given an input image $I$, object detection aims to identify all instances of objects from a predefined set of classes $\mathcal{C} = \{c_1, c_2, \ldots, c_K\}$ and localize each instance with a bounding box. Formally, the output is a set of detections $\mathcal{D} = \{(b_i, c_i, s_i)\}_{i=1}^N$ where $b_i = (x_i, y_i, w_i, h_i)$ represents the bounding box coordinates, $c_i \in \mathcal{C}$ is the predicted class, and $s_i \in [0,1]$ is the confidence score.
\end{definition}

The evolution of object detection architectures can be broadly categorized into two paradigms: two-stage detectors that separate object localization from classification, and one-stage detectors that perform both tasks simultaneously. Additionally, the choice between anchor-based and anchor-free methods represents a fundamental design decision that affects both training dynamics and inference efficiency. Understanding these foundational concepts provides the framework for extending detection principles to 3D scenarios and multi-modal sensor fusion, which we will explore in subsequent sections.

\subsubsection{Two-Stage Detection: R-CNN to Fast R-CNN}

The two-stage detection paradigm emerged as a natural approach to the object detection problem by decomposing it into two sequential sub-problems: first generating a set of object proposals that likely contain objects, then classifying these proposals while refining their locations. This divide-and-conquer strategy proved highly effective, establishing the foundation for many subsequent detection architectures.

\begin{figure}[ht]
\centering
\includegraphics[width=0.75\textwidth]{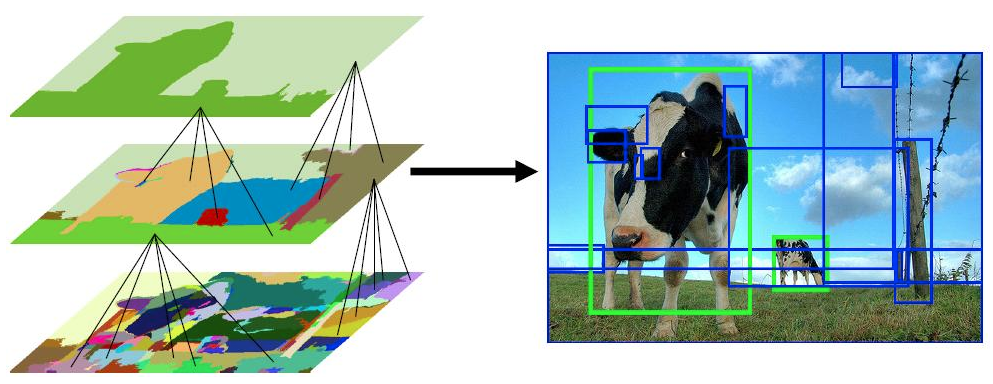}
\caption{Selective search hierarchical grouping process figure from Uijlings et al. (2013)\nocite{uijlings2013selective}. Starting from initial superpixel segmentation (left), regions are progressively merged based on color, texture, size, and shape compatibility, ultimately generating diverse object proposals of varying scales and locations (right).}
\label{fig:selective-search}
\end{figure}

\paragraph{R-CNN: establishing the two-stage paradigm.}
The original R-CNN (Regions with CNN features) architecture\sidenote{Proposed by Girshick et al. (2014), R-CNN demonstrated that CNN features could dramatically improve object detection performance when combined with traditional region proposal methods.} established the two-stage paradigm through a three-step process that combined classical computer vision techniques with modern deep learning. The architecture begins with selective search\sidenote{Selective search is a graph-based segmentation algorithm that generates object proposals by hierarchically grouping superpixels based on color, texture, size, and shape compatibility.}, which generates approximately 2,000 region proposals per image by identifying regions likely to contain objects based on low-level visual cues. The process is illustrated in \cref{fig:selective-search}. 

Each proposed region is then warped to a fixed size of $227 \times 227$ pixels and processed independently through a pre-trained CNN (originally AlexNet) to extract a 4096-dimensional feature vector. This feature extraction step leverages the powerful representations learned by CNNs on large-scale image classification datasets, transferring this knowledge to the detection task. Finally, these CNN features are fed to class-specific Support Vector Machine (SVM) classifiers for object classification and linear regressors for bounding box refinement.

Mathematically, for a region proposal $r$ with extracted CNN features $\phi(r)$, the classification score for class $c$ is computed as:
\begin{equation*}
    s_c(r) = w_c^T \phi(r) + b_c,
\end{equation*}
where $w_c$ and $b_c$ are the learned SVM parameters for class $c$. The bounding box regression predicts corrections $(\Delta x, \Delta y, \Delta w, \Delta h)$ to transform the proposal coordinates $(x, y, w, h)$ to better align with the ground truth:
\begin{equation*}
    \begin{aligned}
    \Delta x &= w_x^T \phi(r) + b_x, \\
    \Delta y &= w_y^T \phi(r) + b_y, \\
    \Delta w &= w_w^T \phi(r) + b_w, \\
    \Delta h &= w_h^T \phi(r) + b_h.
    \end{aligned}
\end{equation*}

While R-CNN achieved breakthrough detection performance on benchmark datasets, it suffered from significant computational inefficiencies and training complexity. Each of the ~2,000 proposals required a separate forward pass through the CNN, making both training and inference extremely slow—processing a single image could take minutes. The multi-stage training process required pre-training the CNN on ImageNet, training SVMs for classification, and training linear regressors for bounding box refinement, making the pipeline complex and difficult to optimize end-to-end.

\paragraph{Fast R-CNN: shared computation breakthrough.}
Fast R-CNN\sidenote{Introduced by Girshick (2015), Fast R-CNN addressed the computational bottlenecks of R-CNN while maintaining the two-stage paradigm and improving detection accuracy.} addressed these limitations through a key architectural innovation: shared computation across all proposals. Instead of processing each proposal independently through the CNN, Fast R-CNN processes the entire input image once through a convolutional backbone to generate a feature map, then extracts proposal-specific features from this shared representation.

\begin{figure}[ht]
\centering
\includegraphics[width=0.6\textwidth]{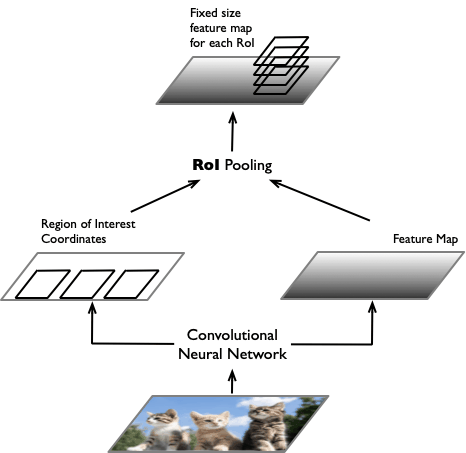}
\caption{RoI pooling mechanism. A region of interest proposal of arbitrary size (left) is used to apply max pooling within the feature map (right) to produce a fixed-size feature representation per region (top) for subsequent classification and regression layers.}
\label{fig:roi-pooling}
\end{figure}

The central innovation is the Region of Interest (RoI) pooling layer, which extracts fixed-size feature representations from variable-sized proposal regions on the shared feature map. For a proposal with coordinates $(x, y, w, h)$ on a feature map $F$ with spatial dimensions $H \times W$, RoI pooling divides the proposal region into a regular $k \times k$ grid (typically $7 \times 7$) and applies max pooling within each grid cell, as shown in \cref{fig:roi-pooling}.

\begin{equation*}
    \text{RoI}_{i,j} = \maximize[(x',y') \in \text{bin}_{i,j}] F_{x',y'},
\end{equation*}
where $\text{bin}_{i,j}$ represents the spatial extent of the $(i,j)$-th grid cell. This operation ensures that regardless of the input proposal size, the output is always a fixed $k \times k \times d$ feature tensor, where $d$ is the number of feature channels.

Fast R-CNN also unified the training process through a multi-task loss function that jointly optimizes classification and bounding box regression:
\begin{equation*}
    L = L_{\text{cls}}(p, u) + \lambda [u \geq 1] L_{\text{bbox}}(t^u, v),
\end{equation*}
where $L_{\text{cls}}(p, u) = -\log p_u$ is the log loss for true class $u$ with predicted class probabilities $p$, and $L_{\text{bbox}}$ is the smooth L1 loss for bounding box regression. The indicator function $[u \geq 1]$ ensures that bounding box loss is only computed for positive examples (background class has $u = 0$), and $\lambda$ balances the two loss terms.

This architectural change provided substantial improvements in both computational efficiency and detection accuracy. By sharing CNN computation across all proposals, Fast R-CNN reduced training time by an order of magnitude while achieving higher mean Average Precision (mAP) on standard benchmarks. The end-to-end training also eliminated the complex multi-stage optimization procedure, making the system more practical for real-world deployment.
However, Fast R-CNN still relied on external region proposal algorithms like selective search, which remained a computational bottleneck and prevented the entire detection pipeline from being truly end-to-end learnable. This limitation motivated the development of learnable region proposal methods, which we will explore in the next section.

\subsubsection{Learnable Proposals: RPN and Faster R-CNN}

While Fast R-CNN significantly improved computational efficiency through shared CNN computation, it still relies on external region proposal algorithms like selective search. These traditional methods suffered from several fundamental limitations: they were computationally expensive, requiring seconds per image; they were not learned from data and thus could not adapt to specific datasets or tasks; and they created a bottleneck that prevented the entire detection pipeline from being optimized end-to-end. The Region Proposal Network (RPN) innovation addressed these limitations by making region proposal generation a learnable component within the detection framework.

\paragraph{Limitations of selective search.}
Selective search and similar traditional proposal methods operate using hand-crafted features and heuristics that remain fixed regardless of the detection task or dataset. These algorithms typically generate thousands of proposals per image using expensive graph-based operations, with processing times often exceeding the CNN inference itself. More critically, since these methods are not learnable, they cannot benefit from the supervision available during detection training—they cannot learn which types of regions are most likely to contain objects for a specific application domain.

\paragraph{Region proposal network innovation.}
The Region Proposal Network (RPN) represents a paradigm shift by treating proposal generation as a learned prediction task. The RPN is essentially a fully convolutional network that slides a small network over the convolutional feature map produced by the backbone CNN. At each sliding window position, the RPN simultaneously predicts multiple region proposals using a set of reference boxes called anchors.

\paragraph{Anchor box design principles.}
Anchors serve as reference templates that cover different scales and aspect ratios at each spatial location in the feature map. For a feature map of size $H \times W$, the RPN generates $H \times W \times k$ potential proposals, where $k$ is the number of anchor templates per location. Common anchor designs use 3 scales (e.g., $128^2$, $256^2$, $512^2$ pixels) and 3 aspect ratios (e.g., 1:1, 1:2, 2:1), resulting in $k = 9$ anchors per location.

Mathematically, for an anchor centered at position $(x_a, y_a)$ with width $w_a$ and height $h_a$, the RPN predicts refinements $(\Delta x, \Delta y, \Delta w, \Delta h)$ to produce a final proposal:
\begin{equation*}
    \begin{aligned}
    x &= \Delta x \cdot w_a + x_a, \\
    y &= \Delta y \cdot h_a + y_a, \\
    w &= w_a \cdot \exp(\Delta w), \\
    h &= h_a \cdot \exp(\Delta h).
    \end{aligned}
\end{equation*}

The exponential transformation for width and height ensures positive values and provides scale-invariant parameterization.

\paragraph{Objectness scoring.}
Unlike traditional proposal methods that use complex heuristics, the RPN performs binary classification to determine ``objectness''—whether each anchor location contains an object of any class versus background. This objectness score $p^*$ is simpler than full multi-class classification but captures the essential information needed for proposal generation. The RPN learns to distinguish object-like regions from background using the same convolutional features that will later be used for detailed classification.

\paragraph{RPN loss function.}
The RPN is trained using a multi-task loss that combines objectness classification and bounding box regression:
\begin{equation*}
    L_{\text{RPN}} = \frac{1}{N_{\text{cls}}} \sum_i L_{\text{cls}}(p_i, p_i^*) + \lambda \frac{1}{N_{\text{box}}} \sum_i p_i^* L_{\text{box}}(t_i, t_i^*),
\end{equation*}
where $L_{\text{cls}}$ is the log loss for binary classification, $L_{\text{box}}$ is the smooth L1 loss for box regression, $N_{\text{cls}}$ and $N_{\text{box}}$ are normalization terms, and $\lambda$ balances the two losses. The box regression loss is only computed for positive anchors (those with $p_i^* = 1$), indicated by the multiplication with $p_i^*$.

During training, anchors are assigned positive labels if they have Intersection over Union (IoU) > 0.7 with any ground truth box, or if they are the highest IoU anchor for a ground truth box. Anchors with IoU < 0.3 are assigned negative labels, while those with intermediate IoU values are ignored to avoid ambiguous supervision.

\paragraph{Faster R-CNN: integration with Fast R-CNN.}
Faster R-CNN combines the RPN with Fast R-CNN into a single, unified network that shares convolutional features between proposal generation and detection. The architecture consists of a shared CNN backbone (e.g., VGG-16 or ResNet), followed by two sibling branches: the RPN for generating proposals and the Fast R-CNN detection head for classifying proposals and refining their locations.

The shared backbone is crucial for computational efficiency—rather than running separate CNNs for proposal generation and detection, both tasks operate on the same feature representation. This sharing also enables the network to learn features that are beneficial for both tasks simultaneously.

\paragraph{Training strategies.}
Training Faster R-CNN requires careful coordination between the RPN and detection components. Initial approaches alternated between training the RPN and the detection network. First, the RPN is trained using ImageNet-pretrained features. Then, the detection network is trained using proposals from the trained RPN, fine-tuning the shared convolutional layers. This process can be repeated, though diminishing returns are typically observed after the first iteration.

Today, most training of both components is done simultaneously, using a combined loss function:
\begin{equation*}
    L_{\text{total}} = L_{\text{RPN}} + L_{\text{Fast R-CNN}}.
\end{equation*}

Joint training is more efficient and often achieves better performance, as it allows the RPN and detection network to adapt to each other during learning.

\paragraph{Non-maximum suppression and post-processing.}
After the RPN generates proposals, Non-Maximum Suppression (NMS) removes redundant detections. The algorithm sorts proposals by objectness score and iteratively removes proposals that have high IoU (typically > 0.7) with higher-scored proposals. This reduces the number of proposals fed to the detection stage from thousands to hundreds, improving computational efficiency while maintaining detection quality.

The complete Faster R-CNN pipeline processes an image through the shared backbone, generates scored proposals via RPN with NMS post-processing, extracts RoI features for the top proposals, and produces final classifications and refined bounding boxes. This end-to-end learnable system achieved significant improvements in both speed and accuracy over previous two-stage methods, establishing the foundation for modern object detection architectures.

\subsubsection{One-Stage Detection: YOLO}

The Region Proposal Network represented a major breakthrough by making proposal generation learnable, but it still required a two-stage pipeline where proposals were generated first and then classified separately. This sequential approach, while effective, created computational bottlenecks that limited real-time performance in robotics applications. As autonomous vehicles, drones, and mobile robots demanded faster detection systems for dynamic environments, a fundamental question emerged: could object detection be reformulated to predict bounding boxes and classes directly from image features in a single forward pass?

You Only Look Once (YOLO)\sidenote{Introduced by Redmon et al. (2016), YOLO revolutionized object detection by demonstrating that competitive detection performance could be achieved through direct single-stage prediction, enabling real-time performance for robotics applications.} provided a radical answer to this question. Rather than decomposing detection into proposal generation followed by classification, YOLO treats object detection as a single regression problem, directly predicting bounding box coordinates and class probabilities from image pixels in one evaluation of the network. This paradigm shift eliminated the computational overhead of generating and processing thousands of proposals, enabling genuine real-time object detection suitable for robotics systems operating in dynamic environments.

\paragraph{Core YOLO innovation: grid-based direct detection.}
YOLO's central innovation lies in its spatial decomposition of the detection problem through a grid-based approach. The method divides the input image into an $S \times S$ grid (typically $7 \times 7$ for the original YOLO), where each grid cell becomes responsible for detecting objects whose center points fall within that cell's spatial region. This responsibility assignment creates a natural spatial organization that eliminates the need for separate proposal generation.

Each grid cell simultaneously predicts multiple bounding boxes along with their associated confidence scores and class probabilities. The key insight is that this grid-based spatial division provides sufficient spatial coverage while maintaining computational tractability—rather than evaluating thousands of potential object locations as in proposal-based methods, YOLO evaluates a fixed number of predictions per grid cell, resulting in a manageable total number of predictions regardless of scene complexity.
The elimination of the proposal generation stage represents more than just a computational optimization; it fundamentally changes how the network approaches object detection. Rather than learning to generate good proposals and then classify them, the network must learn to directly map from image features to final detection outputs. This end-to-end learning enables the network to optimize the entire detection pipeline jointly, potentially leading to better coordination between localization and classification components.

\paragraph{YOLO architecture and predictions.}
The YOLO architecture consists of a single CNN backbone followed by fully connected layers that produce the final detection tensor. The original implementation used a modified GoogLeNet architecture as the backbone, processing input images of size $448 \times 448$ pixels through convolutional layers that progressively reduce spatial resolution while increasing feature depth. The final convolutional features are flattened and processed through fully connected layers to produce a structured output tensor.

The network's output is a tensor of size $S \times S \times (B \times 5 + C)$, where $S$ is the grid size, $B$ is the number of bounding boxes predicted per cell, and $C$ is the number of object classes. For the original YOLO trained on PASCAL VOC, this results in a $7 \times 7 \times 30$ tensor, with $B = 2$ bounding boxes and $C = 20$ classes. Each bounding box prediction consists of five values: $(x, y, w, h, \text{confidence})$, where $(x, y)$ represents the box center relative to the grid cell boundaries, $(w, h)$ represents the box dimensions relative to the entire image, and confidence represents the model's certainty that the box contains an object.
The class predictions are formulated as conditional probabilities $P(\text{Class}_i | \text{Object})$, representing the probability of each class given that an object is present in the cell. This conditional formulation is crucial—each grid cell predicts only one set of class probabilities regardless of the number of bounding boxes, reflecting the assumption that each cell is responsible for at most one object class.

The final detection confidence for each bounding box is computed by multiplying the conditional class probabilities with the bounding box confidence scores:
\begin{equation*}
\text{Detection Confidence} = P(\text{Class}_i | \text{Object}) \times P(\text{Object}) \times \text{IoU}(\text{pred}, \text{truth}).
\end{equation*}
This formulation ensures that high detection scores require both confident object presence and accurate localization.

\paragraph{Loss function and training.}
YOLO's loss function addresses the multi-task nature of the detection problem by combining coordinate regression, confidence prediction, and classification into a unified objective. The loss function consists of several components with different weights to balance their relative importance:
\begin{equation*}
\begin{aligned}
L &= \underbrace{\lambda_{\text{coord}} \sum_{i=0}^{S^2} \sum_{j=0}^{B} \mathbb{1}_{ij}^{\text{obj}} [(x_i - \hat{x}_i)^2 + (y_i - \hat{y}_i)^2]}_{\text{bounding box center coordinates}} \\
&\quad + \underbrace{\lambda_{\text{coord}} \sum_{i=0}^{S^2} \sum_{j=0}^{B} \mathbb{1}_{ij}^{\text{obj}} [(\sqrt{w_i} - \sqrt{\hat{w}_i})^2 + (\sqrt{h_i} - \sqrt{\hat{h}_i})^2]}_{\text{bounding box dimensions}} \\
&\quad + \underbrace{\sum_{i=0}^{S^2} \sum_{j=0}^{B} \mathbb{1}_{ij}^{\text{obj}} (C_i - \hat{C}_i)^2}_{\text{confidence for cells with objects}} \\
&\quad + \underbrace{\lambda_{\text{no-obj}} \sum_{i=0}^{S^2} \sum_{j=0}^{B} \mathbb{1}_{ij}^{\text{no-obj}} (C_i - \hat{C}_i)^2}_{\text{confidence for cells without objects}} \\
&\quad + \underbrace{\sum_{i=0}^{S^2} \mathbb{1}_i^{\text{obj}} \sum_{c \in \text{classes}} (p_i(c) - \hat{p}_i(c))^2}_{\text{class probabilities}}.
\end{aligned}
\end{equation*}
The loss function uses different weights for different components: $\lambda_{\text{coord}} = 5$ increases the importance of coordinate predictions, while $\lambda_{\text{no-obj}} = 0.5$ decreases the weight of confidence predictions for cells without objects. The square root transformation for width and height helps the loss function treat errors in small and large boxes more equally, since a small absolute error in a small box represents a larger relative error than the same absolute error in a large box.

The indicator function $\mathbb{1}_{ij}^{\text{obj}}$ denotes whether cell $i$ contains an object and bounding box $j$ is responsible for that prediction (determined by which predicted box has the highest IoU with the ground truth). This responsibility assignment is crucial for training stability, as it ensures each ground truth object is associated with exactly one predicted bounding box.
Training YOLO follows a two-stage approach: the convolutional layers are first pre-trained on ImageNet for classification, then the entire network is fine-tuned on detection data. The classification pre-training provides the network with strong feature representations that are then adapted for the detection task. During detection training, the learning rate is carefully adjusted to balance the different loss components and ensure stable convergence.

\paragraph{Speed versus accuracy trade-offs and robotics impact.}
YOLO's architectural design prioritizes computational efficiency, achieving detection speeds that were unprecedented at the time of its introduction. The original YOLO processes images at 45 frames per second (FPS) on contemporary GPU hardware, while a faster variant (Fast YOLO) achieved 155 FPS by using a smaller network architecture. These speeds represent order-of-magnitude improvements over contemporary two-stage methods like Fast R-CNN, which operated at approximately 7 FPS.
However, this speed comes with accuracy trade-offs. YOLO's grid-based approach struggles with small objects, since multiple small objects within the same grid cell cannot be detected independently. The method also has difficulty with objects that appear in unusual aspect ratios, as the fixed number of bounding box predictors per cell limits the diversity of detectable shapes. Additionally, the coarse spatial quantization imposed by the grid structure can lead to less precise localization compared to methods that can place proposals at arbitrary locations.

For robotics applications, these trade-offs often represent acceptable compromises. Autonomous vehicles operating in real-time require detection systems that can process sensor data fast enough to support control decisions, even if absolute detection accuracy is somewhat reduced. The ``good enough'' detection philosophy embodied by YOLO aligns well with robotics applications where timely decisions often matter more than perfect perception.
The impact of YOLO on the robotics field extends beyond its specific technical contributions. By demonstrating that real-time object detection was achievable with modest computational resources, YOLO democratized object detection for resource-constrained robotics platforms. Mobile robots, drones, and embedded systems could now incorporate sophisticated visual understanding capabilities without requiring expensive computational hardware.

The evolution of YOLO through subsequent versions (YOLOv2, YOLOv3, YOLOv4, YOLOv5, and beyond) has addressed many of the original accuracy limitations while maintaining the core computational advantages. Modern YOLO variants incorporate multi-scale feature processing, improved loss functions, and architectural refinements that close much of the accuracy gap with two-stage methods while preserving real-time performance. This progression demonstrates the enduring value of the single-stage detection paradigm for robotics applications where speed and efficiency are paramount.

\subsubsection{Transformer-Based Object Detection}
\begin{figure}[ht] 
    \centering
    \includegraphics[width=1.0\textwidth]{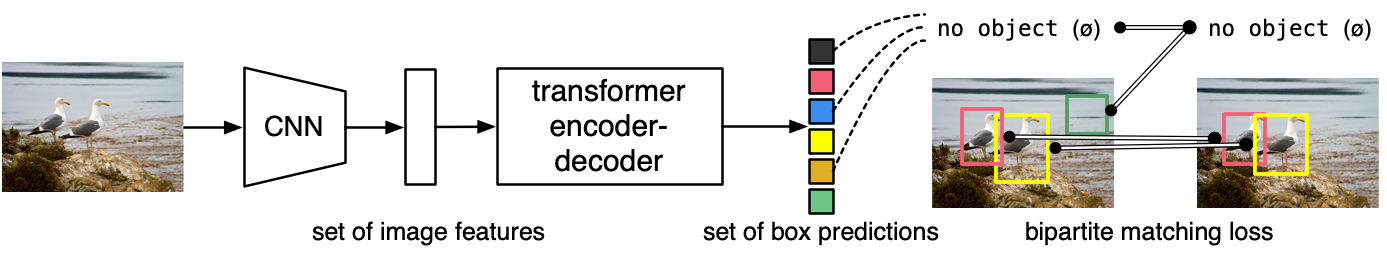}
    \caption{Detection Transformers (DETR) from Carion, Massa, et al. (2020)\nocite{CarionEtAl2020}}
    \label{fig:detr}
\end{figure}

Transformers have recently been adapted to tackle object detection, and their performance shows several benefits over CNN-based models. Detection Transformers (DETR)\cite{CarionEtAl2020}, illustrated in~\Cref{fig:detr}, propose to formulate the object detection problem as a direct set prediction problem, largely streamlining the detection pipeline through its end-to-end structure. In its most basic form, DETR is an end-to-end Transformer model that takes in images as inputs and predicts a fixed set of potential bounding boxes. DETR removes many hand-designed components, including ``region proposal'' and ``non-maximum suppression'' that are commonly used in CNN-based models.

\begin{figure}[ht] 
    \centering
    \includegraphics[width=1.0\textwidth]{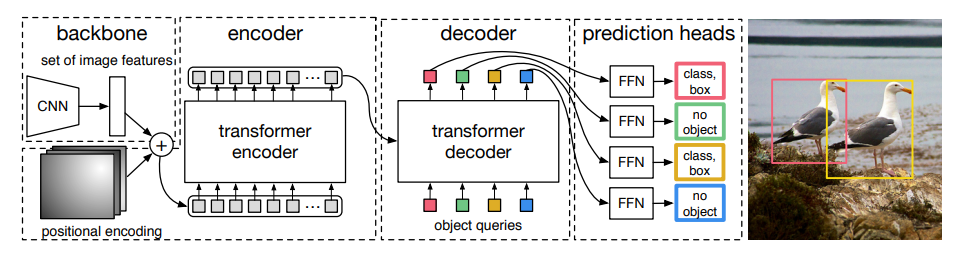}
    \caption{Detailed DETR architecture, from Carion, Massa, et al. (2020)\nocite{CarionEtAl2020}.}
    \label{fig:detr_details}
\end{figure}

Specifically, DETR uses a conventional CNN backbone to learn 2D feature maps from an input image, as shown on the left side of \cref{fig:detr_details}. The model then converts the 2D feature maps into a sequence of feature tokens, similarly to Vision Transformers. These tokenized features are further fed into a Transformer encoder, comprised of a stack of self-attention mechanisms and multi-layer perceptrons, for further feature encoding. The encoded sequence is processed by a Transformer decoder that relates the feature sequence with a set of ``learnable object queries''. These object queries encode the distribution of object information, including size, location, and category, over an image.

Note that Transformer decoders have some key differences from encoders. For example, they can use what we refer to as \emph{cross-attention layers} and \emph{masked attention layers}. We use cross-attention layers to allow a sequence to get contextual information from another sequence, unlike self-attention layers which gather contextual information from within a single sequence. In the context of DETR, this allows the decoder to relate the encoder's feature embeddings to the object queries, which are two different input sets.

Finally, each object query, after absorbing image features, is processed by shared fully connected layers to predict class labels, bounding box centers, and bounding box sizes. A ``no object'' label is assigned to queries without true objects detected, allowing the model to handle a variable number of objects in an image.
In contrast to CNN-based object detectors, Transformer-based object detectors do not have one-to-one matching between the prediction set and the ground-truth set. Therefore, a set-based loss is used to produce an optimal bipartite matching between predicted and ground-truth objects, followed by optimizing object-centric (bounding box) losses.

\section{3D Object Detection}
\label{sec:ch12_3dobjectdetection}
While 2D object detection provides valuable information about what objects are present and their approximate locations in images, many robotics applications require understanding the full 3D structure and pose of objects in the physical world. Consider a robotic arm performing pick-and-place operations—knowing that a cup appears in a specific region of an image is insufficient for grasping; the robot needs the cup's precise 3D location, orientation, and dimensions to plan a successful grasp trajectory. Similarly, autonomous vehicles must understand the 3D positions and velocities of surrounding cars, pedestrians, and obstacles to make safe navigation decisions in real-world coordinates rather than image pixels.
The transition from 2D to 3D detection introduces changes in problem formulation, data representation, and evaluation metrics while preserving many of the core architectural principles developed for 2D detection. Understanding these extensions provides the foundation for building robust 3D detection systems using the point cloud and voxel processing architectures from the previous chapter.

\subsubsection{Extending Object Detection to 3D}
\begin{figure}[ht]
\centering
\includegraphics[width=0.8\textwidth]{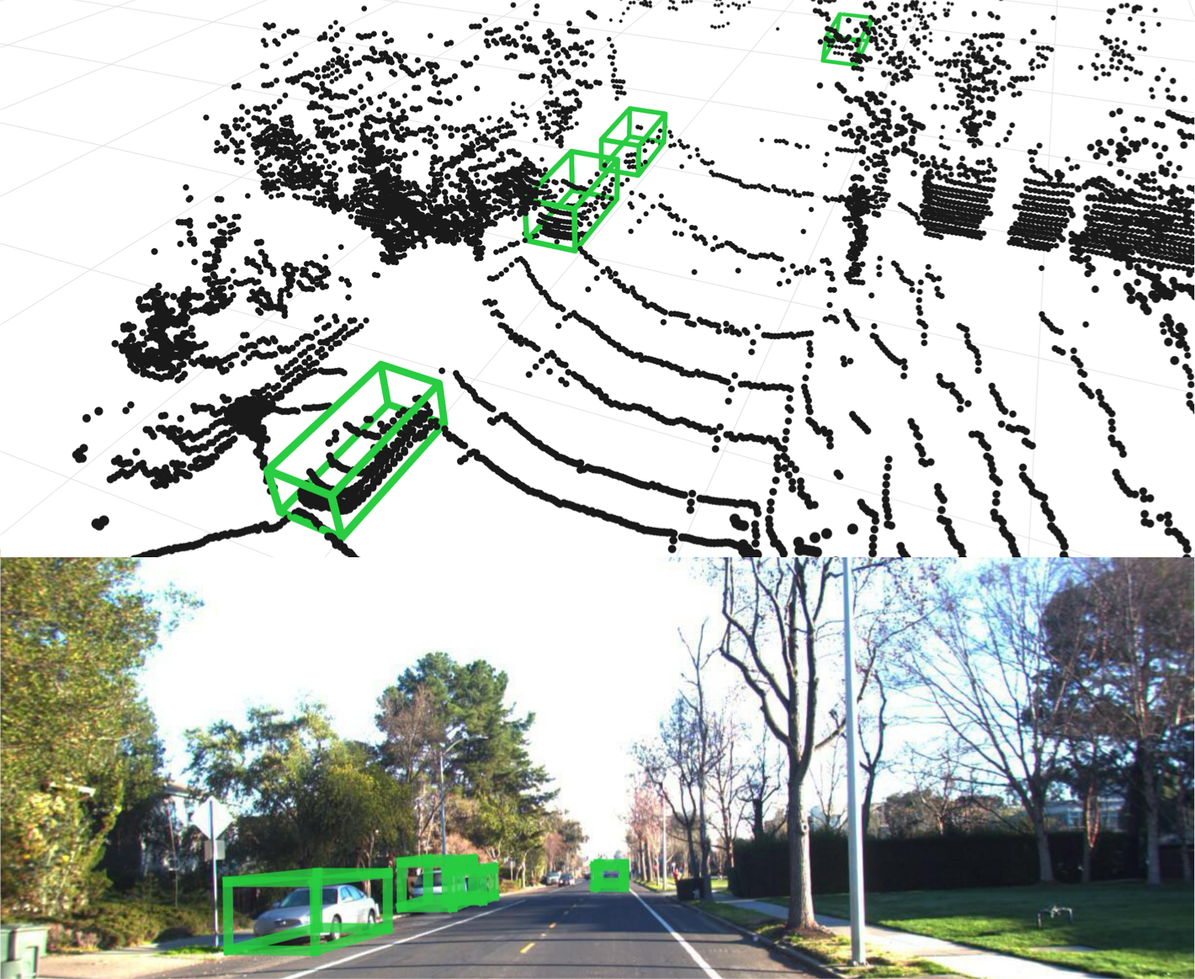}
\caption{3D object detection in a self-driving scene with 3D bounding boxes overlaid on LiDAR point cloud and camera inputs.}
\label{fig:object-detection}
\end{figure}

3D object detection extends the 2D formulation by instead predicting 3D bounding boxes to represent objects in three-dimensional space. While 2D detection outputs bounding boxes parameterized by $(x, y, w, h)$ in image coordinates, 3D detection requires additional parameters to specify the object's full pose and extent.
Examples of 3D object detection in autonomous driving scenarios are shown in \cref{fig:object-detection}.

\begin{definition}[3D object detection]
Given 3D sensor data (point cloud, voxel grid, or RGB-D), 3D object detection aims to identify all instances of objects from a predefined set of classes and localize each instance with a 3D bounding box. The output is a set of 3D detections $\mathcal{D}_{3D} = \{(b_i^{3D}, c_i, s_i)\}_{i=1}^N$ where $b_i^{3D} = (x, y, z, l, w, h, \theta)$ represents the 3D bounding box with center coordinates $(x, y, z)$, dimensions $(l, w, h)$ for length, width, and height, and orientation $\theta$.
\end{definition}

\paragraph{Extending anchor design to 3D space.}
For 3D anchor design, each anchor is parameterized by seven values: $(x_a, y_a, z_a, l_a, w_a, h_a, \theta_a)$ representing the center coordinates, dimensions, and orientation. Common 3D anchor designs use aspect ratios appropriate for the target object classes, and discrete orientation bins (e.g., 0°, 90°, 180°) to handle rotation invariance. 

The anchor refinement process follows similar principles to 2D detection, with the network predicting corrections $(\Delta x, \Delta y, \Delta z, \Delta l, \Delta w, \Delta h, \Delta \theta)$ to transform anchor parameters into final detections:

\begin{equation*}
\begin{aligned}
x &= \Delta x \cdot l_a + x_a, \\
y &= \Delta y \cdot w_a + y_a, \\
z &= \Delta z \cdot h_a + z_a, \\
l &= l_a \cdot \exp(\Delta l), \\
w &= w_a \cdot \exp(\Delta w), \\
h &= h_a \cdot \exp(\Delta h), \\
\theta &= \theta_a + \Delta \theta.
\end{aligned}
\end{equation*}

Similar to 2D setting, the exponential transformation ensures positive dimensions, while orientation is handled through additive corrections with appropriate normalization to handle angle wraparound.

\paragraph{Two-stage versus one-stage paradigms in 3D.}
The two-stage and one-stage detection paradigms from 2D systems transfer directly to 3D detection, with each approach offering distinct advantages for different 3D data modalities and applications.
\emph{Two-stage 3D detectors} follow the proposal-then-classification paradigm, first generating 3D object proposals from point clouds or voxel grids, then refining these proposals through dedicated classification and regression heads. This approach works particularly well with point-based representations, where the first stage can identify promising object centers using techniques like Hough voting, and the second stage can perform detailed classification using local point features.
\emph{One-stage 3D detectors} perform classification and localization simultaneously, making them better suited for real-time robotics applications where latency is critical. These methods work well with regular voxel or pillar representations that enable efficient convolutional processing across the entire 3D space.
The choice between paradigms often depends on the input data modality: point-based methods naturally lend themselves to two-stage approaches due to the irregular nature of point clouds, while voxel-based methods can efficiently implement one-stage detection using 3D CNNs.

\paragraph{Non-maximum suppression in 3D.}
Non-Maximum Suppression extends to 3D by replacing 2D IoU calculations with 3D IoU or Bird's Eye View (BEV) IoU metrics. 3D IoU computes the overlap between two 3D bounding boxes in full 3D space, accounting for differences in position, size, and orientation:
\begin{equation*}
\text{IoU}_{3D}(b_1, b_2) = \frac{\text{Volume}(b_1 \cap b_2)}{\text{Volume}(b_1 \cup b_2)}.
\end{equation*}
Computing 3D IoU requires determining the intersection volume between two oriented 3D boxes, which is more complex than the 2D case but essential for accurate duplicate removal.

For autonomous driving applications, BEV IoU is often preferred as it focuses on the ground plane where most objects interact:
\begin{equation*}
\text{IoU}_{BEV}(b_1, b_2) = \frac{\text{Area}(b_1^{BEV} \cap b_2^{BEV})}{\text{Area}(b_1^{BEV} \cup b_2^{BEV})},
\end{equation*}
where $b^{BEV}$ represents the projection of the 3D bounding box onto the ground plane. BEV IoU is computationally simpler and often more relevant for navigation tasks.

\paragraph{Evaluation metrics for 3D detection.}
3D object detection uses specialized metrics that account for spatial dimensions and orientation accuracy. The standard metric is 3D Average Precision (AP) computed using 3D IoU thresholds (typically 0.5 and 0.7). For autonomous driving, evaluation often focuses on Bird's Eye View (BEV) metrics that emphasize horizontal plane accuracy, with benchmarks like KITTI providing difficulty-based analysis. Orientation accuracy is measured through angular error between predicted and ground truth orientations, with some metrics requiring joint spatial and angular tolerance for correct detections.

\subsubsection{3D Detection from Point Clouds and Voxel Representations}

Building effective 3D object detection systems requires leveraging the specialized architectures for 3D data processing developed in the previous chapter. The choice between point-based and voxel-based representations fundamentally shapes the detection architecture, with each approach offering distinct advantages for different robotics applications. Point-based methods preserve the geometric precision of the original sensor data and handle irregular point distributions naturally, making them well-suited for applications requiring precise object localization. Voxel-based methods trade some geometric precision for computational efficiency by imposing regular grid structures that enable optimized convolutional operations, making them preferred for real-time robotics applications.

Rather than being mutually exclusive, these representations often complement each other within detection pipelines. Many successful 3D detection systems combine the efficiency of voxel processing for initial feature extraction with the precision of point-based refinement for final object localization. Understanding how different detection paradigms—two-stage, one-stage, and transformer-based—can be adapted to work with these 3D representations provides the foundation for building robust detection systems.

\paragraph{Leveraging 3D feature representations.}
The 3D detection architectures we will explore build directly upon the feature extraction capabilities of PointNet, VoxelNet, and PointPillars discussed in the previous chapter. These architectures provide learned feature representations that encode geometric patterns, spatial relationships, and semantic information from 3D sensor data. The key insight is that these features can serve as input to detection heads that predict object classifications and 3D bounding box parameters.

Point-based representations excel at preserving fine-grained geometric details and handling the irregular structure of sensor data. PointNet++ features capture multi-scale geometric patterns through hierarchical set abstraction, enabling detection of objects at different scales and levels of detail. These features are particularly valuable for detecting small objects or distinguishing between closely spaced instances where geometric precision is critical.
Voxel-based representations provide computational advantages through regular grid structures that enable efficient 3D convolutions and parallel processing. VoxelNet features encode local geometric patterns within voxels while maintaining spatial relationships across the scene. PointPillars features offer a hybrid approach, encoding vertical structure within pillars while enabling efficient 2D processing for large-scale scenes. The choice between these representations often depends on the computational constraints and accuracy requirements of the specific robotics application.

\subsubsection{Two-Stage 3D Detection: Extending Faster R-CNN}
\begin{figure}[ht]
\centering
\includegraphics[width=\textwidth]{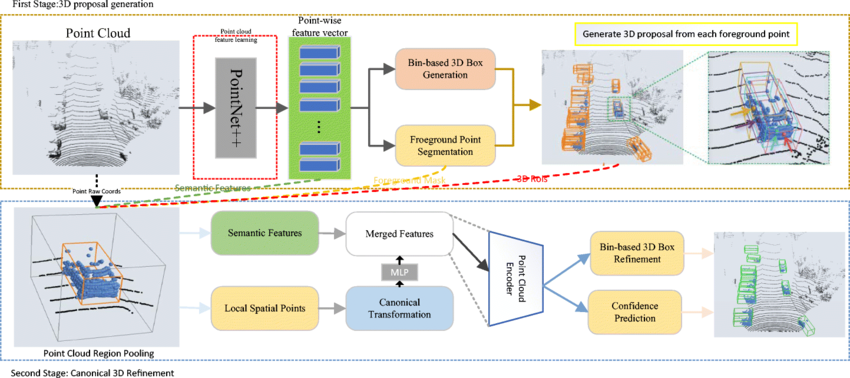}
\caption{PointRCNN architecture showing bottom-up 3D proposal generation from point-wise features, followed by canonical coordinate refinement and final detection heads from Shi et al. (2019)\nocite{ShiEtAl2019}.}
\label{fig:pointrcnn-architecture}
\end{figure}

The two-stage detection paradigm extends naturally to 3D by first generating object proposals in 3D space, then refining these proposals through dedicated classification and regression networks. This approach works particularly well when combined with the hierarchical feature representations from PointNet++ or the structured features from VoxelNet.
The PointRCNN architecture is illustrated in \cref{fig:pointrcnn-architecture}.

\paragraph{PointRCNN: point-based two-stage detection.}
PointRCNN demonstrates how the Faster R-CNN paradigm can be adapted to work directly with point cloud data using PointNet++ features. The architecture follows a bottom-up approach where object proposals are generated directly from point-wise features rather than through dense sliding window approaches used in image detection.

The first stage leverages PointNet++ hierarchical features to perform point-wise binary classification, identifying points that likely belong to foreground objects versus background. Rather than generating proposals at regular grid locations, PointRCNN generates 3D proposals centered at high-confidence foreground points. This approach is computationally efficient because it only considers a subset of points for proposal generation, and it preserves the geometric precision of the original point cloud.
For each proposal, the second stage extracts local point features within the proposed 3D region and applies canonical coordinate transformation to normalize the local geometry. This transformation aligns the object coordinate system with a canonical orientation, making the subsequent classification and regression tasks more robust to object orientation variations. The canonical transformation is particularly important for 3D detection because objects can appear in arbitrary orientations in the sensor coordinate system.

The mathematical formulation for the canonical transformation involves rotating and translating the local point coordinates so that the object's principal axes align with canonical directions:
\begin{equation*}
p_{\text{canonical}} = R^{-1}(p_{\text{local}} - t),
\end{equation*}
where $R$ is the estimated object rotation and $t$ is the estimated object center. This transformation enables the network to learn object-centric features that are invariant to the object's pose in the world coordinate system.

\paragraph{VoxelNet: voxel-based two-stage detection.}
VoxelNet adapts the two-stage paradigm to work with voxel-based representations by integrating Voxel Feature Encoding with Region Proposal Network concepts. The architecture processes point clouds through VFE layers to generate voxel-wise features, then applies 3D convolutional layers to build hierarchical representations across the voxelized space.
The proposal generation stage adapts the RPN concept to 3D by sliding 3D anchor templates across the feature volume. At each spatial location in the 3D feature map, the network predicts objectness scores and 3D bounding box refinements for multiple anchor templates covering different object sizes and orientations. The 3D RPN loss combines objectness classification with 3D bounding box regression:
\begin{equation*}
L_{\text{3D-RPN}} = \frac{1}{N_{\text{cls}}} \sum_i L_{\text{cls}}(p_i, p_i^*) + \lambda \frac{1}{N_{\text{box}}} \sum_i p_i^* L_{\text{3D-box}}(b_i, b_i^*),
\end{equation*}
where $L_{\text{3D-box}}$ incorporates losses for all seven parameters of the 3D bounding box: center coordinates, dimensions, and orientation.

The second stage performs 3D RoI pooling to extract fixed-size features for each proposal, followed by classification and bounding box refinement. The 3D RoI pooling operation extends the 2D concept by pooling features from 3D regions of the feature volume, maintaining spatial relationships in all three dimensions.

\subsubsection{One-Stage 3D Detection: Direct Prediction}
One-stage 3D detection methods perform object classification and localization simultaneously, eliminating the separate proposal generation stage. These approaches are particularly well-suited for real-time robotics applications where detection latency must be minimized.

\paragraph{CenterPoint: treating 3D objects as points.}
CenterPoint represents objects as points in Bird's Eye View (BEV) space and performs detection through keypoint estimation, similar to 2D anchor-free methods like CenterNet. The approach builds on PointPillars pillar-based representation to efficiently process large-scale point clouds while maintaining real-time performance.

The key insight is that 3D objects can be effectively represented by their center points when projected into BEV space, particularly for autonomous driving scenarios where objects primarily move on the ground plane. CenterPoint predicts a heatmap in BEV coordinates where peaks correspond to object centers, along with regression maps that predict 3D bounding box parameters for each detected center.
The detection pipeline processes point clouds through PointPillars to generate BEV feature maps, then applies 2D convolutional networks to predict center heatmaps and regression targets:
\begin{equation*}
\begin{aligned}
\mathbf{Y}_{\text{heatmap}} &= \sigma(\text{Conv}_{2D}(\mathbf{F}_{\text{BEV}})), \\
\mathbf{Y}_{\text{regression}} &= \text{Conv}_{2D}(\mathbf{F}_{\text{BEV}}),
\end{aligned}
\end{equation*}
where $\mathbf{F}_{\text{BEV}}$ represents the BEV feature map from PointPillars processing, and $\sigma$ is the sigmoid activation for heatmap prediction. The regression targets include 3D center offsets, object dimensions, and orientation angles.

The loss function combines center point detection with regression objectives:
\begin{equation*}
L_{\text{CenterPoint}} = L_{\text{heatmap}} + \lambda_{\text{reg}} L_{\text{regression}},
\end{equation*}
where $L_{\text{heatmap}}$ uses focal loss to handle the extreme imbalance between center points and background, and $L_{\text{regression}}$ uses smooth L1 loss for the continuous regression targets.
CenterPoint extends beyond basic detection by incorporating velocity estimation for tracking applications. By processing consecutive frames, the network can predict object velocities directly as part of the regression targets, enabling seamless integration with multi-object tracking systems essential for autonomous navigation.

\subsubsection{Transformer-Based 3D Detection}
Transformer architectures have been successfully adapted to 3D detection by treating object detection as a set prediction problem, eliminating the need for hand-designed anchors and complex post-processing steps like non-maximum suppression.


\paragraph{3DETR: set-to-set prediction in 3D.}
3DETR extends the DETR paradigm to 3D object detection by using transformer architectures to directly predict sets of 3D bounding boxes from point cloud or voxel features. The approach uses learnable object queries that attend to 3D scene features through cross-attention mechanisms, enabling end-to-end learning from raw 3D data to final detections.
The architecture processes 3D input data through feature extraction networks (PointNet++ for point clouds or 3D CNNs for voxel grids) to generate scene feature representations. These features are then processed by a transformer encoder to build contextual representations that capture long-range dependencies across the 3D scene. The transformer decoder uses a fixed set of learnable object queries to attend to the encoded scene features and predict object detections.

Each object query learns to specialize in detecting objects with particular characteristics or in specific spatial regions. The cross-attention mechanism allows queries to gather relevant information from across the entire scene, enabling detection of partially occluded objects or objects that extend across multiple local regions. The self-attention within the decoder enables queries to coordinate with each other, reducing duplicate detections without explicit post-processing.
The final prediction heads convert each object query's representation into 3D bounding box parameters and class predictions:
\begin{equation*}
\begin{aligned}
\mathbf{b}_i &= \text{MLP}_{\text{box}}(\mathbf{q}_i), \\
\mathbf{c}_i &= \text{MLP}_{\text{class}}(\mathbf{q}_i),
\end{aligned}
\end{equation*}
where $\mathbf{q}_i$ is the $i$-th object query after transformer processing. The training uses Hungarian matching to establish optimal assignment between predicted and ground truth objects, followed by standard detection losses.

The set-based prediction eliminates the need for anchor design, anchor assignment strategies, and non-maximum suppression, simplifying the detection pipeline while achieving competitive performance. This approach is particularly attractive for complex 3D scenes where traditional anchor-based methods struggle with the high-dimensional anchor space and complex object interactions.

\section{Semantic and Instance Segmentation}
\label{sec:ch12_segmentation}
\begin{figure}[ht]
\centering
\includegraphics[width=\textwidth]{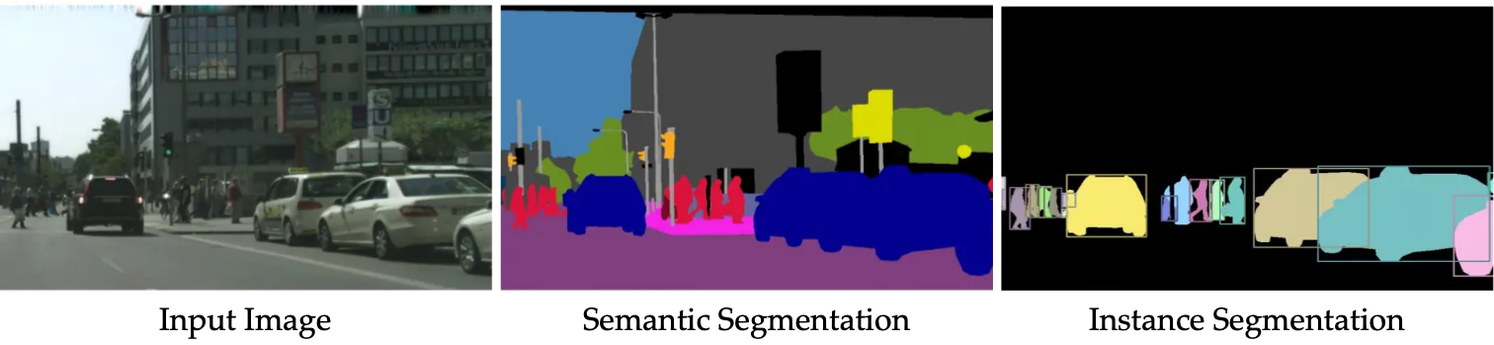}
\caption{An input image and the corresponding semantic and instance segmentation outputs.}
\label{fig:segmentation}
\end{figure}

Segmentation extends object detection by providing pixel-level or point-level understanding of scenes, enabling robots to understand not just where objects are located, but precisely which pixels belong to each object or scene category. While object detection provides coarse spatial understanding through bounding boxes, segmentation offers fine-grained spatial reasoning essential for tasks like autonomous navigation on complex terrain, precise robotic manipulation, and detailed scene understanding.
\cref{fig:segmentation} illustrates an example of semantic and instance segmentation outputs.

\subsubsection{Semantic Segmentation}
Semantic segmentation extends image understanding beyond object detection by classifying every pixel in an image into predefined semantic categories, providing dense spatial understanding of the scene. Unlike object detection which outputs sparse bounding boxes, semantic segmentation produces pixel-level predictions that preserve the precise boundaries and spatial extent of different scene elements. This fine-grained understanding is essential for robotics applications where precise spatial reasoning is required.

\paragraph{Problem definition and robotics applications.}
Semantic segmentation performs pixel-level classification without distinguishing between different instances of the same class. For an input image $I$ of size $H \times W$, the output is a segmentation map $S$ of the same spatial dimensions, where each pixel $(i,j)$ is assigned a class label $S_{i,j} \in \mathcal{C}$ from the predefined set of semantic categories.
For robotics systems, autonomous vehicles use segmentation to identify drivable road surfaces, distinguish between different types of terrain, and understand scene layout for path planning. Mobile robots navigating indoor environments use segmentation to identify floors, walls, furniture, and obstacles, enabling more sophisticated spatial reasoning for navigation planning.
In each case, the pixel-level precision enables robots to make more informed decisions about how to interact with their environment.

\paragraph{Fully convolutional networks (FCNs).}
Semantic segmentation can be viewed as dense classification where standard CNN architectures are adapted to produce spatial output maps rather than single classification scores. Fully Convolutional Networks (FCNs) build on the convolutional feature extraction capabilities of CNNs while replacing the fully connected classification layers with convolutional layers that preserve spatial structure.
These models replace the fully connected layers typically used for classification with convolutional layers that can accept images of arbitrary size and produce correspondingly sized output maps. For a CNN backbone that produces feature maps of size $H/32 \times W/32$ (due to pooling operations), FCN applies $1 \times 1$ convolutions to produce class score maps, then upsamples these maps back to the original image resolution.
The upsampling process uses transposed convolutions (also called deconvolutions) to increase spatial resolution:
\begin{equation*}
y_{i,j} = \sum_{m,n} x_{\lfloor i/s \rfloor + m, \lfloor j/s \rfloor + n} \cdot w_{m,n},
\end{equation*}
where $s$ is the upsampling stride and $w$ represents the learned transposed convolution weights. This operation is the mathematical inverse of convolution with stride $s$, enabling learnable upsampling that can recover spatial details.

FCN introduces skip connections that combine features from different layers of the encoder to recover fine-grained spatial information lost during downsampling. These connections add feature maps from earlier layers (with higher spatial resolution) to upsampled feature maps from deeper layers (with richer semantic information):
\begin{equation*}
F_{\text{fused}} = \text{Upsample}(F_{\text{deep}}) + F_{\text{shallow}}.
\end{equation*}
This fusion enables the network to combine high-level semantic understanding with low-level spatial precision, crucial for accurate boundary delineation in robotics applications.

\begin{figure}[ht]
\centering
\includegraphics[width=0.6\textwidth]{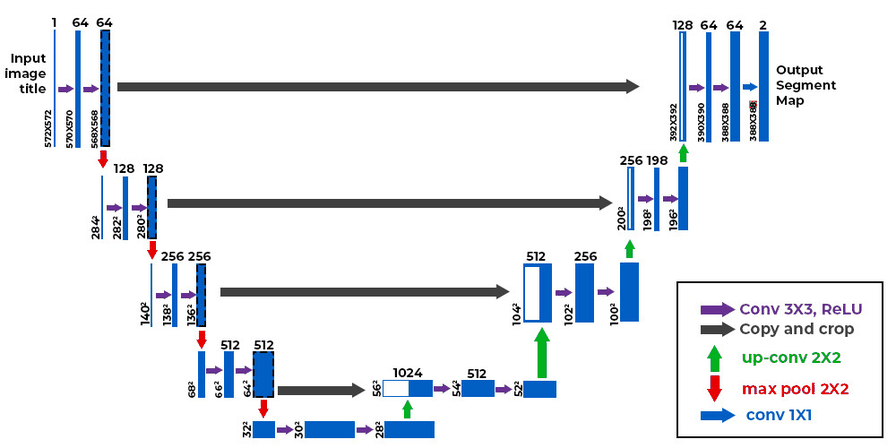}
\caption{U-Net architecture showing the symmetric encoder-decoder design with skip connections at multiple scales, enabling combination of high-resolution spatial information with high-level semantic features from Ronneberger et al. (2015)\nocite{RonnebergerFischerBrox2015}.}
\label{fig:unet-architecture}
\end{figure}

\paragraph{U-Net and encoder-decoder architectures.}
U-Net represents a systematic approach to encoder-decoder architectures that has become foundational for semantic segmentation across many domains.
The U-Net architecture, illustrated in \cref{fig:unet-architecture}, consists of a contracting path (encoder) that progressively reduces spatial resolution while increasing feature depth, followed by an expansive path (decoder) that gradually recovers spatial resolution while combining features across scales. For each decoder layer, skip connections concatenate features from corresponding encoder layer:
\begin{equation*}
F_{\text{decoder}}^{(i)} = \text{UpConv}(F_{\text{decoder}}^{(i-1)}) \oplus F_{\text{encoder}}^{(i)},
\end{equation*}
where $\oplus$ denotes concatenation and $\text{UpConv}$ represents upsampling convolution operations. These skip connections preserve fine-grained spatial details that would otherwise be lost during the encoding process.

The symmetric design ensures that the decoder has access to features at multiple scales, enabling accurate segmentation of both large objects (captured by deep, low-resolution features) and fine details (preserved through skip connections from high-resolution features). This multi-scale feature combination is particularly important for robotics applications where accurate boundary detection affects safety and task performance.

\paragraph{Training loss: classification cross-entropy per pixel.}
Semantic segmentation networks are trained using pixel-wise classification loss, treating each pixel as an independent classification problem. The standard loss function is cross-entropy computed across all pixels:
\begin{equation*}
L_{\text{seg}} = -\frac{1}{HW} \sum_{i=1}^{H} \sum_{j=1}^{W} \sum_{c=1}^{C} y_{i,j,c} \log(\hat{y}_{i,j,c}),
\end{equation*}
where $y_{i,j,c}$ is the ground truth one-hot encoding for pixel $(i,j)$ and class $c$, and $\hat{y}_{i,j,c}$ is the predicted probability. This formulation treats each pixel independently, enabling efficient batch processing and straightforward optimization.

However, pixel-wise cross-entropy can struggle with class imbalance, which is common in robotics scenarios where background pixels often dominate the scene. Various modifications address this challenge, including weighted cross-entropy that assigns different weights to different classes based on their frequency, and focal loss that emphasizes hard examples by down-weighting well-classified pixels.

\subsubsection{Instance Segmentation}

Instance segmentation combines object detection and semantic segmentation by identifying individual object instances and their precise pixel-level boundaries. Unlike semantic segmentation which treats all objects of the same class identically, instance segmentation distinguishes between separate instances—for example, identifying three individual cars rather than just ``car pixels''. This capability is critical for robotics applications where understanding individual objects enables targeted interaction and manipulation.

\paragraph{Problem definition and distinction from semantic segmentation.}
Instance segmentation extends semantic segmentation by assigning unique instance identifiers to pixels belonging to distinct objects. For an input image, the output includes both semantic labels and instance masks, where each instance mask $M_k$ defines the pixel-level extent of the $k$-th detected object instance.
The key distinction is that semantic segmentation answers ``what is this pixel?'' while instance segmentation answers ``what is this pixel and which specific object does it belong to?'' For a robotic arm grasping objects from a bin, semantic segmentation might identify all pixels as ``tool,'' but instance segmentation identifies individual wrenches, screwdrivers, and hammers, enabling the robot to select and grasp specific items.

\begin{figure}
\centering
\includegraphics[width=0.8\textwidth]{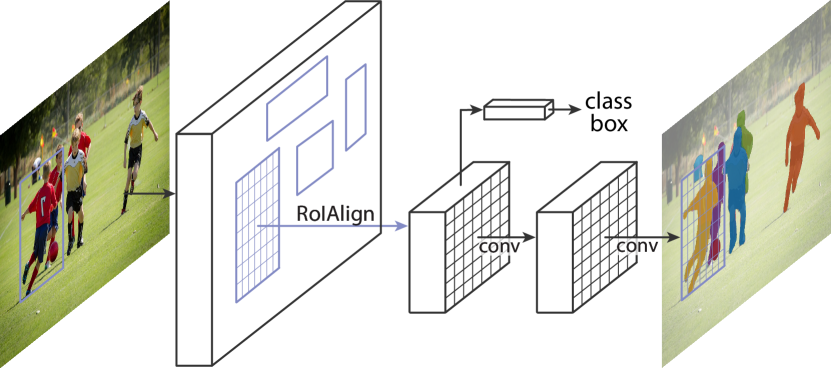}
\caption{Mask R-CNN architecture showing the addition of a mask prediction branch to Faster R-CNN, with RoI Align replacing RoI pooling for improved spatial alignment from He et al. (2017)\nocite{HeEtAl2017}.}
\label{fig:mask-rcnn-architecture}
\end{figure}

\paragraph{Mask R-CNN: extending detection with segmentation.}
Mask R-CNN, illustrated in \cref{fig:mask-rcnn-architecture}, extends Faster R-CNN by adding a segmentation branch that predicts pixel-level masks for each detected object. The architecture maintains the two-stage paradigm: the RPN generates object proposals, and the detection head performs classification, bounding box regression, and mask prediction.
The key innovation is RoI Align, which replaces RoI pooling to address spatial misalignment issues. While RoI pooling quantizes proposal coordinates to discrete feature map positions, RoI Align uses bilinear interpolation to sample features at exact locations:
\begin{equation*}
F_{\text{aligned}}(x, y) = \sum_{i,j} I(i, j) \cdot \max(0, 1-|x-i|) \cdot \max(0, 1-|y-j|).
\end{equation*}
This precise alignment is essential for accurate mask prediction, as small spatial misalignments can significantly degrade segmentation quality.

The mask prediction branch applies a small FCN to each RoI-aligned feature to produce a binary mask for the predicted object class. The mask loss is computed only for the predicted class to avoid competition between classes:
\begin{equation*}
L_{\text{mask}} = -\frac{1}{m^2} \sum_{i,j} [y_{i,j} \log(\hat{y}_{i,j}^{k^*}) + (1-y_{i,j}) \log(1-\hat{y}_{i,j}^{k^*})],
\end{equation*}
where $k^*$ is the predicted class, $y_{i,j}$ is the ground truth mask, and $\hat{y}_{i,j}^{k^*}$ is the predicted mask for class $k^*$.

\paragraph{Panoptic segmentation.}
Panoptic segmentation unifies semantic and instance segmentation by providing complete scene understanding. The task divides semantic categories into "things" (countable objects like cars, people) and "stuff" (amorphous regions like sky, road), performing instance segmentation for things and semantic segmentation for stuff.
For robotics applications, panoptic segmentation provides comprehensive scene understanding. An autonomous vehicle can simultaneously understand the road surface (stuff), individual vehicles and pedestrians (thing instances), and background elements like buildings and vegetation (stuff), enabling holistic reasoning about the driving environment.

\paragraph{Bottom-up approaches.}
Bottom-up instance segmentation methods first perform pixel-level feature learning, then group pixels into instances based on learned embeddings. These approaches contrast with top-down methods like Mask R-CNN that first detect objects then segment them.
Associative embedding learns pixel-level features where pixels belonging to the same instance have similar embedding vectors, while pixels from different instances have dissimilar embeddings. Instance masks are then generated by clustering pixels in the embedding space:
\begin{equation*}
d(e_i, e_j) = ||e_i - e_j||_2,
\end{equation*}
where $e_i$ and $e_j$ are embedding vectors for pixels $i$ and $j$. Pixels with distances below a threshold are grouped into the same instance.

These methods can handle arbitrary numbers of instances without predefined proposals but require robust clustering algorithms to separate instances reliably. They are particularly useful for robotics scenarios with dense object arrangements where proposal-based methods might struggle.

\subsubsection{3D Segmentation}

3D segmentation extends pixel-level understanding to volumetric data, providing precise spatial reasoning for robotics applications that require detailed 3D scene understanding. While 2D segmentation enables robots to understand image content, 3D segmentation allows reasoning about the full spatial extent and structure of objects in the physical world. This capability is essential for manipulation tasks requiring grasp planning, navigation in complex 3D environments, and understanding object affordances based on geometric structure.

\paragraph{Point cloud segmentation.}
Point cloud segmentation assigns semantic labels or instance identifiers to individual points in 3D space. The formulation extends 2D segmentation concepts to irregular point data, where each point $\mathbf{p}_i = (x_i, y_i, z_i)$ receives a label $l_i \in \mathcal{C}$ for semantic segmentation or instance identifier $I_i$ for instance segmentation.
Semantic segmentation of point clouds using PointNet++ leverages the hierarchical set abstraction layers from the previous chapter. The architecture processes points through multiple scales of local feature extraction and aggregation, then applies classification heads to predict semantic labels for each point:
\begin{equation*}
l_i = \text{MLP}_{\text{seg}}(\mathbf{f}_i^{(L)}),
\end{equation*}
where $\mathbf{f}_i^{(L)}$ represents the final point-wise feature after $L$ layers of hierarchical processing. The multi-scale feature extraction enables accurate segmentation of objects at different sizes and levels of detail.

\emph{Instance segmentation} in point clouds requires additional mechanisms to group points into distinct object instances. Methods like PointGroup combine semantic segmentation with learned offset vectors that point toward instance centers, enabling clustering of points belonging to the same object. The training loss combines semantic classification with offset regression:
\begin{equation*}
L_{\text{point-instance}} = L_{\text{semantic}} + \lambda L_{\text{offset}} + \gamma L_{\text{clustering}},
\end{equation*}
where $L_{\text{offset}}$ encourages points to predict vectors pointing toward their instance centers, and $L_{\text{clustering}}$ promotes tight clustering within instances and separation between instances.

\begin{figure}[ht]
\centering
\includegraphics[width=0.85\textwidth]{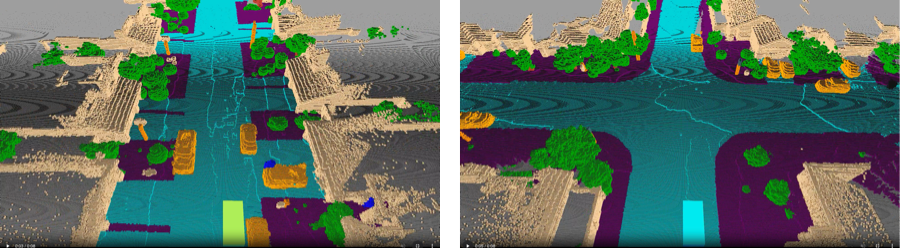}
\caption{3D voxel-based segmentation example from the Occ3D dataset \nocite{tian2023occ3d} for self-driving scenes, specifying what object category occupies each 3D location.}
\label{fig:point-segmentation}
\end{figure}

\paragraph{Voxel-based segmentation.}
Voxel-based segmentation processes regular 3D grids where each voxel represents a volumetric unit in 3D space. The formulation treats segmentation as 3D dense classification, where each voxel $v_{i,j,k}$ receives a semantic label or occupancy prediction.
Occupancy grids represent a fundamental approach where each voxel indicates whether that region of space is occupied by an object. This binary classification provides essential information for navigation and collision avoidance:
\begin{equation*}
o_{i,j,k} = \sigma(\text{MLP}(\mathbf{f}_{i,j,k})),
\end{equation*}
where $o_{i,j,k} \in [0,1]$ represents the occupancy probability for voxel $(i,j,k)$.

3D U-Net architectures extend the encoder-decoder paradigm to volumetric data for detailed semantic segmentation. The architecture applies 3D convolutions throughout the encoding and decoding paths, with 3D skip connections preserving spatial details:
\begin{equation*}
\mathbf{F}_{\text{decoder}}^{(i)} = \text{UpConv3D}(\mathbf{F}_{\text{decoder}}^{(i-1)}) \oplus \mathbf{F}_{\text{encoder}}^{(i)}.
\end{equation*}
The 3D convolutions capture volumetric patterns and spatial relationships essential for accurate 3D segmentation, while skip connections ensure fine-grained geometric details are preserved in the final predictions.
An example of 3D voxel-based segmentation is shown in \cref{fig:point-segmentation}.

\subsubsection{Robotics-Specific Applications}
Segmentation enables several critical robotics capabilities that require detailed geometric understanding of objects and environments. We explore a few examples below.

\begin{example}
    Grasp point prediction through part segmentation identifies functional regions of objects that are suitable for robotic grasping. By segmenting objects into semantic parts (handles, graspable surfaces, fragile regions), robots can plan grasps that are both mechanically sound and functionally appropriate. For example, segmenting a mug into handle, rim, and body regions enables the robot to choose appropriate grasp locations based on the intended manipulation task.
\end{example}

\begin{example}
    Terrain traversability analysis uses 3D segmentation to classify different terrain types and their suitability for robot navigation. Outdoor mobile robots use segmentation to distinguish between solid ground, obstacles, vegetation, and hazardous terrain, enabling safe path planning in complex outdoor environments. The 3D understanding allows reasoning about terrain slope, roughness, and stability that would be impossible with 2D analysis alone.
\end{example}

\begin{example}
    Object affordance understanding through part-based analysis enables robots to reason about how objects can be used based on their geometric structure. By segmenting objects into functional parts and understanding the spatial relationships between parts, robots can infer possible interactions and manipulation strategies. A segmented chair with identified seat, backrest, and legs enables the robot to understand both the object's function and how to manipulate it safely.
\end{example}

These applications demonstrate how 3D segmentation provides the detailed spatial understanding necessary for robots to interact effectively with complex 3D environments, going beyond simple object detection to enable sophisticated reasoning about object structure, function, and manipulation possibilities.

\section{Summary}
\label{sec:obj-detection-summary}
In this chapter, we explored the essential robotic perception tasks of object detection and segmentation, which provide the spatial understanding necessary for robots to interact intelligently with their environments.

We began with the foundations of 2D object detection, contrasting the two-stage paradigm—exemplified by the evolution from R-CNN to the efficient, learnable proposals of Faster R-CNN—with the one-stage paradigm of YOLO, which prioritizes speed for real-time applications. We also discussed how the Transformer architecture has been adapted for detection with models like DETR, which simplify the pipeline by framing detection as a direct set prediction problem.
We then extended these concepts to 3D, detailing how detection paradigms adapt to point cloud and voxel data. We covered two-stage point-based methods like PointRCNN, one-stage voxel-based methods like CenterPoint, and Transformer-based approaches like 3DETR, each offering different trade-offs between precision and computational efficiency for processing 3D sensor data.

Finally, we delved into segmentation, which provides pixel- and point-level understanding. We covered semantic segmentation with architectures like FCN and U-Net for categorizing every pixel, instance segmentation with Mask R-CNN for identifying individual objects, and their unification in panoptic segmentation. We further extended these ideas to 3D point cloud and voxel-based segmentation, highlighting their critical role in applications requiring detailed geometric reasoning, such as grasp point prediction and terrain analysis.

\paragraph{To learn more.}
For a deeper exploration of the topics covered in this chapter, several key resources are available.
The seminal papers on two-stage detection \citep{Girshick2015} and one-stage detection \citep{RedmonEtAl2016} are foundational to the field.
For Transformer-based detection, the DETR paper \citep{CarionEtAl2020} introduced the set prediction paradigm.
In 3D perception, the original papers on PointRCNN \citep{ShiEtAl2019} and VoxelNet \citep{ZhouTuzel2018} are essential reading for point-based and voxel-based detection, respectively.
For segmentation, the works on Mask R-CNN \citep{HeEtAl2017} and U-Net \citep{RonnebergerFischerBrox2015} provide the basis for modern instance and semantic segmentation techniques.

\section{Exercises}
The starter code for the exercises provided below is available online through GitHub. 
To get started, download the code by running in a terminal window:

\begin{tcolorbox}[colback=gray!10]
\begin{minted}{bash}
    git clone https://github.com/StanfordASL/pora-exercises.git
\end{minted}
\end{tcolorbox}

We denote Problems requiring hand-written solutions and coding in Python with \adjustbox{height=2ex, valign=c}{\includegraphics{figs/write.png}} and \adjustbox{height=2ex, valign=c}{\includegraphics{figs/code.png}}, respectively.

\subsection*{\adjustbox{height=2ex, valign=c}{\includegraphics{figs/code.png}}\ Problem 1: Object Detection Using Pre-trained Models}
In this exercise, you will get to experiment with pre-trained computer vision models for image object detection.
Using the provided notebook \\\noindent\colorcode{ch10/exercises/object\_detection.ipynb}:
\begin{enumerate}
\item Implement the code to load and evaluate a pre-trained model for object detection.
\item Implement the function \colorcode{draw\_result} to create an image with the bounding boxes, labels, and scores overlaid.
\item Implement the function \colorcode{filter} to filter the boxes, labels, and scores based on a score threshold.
\end{enumerate}

\newpage
\printbibliography[segment=\therefsegment,heading=subbibliography,title={References}]

\part{Robot Localization and Mapping}
\chapter{Introduction to Localization and Filtering}
\label{ch:intro-to-localization}
\newrefsegment
We have already discussed the robot motion planning problem and surveyed common algorithms for it, ranging from optimal control to sampling-based methods.
All of these approaches implicitly assume access to the robot's current state, for example for initializing trajectory optimization methods or for closing the loop in feedback control.
In practice, however, this state cannot be read directly; it must be \emph{estimated} from noisy, partial sensor data.

Robot perception, as introduced in previous chapters, tackles the challenge of extracting semantic and geometric information from raw sensor streams.
These methods are indispensable for local, instantaneous awareness.
For instance, detecting nearby obstacles with a laser scanner or identifying objects in view with a camera.
Yet this information is inherently \emph{local} and relative to the robot's current position.
It suffices for collision avoidance, but not for the global reasoning required by full planning and control schemes.

\begin{figure}[ht]
	\centering
    \includegraphics[width=0.9\textwidth]{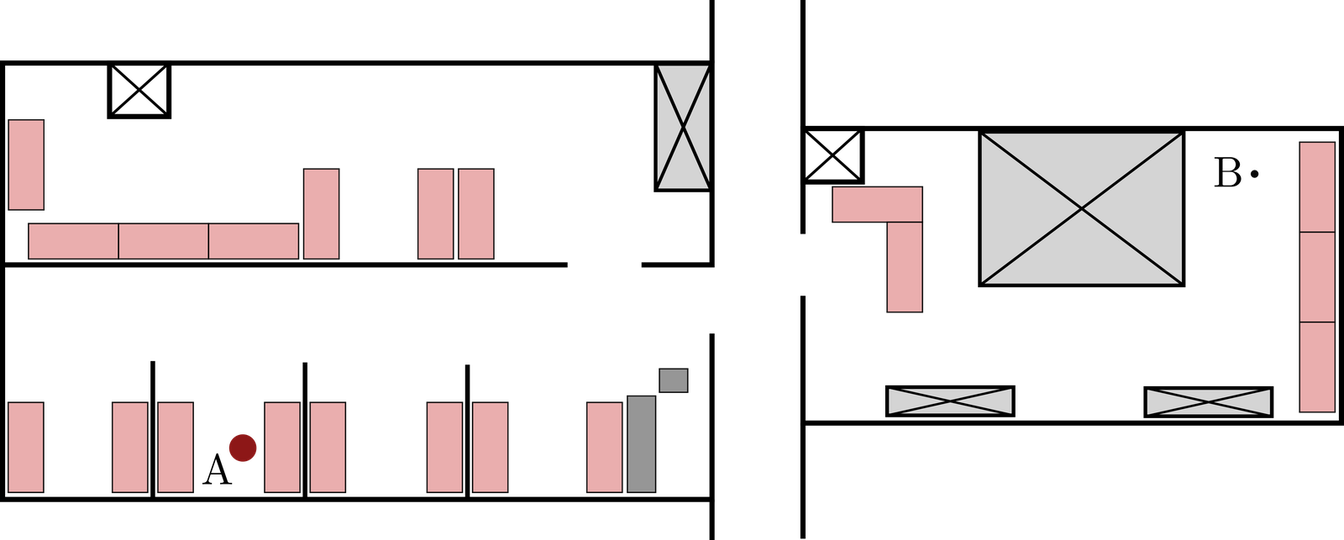}
    \caption{Localization is crucial for autonomy: to move from A to B, the robot must know which room it occupies, and that the only path to B runs through the hallway. 
    Inferring such global information from local range measurements requires specialized algorithms.}
    \label{fig:sample-room}
\end{figure}

This gap is addressed by \emph{robot localization and mapping}, one of the core components of the ``think'' stage in the classical ``see-think-act'' cycle.
The goal of robot localization and mapping is to synthesize local sensor data into a coherent \emph{global estimate} of the robot's state and map the surrounding environment.
In this chapter, we focus on \emph{localization}, which is the ability to infer the robot's current state with respect to a global frame or map~\cite{ThrunBurgardEtAl2005,SiegwartNourbakhshEtAl2011}.
For instance, before a robot can navigate to a target room on the floor plan shown in \cref{fig:sample-room}, it must first establish where in the building it is located. 

A central challenge in localization is \emph{uncertainty}.
Sensor data is noisy, incomplete, and sometimes ambiguous.
To handle this, localization is cast in a \emph{probabilistic} framework: instead of maintaining a single guess of the robot's state, we maintain a \emph{belief distribution} over possible states.
This representation allows us to extract both point estimates and measures of uncertainty.
Uncertainty quantification is vital for downstream tasks.
For example, a planner may avoid high-risk trajectories under localization uncertainty, or even select actions that deliberately reduce uncertainty through information gathering.

The rest of this chapter is organized as follows.
In~\cref{sec:prob_foundations}, we review key concepts in probability theory, including random variables, probability distributions, conditional probabilities, and Bayes' rule.
In~\cref{sec:markov_models}, we introduce Markov models as a probabilistic representation of robot motion and sensing.
Finally, in~\cref{sec:bayes_filter}, we derive the Bayes filter, a recursive algorithm for maintaining and updating a belief distribution as controls are applied and new measurements arrive.
\subsection{Preliminary Concepts in Probability}
\label{sec:prob_foundations}
Tools from probability theory provide us with a way to systematically reason about uncertainty in robotics.
Specifically, these tools provide the language to model noisy sensor measurements, uncertain robot states, and stochastic environments. 
In this section, we review several key building blocks that form the foundation for probabilistic filtering algorithms such as the Bayes filter, namely random variables, probability distributions, conditional probabilities, and Bayes’ rule.

\subsubsection{Random Variables}
We can model uncertain quantities in robotics, such as sensor measurements, robot states, or environmental properties, as \emph{random variables}. 
Depending on the domain of possible values, random variables are classified as discrete or continuous.
\begin{definition}[Discrete random variable]
    A \emph{discrete random variable}~$X$ takes values from a countable set. 
    The probability that~$X$ takes on a specific value~$x$ is denoted by~$p(X=x)$, or more compactly~$p(x)$.
    The function~$p(x)$ is called the \emph{probability mass function} (PMF), and it must satisfy:
    \begin{equation*}
        \sum_{x} p(x) = 1,
    \end{equation*}
    where the sum is over all possible values of~$X$.
\end{definition}

\begin{definition}[Continuous random variable]
    A \emph{continuous random variable}~$X$ takes values in an uncountable set, typically a subset of~$\mathbb{R}^n$. 
    Its distribution is characterized by a \emph{probability density function} (PDF)~$p(x)$, which satisfies:
    \begin{equation*}
        \int_{-\infty}^\infty p(x)\,\d x = 1.
    \end{equation*}
    Unlike the discrete case, the probability that a continuous random variable takes on any \emph{exact} value is zero:
    \begin{equation*}
        P(X = x) = 0.
    \end{equation*}
    Intuitively, this is because a single point has zero width and probability mass only accumulates over intervals. 
    Consequently, probabilities are defined over regions rather than individual values. 
    For example, the probability that~$X$ lies in the interval~$[a,b]$ is:
    \begin{equation*}
        P(a \leq X \leq b) = \int_{a}^{b} p(x)\,\d x.
    \end{equation*}
\end{definition}

\begin{example}[Discrete vs. continuous random variables]
    A coin flip is a discrete random variable,~$X \in \{\text{heads}, \text{tails}\}$, with a probability mass function:
    \begin{equation*}
    p(\text{heads}) = \tfrac{1}{2}, \qquad p(\text{tails}) = \tfrac{1}{2}.
    \end{equation*}
    In robotics, we typically model the robot's pose as a continuous random variable. 
    For example, a planar robot pose~$\x \in SE(2)$\sidenote{$SE(2)$ is the \emph{special Euclidean group in two dimensions}. 
    It represents planar rigid-body transformations and consists of a position $(x,y)$ and an orientation $\theta$. 
    For robots operating in three-dimensional space, the pose lies in $SE(3)$, which represents 3D position and orientation.} 
    can take infinitely many values because both its position and orientation vary continuously.
\end{example}

\paragraph{Probability distributions.} The probability mass function for discrete random variables and probability density function for continuous random variables are often collectively referred to as \emph{probability distributions}.
There are many ways to parameterize a probability distribution.
For discrete variables, the distribution can be specified explicitly by assigning a probability to each possible value. 
For continuous variables, the distribution is often described by a parametric function defined by a small number of parameters. 
Choosing an appropriate representation is important in robotics, since it determines both how uncertainty is modeled and how efficiently algorithms can reason about it.

\subsubsection{Joint Distributions, Independence, and Conditioning}
Many robotics problems involve more than one uncertain quantity at a time. 
For example, a robot might simultaneously reason about its pose, the position of an obstacle, and a sensor reading. 
In such cases, it is useful to describe the probabilities of multiple random variables together using a \emph{joint distribution}.
\begin{definition}[Joint distribution]
The \emph{joint distribution} of two random variables~$X$ and~$Y$ specifies the probability that both take on specific values simultaneously. 
It is denoted by~$p(X=x,\,Y=y)$, or more compactly~$p(x,y)$.
\end{definition}
\paragraph{Independence.}
Random variables can be related to each other in important ways. 
For example, the random variables~$X=$ ``today is cloudy'' and~$Y=$ ``today it is raining'' are correlated: if there are no clouds, it is unlikely to rain. 
In contrast, two random variables are \emph{probabilistically independent} if the value of one does not provide any information about the other.

\begin{definition}[Probabilistic independence]
Two random variables~$X$ and~$Y$ are probabilistically \emph{independent} if and only if:
    \begin{equation}
    \label{eq:indep}
        p(x,y) = p(x)\,p(y).
    \end{equation}
\end{definition}

\begin{example}[Independent sensor measurements]
    \label{ex:sensors-prob}
    Suppose a robot uses a proximity sensor and a temperature sensor, modeled by random variables~$X$ and~$Y$. 
    Let~$X \in \{\text{close}, \text{medium}, \text{far}\}$ and~$Y \in \{\text{low}, \text{med}, \text{high}\}$, with~$p(X=\text{close})=1/3$ and~$p(Y=\text{med})=1/5$. 
    If the two sensors are independent, the joint probability of observing ``close'' and ``med'' is:
    \begin{equation*}
    p(X=\text{close},\,Y=\text{med}) = \tfrac{1}{3}\cdot\tfrac{1}{5} = \tfrac{1}{15}.
    \end{equation*}
\end{example}

\paragraph{Conditional probability.}
Another key concept is the probability of one random variable given that another has already been observed.

\begin{definition}[Conditional probability]
    The \emph{conditional probability} of a random variable~$X$ taking value~$x$, given that~$Y$ took value~$y$, is:
    \begin{equation}
    \label{eq:condprob}
        p(x \mid y) \definedas \frac{p(x,y)}{p(y)}.
    \end{equation}
\end{definition}
Conditional probabilities allow us to update beliefs when new information becomes available. 
If~$X$ and~$Y$ are independent, then~$p(x \mid y) = p(x)$, meaning that knowing~$Y$ provides no additional information about~$X$.

\begin{example}[Sensor conditional probabilities]
    Building on \cref{ex:sensors-prob}, consider an obstacle detection variable~$Z \in \{\text{detected}, \text{not detected}\}$. 
    Assume that the detection probability depends on the proximity sensor value:
\begin{equation*}
\begin{split}
p(Z=\text{detected} \mid X=\text{close}) &= \tfrac{5}{6},\\
    p(Z=\text{detected} \mid X=\text{medium}) &= \tfrac{1}{3},\\
    p(Z=\text{detected} \mid X=\text{far}) &= \tfrac{1}{5}.
\end{split}
\end{equation*}
If~$p(X=\text{close})=\frac{1}{3}$, then the probability that the robot both detects an obstacle and registers ``close'' is:
\begin{equation*}
\begin{split}
p(Z=\text{detected},\,X=\text{close}) &= p(Z=\text{detected}\mid X=\text{close})\,p(X=\text{close}) \\
&= \tfrac{5}{18}.
\end{split}
\end{equation*}
\end{example}

\paragraph{Conditional independence.}
Finally, independence can also hold \emph{given} the outcome of another variable. 
This concept is known as \emph{conditional independence} and plays a central role in probabilistic modeling.

\begin{definition}[Conditional independence]
Two random variables $X$ and $Y$ are said to be \emph{conditionally independent} given a third variable $Z$ if:
\begin{equation}
    p(x,y \mid z) = p(x \mid z)\,p(y \mid z),
\end{equation}
for all values of $x$, $y$, and $z$. 
Equivalently:
\begin{equation}
    p(x \mid y,z) = p(x \mid z).
\end{equation}
We denote conditional independence as:
\begin{equation*}
X \perp Y \mid Z.
\end{equation*}
\end{definition}

Intuitively, conditional independence means that once the value of $Z$ is known, learning the value of $Y$ provides no additional information about $X$. 
However, if $Z$ is not known, the variables $X$ and $Y$ may still appear correlated. 
In other words, the variable $Z$ explains the dependence between $X$ and $Y$.
    
\begin{example}[Conditional independence in robotics]
    A mobile robot equipped with two wheel encoders produces measurements of traveled distance: one from the left wheel ($X$) and one from the right wheel ($Y$). 
    At first glance, these two measurements may seem correlated, since the robot’s motion affects both. 
    However, if we condition on the underlying hidden variable~$Z$ = ``true distance traveled,'' the two encoder readings are independent:
    \begin{equation*}
    p(x,y \mid z) = p(x \mid z)\,p(y \mid z).
    \end{equation*}
    That is, once the actual distance traveled is known, the left and right encoder readings do not provide additional information about each other.
    This is a typical use of conditional independence in probabilistic sensor models.
\end{example}

\subsubsection{Law of Total Probability}
The \emph{law of total probability} links marginal, joint, and conditional probabilities. 
It provides a systematic way to compute the probability of one random variable by accounting for all possible outcomes of another.

\begin{definition}[Law of total probability]
For discrete random variables $X$ and $Y$:
\begin{equation*}
    p(x) = \sum_y p(x,y) = \sum_y p(x \mid y)\,p(y).
\end{equation*}
For continuous random variables:
\begin{equation*}
    p(x) = \int p(x,y)\,dy = \int p(x \mid y)\,p(y)\,\d y.
\end{equation*}
\end{definition}

\noindent Intuitively, the law of total probability states that to find the probability of $X$ taking value $x$, we can sum (or integrate) over all possible values of $Y$, weighting the conditional probability of $X$ given each value of $Y$ by the probability of that value of $Y$ itself.
This process is known as \emph{marginalization}, and~$p(x)$ is called the \emph{marginal probability} of~$X$. 

\begin{example}[Robot localization via marginalization]
Suppose a robot’s position~$X$ depends on which hallway~$Y$ it is currently in. 
We can compute the probability of being at a particular location~$x$ by considering every possible hallway~$y$: 
\begin{equation*}
p(x) = \sum_y p(x \mid y)\,p(y).
\end{equation*}
In practice, this means we marginalize over the possible hallways, combining both the likelihood of being in each hallway and the probability of observing~$x$ given that hallway.
\end{example}

\subsubsection{Bayes' Rule}
The joint probability,~$p(x,y)$, between two random variables,~$X$ and~$Y$, is related to the conditional probabilities,~$p(x \given y)$ and~$p(y \given x)$, from the definition of a conditional probability in \cref{eq:condprob}. 
Since we can express the joint probability using either conditional probability, we have:
\begin{equation*}
    p(x,y) = p(x \given y)p(y) = p(y \given x) p(x).
\end{equation*}
This relationship is commonly referred to as \emph{Bayes' rule}\sidenote{Sometimes also referred to as \emph{Bayes' theorem}.}.
\begin{definition}[Bayes' rule]
For discrete random variables,~$X$ and~$Y$, Bayes' rule states that:
\begin{equation} 
\label{eq:bayes}
    p(x \given y) = \frac{p(y \given x) p(x)}{p(y)}. 
\end{equation}
\end{definition}
Bayes' rule is useful because it provides a relationship between the ``inverse'' conditional probabilities,~$p(x \given y)$ and~$p(y \given x)$. 
This is particularly important for \emph{probabilistic inference} problems where we need to infer the value of one random variable from another.
For example, suppose we have a good initial guess of the probability distribution\sidenote{When we have an estimate of the probability distribution~$p(x)$ before any new information is used to update it, we will refer to it as the \emph{prior} probability.},~$p(x)$, for a random variable,~$X$.
Given new information about the outcome of a second random variable,~$Y$, that is related to~$X$, we can use Bayes' rule to update our belief about the probability distribution of~$X$ by computing~$p(x \given y)$\sidenote{This new distribution, which we obtained by updating the prior distribution~$p(x)$ with the new information about~$Y$, is commonly referred to as the \emph{posterior} probability.}.
Bayes' rule also extends to cases with additional random variables. 
For example, with three random variables,~$X$,~$Y$, and~$Z$, Bayes' rule is:
\begin{equation*}
    p(x \given y, z) = \frac{p(y \given x, z) p(x \given z)}{p(y \given z)}. 
\end{equation*}
\begin{example}[Bayes' rule]
Consider a scenario where a robot is trying to figure out if it is in room A or room B inside of a building.
The robot has an initial guess that the probability it is in room A is~$p(A) = \frac{3}{4}$, and the robot has a camera that can be used to improve the estimate.
Suppose that a single image, $I$, is captured and the features extracted from the image are compared to the known room features which gives the conditional probabilities:
\begin{equation*}
p(I \given A) = \frac{3}{4}, \quad p(I \given B) = \frac{1}{2}.
\end{equation*}
We can use Bayes' rule to compute the posterior probability:
\begin{equation*}
p(A \given I) = \frac{p(I \given A)p(A)}{p(I)},
\end{equation*}
where we use the law of total probability to compute:
\begin{equation*}
p(I) = p(I, A) + p(I, B) = p(I \given A)p(A) + p(I \given B)p(B),
\end{equation*}
and using~$p(B) = 1 - p(A)$.
\end{example}

\subsubsection{Expectation, Variance, and Covariance}
Probability distributions describe uncertainty in full detail by assigning probabilities to every possible outcome of a random variable. 
In practice, however, we often summarize a distribution using more compact statistics. 
Some of the most common statistics used to summarize a distribution include the \emph{expected value}, \emph{variance}, and \emph{covariance}.

\paragraph{Expectation.} The expectation of a random variable is a measure of the central tendency of its distribution.
\begin{definition}[Expectation]
    The \emph{expectation}\sidenote{Also referred to as the \emph{mean} or the \emph{first moment} of a distribution.} of a random variable $X$ is denoted by $\expected{}{X}$. 
    For discrete random variables:
    \begin{equation*}
        \expected{}{X} = \sum_x x\,p(x),
    \end{equation*}
    where the sum is over all outcomes of~$X$. 
    For continuous random variables:
    \begin{equation*}
        \expected{}{X} = \int x\,p(x)\,\d x.
    \end{equation*}
    \end{definition}
\noindent The expected value can be interpreted as the average outcome obtained if the random variable were sampled repeatedly an infinite number of times. 

The expectation has several useful properties. 
One particularly important property is \emph{linearity}, which states that the expectation of a linear transformation of a random variable is equal to the linear transformation of the expectation of the random variable.
Formally, for any random variable~$X$ and constants~$a,b \in \mathbb{R}$, we have:
\begin{equation*}
    \expected{}{aX + b}= a\expected{}{X} + b.
\end{equation*}
This property holds regardless of the distribution of $X$.

For vector-valued random variables $\bX = [X_1,\ldots,X_n]^\top$, the expectation is defined component-wise:
\begin{equation*}
\expected{}{\bX} =
\begin{bmatrix}
\expected{}{X_1}\\
\vdots\\
\expected{}{X_n}
\end{bmatrix}.
\end{equation*}

\paragraph{Variance.}
While the expected value describes the center of a distribution, it does not capture how uncertain the variable is. 
This uncertainty is measured by the \emph{variance}.

\begin{definition}[Variance]
The \emph{variance} of a random variable $X$ is defined as:
\begin{equation*}
    \text{Var}(X) = \expected{}{(X-\expected{}{X})^2}.
\end{equation*}
\end{definition}

The variance measures the average squared deviation of the variable from its mean. 
A large variance indicates that the variable can take values far from the mean, while a small variance indicates that the variable is tightly concentrated around the mean. 
The square root of the variance is called the \emph{standard deviation}, often denoted by $\sigma$.

\paragraph{Covariance.}
When dealing with multiple random variables, it is often important to understand how their uncertainties are related. 
This relationship is captured by the \emph{covariance}.

\begin{definition}[Covariance]
    The \emph{covariance} between two random variables~$X$ and~$Y$ is denoted~$\text{cov}(X,Y)$ and defined as:
    \begin{equation*}
    \begin{split}
        \text{cov}(X,Y) &= \expected{}{(X-\expected{}{X})(Y-\expected{}{Y})^\top} \\
        &= \expected{}{XY^\top} - \expected{}{X}\,\expected{}{Y}^\top .
    \end{split}
    \end{equation*}
\end{definition}

Intuitively, the covariance describes how two random variables vary together.
If the covariance is positive, the variables tend to increase or decrease together. 
If it is negative, one variable tends to increase when the other decreases. 
If the covariance is zero, the variables are uncorrelated.

\begin{example}[Robot motion uncertainty]
Suppose~$X$ represents the forward displacement of a robot and~$Y$ represents its lateral displacement during a single motion step. 
If wheel slip increases as the robot moves farther forward, then larger values of~$X$ tend to be associated with larger sideways deviations, producing a positive covariance between~$X$ and~$Y$. 

Conversely, if the robot's mechanical design or control system tends to stabilize lateral motion during forward travel, the covariance between~$X$ and~$Y$ may be negative.
If forward and lateral displacements arise from unrelated sources, the covariance will be close to zero.
\end{example}
\subsection{Markov Models}
\label{sec:markov_models}
In \cref{ch:model-dyn}, we modeled robot motion using kinematics and dynamics, obtaining a set of first-order differential equations (see \cref{eq:dynamics-ss}) that deterministically describe how the state~$\x$ evolves in time given the current state and control input~$\u$. 
In this section, we generalize this view to a \emph{probabilistic} setting by introducing \emph{Markov models}, which describe how the state evolves under uncertainty. 
Markov models are fundamental to robotics, appearing in localization, mapping, planning, and decision-making under uncertainty problems.

\paragraph{State, controls, and measurements.}
As in \cref{ch:model-dyn}, the state~$\x \in \R^\statedim$ collects all variables relevant to the task at hand. 
In motion planning and control, this typically includes the robot’s physical state (pose, velocity, etc.), while in localization or higher-level planning it may also include environment variables such as landmark positions or object features. 
We work in discrete time, writing~$\x_t$ for the state at time~$t$. 
We also use the shorthand~$\x_{t_1:t_n} \coloneqq \x_{t_1}, \x_{t_2}, \ldots, \x_{t_n}$ for sequences of states, with analogous notation for control inputs~$\u_{t_1:t_n}$ and measurements~$\z_{t_1:t_n}$.\sidenote{Measurements can come from any of the sensors introduced earlier, such as cameras, lidar, or inertial units.}

Unlike deterministic dynamics, Markov models specify probability distributions over possible states and observations. 
In full generality, the state evolution is modeled as:
\begin{equation}
    \label{eq:genprobmod}
        p(\x_t \mid \x_{0:t-1}, \z_{1:t-1}, \u_{1:t}),
\end{equation}
which captures the distribution of the current state~$\x_t$ conditioned on the entire history of past states, controls, and measurements. 
Following the convention used throughout this chapter, the robot first executes the control $\u_t$, then receives the measurement $\z_t$ based on the resulting state $\x_t$. 
The corresponding probabilistic measurement model is:
\begin{equation}
\label{eq:genmeasmod}
    p(\z_t \mid \x_{0:t}, \z_{1:t-1}, \u_{1:t}).
\end{equation}

\paragraph{The Markov property.}
In many applications, we define the state~$\x_t$ to be \emph{complete}, meaning it contains all the information necessary to predict future states. 
Formally, this assumption implies that past states and measurements provide no additional predictive power beyond~$\x_{t-1}$ and~$\u_t$. 
This is known as the \emph{Markov property}, under which the models simplify to:
\begin{equation}
    \label{eq:markovprobmod}
        p(\x_t \mid \x_{t-1}, \u_t),
\end{equation}
for the state transition, and:
\begin{equation}
    \label{eq:markovmeasmod}
        p(\z_t \mid \x_t),
\end{equation}
for the measurement model.

\begin{figure}[tbh]
\begin{center}
    \begin{tikzpicture}[node distance=0cm, ->]
        \tikzstyle{state} = [draw=black, circle, rounded corners, minimum height=3em, minimum width=3em, thick, fill=red!30]
        \tikzstyle{obs} = [draw=black, circle, rounded corners, minimum height=3em, minimum width=3em, thick, fill=white]
        \tikzstyle{action} = [draw=black, rectangle, minimum height=2em, minimum width=2em, thick, fill=gray!30]
        \node[state](s0){$\x_{t-1}$};
        \node[state, right of=s0, xshift=3cm, yshift=0cm](s1){$\x_{t}$};
        \node[state, right of=s1, xshift=3cm, yshift=0cm](s2){$\x_{t+1}$};
        	\node[action, left of=s0, xshift=-1.5cm, yshift=-1.5cm](u0){$\u_{t-1}$};
        	\node[action, left of=s1, xshift=-1.5cm, yshift=-1.5cm](u1){$\u_{t}$};
        	\node[action, left of=s2, xshift=-1.5cm, yshift=-1.5cm](u2){$\u_{t+1}$};
        	\node[obs, right of=s0, xshift=0cm, yshift=-3cm](o0){$\z_{t-1}$};
        	\node[obs, right of=s1, xshift=0cm, yshift=-3cm](o1){$\z_{t}$};
        	\node[obs, right of=s2, xshift=0cm, yshift=-3cm](o2){$\z_{t+1}$};
		\draw[->, thick] (s0) to[] (s1);
		\draw[->, thick] (s1) to[] (s2);
		\draw[->, thick] (u0) to[] (s0);
		\draw[->, thick] (u1) to[] (s1);
		\draw[->, thick] (u2) to[] (s2);
		\draw[->, thick] (s0) to[] (o0);
		\draw[->, thick] (s1) to[] (o1);
		\draw[->, thick] (s2) to[] (o2);
    \end{tikzpicture}
\end{center}
\caption{Graphical representation of a Markov model. 
    At each time step, the control $\u_t$ influences the new state $\x_t$, and the resulting state generates a measurement $\z_t$.}
\label{fig:markov_model}
\end{figure}
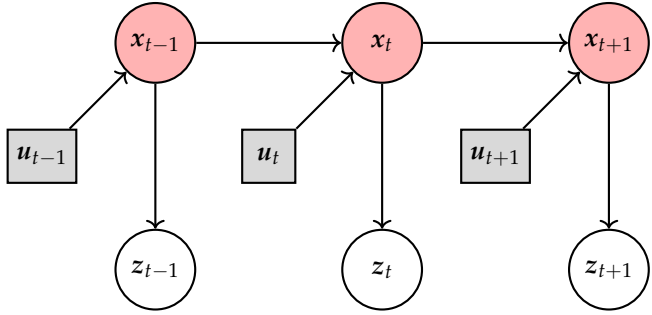

\paragraph{Markov models in robotics.}
A Markov model thus consists of a state transition distribution~\eqref{eq:markovprobmod} and a measurement distribution~\eqref{eq:markovmeasmod}. 
Intuitively, the transition model captures process uncertainty (e.g., wheel slip when applying a control), while the measurement model captures sensor noise (e.g., rangefinder variability). 
Together, these components form the foundation of probabilistic state estimation (\cref{fig:markov_model}). 

The probabilistic model described above belongs to a broader family of models that assume the system state evolves according to the Markov property. 
Several closely related formulations exist, depending on whether control inputs and observability are considered.

In robotics, the state of the system is typically not directly observable. 
Instead, sensors provide indirect and noisy measurements that depend on the underlying state. 
Models with this structure are often referred to as \emph{partially observable Markov models}.

If control inputs are not present, the model reduces to a \emph{hidden Markov model (HMM)}. 
In this case, the state evolves according to a Markov process, and observations provide partial information about that state. 
The term ``hidden'' reflects the fact that the true state~$\x_t$ cannot be observed directly and must instead be inferred from the sequence of measurements.

\subsection{Bayes Filter}
\label{sec:bayes_filter}
Robot localization is a classic instance of a filtering problem\sidenote{Localization is one instance of the more general filtering problem referred to as \emph{state estimation}.} where our goal is to compute a probability distribution over the current state~$\x_t$ given the history of control inputs~$\u_{1:t}$, and measurements~$\z_{1:t}$.
One of the canonical approaches to this filtering problem is known as the \emph{Bayes filter} or \emph{recursive Bayesian estimation}.
The Bayes filter leverages a Markov model to recursively update a belief distribution, which is a probability distribution over~$\x_t$.
Mathematically, we denote the belief distribution as~$\bel(\x_t)$ and define it as:
\begin{equation} 
\label{eq:belief}
\bel(\x_t) \definedas p(\x_t \given \z_{1:t}, \u_{1:t}).
\end{equation}
In other words, the belief $\bel(\x_t)$ is a posterior probability distribution over the state conditioned on the available history information. 
We also define a distribution called the \emph{prediction} distribution as:
\begin{equation} 
\label{eq:predbelief}
\belpred(\x_t) \definedas p(\x_t \given \z_{1:t-1},\u_{1:t}),
\end{equation}
which does not include the most recent measurement~$\z_t$. 
We call the process of using the new measurement~$\z_t$ to compute the belief~$\bel(\x_t)$ from the predicted belief~$\belpred(\x_t)$ a \emph{correction} or \emph{measurement update}.
The Bayes filter consists of a prediction step for computing~$\belpred(\x_t)$ from the prior belief followed by a correction step for computing~$\bel(\x_t)$ given the new measurement~$\z_t$.

\subsubsection{Algorithm}
The recursive structure of the Bayes filter is summarized in \cref{alg:bayes}. 
At each time step, the filter maintains a belief distribution \emph{belief}~$\bel(\x_t)$ over the system state. 
The inputs to the filter are the previous belief\sidenote{In practice, we initialize the prior distribution $\bel(\x_0)$ using either a best guess of the initial state or a uniform distribution when no prior information is available.}, the current control input~$\u_t$, and the latest sensor measurement~$\z_t$.

The algorithm proceeds in two stages. 
First, a \emph{prediction} step computes the distribution $\belpred(\x_t)$ over possible states after applying the control input. 
This step propagates uncertainty forward using the state transition model from \cref{eq:markovprobmod}.
Second, a \emph{correction} step incorporates the new measurement using the measurement model from \cref{eq:markovmeasmod}, adjusting the predicted belief to better match the observed data.

Intuitively, the prediction step estimates where the robot might be after executing the control, while the correction step refines this estimate using information from the sensors.

\begin{algorithm}[ht!]
    \KwData{$\bel(\x_{t-1}), \u_t, \z_t$}
    \KwResult{$\bel(\x_t)$}
    \ForEach{$\x_t$}{
       \tcp{Prediction (motion update)}
       $\belpred(\x_t) = \int p(\x_t \mid \x_{t-1}, \u_t)\,\bel(\x_{t-1})\,\d\x_{t-1}$ \\
       \tcp{Correction (measurement update)}
       $\bel(\x_t) = \eta\, p(\z_t \mid \x_t)\,\belpred(\x_t)$
    }
    \Return $\bel(\x_t)$
    \caption{Bayes Filter}
    \label{alg:bayes}
   \end{algorithm}

In \cref{alg:bayes},~$\eta$ is a normalization constant ensuring~$\bel(\x_t)$ integrates (or sums) to one.\sidenote{In practice,~$\eta = 1 / p(\z_t \mid \z_{1:t-1}, \u_{1:t})$, which follows directly from Bayes' rule.} 
Conceptually, the Bayes filter performs a repeated \emph{predict–correct} cycle, where the motion model spreads the belief forward to account for process uncertainty, while the measurement model reshapes the belief according to how consistent each state is with the observed sensor measurement.

\subsubsection{Derivation}
We now derive the Bayes filter recursion from the definition of the belief distribution introduced in \cref{sec:bayes_filter}. 
Applying Bayes' rule to \cref{eq:belief} yields:
\begin{equation}
\begin{split}
    \bel(\x_t)
    &= p(\x_t \given \z_{1:t}, \u_{1:t}) \\
    &= \eta\, p(\z_t \given \x_t, \z_{1:t-1}, \u_{1:t}) \,
        p(\x_t \given \z_{1:t-1}, \u_{1:t}),
\end{split}
\label{eq:bel_bayes_rule}
\end{equation}
where $\eta$ is a normalization constant:
\begin{equation*}
\eta = \frac{1}{p(\z_t \given \z_{1:t-1}, \u_{1:t})}.
\end{equation*}

\paragraph{Measurement update.}
Using the conditional independence assumptions of the Markov model (\cref{fig:markov_model}), the current measurement depends only on the current state. 
Therefore:
\begin{equation*}
p(\z_t \given \x_t, \z_{1:t-1}, \u_{1:t}) = p(\z_t \given \x_t).
\end{equation*}
Substituting this simplification into \cref{eq:bel_bayes_rule} gives:
\begin{equation*}
\bel(\x_t) = \eta\, p(\z_t \given \x_t)\, p(\x_t \given \z_{1:t-1}, \u_{1:t}).
\end{equation*}
Recall the definition of the prediction belief from \cref{eq:predbelief}:
\begin{equation*}
\belpred(\x_t) \definedas p(\x_t \given \z_{1:t-1}, \u_{1:t}),
\end{equation*}
we obtain:
\begin{equation*}
\bel(\x_t) = \eta\, p(\z_t \given \x_t)\, \belpred(\x_t),
\end{equation*}
which corresponds to the \emph{measurement update} step of the Bayes filter.

\paragraph{Prediction update.}
Next we derive an expression for the prediction belief $\belpred(\x_t)$. 
Starting from its definition in \cref{eq:predbelief}, we apply the law of total probability to marginalize over the previous state $\x_{t-1}$:
\begin{equation*}
\begin{split}
\belpred(\x_t)
&= \int p(\x_t, \x_{t-1} \given \z_{1:t-1}, \u_{1:t}) \d\x_{t-1} \\
&= \int p(\x_t \given \x_{t-1}, \z_{1:t-1}, \u_{1:t})
   p(\x_{t-1} \given \z_{1:t-1}, \u_{1:t}) \d\x_{t-1}.
\end{split}
\end{equation*}

Using the Markov assumption again, the next state depends only on the previous state and the current control input:
\begin{equation*}
p(\x_t \given \x_{t-1}, \z_{1:t-1}, \u_{1:t}) =
p(\x_t \given \x_{t-1}, \u_t).
\end{equation*}

Furthermore, the control input $\u_t$ does not influence the previous state $\x_{t-1}$, so:
\begin{equation*}
p(\x_{t-1} \given \z_{1:t-1}, \u_{1:t}) =
p(\x_{t-1} \given \z_{1:t-1}, \u_{1:t-1}).
\end{equation*}
Recognizing that, by definition:
\begin{equation*}
\bel(\x_{t-1}) = p(\x_{t-1} \given \z_{1:t-1}, \u_{1:t-1}),
\end{equation*}
we obtain the prediction step:
\begin{equation*}
\belpred(\x_t) =
\int p(\x_t \given \x_{t-1}, \u_t)\,
     \bel(\x_{t-1})\, \d\x_{t-1}.
\end{equation*}

\subsubsection{Discrete Bayes Filter}
When the state space is finite, the belief distribution can be represented as a probability mass function over a discrete set of states $\{x_k\}$. 
In this case, the belief at time $t$ is described by a collection of probabilities $\{p_{k,t}\}$, where $p_{k,t}$ denotes the probability that the system is in state $x_k$ at time $t$.
The Bayes filter recursion can then be written in discrete form by replacing the integrals in \cref{alg:bayes} with summations over the possible states. 
The resulting algorithm, shown in \cref{alg:discretebayes}, follows the same two-step procedure as the continuous Bayes filter, with a prediction step using the transition model, followed by a correction step using the measurement model.
\begin{algorithm}[ht!]
 \KwData{$\{p_{k,t-1}\},  \u_{t}, \z_{t}$}
 \KwResult{$\{p_{k,t}\}$}
 \ForEach{$k$}{
  $\overline{p}_{k,t}=\sum_{i} p(\x_{t}\given \x_{i}, \u_{t})\,p_{i,t-1}$\\
  $p_{k,t}=\eta\, p(\z_{t}\given \x_{k})\,\overline{p}_{k,t}$\\
 }
 \Return $\{p_{k,t}\}$
 \caption{Discrete Bayes Filter}
 \label{alg:discretebayes}
\end{algorithm}
Here $\overline{p}_{k,t}$ denotes the predicted probability of state $x_k$ before incorporating the measurement, and $\eta$ is a normalization constant ensuring that the probabilities sum to one.

\subsubsection{Practical Considerations}

The Bayes filter provides a general and principled framework for probabilistic state estimation. 
However, applying it directly is often computationally challenging in realistic robotics problems.

In \emph{continuous state spaces}, the prediction step requires evaluating integrals over the entire state space. 
Computing these integrals exactly is often intractable, resulting in the need for approximations or numerical methods to estimate the belief distribution.
In \emph{discrete state spaces}, the recursion can be computed exactly, but the required summations scale with the number of possible states. 
As the dimensionality of the state increases, the number of discrete states grows rapidly, making exact inference computationally expensive.

Despite these challenges, the Bayes filter serves as the conceptual foundation for many practical state estimation algorithms. 
Widely used filters such as the Kalman filter, extended Kalman filter (EKF), unscented Kalman filter (UKF), and particle filter can all be viewed as specific implementations of the Bayes filter that exploit additional assumptions or approximations to make the computation tractable.

\begin{example}[Robot in a hallway]
    Consider a robot moving along a straight hallway represented by a one-dimensional line. 
    The hidden state $\x_t$ is the robot’s position along this line. 
    At each time step $t$ the robot receives a control input $\u_t$ representing a commanded forward displacement and a measurement $\z_t$ representing the noisy distance to the nearest door in front of the robot, obtained from a range sensor.

    \paragraph{Prediction.}
    Suppose the robot starts at position $\x_{t-1}=2$\,m with a belief concentrated around that location and issues a command $\u_t=+0.5$\,m forward. 
    Due to wheel slip and actuator noise, the true displacement may vary, which we model as:
    \begin{equation*}
    \Delta x \sim \mathcal{N}(u_t,\,0.1^2).
    \end{equation*}
    The prediction step therefore spreads the belief forward, producing belief $\belpred(\x_t)$ centered at $2.5$\,m but with larger variance than the previous belief. 
    
    \paragraph{Correction.}
    At the same step, the robot’s range sensor reports $\z_t=2.4$\,m to the next door. 
    The sensor is noisy and modeled by:
    \begin{equation*}
    p(\z_t \mid \x_t) = \mathcal{N}(\z_t; \text{true distance}(\x_t),\,0.05^2).
    \end{equation*}
    If the predicted belief assigns significant probability to states near $\x_t \approx 2.5$\,m and the hallway map indicates that the next door lies roughly $2.4$\,m ahead from that position, the measurement is consistent with the prediction. 
    The correction step therefore increases the probability of these states and sharpens the belief distribution around them. 
    States that would predict very different measurements are downweighted.

    \paragraph{Recursive operation.}
    Over time, the filter alternates prediction and correction. 
    The prediction step tends to broaden the belief because motion introduces uncertainty, while the correction step can concentrate the belief when informative measurements are received.
    In hallways with repeating structures, several positions may initially produce similar sensor readings, causing the belief to remain multimodal.
    As the robot gathers additional observations, the Bayes filter resolves this ambiguity and the belief collapses around the robot's true location.
    
    In the repository \colorcode{github.com/StanfordASL/pora-exercises}, the notebook \colorcode{ch11/discrete\_bayes.ipynb} provides a concrete implementation of the discrete Bayes filter for this hallway example. 
    Running the notebook illustrates how the belief distribution evolves over time as the robot moves and collects measurements.
\end{example}

\section{Summary}
In this chapter, we introduced the probabilistic foundations underlying robot localization and state estimation. 
We began by motivating why localization is essential for autonomy: robots must infer their pose from noisy and partial sensor data rather than directly observing it. 
To reason systematically about this uncertainty, we reviewed key concepts in probability theory, including random variables, probability distributions, conditional independence, and Bayes’ rule. 
These concepts provide the mathematical framework for representing and updating uncertainty in robotic systems.

Building on these foundations, we introduced Markov models as probabilistic representations of robot motion and sensing. 
These models define how the robot’s state evolves over time and how measurements relate to that state, forming the basis for probabilistic inference in dynamic systems. 
Finally, we derived the Bayes filter, a recursive algorithm for estimating a belief distribution over the robot’s state as it moves and collects new measurements. 
The Bayes filter provides the conceptual foundation for many practical estimation algorithms. 
Widely used methods such as the Kalman filter, extended Kalman filter, unscented Kalman filter, and particle filter can all be understood as specific implementations of this general framework. 
In the following chapters, we build on these ideas to develop practical algorithms for localization, mapping, and simultaneous localization and mapping.

\paragraph{To learn more.}
For a comprehensive introduction to probabilistic robotics, including localization and filtering, readers are referred to the seminal text by~\citet{ThrunBurgardEtAl2005}, which provides an intuitive and rigorous treatment of the Bayes filter and its extensions. 
Additional foundational perspectives on probabilistic reasoning and estimation in robotics can be found in~\citet{SiegwartNourbakhshEtAl2011} and~\citet{maybeck1982stochastic}. 
For readers interested in a more theoretical background in stochastic systems and control,~\citet{gelb1974applied} offers a classical reference on estimation theory. 
Finally,~\citet{slam-handbook} presents a modern, unified treatment of localization, filtering, and mapping under a common probabilistic framework.

\section{Exercises}
The starter code for the exercises provided below is available online through GitHub. 
To get started, download the code by running in a terminal window:

\begin{tcolorbox}[colback=gray!10]
\begin{minted}{bash}
    git clone https://github.com/StanfordASL/pora-exercises.git
\end{minted}
\end{tcolorbox}

We denote Problems requiring hand-written solutions and coding in Python with \adjustbox{height=2ex, valign=c}{\includegraphics{figs/write.png}} and \adjustbox{height=2ex, valign=c}{\includegraphics{figs/code.png}}, respectively.

\subsection*{\adjustbox{height=2ex, valign=c}{\includegraphics{figs/write.png}}\ Problem 1: Airport Security}
Suppose that travelers passing through an airport carry prohibited items $1\%$ of the time.
Each passenger passes through a simple detector that has a true positive detection rate of $92\%$ and false positive detection rate of $8\%$.
Passengers that trigger the detector alarm are sent to a more accurate secondary screening that has a true positive detection rate of $98\%$ and a false positive detection rate of $10\%$.
\begin{enumerate}
\item Given that a randomly selected passenger was flagged by the second screening, what is the probability a passenger was carrying a prohibited item?
\item How does the probability change in this case if \emph{all} passengers were to get the more accurate secondary screening and the first simple detector was not used?
\end{enumerate}

\subsection*{\adjustbox{height=2ex, valign=c}{\includegraphics{figs/write.png}}\ Problem 2: Cookie Machine}
You operate a cookie-making machine that when activated will produce a random number of cookies in the range $[1, \dots, N]$, with each quantity equally likely with probability $\frac{1}{N}$.
Each cookie costs $\$1$ to produce.
\begin{enumerate}
\item Each of your customers will pay a fixed price to activate the machine.
What is the minimum fixed price you should charge each customer to ensure that you don't lose money in the long run?
In other words, what is the expected cost of each customer activation?
\end{enumerate}
Suppose the machine manufacturer produces machines with $N=6$ and $N=10$, and unfortunately they don't know which one they sent you, but the chance you received either machine is equally likely.
However, they did record that in a test run of your machine it produced $5$ cookies.
\begin{enumerate}
\setcounter{enumi}{1}
\item What is the probability that you received a machine configured with $N=6$?
\item What is the expected number of cookies that will be produced the first time you operate the machine?
\end{enumerate}

\newpage
\printbibliography[segment=\therefsegment,heading=subbibliography,title={References}]
\chapter{Approximate Filters for State Estimation}
\label{ch:approximate-filters}
\newrefsegment
In \cref{ch:intro-to-localization}, we introduced the Bayes filter as the canonical framework for state estimation.
The Bayes filter uses a recursive procedure that alternates between \emph{prediction}, using a probabilistic state transition model, and \emph{correction}, using a probabilistic measurement model. 
Intuitively, the prediction step answers the question, ``Where do we expect to be now, given where we were and how we moved?'', while the correction step asks, ``How should we revise that expectation in light of the new sensor data?''

While conceptually elegant, the Bayes filter is rarely tractable to implement in its full generality. 
The integrals in the prediction step and the normalization in the correction step can be computationally intractable for continuous, high-dimensional state spaces.
In discrete domains, exact enumeration is possible in principle but becomes impractical as the number of states grows. 
As a result, practical state estimation algorithms rely on approximations of the belief distribution.
Over time, two broad families of approximations have emerged:

\begin{itemize}
    \item \emph{Parametric filters:} parametric filters assume that the belief distribution belongs to a specific parametric family (most commonly Gaussian), characterized by a fixed set of parameters such as mean and covariance~\citep{ThrunBurgardEtAl2005}. 
    By exploiting the structure of this representation, the belief can be updated efficiently at each time step. 
    The Kalman filter and its variants, including the extended Kalman filter and unscented Kalman filter, are prominent examples.
    
    \item \emph{Non-parametric filters:} non-parametric filters do not assume a fixed functional form for the belief distribution. 
    Instead, the distribution is approximated directly, either through discretization as in histogram filters or through sampling as in particle filters. 
    This flexibility allows non-parametric methods to represent multimodal and highly irregular belief distributions, although often at a higher computational cost.
\end{itemize}

Viewed together, parametric and non-parametric filters represent two ends of a spectrum. 
Parametric filters trade representational flexibility for computational efficiency, while non-parametric filters trade efficiency for expressiveness. 
In robotics practice, both families play a critical role.
Parametric filters often suffice when the problem structure is close to Gaussian and unimodal, while non-parametric filters are indispensable when ambiguity, multimodality, or strong nonlinearities are present.
It is helpful to view them as \emph{complementary} rather than mutually exclusive. 
Many systems use non-parametric methods for global reasoning, then switch to parametric filters for fast local tracking once a unique hypothesis has been identified.
For instance, imagine a mobile robot navigating a building.
Early in the mission, it may be unsure which corridor or even which floor it is on.
In this case, its belief is naturally multimodal and non-parametric methods shine.
As the robot gathers more information and locks onto a unique hypothesis, its uncertainty becomes locally well-approximated by a single Gaussian, and parametric filters become attractive for their speed and simplicity.

In the remainder of this chapter, we develop both approaches within a unified narrative.
We begin by reviewing the Gaussian distribution in \cref{subsec:gaussian}, which forms the foundation of parametric filters.
We then introduce the Kalman filter and its variants in \cref{subsec:kalman-filter} and \cref{subsec:kalman-filter-extensions}, which are the most widely used parametric filters in robotics.
Next, we turn to non-parametric filters in \cref{subsec:nonparametric}, where we relax these assumptions and represent beliefs more directly through discretization (\cref{subsec:histogram-filter}) or sampling (\cref{subsec:particle-filter}).

\subsection{The Gaussian Distribution}
\label{subsec:gaussian}
Before we introduce specific filters, it is worth pausing to review the Gaussian distribution, which is the workhorse of parametric state estimation.
The Gaussian distribution\sidenote{Also referred to as the Normal distribution.} is one of the most widely used probability distributions in science and engineering and plays a central role in robotics state estimation. 
Its importance arises not only from its frequent appearance in natural noise processes, but also from its favorable mathematical properties that make recursive filtering tractable.

Informally, in one dimension, a Gaussian distribution resembles the familiar ``bell curve'': it is centered at its mean,~$\mu$, and its spread is controlled by its variance,~$\sigma^2$.
In higher dimensions, a Gaussian describes an ellipsoidal cloud of probability mass in the state space. 
It is high near the mean, low far away, and its covariance matrix tells us in which directions the uncertainty is large or small.

\paragraph{Univariate case.}
The probability density function of a one-dimensional\sidenote{We refer to a one-dimensional Gaussian as \emph{univariate} and to higher-dimensional cases as \emph{multivariate}.} Gaussian random variable $X$ with mean $\mu$ and variance $\sigma^2$ is:
\begin{equation}
    p(x) = \frac{1}{\sqrt{2\pi \sigma^2}} 
           \exp\!\left(-\tfrac{1}{2}\tfrac{(x-\mu)^2}{\sigma^2}\right).
    \end{equation}
We write this compactly as~$X \sim \mathcal{N}(\mu, \sigma^2)$, and say that ``$X$ is distributed as a Gaussian with mean~$\mu$ and variance~$\sigma^2$''.
The mean,~$\mu$ indicates the center of mass of the distribution, and the variance,~$\sigma^2$, measures how spread out the distribution is around that center.

\paragraph{Multivariate case.}
For an~$n$-dimensional random vector~$\bX \in \R^n$ with mean~$\bmu \in \R^n$ and covariance matrix~$\Sigma \in \R^{n \times n}$, the multivariate Gaussian distribution is defined by:
\begin{equation}
p(\x) = \frac{1}{\sqrt{\det(2\pi \Sigma)}}
\exp\!\left(-\tfrac{1}{2}(\x-\bmu)^\top\Sigma^{-1}(\x-\bmu)\right),
\end{equation}
and we compactly write~$\bX \sim \mathcal{N}(\bmu,\Sigma)$. 
The covariance~$\Sigma$ captures both the spread of each component of~$\x$ and their pairwise correlations.
Geometrically, the level sets of a multivariate Gaussian\sidenote{Each level set contains points of equal probability density.} are ellipsoids centered at~$\bmu$, with shape and orientation determined by the covariance matrix~$\Sigma$.

The Gaussian distribution exhibits several important mathematical properties related to affine transformations, addition, and multiplication, which make it particularly attractive for use in filtering algorithms.
We highlight three that will be used repeatedly in what follows.

\paragraph{Affine transformations.}
The first useful property of the Gaussian distribution is that an affine transformation of a Gaussian random variable is also a Gaussian random variable. 
If the random vector~$\bX$ has a multivariate Gaussian distribution with mean~$\bmu$ and covariance~$\Sigma$, then the random variable~$\bm{Y}$ computed from an affine transformation:
\begin{equation*}
    \bm{Y} = A\bX + b,
\end{equation*}
also has a multivariate Gaussian distribution with mean $A\bmu + b$ and covariance $A\Sigma A^\top $. 
In other words, if~$\bX \sim \mathcal{N}(\bmu, \Sigma)$, then~$\bm{Y} \sim \mathcal{N}(A\bmu+b, A\Sigma A^\top )$.

In the context of robotics, this tells us that if our belief over the current state is Gaussian and the dynamics are linear with additive Gaussian noise, then the predicted state is also Gaussian.
We can therefore keep track of just the mean and covariance instead of an entire arbitrary density.

\paragraph{Sum.}
The next useful property of Gaussians is that the sum of two independent Gaussian random variables is also a Gaussian random variable. 
Suppose~$\bX_1$ and~$\bX_2$ have multivariate Gaussian distributions with means~$\bmu_1$ and~$\bmu_2$ and covariances~$\Sigma_1$ and~$\Sigma_2$. 
Then, the random variable~$\bm{Y}$ computed by the sum:
\begin{equation*}
	\bm{Y} = \bX_1 + \bX_2,
\end{equation*}
also has a multivariate Gaussian distribution with mean~$\bmu_1 + \bmu_2$ and covariance~$\Sigma_1 + \Sigma_2$. 
In other words, if $\bX_1 \sim \mathcal{N}(\bmu_1, \Sigma_1)$ and $\bX_2 \sim \mathcal{N}(\bmu_2, \Sigma_2)$, then $\bm{Y} \sim \mathcal{N}(\bmu_1 + \bmu_2, \Sigma_1 + \Sigma_2)$.

In robotics, this property commonly appears when modeling additive noise. 
For example, if a robot's predicted state is Gaussian and we add independent Gaussian process noise, the resulting state distribution remains Gaussian with covariance equal to the sum of the individual covariances.

\paragraph{Product.}
The product of two Gaussian probability density functions is also a Gaussian probability density function.
Consider two Gaussian probability density functions:
\begin{equation*}
    \begin{split}
    p_1(\x) &= \frac{1}{\sqrt{\det(2\pi \Sigma_1)}} \exp\big( -\frac{1}{2}(\x-\bmu_1)^\top  \Sigma_1^{-1} (\x-\bmu_1) \big)\\
    p_2(\x) &= \frac{1}{\sqrt{\det(2\pi \Sigma_2)}} \exp\big( -\frac{1}{2}(\x-\bmu_2)^\top  \Sigma_2^{-1} (\x-\bmu_2) \big).
    \end{split}
\end{equation*}
Their product is:
\begin{equation*}
    \begin{split}
    p(\x)&=p_1(\x)\cdot p_2(\x)\\
    &=\frac{\exp\big( -\frac{1}{2}(\x-\bmu_1)^\top  \Sigma_1^{-1} (\x-\bmu_1)-\frac{1}{2}(\x-\bmu_2)^\top  \Sigma_2^{-1} (\x-\bmu_2) \big)}{(2\pi)^d\sqrt{\det(\Sigma_1)}\sqrt{\det(\Sigma_2)}}\\
    &=\frac{\exp\big(-\frac{1}{2}(\x-\bmu)^\top \Sigma^{-1}(\x-\bmu)\big)\exp\big(-\frac{1}{2}(\bmu_1^\top\Sigma_1^{-1}\bmu_1+\bmu_2^\top\Sigma_2^{-1}\bmu_2-\bmu^\top\Sigma^{-1}\bmu)\big)}{(2\pi)^d\sqrt{\det(\Sigma_1)}\sqrt{\det(\Sigma_2)}},
    \end{split}
\end{equation*}
where~$d$ is the dimension of the covariance matrices, and we can see that the second exponential is constant with respect to~$\x$.
Therefore, the product is a Gaussian probability density function with mean~$\bmu$ and covariance~$\Sigma$:
\begin{equation*}
    \begin{split}
        \Sigma&=(\Sigma_1^{-1}+\Sigma_2^{-1})^{-1},\\
        \bmu&=\Sigma(\Sigma_1^{-1}\bmu_1+\Sigma_2^{-1}\bmu_2).
    \end{split}
\end{equation*}

This property underlies the Bayes filter measurement update, where multiplying a Gaussian prior by a Gaussian likelihood yields a Gaussian posterior.
The update simply shifts the mean and shrinks or expands the covariance according to how informative and reliable the measurement is.

\paragraph{Why Gaussians in filtering?}
These properties ensure that when both the transition and measurement models are linear with Gaussian noise, the Bayes filter reduces to simple recursive updates of the mean and covariance. 
This leads directly to the family of \emph{Kalman filters}, which we will introduce in the next section.
From a computational point of view, this is extremely attractive.
Instead of carrying around an entire function~$p(\x_t)$, we only need to carry a vector~$\bmu_t$ and a matrix~$\Sigma_t$ and update them at each time step.

\subsection{Kalman Filter}
\label{subsec:kalman-filter}
The Kalman filter is the canonical \emph{parametric} realization of the Bayes filter for systems with linear dynamics and Gaussian noise. 
Specifically, the Kalman filter uses a multivariate Gaussian distribution to parameterize the belief distribution over possible states.
In other words, we assume~$\x_t \sim \mathcal{N}(\bmu_t, \Sigma_t)$, so that:
\begin{equation*}
\bel(\x_t) = \frac{1}{\sqrt{\det(2\pi \Sigma_t)}} \exp\big( -\frac{1}{2}(\x_t-\bmu_t)^\top  \Sigma_t^{-1} (\x_t-\bmu_t) \big).
\end{equation*}

\begin{example}[Constant-velocity motion in one dimension]
\label{ex:kf-1d}  
Consider a robot moving along a straight corridor. 
The state encodes position and velocity:
\begin{equation*}
  \x_t =
  \begin{bmatrix}
    p_t \\ v_t
  \end{bmatrix},
\end{equation*}
and the robot receives noisy position measurements from a range sensor.
A common linear-Gaussian model is:
\begin{align*}
    \x_t &= A_t \x_{t-1} + B_t \u_t + \bm{\epsilon}_t, \\
    \z_t &= C_t \x_t + \bm{\delta}_t,
\end{align*}
where:
\begin{equation*}
  A_t =
  \begin{bmatrix}
    1 & \Delta t \\
    0 & 1
  \end{bmatrix}, \quad
  B_t =
  \begin{bmatrix}
    \tfrac{1}{2}\Delta t^2 \\
    \Delta t
  \end{bmatrix}, \quad
  C_t =
  \begin{bmatrix}
    1 & 0
  \end{bmatrix},
\end{equation*}
and where the random variables~$\bm{\epsilon}_t \sim \mathcal{N}(0,Q_t)$,~$\bm{\delta}_t \sim \mathcal{N}(0,R_t)$ model process noise and measurement noise, respectively. 
In this setting, the Kalman filter provides the optimal recursive estimator of~$\x_t$ in the mean-squared error sense.
\end{example}

Like the Bayes filter, the Kalman filter is split up into two steps, a prediction step and measurement update step.
Both steps update the mean~$\bmu$ and covariance~$\Sigma$ of the Gaussian belief and rely on several structural assumptions about the system.

We first assume that the initial belief $\bel(\x_0)$ is Gaussian with~$\x_0 \sim \mathcal{N}(\bmu_0, \Sigma_0)$.
We also assume that the state transition model is linear and evolves according to:
\begin{equation} 
\label{eq:KFdynamics}
    \x_t = A_t \x_{t-1} + B_t \u_t + \bm{\epsilon}_t,
\end{equation}
where~$\x_{t-1}$ denotes the previous state,~$\u_t$ is the current control input,~$\bm{\epsilon}_{t}$ is an independent process noise that is normally distributed according to~$\bm{\epsilon}_t \sim \mathcal{N}(\bm{0},\stateNoise_t)$, and~$A_t$ and~$B_t$ are time-varying matrices that define the dynamics.
The matrix~$Q_t \in \mathbb{R}^{n\times n}$ is the process noise covariance and captures uncertainty in the motion model\sidenote{Examples include unmodeled accelerations and environmental disturbances.}

The affine structure of the dynamics together with Gaussian noise ensures that the distribution of the next state remains Gaussian. 
That is, if $\x_{t-1}$ is Gaussian, then $\x_t$ is also Gaussian. 
The corresponding probabilistic transition model can therefore be written as:
\begin{equation*}
\begin{split}
p(\x_t &\mid \x_{t-1}, \u_t) = \\
&\frac{1}{\sqrt{\det(2\pi \stateNoise_t)}} \exp\big( -\frac{1}{2}(\x_t-A_t \x_{t-1} - B_t \u_t)^\top  \stateNoise_t^{-1} (\x_t-A_t \x_{t-1} - B_t \u_t) \big),
\end{split}
\end{equation*}
and therefore the next state is normally distributed with:
\begin{equation*}
\x_t \sim \mathcal{N}(A_t \x_{t-1} + B_t \u_t, \: \stateNoise_t).
\end{equation*}

We also assume that the measurement model is linear and of the form:
\begin{equation} 
\label{eq:KFmeasure}
\z_t = C_t \x_t + \bm{\delta}_t,
\end{equation}
where~$\bm{\delta}_t$ is an independent measurement noise that is normally distributed according to~$\bm{\delta}_t \sim \mathcal{N}(\bm{0},\measNoise_t)$, and~$C_t$ is a time-varying matrix that defines how the state maps to measurements.
The matrix~$R_t \in \mathbb{R}^{m\times m}$ is the measurement noise covariance that describes the uncertainty in the sensor measurements.

Under these assumptions, the probabilistic measurement model can be expressed as:
\begin{equation*}
p(\z_t \mid \x_t) = \frac{1}{\sqrt{\det(2\pi \measNoise_t)}} \exp\big( -\frac{1}{2}(\z_t-C_t\x_t)^\top  \measNoise_t^{-1} (\z_t-C_t\x_t) \big),
\end{equation*}
which implies that $\z_t \given \x_t \sim \mathcal{N}(C_t \x_{t}, \: \measNoise_t)$.
In many robotics applications, the matrix $C_t$ selects a subset of the state variables.
For example a sensor may measure position but not velocity.

\medskip
To summarize, the Kalman filter assumes that the initial belief is Gaussian and that both the state transition model and measurement model are linear with additive Gaussian noise. 
These assumptions guarantee that the belief distribution remains Gaussian after each prediction and measurement update. 
As a result, the algorithm only needs to propagate the mean $\bmu$ and covariance $\Sigma$ rather than a full probability distribution.
Therefore, while this property makes the Kalman filter computationally efficient, it also limits its applicability to systems that satisfy the assumptions of linearity and Gaussian noise.

\subsubsection{Algorithm (Predict--Correct Form)}
For clarity, we distinguish between the \emph{predicted} belief, obtained after applying the control but before incorporating the new measurement, and the \emph{corrected} belief, obtained after the measurement update. 
\cref{alg:KF} details the full algorithm, where we denote the predicted mean and covariance by $\bar{\mu}_t$ and $\bar{\Sigma}_t$.

\begin{algorithm}[tbh]
    \KwData{$\bmu_{t-1},\,\Sigma_{t-1},\,\u_t,\,\z_t$}
    \KwResult{$\bmu_t,\,\Sigma_t$}
    \tcp{Prediction (motion update)}
    $\bar{\bmu}_t \;\leftarrow\; A_t\,\bmu_{t-1} + B_t\,\u_t$\\
    $\bar{\Sigma}_t \;\leftarrow\; A_t\,\Sigma_{t-1}A_t^{\top} + \,\stateNoise_t$\\
    \tcp{Innovation (measurement residual)}
    $\tilde{\z}_t \;\leftarrow\; \z_t - C_t\,\bar{\bmu}_t$\\
    $S_t \;\leftarrow\; C_t\,\bar{\Sigma}_t\,C_t^{\top} + \measNoise_t$\\
    \tcp{Kalman gain}
    $K_t \;\leftarrow\; \bar{\Sigma}_t\,C_t^{\top}\,S_t^{-1}$\\
    \tcp{Correction (measurement update)}
    $\bmu_t \;\leftarrow\; \bar{\bmu}_t + K_t\,\tilde{\z}_t$\\
    \tcp{Covariance update (Joseph form for numerical stability)}
    $\Sigma_t \;\leftarrow\; (I-K_t C_t)\,\bar{\Sigma}_t\,(I-K_t C_t)^{\top} + K_t\,\measNoise_t\,K_t^{\top}$\\
    \Return $\bmu_t,\,\Sigma_t$
    \caption{Kalman Filter (linear-Gaussian)}
    \label{alg:KF}
   \end{algorithm}

\paragraph{Intuition.} The Kalman filter can be understood as a repeated negotiation between the model prediction and the sensor measurement. 
The prediction step propagates the belief through the system dynamics while adding process uncertainty.
The measurement update then adjusts this prediction using the newly observed measurement, where the amount of adjustment depends on how reliable the model prediction and the sensor reading are relative to one another.
Specifically, the residual $\tilde{\z}_t$ measures the discrepancy between the predicted measurement and the actual observation. 
The matrix $S_t$ describes the expected uncertainty of this residual and is known as the \emph{innovation covariance}. 
The \emph{Kalman gain} $K_t$ determines how strongly the estimate should respond to the measurement: when sensor noise is large, $K_t$ becomes small and the filter relies more heavily on the prediction; when the model uncertainty is large but the sensor is precise, $K_t$ increases and the estimate is pulled more strongly toward the measurement.

\subsubsection{Derivation}
One way to derive the Kalman filter algorithm is by explicitly evaluating the Bayes filter updates from \cref{ch:intro-to-localization} with the Gaussian belief structure and probabilistic transition and measurement models.
This would involve explicitly computing an integral of~$p(\x_t \given \x_{t-1}, \: \u_t) p(\x_{t-1})$ for the prediction step. 
Instead, we consider a more intuitive approach that directly leverages the properties of Gaussians presented in \cref{subsec:gaussian} to show that the familiar update equations in \cref{alg:KF} are an \emph{exact} consequence of the linear-Gaussian assumptions.

First, from the prior belief distribution,~$\bel(\x_{t-1}) \sim \mathcal{N}(\bmu_{t-1},\Sigma_{t-1})$, we compute the predicted belief,~$\overline{\bel}(\x_{t-1})$, by using the affine transformation property of Gaussian random variables and the property concerning the sum of two independent Gaussian random variables. 
Specifically, we apply these properties to the linear state transition model in \cref{eq:KFdynamics} to give the predicted mean:
\begin{equation*}
\begin{split}
\bar{\bmu}_t &= A_t \bmu_{t-1} + B_t \u_t + \bm{0}, \\
\end{split}
\end{equation*}
where the~$\bm{0}$ comes from the mean of the independent Gaussian process noise,~$\bm{\epsilon}_t \sim \mathcal{N}(\bm{0},\stateNoise_t)$. 
The predicted covariance is then:
\begin{equation*}
\begin{split}
\bar{\Sigma}_t &= A_t \Sigma_{t-1}A_t^\top  + \stateNoise_t. \\
\end{split}
\end{equation*}
For the measurement update step of the Bayes filter, we have:
\begin{equation*}
\begin{split}
\bel(\x_t) \propto p(\z_t \given \x_t)\overline{\bel}(\x_t), \\
\end{split}
\end{equation*}
where~$p(\z_t \given \x_t) \sim \mathcal{N}(C_t\x_t,\: \measNoise_t)$ and $\overline{\bel}(\x_t) \sim \mathcal{N}(\bar{\bmu}_{t},\: \bar{\Sigma}_{t})$.
We can therefore use the fact that the product of two Gaussians probability density functions is also a Gaussian probability density function to compute:
\begin{equation*}
\begin{split}
\bel(\x_t) = \eta\: \exp\big(-\frac{1}2 J_t \big),
\end{split}
\end{equation*}
where~$\eta$ is a normalization constant and:
\begin{equation*}
    J_t = (\z_t-C_t\x_t)^\top  \measNoise_t^{-1} (\z_t-C_t\x_t) + (\x_t-\bar{\bmu}_t)^\top  \bar{\Sigma}_t^{-1} (\x_t-\bar{\bmu}_t).
\end{equation*}
We compute the mean, $\bmu_t$, for this new probability density function by finding where the first derivative of $\bel(\x_t)$ with respect to $\x_t$ is zero, which occurs when the derivative of $J_t$ with respect to $\x_t$ is zero.
Similarly, we compute the new covariance, $\Sigma_t$, as the inverse of the second derivative of $J_t$ with respect to $\x_t$.
Therefore, we have the conditions:
\begin{equation*}
\begin{split}
0 &= -C_t^\top  \measNoise_t^{-1}(\z_t - C_t \bmu_t) + \bar{\Sigma}_t^{-1}(\bmu_t - \bar{\bmu}_t), \\   
\Sigma_t^{-1} &= C_t^\top  \measNoise_t^{-1} C_t + \bar{\Sigma}_t^{-1},
\end{split}
\end{equation*}
which give:
\begin{equation*}
\begin{split}
\Sigma_t &= (C_t^\top  \measNoise_t^{-1} C_t + \bar{\Sigma}_t^{-1})^{-1}.
\end{split}
\end{equation*}
Through algebraic manipulation, we write the mean in terms of the covariance~$\Sigma_t$:
\begin{equation*}
\begin{split}
\bmu_t &= \bar{\bmu}_t + \Sigma_t C_t^\top  \measNoise_t^{-1}(\z_t - C_t \bar{\bmu}_t). \\
\end{split}
\end{equation*}
With a few additional algebraic steps, we now transform these equations in the form of the Kalman filter equations in \cref{alg:KF}.
From the matrix inversion lemma, we write:
\begin{equation*}
(C_t^\top  \measNoise_t^{-1} C_t + \bar{\Sigma}_t^{-1})^{-1} = \bar{\Sigma}_t - \bar{\Sigma}_tC_t^\top (C_t\bar{\Sigma}_tC_t^\top  + \measNoise_t)^{-1}C_t\bar{\Sigma}_t,
\end{equation*}
and then we define the Kalman gain as~$K_t \coloneqq \bar{\Sigma}_{t}C_t^{\top}(C_t\bar{\Sigma}_{t}C_t^\top+\measNoise_t)^{-1}$ so that the covariance is given by:
\begin{equation*}
\begin{split}
\Sigma_t &= \bar{\Sigma}_t - K_tC_t\bar{\Sigma}_t.
\end{split}
\end{equation*}
Through some additional algebraic manipulations, we express the mean in terms of the Kalman gain to get:
\begin{equation*}
\bmu_t = \bar{\bmu}_t + K_t(\z_t-C_t\bar\bmu_{t}).
\end{equation*}
Further details on this derivation and the algebraic steps involved can be found in \citet{ThrunBurgardEtAl2005}.

\subsubsection{Practical Considerations}
The Kalman filter exploits the structure of the Gaussian distribution, which makes it a computationally efficient algorithm for filtering in a continuous state space.
However, the use of Gaussian beliefs also restricts the flexibility of the probabilistic model, since we have to assume the sufficiency of linear state transition and measurement models.
In practice, this linearity assumption may not be very accurate with respect to the real world behavior of the robot and sensors.
The structure also limits the belief distribution to be unimodal, which may limit performance in some applications\sidenote{For example, in robot localization tasks, a multimodal distribution can better capture the global distribution.}. 

Despite these limitations, Kalman filters are ubiquitous in robotics.
They are widely used for fusing inertial sensors, tracking moving objects, estimating velocities from position-only measurements, and many other tasks where the state is reasonably well modeled as evolving linearly with Gaussian noise.
In the next section, we introduce extensions to the Kalman filter that relax the linearity assumptions while retaining the Gaussian belief structure, which further expands the applicability of this family of filters to a wider range of robotics problems.

\subsection{Kalman Filter Extensions}
\label{subsec:kalman-filter-extensions}
The Kalman filter provides an exact recursive solution to the Bayes filter when the dynamics and measurement models are linear and all uncertainty is Gaussian.
In robotics, however, these assumptions are often only approximately true, since robot motion and sensing commonly involve nonlinear relationships.
To address this limitation, several extensions have been developed that preserve the Gaussian belief representation while accommodating nonlinear models. 

In this section, we introduce two of the most widely used approaches: the \emph{extended Kalman filter (EKF)} and the \emph{unscented Kalman filter (UKF)}.
The EKF handles nonlinearities by linearizing the dynamics and measurement models around the current belief mean, while the UKF propagates a carefully chosen set of sample points through the nonlinear models to capture how the belief distribution transforms.
Together, these methods provide practical and widely used tools for nonlinear state estimation when a unimodal Gaussian approximation remains appropriate.

\subsubsection{Extended Kalman Filter (EKF)}
\label{subsec:ekf}
The EKF generalizes the Kalman filter to nonlinear process and measurement models while retaining a Gaussian belief parameterization. 
It does so by locally linearizing the nonlinear models and then applying Kalman-style predict--correct updates to the mean and covariance. 
This makes the EKF a practical default for many robotics state estimation tasks with smooth nonlinear dynamics and sensing.

Instead of the linear models in \cref{eq:KFdynamics} and \cref{eq:KFmeasure} used by the Kalman filter, the EKF considers general nonlinear state transition and measurement models of the form:
\begin{equation}
\begin{split}
\x_t &= \dynmodel(\x_{t-1}, \u_t) + \bm{\epsilon}_{t}, \\
\z_t &= \measmodel(\x_t) + \bm{\delta}_{t},
\end{split}
\end{equation}
where~$\bm{\epsilon}_{t}\sim \mathcal{N}(\bm{0},\stateNoise_t)$ and $\bm{\delta}_{t}\sim \mathcal{N}(\bm{0},\measNoise_t)$ are normally distributed process and measurement noise terms.

The EKF incorporates these nonlinear models into the prediction and measurement update steps of the filter in two ways.
First, by evaluating the nonlinear models directly, and second, by linearizing them via a first-order Taylor series expansion.
We perform the first-order Taylor series expansion of the state transition model,~$\dynmodel(\x_{t-1}, \u_t)$, about the \emph{most likely state} from the current belief distribution, which is the expected value,~$\bmu_{t-1}$:
\begin{equation*}
\begin{split}
\dynmodel(\x_{t-1}, \u_t) &\approx \dynmodel(\bmu_{t-1}, \u_t) + \dynJac_t (\x_{t-1}-\bmu_{t-1}), \\
\end{split}
\end{equation*}
where $\dynJac_t = \nabla_{\x} \dynmodel(\bmu_{t-1}, \u_t)$ is the Jacobian of $\dynmodel(\x_{t-1}, \u_t)$ evaluated at $\bmu_{t-1}$. 
Using this linear approximation, we write the probabilistic state transition model as:
\begin{equation*}
p(\x_t \given \x_{t-1}, \u_t) = \frac{1}{\sqrt{\det(2\pi \stateNoise_t)}} \exp\big( -\frac{1}{2}\Delta \x_t^\top  \stateNoise_t^{-1} \Delta \x_t \big),
\end{equation*}
where:
\begin{equation*}
\Delta \x_t = \x_t - \dynmodel(\bmu_{t-1}, \u_t) - \dynJac_t (\x_{t-1}-\bmu_{t-1}).
\end{equation*}
The prediction step of the EKF leverages the nonlinear state transition model and the linearized model to update the mean and covariance as:
\begin{equation*}
\begin{split}
\bar{\bmu}_t &= \dynmodel(\bmu_{t-1}, \u_t), \\
\bar{\Sigma}_t &= \dynJac_t\Sigma_{t-1} \dynJac_t^\top + \stateNoise_t,
\end{split}
\end{equation*}
which exhibits a strong similarity to the Kalman filter prediction step.

We use a similar procedure for the measurement corrections.
Specifically, we approximate the measurement model using a first-order Taylor series expansion about the \emph{predicted} point,~$\bar{\bmu}_t$, to yield:
\begin{equation*}
\begin{split}
\measmodel(\x_t) \approx \measmodel(\bar{\bmu}_t) + \measJac_t (\x_{t}-\bar{\bmu}_{t}), \\
\end{split}
\end{equation*}
where~$\measJac_t = \nabla_{\x} \measmodel(\bar{\bmu}_{t})$ is the Jacobian of~$\measmodel(\x_t)$ evaluated at~$\bar{\bmu}_{t}$. 
We then write the probabilistic measurement model using this approximation as:
\begin{equation*}
p(\z_t \given \x_t) = \frac{1}{\sqrt{\det(2\pi \measNoise_t)}} \exp\big( -\frac{1}{2}\Delta \z_t^\top  \measNoise_t^{-1} \Delta \z_t \big),
\end{equation*}
where~$\Delta \z_t = \z_t-\measmodel(\bar{\bmu}_t) - \measJac_t (\x_{t}-\bar{\bmu}_{t})$. 
The measurement update step of the EKF uses the nonlinear measurement model and the linear approximation to compute:
\begin{equation*}
\begin{split}
\bmu_t &= \bar{\bmu}_t + K_t(\z_t - \measmodel(\bar{\bmu}_{t})), \\
\Sigma_t &= (I - K_t \measJac_t)\bar{\Sigma}_t,
\end{split}
\end{equation*}
where the Kalman gain is~$K_t = \bar{\Sigma}_{t}\measJac_t^\top(\measJac_t\bar{\Sigma}_{t}\measJac_t^\top+\measNoise_t)^{-1}$.
Again, we can see that this is very similar to the Kalman filter measurement update step.

We combine the EKF prediction and measurement update steps together in the overall EKF algorithm definition in \cref{alg:ekf}.
Compare this to \cref{alg:KF} and you will notice only the small difference that the EKF uses a combination of the nonlinear models and linear approximations from their Jacobians.

\begin{algorithm}[ht!]
    \KwData{$\bmu_{t-1},\,\Sigma_{t-1},\,\u_t,\,\z_t$}
    \KwResult{$\bmu_t,\,\Sigma_t$}
    \tcp{Linearize dynamics at $(\bmu_{t-1},\u_t)$}
    $\dynJac_t \leftarrow \partial \dynmodel/\partial \x \;\big|_{(\bmu_{t-1},\,\u_t)}$\\
    $L_t \leftarrow \partial \dynmodel/\partial \bm{\epsilon} \;\big|_{(\bmu_{t-1},\,\u_t)}$\\
    \tcp{Prediction}
    $\bar{\bmu}_t \leftarrow \dynmodel(\bmu_{t-1}, \u_t)$\\
    $\bar{\Sigma}_t \leftarrow \dynJac_t\,\Sigma_{t-1}\,\dynJac_t^\top \;+\; L_t\,\stateNoise_t\,L_t^\top$\\
    \tcp{Linearize measurement at $\bar{\bmu}_t$}
    $\measJac_t \leftarrow \partial \measmodel/\partial \x \;\big|_{\bar{\bmu}_t}$\\
    \tcp{Innovation and gain}
    $\tilde{\z}_t \leftarrow \z_t - \measmodel(\bar{\bmu}_t)$\\
    $S_t \leftarrow \measJac_t\,\bar{\Sigma}_t\,\measJac_t^\top + \measNoise_t$\\
    $K_t \leftarrow \bar{\Sigma}_t\,\measJac_t^\top\,S_t^{-1}$\\
    \tcp{Correction}
    $\bmu_t \leftarrow \bar{\bmu}_t + K_t\,\tilde{\z}_t$\\
    \tcp{Joseph-form covariance update (numerically robust)}
    $\Sigma_t \leftarrow (I-K_t \measJac_t)\,\bar{\Sigma}_t\,(I-K_t \measJac_t)^\top + K_t\,\measNoise_t\,K_t^\top$\\
    \Return $\bmu_t,\,\Sigma_t$
    \caption{Extended Kalman Filter (EKF)}
    \label{alg:ekf}
\end{algorithm}

\paragraph{Intuition.} 
The EKF carries out a first-order approximation of the nonlinear models around the mean estimate, which is the most plausible operating point. 
The Kalman filter formulas then apply to this locally linear surrogate. 
The quality of the update hinges on the local linearity of~$\dynmodel$ and~$\measmodel$ around the chosen linearization points and on the fidelity of the noise covariances.
If the robot frequently operates in regions where the models are strongly nonlinear over the uncertainty region, or where uncertainty is large, the linearization may be poor and the EKF can become inconsistent or even diverge.

\paragraph{Practical considerations.}
The EKF provides more accurate results than the Kalman filter in many applications due to its ability to consider more general nonlinear models. 
However, because it relies on a first-order Taylor expansion, it can perform poorly or diverge if the models are highly nonlinear over the region of uncertainty or if the linearization point is far from the true state.
The EKF also inherits the unimodal Gaussian belief representation from the Kalman filter and thus cannot represent multi-hypothesis uncertainty.

In practice, successful EKF deployments rely on good initializations, careful tuning of the noise covariances, and models that are only mildly nonlinear over the region of interest.

\subsubsection{Unscented Kalman Filter (UKF)}
\label{subsec:ukf-essentials}
The UKF improves upon a core weakness of the EKF that first-order linearization can poorly capture how nonlinear transformations distort a Gaussian belief. 
Instead of linearizing the nonlinear functions $\dynmodel$ and $\measmodel$, the UKF approximates the belief distribution itself.
It does so by propagating a carefully chosen set of deterministically sampled points called \emph{sigma points} through the true nonlinear functions, and then recomputing the mean and covariance from the transformed points.

Intuitively, if we start from a Gaussian belief, we can summarize it with a small cloud of representative points that capture its shape.
If we pass those points through the nonlinear dynamics and sensing models, we can then reconstruct a new mean and covariance that more accurately reflect how the nonlinearity distorts the belief.
This idea is formalized by the \emph{unscented transform}.

\paragraph{Unscented transform (UT).}
Consider an~$n$-dimensional Gaussian~$\mathcal{N}(\bmu,\Sigma)$ and a nonlinear function~$g(\cdot)$.
The UT constructs a deterministic set of~$2n+1$ sigma points $\{\mathcal{X}^{(i)}\}$ that capture the mean and covariance of the Gaussian:

\begin{equation*}
\mathcal{X}^{(0)}=\bmu,\qquad
\mathcal{X}^{(i)}=\bmu + \bm{c}_i,\qquad
\mathcal{X}^{(i+n)}=\bmu - \bm{c}_i,\;\; i=1,\dots,n,
\end{equation*}
where the vectors~$\bm{c}_i$ are the columns of a matrix square root of~$(n+\lambda)\Sigma$ (for example, a Cholesky factor).
The scalar~$\lambda$ controls the spread of the sigma points via tuning parameters $(\alpha,\beta,\kappa)$, with:
\begin{equation*}
\lambda=\alpha^2(n+\kappa)-n.
\end{equation*}
Each sigma point is assigned an associated weight for the mean and covariance,~$\{W_m^{(i)}, W_c^{(i)}\}$.
A common choice is:
\begin{equation*}
W_m^{(0)} = \frac{\lambda}{n + \lambda},\qquad
W_c^{(0)} = \frac{\lambda}{n + \lambda} + (1- \alpha^2 + \beta),
\end{equation*}
and for $i>0$:
\begin{equation*}
W_m^{(i)} = W_c^{(i)} = \frac{1}{2(n+\lambda)}.
\end{equation*}
Here,~$\alpha\in(0,1]$ controls how far the sigma points spread from the mean,~$\beta$ incorporates prior knowledge about the distribution, and~$\kappa$ is an additional scaling parameter.
To apply the UT to a nonlinear transformation $g(\cdot)$, we simply pass the sigma points through $g$:
\begin{equation*}
\mathcal{Y}^{(i)}=g(\mathcal{X}^{(i)}),
\end{equation*}
and then reconstruct the mean and covariance of the transformed variable as:
\begin{equation*}
\bar{\y}=\sum_i W_m^{(i)}\mathcal{Y}^{(i)}, \qquad
P_{yy}= \sum_i W_c^{(i)}\big(\mathcal{Y}^{(i)}-\bar{\y}\big)\big(\mathcal{Y}^{(i)}-\bar{\y}\big)^\top.
\end{equation*}

The cross-covariance between~$\x$ and~$\y=g(\x)$ is computed analogously.
This cross-covariance quantifies how deviations in the state around~$\bmu$ correlate with deviations in the transformed quantity around~$\bar{\y}$, and it is exactly the object needed to form the Kalman gain in the UKF.

\paragraph{Accuracy of the UT.}
For Gaussian priors, the UT matches the mean and covariance of~$\y = g(\x)$ to at least second order\sidenote{For certain choices of parameters, can be accurate to third order.} in a Taylor expansion of~$g$, without evaluating any Jacobians. 
In contrast, the EKF's linearization is only first-order accurate. 
Intuitively, the symmetry of the sigma points around~$\bmu$ causes many first- and second-order terms in the Taylor series to cancel in the weighted sums, leaving high-order terms as the dominant approximation error. 
We refer the reader to Julier and Uhlmann~\citep{julier1997new} for a detailed analysis.

From a practical point of view, the UT gives us a plug-and-play tool.
We can feed in a Gaussian belief and a nonlinear function, and it returns an updated Gaussian that better reflects the nonlinear mapping than a first-order Taylor series.

\begin{algorithm}[ht!]
    \KwData{$\bmu_{t-1},\,\Sigma_{t-1},\,\u_t,\,\z_t,\,(\alpha,\,\beta,\,\kappa)$}
    \KwResult{$\bmu_t,\,\Sigma_t$}
    \tcp{Build sigma point set from $(\bmu_{t-1},\Sigma_{t-1})$}
    $\lambda \leftarrow \alpha^2(n+\kappa)-n$\\
    $\mathcal{X}^{(0)} \leftarrow \bmu_{t-1}$\\
    $W_m^{(0)} \leftarrow \frac{\lambda}{n + \lambda}$\\
    $W_c^{(0)} \leftarrow \frac{\lambda}{n + \lambda} + (1- \alpha^2 + \beta)$\\
    $W_m^{(>0)} \leftarrow W_c^{(>0)} \leftarrow \frac{1}{2(n + \lambda)}$\\
    $\bm{C} \leftarrow \text{Cholesky}((n+\lambda)\Sigma_{t-1})$\\
    \For{$i = 1$ \KwTo $n$}{
    		$\mathcal{X}^{(i)} \leftarrow \bmu_{t-1} + \bm{c}_i$\\
    		$\mathcal{X}^{(i+n)} \leftarrow \bmu_{t-1} - \bm{c}_i$\\
    }
    \tcp{Propagate sigma points through dynamics}
    \For{$i = 0$ \KwTo $2n$}{
        $\mathcal{Y}^{(i)} \leftarrow \dynmodel(\mathcal{X}^{(i)}, \u_t)$\\
    }
    \tcp{Predicted mean and covariance}
    $\bar{\bmu}_t \leftarrow \sum_i W_m^{(i)}\mathcal{Y}^{(i)}$\\
    $P_{ff}\leftarrow \sum_i W_c^{(i)}\big(\mathcal{Y}^{(i)}-\bar{\bmu}_t\big)\big(\mathcal{Y}^{(i)}-\bar{\bmu}_t\big)^\top$\\
    $\bar{\Sigma}_t \leftarrow P_{ff} + \stateNoise_t$\\
    \tcp{Propagate sigma points through measurement model}
    \For{$i = 0$ \KwTo $2n$}{
        $\mathcal{Z}^{(i)} \leftarrow \measmodel(\mathcal{Y}^{(i)})$\\
    }
    \tcp{Predicted measurement mean and covariances}
    $\bar{\mathbf{z}} \leftarrow \sum_i W_m^{(i)} \mathcal{Z}^{(i)}$\\
    $P_{hh} \leftarrow \sum_i W_c^{(i)}\big(\mathcal{Z}^{(i)} - \bar{\mathbf{z}}\big)\big(\mathcal{Z}^{(i)}-\bar{\mathbf{z}}\big)^\top$\\
    $P_{fh} \leftarrow \sum_i W_c^{(i)}\big(\mathcal{Y}^{(i)} - \bar{\bmu}_t\big)\big(\mathcal{Z}^{(i)}-\bar{\mathbf{z}}\big)^\top$\\
    \tcp{Compute gain and correction}
    $S_t \leftarrow P_{hh} + \measNoise_t$\\
    $K_t \leftarrow P_{fh} S_t^{-1}$\\
    $\bmu_t \leftarrow \bar{\bmu}_t + K_t(\z_t-\bar{\mathbf{z}})$\\
    \tcp{Covariance update}
    $\Sigma_t \leftarrow \bar{\Sigma}_t - K_t S_t K_t^\top$\\
    \Return $\bmu_t,\,\Sigma_t$\\
    \caption{Unscented Kalman Filter (UKF)}
    \label{alg:ukf}
\end{algorithm}

\paragraph{UKF recursion (additive noise).}
\cref{alg:ukf} can be read as a Kalman filter where the prediction and measurement steps are implemented via the unscented transform.
For the common case with additive process and measurement noise, the UKF proceeds as follows:
\begin{enumerate}
    \item \emph{Sigma point generation:} from the current Gaussian belief $(\bmu_{t-1},\Sigma_{t-1})$, construct a set of sigma points and weights using the UT.
    \item \emph{Dynamics propagation:} pass each sigma point through the nonlinear state transition model $\dynmodel(\cdot,\u_t)$ to obtain predicted points.
    Compute the predicted mean $\bar{\bmu}_t$ and covariance $\bar{\Sigma}_t$ by weighted averaging, and add the process noise covariance $\stateNoise_t$.
    \item \emph{Measurement prediction:} apply the measurement model $\measmodel(\cdot)$ to each predicted sigma point.
    Compute the predicted measurement mean $\hat{\z}_t$, the measurement covariance $S_t$, and the cross-covariance between state and measurement, $P_{fh}$.
    \item \emph{Update:} form the Kalman gain $K_t = P_{fh} S_t^{-1}$, update the mean via $\bmu_t = \bar{\bmu}_t + K_t(\z_t-\hat{\z}_t)$, and update the covariance $\Sigma_t = \bar{\Sigma}_t - K_t S_t K_t^\top$.
\end{enumerate}
For non-additive noise models, we typically augment the state vector with noise variables and construct sigma points in this augmented space.

\paragraph{EKF vs.\ UKF: when to use each.} The choice between EKF and UKF depends on the specific problem characteristics, including the degree of nonlinearity, the availability and reliability of Jacobians, and computational constraints.
\begin{itemize}
    \item The \emph{EKF} algorithm is appropriate when the dynamics and measurement models~$f$ and~$h$ are mildly nonlinear, their Jacobians are straightforward to compute and accurate, and the computational budget is tight. 
    The EKF is \emph{first-order accurate} in the sense that it relies on a linear approximation of~$f$ and~$h$ around the mean and can be biased when nonlinearities are strong over the uncertainty region.
    \item The \emph{UKF} algorithm excels when nonlinearities in the dynamics or sensing are substantial, derivatives are difficult to obtain or unreliable, or measurements are strongly nonlinear (for example, bearing-only measurements). 
    For Gaussian priors, the UKF is \emph{second-order accurate} in the mean and covariance, and often higher, and typically captures the belief's evolution more accurately than the EKF in these regimes, at the cost of evaluating the dynamics and measurement models at~$2n+1$ sigma points per step.
\end{itemize}

\begin{example}[Range--bearing update (nonlinear sensing)]
\label{ex:range_bearing}
To see the EKF and UKF in action, consider a planar robot whose state is its pose~$\x_t=[p_x,\,p_y,\,\theta]^\top$ and that observes a fixed landmark at position~$\ell=[\ell_x,\ell_y]^\top$.
The robot carries a sensor that measures the range and bearing to the landmark:
\begin{equation*}
\measmodel(\x_t)=
\begin{bmatrix}
\sqrt{(\ell_x-p_x)^2+(\ell_y-p_y)^2}\\[6pt]
\operatorname{atan2}(\ell_y-p_y,\;\ell_x-p_x)-\theta
\end{bmatrix}.
\end{equation*}
Both the square root and the $\operatorname{atan2}$ function make this a strongly nonlinear measurement model, especially when the robot is close to the landmark or uncertain in orientation.

\paragraph{EKF approach.}
In the EKF, we handle this nonlinearity by linearizing~$\measmodel$ around the predicted mean~$\bar{\bmu}_t=[\bar{p}_x,\,\bar{p}_y,\,\bar{\theta}]^\top$.  
The resulting Jacobian:
\begin{equation*}
\measJac_t = \frac{\partial h}{\partial \x}\Big|_{\bar{\bmu}_t},
\end{equation*}
captures how small changes in pose affect the range and bearing.
Intuitively, moving the robot towards the landmark shortens the range, while rotating the robot changes the bearing.
The measurement innovation is:
\begin{equation*}
\tilde{\z}_t = \z_t - \measmodel(\bar{\bmu}_t),
\end{equation*}
with the bearing residual wrapped into~$(-\pi,\pi]$ to avoid jumps across the angle discontinuity.
The standard EKF correction step then uses~$\measJac_t$, $\measNoise_t$, and the predicted covariance to compute the Kalman gain and update the state.

Algorithm~\ref{alg:range_bearing_ekf} shows a Python implementation of this EKF update.
The structure closely mirrors the math: predict, linearize, compute the innovation and its covariance, compute the gain, and correct.

\begin{listing}[]
\begin{tcolorbox}[colback=gray!10, colframe=gray!50,
    title=Range-Bearing Robot EKF Update, boxrule=0.5mm, arc=0mm]
\begin{minted}[escapeinside=||]{python}
def ekf_update(prior_mean, prior_cov, z, R, landmark):
    # Prediction
    F = robot_jacobian(prior_mean)
    pred_mean = robot_dynamics(prior_mean)
    pred_cov = F @ prior_cov @ F.T

    # Innovation and gain
    H = range_bearing_jacobian(prior_mean, landmark)
    z_hat = range_bearing(pred_mean, landmark)
    innov = np.array([z[0] - z_hat[0], 
                      wrap_angle(z[1] - z_hat[1])])
    S = H @ prior_cov @ H.T + R
    K = pred_cov @ H.T @ np.linalg.inv(S)

    # Correction
    mean = pred_mean + K @ innov
    I = np.eye(3)
    cov = (I - K @ H) @ pred_cov @ (I - K @ H).T + K @ R @ K.T
    mean[2] = wrap_angle(mean[2])
    return mean, cov
\end{minted}
\end{tcolorbox}
\caption{EKF update for the robot with a range-bearing sensor described in \cref{ex:range_bearing}.
The code for this example is available in the repository \colorcode{github.com/StanfordASL/pora-exercises} in the notebook \colorcode{ch12/ekf\_ukf\_range\_bearing.ipynb}.} 
\label{alg:range_bearing_ekf}
\end{listing}

\paragraph{UKF approach.}
The UKF handles the same problem by avoiding Jacobians altogether.
Starting from the current Gaussian belief, it constructs sigma points in the pose space, propagates them through the nonlinear motion model and the nonlinear range--bearing sensor model, and then recomputes the predicted mean, covariance, and cross-covariance from the transformed sigma points.

Because the UKF sees the full curvature of the measurement function through these propagated points, it can more accurately capture how the range--bearing observation tightens or shifts the belief, especially in regimes where the EKF's linear approximation is poor.

Algorithm~\ref{alg:range_bearing_ukf} provides a Python implementation of the UKF update for this example.
Its structure is analogous to the EKF code, but with explicit sigma-point generation, propagation, and weighted recombination in place of Jacobian calculations.

\begin{listing}[]
\begin{tcolorbox}[colback=gray!10, colframe=gray!50,
    title=Range-Bearing Robot UKF Update, boxrule=0.5mm, arc=0mm]
\begin{minted}[escapeinside=||]{python}
def ukf_update(prior_mean, prior_cov, z, R, landmark):
    # Compute sigma points and predicted mean/covariance
    X, Wm, Wc = compute_sigma_points(prior_mean, prior_cov)
    Y = np.array([robot_dynamics(x) for x in X])
    pred_mean = np.zeros(3)
    pred_mean[0] = np.sum(Wm * Y[:,0])
    pred_mean[1] = np.sum(Wm * Y[:,1])
    pred_mean[2] = angle_mean(Wm, Y[:,2])

    # Compute predicted measurement
    Z = np.array([range_bearing(y, landmark) for y in Y])
    pred_z = np.zeros(2)
    pred_z[0] = np.sum(Wm * Z[:,0])
    pred_z[1] = angle_mean(Wm, Z[:,1])

    # Compute predicted covariances
    Pff = np.zeros((3,3))
    Phh = np.zeros((2,2))
    Pfh = np.zeros((3,2))
    for i in range(Y.shape[0]):
        dz = np.array([Z[i,0] - pred_z[0], 
                       wrap_angle(Z[i,1] - pred_z[1])])
        dy = Y[i] - pred_mean
        dy[2] = wrap_angle(dy[2])
        Pff += Wc[i] * np.outer(dy, dy)
        Phh += Wc[i] * np.outer(dz, dz)
        Pfh += Wc[i] * np.outer(dy, dz)

    # Compute gain and correction
    pred_cov = Pff
    S = Phh + R
    K = Pfh @ np.linalg.inv(S)
    innov = np.array([z[0] - pred_z[0], 
                      wrap_angle(z[1] - pred_z[1])])
    mean = pred_mean + K @ innov
    mean[2] = wrap_angle(mean[2])
    cov = pred_cov - K @ S @ K.T
    return mean, cov
\end{minted}
\end{tcolorbox}
\caption{UKF update for the robot with a range-bearing sensor described in \cref{ex:range_bearing}.
The code for this example is available in the repository \colorcode{github.com/StanfordASL/pora-exercises} in the notebook \colorcode{ch12/ekf\_ukf\_range\_bearing.ipynb}.} 
\label{alg:range_bearing_ukf}
\end{listing}

\paragraph{Comparing the EKF and UKF.}
The two filters have the same high-level structure: prediction, measurement prediction, computation of an innovation and its covariance, formation of a Kalman gain, and correction of the mean and
covariance.
The key differences are:
\begin{enumerate}
  \item The EKF uses Jacobians ($F_t,H_t$) and linear formulas for the
        prediction and update, whereas the UKF uses explicit sigma-point
        generation, propagation, and weighted recombination.
  \item The UKF recomputes predicted covariances and cross-covariances by
        weighted outer products of transformed sigma points instead of using
        linear covariance propagation.
\end{enumerate}

\paragraph{Discussion.}
The EKF and UKF are two sides of the same \emph{Gaussian belief} coin. 
The EKF linearizes the models and keeps the belief exact as a Gaussian, while the UKF keeps the models exact and approximates the belief via sigma points.
Both are grounded in the Bayes filter and both maintain a single Gaussian over the state.

This example also highlights when the extra effort of the UKF is worthwhile.
If the robot's pose uncertainty is small and the landmark is far away, the measurement model is nearly linear locally and the EKF performs well at lower computational cost.
If the robot is close to the landmark, has substantial heading uncertainty, or the geometry is strongly nonlinear in other ways, the UKF more faithfully captures how the observation reshapes the belief.

In the repository \colorcode{github.com/StanfordASL/pora-exercises}, the notebook\\\noindent\colorcode{ch12/ekf\_ukf\_range\_bearing.ipynb} visualizes the EKF and UKF updates for this problem and lets you experiment with different levels of nonlinearity and uncertainty.
\end{example}

\subsection{Non-parametric Filters: From Grids to Particles}
\label{subsec:nonparametric}
Parametric filters gain efficiency by committing to a fixed belief shape\sidenote{As discussed earlier in this chapter, usually a Gaussian distribution.}. 
This commitment is powerful when the world behaves roughly as assumed, but it can be too rigid when uncertainty is multimodal, the dynamics or sensing are strongly nonlinear, or data association is ambiguous. 

\emph{Non-parametric} filters remove this structural assumption and approximate the belief directly.
They typically do so either by:
\begin{enumerate}
    \item discretizing the state space into bins (histogram filters), or
    \item representing the belief with samples (particle filters).
\end{enumerate}
Both methods are direct approximations of the Bayes filter and trade additional computation for representational flexibility.

Up to this point, we have mostly assumed that beliefs can be summarized by a single Gaussian.
In many robotics problems this is not realistic\sidenote{For example, in global localization a mobile robot may initially have no idea which part of a building it is in, leading to several widely separated hypotheses.}.
A unimodal Gaussian cannot represent such a situation well because it spreads probability mass between the plausible hypotheses rather than concentrating it at those locations.
Non-parametric filters are designed to address precisely this kind of scenario.

In the remainder of this section, we introduce two widely used non-parametric filters that build on the Bayes filter from \cref{ch:intro-to-localization}.
We first discuss the \emph{histogram filter} in \cref{subsec:histogram-filter}, which discretizes the state space, and then the \emph{particle filter} in \cref{subsec:particle-filter}, which represents the belief with samples.

\subsubsection{Histogram Filter}
\label{subsec:histogram-filter}
The histogram filter is a non-parametric filter that can be viewed as an extension of the discrete Bayes filter from \cref{ch:intro-to-localization} to continuous state spaces. 
It proceeds by discretizing the continuous state space into a finite number of regions and then representing the belief distribution as a set of probabilities over these regions.
Conceptually, we overlay a grid on the state space and store one number per grid cell representing the probability that the state lies in that cell.

Mathematically, for the random state vector~$\bX$, we discretize the continuous state space,~$\statespace$, into a finite set of regions, or \emph{bins}, such that:
\begin{equation*}
\statespace = b_{1} \cup b_{2} \cup \ldots \cup b_{K},
\end{equation*}
where~$b_{k}$ is the $k$-th bin and~$K$ is the total number of bins.
For example, if a one-dimensional state variable takes values in the interval $[0,1]$, the interval can be divided into a set of equally spaced sub-intervals.
The belief distribution $\bel(\x_t)$ is then represented by assigning a probability mass $p_{k,t}$ to each bin $b_k$. 
This quantity represents the probability that the state $\x_t$ lies inside bin $b_k$ at time $t$.
The histogram representation can also be interpreted as a piecewise-constant probability density function, where the density is constant within each bin and given by:
\begin{equation*}
p(\x_t) = \frac{p_{k,t}}{\lvert b_k \rvert}, \quad \x_t \in b_k,
\end{equation*}
where~$\lvert b_k \rvert$ denotes the area or volume of the bin.\sidenote{By construction $\sum_k p_{k,t} = 1$. In practice, the probabilities are renormalized after the correction step to maintain this property.}

To connect the histogram filter to the underlying continuous models, it is useful to define the binwise transition and likelihood that the filter approximates.
Let the continuous transition and measurement models be~$p(\x_t\mid \x_{t-1},\u_t)$ and~$p(\z_t\mid \x_t)$, respectively.
We define the corresponding binwise quantities as:
\begin{align}
    T^{(t)}_{k i}
    &\coloneqq p(\x_t\in b_k \mid \x_{t-1}\in b_i, \u_t)
    = \frac{1}{|b_i|}\!\int_{\x_{t-1}\in b_i}\!\int_{\x_t\in b_k} p(\x_t\mid \x_{t-1},\u_t)\,\d\x_t\,\d\x_{t-1},
    \label{eq:hist-precise-T}
    \\
    L^{(t)}_{k}
    &\coloneqq p(\z_t\mid \x_t\in b_k)
    = \frac{1}{|b_k|}\int_{\x_t\in b_k} p(\z_t\mid \x_t)\,\d\x_t.
    \label{eq:hist-precise-L}
\end{align}
In principle this formulation is exact, however, in practice, directly evaluating the integrals is usually intractable.

A common approximation associates each bin $b_k$ with a representative state defined as the bin mean:
\begin{equation}
    \hat{\x}_{k} \definedas \frac{1}{\lvert b_k \rvert} \int_{b_k} \x \: \d \x.
\end{equation}

Using these mean states, we approximate the binwise transition model~$p(b_{k,t} \given b_{i,t-1}, \: \u_t)$ by:
\begin{equation}
    p(b_{k,t} \given b_{i,t-1}, \: \u_t) \approx \eta \lvert b_k \rvert\, p(\hat{\x}_{k,t} \given \hat{\x}_{i,t-1}, \: \u_t),
\end{equation}
where~$p(\hat{\x}_{k,t} \given \hat{\x}_{i,t-1}, \: \u_t)$ is the original (continuous) state transition model evaluated at the mean bin states, and $\eta$ is a normalization constant.\sidenote{If the bin areas~$\lvert b_k \rvert$ are equal, we can absorb this term into the normalization constant~$\eta$. For higher fidelity, we may sample multiple points per bin and average.}
We discretize the probabilistic measurement model in a similar way:
\begin{equation}
    p(\z_t \given b_{k,t}) \approx p(\z_t \given \hat{\x}_{k,t}),
\end{equation}
so that the measurement probability associated with bin~$b_k$ is approximated by the measurement probability at the representative state~$\hat{\x}_{k}$.\sidenote{For angular variables such as headings, ensure the representative state respects periodicity and wrap residuals appropriately.}

Once we have discretized the state space with bins~$b_k$ and approximated the transition and measurement models using the bin mean states, the histogram filter follows the same structure as the discrete Bayes filter in \cref{alg:discretebayes}.
We summarize it in \cref{alg:histogramfilter}.

\begin{algorithm}[tb]
    \KwData{$\{p_{k,t-1}\}, \u_{t}, \z_{t}$}
    \KwResult{$\{p_{k,t}\}$}
    \ForEach{k}{
     $\overline{p}_{k,t} \leftarrow \sum_{i} p(b_{k,t} \given b_{i,t-1}, \: \u_t)\, p_{i,t-1} $\\
     $p_{k,t} \leftarrow p(\z_t \given b_{k,t}) \, \overline{p}_{k,t}$\\
    }
    \tcp{Normalization (to enforce $\sum_k p_{k,t}=1$; use log-weights if probabilities are very small)}
    $\eta \leftarrow \left(\sum_k p_{k,t}\right)^{-1}$\\
    \ForEach{k}{ $p_{k,t} \leftarrow \eta\, p_{k,t}$ }
    \Return $\{p_{k,t}\}$
    \caption{Histogram Filter}
    \label{alg:histogramfilter}
\end{algorithm}

A few implementation insights make the histogram filter more practical:
\begin{itemize}
    \item The transition matrix is typically \emph{sparse}. Most motion models move probability only to nearby bins. Rather than looping over all bin pairs $(i,k)$, it is often more efficient to loop over each $b_i$ and distribute its mass to a small set of neighboring $b_k$.
    \item If~$p(\x_t\mid \x_{t-1},\u_t)$ is shift-invariant\sidenote{For example, additive Gaussian motion in a grid.}, the prediction step becomes a discrete \emph{convolution}. In such cases the computation can be accelerated using separable kernels or fast Fourier transforms in one- or two-dimensional grids.
    \item At domain boundaries, boundary conditions must be chosen to match the physical system. Common choices include reflecting, absorbing, or wrap-around boundaries.\sidenote{For example, orientation variables on $S^1$ are naturally periodic and therefore use wrap-around boundaries.}
\end{itemize}

Like the discrete Bayes filter, the main disadvantage of the histogram filter is that it can become computationally intractable when the number of bins grows large. 
This occurs when high spatial resolution is required for accuracy or when the state space is high dimensional.
For example, in a robot localization problem where we estimate a planar pose $(x,y,\theta)$, discretizing $x$ and $y$ at $1$ meter resolution in a modestly sized building and $\theta$ at $1^\circ$ resolution can easily yield hundreds of thousands of bins.\sidenote{Cost scales roughly as $O(K\,n_\text{nbr})$ per step, where $n_\text{nbr}$ is the number of motion-neighbor bins (often small). Memory is $O(K)$. This “curse of dimensionality” is a primary motivation for particle filters.}

\begin{example}[One-dimensional motion with Gaussian noise + range sensing]
    Consider a robot moving along a line segment~$[0,L]$. 
    We discretize this interval into~$K$ equal bins of width $\Delta=L/K$, with bin $k$ spanning $[x_k^-,x_k^+]$ and centroid $\hat{x}_k=(k-\tfrac{1}{2})\Delta$. 
    The robot follows a simple additive motion model and receives a noisy range measurement to the wall at the origin:
    \begin{equation*}
    x_t = x_{t-1}+u_t+\epsilon_t,\quad \epsilon_t\sim\mathcal{N}(0,\sigma_u^2),
    \qquad
    z_t = x_t+\nu_t,\quad \nu_t\sim\mathcal{N}(0,\sigma_z^2).
    \end{equation*}
    
    \paragraph{Prediction.}
    After applying control~$u_t$, the robot’s position distribution is shifted and blurred according to motion noise. 
    For each pair of bins~$(i,k)$, the probability of moving from bin~$i$ to bin~$k$ can be approximated by the Gaussian mass over~$b_k$ centered at~$\hat{x}_i+u_t$:
    \begin{equation*}
    T^{(t)}_{k i} \approx 
    \Phi\!\Big(\tfrac{x_k^+-(\hat{x}_i+u_t)}{\sigma_u}\Big) -
    \Phi\!\Big(\tfrac{x_k^--(\hat{x}_i+u_t)}{\sigma_u}\Big),
    \end{equation*}
    where $\Phi$ is the standard normal CDF. 
    The predicted belief $\overline{p}_{k,t}$ is obtained by summing these contributions over all bins $i$.
    
    \paragraph{Correction.}
    Given a measurement $z_t$, each bin is reweighted according to how likely its centroid $\hat{x}_k$ is under the measurement model:
    \begin{equation*}
    L^{(t)}_k \approx \mathcal{N}\!\big(z_t;\,\hat{x}_k,\sigma_z^2\big).
    \end{equation*}
    The posterior is then $p_{k,t}\propto L^{(t)}_k\,\overline{p}_{k,t}$, followed by normalization so that $\sum_k p_{k,t}=1$.
    
    \paragraph{Discussion.}
    This process naturally captures both unimodal and multimodal beliefs. 
    Starting from a uniform prior, the filter may, for some time, maintain several peaks if the measurement is ambiguous\sidenote{For example, if the environment has repeated structures.}. 
    As more controls and measurements accumulate, inconsistent modes gradually lose probability, and the posterior collapses to a single sharp peak near the true location. 
    In this way, the histogram filter provides a simple yet powerful tool for global localization in low-dimensional settings.
\end{example}

\subsubsection{Particle Filter}
\label{subsec:particle-filter}
The \emph{particle filter} is a non-parametric filter that is often more computationally tractable than the histogram filter for continuous, higher-dimensional state spaces. 
Rather than discretizing the state space a priori, it represents the belief distribution by a finite set of samples from the state space, called \emph{particles}.\sidenote{The particle filter is sometimes referred to as a \emph{Monte Carlo} algorithm due to its sampling-based nature.}
The key idea is to place more particles in regions of high probability and fewer in regions of low probability so that the particle set adapts to the shape of the belief.

We define the set of particles at time~$t$ as:
\begin{equation}
    \particleset_t \coloneqq \{\x_t^{[1]}, \x_t^{[2]},\ldots, \x_t^{[K]}\},
\end{equation}
where~$\x_t^{[k]}$ is the $k$-th particle.
Each particle~$\x_t^{[k]}$ represents a hypothesis about the true state~$\x_t$, and regions of the state space with more particles correspond to regions of higher probability.
Ideally, the particles are distributed according to the current belief:
\begin{equation*}
    \x_t^{[k]} \sim \bel(\x_t),
\end{equation*}
but in practice this holds only approximately for finite $K$.\sidenote{As $K\rightarrow\infty$, the empirical distribution of the particles converges to the true belief under mild conditions. In many applications, on the order of $K\approx 10^3$ particles already provides useful approximations.}

Following the Bayes filter paradigm, the particle filter updates the prior belief, represented by~$\particleset_{t-1}$, via a prediction step and a measurement update step.

\paragraph{Prediction (sampling through the dynamics).}
For each particle~$\x_{t-1}^{[k]}$ in the prior set, we draw a new sample from the state transition model:
\begin{equation*}
	\bar{\x}_{t}^{[k]} \sim p(\x_t \given \x_{t-1}^{[k]}, \u_t).
\end{equation*}
This step plays the role of the motion update where the cloud of particles is pushed forward according to the control and process noise.
The resulting set~$\{\bar{\x}_t^{[k]}\}$ approximates the predicted belief~$\overline{\bel}(\x_t)$.

\paragraph{Measurement weighting and resampling.}
We next assess how well each predicted particle is supported by the new measurement~$\z_t$.
For each predicted particle, we compute a weight:
\begin{equation*}
	w_t^{[k]} = p(\z_t \given \bar{\x}_t^{[k]}),
\end{equation*} 
so that particles that are more consistent with the measurement receive larger weights.
We collect the predicted particles and their weights into a weighted set~$\bar{\particleset}_t = \{(\bar{\x}_t^{[k]}, w_t^{[k]})\}$, which approximates the unnormalized posterior.

\begin{algorithm}[t]
    \KwData{$\particleset_{t-1}, \u_{t}, \z_{t}$}
    \KwResult{$\particleset_{t}$}
    $\bar{\particleset}_{t} \leftarrow \emptyset$\\
    \For{$k=1$ \KwTo $K$}{
     Sample $\bar{\x}_{t}^{[k]} \sim p(\x_t\given \x_{t-1}^{[k]}, \u_t)$\\
     $w_t^{[k]} \leftarrow p(\z_t \given \bar{\x}_{t}^{[k]})$\\
     $\bar{\particleset}_{t} \leftarrow \bar{\particleset}_{t} \cup \big(\bar{\x}_{t}^{[k]}, w_t^{[k]} \big)$\\
    }
    \tcp{Resampling step (with replacement) according to weights}
    $\particleset_{t} \leftarrow \emptyset$\\
    \For{$k=1$ \KwTo $K$}{
     Draw index $i$ with probability $\propto w_t^{[i]}$\\
     $\particleset_{t} \leftarrow \particleset_{t} \cup \{ \bar{\x}_{t}^{[i]} \}$\\
    }
    \Return $\particleset_t$
    \caption{Particle Filter}
    \label{alg:particle}
\end{algorithm}
The particle filter's measurement update step then consists of \emph{resampling} (with replacement) a new set of~$K$ particles from~$\bar{\particleset}_t$, according to the normalized weights~$w_t^{[k]}$.
Particles with large weights are likely to be selected many times, while particles with negligible weights may disappear.
The resampled set~$\particleset_t$ approximates the updated belief~$\bel(\x_t)$.

\begin{figure}[ht!]
\centering
\includegraphics[width=0.8\linewidth]{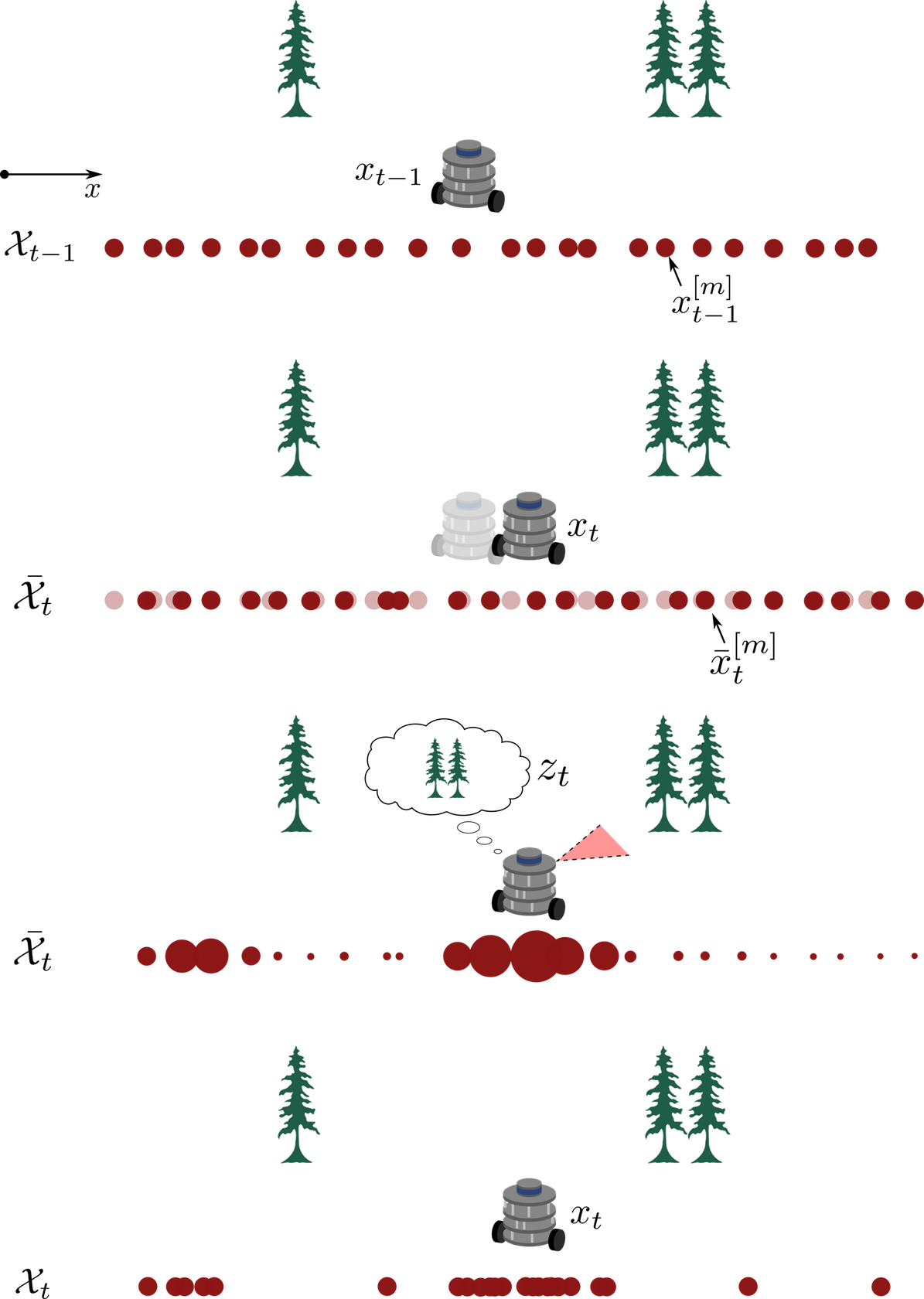}
\caption{A particle filter used for robot localization. 
We first update the initial set of particles according to the transition model, and then weight them according to the observation. 
Finally, we generate a new set of particles through weighted resampling.}
\label{fig:Particle_filter}
\end{figure}

We summarize the algorithm in \cref{alg:particle} and illustrate a few iterations of the particle filter for a simple robot localization problem in \cref{fig:Particle_filter}.

The resampling step is important for more than just incorporating the measurement.
Without resampling, repeated multiplication of weights would gradually concentrate probability on a tiny subset of particles, while most particles would have negligible weight and effectively be wasted.
This phenomenon is known as \emph{particle degeneracy}.
Resampling combats degeneracy by repeatedly discarding low-weight particles and replicating high-weight particles, keeping the effective sample size roughly constant.

This effect has a nice interpretation in evolutionary terms: particles that explain the data well ``survive and reproduce'', while those that do not are gradually removed.
From a computational perspective, resampling focuses particles in high-probability regions of the state space and reduces the number of particles required for a given level of accuracy.

\section{Summary}
In this chapter, we examined practical approximations to the Bayes filter for continuous and high-dimensional state estimation problems in robotics. 
We began by motivating the need for tractable filtering methods and introduced two broad families of approximate filters, parametric and non-parametric, each with its own strengths and weaknesses.

Parametric filters, such as the Kalman filter and its nonlinear extensions, the EKF and the UKF, represent the belief as a Gaussian distribution, enabling efficient recursive updates of mean and covariance. 
We discussed how these filters exploit linear or locally linear models to achieve computational efficiency, along with their limitations in representing multimodal or highly nonlinear beliefs.
    
We then introduced non-parametric filters, including the histogram and particle filters, which relax parametric assumptions and directly approximate the belief distribution through discretization or sampling. 
These methods trade additional computation for representational flexibility, making them effective in situations involving global uncertainty, nonlinear dynamics, or non-Gaussian noise. 

\begin{table}[t]
    \centering
    \caption{Parametric vs.\ Non-parametric Filters (at a glance)}
    \label{tab:pf_compare}
    \begin{tabular}{@{}p{2.5cm}p{5cm}p{5cm}@{}}
    \toprule
     & \textbf{Parametric (KF / EKF / UKF)} & \textbf{Non-parametric (Histogram / Particle)} \\
    \midrule
    Belief & Single Gaussian (mean, covariance) & Grid masses or weighted samples \\
    Nonlinearity & EKF (linearize), UKF (sigma points) & Native; no Jacobians needed \\
    Multimodality & Poor (unimodal) & Natural (multi-peak) \\
    Dimensionality & Scales well with state dim & Histogram: suffers; Particle: scalable with $K$ \\
    Computation & Cheap per step & Histogram: grows with bins; Particle: $O(K)$ \\
    When it shines & Smooth models, near-Gaussian noise, good observability & Ambiguity, strong nonlinearities, non-Gaussian noise, global localization \\
    Pitfalls & Linearization bias (EKF), covariance inconsistency & Degeneracy without resampling; sample impoverishment \\
    \bottomrule
    \end{tabular}
\end{table}
\cref{tab:pf_compare} summarizes the main characteristics of these approaches and highlights their complementary strengths. 
In practice, robotic systems often combine elements of both families in order to balance accuracy, robustness, and computational efficiency.

\paragraph{High-level decision guide.}
It is useful to relate common robotics scenarios to appropriate filter architectures. 
Kalman-style filters are well suited for local tracking problems with smooth motion and reliable sensors, whereas histogram or particle filters are better suited for global localization, ambiguous data association, or highly nonlinear environments.
\begin{enumerate}
    \item If the robot's dynamics or sensing models are mildly nonlinear and beliefs stay near unimodal $\Rightarrow$ \emph{EKF}.
    \item If the Jacobians of the dynamics or sensor models are hard to compute or their nonlinearities are significant $\Rightarrow$ \emph{UKF} (or square-root UKF).
    \item If there is global or multimodal uncertainty, or severe non-Gaussian noise $\Rightarrow$ \emph{Particle filter}.
    \item If the state is low-dimensional and a map or grid is available $\Rightarrow$ \emph{Histogram filter}.
\end{enumerate}

\paragraph{To learn more.}
For a comprehensive and accessible treatment of probabilistic state estimation and filtering methods, readers are encouraged to consult the classic text by~\citet{ThrunBurgardEtAl2005}. 
Detailed derivations and discussions of Kalman filtering theory can also be found in~\citet{maybeck1982stochastic}, while~\citet{julier1997new} provide the seminal introduction to the Unscented Kalman Filter. 
For further reading on non-parametric filters and their applications to robot localization, see~\citet{dellaert1999monte}. 
A modern perspective that unifies filtering, smoothing, and mapping under a probabilistic framework is presented in~\citet{slam-handbook}, which connects the foundations of state estimation to contemporary SLAM and spatial-AI systems.

\section{Exercises}
The starter code for the exercises provided below is available online through GitHub. 
To get started, download the code by running in a terminal window:

\begin{tcolorbox}[colback=gray!10]
\begin{minted}{bash}
    git clone https://github.com/StanfordASL/pora-exercises.git
\end{minted}
\end{tcolorbox}

We denote Problems requiring hand-written solutions and coding in Python with \adjustbox{height=2ex, valign=c}{\includegraphics{figs/write.png}} and \adjustbox{height=2ex, valign=c}{\includegraphics{figs/code.png}}, respectively.

\subsection*{\adjustbox{height=2ex, valign=c}{\includegraphics{figs/code.png}}\ Problem 1: Kalman Filter for Landmark Localization}
In this exercise, you will implement a Kalman Filter to localize a set of fixed landmarks given knowledge about a robot's motion.
Specifically, consider an environment where there are four landmarks, and we define the state of landmark positions as the vector:
\begin{equation*}
\begin{split}
\x_t^{m} = \begin{bmatrix}
x_t^{m_1} & y_t^{m_1} & x_t^{m_2} & y_t^{m_2} & x_t^{m_3} & y_t^{m_3} & x_t^{m_4} & y_t^{m_4}
\end{bmatrix}^\top,
\end{split}
\end{equation*}
which we assume we can directly measure (i.e. $\z_t = \x_t$) with a noisy sensor.

In the file \colorcode{ch12/exercises/kalman\_filter.ipynb}, perform the following tasks:
\begin{enumerate}
\item Define the state transition matrix $A$ from \cref{eq:KFdynamics} and observation matrix $C$ from \cref{eq:KFmeasure} for the Kalman filter algorithm.
The matrix $A$ should correspond to the motion of the landmarks, which we assume are stationary, and $C$ should represent the direct measurement of the landmark positions. 
\item Implement the Kalman filter predict and update steps to compute the means and covariances at each time step.
\item Run the provided code to see the results.
\end{enumerate}

\newpage
\printbibliography[segment=\therefsegment,heading=subbibliography,title={References}]
\chapter{Robot Localization}
\label{ch:robot-localization}
\newrefsegment
The filtering algorithms developed in \cref{ch:intro-to-localization}  and \cref{ch:approximate-filters} provide general tools for estimating the state of a dynamical system from noisy measurements. 
In mobile robotics, one of the most important instances of this general problem is \emph{localization}: estimating a robot's pose with respect to a map of the environment.

In this chapter, we specialize the Bayesian filtering framework to the robot localization problem. 
We begin in \cref{subsec:localization-taxonomy} with a taxonomy that organizes different localization scenarios---pose tracking versus global localization, static versus dynamic environments, active versus passive sensing, and single-robot versus multi-robot settings. 
This taxonomy provides a mental model for the kinds of problems that arise in practice.

In \cref{subsec:loc-bayes}, we then express localization formally as a Bayesian filtering problem in which the state is the robot pose and the map is an additional, known variable that influences both motion and measurements. 
This leads to map-aware state transition and measurement models, which we discuss in \cref{sec:map-transition}.

Building on this formulation, in \cref{subsec:markov-loc} we introduce \emph{Markov localization}, a general Bayes filter for estimating a robot's pose that explicitly incorporates a map. Markov localization is conceptually clean but rarely implemented directly; instead, it serves as a reference from which more specialized algorithms are derived.

We then present two specializations of Markov localization. 
In \cref{subsec:ekf-localization}, we develop an EKF localization algorithm for feature-based maps that are well suited to pose tracking with moderate uncertainty. 
Finally, in \cref{subsec:monte-carlo-localization} we introduce Monte Carlo Localization (MCL), a particle filter–based method that can handle global localization and strongly multimodal beliefs. 
Along the way, we discuss data association and practical considerations that arise in map-based localization.

\subsection{A Taxonomy of Robot Localization Problems}
\label{subsec:localization-taxonomy}

\begin{table}[tb]
    \centering
    \caption{A taxonomy of robot localization problems along four key axes.}
    \label{tab:loc-taxonomy}
    \begin{tabular}{p{2.5cm}p{4cm}p{5.5cm}}
      \toprule
      Axis & Representative cases & Typical algorithmic implications \\
      \midrule
      Initial pose & Pose tracking (local), global localization, kidnapped robot &
      Tracking often amenable to unimodal Gaussian filters; global and kidnapped
      scenarios typically require non-parametric, multimodal beliefs (e.g.,
      particle filters). \\[0.3em]
      Environment & Static, slowly varying, strongly dynamic &
      Static maps simplify modeling; dynamic elements motivate robust measurement
      models and possibly explicit dynamic object tracking. \\[0.3em]
      Action selection & Passive, active (information-seeking) &
      Passive use treats controls as given; active localization couples the
      filter with planning methods that trade off task progress and information
      gain. \\[0.3em]
      Number of robots & Single robot, multi-robot (cooperative) &
      Multi-robot settings introduce shared maps and inter-robot observations,
      leading to coupled estimation problems and opportunities for improved
      robustness. \\
      \bottomrule
    \end{tabular}
  \end{table}

Robot localization problems can differ substantially depending on how much is known about the initial pose, how the environment evolves, how actions are chosen, and how many robots are involved. 
Before turning to specific algorithms, it is helpful to organize these variations using a taxonomy.

We will use the term \emph{pose} to refer to the robot's position and orientation in a global coordinate frame. 
For a planar mobile robot, the pose is typically~$x_t = (x_t, y_t, \theta_t)$ at time~$t$, where $(x,y)$ is the planar position and $\theta$ is the heading angle.

\paragraph{Pose tracking and global pose localization.}
One of the first distinctions among robot localization problems concerns what we assume we know about the initial pose:

\begin{itemize}
    \item \emph{Pose tracking (local localization).} In pose tracking problems,
    the initial pose is known with reasonably small uncertainty. For example, a
    robot may be placed at a known charging station or docking location, with
    a prior belief concentrated in a small neighborhood. The goal is to maintain
    an accurate estimate of pose over time as the robot moves and senses, despite
    process and measurement noise.
  
    \item \emph{Global localization.} In global localization, the robot has very
    little prior information about its initial pose. For instance, it may know
    only that it is somewhere within a building, but not which floor, corridor,
    or room. The belief over poses must therefore represent multiple plausible
    hypotheses, often spread over a large region of the map.
  
    \item \emph{Kidnapped robot problem.} A challenge related to global localization arises when a
    robot that has been successfully tracking its pose is suddenly transported
    to a different, unknown location without its sensors or localization
    algorithm being aware of the event. This \emph{kidnapped robot problem}
    requires the localization method to recover from a grossly incorrect belief
    and reinitialize globally.
  \end{itemize}


Pose tracking problems are often amenable to unimodal Gaussian approximations, such as EKF-based methods, whereas global localization and kidnapped robot scenarios typically require non-parametric methods capable of representing multiple hypotheses, such as histogram or particle filters.



\paragraph{Static and dynamic environment localization.}
A second axis in the taxonomy of robot localization concerns how the environment changes over time:

\begin{itemize}
    \item \emph{Static environments.} In the simplest case, the environment map
    is fixed and does not change. Walls and landmarks remain in place, and any
    moving objects (such as people) are ignored or treated as noise. Many
    foundational localization algorithms are developed under this assumption.
  
    \item \emph{Dynamic environments.} In more realistic settings, parts of the
    environment change over time. Examples include doors that open and close,
    furniture that moves, or other agents that occupy the same space. In these
    cases, the map may be time varying or may explicitly model dynamic objects.
  \end{itemize}

In this chapter, we primarily consider localization with respect to a static map, while allowing for noisy measurements that may occasionally be corrupted by dynamic elements. 
Handling fully dynamic maps is closely related to SLAM and tracking, and will be revisited in \cref{ch:slam}.


\paragraph{Active and passive localization.}

Localization algorithms also differ in how they interact with the environment:

\begin{itemize}
    \item \emph{Passive localization.} In passive localization, the algorithm
    treats the sequence of controls $u_{1:t}$ as given; it does not attempt to
    choose actions to improve localization performance. Many navigation systems
    fall into this category when localization is treated as a background process.
  
    \item \emph{Active localization.} In active localization, the robot chooses
    its actions with the explicit goal of reducing uncertainty about its pose. For
    example, the robot may move to viewpoints that disambiguate map symmetries
    or collect measurements that are expected to be highly informative.
  \end{itemize}

The algorithms developed in this chapter apply in both settings. 
When used in an active framework, they provide the state and uncertainty estimates needed for planning information-seeking actions.



\paragraph{Single and multi-robot localization.}
Finally, localization problems can involve one or multiple robots:

\begin{itemize}
    \item \emph{Single-robot localization.} A single robot must estimate its
    pose using on-board sensors and possibly an external reference, such as GPS,
    relative to a map that we assume is given.
  
    \item \emph{Multi-robot localization.} In multi-robot settings, several
    robots localize simultaneously, often sharing information. Robots may observe
    each other, exchange measurements, or maintain a shared map. This introduces
    additional structure and opportunities for cooperation, but also coupling
    between the individual localization problems.
  \end{itemize}

Throughout this chapter, we focus on single-robot localization in a known, largely static map. 
The methods we develop form the basis for more advanced multi-robot and SLAM systems.

\subsection{Robot Localization via Bayesian Filtering}
\label{subsec:loc-bayes}
In previous chapters, we introduced several well-known variations of the Bayes filter, including the parametric EKF and the non-parametric particle filter in \cref{ch:approximate-filters}.
These algorithms propagate a belief distribution over a state $\x_t$ using a probabilistic Markov state transition model and a probabilistic measurement model.
Recall that for a general state-space model, the belief at time $t$ is defined as:
\begin{equation}
  \bel(\x_t) \definedas p(\x_t \mid \z_{1:t}, \u_{1:t}),
  \label{eq:loc-bel-def}
\end{equation}
where $\z_{1:t}$ are all measurements up to time $t$ and $\u_{1:t}$ are the applied controls.

In map-based localization, the state~$\x_t$ represents the robot pose, and we assume that a map~$\m$ of the environment is given. 
The map encodes information about the environment that constrains both how the robot can move and what it can measure. 
Our goal is to compute the posterior over robot poses conditioned on the map:
\begin{equation*}
  \bel(\x_t) \definedas p(\x_t \mid \z_{1:t}, \u_{1:t}, \m).
\end{equation*}

To incorporate the map into the filtering framework, we modify both the state transition model and the measurement model.
Specifically, in \cref{sec:map-transition}, we introduce a \emph{map-aware} state transition model $p(\x_t \mid \x_{t-1}, \u_t, \m)$ that captures the fact that some poses are inconsistent with the environment.
Then, in \cref{sec:map-measurement}, we introduce a \emph{map-aware} measurement model $p(\z_t \mid \x_t, \m)$ that reflects how local sensor readings depend on nearby structures in the map.
The combination of these two models leads directly to the localization algorithms we introduce in \cref{subsec:ekf-localization} and \cref{subsec:monte-carlo-localization}. 

  \paragraph{Map representation.}
  Before modifying the probabilistic models, it is useful to briefly review how we represent maps in this chapter. 
  We assume that the environment can be described by a collection of spatial entities---either discrete landmarks or volumetric cells---with associated properties.
Formally, we denote the map as:
\begin{equation*}
    \m = \{ m_1, m_2, \dots, m_N \},
\end{equation*}
  where each element~$m_i$ describes a portion of the environment. 
  
  Depending on the application, $m_i$ may represent the occupancy state of a grid cell (occupied/free/unknown), the position of a point landmark, $m_i = (m_{i,x}, m_{i,y})$ in a global frame, or more complex attributes such as semantic labels or reflectivity.
  We will use the generic term ``map element'' for $m_i$, and refer informally to these as ``objects'' or ``cells'' when helpful.

  \begin{figure}[t]
    \centering
    \includegraphics[width=.75\linewidth]{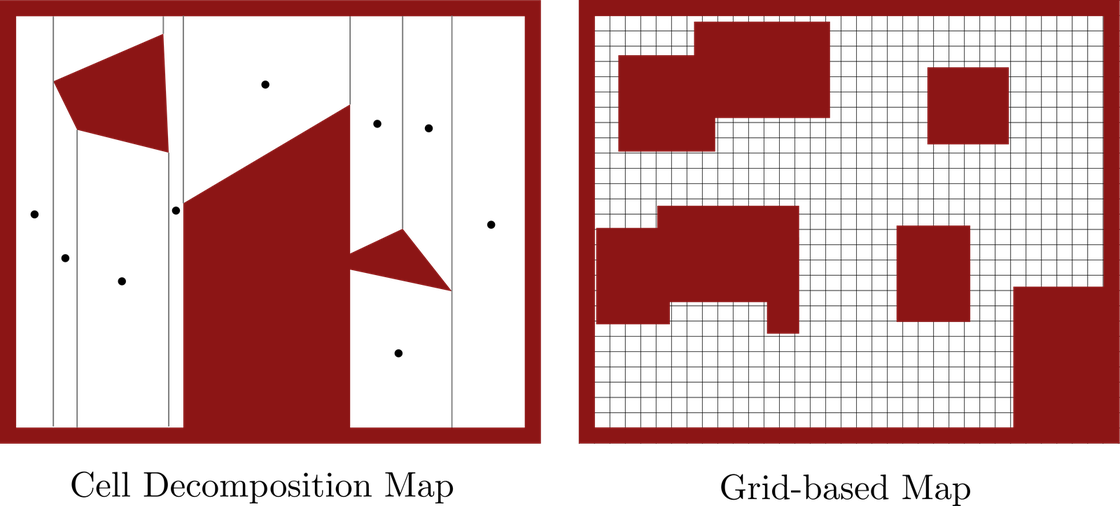}
    \caption{Two examples of location-based maps. 
    Both represent the map as a set of volumetric objects, which in these examples are cells.}
    \label{fig:LocationBasedMaps}
    \end{figure}
    \begin{figure*}[t]
        \centering
        \includegraphics[width=0.8\linewidth]{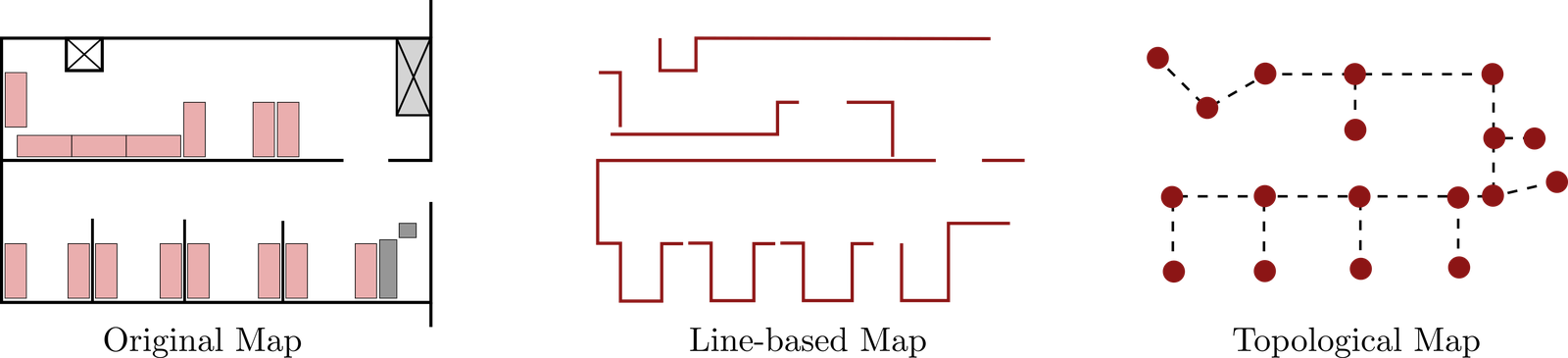}
        \caption{Two examples of feature-based maps. 
        One uses a collection of lines and the other uses a graph representation of the empty spaces.}
        \label{fig:FeatureBasedMaps}
    \end{figure*}

  Two important map families arise frequently in mobile robotics:
  \begin{itemize}
    \item \emph{Feature-based maps}, where each $m_i$ is a compact feature, such
    as a point landmark or line segment, with precisely defined geometry.
    \cref{fig:FeatureBasedMaps} shows two examples of feature-based maps, one represented by lines and another by a graph of nodes and edges.\sidenote{Graph-based maps are generally referred to as \emph{topological maps}.}
    \item \emph{Cell-based (occupancy grid) maps}, where the environment is
    discretized into a grid of cells, each with an associated occupancy
    probability. 
    This is closely related to the grid-based motion planning representations discussed in \cref{subsec:grid-based-planning}.
    \cref{fig:LocationBasedMaps} shows examples where the environment is represented by a set of volumetric cells.
  \end{itemize}

Cell-based maps trade off resolution and computational cost: smaller cells provide finer spatial detail but increase memory usage and computation time, while larger cells reduce cost but blur detail and can misrepresent narrow passages. 
We refer back to the grid-based planning discussion in \cref{subsec:grid-based-planning} for a deeper treatment of these trade-offs. 
In this chapter, we treat the choice of map representation as given, and focus on how the map enters the filtering equations.

\subsection{Map-Aware State Transition Model}
\label{sec:map-transition}
The map influences how the robot can move. 
For example, a mobile robot cannot pass through walls or leave the known workspace. 
A purely kinematic transition model~$p(\x_t \mid \x_{t-1}, \u_t)$ does not capture these constraints: it might assign nonzero probability to states that lie inside obstacles or outside the mapped area.

To incorporate the map, we consider the map-aware transition model~$p(\x_t \mid \x_{t-1}, \u_t, \m)$. 
Directly modeling this distribution for arbitrary maps is difficult, so we adopt an approximation that reuses the original transition model and a separate, map-induced prior. 
Specifically, we define~$p(\x_t \mid \m)$ as the probability of a pose given only the map. 
This term characterizes how consistent a pose is with the environment. 
For example,~$p(\x_t \mid \m) = 0$ for a pose inside a wall or outside the mapped region.

Using Bayes' rule, we can approximate the map-aware transition model as:
\begin{equation}
  p(\x_t \mid \x_{t-1}, \u_t, \m)
  \approx \eta \,
    \frac{p(\x_t \mid \x_{t-1}, \u_t)\, p(\x_t \mid \m)}{p(\x_t)},
  \label{eq:map-transition}
\end{equation}
where $\eta$ is a normalization constant that ensures the left-hand side integrates to one. 
\cref{eq:map-transition} can be derived step-by-step by starting from Bayes' rule:
\begin{equation*}
  p(\x_t \mid \x_{t-1}, \u_t, \m)
  = \frac{p(\m \mid \x_t, \x_{t-1}, \u_t)\, p(\x_t \mid \x_{t-1}, \u_t)}
         {p(\m \mid \x_{t-1}, \u_t)}.
\end{equation*}
Since~$p(\m \mid \x_{t-1}, \u_t)$ does not depend on~$\x_t$, we can absorb it into a constant $\eta'$:
\begin{equation*}
  p(\x_t \mid \x_{t-1}, \u_t, \m)
  = \eta' \, p(\m \mid \x_t, \x_{t-1}, \u_t)\, p(\x_t \mid \x_{t-1}, \u_t).
\end{equation*}

Next, we assume that the influence of the past on the map is captured entirely through the current pose~$\x_t$:
\begin{equation*}
  p(\m \mid \x_t, \x_{t-1}, \u_t) \approx p(\m \mid \x_t).
\end{equation*}

Intuitively, once we know the current pose, the specific path taken to get there does not provide additional information about the static map.\footnote{%
This approximation becomes more accurate as the time step between $t-1$ and $t$ shrinks, so that the robot moves only a short distance between successive poses.}
With this assumption:
\begin{equation*}
  p(\x_t \mid \x_{t-1}, \u_t, \m)
  \approx \eta' \, p(\m \mid \x_t) \, p(\x_t \mid \x_{t-1}, \u_t).
\end{equation*}
Applying Bayes' rule again to $p(\m \mid \x_t)$ gives:
\begin{equation*}
  p(\m \mid \x_t) = \frac{p(\x_t \mid \m) p(\m)}{p(\x_t)}.
\end{equation*}
Substituting into the previous expression and absorbing constant factors that do not depend on $\x_t$ into a new normalization constant $\eta$ yields:
\begin{equation*}
  p(\x_t \mid \x_{t-1}, \u_t, \m)
  \approx \eta \, \frac{p(\x_t \mid \x_{t-1}, \u_t)\, p(\x_t \mid \m)}{p(\x_t)},
\end{equation*}
which is Equation~\eqref{eq:map-transition}.

In this approximation, $p(\x_{t} \given \m )$ is the probability of a state given only the map and captures the \emph{geometric consistency} of that state with respect to the environment\sidenote{For example, $p(\x_{t} \given \m ) = 0$ for a state $\x_t$ that lies inside a wall or outside the known map.}.
The factor $p(\x_{t}\given \x_{t-1}, \u_t)$ encodes the motion model, and $p(\x_t \given \m)$ acts as a map-based correction that suppresses physically implausible states.

From a computational perspective, Equation~\eqref{eq:map-transition} is attractive because it allows us to reuse existing implementations of the motion model and incorporate the map through a relatively inexpensive correction:
\begin{itemize}
  \item We first generate a \emph{kinematic prediction} of the new pose using
  $p(\x_t \mid \x_{t-1}, \u_t)$, which depends only on the robot model and control.

  \item We then reweight or prune these predictions using $p(\x_t \mid \m)$,
  which often reduces to simple operations such as checking whether the pose
  lies in free space (for occupancy grids) or near a corridor (for semantic
  maps).
\end{itemize}
Thus, the map-aware transition model modifies the likelihood of proposed poses without requiring a complete redesign of the underlying motion model.

\subsubsection{Map-Aware Measurement Model}
\label{sec:map-measurement}
The map also strongly influences sensor measurements. 
For instance, the range returned by a lidar beam depends on where that beam first intersects an obstacle in the map, and the bearing to a visual landmark depends on the landmark's position in the global frame. 
To reflect this dependence, we introduce a map-aware measurement model:
\begin{equation*}
  p(\z_t \mid \x_t, \m).
\end{equation*}

In many systems, a measurement vector $\z_t \in \mathbb{R}^\outputdim$ consists of individual components, such as individual range beams or landmark observations.
A common simplifying assumption is that the individual components of the measurement vector $\z_t \in \R^\outputdim$ are conditionally independent given the state and map.
Under this assumption, we can factor the measurement model as:
\begin{equation} 
\label{eq:condind}
    p(\z_{t} \given \x_{t}, \m) = \prod_{i=1}^{\outputdim}p(\z_{t}^{i}\given \x_{t}, \m).
\end{equation}

The conditional independence assumption is not exact.
For example, nearby range beams may be correlated if they hit the same object.
However, this assumption greatly simplifies inference and works well in many practical settings. 
It allows us to process multiple sensor readings either in a batch or sequentially, using the same underlying model $p(\z_t^i \mid \x_t, \m)$ for each component.

\subsection{Markov Localization}
\label{subsec:markov-loc}
The first map-based localization algorithm we introduce is \emph{Markov localization}.
Markov localization applies the Bayes filter from \cref{ch:intro-to-localization} to the map-aware models introduced above. 
The belief over robot pose at time $t$ is:
\begin{equation*}
  \bel(\x_t) \definedas p(\x_t \mid \z_{1:t}, \u_{1:t}, \m),
\end{equation*}
and the map $\m$ is treated as known and fixed.

Using the map-aware transition and measurement models~$p(\x_t \mid \x_{t-1}, \u_t, \m)$ and~$p(\z_t \mid \x_t, \m)$, the Bayes filter recursion becomes:
\begin{align*}
  \text{Prediction:} \qquad 
  \belpred(\x_t)
  &= \int p(\x_t \mid \x_{t-1}, \u_t, \m)\, \bel(\x_{t-1}) \, \d\x_{t-1}, \\
  \text{Correction:} \qquad
  \bel(\x_t)
  &= \eta \, p(\z_t \mid \x_t, \m)\, \belpred(\x_t),
\end{align*}
where $\belpred(\x_t)$ denotes the predicted belief and $\eta$ is a normalization constant chosen so that $\int \bel(\x_t)\, \d\x_t = 1$.
We summarize this recursion in \cref{alg:markov-loc}.

\begin{algorithm}[ht]
    \KwData{$\bel(\x_{t-1}), \u_{t},\z_{t}, \m$}
    \KwResult{$\bel(\x_{t})$}
    \ForEach{$\x_t$}{
       $\overline{\bel}(\x_t) = \int p(\x_t\given \x_{t-1}, \u_{t}, \m) \bel(\x_{t-1}) \d\x_{t-1}$ \\
       $\bel(\x_t) = \eta p(\z_t\given \x_{t},\m)\overline{\bel}(\x_t)$
    }
    \Return $\bel(\x_{t})$
    \caption{Markov Localization}
    \label{alg:markov-loc}
\end{algorithm}

   Conceptually, Markov localization has the same structure as the Bayes filter from \cref{ch:intro-to-localization}: a prediction step that propagates the belief through the motion model, followed by a correction step that incorporates the latest measurement.
   The only difference is that both the prediction and correction now depend explicitly on the map $\m$.

   In its most general form, however, Markov localization is not directly computationally tractable. 
   The integral over all possible poses and the need to maintain an arbitrary belief function are prohibitive in high-dimensional or continuous state spaces. 
   As in \cref{ch:approximate-filters}, the key to practicality is to choose a \emph{representation} for the belief that is expressive enough for the problem at hand, yet structured enough to admit efficient computation.

   In \cref{subsec:ekf-localization}, we adopt a unimodal Gaussian belief and derive an EKF-based localization algorithm, which is well suited for pose tracking with moderate uncertainty. 
   In \cref{subsec:monte-carlo-localization}, we adopt a particle-based representation and derive Monte Carlo Localization, capable of handling global localization and multimodal beliefs.

\subsection{EKF Localization}
\label{subsec:ekf-localization}
We now develop an EKF-based realization of Markov localization. 
The EKF localization algorithm assumes that the belief over the robot's pose can be approximated by a single Gaussian:
\begin{equation*}
\bel(\x_t) \sim \mathcal{N}(\bmu_t, \Sigma_t),
\end{equation*}
and applies the EKF prediction and update equations from \cref{ch:approximate-filters} using map-aware models. 
This Gaussian structure significantly improves computational efficiency relative to the full Markov localization algorithm, at the cost of not being able to represent multiple well-separated pose hypotheses.\footnote{%
As in \cref{ch:approximate-filters}, this unimodality assumption is well suited to pose tracking but generally insufficient for global localization.}

We assume the same nonlinear state transition model as in \cref{subsec:ekf}:
\begin{equation*}
    \x_t = \dynmodel(\x_{t-1}, \u_t) + \bm{\epsilon}_t,
    \end{equation*}
where $\bm{\epsilon}_t \sim \mathcal{N}(\bm{0}, \stateNoise_t)$ is zero-mean Gaussian process noise.
The corresponding state transition Jacobian is:
\begin{equation}
	\dynJac_t = \nabla_{\x}\dynmodel(\bmu_{t-1}, \u_t),
\end{equation}
where $\bmu_{t-1}$ is the mean of the previous belief $\bel(\x_{t-1})$.

The main difference between the general EKF and EKF localization is the presence of a feature-based map and a measurement model that relates robot poses to observed landmarks.
We assume we have a map $\m$ of $N$ point landmarks:
\begin{equation*}
\m = \{m_1, m_2, \dots, m_N\}, \quad m_j = (m_{j,x}, m_{j,y}),
\end{equation*}
where each landmark $m_j$ is given by its two-dimensional location $(m_{j,x}, m_{j,y})$ in the global coordinate frame.
At time $t$ the robot obtains a set of landmark measurements:
\begin{equation*}
  \z_t = \{ \z_t^1, \z_t^2, \dots \},
\end{equation*}
where each $\z_t^i$ is associated (implicitly or explicitly) with one landmark.

Given a pose $\x_t$ and the index $j$ of the corresponding landmark, the measurement model is:
\begin{equation*}
  \z_t^i = \measmodel(\x_t, j, \m) + \bm{\delta}_t,
\end{equation*}
where $\bm{\delta}_t \sim \mathcal{N}(\bm{0}, \measNoise_t)$ models zero-mean Gaussian sensor noise, and $\measmodel(\cdot)$ encodes the expected range, bearing, or other features to landmark $m_j$. 
For each measurement, the Jacobian of the measurement model with respect to the state is:
\begin{equation}
  H^{c^i_t}_t = \nabla_{\x}h(\bar{\bmu}_t,c^i_t,\m),
  \label{eq:H-ekf-loc}
\end{equation}
where $\bar{\bmu}_t$ is the predicted mean from the EKF prediction step, and $c_t^i$ is the index of the map feature associated with measurement $i$.

A new challenge in feature-based localization is \emph{data association}. 
Given a set of measurements $\z_t$, which landmark does each measurement correspond to?
We denote the correspondence for measurement $i$ at time $t$ by $c_t^i \in \{1, \dots, N+1\}$, where $c_t^i = j$ means that measurement $i$ corresponds to landmark $m_j$, and $c_t^i = N+1$ indicates that measurement $i$ does not correspond to any known landmark\sidenote{For example, due to a spurious detection.}.

We first consider the simpler case where we assume the correspondences $c_t^i$ are known.

\subsubsection{EKF Localization with Known Correspondences}
\label{sec:ekf-known-corr}
Assume for the moment that the correspondences $\bc_t = \{ c_t^1, c_t^2, \dots \}$ are known. 
The EKF localization algorithm then resembles the standard EKF from \cref{ch:approximate-filters}, with two key modifications:
\begin{enumerate}
  \item The measurement model relates the pose to map features via $\measmodel(\x_t, j, \m)$.
  \item Multiple landmark measurements are processed at each time step.
\end{enumerate}

Given the pose $\x_t$ and the map $\m$, and given known correspondences $\bc_t$, the joint likelihood of the measurement set $\z_t$ can be written as:
\begin{equation*}
  p(\z_t \mid \x_t, \bc_t, \m).
\end{equation*}
Using the conditional independence assumption from \cref{eq:condind}---now applied at the level of individual landmark measurements---we have:
\begin{equation*}
  p(\z_t \mid \x_t, \bc_t, \m)
  = \prod_i p(\z_t^i \mid \x_t, c_t^i, \m).
\end{equation*}
Each factor $p(\z_t^i \mid \x_t, c_t^i, \m)$ is a Gaussian distribution whose mean is given by $\measmodel(\x_t, c_t^i, \m)$ and whose covariance is $\measNoise_t$.

Because the prior over $\x_t$ is Gaussian and each measurement likelihood term is Gaussian, the posterior remains Gaussian. 
Moreover, under the conditional independence assumption, we can apply the standard EKF measurement update sequentially for each $\z_t^i$ and obtain the same result as if we had processed all measurements in a single stacked update. 
This is a consequence of the fact that, for Gaussian models, multiplication of likelihood terms can be carried out in any order.

\begin{algorithm}[tb]
 \KwData{$\bmu_{t-1}, \Sigma_{t-1}, \u_{t},\z_{t}, \bc_t, \m$}
 \KwResult{$\bmu_t, \Sigma_t$}
 \tcp{Prediction (motion update)}
 $\bar{\bmu}_t = \dynmodel(\bmu_{t-1}, \u_t)$\\
 $\bar{\Sigma}_t = \dynJac_t\Sigma_{t-1} \dynJac_t^\top + \stateNoise_t$\\
 \tcp{Correction: loop over landmark measurements}
 \ForEach{$\z_t^i$}{
  $j = c_t^i$\\
  $S_t^i = H_t^j\bar{\Sigma}_{t}[H^j_t]^\top+\measNoise_t$\\
  $K^i_t = \bar{\Sigma}_{t}[H^j_t]^\top[S_t^i]^{-1}$\\
  $\bar{\bmu}_t = \bar{\bmu}_t + K^i_t(\z^i_t - h(\bar{\bmu}_{t}, j, \m))$\\
  $\bar{\Sigma}_t = (I - K^i_t H^j_t)\bar{\Sigma}_t$\\
 }
 \tcp{Final posterior}
 $\bmu_t = \bar{\bmu}_t$\\
 $\Sigma_t = \bar{\Sigma}_t$\\
 \Return $\bmu_t, \Sigma_t$
 \caption{EKF Localization with Known Correspondences}
 \label{alg:ekflocal}
\end{algorithm}
\cref{alg:ekflocal} summarizes the EKF localization recursion in the known-correspondence case.
This algorithm provides a useful baseline: if we knew which measurement came from which landmark, EKF localization would be a straightforward specialization of the standard EKF. 
The main additional difficulty in practice is that the correspondences $c_t^i$ are typically unknown and must be estimated jointly with the robot state.

\subsubsection{EKF Localization with Unknown Correspondences}
\label{sec:ekf-unknown-corr}
In realistic scenarios, the correspondences $c_t^i$ are not given and must be inferred from the data. 
One common approach is to estimate the correspondences using a maximum likelihood (MLE) criterion. 
At each time step, we choose $\bc_t = \{c_t^i\}$ to maximize the likelihood of the current measurements:
\begin{equation*}
    \hat{\bc}_{t} = \arg \max_{\bc_{t}}p(\z_{t}\given \bc_{1:t},\m,\z_{1:t-1},\u_{1:t}).
\end{equation*}
In words, we select the assignment of measurements to landmarks that makes the observed data most probable, given the map and the past history.

To make this optimization tractable, we first marginalize over the unknown pose $\x_t$:
\begin{equation*}
\begin{split}
p(\z_{t}\given \bc_{1:t},\m,\z_{1:t-1},\u_{1:t}) 
&= \int p(\z_{t}\given \x_t, \bc_{1:t}, \m, \z_{1:t-1}, \u_{1:t})\, p(\x_{t}\given \bc_{1:t},\m,\z_{1:t-1},\u_{1:t}) \d\x_t, \\
&= \int p(\z_{t}\given \x_t, \bc_{t}, \m)\,\overline{\bel}(\x_t) \d\x_t,
\end{split}
\end{equation*}
where we used the Markov assumption to simplify $p(\z_{t}\given \x_t, \bc_{1:t}, \m, \z_{1:t-1}, \u_{1:t}) =  p(\z_{t}\given \x_t, \bc_{t}, \m)$, and defined the predicted belief:
\begin{equation*}
	\overline{\bel}(\x_t) \definedas p(\x_{t}\given \bc_{1:t},\m,\z_{1:t-1},\u_{1:t}).
\end{equation*}

The term $p(\z_t \mid \x_t, \bc_t, \m)$ is the measurement model with known correspondences. 
Using the conditional independence assumption from \cref{eq:condind}, we can factor it as:
\begin{equation*}
  p(\z_t \mid \x_t, \bc_t, \m)
  = \prod_i p(\z_t^i \mid \x_t, c_t^i, \m).
\end{equation*}
Substituting into the integral, we obtain:
\begin{equation*}
  p(\z_t \mid \bc_{1:t}, \m, \z_{1:t-1}, \u_{1:t})
  = \int \overline{\bel}(\x_t) \prod_i p(\z_t^i \mid \x_t, c_t^i, \m)\, \d\x_t.
\end{equation*}

Although the measurements are conditionally independent given $\x_t$, they become coupled after marginalizing over $\x_t$. 
A common approximation is therefore to select correspondences independently by scoring each measurement-landmark pairing under the predicted belief:

\begin{equation*}
\begin{split}
\hat{c}^i_{t} = \arg \max_{c^i_{t}} \int p(\z^i_{t} \given \x_t, c^i_{t}, \m)\, \overline{\bel}(\x_t) \, \d\x_t.
\end{split}
\end{equation*}

Under the Gaussian assumptions on the belief and measurement model, the integral above is a Gaussian distribution with mean and covariance:
\begin{equation*}
  \int p(\z_t^i \mid \x_t, c_t^i, \m) \, \overline{\bel}(\x_t)\, \d\x_t
  \sim \mathcal{N}\bigl( \hat{\z}_t^{c_t^i},\, S_t^{c_t^i} \bigr),
\end{equation*}
where:
\begin{equation*}
  \hat{\z}_t^j = \measmodel(\bar{\bmu}_t, j, \m),
  \qquad
  S_t^j = H_t^j \bar{\Sigma}_t (H_t^j)^\top + \measNoise_t.
\end{equation*}

Maximizing the likelihood is therefore equivalent to choosing $c_t^i$ to maximize the Gaussian density $\mathcal{N}(\z_t^i \mid \hat{\z}_t^j, S_t^j)$, or equivalently to minimize the associated Mahalanobis distance:
\begin{equation}
    \hat{c}_t^i
    = \arg\min_{c_t^i} d_t^{i, c_t^i},
    \label{eq:c-ml}
\end{equation}
where:
\begin{equation}
    d_t^{ij}
    = (\z_t^i - \hat{\z}_t^j)^\top (S_t^j)^{-1} (\z_t^i - \hat{\z}_t^j),
    \label{eq:mahalanobis}
\end{equation}
is the Mahalanobis distance between the actual measurement $\z_t^i$ and its prediction $\hat{\z}_t^j$ under landmark $j$.

  In practice, additional \emph{validation gates} are often imposed: if the minimum distance $d_t^{i \hat{c}_t^i}$ exceeds a threshold, the measurement is treated as unmatched (assigned to the ``no landmark'' index $N+1$) and excluded from the EKF update. 
  Once the correspondences $\hat{\bc}_t$ have been determined, they are treated as known in \cref{alg:ekflocalunknowncorr}, yielding the full EKF localization algorithm with unknown correspondences.

\begin{algorithm}[tb!]
 \KwData{$\bmu_{t-1}, \Sigma_{t-1}, \u_{t},\z_{t}, \m$}
 \KwResult{$\bmu_t, \Sigma_t$}
 $\bar{\bmu}_t = \dynmodel(\bmu_{t-1}, \u_t)$\\
 $\bar{\Sigma}_t = \dynJac_t\Sigma_{t-1} \dynJac_t^\top + \stateNoise_t$\\
 \ForEach{$\z_t^i$}{
    \ForEach{landmark $k$ in the map}{
      $\hat{\z}_t^k = h(\bar{\bmu}_{t}, k, \m)$\\
      $S_t^k = H_t^k\bar{\Sigma}_{t}[H^k_t]^\top+\measNoise_t$\\
     }
  $j = \arg\min_k \:\: (\z_t^i-\hat{\z}^{k}_t)^\top  [S_t^{k}]^{-1} (\z_t^i-\hat{\z}^{k}_t)$\\
  $K^i_t = \bar{\Sigma}_{t}[H^j_t]^\top[S_t^j]^{-1}$\\
  $\bar{\bmu}_t = \bar{\bmu}_t + K^i_t(\z^i_t - \hat{\z}_t^j)$\\
  $\bar{\Sigma}_t = (I - K^i_t H^j_t)\bar{\Sigma}_t$\\
 }
 $\bmu_t = \bar{\bmu}_t$\\
 $\Sigma_t = \bar{\Sigma}_t$\\
 \Return $\bmu_t, \Sigma_t$
 \caption{EKF Localization with Unknown Correspondences}
 \label{alg:ekflocalunknowncorr}
\end{algorithm}

\begin{example}[Differential drive robot with range and bearing measurements] 
\label{ex:rangeandbearing}
Consider a differential drive robot with state $\x = \vCol{x, y, \theta}$ and a sensor that measures the range $r$ and bearing $\phi$ to landmarks $m_j \in \m$ in the robot’s local coordinate frame. 
We assume that at each time step the robot collects multiple measurements corresponding to different features:
\begin{equation*}
\z_t = \{[r_t^1,\phi_t^1]^\top , [r_t^2,\phi_t^2]^\top , \dots\},
\end{equation*}
where each measurement $\z_t^i$ contains the range $r_t^i$ and bearing $\phi_t^i$. 

Assuming the correspondences are known, the measurement model for the range and bearing of landmark $j$ is:
\begin{equation}
\measmodel(\x_t, j, \m)  = \begin{bmatrix}
\sqrt{(m_{j,x} - x)^{2} + (m_{j,y}- y)^{2}} \\
\text{atan2}(m_{j,y}- y, m_{j,x} - x) - \theta
\end{bmatrix}.
\end{equation}
The measurement Jacobian $H^j_t$ corresponding to a measurement from landmark $j$ is therefore:
\begin{equation}
H^j_t = \begin{bmatrix}
-\frac{m_{j,x} - \bar{\mu}_{t,x}}{\sqrt{(m_{j,x} - \bar{\mu}_{t,x})^2 + (m_{j,y} - \bar{\mu}_{t,y})^2}} & -\frac{m_{j,y} - \bar{\mu}_{t,y}}{\sqrt{(m_{j,x} - \bar{\mu}_{t,x})^2 + (m_{j,y} - \bar{\mu}_{t,y})^2}} & 0 \\
\frac{m_{j,y} - \bar{\mu}_{t,y}}{(m_{j,x} - \bar{\mu}_{t,x})^2 + (m_{j,y} - \bar{\mu}_{t,y})^2} & -\frac{m_{j,x} - \bar{\mu}_{t,x}}{(m_{j,x} - \bar{\mu}_{t,x})^2 + (m_{j,y} - \bar{\mu}_{t,y})^2} & -1
\end{bmatrix}.
\end{equation}
It is also common to assume a diagonal covariance for the measurement noise:
\begin{equation*}
\measNoise_t = \begin{bmatrix}
\sigma_r^2 & 0 \\ 0 & \sigma_\phi^2
\end{bmatrix},
\end{equation*}
where $\sigma_r$ is the standard deviation of the range measurement noise and $\sigma_\phi$ is the standard deviation of the bearing measurement noise. 
This reflects the assumption that range and bearing errors are uncorrelated.
\end{example}

\subsection{Monte Carlo Localization (MCL)}
\label{subsec:monte-carlo-localization}
The second practical realization of Markov localization we consider is \emph{Monte Carlo localization (MCL)}.
This algorithm leverages the non-parametric particle filter framework from \cref{ch:approximate-filters} and is particularly well suited to \emph{global pose} localization and the kidnapped robot problem.\sidenote{We can also use MCL to solve the kidnapped robot problem by injecting new randomly sampled particles at each step so that the filter remains sensitive to unexpected measurements.}

Like the particle filter, MCL represents the belief distribution $\bel(\x_t)$ by a set of $K$ particles:
\begin{equation*}
\particleset_t \coloneqq \{\x_t^{[1]}, \x_t^{[2]},..., \x_t^{[K]}\},
\end{equation*}
where each particle $\x_t^{[k]}$ represents a hypothesis about the true state $\x_t$. 
Regions of the state space with more particles correspond to higher belief.

At each step of the algorithm, we:
\begin{enumerate}
    \item propagate particles forward using the map-aware state transition model, and
    \item weight and resample the particles using the map-aware measurement model.
\end{enumerate}
This is summarized in \cref{alg:montecarlo}, which is nearly identical to the particle filter algorithm in \cref{alg:particle}, the only difference being that the map $\m$ now appears in the probabilistic state transition and measurement models.

\begin{algorithm}[tb!]
 \KwData{$\particleset_{t-1}, \u_{t}, \z_{t}, \m$}
 \KwResult{$\particleset_{t}$}
 $\bar{\particleset}_{t} = \particleset_t = \emptyset$\\
 \For{$k=1$ \KwTo $K$}{
  Sample $\bar{\x}_{t}^{[k]} \sim p(\x_t \given \x_{t-1}^{[k]}, \u_t,\m)$\\
  $w_t^{[k]} = p(\z_t \given \bar{\x}_{t}^{[k]},\m)$\\
  $\bar{\particleset}_{t} = \bar{\particleset}_{t} \cup \big(\bar{\x}_{t}^{[k]}, w_t^{[k]} \big)$\\
 }
 \For{$k=1$ \KwTo $K$}{
  Draw $i$ with probability $\propto w_t^{[i]}$\\
  Add $\bar{\x}_{t}^{[i]}$ to $\particleset_t$
 }
 \Return $\particleset_t$
 \caption{Monte Carlo Localization}
 \label{alg:montecarlo}
\end{algorithm}

In global localization, MCL typically starts from a broad prior, such as having particles spread over the entire map.
As the robot moves and gathers measurements, inconsistent hypotheses are down-weighted and vanish during resampling, while particles near the true pose accumulate.
This makes MCL a powerful tool for environments with strong ambiguities, repeated structures, or very uncertain initial conditions.

\section{Summary}
In this chapter, we examined the fundamental concepts and algorithms underlying robot localization—the process of estimating a robot's pose relative to a known map of the environment. 
We began with a taxonomy of localization problems, distinguishing between pose tracking and global localization, static and dynamic environments, active and passive strategies, and single-robot versus multi-robot settings. 
This taxonomy provides a conceptual checklist for matching real-world scenarios to appropriate algorithmic tools.

Building on the Bayesian estimation framework introduced earlier, we extended the state transition and measurement models to explicitly incorporate map information.
This led to the \emph{Markov localization} algorithm, a general Bayesian filter that serves as a conceptual foundation for map-aware localization.

We then explored two important realizations of this framework: EKF localization, which represents the belief as a unimodal Gaussian and is effective for pose tracking with moderate uncertainty; and MCL, which uses a particle-based, non-parametric representation suitable for multi-hypothesis and global localization problems.
We discussed data association—the challenge of determining correspondences between sensor measurements and known map features—and techniques such as maximum-likelihood matching and validation gates to improve robustness.

Together, these methods form the core of modern localization systems, enabling robots to estimate their position and orientation using noisy sensors, partial maps, and uncertain environments.
In later chapters, these ideas will reappear in more complex settings such as simultaneous localization and mapping (SLAM), where the robot must estimate \emph{both} its pose and the map at the same time.

\paragraph{To learn more.}
For a detailed and foundational treatment of probabilistic localization, \citet{ThrunBurgardEtAl2005} remains the canonical reference. 
The original formulation of the Monte Carlo Localization algorithm is presented by \citet{dellaert1999monte}, while \citet{LeonardDurrantWhyte1991} provide an early and influential treatment of EKF-based localization with geometric landmarks. 
Readers interested in recent advances and the connection between localization and SLAM may refer to \citet{cadena2017past} and the comprehensive modern text by \citet{slam-handbook}, which unify localization and mapping under a common probabilistic and optimization-based framework.

\section{Exercises}
The starter code for the exercises provided below is available online through GitHub. 
To get started, download the code by running in a terminal window:

\begin{tcolorbox}[colback=gray!10]
\begin{minted}{bash}
    git clone https://github.com/StanfordASL/pora-exercises.git
\end{minted}
\end{tcolorbox}

We denote Problems requiring hand-written solutions and coding in Python with \adjustbox{height=2ex, valign=c}{\includegraphics{figs/write.png}} and \adjustbox{height=2ex, valign=c}{\includegraphics{figs/code.png}}, respectively.

\subsection*{\adjustbox{height=2ex, valign=c}{\includegraphics{figs/code.png}}\ Problem 1: Extended Kalman Filter Localization}
In this problem, you will implement an extended Kalman filter (EKF) for robot localization in an environment where the robot can collect relative position measurements to a set of four landmarks.
We will consider a robot with a discrete-time dynamics model $\x_{t+1} = f(\x_t, \u_t) + \bm{\epsilon}_t$ defined by:
\begin{equation*}
\begin{split}
x_{t+1} &= x_{t} + V_{t} \cos(\theta_{t}) \Delta t + \epsilon_t^x, \\
y_{t+1} &= y_{t} + V_{t} \sin(\theta_{t}) \Delta t + \epsilon_t^y, \\
\theta_{t+1} &= \theta_{t} + \omega_t \Delta t + \epsilon_t^{\theta},
\end{split}
\end{equation*}
and the goal is to estimate the unknown robot pose, $\x_{t} = [x_{t}, y_t, \theta_t]^\top$. 
The noise vector $\bm{\epsilon}_t = [\epsilon_t^x, \epsilon_t^y, \epsilon_t^{\theta}]^\top$ is a random variable with a zero mean Gaussian distribution $\bm{\epsilon}_t \sim \mathcal{N}(\bm{0}, \stateNoise)$, where $\stateNoise = 0.1 \Delta t^2 I$. 
In this problem, we assume the global ground truth positions of the four landmarks are known. 
The landmarks are stationary objects in the environment, and their combined state vector is:
\begin{equation*}
\begin{split}
\m = \begin{bmatrix}
m_{1,x} & m_{1,y} & m_{2,x} & m_{2,y} & m_{3,x} & m_{3,y} & m_{4,x} & m_{4,y}
\end{bmatrix}^\top.
\end{split}
\end{equation*}
As the robot navigates through its environment, it receives noisy measurements of the positions of four landmarks in the environment relative to the robot's current pose. 
The measurement for landmark $i$ is the relative position with the measurement model:
\begin{equation*}
\begin{split}
\z_t^{i} = \measmodel(\x_t, i, \m) + \bm{\delta}_t = 
\begin{bmatrix}
    \cos(\theta_{t}) & \sin(\theta_{t}) \\
    -\sin(\theta_{t}) & \cos(\theta_{t})
\end{bmatrix}
\Big(\begin{bmatrix}
    m_{i,x} \\ m_{i,y}
\end{bmatrix} - \begin{bmatrix}
    x_t \\ y_t
\end{bmatrix}\Big) + \bm{\delta}_t,
\end{split}
\end{equation*}
where the measurements have an associated measurement noise with $\bm{\delta}_t \sim \mathcal{N}(\bm{0}, \measNoise)$, where $\measNoise = 0.25 I$. 
The full measurement vector of all landmarks is:
\begin{equation*}
\begin{split}
\z_t = \begin{bmatrix} \z_t^{1} \\  \z_t^{2} \\ \z_t^{3} \\ \z_t^{4} \end{bmatrix}.
\end{split}
\end{equation*}

In the file \colorcode{ch13/exercises/ekf\_localization.ipynb}, complete the following:
\begin{enumerate}
\item Implement the functions \colorcode{robot\_dynamics} and \colorcode{robot\_measurement} to match the dynamics and measurement models described above.
\item Implement the function \colorcode{dynamics\_jacobian} to compute the dynamics Jacobian $\dynJac_t = \nabla_{\x}f(\x_t, \u_t)$.
\item Implement the function \colorcode{measurement\_jacobian} to compute the measurement model Jacobian $H_t = \nabla_{\x}h(\x_t, \m)$ for the model that computes the full measurement vector $\z_t$.
\item Implement the function \colorcode{ekf\_localization\_update} to implement the EKF localization update described in \cref{alg:ekflocal}.
Note that you won't need to explicitly have a for loop in this function for each measurement since we have defined the update to be vectorized for all measurements at once.
\item Run the provided code to see how the algorithm performs for the simulated robot.
\end{enumerate}

\subsection*{\adjustbox{height=2ex, valign=c}{\includegraphics{figs/code.png}}\ Problem 2: Particle Filter Localization}
In this problem, we consider the same problem setup defined in Problem 1, where we are localizing a robot given relative position measurements to a set of four known landmarks.
In the file \colorcode{ch13/exercises/particle\_filter\_localization.ipynb}, complete the following:
\begin{enumerate}
\item Implement the functions \colorcode{robot\_dynamics} and \colorcode{robot\_measurement} to match the dynamics and measurement models described above. Note: you can reuse your solution from the previous problem.
\item Implement the particle filter algorithm update function \colorcode{particle\_filter\_update} as described in \cref{alg:montecarlo}.
\item Run the provided code to see how the algorithm performs for the simulated robot.
\end{enumerate}
\newpage
\printbibliography[segment=\therefsegment,heading=subbibliography,title={References}]
\chapter{Simultaneous Localization and Mapping (SLAM)}
\label{ch:slam}
\newrefsegment
In \cref{ch:robot-localization}, we studied robot \emph{localization} under the assumption that a map of the environment,~$\m$, was known.
While this assumption simplifies the estimation process, it is rarely met in practice.
Robots often operate in \emph{a priori unknown} or partially known environments.
Examples include autonomous search-and-rescue in collapsed buildings, planetary exploration, and mapping of underwater structures.
In such cases, a robot must concurrently infer its own state and build a map from noisy sensor data.
This joint estimation problem is known as \emph{simultaneous localization and mapping (SLAM)}.
The term SLAM and many classical algorithms were popularized by \citet[Chs.~10--13]{ThrunBurgardEtAl2005}, which remains a standard reference.

SLAM plays a central role in the perception and navigation layers of an autonomy stack.
By providing a consistent, evolving spatial frame of reference, it allows downstream planning, control, and decision-making components to operate effectively in previously unseen environments.

The SLAM problem emerged in the late 1980s, with seminal contributions by Hugh Durrant-Whyte, John Leonard, and others, who framed it as a probabilistic joint estimation of pose and map.
Early approaches treated SLAM as a large filtering problem.
The EKF\cite{SmithSelfEtAl1990, LeonardDurrantWhyte1991} provided a conceptually elegant way to maintain a joint Gaussian distribution over all robot poses and landmarks.
In this formulation, the state vector contains the robot pose together with the positions of all $N$ landmarks, and the EKF maintains a $(3+2N)\times(3+2N)$ covariance matrix that encodes correlations between every pair of variables.
As the map grows, storing and updating this dense covariance incurs $\mathcal{O}(N^2)$ memory and computation, which quickly becomes prohibitive for large-scale mapping.
Moreover, repeated linearization of nonlinear motion and measurement models around a single mean can introduce inconsistency and overly optimistic uncertainty estimates.

In the early 2000s, particle-filter methods, most notably FastSLAM\cite{MontemerloThrunEtAl2002} offered a different perspective. 
By combining a particle-based representation of the robot’s trajectory with separate Gaussian estimates for individual landmarks (a technique known as \emph{Rao--Blackwellization}), FastSLAM improved scalability and handled data association more flexibly. 
Here, \emph{data association} refers to the problem of determining which previously mapped landmark, if any, corresponds to each new sensor measurement, an ambiguity that arises when multiple landmarks look similar or when sensing is noisy. 
At the same time, particle-filter approaches are susceptible to \emph{particle depletion}. 
Over long trajectories, repeated weighting and resampling cause most particles to carry negligible weight, so that only a few distinct hypotheses effectively remain. 
When this happens, the particle set no longer provides a good approximation of the posterior, and the algorithm may become brittle unless many particles are used or more sophisticated proposal distributions are designed.

Subsequent work in the mid-2000s adopted an optimization-based view.
Rather than updating a filter one step at a time, researchers formulated SLAM as a sparse nonlinear least-squares problem over a \emph{pose graph}. 
In this representation, nodes correspond to robot poses and edges represent relative pose constraints obtained from odometry, loop closures, or scan matching. 
This \emph{graph-based SLAM} perspective made it possible to exploit sparsity in the underlying problem, leverage robust cost functions to manage outliers, and apply incremental solvers for real-time operation\sidenote{Representative systems include early pose-graph formulations by \citet{LuMilios1997}, as well as modern optimization and incremental smoothing approaches such as square-root SAM~\citep{DellaertKaess2006}, iSAM~\citep{KaessRanganathanDellaert2008}, and g2o~\citep{KummerleEtAl2011}.}.
The \emph{factor-graph} formalism further generalizes this idea: both robot poses and map variables are represented as nodes, while each measurement or prior contributes a \emph{factor}, that is, a local term in the joint probability density that depends only on a small subset of variables. 
Factor graphs provide a unified way to encode heterogeneous constraints (from odometry to landmark observations and calibration parameters) in a single probabilistic model and now underpin most modern SLAM back-ends\cite{dellaert2012factor,dellaert2021factor}.

Recent years have seen SLAM move beyond purely geometric mapping.
Visual and visual--inertial SLAM, dense 3D reconstruction, semantic mapping, neural implicit representations (e.g., NeRF and Gaussian splatting), and learning-based front-ends all build on the same probabilistic foundations while extending the range of environments and tasks for which SLAM is viable. 
Modern SLAM systems combine probabilistic estimation, large-scale optimization, and learned representations, and are increasingly viewed as core components of broader “spatial AI” systems.

In the remainder of this chapter, we develop a unified view of SLAM that connects these ideas. 
\Cref{sec:slam-paradigms} introduces the main algorithmic paradigms for SLAM, distinguishing filter-based and smoothing-based back-ends and clarifying the role each plays in modern systems. 
Next, we turn to the SLAM front-end in \cref{sec:slam-front-end}, describing how raw sensor data are converted into constraints, and then to modality-specific pipelines in \cref{sec:slam-modality-examples}, where we illustrate representative designs for visual, lidar, and radar SLAM. 
Building on this intuition, \cref{sec:slam-math-foundations} formalizes SLAM as a Bayesian state estimation problem, introduces the underlying state-space models for motion and measurement, and derives the online and full SLAM formulations. 
The following sections instantiate this framework in concrete algorithms: EKF-SLAM in \cref{sec:ekf-slam}, particle-filter-based methods such as FastSLAM in \cref{sec:pf-slam}, and optimization-based approaches including pose-graph SLAM and factor-graph SLAM in \cref{sec:graph-slam,sec:factor-graph-slam}. 
Finally, in \cref{sec:slam-advanced} we survey advanced and emerging methods, dense and semantic mapping, neural implicit representations, and learning-based front-ends, and connect them to the broader vision of spatial AI.

\subsection{SLAM Paradigms}
\label{sec:slam-paradigms}
Modern SLAM systems are typically organized into two conceptual layers.
The \emph{front-end} processes raw sensor data and extracts a set of constraints between states and map elements: features, correspondences, relative poses, and \emph{loop closures}. 
Loop closures occur when the robot revisits a previously seen place and obtains a measurement that links two non-consecutive poses\sidenote{For example, recognizing a corridor or room visited much earlier.}, and these constraints are crucial for correcting accumulated drift. 
The \emph{back-end} takes all these constraints and solves the underlying estimation problem, either recursively (filtering) or by optimizing over a window or the full trajectory (smoothing).

In this section, we focus on the back-end and distinguish two main paradigms:
\emph{filter-based} approaches and \emph{smoothing-based} approaches. 
Both are grounded in the same probabilistic models introduced later in \cref{sec:slam-math-foundations}, but they make different choices about which variables to estimate explicitly and how to use past measurements.

\subsubsection{Filter-based Approaches}
Filter-based SLAM maintains a compact state estimate that is updated online as new sensor measurements arrive.
The key idea is to propagate a belief over the current robot state and map using a recursive Bayes filter, without explicitly revisiting all past data at each step.

In \emph{EKF-SLAM}, the robot state and map are stacked into a single augmented state vector, and the joint belief is modeled as a multivariate Gaussian.
Motion and measurement models are linearized around the current estimate, and the EKF prediction--correction equations are applied at every time step (see \cref{subsec:ekf} for the underlying EKF machinery). 
This yields an online algorithm that fuses information incrementally, but the covariance matrix is dense and of size $(n+2N) \times (n+2N)$ for $N$ landmarks and $n$ robot state variables. 
Storing and updating this matrix incurs $\mathcal{O}(N^2)$ memory and computational cost, which becomes prohibitive as the map grows. 
Moreover, because non-linear motion and measurement models are repeatedly linearized around a single mean, linearization errors can accumulate over time and lead to inconsistent uncertainty estimates (see, for example, \citet[Chs.~10--13]{ThrunBurgardEtAl2005}).

\emph{Rao--Blackwellized particle filters}, such as FastSLAM\cite{MontemerloThrunEtAl2002}, adopt a hybrid strategy. 
A particle filter represents the distribution over robot trajectories, while each particle carries separate Gaussian estimates for the landmarks. 
Conditioned on a sampled trajectory, the landmark estimates become conditionally independent, so that the map can be updated by running a set of tractable Kalman filters, specifically one per landmark, instead of one huge joint filter.
This factorization dramatically reduces the cost per update and makes it easier to handle ambiguous data association. 
At the same time, particle-filter-based methods introduce new practical challenges, such as designing good proposal distributions, avoiding particle depletion over long trajectories, and maintaining global consistency.
A detailed treatment of FastSLAM and Rao--Blackwellized particle filters can be found in Thrun et al.~\citep[Ch.~13]{ThrunBurgardEtAl2005}.

Filter-based methods are appealing when computation and memory budgets are tight, and when online operation with bounded per-step cost is a primary requirement. 
However, when long-term consistency, large-scale mapping, and aggressive loop closing are critical, they may struggle to fully exploit all available measurements, especially those that create strong constraints between distant poses.

\subsubsection{Smoothing-based Approaches}
Smoothing-based SLAM treats the entire robot trajectory and map as variables in a global estimation problem. 
Rather than maintaining only the current posterior~$p(\y_t \mid \z_{1:t}, \u_{1:t})$, smoothing methods aim to recover either (i) the full posterior $p(\y_{1:t} \mid \z_{1:t}, \u_{1:t})$ or (ii) a maximum a posteriori (MAP) estimate of the entire trajectory and map.

A convenient way to express this is through a \emph{graph} representation with nodes and edges/factors.
\begin{itemize}
  \item \emph{Nodes} represent unknown variables, such as robot poses at
        different times and landmark positions.
  \item \emph{Edges} or \emph{factors} represent measurements or priors that
        relate a small subset of these variables: odometry constraints between
        consecutive poses, loop-closure constraints between non-consecutive
        poses, landmark observations that couple a pose to a landmark, and
        prior terms that anchor the map.
\end{itemize}

In a \emph{pose graph}, nodes are pose variables and edges are relative pose constraints (odometry and loop closures). 
In a more general \emph{factor graph}, both poses and map variables are nodes, and each measurement contributes a factor\sidenote{A local term in the joint probability density that depends only on the variables involved in that measurement.}.

From a probabilistic standpoint, each factor corresponds to a likelihood or prior term, and the overall objective (for MAP estimation) is to find the trajectory and map that best satisfy all factors simultaneously. 
Under Gaussian assumptions, this leads to a sparse nonlinear least-squares problem. 
The sparsity arises because each measurement involves only a few variables\sidenote{For example, a single relative pose constraint involves two poses, and a landmark observation involves one pose and one landmark.}, so the corresponding Jacobian and Hessian matrices contain many zeros. 
Modern sparse solvers exploit this structure to solve very large SLAM problems efficiently.
References for graph- and factor-graph-based SLAM include~\citet{DellaertKaess2006}, \citet{KaessRanganathanDellaert2008}, and \citet{KummerleEtAl2011}.

Smoothing has two main advantages over pure filtering:
\begin{enumerate}
  \item It can \emph{re-linearize} constraints in light of new data. When new
        measurements arrive, such as from a loop closure, the entire
        trajectory and map can be re-optimized, improving consistency relative
        to filters that only update the current state.
  \item It naturally incorporates loop closures by explicitly adding
        constraints between non-consecutive poses and adjusting the entire
        trajectory to satisfy them. This allows accumulated drift to be
        redistributed along the path when the robot revisits known areas.
\end{enumerate}

\subsubsection{Choosing a Paradigm}
In practice, the choice between filtering and smoothing depends on several factors:

\begin{itemize}
  \item \emph{Real-time constraints.}
  If strict real-time updates with tightly bounded per-step latency are required
  and computational resources are limited, filter-based approaches (EKF-SLAM,
  FastSLAM) are attractive.

  \item \emph{Problem scale and loop closures.}
  For large environments with many loop closures and long missions, smoothing-
  based methods and pose- or factor-graph formulations tend to yield more accurate
  and consistent maps, since they can re-optimize the full trajectory when new
  constraints arrive.

  \item \emph{Implementation complexity.}
  Filters are conceptually straightforward and often easier to implement for
  small systems. Graph-based methods require additional mathematical and
  software infrastructure\sidenote{For example, sparse linear algebra and nonlinear optimization libraries.}, but they are highly modular and extensible once in place.
  \item \emph{Application priorities.}
  Short-term navigation may prioritize fast local odometry and modest map
  maintenance, while long-term mapping, multi-session operation, and multi-robot
  scenarios benefit from the global consistency offered by smoothing and graph
  optimization.
\end{itemize}

In modern systems, the boundary between these paradigms is increasingly blurred. 
Incremental smoothing methods, such as iSAM\cite{KaessRanganathanDellaert2008}, provide real-time updates while retaining many of the advantages of batch optimization, and are widely used as SLAM back-ends in contemporary robotics libraries.

\subsection{Front-End}
\label{sec:slam-front-end}
So far, we have focused on back-end formulations: how to structure and solve the estimation problem once constraints are given. 
In a SLAM system, however, those constraints do not appear magically. 
They are distilled from raw sensor data by the \emph{front-end}, which transforms pixels, point clouds, and other measurements into discrete relationships—such as relative poses, landmark observations, and loop closures—that the back-end subsequently enforces.

A useful way to think about this division of labor is the following.
The back-end is the inference engine that reasons about the history of the robot’s motion and the structure of the environment, while the front-end is the perceptual pipeline that decides \emph{what} information the back-end sees and \emph{how} it is packaged. 
A strong back-end cannot compensate for a front-end that produces systematic outliers or weak, uninformative constraints.

At a high level, most SLAM front-ends can be understood as variants of the same conceptual pipeline:

\begin{enumerate}
  \item \emph{Feature extraction.} Identify salient structures in the sensor data, such as keypoints, edges, geometric primitives, and learned descriptors.
  \item \emph{Data association.} Decide which current features correspond to
        which previously seen features or map elements.
  \item \emph{Outlier rejection.} Detect and remove inconsistent or spurious
        associations before they contaminate the estimate.
  \item \emph{Loop closure detection.} Recognize previously visited places in
        order to add long-range constraints that correct drift.
  \item \emph{Scan alignment and registration.} For range data, estimate relative
        poses between overlapping point clouds.
\end{enumerate}

The detailed implementation of each stage depends strongly on the sensing modality\sidenote{For example, cameras, lidar, radar, or others.}, but their \emph{role in the SLAM system} is shared: they determine which parts of the environment become landmarks or edges in the graph, how confidently we relate different poses, and when we introduce powerful loop-closure constraints. 
In the rest of this subsection, we briefly revisit these stages, connecting them to the perception tools developed in earlier chapters and emphasizing their global impact on SLAM.

\subsubsection{Feature Extraction}
From the SLAM perspective, feature extraction is the step where we decide which aspects of the environment will serve as “handles” for localization and mapping.
This builds directly on the classical perception techniques of \cref{ch:classical_perception}, which developed image filtering, edge and corner detection, descriptors, and geometric feature extraction, as well as the point cloud registration tools of \cref{subsec:pointcloud-reg}.\sidenote{Deep learning architectures for perception, introduced in
\cref{ch:vision-networks}, provide learned alternatives to hand-designed features and are increasingly used in modern SLAM front-ends.}

In \emph{visual SLAM}, the front-end typically detects and describes repeatable image features---for example corners, blobs, or local patches---using classical methods such as Harris corners and SIFT-like descriptors, or modern learned features.
These features are chosen to be robust to viewpoint and illumination changes so that they can be matched reliably across time and across cameras. 
The resulting 2D keypoints and descriptors form the basis for constructing geometric constraints—such as epipolar relations and reprojection errors—between camera poses and 3D landmarks, as discussed in \cref{ch:cameras} and \cref{ch:stereo-vision}.

In \emph{lidar-based SLAM}, the analogous role is played by geometric primitives in point clouds: planes, edges, corner-like structures, or local surface patches. 
Extracting these primitives\sidenote{For instance, using segmentation and model fitting methods from \cref{ch:classical_perception}.} provides stable anchor points that can be tracked across scans and used for scan-to-scan or scan-to-map registration.

For radar and other modalities, feature extraction must cope with lower resolution, clutter, and multipath. 
A common representation for automotive radars is the \emph{range--Doppler image}: a 2D array obtained by applying Fourier transforms to the raw radar chirps, whose horizontal axis indexes \emph{range} bins (distance from the sensor) and whose vertical axis indexes \emph{Doppler} bins (radial velocity). 
Each pixel intensity reflects the strength of the return from targets at a given distance and relative speed, so a range--Doppler image can be viewed as a grayscale image that encodes both where objects are and how fast they are moving. 
In this representation, hand-crafted geometric cues are often combined with learned representations\sidenote{For example, CNNs operating directly on range--Doppler images.} to produce robust, repeatable signatures that can be matched over time.

\subsubsection{Data Association}
Once features have been extracted, the next crucial step is \emph{data association}: determining whether a feature observed at time~$t$ corresponds to a previously observed feature or map element.
This is the mechanism by which we translate “this corner in the current image” into “the same corner we saw two seconds ago” or “landmark~$m_j$ in the map”.

As in the localization setting in \cref{ch:robot-localization}, data association is challenging: perceptual aliasing, sensor noise, occlusions, and environmental changes all make different places look similar, and the same place can look different over time. 
A correct association introduces a valid constraint between poses and landmarks, whereas an incorrect association can inject a large, systematically biased constraint that corrupts the entire estimate.

In practice, SLAM systems use a combination of techniques for data association, including:
\begin{itemize}
  \item Descriptor similarity, such as the distance in feature space for visual keypoints.
  \item Geometric predictions from the current state estimate, such as projecting a
        landmark into the image or into a scan.
  \item Probabilistic \emph{gating}, where candidate matches are accepted only if
        they lie within an uncertainty-dependent region (often using Mahalanobis
        distance).
  \item Higher-level cues, such as semantic labels like whether a feature belongs to a
        “door” object, or temporal consistency.
\end{itemize}

In many systems, data association is the most fragile component, but designing strategies with high recall (few missed true matches) and high precision (few false matches) is nonetheless critical for robust operation.

\subsubsection{Outlier Rejection}

Even with careful association, some matches will be wrong. 
The front-end must therefore include explicit mechanisms for identifying and rejecting \emph{outliers} before they reach the back-end. 
This mirrors the geometric verification and model fitting pipelines discussed in \cref{ch:classical_perception}\sidenote{For example, RANSAC for line fitting or object pose estimation.}, but now the models are camera geometry, relative poses, or scan alignment.
Common strategies for outlier rejection include:
\begin{itemize}
  \item \emph{Geometric consistency checks}, such as verifying that matched
        features satisfy epipolar constraints or that 3D-to-2D correspondences
        are compatible with a plausible camera pose.
  \item \emph{Robust model fitting}, for example using RANSAC and its variants to infer
        a relative pose or homography from putative matches while discarding outliers.
  \item \emph{Statistical tests} on residuals, where measurements whose errors
        are inconsistent with the expected covariance are discarded or downweighted.
  \item \emph{Consistency with the current map}, for instance requiring that new
        measurements be compatible with already-estimated landmarks and surfaces.
\end{itemize}

From the standpoint of the SLAM back-end, outlier rejection is not an optional cleanup step: a handful of large, unmodeled outliers can easily overwhelm even a sophisticated optimization or filtering algorithm.

\subsubsection{Loop Closure Detection}
A defining capability of SLAM systems is \emph{loop closure detection}.
Loop closure detection involves recognizing that the robot has returned to a previously visited place and generating a constraint that links the corresponding poses. 
From a global perspective, loop closures are the main tool for correcting accumulated drift: they tie together distant parts of the trajectory and allow the back-end to redistribute errors along the path.

Conceptually, loop closure detection extends data association from individual features to whole \emph{places}. 
The front-end must answer questions like “does the current view correspond to some earlier pose $x_k$?” and, if so, “what is the relative transformation between $x_t$ and $x_k$?”. 
Typical loop closure pipelines involve the following steps:
\begin{enumerate}
  \item Constructing a compact descriptor for each place. For example, a bag-of-words
        representation over visual features, a global descriptor for a lidar scan,
        or a learned embedding from a neural network.
  \item Using this descriptor to retrieve a small set of candidate earlier poses
        that are likely to correspond to the same place.
  \item Verifying these candidates geometrically, such as by estimating a relative
        pose and checking the consistency of reprojection or registration errors.
\end{enumerate}

Visual place-recognition methods (both classical and learned) and lidar scan descriptors (such as scan-context-style approaches) implement these ideas with different design choices and trade-offs between robustness, invariance, and computational cost.

Because each accepted loop closure introduces a long-range constraint in the graph, the stakes are high: a true loop closure can dramatically improve global consistency, whereas a false one can severely distort the map. For this reason, loop closure modules are often conservative, preferring to miss some potential closures rather than accept unreliable ones.

\subsubsection{Scan Alignment and Iterative Closest Point (ICP)}
In range-based SLAM, an additional central task is \emph{scan alignment}: estimating the relative pose between overlapping point clouds or depth images.
This aligns successive lidar sweeps, RGB-D frames, or accumulated submaps and provides relative pose constraints to the back-end.

\cref{subsec:pointcloud-reg} introduced the point cloud registration problem and the Iterative Closest Point (ICP) algorithm in detail. 
In the SLAM front-end, ICP and its many variants are used as a building block:

\begin{itemize}
  \item Given an initial guess for the relative pose between two scans (from
        odometry, IMU integration, or feature-based matching), ICP refines this
        guess by iteratively pairing points or primitives and solving for the rigid
        transformation that best aligns them.
  \item The resulting transformation and its estimated uncertainty are then
        passed to the back-end as a relative pose constraint between the
        corresponding robot poses.
\end{itemize}

While conceptually simple and powerful, ICP is sensitive to initialization, can converge to local minima, and can be affected by outliers or partial overlap between scans. 
Modern SLAM systems therefore rarely rely on ICP alone: feature-based methods or learned global descriptors are used to obtain robust initial estimates and to reject poor matches, with ICP providing fine geometric refinement within a multi-stage registration pipeline.

\subsubsection{Summary}
Putting these pieces together, the front-end can be viewed as a modular pipeline that starts from raw sensor streams and outputs a set of carefully curated constraints: local feature-based measurements, loop-closure links, and scan-to-scan registrations, each with an associated uncertainty. 
In \cref{sec:slam-modality-examples}, we will see how this conceptual pipeline adapts to different sensing modalities, and in \cref{sec:slam-math-foundations} we will formalize how these constraints enter the probabilistic SLAM model and drive the back-end algorithms developed in the remainder of the chapter.

\subsection{SLAM Across Sensing Modalities}
\label{sec:slam-modality-examples}
Up to this point, we have separated the SLAM problem into a \emph{front-end}, which converts raw sensor observations into constraints, and a \emph{back-end}, which estimates the trajectory and map given those constraints. 
In practice, the design of both components is strongly shaped by the available sensors. 
Cameras, lidar, and radar provide very different raw data, and thus require different feature extraction, data association, and loop-closure strategies, even though they ultimately feed the same types of constraints into the back-end.

In this section, we illustrate how the general front-end pipeline from \cref{sec:slam-front-end} (feature extraction, data association, outlier rejection, loop closure, and scan alignment) specializes to three widely used sensing modalities: cameras (visual SLAM), lidar (lidar SLAM), and radar (radar SLAM).
Our goal is not to cover each modality exhaustively, but to highlight how the same probabilistic ideas manifest in different front-end designs and what this implies for the back-end.

\subsubsection{Vision-Based SLAM}
Vision-based SLAM uses cameras as the primary source of information. 
Common configurations include monocular, stereo, and visual--inertial\sidenote{Visual--inertial setups combine cameras with IMUs in a complementary way.} setups. 
The raw input is a stream of images that the front-end must transform into geometric constraints between camera poses and 3D structure.

\paragraph{Front-end.}
Building on the feature extraction and multi-view geometry tools from \cref{ch:cameras}--\cref{ch:classical_perception}, a typical visual front-end:

\begin{itemize}
  \item Detects and describes repeatable 2D features, such as corners, blobs, or local
        patches, using classical descriptors or learned features.
  \item Matches features across frames to obtain 2D--2D or 2D--3D correspondences.
  \item Estimates relative camera motion using geometric relations such as the
        \emph{essential matrix} and the \emph{Perspective-$n$-Point (PnP)} model.
  \item Performs outlier rejection, for example with RANSAC on the essential matrix or
        PnP residuals, and passes only geometrically consistent constraints to
        the back-end.
  \item Runs place-recognition and geometric verification to propose and confirm
        loop closures.
\end{itemize}

Recall from \cref{ch:stereo-vision} that, for calibrated cameras, the essential matrix $E = [t]_\times R$ encodes the epipolar constraint between two views:
$p^\top E p' = 0$ for corresponding normalized image points $p, p'$. 
Estimating $E$ from feature matches allows recovery of the relative rotation $R$ and translation direction $t$ between frames, up to scale. 
Similarly, the PnP problem (see \cref{ch:stereo-vision}) recovers the camera pose that best explains a set of 2D--3D correspondences between image points and map landmarks. 
In a SLAM front-end, these tools provide the relative pose edges and reprojection constraints that will later appear in the pose or factor graph.

For monocular cameras, the front-end must handle inherent scale ambiguity: relative motion can be recovered only up to an unknown scale until additional information resolves it\sidenote{For example, from motion parallax, priors, or other sensors.}. 
Stereo and visual--inertial setups reduce or remove this ambiguity by providing direct depth estimates or inertial motion constraints.

\paragraph{Back-end.}
Visual SLAM back-ends commonly use sparse bundle adjustment or pose-graph optimization: frame-to-frame relative pose estimates, landmark reprojection errors, and inertial constraints are assembled into a single optimization problem. 
Systems such as PTAM\cite{klein2007parallel}, ORB-SLAM\cite{mur2015orb}, and VINS-Mono\cite{qin2018vins} follow this design: a front-end that maintains a set of keyframes, tracks features, and proposes loop closures, coupled with a back-end that optimizes over camera poses (and possibly landmark positions) using the factor-graph machinery discussed in \cref{sec:slam-paradigms}.

\paragraph{Perspective.}
Vision-based SLAM excels at capturing rich appearance information and can be implemented with low-cost hardware, but it is sensitive to lighting changes, motion blur, and textureless scenes. 
These weaknesses motivate the use of robust front-end design and additional sensors such as IMUs and lidar.

\subsubsection{Lidar-Based SLAM}
Lidar sensors provide dense or semi-dense 3D point clouds that directly encode geometry. 
They are widely used in autonomous driving and field robotics, where accurate metric localization is required over large areas and in diverse conditions.

\paragraph{Front-end.}
A typical lidar SLAM front-end includes:
\begin{itemize}
  \item \emph{Scan preprocessing}, including deskewing and motion compensation,
        filtering, and ground removal.
  \item \emph{Feature extraction}, identifying edge-like or planar structures or
        small surface patches in the point cloud.
  \item \emph{Scan alignment}, using ICP or feature-based registration (building
        on the point cloud registration methods introduced in
        \cref{subsec:pointcloud-reg}).
  \item \emph{Loop closure detection}, often via global scan descriptors or
        learned embeddings that summarize the shape of a scan or local map.
\end{itemize}

ICP-based alignment provides relative pose constraints between consecutive scans (scan-to-scan) or between a scan and an accumulated submap (scan-to-map).
Combining these with loop-closure constraints yields a rich set of geometric relationships between poses that the back-end can exploit.

\paragraph{Back-end.}
On the back-end, pose-graph optimization is standard: nodes represent vehicle poses, and edges encode relative poses from scan registration and loop closures. 
Because lidar provides high-accuracy range data, even small misalignments can accumulate into significant drift if not corrected. 
Robust cost functions and outlier-resistant optimization strategies are therefore crucial. 
Incremental solvers allow these graphs to be updated in real time as new scans arrive.

Classic systems such as LOAM (lidar Odometry and Mapping)\cite{zhang2014loam} demonstrate how a carefully designed lidar front-end, coupled with a graph-based back-end, can achieve centimeter-level accuracy in real time.

\paragraph{Perspective.}
Compared to cameras, lidar front-ends work with sparser but metrically precise data. 
They avoid some of the perceptual aliasing issues of vision, but introduce their own challenges in scan registration, handling dynamic objects, and coping with adverse weather or sensor artifacts.

\subsubsection{Radar SLAM}
Radar sensors, historically used in autonomy mainly for collision avoidance, are increasingly used for SLAM due to their robustness to fog, rain, dust, and poor lighting, and their long-range detection capabilities.

\paragraph{Front-end.}
Radar returns are noisy, have lower angular resolution than cameras or lidar, and are degraded by multipath effects. 
As a result, feature extraction and data association are more challenging. 
Typical radar front-ends:
\begin{itemize}
  \item Compute range--Doppler or range--angle images from raw radar returns.
  \item Extract prominent reflectors or local patterns as features, often using
        filtering and thresholding techniques adapted from \cref{ch:classical_perception}.
  \item Build global descriptors of scans or trajectories, or use learned
        embeddings tailored to radar data, to support place recognition and loop
        closure detection.
  \item Estimate relative motions using registration in range--Doppler space,
        or by aligning reconstructed point clouds (when available), often in
        combination with inertial data.
\end{itemize}

Because of the high false-positive rate and spurious reflections, strong outlier rejection and conservative loop closure validation are especially important.

\paragraph{Back-end.}
Back-end formulations for radar SLAM often mirror those of lidar SLAM: pose graphs whose edges come from radar-based relative pose estimates and loop closures. 
Recent systems frequently combine radar with other modalities, such as radar--inertial or radar--lidar odometry, to compensate for radar’s lower spatial resolution and to operate reliably in conditions where cameras and lidar are degraded.

\paragraph{Perspective.}
Radar front-ends emphasize robustness over raw information density. 
When fused with other modalities, radar can significantly improve reliability in adverse conditions, which is increasingly important in safety-critical applications.

\subsubsection{Discussion}
Each sensing modality offers a different balance between richness of data, robustness, and environmental adaptability. 
Cameras provide dense appearance cues but can fail under challenging lighting or in texture-poor scenes. 
Lidar yields precise geometry but is more expensive and can be affected by adverse weather. 
Radar offers robustness and long range at the cost of sparser, noisier measurements and more difficult data association.

Modern SLAM systems therefore rarely rely on a single sensor. 
Instead, they combine complementary modalities---for example, visual--inertial odometry with lidar, or lidar with radar---so that the strengths of one sensor compensate for the weaknesses of another. 
This trend underscores the importance of the modular front-end and back-end architecture developed in \cref{sec:slam-front-end,sec:slam-paradigms}: front-ends can be adapted or extended as sensors change, while back-ends operate on a common abstraction of constraints.

In the next section, \cref{sec:slam-math-foundations}, we temporarily abstract away these modality-specific details and formalize SLAM as a Bayesian state estimation problem. 
There, we introduce the motion and measurement models that underlie both filtering and smoothing formulations, and show how the constraints produced by the various front-ends enter the probabilistic SLAM framework.

\subsection{Mathematical Foundations of SLAM}
\label{sec:slam-math-foundations}
Equipped with an understanding of the front-end, we can now formalize the SLAM problem as a Bayesian state estimation problem. 
This formal viewpoint captures, in a single probabilistic model, how the robot state evolves over time and how the map of the environment is refined as new data arrive. 
It also provides the bridge between the front-end constraints and the filtering- and smoothing-based back-ends discussed in \cref{sec:slam-paradigms}.

Throughout this section we assume that the map $\m$ is \emph{static} over the time horizon of interest: walls do not move, landmarks remain fixed, and the environment does not change in ways that must be explicitly modeled. 
This assumption is appropriate for many indoor and urban scenarios and keeps the notation manageable. 
Extensions to dynamic maps---for example, modeling moving objects or slowly changing geometry—typically augment the state with additional variables for dynamic entities and introduce explicit time-dependence in the map; we briefly return to these ideas in later chapters.

Formally, given a sequence of control inputs $\u_{1:t}$ and sensor measurements $\z_{1:t}$, the SLAM problem asks the robot to estimate both its trajectory $\x_{1:t}$ (or at least its current state $\x_t$) and the map $\m$ of the environment. 
As in \cref{ch:robot-localization}, we treat $\x_t \in \reals^n$ as the robot state at time $t$, typically including pose and possibly velocity or other motion-related variables. 
The map $\m$ encodes properties of the environment according to a chosen representation. 
We reuse the same families of map models introduced in \cref{ch:robot-localization}, namely feature-based, dense/grid-based, and hybrid maps, which can include sparse sets of landmarks, occupancy grids or signed distance fields, and combinations thereof for systems that require both long-range localization and detailed local geometry.

Since the goal is to estimate both the robot state and the map, it is convenient to combine them into a single \emph{joint state} at time $t$:
\begin{equation*}
  \y_t \coloneqq \tup{\x_t, \m},
\end{equation*}
and we denote the joint state trajectory by:
\begin{equation*}
  \y_{1:t} \coloneqq \tup{\y_1,\ldots,\y_t}.
\end{equation*}

Two closely related formulations are particularly important in practice.

\begin{definition}[Online SLAM.]
  The goal of online SLAM is to estimate the \emph{current} robot state together with the map.
  Mathematically, this corresponds to the belief:
  \begin{equation}
  \label{eq:online-slam}
  \bel(\y_t) = p(\y_t \mid \z_{1:t}, \u_{1:t}).
  \end{equation}
  \end{definition}
  
  \begin{definition}[Full SLAM.]
  The goal of full SLAM is to estimate the \emph{entire} trajectory of the robot together with the map.
  Mathematically, this corresponds to the belief:
  \begin{equation}
  \label{eq:full-slam}
  \bel(\y_{1:t}) = p(\y_{1:t} \mid \z_{1:t}, \u_{1:t}).
  \end{equation}
  \end{definition}

\begin{figure}[tb]
  \centering
  \includegraphics[width=.8\textwidth]{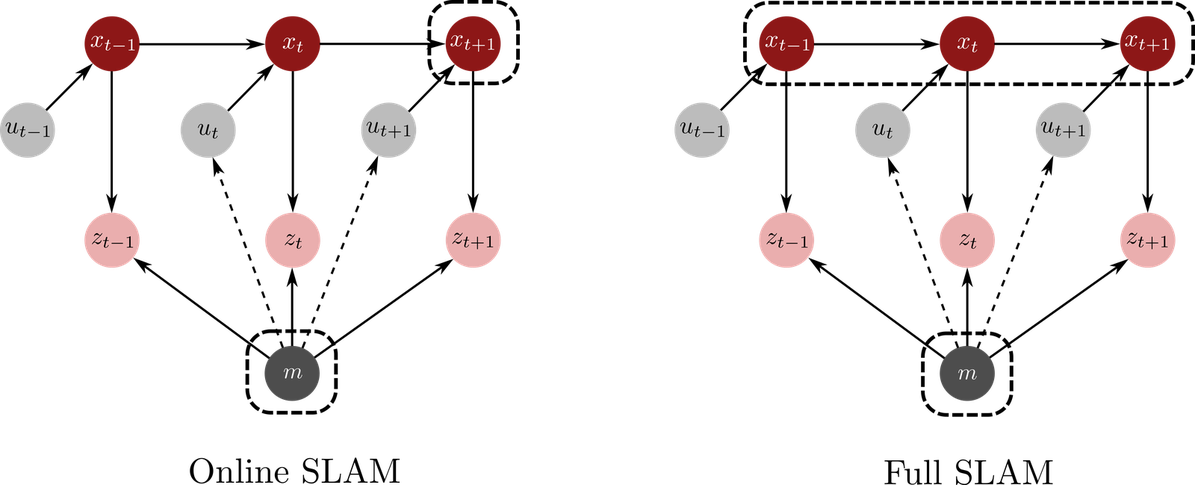}
  \caption{Online SLAM problems estimate only the current robot state together with the map, whereas full SLAM problems estimate the entire trajectory of past robot states (the state history) along with the map.}
  \label{fig:online_full_slam}
\end{figure}

  The distinction between these two viewpoints is illustrated in \cref{fig:online_full_slam}. 
  Online SLAM focuses on the most recent pose and the map, which is often what is needed for real-time control and navigation. 
  Full SLAM retains the entire pose history, which is particularly useful for building globally consistent maps, enforcing loop closures, or performing offline analysis over long missions.

  In both cases, localization and mapping are tightly coupled: accurate localization requires an accurate map, and accurate mapping depends on reliable localization. 
  This interplay makes SLAM sensitive to drift, data association errors, and outliers—issues that modern algorithms address via loop closure detection, robust front-ends, and global optimization in the back-end.

\subsubsection{Motion and Measurement Models}
To connect controls, states, and measurements in a probabilistic framework, we must first specify models for how the robot moves and how its sensors behave.

At each time step~$t$, the robot receives:
\begin{itemize}
  \item A control input~$\u_t$, such as wheel velocities, steering commands, thrust forces.
  \item Sensor observations~$\z_t$, such as lidar scans, images, range–bearing measurements.
\end{itemize}
The full histories up to time~$t$ are denoted by~$\u_{1:t}=\tup{\u_1,\ldots, \u_t}$ and~$\z_{1:t}=\tup{\z_1,\ldots,\z_t}$.

The robot’s motion is captured by a state transition model:
\begin{equation*}
  \x_{t+1}=\dynmodel(\x_{t},\u_t) + \bm{\epsilon}_t,
\end{equation*}
where~$\dynmodel(\cdot)$ encodes the deterministic dynamics and~$\bm{\epsilon}_t$ is stochastic process noise with distribution~$p(\bm{\epsilon}_t)$.
The term~$\bm{\epsilon}_t$ accounts for unmodeled effects such as slippage, disturbances, or modeling errors.

Sensor observations are modeled by a measurement function:
\begin{equation*}
  \z_{t}=\measmodel(\x_t,\m)+\bm{\delta}_t,
\end{equation*}
where~$\measmodel(\cdot)$ maps the robot state and map to an ideal, noise-free measurement, and~$\bm{\delta}_t$ is measurement noise with distribution~$p(\bm{\delta}_t)$.
This noise captures sensor imperfections and environmental effects, such as reflections, lighting changes, and occlusions.

Both motion and measurement models may be \emph{linear} or \emph{nonlinear}, depending on the sensing and actuation setup.
A few simple examples illustrate this variety:

\begin{example}[Linear motion model (odometry).]
  For a differential-drive robot with small wheel slippage, the dynamics can be approximated as:
  \begin{equation*}
    \x_{t+1}=\x_t + B\u_t + \bm{\epsilon}_t,
  \end{equation*}
  where~$B$ is a constant matrix mapping wheel velocities~$\u_t$ to pose increments.
\end{example}

\begin{example}[Nonlinear motion model (unicycle).]
  A more realistic planar model uses the robot’s heading:
  \begin{equation*}
    \begin{aligned}
    x_{t+1}&=x_t+V_t\cos(\theta_t)\Delta t+w_t^x,\\
    y_{t+1}&=y_t+V_t\sin(\theta_t)\Delta t+w_t^y,\\
    \theta_{t+1}&=\theta_t+\omega_t\Delta t+w_t^\theta,
    \end{aligned}
  \end{equation*}
  where~$V_t$ and~$\omega_t$ are commanded linear and angular velocities and~$w_t^x,w_t^y,w_t^\theta$ represent process noise.
\end{example}

\begin{example}[Linear measurement model (1D range).]
  If a robot moves along a line and measures the distance to a fixed landmark at position~$m$, then:
  \begin{equation*}
    z_t = m - x_t + v_t,
  \end{equation*}
  which is linear in~$x_t$.
\end{example}

\begin{example}[Nonlinear measurement model (range--bearing in 2D).]
  For a landmark at position~$\tup{m_x,m_y}$, the sensor might return range and bearing:
  \begin{equation*}
  \begin{aligned}
  r_t&=\sqrt{(m_x-x_t)^2+(m_y-y_t)^2}+\delta_t^r,\\
  \phi_t&=\operatorname{atan2}(m_y-y_t, m_x-x_t)-\theta_t+\delta_t^\phi,
  \end{aligned}
  \end{equation*}
  which is nonlinear in the robot pose and landmark coordinates.
\end{example}

\begin{example}[Nonlinear measurement model (camera projection).]
  A 3D landmark~$\tup{X,Y,Z}$ projects to image coordinates via:
  \begin{equation*}
    \begin{bmatrix} u \\ v \end{bmatrix} =
    \frac{1}{Z}
    \begin{bmatrix} f_x & 0 & c_x \\ 0 & f_y & c_y \end{bmatrix}
    \begin{bmatrix} X \\ Y \\ Z \end{bmatrix} + \bm{\delta}_t,
  \end{equation*}
  where~$f_x,f_y$ are focal lengths and~$c_x,c_y$ the principal point.
  The division by~$Z$ makes the model nonlinear.
\end{example}

These models, together with the choice of map representation, fully specify the probabilistic SLAM problem.

\subsubsection{Bayesian Formulation of SLAM}
From a Bayesian perspective, SLAM is a problem of inferring unknown quantities (the trajectory and map) from known data (controls and measurements), given probabilistic models for motion and sensing.

Using Bayes’ rule, the full SLAM posterior can be written as:
\begin{equation}
  p(\y_{1:t} \mid \z_{1:t}, \u_{1:t})
  \propto
  p(\z_{1:t} \mid \y_{1:t}, \u_{1:t}) \;
  p(\y_{1:t} \mid \u_{1:t}),
\label{eq:slam-bayes-posterior}
\end{equation}
where:
\begin{itemize}
  \item $p(\z_{1:t} \mid \y_{1:t}, \u_{1:t})$ is the {likelihood}, which encodes how probable the sensor data are given a particular trajectory and map; and
  \item $p(\y_{1:t} \mid \u_{1:t})$ is the {prior}, which captures knowledge about how the system evolves under the controls, as well as any prior assumptions on the map.
\end{itemize}

The distinction between online and full SLAM discussed above is reflected in where we place the focus: online SLAM is concerned with the marginal $p(\y_t \mid \z_{1:t}, \u_{1:t})$, obtained by integrating out past states from \cref{eq:slam-bayes-posterior}, whereas full SLAM keeps the entire path $p(\y_{1:t} \mid \z_{1:t}, \u_{1:t})$. 
In the remainder of the chapter, we start from this Bayesian formulation and derive classical solutions:
\emph{EKF-SLAM} for online SLAM and \emph{FastSLAM} for full SLAM.
We then move on to modern graph-based approaches, robust optimization, and registration methods such as \emph{ICP} and bundle adjustment, which can be interpreted as optimization-based approximations to the same underlying posterior.

\subsubsection{State-space Factorization for SLAM}
Direct computation of the SLAM posterior is intractable in all but the simplest cases.
However, the problem becomes manageable once we exploit conditional independencies implied by the motion and measurement models.

The key assumption is the {Markov property} of the dynamics: the future depends on the past only through the present state.
For the motion model, this gives:
\begin{equation*}
  p(\x_{t+1} \mid \x_{1:t}, \m, \u_{1:t}) = p(\x_{t+1} \mid \x_t, \u_t),
\end{equation*}
which states that the next state depends only on the current state and the latest control.
Similarly, for the measurement model we assume:
\begin{equation*}
  p(\z_t \mid \x_{1:t}, \m, \z_{1:t-1}, \u_{1:t}) = p(\z_t \mid \x_t, \m),
\end{equation*}
meaning that the current measurement depends only on the current state and the map, not on the full history.

Under these assumptions, the posterior for full SLAM factors as:
\begin{equation}
  p(\x_{1:t}, \m \mid \z_{1:t}, \u_{1:t})
  \propto \eta\, p(\x_{1})p(\m)
  \prod_{i=1}^{t} p(\z_i \mid \x_i, \m)
  \prod_{i=1}^{t-1} p(\x_{i+1} \mid \x_{i}, \u_{i}),
\label{eq:slam-factorization}
\end{equation}
where~$p(\x_1)$ is the prior over the initial state,~$p(\m)$ is the prior over the map, and~$\eta$ is a normalization constant.
Each factor in this expression corresponds directly to one of the components of the SLAM model:
\begin{itemize}
  \item $p(\x_{i+1} \mid \x_i, \u_i)$ is a motion factor (odometry, IMU, \dots).
  \item $p(\z_i \mid \x_i, \m)$ is a measurement factor (range–bearing, image
        reprojection, scan alignment, \dots).
  \item $p(\x_1)$ and $p(\m)$ are prior factors that anchor the solution.
\end{itemize}

This factorization is the state-space version of the graphical models discussed in \cref{sec:slam-paradigms}. 
It underlies both filtering-based and smoothing-based methods:

\begin{itemize}
  \item \emph{Filtering.}  
  Filtering-based methods marginalize out older states to maintain a belief over
  the current state and map:
  \begin{equation*}
    \bel(\x_t, \m) \;\coloneqq\; p(\x_t, \m \mid \z_{1:t}, \u_{1:t}),
  \end{equation*}
  and update it as new controls and measurements arrive. 
  Applying Bayes’ rule
  and the Markov assumptions yields the familiar Bayes-filter recursion:

  \emph{Prediction (motion update):}
  \begin{equation*}
    \overline{\bel}(\x_{t+1}, \m)
    = \int p(\x_{t+1} \mid \x_t, \u_t) \;
      \bel(\x_t, \m) \; \d\x_t,
  \end{equation*}

  \emph{Correction (measurement update):}
  \begin{equation*}
    \bel(\x_{t+1}, \m)
    = \eta \, p(\z_{t+1} \mid \x_{t+1}, \m) \;
      \overline{\bel}(\x_{t+1}, \m),
  \end{equation*}
  where $\eta$ normalizes the belief to integrate to $1$.

  \item \emph{Smoothing.}  
  Smoothing-based methods operate directly on
  \cref{eq:slam-factorization}, keeping the entire trajectory $\x_{1:t}$ and
  map $\m$ as variables. Under Gaussian assumptions, maximizing the posterior
  $p(\x_{1:t}, \m \mid \z_{1:t}, \u_{1:t})$ is equivalent to solving a sparse
  nonlinear least-squares problem whose terms correspond to the factors in
  \cref{eq:slam-factorization}—this is the optimization viewpoint developed in
  \cref{sec:graph-slam,sec:factor-graph-slam}.
\end{itemize}

The rest of this chapter introduces what can be interpreted as different ways of approximating and exploiting this factorization:
\begin{itemize}
  \item EKF-SLAM (\cref{sec:ekf-slam}) implements the Bayes-filter
        recursion directly under Gaussian noise and first-order linearization.
  \item FastSLAM and related particle-filter methods
        (\cref{sec:pf-slam}) use sampling over trajectories combined with
        conditional Gaussian subproblems for map features.
  \item Graph- and factor-graph-based SLAM
        (\cref{sec:graph-slam,sec:factor-graph-slam}) encode the factors in
        \cref{eq:slam-factorization} explicitly in a graphical model and solve
        for a MAP estimate via sparse nonlinear optimization.
\end{itemize}
In each case, the underlying probabilistic structure is the same; what differs is how we represent the belief and which approximations we make to render computation tractable.

\subsection{Extended Kalman Filter SLAM}
\label{sec:ekf-slam}
We have already encountered the EKF in two contexts: first as a general state estimation algorithm in \cref{ch:approximate-filters}, and then as a localization method when a map is known in \cref{ch:robot-localization}.
We now extend this idea to the SLAM setting.

In \emph{EKF-SLAM}, the map is treated as part of an augmented state vector, and the joint posterior over robot pose and map is updated recursively under Gaussian noise assumptions and first-order linearizations of the process and measurement models\cite{SmithSelfEtAl1990,LeonardDurrantWhyte1991}.
This generalizes the earlier uses of the EKF from estimating only the robot’s state to simultaneously estimating a static feature-based map and the evolving trajectory.

As in \cref{ch:robot-localization}, we assume that the map is feature-based:
\begin{equation*}
\m = \{m_1,m_2,\dots ,m_N\},
\end{equation*}
where $m_i$ is the $i$-th feature with coordinates $\tup{m_{i,x}, m_{i,y}}$ in a 2D environment.
The joint state vector at time~$t$ is then:
\begin{equation} 
  \label{eq:augmentedstate}
  \y_t \definedas \begin{bmatrix}
  \x_t \\ \m
  \end{bmatrix},
\end{equation}
and the online SLAM goal is to compute the posterior belief:
\begin{equation*}
  \bel(\y_t)=p(\x_t,\m\mid \z_{1:t},\u_{1:t}).
\end{equation*}

We consider a state transition model for the augmented state~$\y_t$ of the form:
\begin{equation*}
  \y_t=g(\y_{t-1},\u_t)+\bm{\epsilon}_t,
\end{equation*}
with additive Gaussian process noise~$\bm{\epsilon}_t\sim \mathcal{N}(\bm{0},\stateNoise_t)$.
The nonlinear function~$g$ is defined as:
\begin{equation*}
  g(\y_{t-1}, \u_t) = \begin{bmatrix}
  \dynmodel(\x_{t-1}, \u_t) \\ m_{1,t-1} \\ \vdots \\ m_{N,t-1}
  \end{bmatrix},
\end{equation*}
where~$\dynmodel$ denotes the robot motion model, and we assume that each map feature~$m_i$ is static, so its process model is the identity.
The process noise covariance has block structure:
\begin{equation*}
\stateNoise_t = \begin{bmatrix}
\tilde{\stateNoise}_t & 0 \\
0 & 0
\end{bmatrix},
\end{equation*}
where~$\tilde{\stateNoise}_t$ is the motion model noise covariance for the robot, and the zero blocks reflect the assumption of no process noise for the map features.

The Jacobian of the augmented motion model is:
\begin{equation*}
  G_t = \nabla_{\y} g(\y, \u_t)\big|_{\y=\bmu_{t-1}},
\end{equation*}
that is, the derivative of~$g$ with respect to the augmented state, evaluated at the current mean estimate~$\bmu_{t-1}$.

The measurement model mirrors the localization setting in \cref{ch:robot-localization}:
\begin{equation*}
\z^i_t = h(\y_t, j) + \bm{\delta}_t,
\end{equation*}
where~$\bm{\delta}_t \sim \mathcal{N}(\bm{0}, \measNoise_t)$ is zero-mean Gaussian noise and~$j$ is the index of the map feature~$m_j \in \m$ associated with measurement $i$.
The Jacobian of the measurement model is:
\begin{equation*}
  H^{j}_t = \nabla_{\y} h(\y, j)\big|_{\y = \bar{\bmu}_t},
\end{equation*}
the derivative of the measurement function with respect to the augmented state, evaluated at the predicted mean~$\bar{\bmu}_t$ obtained from the EKF prediction step.

\subsubsection{EKF-SLAM with Known Correspondences}
As in EKF localization, it is instructive to first consider the case in which the data associations are known.
Let~$\bc_t = [c_t^1, \dots ]^\top$ denote the correspondence vector, where~$c_t^i$ is the index of the map feature associated with measurement~$\z_t^i$.
With known correspondences, the EKF-SLAM algorithm shown in \cref{alg:ekfslam} is nearly identical to the EKF localization algorithm in \cref{alg:ekflocal}, except that it operates on the augmented state~$\y$ containing both robot pose and map.

\begin{algorithm}[tb!]
  \KwData{$\bmu_{t-1}, \Sigma_{t-1}, \u_{t}, \z_{t}, \bc_t$}
  \KwResult{$\bmu_t, \Sigma_t$}
  \tcp{Prediction step: propagate belief with motion model}
  $\bar{\bmu}_t = g(\bmu_{t-1}, \u_t)$\\
  $\bar{\Sigma}_t = G_t\Sigma_{t-1} G_t^{T} + \stateNoise_t$\\
  \tcp{Correction step: process each measurement}
  \ForEach{$\z_t^i$}{
   $j = c_t^i$ \tcp{index of associated map feature}
   \If{feature $j$ has never been seen before}{
     Initialize $\begin{bmatrix}\bar{\mu}_{j,x} \\ \bar{\mu}_{j,y} \end{bmatrix}$ 
     as the expected position based on~$\z_t^i$
   }
   \tcp{Innovation covariance}
   $S_t^i = H_t^j\bar{\Sigma}_{t}[H^j_t]^{T}+\measNoise_t$\\
   \tcp{Kalman gain}
   $K^i_t = \bar{\Sigma}_{t}[H^j_t]^{T}[S_t^i]^{-1}$\\
   \tcp{State update using measurement residual}
   $\bar{\bmu}_t = \bar{\bmu}_t + K^i_t(\z^i_t - h(\bar{\bmu}_{t}, j))$\\
   \tcp{Covariance update}
   $\bar{\Sigma}_t = (I - K^i_t H^j_t)\bar{\Sigma}_t$\\
  }
  \tcp{Final posterior belief}
  $\bmu_t = \bar{\bmu}_t$\\
  $\Sigma_t = \bar{\Sigma}_t$\\
  \Return $\bmu_t, \Sigma_t$
  \caption{EKF Online SLAM with Known Correspondences}
  \label{alg:ekfslam}
 \end{algorithm}

A typical initialization for the belief~$\bel(\y_0)$ places the robot at the origin of the map frame with high confidence and assigns very weak priors to the features:
\begin{equation*}
\bmu_0 = \begin{bmatrix}
\x_0 \\ 0 \\ \vdots \\ 0
\end{bmatrix}, \quad \Sigma_0 = \begin{bmatrix}
\tilde{\Sigma}_0 & 0 & \cdots & 0 \\
    0 & \infty & \cdots & 0 \\
     \vdots & \vdots & \ddots & \vdots \\
    0 & 0 & \cdots & \infty \\
\end{bmatrix},
\end{equation*}
where:
\begin{equation*}
\x_0 = \begin{bmatrix}
0 \\ \vdots \\ 0
\end{bmatrix}, \quad \tilde{\Sigma}_0 = \begin{bmatrix}
0 & \cdots & 0 \\ \vdots & \ddots & \vdots \\ 0 & \cdots & 0
\end{bmatrix},
\end{equation*}
and $\x_0$ and $\tilde{\Sigma}_0$ are the initial robot state and its covariance.
The large feature covariance terms—conceptually infinite—express complete lack of prior knowledge about landmark locations.
When a feature is first observed, the algorithm reinitializes its mean using the corresponding measurement, rather than linearizing the measurement function about an arbitrary initial guess such as the origin.

For a range–bearing sensor in 2D, the geometry is simple. 
Suppose the robot pose is $(x_t, y_t, \theta_t)$ in the global frame, and the sensor returns a range $r$ and bearing $\varphi$ to a previously unseen feature in the {robot frame}. 
In the robot frame, the feature lies at:
\begin{equation*}
  \begin{bmatrix}
    r\cos\varphi \\
    r\sin\varphi
  \end{bmatrix}.
\end{equation*}
To express this in the global frame, we first rotate by the robot heading $\theta_t$ and then translate by the robot position:
\begin{equation*}
  \begin{bmatrix}
    m_x \\[0.2em] m_y
  \end{bmatrix}
  =
  \begin{bmatrix}
    x_t \\[0.2em] y_t
  \end{bmatrix}
  +
  \begin{bmatrix}
    \cos\theta_t & -\sin\theta_t \\
    \sin\theta_t & \cos\theta_t
  \end{bmatrix}
  \begin{bmatrix}
    r\cos\varphi \\[0.2em]
    r\sin\varphi
  \end{bmatrix}.
\end{equation*}
Using trigonometric identities, this simplifies to:
\begin{equation*}
  m_x = x_t + r\cos(\theta_t+\varphi),
  \qquad
  m_y = y_t + r\sin(\theta_t+\varphi).
\end{equation*}
This is the formula used in the algorithm to initialize a landmark from a single range–bearing observation. 
Because the sensor provides both range and bearing, a single measurement suffices to place the landmark (up to measurement noise).
By contrast, for pure-bearing sensors such as a monocular camera without depth information, multiple views and robot motion are required to infer landmark positions, as discussed in the visual SLAM examples of \cref{sec:slam-modality-examples}.

\subsubsection{EKF-SLAM with Unknown Correspondences}
So far, we have assumed that each measurement~$\z_t^i$ is already matched to a map feature via a correspondence index~$c_t^i$.
In reality, these correspondences are rarely known and must be inferred online.
This makes SLAM substantially harder than the known-map localization problem in \cref{ch:intro-to-localization} because the map itself is uncertain.

Conceptually, the main new task is to decide, for every incoming measurement, whether it should be assigned to an \emph{existing} feature (data association) or interpreted as a \emph{new} feature (map expansion).
Both choices affect subsequent estimates: an incorrect association can corrupt the map, while an unnecessary new feature increases complexity.

A common strategy is to choose correspondences by {maximum likelihood}.
For each measurement $\z_t^i$, we evaluate its likelihood under each possible landmark hypothesis and select the most plausible one.

Given a predicted belief $\overline{\bel}(\y_t)$ after the motion update, the \emph{predictive distribution} for a measurement $\z_t^i$ associated with landmark $j$ is the distribution of $\z_t^i$ implied by the current uncertainty over $\y_t$ and the measurement model. 
Formally:
\begin{equation*}
  p(\z_t^i \mid \z_{1:t-1}, \u_{1:t}, c_t^i = j)
  = \int p(\z_t^i \mid \y_t, c_t^i = j)\;
        \overline{\bel}(\y_t)\, \d\y_t.
\end{equation*}
Under the EKF assumptions of a Gaussian belief and linearized measurement model, this integral can be computed in closed form and yields a likelihood Gaussian distribution $\mathcal{N}(\z_t^i \mid \hat{\z}_t^j, S_t^j)$ with mean:
\begin{equation*}
  \hat{\z}_t^j = \measmodel(\bar{\bmu}_t, m_j),
\end{equation*}
and covariance:
\begin{equation*}
  S_t^j = H_t^j \bar{\Sigma}_t [H_t^j]^\top + \measNoise_t,
\end{equation*}
exactly as in the EKF localization case.

Maximizing this likelihood with respect to $j$ is equivalent (see \cref{ch:intro-to-localization}) to minimizing the \emph{Mahalanobis distance}:
\begin{equation}
  \label{eq:mahalanobis-ekf-slam}
  d_t^{ij}
  =
  (\z_t^i-\hat{\z}_t^j)^\top [S_t^j]^{-1}(\z_t^i-\hat{\z}_t^j).
\end{equation}
Intuitively, $d_t^{ij}$ measures how many “standard deviations” the actual
measurement $\z_t^i$ lies from the predicted measurement $\hat{\z}_t^j$, taking
into account the full covariance. Compared to Euclidean distance, the
Mahalanobis distance automatically downweights directions of high uncertainty
and emphasizes directions of low uncertainty.

\paragraph{Decision rule and $\chi^2$ gating.}
The unknown-correspondence EKF-SLAM loop adds one more decision layer on top of \cref{alg:ekfslam}:

\begin{enumerate}
  \item For each measurement $\z_t^i$, hypothesize a potential new feature
        position, for example by triangulating from range–bearing measurements, which
        would increase the feature count from $N_{t-1}$ to $N_t = N_{t-1} + 1$.
  \item For all existing features $k = 1,\ldots,N_t$, compute the Mahalanobis
        distance $d_t^{ik}$ between $\z_t^i$ and the prediction for feature $k$:
        \begin{equation*}
          \hat{\z}_t^k = \measmodel(\bar{\bmu}_t, k),
          \qquad
          S_t^k = H_t^k\bar{\Sigma}_t[H_t^k]^{T}+\measNoise_t,
        \end{equation*}
        and:
        \begin{equation*}
          d_t^{ik}
          = (\z_t^i-\hat{\z}_t^k)^\top [S_t^k]^{-1}(\z_t^i-\hat{\z}_t^k).
        \end{equation*}
  \item If all $d_t^{ik}$ are “too large”, treat $\z_t^i$ as a new feature;
        otherwise, assign the measurement to the feature with the smallest
        Mahalanobis distance.
\end{enumerate}

To formalize the notion of “too large”, it is common to use a \emph{$\chi^2$ gate}. 
For a $d$-dimensional measurement and a correctly specified Gaussian model, the Mahalanobis distance $d_t^{ij}$ follows a $\chi^2$ distribution with $d$ degrees of freedom. 
This means that if a landmark hypothesis is correct, $d_t^{ij}$ will lie below a chosen threshold most of the time. 
We therefore pick a threshold $\alpha$ such that:
\begin{equation*}
  \Pr\bigl[\chi^2_d \le \alpha\bigr] = p,
\end{equation*}
where $p$ is a desired confidence level, such as $p = 0.95$. 
The region $d_t^{ij} \le \alpha$ is then a $p$-confidence “ellipse” in measurement space: we accept an association only if the measurement falls inside this ellipse. 
In practice:
\begin{itemize}
  \item If $d_t^{ik} \le \alpha$ for some $k$, we select the feature with the
        smallest $d_t^{ik}$ as the best match.
  \item If $d_t^{ik} > \alpha$ for all existing features, we consider $\z_t^i$
        to be a new landmark and initialize it accordingly.
\end{itemize}

The complete EKF-SLAM algorithm for unknown correspondences is summarized in
\cref{alg:ekfslamunknowncorr}.

\begin{algorithm}[tb!]
 \KwData{$\bmu_{t-1}, \Sigma_{t-1}, \bu_{t},\z_{t}, N_{t-1}$}
 \KwResult{$\bmu_t, \Sigma_t$}
 $N_t = N_{t-1}$\\
 \tcp{Prediction}
 $\bar{\bmu}_t = g(\bmu_{t-1}, \u_t)$\\
 $\bar{\Sigma}_t = G_t\Sigma_{t-1} G_t^{T} + \stateNoise_t$\\
 \tcp{Process each measurement}
 \ForEach{$\z_t^i$}{
  Estimate position $\begin{bmatrix}\bar{\mu}_{N_t + 1,x} \\ \bar{\mu}_{N_t + 1,y} \end{bmatrix}$ from $\z_t^i$ \\
    \ForEach{$k=1$ \KwTo $N_t+1$}{
      $\hat{\z}_t^k = h(\bar{\bmu}_{t}, k)$\\
      $S_t^k = H_t^k\bar{\Sigma}_{t}[H^k_t]^{T}+\measNoise_t$\\
      $d_t^{ik} = (\z_t^i-\hat{\z}^{k}_t)^\top  [S_t^{k}]^{-1} (\z_t^i-\hat{\z}^{k}_t)$\\
     }
  $d_t^{i(N_t+1)} = \alpha$\\
  $j = \arg\min_k \:\: d_t^{ik}$\\
  $N_t = \max\{N_t, j\}$\\
  $K^i_t = \bar{\Sigma}_{t}[H^j_t]^{T}[S_t^j]^{-1}$\\
  $\bar{\bmu}_t = \bar{\bmu}_t + K^i_t(\z^i_t - \hat{\z}_t^j)$\\
  $\bar{\Sigma}_t = (I - K^i_t H^j_t)\bar{\Sigma}_t$\\
 }
 $\bmu_t = \bar{\bmu}_t$\\
 $\Sigma_t = \bar{\Sigma}_t$\\
 \Return $\bmu_t, \Sigma_t$
 \caption{EKF Online SLAM with Unknown Correspondences}
 \label{alg:ekfslamunknowncorr}
\end{algorithm}

Although conceptually straightforward, EKF online SLAM with unknown correspondences is rarely robust enough for large, cluttered environments.
Spurious measurements can create false landmarks that persist indefinitely, and errors in association may contaminate both the map and the pose estimates.
Mitigation strategies include stronger outlier rejection in the front-end, more distinctive feature descriptors, and conservative validation gates. 
A further limitation is that the computational and memory requirements of EKF-SLAM scale quadratically with the number of features $N$, making it challenging to scale to very large maps.

\begin{example}[Differential drive robot with range and bearing measurements.] 
\label{ex:rangeandbearingEKFSLAM}
\theoremstyle{definition}
Consider a differential drive robot with state consisting of two-dimensional position and heading,~$\x = [x, y, \theta]^\top$.
Suppose a sensor is available that measures the range,~$r$, and bearing,~$\phi$, to features~$m_j \in \m$ relative to the robot’s local frame. 
At each time step, multiple measurements are collected:
\begin{equation*}
\z_t = \{[r_t^1,\phi_t^1]^\top , [r_t^2,\phi_t^2]^\top , \dots\},
\end{equation*}
where each measurement~$\z_t^i=[r_t^i,\phi_t^i]^\top$. 

For SLAM, define the augmented state:
\begin{equation*}
\y_t \definedas
    \begin{bmatrix}
    \x_t \\
    m_1 \\ \vdots \\ m_N
    \end{bmatrix}
    = \begin{bmatrix}
    x & y & \theta & m_{1,x} &m_{1,y} & \dots & m_{N,x} &m_{N,y}
    \end{bmatrix}^\top .
\end{equation*}
With known correspondences, the measurement model for feature~$j$ is:
\begin{equation*}
\measmodel(\y_t, j)  = \begin{bmatrix}
\sqrt{(m_{j,x} - x)^{2} + (m_{j,y}- y)^{2}} \\
\text{atan2}(m_{j,y}- y, m_{j,x} - x) - \theta
\end{bmatrix}.
\end{equation*}
The associated Jacobian~$H^j_t$, corresponding to a measurement from feature $j$, is:
\begin{equation*}
H^j_t = \begin{bmatrix}
-\frac{\bar{\mu}_{j,x} - \bar{\mu}_{t,x}}{\sqrt{q_{t,j}}} & -\frac{\bar{\mu}_{j,y} - \bar{\mu}_{t,y}}{\sqrt{q_{t,j}}} & 0 & 0 & \dots & 0 & \frac{\bar{\mu}_{j,x} - \bar{\mu}_{t,x}}{\sqrt{q_{t,j}}} & \frac{\bar{\mu}_{j,y} - \bar{\mu}_{t,y}}{\sqrt{q_{t,j}}} & 0 & \dots \\
\frac{\bar{\mu}_{j,y} - \bar{\mu}_{t,y}}{q_{t,j}} & -\frac{\bar{\mu}_{j,x} - \bar{\mu}_{t,x}}{q_{t,j}} & -1 & 0 & \dots & 0 & -\frac{\bar{\mu}_{j,y} - \bar{\mu}_{t,y}}{q_{t,j}} & \frac{\bar{\mu}_{j,x} - \bar{\mu}_{t,x}}{q_{t,j}} & 0 & \dots 
\end{bmatrix},
\end{equation*}
where:
\begin{equation*}
    q_{t,j} = (\bar{\mu}_{j,x} - \bar{\mu}_{t,x})^{2} + (\bar{\mu}_{j,y} - \bar{\mu}_{t,y})^{2},
\end{equation*}
and $\bar{\mu}_{j,x}$ and $\bar{\mu}_{j,y}$ are the estimates of the $x$ and $y$ coordinates of feature $m_j$ extracted from $\bar{\bmu}_t$.

Given both range and bearing measurements, we can initialize the estimated position of feature $m_j$ using:
\begin{equation*}
    \begin{bmatrix}
    \bar{\mu}_{j,x} \\
    \bar{\mu}_{j,y}
    \end{bmatrix}
    =
    \begin{bmatrix}
    \bar{\mu}_{t,x}\\
    \bar{\mu}_{t,y}
    \end{bmatrix}
    +
    \begin{bmatrix}
    r^{i}_{t}\cos(\phi^{i}_{t} + \bar{\mu}_{t,\theta}) \\
    r^{i}_{t}\sin(\phi^{i}_{t} + \bar{\mu}_{t,\theta})
    \end{bmatrix},
\end{equation*}
which can be used in the known-correspondence EKF-SLAM algorithm in \cref{alg:ekfslam} to initialize feature positions.
In the unknown-correspondence case of \cref{alg:ekfslamunknowncorr}, similar triangulation can be used to hypothesize new features.
Interactive code for this example (with known correspondences) is available in the repository \colorcode{github.com/StanfordASL/pora-exercises} in the notebook \colorcode{ch14/range\_bearing\_ekf\_slam.ipynb}.
\end{example}

While EKF-SLAM provides a principled probabilistic framework for joint pose and map estimation, its quadratic scaling in the number of features and its dependence on Gaussian assumptions and linearization limit performance in large or highly nonlinear environments.
These limitations have motivated more scalable and flexible alternatives, particularly particle filter-based and graph-based approaches, which we now discuss.

\subsection{Particle Filter-Based SLAM}
\label{sec:pf-slam}
The SLAM problem can also be tackled with nonparametric particle filters. 
A major advantage of this family of methods is that it fits naturally with the {full SLAM} formulation: a particle can represent an entire {trajectory} $\x_{1:t}$ of the robot, not just its current pose. 
In other words, the state of the particle filter is the whole path history, and the filter approximates the path posterior:
\begin{equation*}
  p(\x_{1:t} \mid \z_{1:t},\u_{1:t},\bc_{1:t}),
\end{equation*}
which is exactly the object of interest in full SLAM (see \cref{sec:slam-math-foundations}).
Once we have a set of sampled trajectories, we can condition on each trajectory and reason about the map. 
This stands in contrast to EKF-SLAM, which works directly in the online SLAM setting by maintaining a single Gaussian belief over the {current} pose and map.

A naive approach would treat the entire augmented state $\y_t$ from \cref{eq:augmentedstate} as the state of a particle filter, in analogy with MCL in \cref{ch:robot-localization}. 
In practice, however, this is infeasible: the number of particles required to approximate the belief grows rapidly with the state dimension, and a realistic map may contain hundreds or thousands of features.

The key insight behind particle-based SLAM (and FastSLAM in particular) is that, given the full robot path and known correspondences, the locations of individual map features become {conditionally independent}. 
Formally, the SLAM posterior over $\y_{1:t} = (\x_{1:t}, \m)$ can be factored as:
\begin{equation}
  p(\y_{1:t} \given \z_{1:t}, \u_{1:t}, \bc_{1:t}) 
  = p(\x_{1:t} \given \z_{1:t}, \u_{1:t}, \bc_{1:t})
    \prod_{i=1}^N p(m_i \given \x_{1:t}, \z_{1:t}, \bc_{1:t}),
  \label{eq:factored-short}
\end{equation}
whose derivation we present in more detail in \cref{eq:factored}.

This factorization separates the SLAM posterior into:
\begin{itemize}
  \item a \emph{path posterior} $p(\x_{1:t} \mid \z_{1:t}, \u_{1:t}, \bc_{1:t})$ over robot trajectories, and
  \item individual \emph{feature posteriors} $p(m_i \mid \x_{1:t}, \z_{1:t}, \bc_{1:t})$ for each map element.
\end{itemize}
The idea behind particle-based SLAM is to approximate the path posterior with a particle filter while maintaining each feature posterior with a parametric estimator conditioned on the sampled path\sidenote{The feature posterior is usually Gaussian.}. 
This reduces the effective dimensionality of the state space represented by particles: the map variables are handled analytically, and sampling is required only for the robot trajectory.

As with other particle-based methods, particle SLAM can:
(i) handle nonlinear process and measurement models without explicit linearization,
(ii) represent multimodal distributions, and
(iii) avoid computing Jacobians.
On the other hand, particle methods may require a large number of samples to avoid degeneracy in higher dimensions, and their performance depends heavily on the choice of proposal distributions and resampling strategies.

\paragraph{Factoring the posterior.}
Let the full augmented state be~$\y_{1:t}=\tup{\x_{1:t},\m}$ and assume a single measurement per time step with a known correspondence~$c_{1:t}$.
The factorization in \cref{eq:factored-short} can be written more explicitly as:
\begin{equation} 
\label{eq:factored}
p(\y_{1:t} \given \z_{1:t}, \u_{1:t}, c_{1:t}) = p(\x_{1:t} \given \z_{1:t}, \u_{1:t}, c_{1:t}) \prod_{i=1}^N p(m_i \given \x_{1:t}, \z_{1:t},  c_{1:t}),    
\end{equation}
where~$m_i$ is the $i$-th feature in the map $\m$, the term~$p(\x_{1:t} \given \z_{1:t}, \u_{1:t}, c_{1:t})$ is the \emph{path posterior}, and the terms $p(m_i \given \x_{1:t}, \z_{1:t},  c_{1:t})$ are the \emph{feature posteriors}. 

We derive this factorization as follows.
First, by Bayes’ rule:
\begin{equation*}
p(\y_{1:t} \given \z_{1:t}, \u_{1:t}, c_{1:t}) = p(\x_{1:t} \given \z_{1:t}, \u_{1:t}, c_{1:t}) p(\m \given \x_{1:t}, \z_{1:t}, \u_{1:t},  c_{1:t}).
\end{equation*}
Conditioning the feature posterior on~$\x_{1:t}$ renders the past controls redundant, so:
\begin{equation*}
p(\y_{1:t} \given \z_{1:t}, \u_{1:t}, c_{1:t}) = p(\x_{1:t} \given \z_{1:t}, \u_{1:t}, c_{1:t}) p(\m \given \x_{1:t}, \z_{1:t},  c_{1:t}).
\end{equation*}

Next, consider the feature posterior $p(\m \given \x_{1:t}, \z_{1:t}, c_{1:t})$ and focus on a particular feature~$m_i$.
We distinguish two cases according to whether this feature is observed at time~$t$:
if $i \neq c_t$, feature~$m_i$ is not observed, whereas if $i = c_t$, it is.
Under these two cases we get:
\begin{equation*}
p(m_i \given \x_{1:t}, \z_{1:t}, c_{1:t}) = \begin{cases}
p(m_i \given \x_{1:t-1}, \z_{1:t-1},  c_{1:t-1}), & i \neq c_t, \\
\frac{p(\z_t \given m_i, \x_{t}, c_{t})p(m_i \given \x_{1:t-1}, \z_{1:t-1}, c_{1:t-1})}{p(\z_t \given \x_{1:t}, \z_{1:t-1}, c_{1:t})}, & i  = c_t,
\end{cases}
\end{equation*}
where the first case simply states that an unobserved feature cannot be updated by the latest measurement, and the second follows from Bayes’ rule together with conditional independence of features given the trajectory.

For the observed feature ($i=c_t$), we may also write:
\begin{equation*}
p(m_{c_t} \given \x_{1:t-1}, \z_{1:t-1}, c_{1:t-1}) =\frac{p(\z_t \given \x_{1:t}, \z_{1:t-1}, c_{1:t})p(m_{c_t} \given \x_{1:t}, \z_{1:t}, c_{1:t})}{p(\z_t \given m_{c_t}, \x_{t}, c_{t})}.
\end{equation*}

We now show that the factorization in \cref{eq:factored} holds by induction.
Assume that at time~$t-1$ the feature posterior factors as\sidenote{This is trivially true at the first time step because there is not yet any information coupling the features.}:
\begin{equation*}
p(\m \given \x_{1:t-1}, \z_{1:t-1},  c_{1:t-1}) = \prod_{i=1}^N p(m_i \given \x_{1:t-1}, \z_{1:t-1},  c_{1:t-1}).
\end{equation*}
Then:
\begin{equation*}
\begin{split}
p(\m \given \x_{1:t}, \z_{1:t}, c_{1:t})
&= \frac{p(\z_t \given \m, \x_{t}, c_{t})p(\m \given \x_{1:t-1}, \z_{1:t-1}, c_{1:t-1})}{p(\z_t \given \x_{1:t}, \z_{1:t-1}, c_{1:t})}, \\
&= \frac{p(\z_t \given m_{c_t}, \x_{t}, c_{t})}{p(\z_t \given \x_{1:t}, \z_{1:t-1}, c_{1:t})}
\prod_{i=1}^N p(m_i \given \x_{1:t-1}, \z_{1:t-1},  c_{1:t-1}). \\
\end{split}
\end{equation*}
Substituting the two cases of $i\neq c_t$ and $i=c_t$ for $p(m_i \given \x_{1:t}, \z_{1:t}, c_{1:t})$ yields:
\begin{equation*}
\begin{split}
p(\m \given \x_{1:t}, \z_{1:t}, c_{1:t})
&= p(m_{c_t} \given \x_{1:t}, \z_{1:t},  c_{1:t}) \prod_{i \neq c_t} p(m_i \given \x_{1:t}, \z_{1:t},  c_{1:t}) \\
&= \prod_{n=1}^N p(m_n \given \x_{1:t}, \z_{1:t},  c_{1:t}),
\end{split}
\end{equation*}
which proves the factorization by induction.

The factorization in \cref{eq:factored} says that once we fix a particular trajectory $\x_{1:t}$, each landmark can be estimated independently from the others. 
This is precisely what FastSLAM exploits: particles are used only to represent different hypotheses over the {trajectory}, while each particle carries an analytical estimate of every landmark conditioned on that trajectory.

\subsubsection{FastSLAM with Known Correspondences}
The factorization in \cref{eq:factored} forms the basis of \emph{FastSLAM}, a particle-based SLAM algorithm that exploits this structure for computational efficiency.
FastSLAM uses a particle filter to represent the path posterior $p(\x_{1:t} \given \z_{1:t}, \u_{1:t}, \bc_{1:t})$ and, for each particle, maintains a separate EKF for each map feature representing $p(m_i \given \x_{1:t}, \z_{1:t}, \bc_{1:t})$.
The complete procedure is summarized in \cref{alg:fastslam}.

In this scheme, the set of particles is:
\begin{equation*}
\particleset_t \definedas \{P_t^{[1]}, P_t^{[2]},\ldots, P_t^{[K]}\},
\end{equation*}
where the~$k$-th particle is:
\begin{equation*}
P_t^{[k]} \definedas \{\x_t^{[k]}, \bmu_{1,t}^{[k]}, \Sigma_{1,t}^{[k]}, \dots, \bmu_{N,t}^{[k]}, \Sigma_{N,t}^{[k]} \},
\end{equation*}
where $\x_t^{[k]}$ denotes a trajectory hypothesis for the robot state and $(\bmu_{i,t}^{[k]}, \Sigma_{i,t}^{[k]})$ the EKF mean and covariance for feature~$m_i$ under that trajectory.
For each particle, we thus maintain \emph{one EKF per feature}; with $K$ particles and $N$ features, there are $N K$ independent EKFs in total. 
Each EKF operates in a low-dimensional state space\sidenote{Typically 2D or 3D for a landmark.}, so these updates remain inexpensive even when the full map is large.

\paragraph{Algorithmic structure.}
The FastSLAM recursion closely resembles a particle filter, augmented with EKF feature updates:

\begin{enumerate}
  \item \emph{Prediction (motion update).}  
  For each particle, sample a new robot pose~$\x_t^{[k]}$ from the state transition model given the control input~$\u_t$:
  \begin{equation*}
    \x_t^{[k]} \sim p(\x_t \mid \x_{t-1}^{[k]}, \u_t).
  \end{equation*}

  \item \emph{Feature update (measurement correction).}  
  For the observed feature~$j=c_t$, update the EKF mean and covariance in each particle:
\begin{equation*}
  \begin{aligned}
  \hat{\z}^{[k]} &= h(\bmu_{j,t-1}^{[k]}, \x_t^{[k]}),\\
    S &= H^j \Sigma_{j,t-1}^{[k]} [H^j]^\top + Q_t,\\
    K &= \Sigma_{j,t-1}^{[k]} [H^j]^\top S^{-1},\\
    \bmu_{j,t}^{[k]} &= \bmu_{j,t-1}^{[k]} + K(\z_t - \hat{\z}^{[k]}), \quad
    \Sigma_{j,t}^{[k]} = (I - K H^j)\Sigma_{j,t-1}^{[k]}.
  \end{aligned}
\end{equation*}

  \item \emph{Weighting.}  
  Assign each particle a weight~$w^{[k]}$ proportional to the measurement likelihood under its map estimate:
  \begin{equation*}
    w^{[k]} \propto \exp\!\left(-\tfrac{1}{2}(\z_t - \hat{\z}^{[k]})^\top S^{-1}(\z_t - \hat{\z}^{[k]})\right).
  \end{equation*}

  \item \emph{Copying unchanged features.}  
  For all features~$n \neq c_t$, keep the corresponding EKF parameters unchanged:
  \begin{equation*}
    \bmu_{n,t}^{[k]} = \bmu_{n,t-1}^{[k]}, \quad
    \Sigma_{n,t}^{[k]} = \Sigma_{n,t-1}^{[k]}.
  \end{equation*}

  \item \emph{Resampling.}  
  Draw a new particle set~$\particleset_t$ by resampling from the weighted particles, favoring those that explain the measurements well.
\end{enumerate}

\begin{algorithm}[tb!]
  \KwData{$\particleset_{t-1}, \u_{t}, \z_{t}, c_t$}
  \KwResult{$\particleset_{t}$}
  \For{$k=1$ \KwTo $K$}{
    \tcp{Prediction: Sample new robot pose}
    Sample $\x_{t}^{[k]} \sim p(\x_t \given \x_{t-1}^{[k]}, \u_t)$\\
    
    \tcp{Measurement update for observed feature}
    $j = c_t$\\
    \eIf{feature $j$ never seen before}{
      Initialize feature: $(\bmu^{[k]}_{j,t-1}, \Sigma^{[k]}_{j,t-1})$
    }
    {
      $\hat{\z}^{[k]} = h(\bmu^{[k]}_{j,t-1}, \x_{t}^{[k]})$\\
      $S = H^j \Sigma^{[k]}_{j,t-1} [H^j]^{T} + \measNoise_t$\\
      $K = \Sigma^{[k]}_{j,t-1} [H^j]^{T} [S]^{-1}$\\
      $\bmu^{[k]}_{j,t} = \bmu^{[k]}_{j,t-1} + K(\z_t - \hat{\z}^{[k]})$\\
      $\Sigma^{[k]}_{j,t} = (I - K H^j)\Sigma^{[k]}_{j,t-1}$\\
      
      \tcp{Weighting: compute importance weight}
      $w^{[k]} = \bigl(\det(2\pi S)\bigr)^{-1/2}\exp\!\left(-\tfrac{1}{2}(\z_t - \hat{\z}^{[k]})^\top S^{-1} (\z_t - \hat{\z}^{[k]})\right)$\\
    }
    
    \tcp{Carry over unchanged features}
    \For{$n \in \{1,\dots,N\}, \, n \neq c_t$}{
      $\bmu^{[k]}_{n,t} = \bmu^{[k]}_{n,t-1}$\\
      $\Sigma^{[k]}_{n,t} = \Sigma^{[k]}_{n,t-1}$\\
    }
  }
  
  \tcp{Resampling: Select new particle set according to weights}
  $\particleset_t = \emptyset$\\
  \For{$i=1$ \KwTo $K$}{
    Draw $k$ with probability $\propto w_t^{[k]}$\\
    $\particleset_t = \particleset_t \cup (\x_{t}^{[k]}, \bmu^{[k]}_{1,t}, \Sigma^{[k]}_{1,t}, \dots, \bmu^{[k]}_{N,t}, \Sigma^{[k]}_{N,t})$\\
  }
  
  \Return $\particleset_t$
  \caption{FastSLAM}
  \label{alg:fastslam}
 \end{algorithm}

FastSLAM is thus a hybrid algorithm: it uses a particle filter to represent the distribution over trajectories and, within each particle, uses EKFs to maintain Gaussian estimates for each feature.
This combination avoids the worst of the curse of dimensionality by sampling only over robot state, not over the entire map.

\paragraph{Unknown correspondences.}
So far we have assumed that correspondences $c_t$ are known. 
In practice, this is rarely the case. 
FastSLAM can be extended to treat correspondences as latent variables as well, leading to algorithms often referred to as \emph{FastSLAM 2.0}\cite{MontemerloThrunEtAl2003} and related variants\cite{ThrunBurgardEtAl2005}.

At a high level, within each particle we can:
\begin{itemize}
  \item Evaluate the likelihood of an observation under multiple existing
        features using, for example, Mahalanobis-distance gating as in EKF-SLAM.
  \item Consider the hypothesis that the observation corresponds to a new
        feature and initialize a new EKF state for it.
  \item Update the particle’s weight by marginalizing over these correspondence
        hypotheses, or by selecting the most likely association within that
        particle.
\end{itemize}
This effectively embeds a data association procedure inside each particle: a particle whose map explains the observations well under some correspondence assignment receives a larger weight and is more likely to survive resampling.

While this strategy increases robustness to data association errors, it also increases computational cost, since each particle maintains its own local map and solves its own correspondence problem. 
In practice, FastSLAM with unknown correspondences is often combined with strong front-end outlier rejection, careful gating, and heuristics to keep the number of candidate associations manageable.
Detailed treatments can be found in \citet[Ch.~13]{ThrunBurgardEtAl2005} and subsequent work on
Rao–Blackwellized particle filters for SLAM.

\subsection{Graph SLAM}
\label{sec:graph-slam}
In the previous sections, we treated SLAM in full generality: the unknowns included both the robot trajectory and a (possibly large) map.
In many applications, however, we are primarily interested in a \emph{consistent trajectory} expressed in a global frame, while the environment is represented implicitly through relative pose constraints between robot states. 
This situation occurs, for example, when dense maps are built from aligned scans outside the optimization loop, or when a separate mapping module consumes the estimated poses.

These specific characteristics of the problem, an environment represented implicitly through relative pose measurements between robot states, motivate an important extension of the general SLAM formulation: \emph{pose-graph SLAM}.
Here, the unknowns are restricted to the sequence of robot poses $\x_{1:t}$, while landmarks and other map elements are either marginalized out or not represented explicitly. 
The resulting model is naturally expressed as a graph:

\begin{itemize}
  \item \emph{Nodes} correspond to robot poses $\x_i$ at discrete times (or
        keyframes).
  \item \emph{Edges} correspond to relative pose measurements between pairs
        of poses, such as odometry constraints between consecutive poses or
        loop-closure constraints between nonconsecutive poses.
\end{itemize}

A generic relative pose measurement between poses $i$ and $j$ can be written as:
\begin{equation*}
  \z_{ij} = \measmodel_{ij}(\x_i, \x_j) + \bm{\delta}_{ij},
\end{equation*}
where $\z_{ij}$ is the measured relative transformation from pose $i$ to pose $j$, $\measmodel_{ij}$ is the measurement function (often a composition of rigid transformations), and $\bm{\delta}_{ij}$ is zero-mean noise with known covariance.

Under Gaussian noise assumptions, the MAP estimate of the poses is obtained by minimizing the sum of squared, information-weighted residuals:
\begin{equation*}
  \x_{1:t}^\star
  =
  \arg\min_{\x_{1:t}}
  \sum_{(i,j) \in \mathcal{E}} 
  \left\| \bm{r}_{ij}(\x_i,\x_j) \right\|_{\Omega_{ij}}^2,
\end{equation*}
where $\mathcal{E}$ is the set of edges in the graph, $\Omega_{ij}$ is the \emph{information matrix} associated with measurement $\z_{ij}$, and:
\begin{equation*}
  \bm{r}_{ij}(\x_i,\x_j)
  \coloneqq
  \z_{ij} \ominus h_{ij}(\x_i,\x_j),
\end{equation*}
is the residual. 
The information matrix is the inverse of the measurement covariance, $\Omega_{ij} = \measNoise_{ij}^{-1}$. 
Directions with high measurement variance correspond to low information (small entries in $\Omega_{ij}$), and conversely, directions with low variance represent high information. 
The weighted norm $\| \bm{r} \|_{\Omega}^2 = \bm{r}^\top \Omega \bm{r}$ therefore penalizes residuals more strongly in directions where the sensor is reliable.

The operator $\ominus$ denotes the \emph{relative pose difference} on $SE(2)$ or $SE(3)$, the spaces of 2D or 3D rigid-body transformations. 
Given two poses $\x_a, \x_b \in SE(3)$, the expression:
\begin{equation*}
  \x_a \ominus \x_b
  = \log\!\big(\x_a^{-1}\x_b\big),
\end{equation*}
maps the transformation from $\x_a$ to $\x_b$ to a vector in a local linear space (a 3D vector for $SE(2)$, or 6D for $SE(3)$). 
Here $\log(\cdot)$ is the inverse of the exponential map used to represent small rotations and translations. 
This construction allows us to compute residuals as ordinary vectors, while still respecting the underlying geometry of rotations and translations.

\paragraph{Linearization and Jacobians.}
Because the residuals are generally nonlinear in $\x_i$ and $\x_j$, we solve the MAP problem iteratively\sidenote{For example, using Gauss--Newton or Levenberg--Marquardt methods.}. 
At each iteration, the residuals are linearized around the current pose estimates.
Denote the Jacobians of $\bm{r}_{ij}$ as:
\begin{equation*}
  A_i = \frac{\partial \bm{r}_{ij}}{\partial \x_i}, 
  \qquad
  A_j = \frac{\partial \bm{r}_{ij}}{\partial \x_j}.
\end{equation*}
These matrices describe how the residual for edge $(i,j)$ changes under small perturbations of the connected poses. 
Collecting these Jacobians over all edges gives the linearized relationship between pose increments and residuals.

\paragraph{Normal equations and sparsity.}
The linearized least-squares problem leads to the (sparse) normal equations:
\begin{equation*}
  H\,\Delta\x = \bm{b},
\end{equation*}
where $\Delta\x$ is the stacked vector of pose increments for all nodes. 
The global information (Hessian) matrix $H$ is obtained by summing contributions from all edges:
\begin{equation*}
  H = \sum_{(i,j)\in\mathcal{E}} J_{ij}^\top \,\Omega_{ij}\, J_{ij},
\end{equation*}
where $J_{ij}$ stacks the Jacobians $A_i$ and $A_j$, and $\bm{b}$ is the corresponding gradient vector:
\begin{equation*}
  \bm{b} = \sum_{(i,j)\in\mathcal{E}} J_{ij}^\top \,\Omega_{ij}\, \bm{r}_{ij}.
\end{equation*}

Each measurement affects only a small number of poses (typically two), so most entries in $H$ are zero: only the blocks corresponding to poses $i$ and $j$ are affected by measurement $(i,j)$. 
This sparsity is what allows large pose graphs with thousands of poses and constraints to be optimized efficiently using sparse linear algebra techniques.

Two practical refinements are crucial in real-world pose-graph SLAM.

\paragraph{Robust kernels.}
Incorrect edges---for example, from wrong loop closures or corrupted sensor data---can strongly pull the solution away from the true trajectory if they are modeled with a simple quadratic loss. 
To mitigate this, we often replace the quadratic term $\|\bm{r}_{ij}\|^2_{\Omega_{ij}}$ with a \emph{robust loss} $\rho(\|\bm{r}_{ij}\|^2_{\Omega_{ij}})$, such as a Huber or Tukey loss. 
These functions behave quadratically for small residuals (so inliers are treated as in ordinary least squares), but grow more slowly for large residuals, effectively downweighting measurements that are inconsistent with the majority of the data.
Robust estimation of this kind is standard in bundle adjustment and pose-graph SLAM; see, for example, the discussions in \citet{HartleyZisserman2002} and \citet{ThrunBurgardEtAl2005}.

\paragraph{Priors and gauge freedom.}
A pose graph contains only \emph{relative} constraints between poses. 
Without additional information, the entire trajectory can be rotated or translated without changing the relative pose errors, so the optimization problem is underdetermined. 
To fix this \emph{gauge freedom}, we add a prior\sidenote{Also sometimes called an \emph{anchor}.} on one pose, typically the first one, such as:
\begin{equation*}
  \x_1 \sim \mathcal{N}(\x_1^{\text{prior}}, \,\Sigma_{\text{prior}}),
\end{equation*}
with $\x_1^{\text{prior}}$ set to the origin and $\Sigma_{\text{prior}}$ a small covariance. 
This prior pins the coordinate frame and renders the solution unique up to small variations consistent with the prior.

\paragraph{Updating poses on the manifold.}
The solution of the linear system yields increments $\Delta\x_i$ that live in the local linear space attached to each pose (the tangent space of $SE(2)$ or $SE(3)$). 
Robot poses themselves, however, must remain valid rigid-body transformations and cannot be updated by simple vector addition.

To perform a valid update, we use an operation often called a \emph{retraction}, which maps a tangent increment back onto the manifold of rigid transformations:
\begin{equation*}
  \x_i \gets \x_i \oplus \Delta \x_i.
\end{equation*}
Concretely, for $SE(2)$ or $SE(3)$, this is typically implemented via the exponential map:
\begin{equation*}
  \x_i \oplus \Delta\x_i
  \;=\;
  \x_i \cdot \Exp(\Delta\x_i),
\end{equation*}
where $\Exp(\cdot)$ maps a small 3D (or 6D) vector $\Delta\x_i$ to a corresponding rigid-body transformation (rotation plus translation), and the product $\x_i \cdot \Exp(\Delta\x_i)$ composes the current pose with this small
increment. 
In practice, the user of a SLAM library does not need to work with the Lie-group details explicitly.
It is enough to understand that:
\begin{itemize}
  \item Optimization computes small incremental motions as vectors.
  \item These increments are “applied” to the current poses using group
        composition rather than plain addition.
\end{itemize}

\begin{example}[Pose-graph SLAM vs. dead reckoning.]
  \label{ex:pose-graph-dead-reckoning}
  In the repository \colorcode{github.com/StanfordASL/pora-exercises}, the notebook \\\noindent\colorcode{ch14/pose\_graph\_slam.ipynb} implements pose-graph SLAM for
  a differential-drive robot. 
  The baseline “dead-reckoning” trajectory is obtained by integrating odometry alone, without loop-closure constraints.
  As a result of dead reckoning, small errors accumulate over time and the path drifts. 
  The pose-graph solution augments odometry edges with loop-closure edges between nonconsecutive poses whenever the robot revisits a known place. 
  Optimizing the graph adjusts the entire trajectory so that both odometry and loop-closure constraints are satisfied as well as possible, dramatically reducing drift compared to dead reckoning.
  \end{example}

\begin{algorithm}[tb!]
  \DontPrintSemicolon
  \KwData{Initial poses $\x_{1:t}^{(0)}$ (e.g., from odometry), edge set $\mathcal{E}$ with measurements $\{\z_{ij}\}$ and information matrices $\{\Omega_{ij}\}$, prior on root pose $\tup{\x_1^{\text{prior}},\Omega_{\text{prior}}}$, max iterations~$K$, damping~$\lambda \ge 0$ (LM), robust kernel~$\rho$ (optional)}
  \KwResult{Optimized poses $\x_{1:t}^\star$}
  $\x \gets \x_{1:t}^{(0)}$ \;
  \For{$k=1$ \KwTo $K$}{
    Initialize normal equations: $H \gets \bm{0}$, $\bm{b} \gets \bm{0}$ \;
    \ForEach{$(i,j)\in\mathcal{E}$}{
      $\hat{\z}_{ij} \gets \measmodel_{ij}(\x_i,\x_j)$ \;
      $\bm{r}_{ij} \gets \z_{ij} \ominus \hat{\z}_{ij}$ \tcp*{pose residual on $SE(2/3)$}
      $A_i,A_j \gets \nabla_{\x_i,\x_j}\bm{r}_{ij}$ \;
      $w_{ij} \gets$ robust weight from $\rho(\|\bm{r}_{ij}\|_{\Omega_{ij}})$ \tcp*{set $w_{ij}=1$ if no $\rho$}
      $\tilde{\Omega}_{ij} \gets w_{ij}\Omega_{ij}$ \;
      \tcp{Scatter-add into sparse $H,\bm{b}$}
      $H_{ii} \pluseq A_i^\top \tilde{\Omega}_{ij} A_i,\quad
       H_{ij} \pluseq A_i^\top \tilde{\Omega}_{ij} A_j,\quad
       H_{jj} \pluseq A_j^\top \tilde{\Omega}_{ij} A_j$ \;
      $\bm{b}_i \pluseq A_i^\top \tilde{\Omega}_{ij}\bm{r}_{ij},\quad
       \bm{b}_j \pluseq A_j^\top \tilde{\Omega}_{ij}\bm{r}_{ij}$ \;
    }
    \tcp{Anchor to fix gauge}
    $H_{11} \pluseq \Omega_{\text{prior}},\quad
     \bm{b}_1 \pluseq \Omega_{\text{prior}}\big(\x_1^{\text{prior}} \ominus \x_1\big)$ \;
    \tcp{Solve for increment}
    Solve $(H + \lambda \bm{I}) \, \Delta\x = \bm{b}$ with sparse Cholesky/QR \;
    \tcp{Retract on the manifold}
    \For{$i=1$ \KwTo $t$}{ $\x_i \gets \x_i \oplus \Delta\x_i$ \tcp*{retraction via $\Exp(\cdot)$ on $SE(2/3)$}}
    \If{$\|\Delta\x\|_\infty < \varepsilon$ \textbf{or} \text{relative cost decrease} $< \tau$}{\textbf{break}}
  }
  \Return $\x$
    \caption{Pose-Graph (GraphSLAM) -- Batch Gauss--Newton / Levenberg--Marquardt}
  \label{alg:graphslam}
 \end{algorithm}

Pose-graph SLAM casts consistent trajectory estimation as a sparse, nonlinear least-squares optimization problem defined over a graph of poses. 
Odometry and other local motion estimates appear as edges between consecutive nodes, while loop closures appear as edges between nonconsecutive nodes corresponding to revisited places. 
These loop-closure edges are particularly powerful: they introduce long-range constraints that “tie together” distant parts of the trajectory and allow accumulated drift to be redistributed along the path.

Because each measurement involves only a small subset of poses, the resulting Hessian matrix is sparse and can be solved efficiently using modern sparse linear algebra and incremental solvers. 
Pose-graph SLAM is widely used when relative pose constraints are the primary information source, and the same formulation extends naturally to multi-robot scenarios (with inter-robot edges) and to hybrid representations in which selected landmarks, sensor extrinsics, or biases are kept as additional variables.

In the next section, we generalize this idea to \emph{factor graphs}, which provide a more flexible and modular representation for SLAM and related estimation problems, and make it convenient to incorporate heterogeneous measurements and additional unknowns within a single unified framework.

\subsection{Factor Graph SLAM}
\label{sec:factor-graph-slam}
In \cref{sec:graph-slam}, we focused on \emph{pose-graph SLAM}, in which the only unknown variables are robot poses and every measurement is expressed as a relative pose constraint between two poses.
While pose-graph SLAM covers many important applications, real SLAM systems often contain additional unknowns: explicit landmark positions, sensor calibration parameters, biases, or even semantic information. 
 We need a representation that can accommodate all of these in a principled way.

\emph{Factor-graph SLAM} provides this generalization. 
Instead of having a graph whose nodes are only robot poses and whose edges are only relative-pose constraints, we consider a graph in which:
\begin{itemize}
  \item Nodes represent \emph{any} unknown variable we want to estimate.
  \item Edges (referred to as factors) represent the probabilistic relation induced by a single measurement or prior on the subset of variables it involves.
\end{itemize}

Pose-graph SLAM is elegant but restrictive: it assumes that every piece of information can ultimately be written as a relative pose between two robot states. 
In practice, this abstraction hides important modeling elements:

\begin{itemize}
  \item \emph{Landmarks.}  
  If we want to maintain and refine explicit landmark locations, such as for
  long-term mapping or semantic reasoning, only representing poses is
  insufficient.
  \item \emph{Sensor parameters and biases.}  
  Camera intrinsics, lidar–IMU extrinsics, time offsets, and slowly varying
  sensor biases often need to be estimated jointly with the trajectory.
  \item \emph{Multi-way constraints.}  
  Some measurements depend on more than two variables at once, such as a stereo
  observation that depends on a pose, a landmark, and stereo calibration.
  These cannot be expressed as simple pairwise pose–pose constraints.
\end{itemize}

Factor graphs overcome these limitations by embedding SLAM into the broader framework of probabilistic graphical models. 
We already met this idea at a high level in \cref{sec:slam-paradigms}, and here we turn it into a concrete optimization problem. 
As we will see, pose-graph SLAM appears as the special case where poses are the only variables and all factors connect at most two poses at a time.
In this case, the factor-graph formulation reduces exactly to the pose-graph formulation and algorithm in \cref{alg:graphslam}.
 
\paragraph{Variables, factors, and the joint posterior.}

In a factor graph, we collect all unknowns into three (possibly overlapping) groups:
\begin{align*}
  X &= \{\x_1, \dots, \x_t\} && \text{robot poses},\\
  L &= \{\ell_1, \dots, \ell_M\} && \text{landmarks},\\
  \Theta &= \{\boldsymbol{\theta}_1, \dots\} && \text{sensor parameters, biases, \dots}
\end{align*}
We denote the set of all variables by:
\begin{equation*}
  Y \coloneqq \{X, L, \Theta\}.
\end{equation*}

Each measurement or prior gives rise to a factor that ties together only the subset of variables it depends on. 
Let $Y_k \subseteq Y$ be the variables affected by the $k$-th measurement, and let $\phi_k(Y_k)$ denote the corresponding factor. 
Under standard conditional-independence assumptions, the joint posterior has the product form:
\begin{equation}
  p(Y \mid Z)
  \;\propto\;
  \prod_{k} \phi_k(Y_k),
  \label{eq:fg-posterior}
\end{equation}
where $Z = \{\z_k\}$ denotes all measurements. 
Each factor $\phi_k$ can be interpreted as a (possibly unnormalized) likelihood term for measurement $k$ given the variables $Y_k$.
 
 \begin{example}[A ternary factor.]
  Suppose a stereo camera at pose $\x_t$ observes a point landmark $\ell_j$. 
  The stereo measurement $z_{t,j}$ depends on:
  \begin{itemize}
    \item The robot pose $\x_t$ (through the camera pose).
    \item The landmark coordinates $\ell_j$.
    \item Stereo calibration parameters $\btheta$, such as the baseline, camera intrinsics, etc.
  \end{itemize}
  In factor-graph language, this is a \emph{ternary} factor:
  \begin{equation*}
    \phi(\x_t, \ell_j, \btheta)
    \;\propto\;
    p(z_{t,j} \mid \x_t, \ell_j, \btheta),
  \end{equation*}
  connecting three variables simultaneously. 
  Such higher-order constraints cannot be represented in a pure pose graph, which allows only pairwise pose–pose edges, but they appear naturally in a factor graph by simply allowing factors to affect more than two nodes.
 \end{example}

 \paragraph{From probabilities to least squares.}
 As in the previous sections, we assume that each measurement $\z_k$ is modeled by a measurement function $\measmodel_k(Y_k)$ with additive Gaussian noise:
\begin{equation*}
  \z_k = \measmodel_k(Y_k) + \bm{\delta}_k,
  \qquad
  \bm{\delta}_k \sim \mathcal{N}(\bm{0}, \measNoise_k).
\end{equation*}
The associated factor is then:
\begin{equation*}
  \phi_k(Y_k)
  \;\propto\;
  \exp\!\Bigl(-\tfrac{1}{2}\,
               \| \bm{r}_k(Y_k) \|_{\measNoise_k^{-1}}^2\Bigr),
\end{equation*}
where:
\begin{equation*}
  \bm{r}_k(Y_k) \coloneqq \z_k \ominus \measmodel_k(Y_k),
\end{equation*}
is the residual\sidenote{Possibly defined in a tangent space, as in the pose-graph case.} and $\| \bm{r} \|_{R^{-1}}^2 = \bm{r}^\top \measNoise^{-1} \bm{r}$ is the information-weighted squared norm.

Substituting this into \cref{eq:fg-posterior} and taking the negative log-likelihood shows that computing a MAP estimate:
\begin{equation*}
  Y^\star
  = \arg\max_Y p(Y \mid Z),
\end{equation*}
is equivalent to solving the nonlinear least-squares problem:
\begin{equation}
  Y^\star
  =
  \arg\min_Y
  \sum_k
  \| \bm{r}_k(Y_k) \|_{\Omega_k}^2,
  \qquad
  \Omega_k \coloneqq \measNoise_k^{-1}.
  \label{eq:fg-least-squares}
\end{equation}

This has exactly the same structure as the pose-graph objective, but now the variables $Y$ include poses, landmarks, calibration parameters, and so on, and factors can involve any subset of them.

\paragraph{Linearization, sparsity, and the normal equations.}
To solve \ref{eq:fg-least-squares}, we use iterative methods such as Gauss--Newton or Levenberg--Marquardt (LM), as in \cref{sec:graph-slam}. 
At each iteration, we linearize every residual around the current estimate, $Y^{(k)}$:
\begin{equation*}
  \bm{r}_k(Y_k)
  \;\approx\;
  \bm{r}_k(Y_k^{(k)})
  + J_k \, \Delta Y_k,
\end{equation*}
where $J_k$ is the Jacobian of $\bm{r}_k$ with respect to the stacked variables $Y_k$, and $\Delta Y_k$ is the stacked increment for those variables.

Collecting the contributions from all factors leads to the linearized normal equations:
\begin{equation}
  H \,\Delta Y = \bm{b},
  \label{eq:fg-normal-eq}
\end{equation}
with:
\begin{align*}
  H &= \sum_k J_k^\top \Omega_k J_k,\\
  \bm{b} &= \sum_k J_k^\top \Omega_k \,\bm{r}_k.
\end{align*}
In other words, each factor contributes a local term $J_k^\top \Omega_k J_k$ to the global Hessian (information matrix) $H$ and a local term $J_k^\top \Omega_k \bm{r}_k$ to the gradient $\bm{b}$.

Crucially, each factor $\phi_k$ depends only on the variables in $Y_k$. 
This means that:
\begin{itemize}
  \item The Jacobian $J_k$ has nonzero columns only for those variables.
  \item The contribution $J_k^\top \Omega_k J_k$ affects only the corresponding
        blocks of $H$.
\end{itemize}

As a result, most entries of $H$ are zero: the matrix is \emph{sparse}. 
This sparsity is \emph{the key to scalability}, since sparse direct solvers with carefully chosen variable orderings can solve very large systems in time that grows almost linearly with the number of variables, rather than cubic in the dimension as in the dense case.

\paragraph{Robust kernels and priors.}
As in pose-graph SLAM, we often replace the simple quadratic term $\|\bm{r}_k\|^2_{\Omega_k}$ with a robust loss $\rho(\|\bm{r}_k\|^2_{\Omega_k})$ to reduce the influence of outliers\sidenote{For example, due to incorrect correspondences or spurious loop closures.}. 
Priors are represented as additional factors, such as a prior on the first pose or on calibration parameters, and they play the same role of fixing gauge freedoms and encoding prior knowledge.

\paragraph{Landmark marginalization and the Schur complement.}
When there are many landmarks, it is often advantageous to eliminate them analytically from the linear system and solve directly for the remaining variables (typically poses and calibration parameters). 
This is achieved via the \emph{Schur complement}, which we now introduce at a high level.

Consider the linear system in \cref{eq:fg-normal-eq} with unknowns separated into two groups: $Y = (X,L)$, where $X$ are pose-like variables and $L$ are landmarks. 
After reordering, the normal equations can be written in block form as:
\begin{equation*}
  \begin{bmatrix}
    H_{XX} & H_{XL} \\
    H_{LX} & H_{LL}
  \end{bmatrix}
  \begin{bmatrix}
    \Delta X \\
    \Delta L
  \end{bmatrix}
  =
  \begin{bmatrix}
    \bm{b}_X \\
    \bm{b}_L
  \end{bmatrix}.
\end{equation*}
The idea of the Schur complement is to first express $\Delta L$ in terms of $\Delta X$ using the second block row:
\begin{equation*}
  H_{LX} \Delta X + H_{LL} \Delta L = \bm{b}_L,
\end{equation*}
which, assuming $H_{LL}$ is invertible, gives:
\begin{equation*}
  \Delta L
  =
  H_{LL}^{-1} (\bm{b}_L - H_{LX} \Delta X),
\end{equation*}
and then substitute this into the first block row. 
The result is a reduced system in the pose variables alone:
\begin{equation}
  \bigl(H_{XX} - H_{XL} H_{LL}^{-1} H_{LX}\bigr) \Delta X
  =
  \bm{b}_X - H_{XL} H_{LL}^{-1} \bm{b}_L.
  \label{eq:schur-reduced}
\end{equation}
The matrix $H_{XX} - H_{XL} H_{LL}^{-1} H_{LX}$ is called the Schur complement of $H_{LL}$ in the full system. 
Once we solve \cref{eq:schur-reduced} for $\Delta X$, we can recover $\Delta L$ from the expression above if needed.

In SLAM problems, $H_{LL}$ is typically block-diagonal or very sparse because landmarks are conditionally independent given the poses and each landmark is observed by a small subset of poses. 
This makes $H_{LL}^{-1}$ cheap to compute or apply. 
The Schur complement therefore allows us to:
\begin{itemize}
  \item Reduce the dimension of the main linear system to solve to the number of pose and
        calibration variables.
  \item Still retain the information carried by landmark observations.
\end{itemize}
This strategy is standard in bundle adjustment and large-scale visual SLAM.

\paragraph{Manifold retraction and variable updates.}

After solving the linear system or its Schur-reduced version, we obtain an increment $\Delta Y$ for all variables. 
As in \cref{sec:graph-slam}, these increments live in the tangent spaces of the corresponding manifolds\sidenote{For example, $SE(3)$ for poses and $\reals^3$ for Euclidean points.} and cannot, in general, be added to the variables with ordinary vector addition.

To update the estimate, each variable $y \in Y$ is updated via a \emph{retraction}:
\begin{equation}
  y \gets y \oplus \Delta y,
  \label{eq:fg-retraction}
\end{equation}
where $\oplus$ maps a small vector $\Delta y$ in the tangent space at $y$ back to a valid point on the manifold. 
Concretely:
\begin{itemize}
  \item For pose variables on $SE(2)$ or $SE(3)$, $\oplus$ is implemented using
        the exponential map:
        \begin{equation*}
          \x \oplus \Delta\x = \x \cdot \Exp(\Delta\x),
        \end{equation*}
        where $\Exp(\Delta\x)$ converts the small 3D/6D vector $\Delta\x$ into
        a rigid-body transform, and the dot denotes composition.
  \item For Euclidean variables, such as landmark positions or scalar biases,
        $\oplus$ reduces to simple addition:
        \begin{equation*}
          \ell \oplus \Delta \ell = \ell + \Delta \ell.
        \end{equation*}
\end{itemize}
Conceptually, the optimization alternates between:
\begin{enumerate}
  \item Solving a linearized problem in a local coordinate system or tangent
        space.
  \item Mapping the resulting update back to the nonlinear manifold via
        \cref{eq:fg-retraction}.
\end{enumerate}
This is exactly the same pattern we saw for pose-graph SLAM, now generalized to arbitrary variable types.

\paragraph{Batch and incremental factor-graph SLAM.}

The batch factor-graph SLAM algorithm is summarized in \cref{alg:factor-graph-slam}. 
Algorithmically, it closely resembles the pose-graph optimizer in \cref{alg:graphslam} but operates on a larger set of variables and factors.

\begin{algorithm}[ht!]
   \DontPrintSemicolon
   \KwData{Variables $Y = \{X,L,\Theta,\dots\}$ with initial guess $Y^{(0)}$, factors $\mathcal{F}=\{\phi_k\}$, each with measurement $\z_k$, information $\Omega_k$, and model $\z_k \approx \measmodel_k(Y_k)$, optional ordering $\pi$, max number of iterations $K$, damping $\lambda \ge 0$, robust kernel $\rho$ (optional)}
   \KwResult{MAP estimate $Y^\star$}
   $Y \gets Y^{(0)}$ \;
   \For{$t=1$ \KwTo $K$}{
     $H \gets \bm{0}$, $\bm{b} \gets \bm{0}$ \;
     \ForEach{$\phi_k \in \mathcal{F}$}{
       $\hat{\z}_k \gets h_k(Y_k)$ \;
       $\bm{r}_k \gets \z_k \ominus \hat{\z}_k$ \tcp*{on tangent space}
       $J_k \gets \nabla_{Y_k}\bm{r}_k$ \;
       $w_k \gets$ robust weight from $\rho(\|\bm{r}_k\|_{\Omega_k})$ \;
       $\tilde{\Omega}_k \gets w_k \Omega_k$ \;
       $H \pluseq J_k^\top \tilde{\Omega}_k J_k,\quad
        \bm{b} \pluseq J_k^\top \tilde{\Omega}_k \bm{r}_k$ \;
     }
     \If{\textnormal{use Schur complement}}{
       Partition $H,\bm{b}$ into pose vs.\ landmark blocks and eliminate $L$ \;
     }
     Solve $(H + \lambda \bm{I}) \Delta Y = \bm{b}$ with sparse Cholesky/QR \;
     \ForEach{variable $y \in Y$}{ $y \gets y \oplus \Delta y$ }
     \If{$\|\Delta Y\|_\infty < \varepsilon$ or relative cost decrease $< \tau$}{\textbf{break}}
   }
   \Return $Y$
   \caption{Factor-Graph SLAM: Batch Gauss--Newton / Levenberg--Marquardt}
   \label{alg:factor-graph-slam}
\end{algorithm}
 
 Gauss--Newton and Levenberg--Marquardt differ mainly in how the damping parameter $\lambda$ is chosen and updated. 
 Gauss--Newton corresponds to $\lambda = 0$ and works well when the initial estimate is close to the optimum.
 Levenberg--Marquardt introduces a positive $\lambda$ to make the linear system better conditioned and to interpolate between Gauss--Newton and gradient descent when far from the solution.
 
 For large-scale and online problems, recomputing and refactorizing $H$ from scratch at every iteration is wasteful. 
 \emph{Incremental smoothing and mapping} algorithms, such as iSAM and iSAM2, exploit the factor-graph structure to update only the affected parts of the solution when new measurements arrive, reusing previous computations and maintaining sparsity.\cite{KaessRanganathanDellaert2008,kaess2012isam2}
 This enables real-time performance in many practical SLAM systems.

 Factor-graph SLAM provides a unifying, modular view of SLAM back-ends:
 \begin{itemize}
   \item It generalizes pose-graph SLAM by allowing arbitrary variables
         (poses, landmarks, calibration, biases, \dots) and factors of any
         arity.
   \item The joint posterior factors into local terms, leading to sparse
         Jacobians and Hessians and enabling scalable optimization.
   \item Landmark marginalization via the Schur complement and incremental
         solvers such as iSAM build directly on this structure to handle
         large-scale, real-time applications.
 \end{itemize}
 Because of these advantages, factor graphs underpin nearly all modern SLAM back-ends and form the conceptual bridge between probabilistic modeling and the efficient numerical algorithms used in practice.

\subsection{Advanced and Emerging Methods}
\label{sec:slam-advanced}
Classical SLAM pipelines rely primarily on geometric features and optimization-based back-ends.
In recent years, however, the field has expanded to incorporate learning-based techniques, richer scene representations, and deeper integration with broader AI systems.
These emerging methods aim not only to improve accuracy and robustness, but also to enable SLAM to operate in environments and applications beyond the reach of purely geometric methods.
 
\paragraph{Deep learning in SLAM.}
 Learning has been used to enhance both the front-end and back-end of SLAM.
 On the front-end, convolutional and transformer-based networks provide robust feature detection, semantic segmentation, and depth estimation even in challenging lighting or texture-poor settings.
 On the back-end, learned priors can regularize optimization, improve loop-closure detection, or better model uncertainty in sensor data.
 
 Beyond such modular uses, end-to-end “neural SLAM” architectures attempt to replace hand-engineered pipelines entirely with learned models that ingest raw sensory streams and output pose and map estimates\cite{chaplot2020learning}.
 While promising, these approaches raise questions about generalization, interpretability, and robustness that remain active research topics.
 
 \paragraph{Neural implicit maps.}
 Traditional mapping approaches—occupancy grids, point clouds, and mesh reconstructions—either scale poorly or lack continuity and compactness.
 Neural implicit representations, such as neural signed distance functions and neural radiance fields, provide continuous, compact encodings of geometry and appearance.
 These maps can be queried at arbitrary resolution, fused across time, and potentially shared among multiple agents.
 
 Although computationally demanding, implicit maps suggest a new paradigm in which SLAM outputs not only metric geometry but also a photorealistic and semantically enriched “digital twin” of the environment.
 
 \subsubsection{Semantic and Dynamic SLAM}
 Classic SLAM often assumes static scenes, but real-world environments are dynamic and populated by moving agents.
 \emph{Semantic SLAM} augments maps with object-level labels, enabling robots to recognize and reason about doors, vehicles, furniture, and other meaningful entities rather than anonymous landmarks.
 \emph{Dynamic SLAM} explicitly models moving objects, separating them from the static background and, in some cases, tracking them jointly with the ego-motion.
 
 These capabilities unlock task-driven autonomy, where maps support higher-level reasoning, interaction, and prediction, not just localization.
 
 \subsubsection{Toward Spatial AI}
 Looking forward, SLAM is evolving beyond trajectory estimation and mapping toward a broader concept sometimes referred to as \emph{Spatial AI}.
 Here, geometry, semantics, and temporal dynamics are tightly integrated into a coherent representation that supports decision-making, planning, and interaction.
 Rather than being a self-contained module, SLAM becomes part of a larger perception–and–action loop, enabling robots to act intelligently in complex, dynamic worlds. 
 
 \section{Summary}
 In this chapter, we developed a unified view of the SLAM problem as a cornerstone of robot perception and autonomy. 
 We began by framing SLAM as a joint state estimation problem in which a robot must concurrently infer its own trajectory and a map of the environment from noisy sensor data.
 We introduced the Bayesian formulation that underlies both classical and modern approaches, emphasizing the Markov assumptions and probabilistic factorizations that make estimation tractable.
 
 We then explored the algorithmic paradigms that have shaped SLAM over the past three decades. 
 Filter-based methods, such as the EKF SLAM and particle-filter approaches like FastSLAM, introduced recursive estimation frameworks capable of operating online. 
 Smoothing-based methods, including graph- and factor-graph SLAM, reframed the SLAM problem as a sparse nonlinear least-squares optimization problem, enabling accurate and scalable solutions through modern sparse solvers and robust cost functions.
 
 The chapter further distinguished between the \emph{front-end}, which extracts constraints from raw sensor data (through feature detection, data association, and loop closure), and the \emph{back-end}, which solves the underlying estimation problem. 
 We discussed representative pipelines for different sensing modalities—including visual, lidar, and radar SLAM—and highlighted how sensor characteristics shape both front-end and back-end design.
 
 Finally, we surveyed recent advances that extend SLAM beyond purely geometric mapping, encompassing learning-based front-ends, neural implicit representations, semantic and dynamic mapping, and the emerging paradigm of \emph{Spatial AI}, which integrates geometry, semantics, and temporal reasoning into unified spatial representations.
 
 \paragraph{To learn more.}
 Comprehensive treatments of probabilistic robotics and SLAM algorithms can be found in~\citet{ThrunBurgardEtAl2005, LeonardDurrantWhyte1991}, which remain foundational references for understanding classical formulations. 
 For a modern perspective emphasizing optimization and factor graphs,~\citet{slam-handbook} provides an accessible and rigorous overview of contemporary SLAM back-ends and the underlying estimation theory. 
 Readers interested in practical implementations and ongoing research frontiers—particularly in visual–inertial, semantic, and learning-based SLAM—are encouraged to consult recent surveys and open-source frameworks such as GTSAM, g2o, and ORB-SLAM.
 
\section{Exercises}
The starter code for the exercises provided below is available online through GitHub. 
To get started, download the code by running in a terminal window:

\begin{tcolorbox}[colback=gray!10]
\begin{minted}{bash}
    git clone https://github.com/StanfordASL/pora-exercises.git
\end{minted}
\end{tcolorbox}

We denote Problems requiring hand-written solutions and coding in Python with \adjustbox{height=2ex, valign=c}{\includegraphics{figs/write.png}} and \adjustbox{height=2ex, valign=c}{\includegraphics{figs/code.png}}, respectively.

\subsection*{\adjustbox{height=2ex, valign=c}{\includegraphics{figs/code.png}}\ Problem 1: EKF SLAM}
In this problem, you will implement an extended Kalman filter (EKF) SLAM algorithm for robot and landmark localization in an environment where the robot can collect relative position measurements to a set of four landmarks\sidenote{Note that this is the same problem setup as Problem 1 and 2 in \cref{ch:robot-localization}.}.
Specifically, we consider a robot with a discrete-time dynamics model $\x_{t+1} = f(\x_t, \u_t) + \bm{\epsilon}_t$ with the state being the robot pose, $\x_{t} = [x_{t}, y_t, \theta_t]^\top$, and the dynamics are defined by:
\begin{equation*}
\begin{split}
x_{t+1} &= x_{t} + V_{t} \cos(\theta_{t}) \Delta t + \epsilon_t^x, \\
y_{t+1} &= y_{t} + V_{t} \sin(\theta_{t}) \Delta t + \epsilon_t^y, \\
\theta_{t+1} &= \theta_{t} + \omega_t \Delta t + \epsilon_t^{\theta}.
\end{split}
\end{equation*}
The noise vector $\bm{\epsilon}_t = [\epsilon_t^x, \epsilon_t^y, \epsilon_t^{\theta}]^\top$ is a random variable with a zero mean Gaussian distribution $\bm{\epsilon}_t \sim \mathcal{N}(\bm{0}, \stateNoise)$, where $\stateNoise = 0.1 \Delta t^2 I$. 

In this problem, we assume there are four stationary landmarks in the environment whose positions are unknown.
We define the state vector for the landmark positions as:
\begin{equation*}
\begin{split}
\m = \begin{bmatrix}
m_{1,x} & m_{1,y} & m_{2,x} & m_{2,y} & m_{3,x} & m_{3,y} & m_{4,x} & m_{4,y}
\end{bmatrix}^\top.
\end{split}
\end{equation*}
As the robot navigates through its environment, it receives noisy measurements of the positions of four landmarks in the environment relative to the robot's current pose. 
The measurement for landmark $i$ is the relative position with the measurement model:
\begin{equation*}
\begin{split}
\z_t^{i} = h(\x_t, i, \m) + \bm{\delta}_t = 
\begin{bmatrix}
    \cos(\theta_{t}) & \sin(\theta_{t}) \\
    -\sin(\theta_{t}) & \cos(\theta_{t})
\end{bmatrix}
\Big(\begin{bmatrix}
    m_{i,x} \\ m_{i,y}
\end{bmatrix} - \begin{bmatrix}
    x_t \\ y_t
\end{bmatrix}\Big) + \bm{\delta}_t,
\end{split}
\end{equation*}
where the measurements have associated noise $\bm{\delta}_t \sim \mathcal{N}(\bm{0}, \measNoise_t)$, with $\measNoise = 0.25 I$. 
The full measurement vector of all landmarks is:
\begin{equation*}
\begin{split}
\z_t = \begin{bmatrix} \z_t^{1} \\  \z_t^{2} \\ \z_t^{3} \\ \z_t^{4} \end{bmatrix}.
\end{split}
\end{equation*}

In this exercise, we consider the SLAM problem of estimating simultaneously the robot state $\x$ and the landmark state $\m$.
We denote the combined state as $\y = [\x, \m]^\top$ and denote the combined dynamics model for this state as $\y_{t+1}=g(\y_t,\u_{t})+\bm{\epsilon}_t$
In the file \colorcode{ch14/exercises/ekf\_slam.ipynb}, complete the following:
\begin{enumerate}
\item Implement the functions \colorcode{robot\_dynamics}, \colorcode{robot\_measurement}, and \\\noindent\colorcode{state\_dynamics} that define the robot's dynamics model and measurement model described above, as well as the dynamics model for the SLAM state $\y$.
\item Implement the function \colorcode{dynamics\_jacobian} to compute the dynamics Jacobian for the combined state $G_t = \nabla_{\y}g(\y_t, \u_t)$.
\item Implement the function \colorcode{measurement\_jacobian} to compute the measurement model Jacobian $H_t = \nabla_{\y}h(\x_t, \m)$ for the model that computes the full measurement vector $\z_t$.
Note we compute the measurement Jacobian with respect to the combined state $\y$.
\item Implement the function \colorcode{ekf\_slam\_update} to implement the EKF SLAM update.
\item Run the provided code to see how the algorithm performs for the simulated robot.
\end{enumerate}
\newpage
\printbibliography[segment=\therefsegment,heading=subbibliography,title={References}]
\chapter{Sensor Fusion and Object Tracking}
\label{ch:sensor-fusion}
\newrefsegment
Individual sensors come with hard-wired limitations in terms of range, field of view, and resolution or quantization. 
On top of these design limits, performance often degrades under certain environmental conditions\cite{hall2001handbook, ThrunBurgardEtAl2005}. 
In practice, sensors also age and occasionally fail: calibration drifts, biases grow, and sometimes a device drops out altogether. 
The aim of \emph{sensor fusion} is to design robotic systems that remain robust to individual sensor weaknesses by combining multiple, often heterogeneous, sensors so that their information collectively reduces uncertainty in perception and localization tasks\cite{Gustafsson2010,Simon2006}.

\medskip

\paragraph{Why fusion helps.}
Three recurring patterns motivate sensor fusion in real systems:
\begin{enumerate}
    \item \emph{Complementarity:} Different sensor modalities observe different aspects of the world, such as geometry or appearance, and combining them reduces ambiguity.
    \item \emph{Redundancy:} Multiple sensors that observe the same latent quantity allow resilience to noise spikes and dropouts via consistency checks and cross-validation.
    \item \emph{Cooperation:} Two weak cues can form a strong one when combined, such as monocular vision paired with an IMU to recover metric scale.
\end{enumerate}

At a high level, sensor fusion treats each sensor output as a probabilistic observation of the underlying state and then combines those observations to obtain a posterior belief that is sharper and more reliable than any single source.

\begin{figure}[ht]
    \centering
    \includegraphics[width=0.85\textwidth]{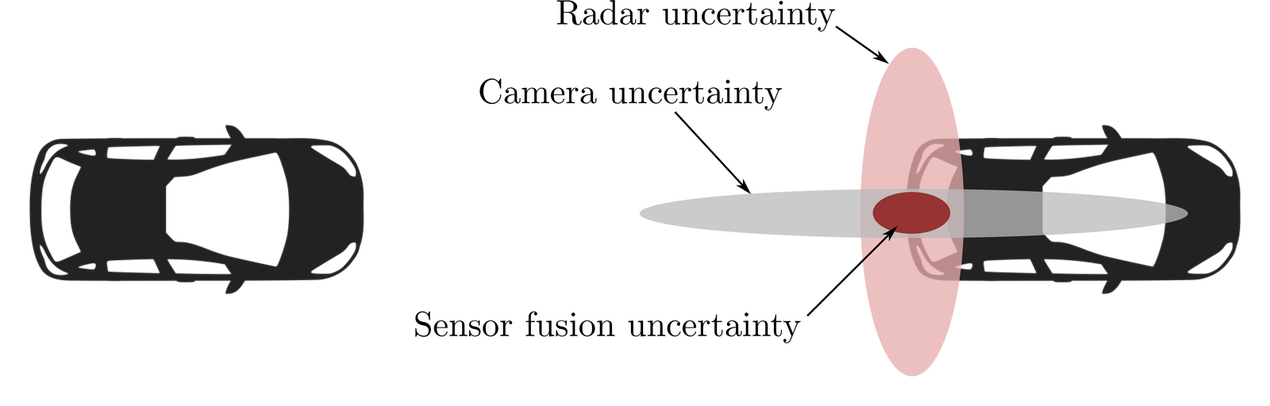}
    \caption[][-1\baselineskip]{Sensor fusion can reduce uncertainty by providing more well-rounded data. 
    For example, a radar sensor may provide good longitudinal distance accuracy but slightly less lateral accuracy, and a camera may provide poor range estimation but good lateral position estimation. 
    By fusing these two signals, the resulting position estimate can be accurate longitudinally and laterally.}
    \label{fig:reduceuncertainty}
\end{figure}

\begin{example}[Perception for autonomous driving]
    A self-driving car typically uses a combination of lidar and radar for distance sensing. 
    Lidar provides high-resolution geometric structure at short to medium range, while radar is more robust at longer ranges and in adverse weather\cite{blackman1999tracking}\sidenote{Radar is also generally more robust than lidar in fog, rain, or snow.}. 
    Cameras complement distance sensors: they provide high spatial density and rich appearance cues for object recognition and semantics, and they supply accurate bearing information to landmarks and obstacles.
    
    \cref{fig:reduceuncertainty} illustrates the idea in 2D.  
    Radar measurements often have strong longitudinal (range) accuracy but weaker lateral resolution; cameras, by contrast, deliver accurate lateral bearing but poor absolute depth. 
    Fusing the two yields a positional estimate that is precise both longitudinally and laterally.
\end{example}

\begin{example}[Localization for ground robots]
A wheeled robot might use GNSS together with wheel encoders to estimate position. 
GNSS keeps global position error bounded, while encoders provide fine-grained, high-rate short-term motion increments. 
When GNSS is degraded or lost, such as in urban canyons or under foliage, encoder odometry can bridge the gap.  
Conversely, when wheel slip corrupts odometry, GNSS can pull drift back into line. 
As we will see, treating these sources probabilistically allows a filter to down-weight whichever one is less reliable at any moment.
\end{example}

\begin{example}[Other common pairings]
Aerial robots often fuse a high-rate IMU with barometer and vision. 
The IMU stabilizes attitude and short-term motion, vision constrains drift and provides scale, and the barometer anchors altitude. 
Mobile manipulators combine joint encoders with depth sensing for precise end-effector placement while maintaining situational awareness. 
In all of these cases, no single sensor is sufficient across all operating regimes, but a carefully engineered combination is.
\end{example}

We will adopt a probabilistic perspective on fusion.
Let~$\x$ denote the latent state and let~$\mathcal{Z}=\{ \z^{(1)},\z^{(2)},\ldots \}$ be the set of measurements from all sensors. 
Under suitable assumptions, fusion is described by the posterior belief:
\begin{equation*}
p(\x\mid \mathcal{Z}) \;\propto\; p(\x) \prod_i p(\z^{(i)} \mid \x),
\end{equation*}
which we will realize with Kalman-family filters and their nonlinear and distributed variants. 
The same viewpoint also underpins modern learning-based systems, where we often fuse intermediate \emph{features} rather than raw measurements.

In this chapter, we first introduce a taxonomy of fusion modes and architectures, then develop Bayesian fusion using linear and nonlinear filters, including bias-aware modeling and conservative strategies for unknown correlations. 
We discuss practical issues such as spatial registration, time alignment, multi-rate/asynchronous updates, and out-of-sequence measurements. 
We then connect fusion to \emph{object tracking}, covering gating and data association (GNN/JPDA/MHT) as well as advanced random finite set (RFS) methods, and finally survey modern practice in feature- and BEV-level fusion, uncertainty calibration for learned modules, and cooperative (V2X) perception. 
Throughout, we tie the abstractions back to real robotic systems using examples like those above.

\subsection{A Taxonomy of Sensor Fusion}
\label{sec:fusion-taxonomy}

We begin by situating sensor fusion along three complementary axes: one describes the \emph{data} we receive from sensors, one characterizes the \emph{fusion problem}, and one describes the \emph{system architecture}. 
Thinking through all three lenses helps identify both an appropriate mathematical formulation and the right engineering design choices for a given application.

\paragraph{Data-related taxonomy.}
Real sensors are imperfect in several simultaneous ways. 
Measurements are uncertain due to stochastic noise, biased because of drifts and miscalibrations, and coarse due to finite resolution or quantization\cite{hall2001handbook}. 
Beyond these first-order issues, real systems often exhibit:
\begin{itemize}
  \item Correlations across streams, for example due to shared timing, mounting, vibrations, or processing pipelines.
  \item Disparities between modalities, such as different range envelopes and failure modes.
  \item Outright inconsistencies, such as outliers, contradictory readings, or disorder due to out-of-sequence arrivals.
\end{itemize}
Robust fusion methods must explicitly acknowledge these realities.  
In this chapter, we will model noise and bias (\cref{sec:ch15-bayes-linear}), guard against cross-stream double-counting when correlations are unknown (\cref{sec:ch15-ci}), and handle asynchrony and delayed packets with fixed-lag smoothing (\cref{sec:ch15-oosm}).

\paragraph{Fusion-related taxonomy.}
It is also useful to classify problems by \emph{what} we fuse.  
At the lowest level, we combine raw or lightly processed time series such as range readings or pixel measurements.
At an intermediate level, we fuse features or representations such as bird's-eye-view (BEV) representations or learned embeddings.
At a high level, we fuse decisions such as detections, tracks, or maps. 
These are often termed \emph{early}, \emph{mid}, and \emph{late} fusion.
All three appear in practice and impose different constraints on bandwidth, latency, and calibration.  

Orthogonal to the representational choice is the relationship among sensors, leading to \emph{competitive} fusion, \emph{complementary} fusion, and \emph{cooperative} fusion.
In competitive fusion, redundant measurements of the same quantity are combined to improve reliability via consistency checks.
In complementary fusion, we combine sensors that see different aspects of the world, such as lidar for short-range geometry and radar for long-range motion, to fill in gaps and reduce ambiguity.
In cooperative fusion, sensors recover information that no single one can provide alone, as when monocular vision acquires metric scale only when paired with inertial sensing\sidenote{For example, GNSS localization and stereo vision can be cooperatively fused because they measure fundamentally different properties of the environment.}.

In this chapter, we give each of these a probabilistic interpretation, treating every stream as a stochastic observation, and show how the choice of level and relationship guides the appropriate update rules and statistical tests (\cref{sec:ch15-bayes,sec:ch15-tracking}).

\paragraph{Architectural taxonomy.}
Finally, we classify fusion by \emph{where} the computation takes place.
In a \emph{centralized} architecture, raw data or features are transported to a fusion center that estimates the state using all information at once. 
This is statistically efficient but demands bandwidth and creates single points of failure.
In a \emph{decentralized} architecture, each platform or subsystem runs a local estimator and transmits summaries, such as tracks with covariances, to a higher layer that fuses posterior distributions. 
This reduces raw bandwidth but raises questions about cross-correlation. 
In a \emph{distributed} architecture, peers exchange beliefs or innovations over a communication graph and seek agreement via consensus or information sharing. 
Such architectures scale naturally to vehicle-to-everything (V2X) and multi-robot scenarios but must preserve \emph{consistency} when cross-covariances are unknown. 
We will see how information-form filters support decentralized and distributed updates, and how conservative schemes such as Covariance Intersection maintain correctness in the face of unknown dependencies (\cref{sec:ch15-architectures})\cite{olfatisaber2007consensus}.

\medskip
These three taxonomies—imperfections in the \emph{data}, the \emph{level} and \emph{relationship} of fusion, and the \emph{architecture} of computation—provide a scaffold for the remainder of the chapter. 
We move from Bayesian formulations and Kalman-family updates to the practicalities of registration and asynchrony, then on to multi-object tracking and modern feature/BEV-level fusion, referring back to this taxonomy to justify design choices along the way.

\subsection{Bayesian Approach to Sensor Fusion}
\label{sec:ch15-bayes}
In previous chapters, we introduced Bayes-filter-based algorithms for state estimation and localization. 
The very same viewpoint naturally solves \emph{sensor fusion}: each sensor contributes a probabilistic observation of the latent state, and Bayes’ rule combines these contributions into a posterior belief that is more precise and more reliable.

\paragraph{Beliefs, likelihoods, and Bayes’ rule.}
We model unknown quantities as random variables and represent knowledge as probability distributions. 
Let $\x_t$ denote the latent state at time $t$ and let $\mathcal{Z}_t=\{\z^{(i)}_t\}_{i=1}^m$ be the set of measurements from $m$ sensors at that time. 
Under the assumption that the measurements are conditionally independent given $\x_t$, the Bayesian update from prior $p(\x_t)$ is:
\begin{equation}
    p(\x_t \mid \mathcal{Z}_t) \;\propto\; p(\x_t)\,\prod_{i=1}^m p(\z^{(i)}_t \mid \x_t).
    \label{eq:bayes-multi-sensor}
\end{equation}
If this conditional independence assumption does not hold, we must either model the joint likelihood (including cross-covariances) or use conservative fusion strategies that remain valid under unknown correlations (\cref{sec:ch15-ci}). 
In either case, the \emph{product of likelihoods} in \cref{eq:bayes-multi-sensor} formalizes the intuition that multiple sensors together reduce uncertainty.

\paragraph{Why the Bayesian approach?}
The Bayesian viewpoint is attractive for several reasons:
\begin{itemize}
  \item It provides a \emph{unified}, interpretable representation of information: any modality that can be probabilistically modeled can be expressed as a likelihood over the state.
  \item It natively handles \emph{uncertainty}\sidenote{For example, the variance of a Gaussian posterior quantifies dispersion. Calibrated posteriors therefore allow principled gating and fault detection.}.
  \item It offers a principled update rule via Bayes’ theorem and naturally handles missing data, delayed packets, and novel observations.
\end{itemize}

\begin{example}[Probabilistic competitive fusion] 
\label{ex:competitive-fusion}
\theoremstyle{definition}
Consider two sensors measuring the same scalar quantity $x\in\mathbb{R}$, producing measurements $z_1$ and $z_2$ with Gaussian noise:
\begin{equation*}
p(z_1\mid x)=\mathcal{N}(z_1;\:x,\sigma_1^2),\qquad
p(z_2\mid x)=\mathcal{N}(z_2;\:x,\sigma_2^2).
\end{equation*}
Treating the likelihood as a function of $x$, their product is proportional to a Gaussian in $x$:
\begin{equation*}
p(z_1,z_2\mid x)\;\propto\; \mathcal{N}\!\left(x;\,\mu,\sigma^2\right),\quad
\mu=\frac{z_1\sigma_2^2+z_2\sigma_1^2}{\sigma_1^2+\sigma_2^2},\quad
\sigma^2=\frac{\sigma_1^2\sigma_2^2}{\sigma_1^2+\sigma_2^2}.
\end{equation*}
The maximum-likelihood estimate of $x$ is therefore a \emph{precision-weighted average} of $z_1$ and $z_2$, and the joint uncertainty strictly decreases: $\sigma^2<\min\{\sigma_1^2,\sigma_2^2\}$. 

If we also incorporate a Gaussian prior $x\sim\mathcal{N}(\mu_0,\sigma_0^2)$, the posterior remains Gaussian with:
\begin{equation*}
\sigma_{\text{post}}^{-2}=\sigma_0^{-2}+\sigma_1^{-2}+\sigma_2^{-2},\qquad
\mu_{\text{post}}=\sigma_{\text{post}}^2\!\left(\tfrac{\mu_0}{\sigma_0^2}+\tfrac{z_1}{\sigma_1^2}+\tfrac{z_2}{\sigma_2^2}\right),
\end{equation*}
and the earlier result is recovered in the uninformative-prior limit $\sigma_0^2\to\infty$.
\end{example}

\subsubsection{Linear-Gaussian Fusion: The Kalman Filter}
\label{sec:ch15-bayes-linear}
The Kalman filter, introduced in \cref{ch:approximate-filters}, is the linear-Gaussian instance of the Bayesian filter and a workhorse for sensor fusion. 
We assume linear dynamics:
\begin{equation*}
    \x_t = A_t \x_{t-1} + B_t \u_t + \bm{\epsilon}_t,\qquad
    \bm{\epsilon}_t \sim \mathcal{N}(\bm{0},\stateNoise_t),
\end{equation*}
and a linear measurement model:
\begin{equation*}
    \z_t = C_t \x_t + \bm{\delta}_t,\qquad
    \bm{\delta}_t \sim \mathcal{N}(\bm{0},\measNoise_t),
\end{equation*}
with a Gaussian belief~$\bel(\x_t)\sim\mathcal{N}(\bmu_t,\Sigma_t)$. 
Here $\stateNoise_t$ is the process-noise covariance and $\measNoise_t$ is the measurement-noise covariance, which may encode multiple sensors.

\paragraph{Prediction update.}
The Kalman filter prediction step is:
\begin{equation*}
\bar{\bmu}_t=A_t \bmu_{t-1}+B_t\u_t,\qquad
\bar{\Sigma}_t=A_t\Sigma_{t-1}A_t^\top+\stateNoise_t,
\end{equation*}
where $\bar{\bmu}_t$ and $\bar{\Sigma}_t$ denote the predicted mean and covariance at time $t$.

\paragraph{Measurement update (stacked multi-sensor form).}
Suppose $m$ sensors report at time $t$. 
We stack their measurements into a single vector:
\begin{equation*}
\z_t =
\begin{bmatrix}\z_t^{(1)}\\ \vdots \\ \z_t^{(m)}\end{bmatrix},\quad
C_t \!=\!
\begin{bmatrix}C_t^{(1)}\\ \vdots \\ C_t^{(m)}\end{bmatrix},\quad
\measNoise_t \!=\! \mathrm{blkdiag}\!\left(\measNoise_t^{(1)},\ldots,\measNoise_t^{(m)}\right),
\end{equation*}
where $C_t^{(i)}$ and $\measNoise_t^{(i)}$ are the measurement matrix and noise covariance for the $i$-th sensor. 
Then:
\begin{equation*}
\tilde{\z}_t=\z_t-C_t\bar{\bmu}_t,\quad
S_t=C_t\bar{\Sigma}_tC_t^\top+\measNoise_t,\quad
K_t=\bar{\Sigma}_tC_t^\top S_t^{-1},
\end{equation*}
and the update becomes:
\begin{equation*}
\bmu_t = \bar{\bmu}_t + K_t \tilde{\z}_t,\qquad
\Sigma_t=(I-K_tC_t)\bar{\Sigma}_t.
\end{equation*}
Sensors with smaller covariance in $\measNoise_t$ are automatically weighted more heavily in the Kalman gain\sidenote{This is visible in $K_t=\bar{\Sigma}_tC_t^\top(C_t\bar{\Sigma}_tC_t^\top+\measNoise_t)^{-1}$.}.

\paragraph{Information form (useful for decentralized/distributed fusion).}
Define the information matrix and vector by $Y_{t}=\Sigma_{t}^{-1}$ and~$\mathbf{y}_{t}=Y_{t}\bmu_{t}$. 
Each independent measurement block contributes:
\begin{equation*}
\Delta Y^{(i)}=C_t^{(i)\top}\measNoise_t^{(i)-1}C_t^{(i)},\qquad
\Delta\mathbf{y}^{(i)}=C_t^{(i)\top}\measNoise_t^{(i)-1}\z_t^{(i)}.
\end{equation*}
Fusion then reduces to summation:
\begin{equation*}
Y^+ = Y^- + \sum_i \Delta Y^{(i)}, \qquad
\mathbf{y}^+ = \mathbf{y}^- + \sum_i \Delta\mathbf{y}^{(i)}.
\end{equation*}
This additive structure makes it natural to combine multi-sensor updates or exchange increments over a network, as we will discuss in \cref{sec:ch15-architectures}.

\paragraph{Innovation tests and gating.}
Define the innovation~$\tilde{\z}_t$ and its covariance~$S_t$ as above. 
The \emph{Normalized Innovation Squared} (NIS):
\begin{equation*}
\text{NIS} = \tilde{\z}_t^\top S_t^{-1}\tilde{\z}_t,
\end{equation*}
follows a $\chi^2$ distribution of appropriate dimension under correct modeling. 
We will use this to gate outliers\sidenote{We revisit gating in \cref{sec:ch15-tracking-gating}.}, detect faults, and monitor consistency.

\paragraph{Augmenting the state (bias-aware fusion).}
A powerful trick is to augment the state $\x$ with nuisance parameters such as sensor biases or calibration terms. 
For an additive bias $\bm{b}$, we define the augmented state:
\begin{equation*}
\x' = \begin{bmatrix}\x\\ \bm{b}\end{bmatrix},\qquad
\bm{b}_{t}=\bm{b}_{t-1}+\bm{w}_t,\ \ \bm{w}_t\sim\mathcal{N}(\mathbf{0},\stateNoise_b),
\end{equation*}
and a measurement~$\z_t=C\x_t+\bm{b}_t+\bm{\delta}_t$ becomes linear in $\x'$ with measurement matrix $[C\ \ I]$. 
This allows the filter to learn biases online and prevents them from masquerading as state errors. 
Scale or misalignment biases can be handled similarly with appropriate parameterizations.

\paragraph{Nonlinear sensors: EKF and UKF.}
For nonlinear measurements $\z_t = h(\x_t)+\bm{\delta}_t$, we can either linearize $\measmodel$ via the Jacobian $H_t=\frac{\partial \measmodel}{\partial \x}\big|_{\bmu_t}$ and apply the \emph{extended Kalman filter} (EKF), or propagate sigma points through $\measmodel$ in the \emph{unscented Kalman filter} (UKF).
Both preserve the Bayesian structure, and the UKF generally performs better on strongly nonlinear problems. 
Bias augmentation and stacked updates carry over unchanged.

\begin{example}[Kalman filter multi-sensor fusion]
    \label{ex:car-sensor-fusion-kf}
    \theoremstyle{definition}
    Consider a self-driving car equipped with an IMU, a GNSS receiver, and a lidar sensor. 
    We estimate longitudinal position $p$, velocity $v$, and acceleration $a$ using a constant-acceleration kinematic model:
    \begin{equation*}
    \dot{p}=v,\qquad \dot{v}=a.
    \end{equation*}
    Discretizing with sampling time $T$ and allowing process noise $\bm{\epsilon}_t\sim\mathcal{N}(\mathbf{0},\stateNoise_t)$ gives
    \begin{equation*}
    \begin{bmatrix} p_{t+1}\\ v_{t+1}\\ a_{t+1}\end{bmatrix} =
    \underbrace{\begin{bmatrix}1 & T & \tfrac{T^2}{2}\\ 0 & 1 & T\\ 0 & 0 & 1\end{bmatrix}}_{A}
    \begin{bmatrix} p_t\\ v_t\\ a_t\end{bmatrix} + \bm{\epsilon}_t.
    \end{equation*}
    Suppose lidar and GNSS measure the position $p$, and the IMU measures the acceleration $a$:
    \begin{equation*}
    \begin{bmatrix} z_{\text{lidar},t}\\ z_{\text{gnss},t}\\ z_{\text{imu},t}\end{bmatrix} =
    \underbrace{\begin{bmatrix}1&0&0\\ 1&0&0\\ 0&0&1\end{bmatrix}}_{C}
    \begin{bmatrix} p_t\\ v_t\\ a_t\end{bmatrix} + \bm{\delta}_t,\quad
    \bm{\delta}_t\sim\mathcal{N}\!\Bigl(\mathbf{0},\
    \measNoise_t=\begin{bmatrix}\sigma_{\text{lidar}}^2&0&0\\[2pt]0&\sigma_{\text{gnss}}^2&0\\[2pt]0&0&\sigma_{\text{imu}}^2\end{bmatrix}\Bigr).
    \end{equation*}
    Here $\measNoise_t$ stacks the sensor variances, $\sigma^2_{\text{lidar}}$, $\sigma^2_{\text{gnss}}$, and $\sigma^2_{\text{imu}}$. 
    \cref{fig:fusionexample} shows that adding a lower-variance GNSS channel tightens the position estimate and reduces noise\sidenote{Even a noisier additional sensor can help by improving observability or by providing redundancy for fault detection, but the effect is more modest.}. 
    The relative weighting is handled automatically by the Kalman gain.
    \begin{figure}[ht!]
        \centering
        \includegraphics[width=0.85\textwidth]{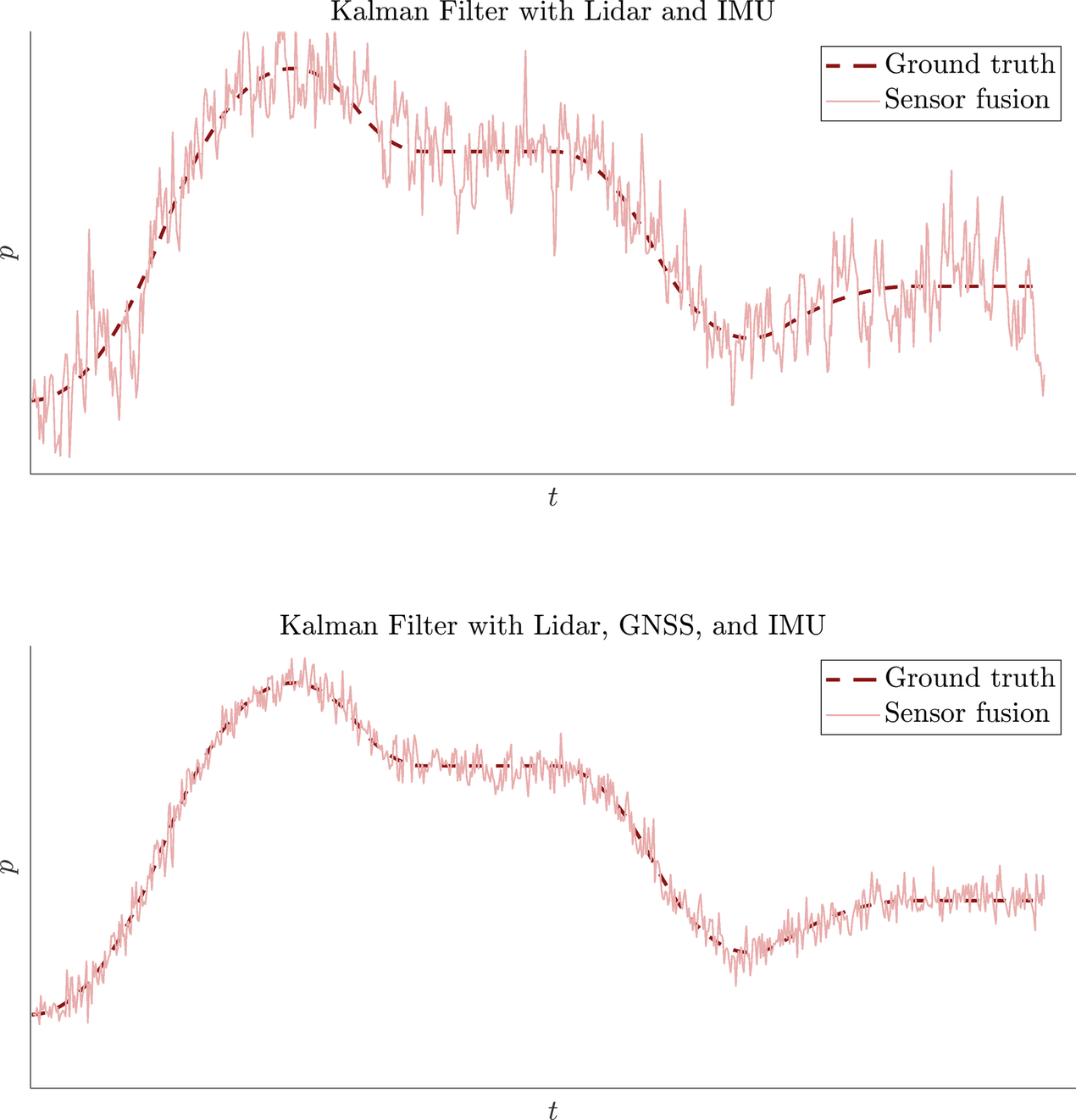}
        \caption{Kalman filter sensor fusion for \cref{ex:car-sensor-fusion-kf}. 
        The position of a vehicle is estimated using noisy lidar, GNSS, and IMU data, and the resulting estimate tracks the ground truth. 
        Adding the GNSS sensor improves the estimate through fusion.}
        \label{fig:fusionexample}
    \end{figure}
\end{example}

\subsubsection{Unknown Correlations and Conservative Fusion}
\label{sec:ch15-ci}
When fusing estimates from separate filters\sidenote{For example, in track-to-track fusion, decentralized networks, or when reusing map priors.}, cross-covariances are often \emph{unknown}. 
Naively fusing them as if they are independent produces overconfident posteriors. 
A principled, conservative alternative is \emph{Covariance Intersection} (CI), which combines Gaussian estimates without requiring cross-covariances. 
Given $(\bmu_1,\Sigma_1)$ and $(\bmu_2,\Sigma_2)$, CI defines:
\begin{equation*}
\begin{split}
\bmu_{\text{CI}}&=\Sigma_{\text{CI}}\big(\omega\,\Sigma_1^{-1}\bmu_1+(1-\omega)\,\Sigma_2^{-1}\bmu_2\big), \\
\Sigma_{\text{CI}}^{-1}&=\omega\,\Sigma_1^{-1}+(1-\omega)\,\Sigma_2^{-1},
\end{split}
\end{equation*}
with $\omega\in[0,1]$ chosen to minimize, for example, $\mathrm{tr}(\Sigma_{\text{CI}})$ or $\log\det \Sigma_{\text{CI}}$. 
CI guarantees \emph{consistency} irrespective of the true, unknown correlation structure and will reappear in our discussion of distributed fusion in \cref{sec:ch15-architectures}.
    
\medskip
In summary, the Bayesian lens turns the problem of sensor fusion into a precise algebra over beliefs and likelihoods. 
Linear-Gaussian assumptions yield closed-form Kalman filters with simple stacked updates and an additive information form, mild nonlinearities are handled by EKF/UKF variants, and practical wrinkles such as biases, asynchrony, and unknown correlations are addressed with state augmentation, smoothing, and conservative fusion. 
We build on this foundation in the next sections.

\subsection{Engineering Realities: From Models to Working Systems}
\label{sec:ch15-engineering}
The Bayesian rules from the previous section tell us how to combine beliefs and likelihoods. 
Real robots, however, add a layer of practical complexity: sensors are mounted somewhere on the body and point in particular directions, they operate at different rates with different latencies, networks drop or reorder packets, and multiple streams may be subtly correlated. 
This section bridges the gap between the clean algebra of Bayes’ rule and systems that survive contact with hardware.

We move from \emph{where} a sensor lives (spatial registration), to \emph{when} it speaks (time alignment and multi-rate updates), to what happens \emph{when it speaks late} (out-of-sequence measurements), and finally to staying \emph{honest} about information (unknown correlations and faults).

\subsubsection{Spatial Registration and Calibration}
\label{sec:ch15-registration}
Before any fusion can succeed, measurements must be made \emph{commensurate} with per-sensor \emph{intrinsics}\sidenote{Such as a camera’s focal length and distortion, an IMU’s scale factors and axis misalignment, or a radar’s range/Doppler scaling.} and \emph{extrinsics}\sidenote{The rigid transform from each sensor frame to the body frame and, where needed, from the body to a global frame.}. 
Calibration is not a one-time event: temperature changes, mechanical wear, and firmware updates can shift sensor parameters over time. 
A filter that appears increasingly overconfident often points to registration that has drifted out of spec.

A useful way to understand the effect of small miscalibration is to linearize it. 
If a measurement depends on the transform from body $B$ to sensor $S$, denoted ${}^{S}\!T_B$, through some function $h$, then around a nominal ${}^{S}\!\bar{T}_B$ we can write:
\begin{equation*}
\z
\;\approx\;
h\!\left({}^{S}\!\bar{T}_B\,\x\right)
\;+\;
J_{\!T}\,\boldsymbol{\eta}
\;+\;
\boldsymbol{\delta},
\end{equation*}
where $\boldsymbol{\eta}$ is a small pose perturbation, $J_{\!T}$ is its sensitivity, and $\boldsymbol{\delta}$ is sensor noise. 
This first-order view suggests two complementary strategies:
\begin{itemize}
  \item Treat extrinsics as known but \emph{uncertain}: inflate the measurement covariance by $J_{\!T}\,\Sigma_T\,J_{\!T}^\top$, where $\Sigma_T$ encodes extrinsic uncertainty. 
        This preserves consistency without changing the state.
  \item Augment the state with a minimal parameterization of the extrinsic and assign it a slow random-walk prior. 
        The filter then \emph{learns} small misalignments online rather than forcing other states to absorb them.
\end{itemize}

\begin{example}[Online camera–lidar extrinsics]
    \theoremstyle{definition}
    A mobile platform relies on a pre-calibrated extrinsic transform ${}^{\text{cam}}\!T_{\text{lidar}}$. 
    Over weeks, thermal drift subtly increases the innovation statistics (NIS) of the camera measurements. 
    By augmenting the filter state with a six-degree-of-freedom perturbation of ${}^{\text{cam}}\!T_{\text{lidar}}$ and adding sparse visual–depth correspondences as pseudo-measurements, the system recenters the alignment online. 
    A small process noise on the extrinsic states prevents overfitting transient effects.
\end{example}
    
\subsubsection{Time Alignment, Latency, and Stamps vs.\ Arrival}
\label{sec:ch15-timesync}
Sensors almost never report in lockstep time. 
They are clocked by different oscillators, traverse different processing pipelines, and communicate through different buses. 
The filter, however, should apply each measurement at the \emph{time it was taken}. 
Accurate \emph{measurement timestamps} are therefore critical; arrival times are, at best, indirect clues.

Hardware triggers or disciplined clocks, such as those using Network Time Protocol (NTP) or Precision Time Protocol (PTP), reduce relative drift.
Additionally, known fixed latencies, such as a camera’s ISP delay, should be subtracted so that time stamps reflect exposure times, not the end of the processing chain.
When an update is due at an intermediate time $\tau$ between filter steps, the state should be \emph{interpolated} to $\tau$ before applying the measurement.
For inertial navigation, IMU data can be \emph{preintegrated} over $(t_i,t_j]$ to produce a relative-motion pseudo-measurement consistent with $SE(3)$ geometry, removing the need to resample at the IMU rate.

\begin{example}[Stamp vs.\ arrival time matters]
    A radar packet arrives at $t=1.030$\,s with a timestamp $\tau=1.000$\,s. 
    If the filter treats the packet as if it were taken at $1.030$\,s, the update is effectively “time-shifted,” biasing the estimate and corrupting the covariance. 
    Maintaining a short, time-ordered buffer of states and applying the update at $\tau$ avoids this problem and sets the stage for proper handling of genuinely late packets in \cref{sec:ch15-oosm}.
\end{example}

\subsubsection{Multi-Rate and Asynchronous Updates}
\label{sec:ch15-multirate}
On a typical platform, the IMU runs at hundreds of Hertz, cameras at a few tens, lidar somewhere in between, and GNSS in single digits. 
The Kalman family naturally accommodates this: we predict forward to the next measurement time, and we update whenever a packet arrives. 

In software, it is convenient to maintain a priority queue keyed by measurement time. 
The filter repeatedly advances (predicts) to the earliest timestamp, next applies\sidenote{Stacking multiple measurements with the same stamp into a single block so that the innovation covariance $S_t$ reflects their joint effect.} all measurements with that timestamp, and then repeats.
This simple discipline—“predict to the stamp, then stack at the stamp”—prevents subtle double counting and preserves the meaning of NIS tests used later.

\subsubsection{Out-of-Sequence Measurements and Fixed-Lag Smoothing}
\label{sec:ch15-oosm}
Even with careful time stamping, networks reorder and delay packets. 
A classical example is GNSS delivered over a congested channel, where a position measurement computed at time $k-\ell$ may arrive at time $k$. 
Applying the measurement at the head of the filter timeline warps the uncertainty and can cause visible jumps in the estimate. 
The remedy is to keep a short history and re-solve the portion of the problem that lies within that window.
A high level procedure for fixed-lag Rauch–Tung–Striebel (RTS)\cite{rauch1965rts} smoothing for out-of-sequence measurements is:

\begin{enumerate}
\item Fix a lag $L$ and maintain a buffer of estimates $\{\bmu_{j|j},\Sigma_{j|j}\}_{j=k-L}^{k}$ together with the dynamics $(A,\stateNoise)$. 
      Suppose a measurement $(\z_{k-\ell},C,\measNoise)$ with $0\le\ell\le L$ arrives.
\item Run an RTS backward pass from $k$ back to $k-\ell$ to compute smoothed estimates $\{\bmu_{j|k},\Sigma_{j|k}\}$ (and, if needed, cross-covariances).
\item Insert the update at $k-\ell$ using $\bmu_{k-\ell|k}$ as the prior.
\item Re-propagate forward from $k-\ell$ to $k$, re-applying any later measurements in the buffer.
\item Commit the corrected states.
\end{enumerate}

The lag $L$ should exceed typical delay jitter.
If a packet falls outside the window, we can either discard it or assimilate it approximately with an inflated, time-shifted update.

\begin{example}[GNSS delay in a lidar–IMU EKF]
    An EKF runs at IMU rate ($200$\,Hz) with lidar odometry updates at $10$\,Hz. 
    GNSS packets occasionally arrive $300$\,ms late. 
    With a fixed lag $L=1$\,s, the filter back-smooths, inserts the GNSS update at its true time, and re-propagates. 
    The NIS distribution tightens, and the trajectory becomes free of the jumps that were present when late packets were applied at the head of the queue.
\end{example}

\subsubsection{Unknown Correlations, Double Counting, and Consistency}
\label{sec:ch15-correlation}

Fusion is not only about more data, it is about \emph{honest} information. 
Measurement streams may be correlated because they share process noise\sidenote{For example, the sensors are mounted on the same body.}, because they reuse the same map or prior, or because one stream already incorporates measurements from the other. 
If cross-covariances are known, they can be modeled explicitly. 
Often they are not, and fusing as if streams were independent yields overconfident covariances: NIS and NEES statistics will then fail $\chi^2$ checks.

Several responses are available:
\begin{itemize}
  \item When possible, exchange \emph{innovations} (residuals and their covariances) rather than full posteriors, since innovations are closer to independent across nodes.
  \item When that is insufficient, conservative fusion methods such as \emph{Covariance Intersection} (\cref{sec:ch15-ci}) combine Gaussians without any knowledge of cross-covariances and guarantee consistency at the price of modest optimality.
  \item In all cases, monitor empirical NIS/NEES against theoretical quantiles and treat persistent deviations as feedback that some correlation, timing error, or noise model has been overlooked.
\end{itemize}

\subsubsection{Fault Detection and Isolation}
\label{sec:ch15-fdi}
Even a well-registered, well-timed system must protect itself against bad data. 
Kalman filters already provide a convenient diagnostic: the innovation $\tilde{\z}$ and its covariance $S$ define the NIS $\tilde{\z}^\top S^{-1}\tilde{\z}$, which under a correct model follows a $\chi^2$ distribution of known dimension. 
Setting a gate at a chosen significance level rejects gross outliers, and tracking moving averages of the NIS per sensor exposes slow degradation.

When a sensor begins to misbehave, there are two complementary reactions:
\begin{itemize}
  \item \emph{Innovation-based adaptive estimation:} inflate that sensor’s measurement covariance $\measNoise$ according to a feedback law derived from recent NIS so that the filter automatically down-weights it.
  \item \emph{Structural response:} if the issue resembles a bias or scale error more than random noise, augment the state with the offending parameter and let the filter learn it.
\end{itemize}
In systems with redundancy\sidenote{Two or more sensors nominally measuring the same quantity.}, it is natural to maintain a per-sensor health score and to prefer the healthier stream while keeping the others alive for fault detection and graceful recovery.

\begin{example}[Rain on a camera, radar nominal]
    During a storm, the camera’s measurements intermittently fail the NIS gate, while radar remains nominal.
    The system inflates the camera’s measurement covariance $\measNoise$ and leans more heavily on radar for range and velocity. 
    Once the rain clears and the camera’s NIS returns to expected quantiles, its weight rises automatically. 
    Throughout, the reported covariance remains conservative, so downstream planners are not surprised.
\end{example}

\subsubsection{Numerical Stability and Practical Monitors}
\label{sec:ch15-numerics}
A few numerical practices are particularly useful.
First is the use of \emph{square-root filters}, which propagate a Cholesky factor of $\Sigma$ to be more stable than propagating $\Sigma$ directly, and help prevent negative-definite covariances due to round-off.
Second, when updating the covariance we can use the \emph{Joseph stabilized form}:
\begin{equation*}
    \Sigma_t=(I-K_tC_t)\,\bar{\Sigma}_t\,(I-K_tC_t)^\top + K_t\,\measNoise\,K_t^\top,
\end{equation*}
which preserves positive semidefiniteness in finite precision.
On the monitoring side, plotting NIS/NEES histograms with $\chi^2$ overlays, tracking the determinant and condition number of $\Sigma$, and logging gate hit-rates and stamp-minus-arrival statistics provide early warnings for timing and calibration regressions.

\paragraph{Takeaway.} 
Registration and timing turn raw measurements into \emph{commensurate evidence}; buffering and smoothing reconcile the past with the present; and attention to correlation and faults keeps the estimator honest.  
With these pieces in place, we can extend from single-state fusion to the challenges of \emph{object tracking} and the demands of \emph{distributed} systems, which we address next.

\subsection{Fusion Architectures}
\label{sec:ch15-architectures}
The Bayesian viewpoint tells us \emph{how} evidence should be combined and the architecture decides \emph{where} the combination happens and \emph{what} is exchanged. 
In practice, three recurring patterns emerge: centralized systems that gather everything in one place, decentralized systems that fuse local tracks at a higher layer, and distributed systems that reach agreement over a network. 
Each involves trade-offs in bandwidth, latency, robustness, and the ease of maintaining \emph{consistent} uncertainty (\cref{sec:ch15-correlation}).

\subsubsection{Centralized Fusion}
\label{sec:ch15-arch-central}
Centralized designs are conceptually simplest: raw measurements or lightly processed features are transported to a single estimator that maintains the posterior over the state. 
Mathematically, this is just the stacked update in \cref{sec:ch15-bayes-linear}, where at time $t$ we form:
\begin{equation*}
\z_t=\begin{bmatrix}\z_t^{(1)}\\[-2pt]\vdots\\[-2pt]\z_t^{(m)}\end{bmatrix},\quad
C_t=\begin{bmatrix}C_t^{(1)}\\[-2pt]\vdots\\[-2pt]C_t^{(m)}\end{bmatrix},\quad
\measNoise_t=\mathrm{blkdiag}\big(\measNoise_t^{(1)},\ldots,\measNoise_t^{(m)}\big),
\end{equation*}
compute the innovation $\tilde{\z}_t$ and its covariance $S_t$, and apply a single Kalman update. 
Centralized fusion is statistically efficient and gives the clearest path to proper \emph{gating}, bias handling, and OOSM smoothing (\cref{sec:ch15-timesync,sec:ch15-oosm}).

Its limitations are operational rather than mathematical.
Communicating raw sensor streams consumes bandwidth, creates central bottlenecks and single points of failure, and can be awkward when sensors live on different platforms, such as across vehicles and infrastructure. 
A common compromise is to centralize only within a platform or robot, fusing camera/lidar/radar/IMU locally and exposing higher-level artifacts—detections, tracks, occupancy—to the outside world.

\subsubsection{Decentralized Track-to-Track Fusion}
\label{sec:ch15-arch-decentral}
In a decentralized system, each subsystem or platform runs its own filter and publishes summarized beliefs—typically a state estimate with covariance and, possibly, a timestamp and a health score. 
A \emph{fusion node} then combines these ``tracks''\cite{zhang2010tracking}. 
The main question is how to fuse them without breaking consistency.

If the contributing tracks are conditionally independent given the true state\sidenote{For example, if they use disjoint raw measurements.}, fusion is straightforward in \emph{information form}. 
If a local node transforms a common prior $(Y^-,\mathbf{y}^-)$ into $(Y^+,\mathbf{y}^+)$, it can transmit its \emph{increment}:
\begin{equation*}
\Delta Y \equiv Y^+ - Y^-,\qquad
\Delta\mathbf{y} \equiv \mathbf{y}^+ - \mathbf{y}^-,
\end{equation*}
and the fusion node simply adds the increments from all sources:
\begin{equation*}
(Y,\mathbf{y}) \leftarrow (Y^-,\mathbf{y}^-) + \sum_i (\Delta Y^{(i)},\Delta\mathbf{y}^{(i)}).
\end{equation*}
In the linear-Gaussian case, these increments equal $C^\top \measNoise^{-1}C$ and $C^\top\measNoise^{-1}\z$ for that node’s local measurements. 
This “common-prior + increment” view keeps the algebra exact, but it requires that all nodes agree on (or communicate) the prior against which those increments were formed.

When two tracks are \emph{not} independent—because they share process noise, maps, or each other’s measurements—their cross-covariance is generally \emph{unknown}. 
Fusing as if independent is then dangerous (\cref{sec:ch15-correlation}). 
A safe default is \emph{Covariance Intersection} (\cref{sec:ch15-ci}), which guarantees consistency at the cost of some conservatism\cite{julier1997non}. 
If cross-covariances \emph{are} known\sidenote{For example, in a carefully engineered multi-radar system with a common process model.}, we can compute the optimal \emph{best linear unbiased estimator} (BLUE) weights using the joint covariance of the two track estimates. 
In practice, the engineering overhead to maintain those cross-terms often outweighs the marginal gain over CI.

\begin{example}[Track-to-track fusion with equivalent information]
Two drones estimate a shared target’s position. 
Each runs a local EKF and, at each second, exports the pair $(\Delta Y,\Delta\mathbf{y})$ computed from its local pre- and post-update information states. 
A ground station maintains a \emph{common prior} and updates it by addition. 
Because increments are additive and tied to a common prior, the result matches exactly what would have been obtained had the ground station received both raw measurement streams centrally. 
If a communications hiccup delays one drone’s packet, the ground station treats it as an out-of-sequence increment and inserts it using the same fixed-lag machinery as in \cref{sec:ch15-oosm}.
\end{example}

\subsubsection{Distributed Fusion by Consensus and Information Exchange}
\label{sec:ch15-arch-distributed}

Truly distributed systems have no fusion center. 
Instead, peers exchange messages over a communication graph and attempt to agree on the posterior. 
The \emph{information form} makes this natural: at each time step, node $i$ computes its local increment $(\Delta Y_i,\Delta\mathbf{y}_i)$ from its own measurements, then the network mixes these contributions so that everyone converges to the same sum.

One simple version is \emph{average consensus on information increments} using a symmetric, doubly stochastic mixing matrix $W=[w_{ij}]$ that respects the communication graph. 
Assume all nodes start from the same prior $(Y^-,\mathbf{y}^-)$. 
Each node initializes its local consensus state from its own increment:
\begin{equation*}
\tilde{\Delta Y}_i^{(0)}=\Delta Y_i,\qquad
\tilde{\Delta\mathbf{y}}_i^{(0)}=\Delta\mathbf{y}_i.
\end{equation*}
Then perform $r$ rounds of neighbor averaging:
\begin{equation*}
\tilde{\Delta Y}_i^{(s+1)}=\sum_{j} w_{ij}\,\tilde{\Delta Y}_j^{(s)},\qquad
\tilde{\Delta\mathbf{y}}_i^{(s+1)}=\sum_{j} w_{ij}\,\tilde{\Delta\mathbf{y}}_j^{(s)}\quad (s=0,\ldots,r-1).
\end{equation*}

When the graph is connected and $W$ is well chosen, $\tilde{\Delta Y}_i^{(s)}$ and $\tilde{\Delta\mathbf{y}}_i^{(s)}$ converge to the network-wide \emph{averages} of the increments. 
An exact centralized posterior is then recovered by multiplying by the number of nodes $n$:
\begin{equation*}
Y_i^+=Y^-+n\,\tilde{\Delta Y}_i^{(r)},\qquad
\mathbf{y}_i^+=\mathbf{y}^-+n\,\tilde{\Delta\mathbf{y}}_i^{(r)}.
\end{equation*}
In directed or time-varying graphs, push-sum or diffusion variants play the same role. 
In practice, a small number of rounds\sidenote{Even $r=1$ or $2$.} often captures most of the benefit.

Correlation again requires care. 
If nodes are connected through shared process models or re-used features, repeated neighbor mixing can re-inject the same evidence multiple times. 
Two mitigations are common.
First, we can exchange \emph{innovations} rather than full posteriors, since these tend to be closer to independent across nodes.
Second, we can replace consensus averaging with \emph{CI-consensus}: combining neighbors’ information conservatively when independence is doubtful so each node’s covariance remains an upper bound on its true error.

\begin{algorithm}[ht]
  \KwData{Common prior $(Y^-,\mathbf{y}^-)$, local measurements $\{\z_i\}$ giving $\Delta Y_i,\Delta\mathbf{y}_i$, neighbor set $\mathcal{N}(i)$, mixing weights $\{w_{ij}\}$ with $\sum_j w_{ij}=1$, number of nodes $n$, number of rounds $r$.}
    \KwResult{Posterior $(Y_i^+,\mathbf{y}_i^+)$.}
    \tcp{Initialize}
    $\tilde{\Delta Y}_i\leftarrow \Delta Y_i$\\ 
    $\tilde{\Delta\mathbf{y}}_i\leftarrow \Delta\mathbf{y}_i$\\
    \For{$k = 1$ \KwTo $r$}{
      	Get $(\tilde{\Delta Y}_j,\tilde{\Delta\mathbf{y}}_j)$ from neighbor node $j\in\mathcal{N}(i)$\\
      	$\tilde{\Delta Y}_i \leftarrow \sum_{j} w_{ij}\tilde{\Delta Y}_j$\\
      	$\tilde{\Delta\mathbf{y}}_i \leftarrow \sum_{j} w_{ij}\tilde{\Delta\mathbf{y}}_j$\\
      }
      $Y_i^+ \leftarrow Y^- + n\,\tilde{\Delta Y}_i$\\
      $\mathbf{y}_i^+ \leftarrow \mathbf{y}^- + n\,\tilde{\Delta\mathbf{y}}_i$\\
    \caption{Consensus information filtering (node $i$).}
    \label{alg:consensus-if}
\end{algorithm}

\subsubsection{What to Send: Raw Data, Features, or Tracks?}
\label{sec:ch15-arch-comm}

Architectures are inseparable from communication budgets. 
A useful mental model is to choose a \emph{bandwidth tier}:
\begin{itemize}
  \item \emph{Raw or ROI data}\sidenote{For example, point clouds or image crops.}: maximal accuracy and flexibility, but highest bandwidth, strict latency requirements, and potential privacy concerns.
  \item \emph{Features or bird's-eye-view (BEV) grids}: a strong trade-off in modern stacks; semantics are preserved, bandwidth is moderate, and time alignment can be handled at the feature level.
  \item \emph{Decisions (detections, tracks, occupancy)}: minimal bandwidth and simplest to distribute; best suited to decentralized and distributed fusion, but with the least flexibility for correcting upstream errors.
\end{itemize}

Two additional practicalities:
\begin{itemize}
  \item Messages should carry \emph{uncertainty} (covariances or credible intervals), not just point estimates, to enable principled fusion at the receiver.
  \item Timing matters as much as content: include measurement timestamps, not just send times, so that receivers can place information correctly on their own timelines (\cref{sec:ch15-timesync,sec:ch15-oosm}).
\end{itemize}

\begin{example}[Cooperative perception at an intersection]
\label{ex:ch15-arch-v2x}
Consider an urban intersection with an instrumented roadside unit (RSU) and vehicles approaching from multiple directions. 
Each vehicle maintains a centralized, on-board fusion stack (camera/lidar/radar/IMU) and publishes a stream of tracks with covariances. 
The RSU runs its own perception stack from elevated cameras and a 4D imaging radar.
A decentralized \emph{fusion server} aggregates vehicle and RSU tracks via equivalent information increments tied to a common prior at $10$\,Hz. 
Vehicles subscribe to this fused track set.

When bandwidth is plentiful, the RSU also publishes mid-level BEV features for regions of interest such as crosswalks. 
Nearby vehicles that can spare compute perform a short, two-round consensus step (\cref{alg:consensus-if}) on the BEV-derived information, which sharpens occupancy in occluded regions. 
During peak congestion, the system falls back to track-level CI fusion to preserve consistency under stronger correlations. 
The result is a layered architecture: centralized fusion \emph{within} each agent, decentralized fusion at the server for robustness, and briefly distributed fusion among peers when conditions allow.
\end{example}

\paragraph{Takeaway.} Centralized fusion is the gold standard when bandwidth and compute permit it. 
Decentralized fusion scales well and is straightforward when independence holds (or CI is used when it does not). 
Distributed fusion achieves resilience and coverage across a network, provided that messages preserve timing, carry uncertainty, and respect the difference between independence and correlation. 
We now turn from \emph{what} and \emph{where} to fuse to the closely related problem of \emph{tracking multiple objects} through time.

\subsection{Object Tracking}
\label{sec:ch15-tracking}
Sensor fusion becomes especially important when the goal is not just an ego state but a changing \emph{population} of objects: vehicles, pedestrians, drones, or landmarks that appear, move, occlude one another, and disappear. 
The task of \emph{object tracking} is to estimate, through time, both the continuous states, such as positions and velocities, and the discrete \emph{identities} of these objects given noisy, partial, and sometimes contradictory measurements. 

In this section, we build a practical tracker from motion and measurement models, add principled \emph{gating} to keep outliers at bay, address \emph{data association} to decide which detection belongs to which track, and define \emph{track management} to handle births, deaths, and occlusions. 
We close with an advanced view based on random finite sets and a brief tour of tracking-by-detection systems and evaluation metrics.

\subsubsection{States, Motion Models, and Measurements}
\label{sec:ch15-tracking-models}

We model each target $k$ at time $t$ with a state vector $\x^k_t$ and a measurement model that connects states to observed quantities. 
A common starting point for ground vehicles is a constant-velocity (CV) model in the plane:
\begin{equation*}
\x_t=\begin{bmatrix}x & y & \dot{x} & \dot{y}\end{bmatrix}^\top,\qquad
F(\Delta t)=
\begin{bmatrix}
1 & 0 & \Delta t & 0\\
0 & 1 & 0 & \Delta t\\
0 & 0 & 1 & 0\\
0 & 0 & 0 & 1
\end{bmatrix},
\end{equation*}
with white-acceleration process noise:
\begin{equation*}
\stateNoise(q,\Delta t)=q\,
\begin{bmatrix}
\frac{\Delta t^3}{3} & 0 & \frac{\Delta t^2}{2} & 0\\
0 & \frac{\Delta t^3}{3} & 0 & \frac{\Delta t^2}{2}\\
\frac{\Delta t^2}{2} & 0 & \Delta t & 0\\
0 & \frac{\Delta t^2}{2} & 0 & \Delta t
\end{bmatrix}.
\end{equation*}
For targets that turn, a coordinated-turn (CT) model augments the state with a turn rate $\omega$ and typically uses an EKF or UKF update. 
A constant-acceleration (CA) model can be more appropriate for pedestrians or stop-and-go traffic. 
The choice of model trades fidelity for simplicity; a good practice is to begin with CV and only move to CT or CA if innovation statistics suggest a benefit.

Measurements take many forms. 
A radar may report range $r$, bearing $\phi$, and sometimes radial velocity $\dot{r}$:
\begin{equation*}
h_{\text{radar}}(\x) =
\begin{bmatrix}
\sqrt{x^2+y^2}\\[2pt]
\mathrm{atan2}(y,x)\\[2pt]
\frac{x\dot{x}+y\dot{y}}{\sqrt{x^2+y^2}}
\end{bmatrix} + \boldsymbol{\delta},\quad \boldsymbol{\delta}\sim\mathcal{N}(\mathbf{0},\measNoise).
\end{equation*}
Camera detections may provide image-space bounding boxes and, if depth is available, 3D centroids in the ego frame. 
Lidar clusters similarly yield centroids and sometimes yaw estimates. 
We fold these into linear or linearized measurement models and apply the Kalman-family update from \cref{sec:ch15-bayes-linear} for each track.

\subsubsection{Gating and the Role of the Innovation}
\label{sec:ch15-tracking-gating}
Before tackling the combinatorics of association, we narrow the search space with \emph{gating}. 
Given a predicted measurement $\bar{\z}$ for a track and its innovation covariance $S$, a detection $\z$ is deemed \emph{compatible} if the (squared) Mahalanobis distance:
\begin{equation*}
m^2(\z) = (\z-\bar{\z})^\top S^{-1}(\z-\bar{\z}),
\end{equation*}
lies below a threshold from the $\chi^2$ distribution at a chosen confidence\sidenote{For example, $\alpha=0.99$ for a conservative, wide gate.}. 
This is exactly the NIS concept from \cref{sec:ch15-engineering}; under the model, $m^2$ follows $\chi^2_d$ for measurement dimension $d$. 

Gating dramatically reduces the clutter presented to the assignment solver: it retains almost all true matches while excluding most spurious ones.

\begin{example}[Ellipsoidal gates in practice]
With radar range–bearing measurements ($d=2$), a threshold near $\chi^2_{2,0.99}\approx 9.21$ carves out an ellipse in measurement space. 
In dense traffic, this can reduce candidates per track from dozens to a handful, turning an intractable assignment into a manageable one while preserving nearly all true associations.
\end{example}

\subsubsection{Data Association: From Nearest Neighbor to Probabilistic}
\label{sec:ch15-tracking-assoc}

After gating, we must decide which detection updates which track. 
The simplest approach is \emph{Global Nearest Neighbor} (GNN): we define a cost matrix $C_{ij}$ between track $i$ and detection $j$ (often using Mahalanobis distance), and solve the 2-D assignment problem with the Hungarian algorithm. Algorithm~\ref{alg:gnn} summarizes a standard gated GNN pipeline with gating and occlusion handling. 
Unassigned tracks receive missed-detection events; unassigned detections may seed new tracks.

GNN is fast and effective in relatively unambiguous scenes, but it commits to a single hypothesis even when multiple associations are plausible. 
Two families of methods hedge this ambiguity.

\paragraph{Probabilistic (joint) data association (PDA/JPDA).}
In Probabilistic Data Association (PDA), each track $i$ considers all detections in its gate and performs a \emph{mixture} update weighted by association probabilities $\{\beta_{ij}\}$ and a missed-detection probability $\beta_{i0}$. 
Joint PDA (JPDA) generalizes this to multiple tracks, computing a consistent set of probabilities across the scene.

For a track with prior $(\bmu^-,\Sigma^-)$ and each candidate detection $j$, we compute the innovation $\tilde{\z}_j$, its likelihood:
\begin{equation*}
L_j \propto \frac{\exp\!\left(-\tfrac{1}{2}\tilde{\z}_j^\top S^{-1}\tilde{\z}_j\right)}{\sqrt{\det S}},
\end{equation*}
and then normalize (with clutter intensity included) to obtain the $\beta_{ij}$. 
For each candidate association $j$, let $(\bmu_j^+,\Sigma_j^+)$ denote the conditional KF/EKF posterior, and let $(\bmu_0^+,\Sigma_0^+)=(\bmu^-,\Sigma^-)$ denote the missed-detection case. 
JPDA then matches the first two moments of this mixture:
\begin{equation*}
\bmu^+=\sum_{j=0}^{M}\beta_{ij}\,\bmu_j^+,
\end{equation*}
\begin{equation*}
\Sigma^+=\sum_{j=0}^{M}\beta_{ij}\left[\Sigma_j^+ + (\bmu_j^+-\bmu^+)(\bmu_j^+-\bmu^+)^\top\right].
\end{equation*}
The second term inside the brackets captures residual association ambiguity, such that even if each conditional posterior is sharp, disagreement among the candidate means keeps the final covariance honest. 
JPDA’s main cost is combinatorial in the worst case, but clustering or pruning unlikely joint events keeps it tractable.

\paragraph{Multiple-hypothesis tracking (MHT).}
Multiple-hypothesis tracking maintains a small forest of competing association histories and prunes them by likelihood. 
Each hypothesis carries its own set of track states. 
When new detections arrive, the tree branches over plausible assignments; pruning and N-scan backtracking keep the tree shallow. 
Deferred decisions across a short horizon can significantly improve performance in crowded, ambiguous scenes.

\begin{algorithm}[ht]
\KwData{Predicted tracks $\{\bmu_i,\Sigma_i\}$, detections $\{\z_j\}$, gating threshold $\tau$, cost function $C_{ij}$.}
\KwResult{Updated tracks, with occluded ones kept alive via motion-only prediction for a limited horizon.}
\For{each track $i$}{
	Compute $\hat{\z}_i$, $S_i$, and gate detections by $m^2\le\tau$.\\
	\For{each detection $j$}{
		Build $C_{ij}$, e.g., set $C_{ij}=m^2_{ij}$ for gated pairs and a large cost otherwise.\\
	}
}
Solve the 2-D assignment with the Hungarian algorithm.\\
Update matched tracks via KF/EKF.\\
For unmatched tracks, perform a \emph{missed-detection} update (increase uncertainty, decrease a survival score).\\
For unmatched detections, run initiation logic (\cref{sec:ch15-track-mgmt}).\\
\caption{GNN with gating and occlusion handling.}
\label{alg:gnn}
\end{algorithm}

\paragraph{Appearance and motion together.}
Motion models constrain \emph{where} a target could go; learned appearance cues, such as embeddings from image crops or lidar shape descriptors, constrain \emph{who} it likely is. 
A common composite cost is:
\begin{equation*}
C_{ij}=\lambda_m\, d^2_{\text{Mahalanobis}}(\z_j,\hat{\z}_i)+\lambda_a\,(1-\mathrm{cosSim}(\phi_j,\psi_i)),
\end{equation*}
where $\phi_j$ is a detection embedding and $\psi_i$ is a track’s appearance model. 
Appearance reduces identity switches during occlusions and interactions, particularly in camera-heavy setups.

\begin{example}[Ambiguity at a pedestrian crossing]
Two pedestrians cross paths inside each other’s gates. 
GNN alone risks swapping identities, but JPDA softens the update for both tracks across the ambiguous frames, while a modest appearance term, such as a color/texture embedding, stabilizes the assignment and preserves identities.
\end{example}

\subsubsection{Track Management: Birth, Death, and Occlusion}
\label{sec:ch15-track-mgmt}
Bookkeeping is a critical function for tracking modules. 
Good \emph{track management} balances eagerness to explain new detections with skepticism that avoids proliferating false tracks.

\paragraph{Initiation.}
Common strategies to initialize tracks include: M/N logic, which confirms a track only after $M$ hits in the last $N$ frames, and a score that accumulates matched updates and decays on misses.
Initial covariances should be broad enough to reflect detector uncertainty and any depth ambiguity.

\paragraph{Maintenance and missed detections.}
When a track receives no compatible detection, propagate its state through the motion model, inflate its covariance, and reduce its survival score. 
During short occlusions, this “coast” allows re-acquisition without identity breaks. 
Keep occluded tracks alive for at most $T_{\text{miss}}$ seconds or $N_{\text{miss}}$ frames, tunable to the scene.

\paragraph{Termination and hygiene.}
Delete tracks that fail confirmation or whose survival score falls below threshold. 
Merge or split tracks when they overlap persistently or when one detection consistently explains two weak tracks better than the reverse. 
Periodically purge stale hypotheses in MHT and stale appearance embeddings.

\subsubsection{Random Finite Set (RFS) Filters}
\label{sec:ch15-rfs}

When the number of targets varies and clutter is heavy, it is natural to treat the \emph{set} of objects as the fundamental random variable. 
RFS filters propagate distributions over \emph{sets} rather than over a fixed list of tracks.

\paragraph{PHD and GM-PHD.}
The Probability Hypothesis Density (PHD) filter evolves the first moment (intensity) of the target-set distribution, whose peaks correspond to likely targets. 
With linear-Gaussian models, the Gaussian-mixture PHD (GM-PHD) filter maintains a set of weighted Gaussians with birth and survival terms. 
This approach is efficient, handles births and deaths gracefully, and works well when targets are numerous but relatively weak, though it does not manage identity explicitly.

\paragraph{Labeled multi-Bernoulli and $\delta$-GLMB.}
For identity-aware tracking, labeled multi-Bernoulli (LMB) and $\delta$-GLMB filters maintain labeled tracks with existence probabilities. 
They offer principled handling of associations, births, and deaths within a Bayesian set framework. 
The cost is increased computation and bookkeeping, but in return they handle combinatorial association in a coherent manner and provide clean uncertainty accounting in heavy clutter.

RFS methods are not always necessary, but they are useful when scenes are dense, clutter rates are high, and identity is secondary to coverage (PHD) or when a full Bayesian treatment of multi-target tracking with identities ($\delta$-GLMB/LMB) is desired.

\subsubsection{Tracking-by-Detection and Modern Practice}
\label{sec:ch15-tbd}
Modern trackers often follow a \emph{tracking-by-detection} paradigm, where a perception module (classical or learned) produces frame-wise detections with uncertainties, a motion model predicts track states, and an association layer links detections to tracks. 
In camera-centric systems, learned appearance embeddings significantly improve identity stability.
In lidar/radar-centric systems, motion and geometry dominate, and appearance plays a smaller role. 

Two themes recur:
\begin{enumerate}
  \item \emph{Uncertainty matters.} Downstream gating and data association work much better when the detector exports calibrated confidence and geometric covariance (or a proxy such as a covariance in BEV cells). Calibrate these heads before fusion, for example with temperature scaling for classification and reliability diagrams for regression.
  \item \emph{Robustness to missing modalities.} In adverse weather or partial failures, radar or thermal imaging may carry the burden. 
        Design the tracker to operate with a subset of sensors by inflating the covariance of the missing modality, adjusting initiation thresholds, and tuning $T_{\text{miss}}$ accordingly.
\end{enumerate}

\subsubsection{Evaluation: Metrics and What They Mean}
\label{sec:ch15-tracking-metrics}

Tracking quality is multi-faceted: we care about detection quality, geometric accuracy, and identity preservation. 
Three widely used metrics capture different aspects:

\paragraph{MOTA/MOTP.}
Multiple Object Tracking Accuracy (MOTA) aggregates missed detections (FN), false positives (FP), and identity switches (IDSW) against the number of ground-truth objects (GT):
\begin{equation*}
\mathrm{MOTA} = 1 - \frac{\mathrm{FN} + \mathrm{FP} + \mathrm{IDSW}}{\mathrm{GT}}.
\end{equation*}
Multiple Object Tracking Precision (MOTP) summarizes localization error for correctly matched pairs. 
MOTA is simple and interpretable, but it can obscure trade-offs between detection and identity.

\paragraph{IDF1.}
The IDF1 metric measures the F1 score of correctly identified detections over all matches, focusing on identity preservation. 
It penalizes identity swaps more explicitly than MOTA and is useful when appearance cues play a central role.

\paragraph{HOTA.}
Higher Order Tracking Accuracy (HOTA) balances localization, detection, and association in a unified measure by scoring matched pairs over a range of thresholds and combining the resulting detection and association accuracies. 
It correlates better with human judgment in crowded scenes where identity stability matters.

A mature evaluation typically reports at least one association-aware metric (IDF1 or HOTA) alongside MOTA/MOTP and includes qualitative sequences that reveal behavior under occlusion, crossing, and sensor degradation. 
For safety-critical systems, also monitor \emph{calibration}: plot innovation/NIS statistics for associated pairs to ensure the tracker remains consistent across conditions.

\paragraph{Takeaway.} A reliable tracker is not a single algorithm but a disciplined combination of realistic motion models, measurements with honest uncertainties, ellipsoidal gating to filter clutter, association that respects ambiguity (using appearance where it helps), and careful track management. 
In dense or high-clutter regimes, RFS methods provide a principled alternative that scales gracefully. 
With these pieces in place, we return to the broader fusion story in the learning era—feature-level fusion in BEV, cooperative perception, and uncertainty calibration—which shape \emph{where} we fuse in modern stacks.

\subsection{Learning-Era Fusion: Features, BEV, and Cooperation}
\label{sec:ch15-learning}

The probabilistic view from earlier sections remains the backbone of modern perception, but practice has shifted in two important ways. 
First, many systems now fuse \emph{features} rather than raw measurements, often in a shared \emph{bird's-eye view} (BEV) representation that aligns modalities in space and time. 
Second, the locus of fusion has moved beyond a single box: vehicles and infrastructure exchange information, and filters increasingly sit downstream of learned modules whose outputs carry (or should carry) uncertainty. 
This section turns these trends into concrete design patterns.

\subsubsection{Where to Fuse: Early, Mid, or Late?}
\label{sec:ch15-learning-where}

Classical fusion, as in \cref{sec:ch15-bayes-linear}, stacked raw measurements and operated directly on likelihoods. 
This remains the gold standard when bandwidth is ample and calibration is impeccable. 
Modern stacks, however, often benefit from two additional levels that better reflect computational and communication realities.

\paragraph{Early fusion (raw space).}
Here we project measurements into a common geometric frame and combine them there.
For example, we may project lidar points colored by camera radiance, radar range–Doppler points registered in the ego frame, and stereo depth “lifted” into 3D. 
The reward is geometric precision and simple physics-based likelihoods. 
The cost is high bandwidth, strict time alignment, and sensitivity to small calibration errors: a milliradian of extrinsic yaw drift that is barely noticeable in an image can become a meter of error at long range when projected onto the ground.

\paragraph{Mid-level fusion (feature space).}
In mid-level fusion, each sensor produces features via a learned backbone, and those features are then fused in a representation designed for downstream tasks. 
BEV has become the workhorse: by lifting multiple camera views, lidar, and radar into a ground-aligned grid, we obtain a scene description that is geometry-aware, compact, and well matched to detection, tracking, and planning. 
BEV removes ego-motion, naturally accommodates occupancy and flow estimates, and plays well with both filters and learned modules.

\paragraph{Late fusion (decision space).}
At the other end of the spectrum we fuse \emph{decisions}: detections, tracks, or occupancy tiles with associated uncertainties. 
Late fusion is bandwidth-efficient and maps neatly onto decentralized and distributed architectures (\cref{sec:ch15-architectures}). 
The trade-off is reduced flexibility: once an upstream detector has committed to boxes or tracks, there are fewer opportunities to correct miscalibration or recover missed evidence.

No single level is universally best. 
A pragmatic recipe is:
\begin{enumerate}
  \item Centralize \emph{within} a platform at mid-level (BEV), where most accuracy-per-byte gains lie.
  \item Export late-level artifacts for cooperation across platforms.
  \item Maintain an early-level path only where the safety case demands it\sidenote{For example, an emergency braking stack that reads raw radar.}.
\end{enumerate}

\subsubsection{BEV and Transformer-Style Fusion}
\label{sec:ch15-learning-bev}

The BEV idea is conceptually simple but rich in practice.
We represent the local scene as a grid aligned with the ground plane and ego pose, and let each modality contribute to that grid in the way most natural for it. 
The payoff is a shared canvas on which geometry, semantics, and motion can be reasoned about jointly.

\paragraph{From cameras to BEV.}
Multi-view image features $\{\mathbf{f}_c\}$ extracted by a backbone are lifted into BEV either through explicit geometry, such as projecting along estimated depth, or attention mechanisms that aggregate image features at positions consistent with BEV queries.
In the geometric case, a pixel $(u,v)$ with depth $\hat{d}$ back-projects to the camera frame and then to the ego frame:
\begin{equation*}
\x_{\text{ego}} = {}^{\text{ego}}\!T_{\text{cam}}\; \hat{d}\,K^{-1}\!\begin{bmatrix}u\\v\\1\end{bmatrix},\quad
\text{bin}(\x_{\text{ego}})\mapsto \text{BEV cell}.
\end{equation*}
In the attention-based case, a BEV query at ground point $\mathbf{g}$ gathers evidence across cameras with weights that depend on viewing geometry and learned compatibility, sidestepping brittle monocular depth estimates.

\paragraph{Lidar and radar to BEV.}
Lidar contributes 3D points whose heights and intensities can be pooled (min/max/mean, or learned pooling) into BEV features. 
Radar adds range–Doppler–angle evidence highlighting long-range velocities and all-weather robustness. 
After alignment, the result is a multi-modal BEV tensor $B\in\mathbb{R}^{H\times W\times C}$ in which nearby cells carry detailed geometry and far cells carry coarser, velocity-centered context.

\paragraph{Temporal fusion.}
Scenes evolve and sensors report at different times. 
BEV benefits from temporal memory in two complementary forms:
\begin{itemize}
  \item \emph{Geometric memory:} warp the previous BEV by the ego motion between frames and aggregate it with the current BEV. This acts as a skip connection that respects kinematics.
  \item \emph{Learned memory:} apply spatiotemporal attention over a short buffer of BEV frames so the model can “remember” moving actors.
\end{itemize}
Both require ego motion used for alignment to be \emph{time-consistent} with feature timestamps (\cref{sec:ch15-timesync}). 
Otherwise, the network ends up compensating for misalignment rather than modeling the scene.

\paragraph{Outputs and uncertainty.}
Heads attached to the multi-modal BEV tensor $B$ predict detections, occupancy, flow, and other downstream quantities. 
To keep fusion principled, these heads should emit \emph{calibrated} confidences and, when possible, geometric covariances or credible intervals for positions and extents (\cref{sec:ch15-learning-uncertainty}). 
These outputs feed directly into gating and association (\cref{sec:ch15-tracking-gating,sec:ch15-tracking-assoc}) and determine how heavily a filter should trust each piece of evidence.

\subsubsection{From Features to Filters: The Adaptor Pattern}
\label{sec:ch15-learning-adaptor}

Learned modules speak in logits, heatmaps, and BEV tensors; filters expect pseudo-measurements with covariances. 
An \emph{adaptor} translates between these languages so that the Bayesian machinery from earlier sections can operate on learned outputs without any sleight of hand.

At its simplest, an adaptor takes a detection $(\hat{\bm{p}}, \Sigma_{\text{net}}, s)$—a position, a covariance proxy, and a confidence—and turns it into a measurement $\z$ with covariance $\measNoise$ suitable for a KF/EKF update. Algorithm~\ref{alg:feature-to-filter} shows a simple per-detection adaptor. 
The crucial step is \emph{calibration}: mapping $\Sigma_{\text{net}}$ to a covariance $\measNoise$ whose empirical NIS statistics match $\chi^2$ quantiles on held-out data. 
Once calibrated, learned detections and classical sensors inhabit the same probabilistic currency.

\begin{example}
    A BEV detector yields a 3D position estimate $\hat{\bm{p}}$ and a covariance proxy $\Sigma_{\text{net}}$ from a covariance head.
    We define $\measmodel(\x)=\bm{p}(\x)$ and update a platform-centric track with $\z=\hat{\bm{p}}$ and $\measNoise=\mathrm{calib}(\Sigma_{\text{net}})$, where $\mathrm{calib}$ is learned or fitted offline so that the NIS aligns with $\chi^2$ on validation sequences. 
    The update then proceeds identically to a classical sensor, including gating and fault handling.
\end{example}

\begin{algorithm}[ht]
\KwData{Network output $(\hat{\bm{p}}, \Sigma_{\text{net}}, s)$ with position $\hat{\bm{p}}$, covariance proxy $\Sigma_{\text{net}}$, confidence $s$, track prior $(\bmu^-,\Sigma^-)$, calibration map $\mathrm{calib}(\cdot)$, minimum confidence $s_{\text{min}}$.}
\KwResult{Updated track and logged NIS for calibration monitoring.}
\If{$s<s_{\text{min}}$}{
	Discard detection and return.\\
}
$\z\leftarrow\hat{\bm{p}}$\\
$\measNoise\leftarrow\mathrm{calib}(\Sigma_{\text{net}})$\\
Compute innovation $\tilde{\z}$ and NIS.\\
\If{\text{NIS} $\leq \chi^2_{d,\alpha}$}{
	Accept detection and update track with KF/EKF.
}
\Else{
	Down-weight or reject detection.
}
\caption{Feature $\rightarrow$ filter adaptor (per detection).}
\label{alg:feature-to-filter}
\end{algorithm}

\subsubsection{Cooperative Perception (V2X)}
\label{sec:ch15-learning-v2x}

A single vehicle’s field of view is limited by its own geometry and occluders such as other vehicles and buildings. 
By contrast, an intersection camera may “see around corners,” and neighboring vehicles can reveal what an ego vehicle cannot. 
Cooperative perception asks three questions: \emph{what} should we send, \emph{when} should we send it, and \emph{how} should the receiver fuse it while maintaining honest uncertainty?

\paragraph{What to send.}
The tiers from \cref{sec:ch15-arch-comm} apply directly. 
Raw or ROI snippets are rich but expensive, mid-level BEV tiles carry useful context at manageable bitrate, and late-level tracks and occupancy are lightweight and easy to fuse. 
In practice, a layered strategy often works best: share BEV features for critical regions, such as blind corners, and share tracks elsewhere.

\paragraph{When and how.}
Messages must carry \emph{measurement timestamps} so receivers can place them correctly in their own timelines (\cref{sec:ch15-timesync}). 
Relative pose between agents should be maintained by a small filter on $SE(3)$ with an associated covariance. 
That pose uncertainty should then propagate either into feature alignment (wider attention kernels, conservative warps) or into track fusion weights. 
Fusing as if poses were perfect is a quick way to become overconfident.

\paragraph{Fusion modes.}
Feature-level cooperation aligns and aggregates BEV tiles—often with a learned attention block that explicitly accounts for pose uncertainty. 
Track-level cooperation uses the decentralized and distributed methods of \cref{sec:ch15-architectures}; when independence is doubtful, \emph{Covariance Intersection} (\cref{sec:ch15-ci}) provides a conservative backstop that preserves consistency.

\begin{example}[Occlusion busting at a four-way stop]
    A vehicle approaches a four-way stop occluded by a truck. 
    The roadside unit (RSU) shares a narrow strip of BEV features covering the blind zone at $10$\,Hz. 
    The vehicle fuses these with its own BEV via attention, then initializes two pedestrian tracks with calibrated covariance. 
    A neighboring vehicle’s track messages arrive a moment later; CI fuses them conservatively with the ego tracks, shrinking uncertainty without overstating confidence. 
    The planner receives a consistent occupancy map and honest covariances.
\end{example}

\subsubsection{Differentiable Filtering and Hybrid Models}
\label{sec:ch15-learning-hybrid}

Filters and networks are complementary rather than competing tools. 
Three hybrid patterns recur in modern systems.

\paragraph{Learned dynamics residuals.}
We retain a physically grounded motion model (CV/CT) but allow a network to predict a residual acceleration or steering term from context (maps, intents, social cues). 
The filter then predicts with the combination of physics and learned residual, and the residual’s variance reflects confidence in the learned component. 
This improves short-term forecasting without abandoning structure.

\paragraph{Learned measurement models.}
Instead of hand-coding the measurement model $\measmodel(\cdot)$, we can learn a mapping from features to pseudo-measurements with a covariance head. 
The adaptor in \cref{sec:ch15-learning-adaptor} ensures that outputs are calibrated before they reach the filter, keeping NIS/NEES in check.

\paragraph{Back-propagating through filters.}
For end-to-end tuning, we can unroll a few filter steps and back-propagate through the Kalman updates to adjust the feeding network. 
However, some safeguards are important: keep gains and covariances positive definite, regularize to avoid collapsing uncertainty, and validate with held-out NIS/NEES so the filter remains a filter rather than a brittle function approximator.

\subsubsection{Uncertainty You Can Trust}
\label{sec:ch15-learning-uncertainty}
Neural modules are powerful, but their confidence estimates are often miscalibrated, and fusion depends on honest uncertainty to avoid gates admitting outliers and associations overcommitting. 
Three classes of tools help keep confidence in line with reality: calibration, deep ensembles (or evidential models), and conformal prediction.

\paragraph{Calibration (post-hoc).}
For classification, \emph{temperature scaling} rescales logits $z$ by a scalar $T$, chosen on a validation set, before applying softmax, improving calibration\cite{guo2017calibration}. 
For regression, we can fit an affine map from raw variance proxies to empirical errors so predicted variances match residuals. 
Expected calibration error (ECE) and reliability diagrams provide simple diagnostics and can be tracked across conditions.

\paragraph{Deep ensembles and evidential models.}
Small ensembles\sidenote{For example, three to five seeds.} average predictions and expose epistemic uncertainty via disagreement\cite{lakshminarayanan2017simple}. 
Evidential models predict parameters of a distribution over distributions\sidenote{For example, a Normal–Inverse-Gamma for scalar regression.}, so uncertainty grows in regions with limited training data\cite{amini2020deep}. 
Either way, we feed the resulting variance through the adaptor so that gates and gains respond quantitatively.

\paragraph{Conformal prediction for finite-sample guarantees.}
Conformal methods wrap any base predictor and produce prediction sets with coverage $1-\alpha$ without distributional assumptions\cite{angelopoulos2023conformal}. 
For detections, a simple nonconformity score is negative log-likelihood or $1-\mathrm{IoU}$ with ground truth on a calibration set. 
Given a $(1-\alpha)$ quantile $q_{1-\alpha}$ of scores, at test time we:
\begin{itemize}
  \item accept only predictions with score $\le q_{1-\alpha}$, or
  \item inflate their covariance until the score would fall below $q_{1-\alpha}$.
\end{itemize}
This yields explicit, finite-sample control of false exclusion at the level of detections flowing into the tracker and keeps downstream gating behavior predictable.

\subsubsection{Asynchrony, Events, and Adverse Weather}
\label{sec:ch15-learning-robust}

Two additional practical issues round out the learning-era picture.

\paragraph{Event cameras and asynchrony.}
Event sensors report brightness changes at microsecond latency with large dynamic range. 
They pair naturally with IMUs and frame cameras: the IMU stabilizes short-term motion, event sensors add blur-free edges, and frame cameras add texture. 
Fusion follows the same timing discipline as in \cref{sec:ch15-timesync}: respect timestamps, interpolate priors, and back-smooth when out-of-sequence events matter.

\paragraph{All-weather complements.}
Rain, fog, and snow degrade cameras and lidar before radar and thermal imagers fail. 
Radar and thermal should therefore be treated as complementary modalities, not afterthoughts. 
During training, \emph{modality dropout} in BEV fusion helps the network succeed when one input is missing. 
Online, the system inflates the covariance of degraded modalities and uses health scores (\cref{sec:ch15-fdi}) to adapt sensor weights. 
The goal is graceful degradation: performance should bend under adverse conditions but not break.

\paragraph{Takeaway.} Learning-era fusion does not replace the Bayesian core; it builds on it. 
BEV provides a geometry-aware workspace where heterogeneous features meet, cooperative perception broadens the field of view, hybrid models let learning fill in what physics leaves out, and calibration and conformal wrappers keep uncertainty honest.

\subsection{Summary}
\label{sec:ch15-summary}

Sensor fusion is not a single algorithm but a way of organizing information. 
We began with a probabilistic lens: sensors provide likelihoods over latent quantities, and Bayes’ rule combines them into a posterior (\cref{sec:ch15-bayes}). 
Under linear–Gaussian assumptions this reduces to the Kalman filter with stacked multi-sensor updates and an additive information form; with mild nonlinearities, we can extend to EKF/UKF. 
Along the way we introduced practical methods: augmenting states to absorb biases, gating with innovation tests to keep outliers from steering the estimate, and—when fusing external estimates—preferring conservative schemes such as Covariance Intersection over fragile independence assumptions.

To implement fusion in practice, we saw the importance of respecting the timing of measurements. 
Registration aligns frames so measurements are commensurate; timing discipline ensures updates occur at the \emph{measurement} timestamp, not arrival time; multi-rate and asynchronous updates are handled by “predicting to the stamp, then stacking at the stamp”; and late packets are reconciled by fixed-lag smoothing (\cref{sec:ch15-engineering}). 
These habits, though not glamorous, are what make filters behave in the real world as theory predicts.

Fusion architectures then decide \emph{where} fusion happens and \emph{what} is exchanged. 
Centralized systems are statistically clean but bandwidth-hungry, while decentralized track-to-track fusion scales more easily provided independence holds—or CI is used to maintain consistency when it does not. 
Distributed systems reach agreement by exchanging information over a graph, with consensus and CI-consensus as workhorses (\cref{sec:ch15-architectures}). 
The same care with timestamps and covariances that serves a single robot becomes the glue that binds cooperating agents.

We then explored object tracking, which reframes fusion as the estimation of entities that move, become occluded, and reappear. 
Practical trackers are built from modest parts: realistic motion models, measurements with honest uncertainty, ellipsoidal gating to reduce clutter, assignment algorithms that hedge ambiguity (GNN when scenes are clear, JPDA/MHT when they are not), and disciplined track management for births, deaths, and occlusions. 
In dense or high-clutter regimes, random-finite-set methods (PHD, LMB/\(\delta\)-GLMB) offer a principled alternative (\cref{sec:ch15-tracking}).

Finally, we examined how modern practice shifts both the \emph{form} of what we fuse and the \emph{locus} of fusion. 
Mid-level feature fusion in BEV provides a geometry-aware canvas where cameras, lidar, and radar contribute according to their strengths.
Cooperative perception extends that canvas across vehicles and infrastructure.
Hybrid and differentiable designs let learned modules supply residual dynamics or measurement models while the filter keeps uncertainty honest. 
Calibration and conformal wrappers bridge learned outputs to the Bayesian core so that confidence is earned, not assumed (\cref{sec:ch15-learning}). 

If there is a single theme to carry forward, it is that \emph{uncertainty is as important a piece of information as state or control}. 
When we model it, track it, and respect it—across sensors, across time, and across machines—fusion becomes not just a way to combine data but a method for building systems that remain useful when conditions are least friendly.

\paragraph{To learn more.}
\label{sec:ch15-reading}
%
%
For a rigorous bridge from Bayesian filtering to multisensor fusion, \citet{barshalom2001estimation} is a classic engineering text.
\citet{Simon2006} provides a comprehensive treatment of KF/EKF/UKF, including numerical issues such as Joseph-form updates. 
\citet{ThrunBurgardEtAl2005} remains a gentle but thorough introduction to probabilistic robotics, Bayes filters, and thinking in likelihoods. 
\citet{Gustafsson2010} offers a compact, practice-oriented overview of statistical sensor fusion. 
For broad taxonomies and systems perspectives, the multisensor fusion handbook by \citet{liggins2017handbook} is invaluable.

For gating, assignment, and multi-target tracking, \citet{blackman1999tracking} is the radar-informed standard that still underpins many engineered systems. 
\citet{stone2013bayesian} give a Bayesian perspective on multiple-target tracking and association that pairs well with JPDA/MHT. 
For the assignment layer itself, Kuhn’s Hungarian method\cite{kuhn1955hungarian} and Bertsekas’s auction algorithm\cite{bertsekas1988auction} are canonical references.
%
When populations vary and clutter is heavy, \citet{mahler2007statistical} develops RFS theory from first principles. 
For practitioners, the Gaussian-mixture PHD filter by \citet{vo2006gaussian} and the labeled RFS/$\delta$-GLMB family by \citet{vo2014labeled} provide concrete algorithms that scale to realistic scenes.

The optimization view of fusion is well covered by \citet{dellaert2012factor} for factor graphs and by \citet{kaess2012isam2} for incremental smoothing and mapping. 
For inertial navigation specifically, \citet{forster2016manifold} develop on-manifold preintegration, which connects cleanly to bias-aware EKF and fixed-lag smoothing.
%
For consensus and distributed information exchange over graphs, the tutorial by \citet{olfatisaber2007consensus} is the right starting point. 
When cross-covariances are unknown and independence is doubtful, Covariance Intersection\cite{julier1997non} provides a principled, conservative backstop.


Calibration of neural confidences is essential before learned outputs are fused with classical sensors. 
\citet{guo2017calibration} introduce temperature scaling for classification; deep ensembles\cite{lakshminarayanan2017simple} offer a robust baseline for epistemic uncertainty; and evidential regression is a compact alternative for heteroscedastic regression\cite{amini2020deep}. 
For distribution-free, finite-sample coverage that plugs directly into gating, \citet{angelopoulos2023conformal} provide a clear tutorial on conformal prediction.


\section{Exercises}
The starter code for the exercises provided below is available online through GitHub. 
To get started, download the code by running in a terminal window:

\begin{tcolorbox}[colback=gray!10]
\begin{minted}{bash}
    git clone https://github.com/StanfordASL/pora-exercises.git
\end{minted}
\end{tcolorbox}

We denote Problems requiring hand-written solutions and coding in Python with \adjustbox{height=2ex, valign=c}{\includegraphics{figs/write.png}} and \adjustbox{height=2ex, valign=c}{\includegraphics{figs/code.png}}, respectively.

\subsection*{\adjustbox{height=2ex, valign=c}{\includegraphics{figs/write.png}}\ Problem 1: Variance Reduction}
Consider the problem from \cref{ex:competitive-fusion} where we have two sensors that measure the same quantity with Gaussian noise.
In this exercise, derive the result from \cref{ex:competitive-fusion} that:
\begin{equation*}
\mu=\frac{z_1\sigma_2^2+z_2\sigma_1^2}{\sigma_1^2+\sigma_2^2},\quad
\sigma^2=\frac{\sigma_1^2\sigma_2^2}{\sigma_1^2+\sigma_2^2}.
\end{equation*}
Additionally, prove that from this result that $\sigma^2<\min\{\sigma_1^2,\sigma_2^2\}$. 

\subsection*{\adjustbox{height=2ex, valign=c}{\includegraphics{figs/code.png}}\ Problem 2: Kalman Sensor Fusion}
In this problem, you will explore using the Kalman filter for sensor fusion for a simple 1D autonomous car.
We will model the car's motion using the kinematic model:
\begin{equation*}
\dot{p} = v, \quad \dot{v} = a, \quad \dot{a} = j,
\end{equation*}
where $p$ is the position, $v$ is the velocity, and $a$ is the acceleration, and the control is the jerk, $j$.
Assuming we apply a constant jerk across each time step, we can discretize this model exactly with sampling time $T$ as:
\begin{equation*}
\begin{split}
p_{t+1} &= p_t + v_t T + \frac{1}{2}a_tT^2 + \frac{1}{6}jT^3, \\ 
v_{t+1} &= v_t + a_t T + \frac{1}{2}jT^2, \\
a_{t+1} &= a_t + j T.
\end{split}
\end{equation*}
We will consider three possible sensors:
\begin{enumerate}
\item An IMU that measures the acceleration, $a$, with zero-mean Gaussian noise with standard deviation $\sigma_{\text{IMU}}$.
\item A lidar sensor that measures the distance to a known object, which provides a measurement of the position $p$ with zero-mean Gaussian noise with standard deviation $\sigma_{\text{lidar}}$.
\item A GNSS sensor that measures the position $p$, but is not operating correctly and has a constant bias error, $b$, such that $z_{\text{GNSS}} = p + b$, and also zero-mean Gaussian noise with standard deviation $\sigma_{\text{GNSS}}$.
\end{enumerate}
In the notebook \colorcode{ch15/exercises/kalman\_sensor\_fusion.ipynb}, complete the following exercises:
\begin{enumerate}
    \item Implement the matrices $A$ and $B$ to define the system dynamics model:
    \begin{equation*}
    	\x_t=A \x_{t-1}+B\u_t,
    \end{equation*}
     based on the discrete time model above for the car.
     Then, implement the Kalman filter algorithm function \colorcode{kalman\_filter\_update}.
     \item For the sensor model:
     \begin{equation*}
    \z_t = C \x_t + \bm{\delta}_t,\qquad
    \bm{\delta}_t \sim \mathcal{N}(\mathbf{0},\measNoise),
\end{equation*}
implement the matrices $C$ and $\measNoise$ for each combination of sensor setups:
\begin{enumerate}
\item IMU only
\item IMU + lidar
\item IMU + lidar + GNSS
\end{enumerate}
How does the RMSE for the position estimate compare among these configurations?
How does the GNSS sensor's bias affect the estimate?
For the IMU only configuration, should we expect this to be able to estimate the position well\sidenote{For a more in-depth study, take a look at the function \colorcode{observable} defined in the utility library. This is a method from linear control systems theory that can give us a theoretical analysis of if a sensor configuration will provide us with sufficient information.}? 
\item Re-define the Kalman estimator matrices for an augmented system state that includes the GNSS sensor bias.
Run the provided code to see how this affects the state estimate when using the GNSS sensor.
\end{enumerate}
\newpage
\printbibliography[segment=\therefsegment,heading=subbibliography,title={References}]

\part{Robot Decision Making}
\chapter{Finite State Machines}
\label{ch:finite-state-machines}
\newrefsegment
The preceding parts of this book have endowed the robot with a comprehensive set of foundational competencies. 
In Part I, we explored how optimal control provides a powerful framework for generating and executing robot motion. 
Part II introduced the perceptual capabilities of the robot, examining the sensors and algorithms required to observe and interpret its environment. 
Part III endowed the robot with a representation of its own state within the world, covering the methods by which it estimates its pose and constructs maps of its surroundings.

In this chapter, and Part IV as a whole, we move to a higher level of abstraction: \emph{robot decision-making}. 
Here, the focus of decision-making shifts from the fine-grained details of physical motion to the strategic choices a robot must make to achieve its long-term goals.
To illustrate this distinction, consider a mission in which a robot must navigate from an initial location A to a pickup location B, retrieve a package, and deliver it to a destination C. 
Executing the motion from A to B relies on the planning and control techniques developed in Part I. 
Detecting and identifying the package requires the perceptual models of Part II, while successful navigation depends on the localization and mapping methods introduced in Part III. 
However, coordinating these capabilities---deciding to navigate first, then grasp the object, then proceed to the delivery location, and to monitor and recover from failures along the way---requires an additional layer of decision-making. 
This layer operates over a discrete set of task modes and action choices, complementing the robot's continuous physical state.

In this chapter, we introduce \emph{finite state machines} as a foundational framework for modeling and implementing discrete decision-making in robots\cite{KaelblingWhiteEtAl2011}.
We begin by providing a mathematical definition of a finite state machine in \cref{sec:fsm}, and then discuss some architecture options, computational challenges, and practical implementation approaches in \cref{sec:fsm-arch}.
Finally, in \cref{sec:fsm-limitations}, we discuss the main limitations of finite state machines that motivate more advanced decision-making frameworks, which we explore in Chapters \ref{ch:sequential-decision-making}-\ref{ch:imitation-learning}.

\section{A Mathematical Model of Discrete Decision-Making}
\label{sec:fsm}
Finite state machines (FSMs) are a computational modeling framework for systems that can be in one of a \emph{finite} number of discrete states at any given time.
This framework is used in a wide variety of disciplines, including electrical engineering, linguistics, computer science, philosophy, biology, and more.
We can use FSMs in several different ways, including to \emph{specify} a desired program or behavior, to \emph{model and analyze} a system's behavior, or to \emph{predict future behavior}.

Formally, we define a FSM by a finite set of states $S$, an input alphabet $I$, an output alphabet $O$, a next–state function $n: S\times I \to S$, and an initial state $s_0 \in S$.
Additionally, we can define an output function according to two standard conventions:
\begin{itemize}
    \item \emph{Mealy machine:} the output depends on the current state and the current input:
    $$
    o: S\times I \to O,\qquad o_t = o(s_t,i_t).
    $$
    \item \emph{Moore machine:} the output depends only on the current state:
    $$
    \tilde{o}: S \to O,\qquad o_t = \tilde{o}(s_t).
    $$
\end{itemize}
Graphically, we represent states $S$ as nodes and admissible transitions as directed edges. 
For a Mealy machine, each edge is typically labeled with an input/output pair $i/o$, indicating that when input $i$ is received in state $s$, the machine moves along that edge to $s^\prime=n(s,i)$ and produces output $o\!=\!o(s,i)$.
Equivalently, the output associated with a Mealy machine can be viewed as an annotation on each state–input pair $(s,i)$.
For a Moore machine, edges are labeled only by inputs\sidenote{Since outputs do not depend on inputs.} and each node is annotated with its output value.

\begin{figure}[t]
    \begin{center}
    \begin{tikzpicture}[node distance=2.4cm,>=stealth',auto]
      \tikzstyle{state}=[draw=black,rounded corners,minimum height=2.2em,minimum width=5.2em,thick,fill=red!30]
    
      \node[state](s0){$s_0$};
      \node[state, right of=s0, xshift=1.6cm, yshift=2.2cm](s1){$s_1$};
      \node[state, right of=s1, xshift=1.6cm, yshift=-2.3cm](s2){$s_2$};
    
      \draw[->,thick] (s0) to[bend left=12]
        node[pos=0.5,sloped,above=6pt,inner sep=1pt,fill=white,fill opacity=0.9,text opacity=1]
          {\small $i_1\,/\,o_1$} (s1);
    
      \draw[->,thick] (s0) to[bend right=8]
        node[pos=0.5,sloped,below=6pt,inner sep=1pt,fill=white,fill opacity=0.9,text opacity=1]
          {\small $i_0\,/\,o_0$} (s2);
    
      \draw[->,thick] (s1) to[bend left=10]
        node[pos=0.5,sloped,above=6pt,inner sep=1pt,fill=white,fill opacity=0.9,text opacity=1]
          {\small $i_0\,/\,o_1$} (s2);
    
      \draw[->,thick] (s2) to[bend left=10]
        node[pos=0.5,sloped,below=6pt,inner sep=1pt,fill=white,fill opacity=0.9,text opacity=1]
          {\small $i_2\,/\,o_0$} (s1);
    \end{tikzpicture}
    \end{center}
    \label{fig:fsm}
    \caption{A graphical representation of a FSM with states $S = \{s_0, s_1, s_2\}$, inputs $I = \{i_0,i_1,i_2\}$ and outputs $O = \{o_0, o_1\}$. 
    The directed edges correspond to the next-state functions and the output associated with each edge is defined by the output function. 
    For example, in this FSM, we show the transition $n(s_0, i_1) \xrightarrow{} s_1$ in the top left, along with the corresponding output $o(s_0, i_1) \xrightarrow{} o_1$.}
\end{figure}
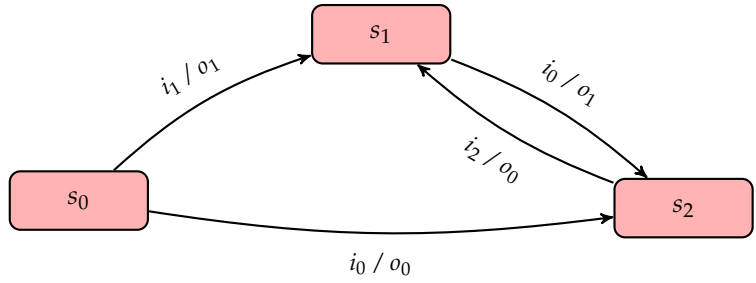

In what follows, we adopt the Mealy convention. 
Unless stated otherwise, we assume a deterministic\sidenote{In a deterministic FSM, each state has only one transition for each possible input.} FSM with initial state $s_0$, discrete time index $t=0,1,2,\dots$, next state $s_{t+1}=n(s_t,i_t)$, and output $o_t=o(s_t,i_t)$. 
This matches the graphical convention in \cref{fig:fsm}, where nodes are states, directed edges encode input-driven transitions, and edge labels include both the triggering input and the resulting output.

\begin{example}[Parking Gate Control] 
    \label{ex:parkinggate}
    \theoremstyle{definition}
    Consider a parking gate control problem where the goal is to raise the gate when a car arrives, and then lower the gate when the car has passed. 
    We assume sensors (or software events) indicate when a car is detected/cleared at the gate and when the gate reaches its end stops. 
    The control actions are raising, lowering, or holding the gate position fixed. 
    Note that in the real world, the position and velocity of the gate can vary continuously between the \emph{down} and \emph{up} positions. However, we use a higher-level discrete abstraction for the overall logic.
    
    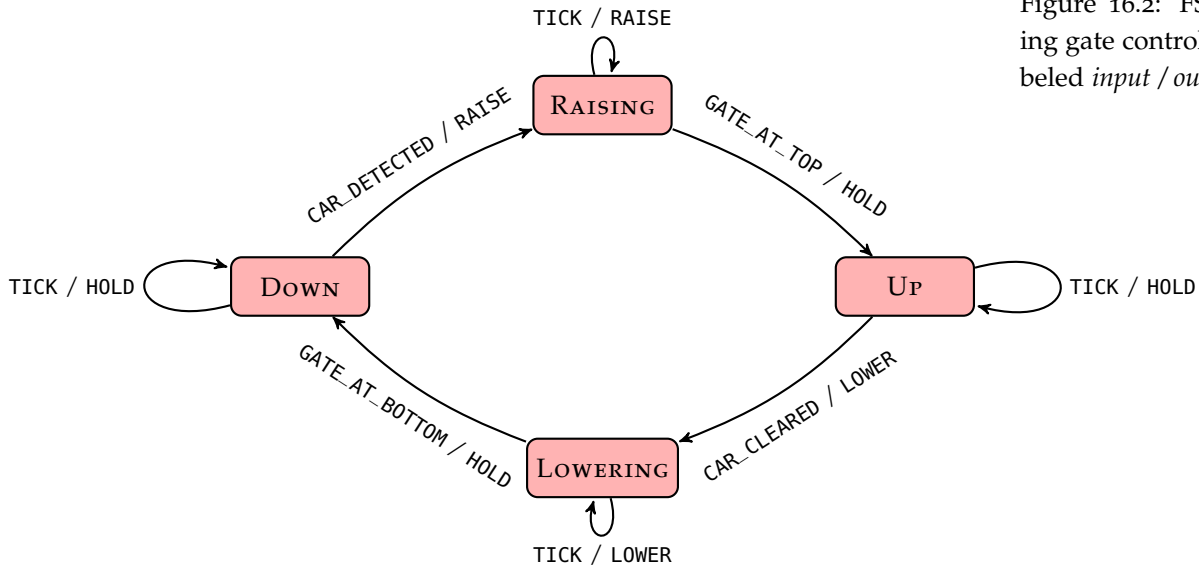
\begin{figure}[t]
        \begin{center}
        \begin{tikzpicture}[node distance=2.4cm,>=stealth',auto]
        \tikzstyle{state}=[draw=black,rounded corners,minimum height=2.2em,minimum width=5.2em,thick,fill=red!30]
        
        \node[state](down){\textsc{Down}};
        \node[state, right of=down, xshift=1.6cm, yshift=2.4cm](raising){\textsc{Raising}};
        \node[state, right of=down, xshift=1.6cm, yshift=-2.4cm](lowering){\textsc{Lowering}};
        \node[state, right of=raising, xshift=1.6cm, yshift=-2.4cm](up){\textsc{Up}};
        
        \draw[->,thick] (down) to[bend left=12]
        node[pos=0.5,sloped,above=6pt,inner sep=1pt]{\small \texttt{CAR\_DETECTED} / \texttt{RAISE}} (raising);
        \draw[->,thick] (raising) to[bend left=12]
        node[pos=0.5,sloped,above=6pt,inner sep=1pt]{\small \texttt{GATE\_AT\_TOP} / \texttt{HOLD}} (up);
        \draw[->,thick] (up) to[bend left=12]
        node[pos=0.5,sloped,below=6pt,inner sep=1pt]{\small \texttt{CAR\_CLEARED} / \texttt{LOWER}} (lowering);
        \draw[->,thick] (lowering) to[bend left=12]
        node[pos=0.5,sloped,below=6pt,inner sep=1pt]{\small \texttt{GATE\_AT\_BOTTOM} / \texttt{HOLD}} (down);
        
        \path[->,thick]
        (raising) edge[loop above] node{\small \texttt{TICK} / \texttt{RAISE}} ()
        (lowering) edge[loop below] node{\small \texttt{TICK} / \texttt{LOWER}} ()
        (down) edge[loop left] node{\small \texttt{TICK} / \texttt{HOLD}} ()
        (up) edge[loop right] node{\small \texttt{TICK} / \texttt{HOLD}} ();
        
        \end{tikzpicture}
        \end{center}
        \caption{FSM for the parking gate controller. Edges are labeled \emph{input / output}.}
        \label{fig:parkinggate}
        \end{figure}

    We model the FSM in Mealy style with states:
    \[
    S \definedas \{\textsc{Down},\textsc{Raising},\textsc{Up},\textsc{Lowering}\},
    \]
    an input alphabet:
    \[
    I \definedas \{\texttt{CAR\_DETECTED},\ \texttt{CAR\_CLEARED},\ \texttt{GATE\_AT\_TOP},\ \texttt{GATE\_AT\_BOTTOM},\ \texttt{TICK}\},
    \]
    and outputs:
    \[
    O \definedas \{\texttt{RAISE},\ \texttt{LOWER},\ \texttt{HOLD}\}.
    \]
    Here \texttt{TICK} is a periodic control-cycle event that lets the machine command continuous motion between end-stop events.
    
    We define the next-state and output maps \(n(s,i)\) and \(o(s,i)\) with representative rules:
    \[
    \begin{aligned}
    &n(\textsc{Down},\texttt{CAR\_DETECTED})=\textsc{Raising}, \quad &&o(\textsc{Down},\texttt{CAR\_DETECTED})=\texttt{RAISE},\\
    &n(\textsc{Raising},\texttt{GATE\_AT\_TOP})=\textsc{Up}, \quad &&o(\textsc{Raising},\texttt{GATE\_AT\_TOP})=\texttt{HOLD},\\
    &n(\textsc{Raising},\texttt{TICK})=\textsc{Raising}, \quad &&o(\textsc{Raising},\texttt{TICK})=\texttt{RAISE},\\
    &n(\textsc{Up},\texttt{CAR\_CLEARED})=\textsc{Lowering}, \quad &&o(\textsc{Up},\texttt{CAR\_CLEARED})=\texttt{LOWER},\\
    &n(\textsc{Lowering},\texttt{GATE\_AT\_BOTTOM})=\textsc{Down}, \quad &&o(\textsc{Lowering},\texttt{GATE\_AT\_BOTTOM})=\texttt{HOLD},\\
    &n(\textsc{Lowering},\texttt{TICK})=\textsc{Lowering}, \quad &&o(\textsc{Lowering},\texttt{TICK})=\texttt{LOWER},\\
    &n(\textsc{Down},\texttt{TICK})=\textsc{Down}, \quad &&o(\textsc{Down},\texttt{TICK})=\texttt{HOLD},\\
    &n(\textsc{Up},\texttt{TICK})=\textsc{Up}, \quad &&o(\textsc{Up},\texttt{TICK})=\texttt{HOLD}.
    \end{aligned}
    \]
    
    \noindent \cref{fig:parkinggate} shows the graphical representation of the FSM.
\end{example}

\section{Finite State Machine Architectures}
\label{sec:fsm-arch}
\begin{figure}[t]
    \centering
    \includegraphics[width=0.65\textwidth]{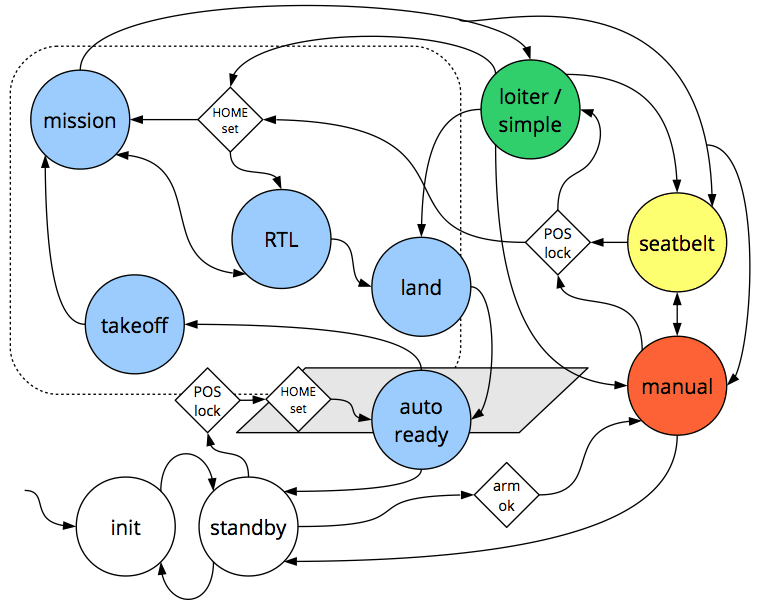}
    \caption{A graphical representation of the FSM for PX4. 
    We can see that even for a relatively small number of states, the FSM is quite complex in order to model the full behavior of the system.
     Image retrieved from diydrones.com.}
    \label{fig:px4fsm}
\end{figure}

One important practical disadvantage of FSMs is that their complexity does not scale well with system complexity, and, generally speaking, it can be time consuming and challenging to design FSMs for practical robotic systems.
To reduce complexity as much as possible, we must carefully choose the appropriate set of states to represent the system, and even with a well-defined set of states the interactions and transitions between states can be complex and hard to specify.
For example, \cref{fig:px4fsm} shows a graphical representation of the FSM for the popular open source flight software PX4\sidenote{PX4 is a flight control software for drones and other unmanned vehicles. See \url{https://px4.io/} for more information.}.
Specifying the full behavior for a system like this can lead to a complex FSM, even if there are not very many states.

At a high level, three complementary techniques help manage this complexity: 
\begin{itemize}
    \item \emph{State Minimization}, which merges behaviorally equivalent states to remove redundancy.
    \item \emph{Hierarchical State Machines}, which allow states to be nested within other states to create a hierarchy of states.
    \item \emph{Composition}, where larger state machines are built from smaller, simpler state machines.
\end{itemize}

\subsubsection{State Minimization}
A standard way to reduce an FSM without changing its input–output behavior is \emph{partition refinement}.
Two states are \emph{equivalent} if, for every input string, they produce the same output sequence (Mealy) or the same state outputs (Moore) and transition to equivalent states.
Partition refinement starts from a coarse partition---states that are immediately distinguishable by their outputs---and repeatedly refines the blocks by splitting states whose next states fall into different blocks for some input.
When the process reaches a fixed point, merging the states within each block yields a minimal machine with the same behavior.
We provide an example of this procedure in \cref{ex:sequence}.
\begin{example}[Finite State Machine State Reduction] 
\label{ex:sequence}
\theoremstyle{definition}
Consider a FSM that detects the input sequences 010 or 110. 
\cref{tab:sequence} lists the next-state function and the output function for each state and input. 
We can see that the states are the partial sequences and a \emph{Reset} state, $S \definedas \{0,1,00,01,10,11,\text{Reset}\}$, the inputs are $I \definedas \{0,1\}$, and the outputs are the booleans $O \definedas \{\text{True}, \text{False} \}$ that indicate if the sequence 010 or 110 has been created. 
For example, if the current partial sequence is 01 and a 0 is input, the next state will be the \emph{Reset} state and the output will be \emph{True}.
\begin{table}[ht]
\centering
\begin{tabular}{|l|
>{\columncolor[HTML]{C0C0C0}}l |l|
>{\columncolor[HTML]{C0C0C0}}l |l|}
\hline
State, $s$ & $n(s,0)$ & $n(s,1)$ & $o(s,0)$ & $o(s,1)$ \\ \hline
Reset      & 0        & 1        & False    & False    \\ \hline
0          & 00       & 01       & False    & False    \\ \hline
1          & 10       & 11       & False    & False    \\ \hline
00         & Reset    & Reset    & False    & False    \\ \hline
01         & Reset    & Reset    & True     & False    \\ \hline
10         & Reset    & Reset    & False    & False    \\ \hline
11         & Reset    & Reset    & True     & False    \\ \hline
\end{tabular}
\caption{FSM for a sequence detector that accepts digits 0 and 1 and outputs True if the sequences 010 or 110 are generated.}
\label{tab:sequence}
\end{table}

We can now simplify this FSM by removing redundant states.
To do so, we begin with the initial partition that groups states based on their output behavior:
\begin{equation*}
\begin{split}
\{\text{Reset}, 0, 1, 00, 10\}&: \text{always leads to a False output},\\
\{01,11\}&: \text{does not always lead to False output}.
\end{split}
\end{equation*}
We then further partition these sets based on the next-state function until we cannot make any further partitions. 
In the first step, we partition the set $\{\text{Reset}, 0, 1, 00, 10\}$ into:
\begin{equation*}
\begin{split}
\{\text{Reset}, 00, 10\}&: \text{cannot transition to \{01,11\}},\\
\{0, 1\} &: \text{can transition to \{01,11\}},
\end{split}
\end{equation*}
and then partition $\{\text{Reset}, 00, 10\}$ into:
\begin{equation*}
\begin{split}
\{\text{Reset}\}&: \text{can transition to } \{0, 1\},\\
\{00, 10\}&: \text{cannot transition to } \{0, 1\}.
\end{split}
\end{equation*}
After applying the partition refinement procedure, the original seven states $\{0,\allowbreak 1,\allowbreak 00,\allowbreak 01,\allowbreak 10,\allowbreak 11,\text{Reset}\}$ are reduced to four states, $S_\text{new} = \{\{01,11\}, \allowbreak\{0, 1\}, \allowbreak \{00, 10\}, \allowbreak \text{Reset}\}$.
The resulting machine, shown in \cref{tab:sequence2}, is therefore an equivalent\sidenote{Equivalent here meaning it has the same input–output behavior.} but reduced FSM.
\begin{table}[ht]
\centering
\begin{tabular}{|l|
>{\columncolor[HTML]{C0C0C0}}l |l|
>{\columncolor[HTML]{C0C0C0}}l |l|}
\hline
State, $s$                        & $n(s,0)$  & $n(s,1)$                          & $o(s,0)$ & $o(s,1)$ \\ \hline
Reset                             & \{0,1\}   & \{0,1\}                           & False    & False    \\ \hline
\{0,1\}                           & \{00,10\} & \cellcolor[HTML]{FFFFFF}\{01,11\} & False    & False    \\ \hline
\cellcolor[HTML]{FFFFFF}\{00,10\} & Reset     & Reset                             & False    & False    \\ \hline
\cellcolor[HTML]{FFFFFF}\{01,11\} & Reset     & Reset                             & True     & False    \\ \hline
\end{tabular}
\caption{Reduced FSM for the 010 or 110 sequence detector.}
\label{tab:sequence2}
\end{table}
\end{example}

\subsubsection{Hierarchical FSMs}
In some cases there are states that are not strictly equivalent but are closely related in behavior.
A common way to manage such structure is to use \emph{hierarchical finite state machines} (HFSMs), also known as \emph{Statecharts}\cite{HarelEtAl1987}.
HFSMs introduce \emph{super-states}\sidenote{Also called \emph{composite states}.} that group together related states into a higher-level state, and \emph{generalized transitions} that allow transitions to and from these super-states.
This reduces diagram clutter and mitigates state explosion by allowing behavior to be factored and reused across related modes.
Compared to flat FSMs, HFSMs support modularity, abstraction, and reuse of shared transitions at higher levels of the hierarchy.
\citet{HarelEtAl1987} and \citet{Alur2015} provide an in-depth treatment of hierarchical FSMs, including formal definitions, semantics, and algorithms for analysis and verification.

\subsubsection{Compositions}
We can also compose individual state machines in a variety of ways depending on their input/output behavior, including \emph{cascade} compositions, \emph{parallel} compositions, and \emph{feedback} compositions. 

\begin{marginfigure}
    \centering
    \includegraphics[width=0.8\textwidth]{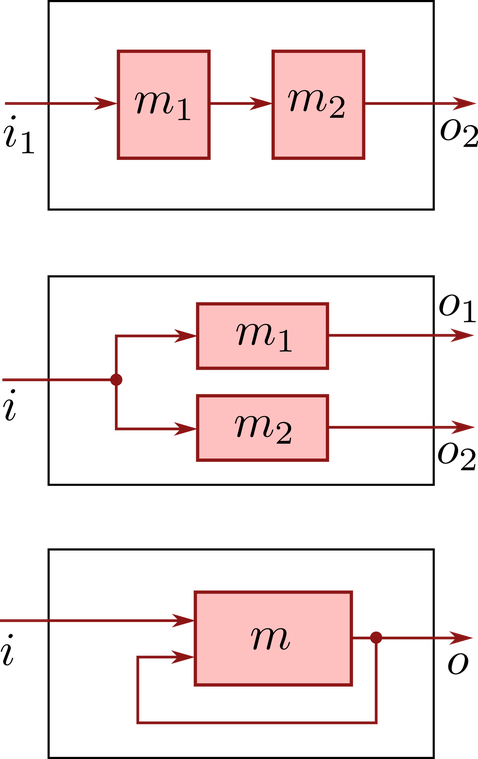}
    \caption{Cascade, parallel, and feedback compositions of FSMs.}
    \label{fig:cascade}
\end{marginfigure}

An example of each of these composition types is shown in \cref{fig:cascade}.

\paragraph{Cascade (serial) composition.}
Let $M_1=(S_1,I_1,O_1,n_1,o_1,s_{1,0})$ and $M_2=(S_2,\allowbreak I_2,\allowbreak O_2,\allowbreak n_2,\allowbreak o_2,\allowbreak s_{2,0})$ be two FSMs and assume a wiring map $\rho:O_1\to I_2$ (often $\rho$ is the identity after renaming).
The cascade composition $M=M_2\circ_\rho M_1$ is defined as the FSM $M=(S,I,O,n,o,s_0)$ where:
\begin{equation}
    \begin{aligned}
        S &= S_1 \times S_2, \\
        I &= I_1, \\
        O &= O_2, \\
        n((s_1,s_2),i) &= (n_1(s_1,i), n_2(s_2,\rho(o_1(s_1,i)))), \\
        o((s_1,s_2),i) &= o_2(s_2,\rho(o_1(s_1,i))), \\
        s_0 &= (s_{1,0},s_{2,0}).
    \end{aligned}
\end{equation}
Intuitively, $M_1$ processes the external input $i$ and produces an output $o_1(s_1,i)$, which is then fed into $M_2$ as input via the wiring map $\rho$.

\paragraph{Parallel (synchronous) composition.}
Parallel compositions combine two FSMs that share the same input alphabet and operate simultaneously on the same input.
Let $M_1=(S_1,I,O_1,n_1,o_1,s_{1,0})$ and $M_2=(S_2,I,O_2,n_2,o_2,s_{2,0})$ be two FSMs with the same input alphabet $I$.
The parallel composition $M=M_1\parallel M_2$ is defined as the FSM $M=(S,I,O,n,o,s_0)$ where:
\begin{equation}
    \begin{aligned}
        S &= S_1 \times S_2, \\
        I &= I, \\
        O &= O_1 \times O_2, \\
        n((s_1,s_2),i) &= (n_1(s_1,i), n_2(s_2,i)), \\
        o((s_1,s_2),i) &= (o_1(s_1,i), o_2(s_2,i)), \\
        s_0 &= (s_{1,0},s_{2,0}).
    \end{aligned}
\end{equation}

\paragraph{Feedback composition.}
Feedback compositions connect (part of) an FSM's output back to its input, creating a closed-loop system.
The closed-loop machine is well-defined if the induced equations have a unique solution for the input given the output.
Otherwise, the feedback composition is said to be ill-formed.

\section{Limitations of Finite State Machines}
\label{sec:fsm-limitations}
FSMs provide a clear, simple, and formally verifiable framework for discrete decision-making, which makes them attractive for implementing basic robot behaviors. 
However, their effectiveness as the primary control architecture for complex autonomous robots is fundamentally limited. 
Although the architectural techniques discussed in \cref{sec:fsm-arch} can alleviate some practical issues, they do not address the core limitations of the FSM paradigm. 
In practice, FSMs are best suited to highly structured environments in which the set of relevant situations and required responses is small, predictable, and can be exhaustively anticipated by a designer.

As the complexity of a robot's task and environment increases, FSM-based systems become difficult to scale. 
The number of states required to accurately represent the system can grow combinatorially with the number of factors that influence decision-making, including both the robot's internal operating mode and aspects of the external world such as object configurations or the behavior of other agents. 

FSMs also tend to exhibit brittle behavior when deployed outside the situations explicitly anticipated by their designers. 
Because all transitions and responses must be hand-specified, the robot can only react meaningfully to inputs for which logic has been defined in advance. 
When confronted with novel objects, unmodeled environmental changes, or unexpected sensor readings, the FSM lacks a mechanism for reasoning about new information or generalizing from prior experience. 
As a result, reliable performance is difficult to achieve in open-ended or dynamic environments without extensive manual engineering.

Another limitation of FSMs is the absence of an intrinsic notion of optimality. 
Standard FSMs describe which behaviors are permissible, but they do not provide a formal way to evaluate or compare alternative actions in terms of cost, reward, or long-term objectives. 
While it is possible to encode heuristically chosen preferences through careful state and transition design, the FSM framework itself does not support principled decision-making based on the optimization of a defined performance criterion.
The system simply executes the logic that has been predefined.

Finally, the deterministic nature of FSMs makes them a poor match for the uncertainty inherent in real-world robotics. 
Sensor measurements are noisy, action outcomes are often stochastic, and the robot's internal representation of the world is typically incomplete or approximate. 
Although designers can introduce states that qualitatively represent uncertainty, such as hypothesized or likely conditions, FSMs do not provide a principled mechanism for updating beliefs or making decisions based on probabilistic information.

Taken together, these limitations---poor scalability, sensitivity to unanticipated situations, and the inability to reason explicitly about optimality and uncertainty---motivate the use of more expressive decision-making frameworks. 
In the next chapters, we will introduce decision-making frameworks which support optimization under stochastic dynamics, as well as extensions that address partial observability, and learning-based approaches that allow robots to acquire complex behaviors from data rather than relying solely on manual specification.

\subsection{Summary}
\label{subsec:fsm-summary}
In this chapter, we introduced finite state machines as a mathematical model for systems with discrete states and transitions.
We defined the components of an FSM, including states, input and output alphabets, state transition functions, and output functions.
We explored how FSMs can be represented using state diagrams and transition tables, providing visual and tabular representations of their behavior.
Recognizing that FSMs can rapidly grow in complexity, we discussed three architectural strategies for managing this complexity: state minimization, hierarchical finite state machines, and mechanisms for composing FSMs.

\paragraph{To learn more.}
For a rigorous and comprehensive introduction to finite state machines within the broader context of computation, a classic resource is the textbook by \citet{Sipser1996} on the theory of computation. 
The seminal paper by \citet{HarelEtAl1987} is essential reading for a deep understanding of hierarchical FSMs, which it introduced as Statecharts. 
For a more modern and formal textbook treatment of hybrid systems, including hierarchical and concurrent state machines, see \citet{Alur2015}. 
Finally, for the application of these concepts in the wider context of AI for robotics, we refer to the course by \citet{KaelblingWhiteEtAl2011}. 

\section{Exercises}
The starter code for the exercises provided below is available online through GitHub. 
To get started, download the code by running in a terminal window:

\begin{tcolorbox}[colback=gray!10]
\begin{minted}{bash}
    git clone https://github.com/StanfordASL/pora-exercises.git
\end{minted}
\end{tcolorbox}

We denote Problems requiring hand-written solutions and coding in Python with \adjustbox{height=2ex, valign=c}{\includegraphics{figs/write.png}} and \adjustbox{height=2ex, valign=c}{\includegraphics{figs/code.png}}, respectively.

\subsection*{\adjustbox{height=2ex, valign=c}{\includegraphics{figs/write.png}}\ Problem 1: State Machine}
In this problem, you will create a finite state machine for a simple autonomous machine of your choice.
For the system of your choice, define the set of states $S$, the input alphabet $I$, the output alphabet $O$, the next–state function $n: S\times I \to S$, and the initial state $s_0 \in S$.
Define the output function using the Mealy machine convention.
Draw a diagram of the state machine similar to \cref{fig:parkinggate}.
\newpage
\printbibliography[segment=\therefsegment,heading=subbibliography,title={References}]
\chapter{Sequential Decision Making and Dynamic Programming}
\label{ch:sequential-decision-making}
\newrefsegment
In \cref{ch:finite-state-machines}, we introduced finite state machines as a structured and explicit way to model the logical flow of a robot's behavior, allowing us to hand-design a set of rules that govern its actions in response to events. 
For well-defined tasks with a limited number of states, a carefully crafted finite state machine is an effective and interpretable way to implement a robot's decision-making logic.
However, the very structure that makes finite state machines clear also imposes fundamental limitations, especially as the complexity and uncertainty of the robot's environment grow.

This chapter introduces a more general, optimization-based framework that directly addresses these limitations.
Rather than manually specifying decision rules, we formulate decision-making as a mathematical optimization problem, allowing optimal actions to be computed automatically for complex, multi-step tasks, even under uncertainty.
The central computational tool explored in this chapter is \emph{dynamic programming}, a powerful algorithmic paradigm for solving sequential decision-making problems.
Dynamic programming operates by decomposing complex, long-horizon problems into a sequence of simpler, nested subproblems that can be solved efficiently. 
As we will see, it also provides the theoretical foundation for modern learning-based approaches like reinforcement learning (\cref{ch:reinforcement-learning}) and imitation learning (\cref{ch:imitation-learning}).

We begin in \cref{subsec:det-decision-making} by applying dynamic programming to deterministic decision-making problems, where the robot's actions have certain and predictable outcomes.
We then extend this framework in \cref{subsec:stoch-decision-making} to stochastic decision-making problems, specifically Markov decision processes (MDPs), which explicitly account for uncertainty in the environment.
Finally, in \cref{subsec:limitations_dp}, we discuss the limitations of classical dynamic programming methods, thereby motivating the learning-based approaches introduced in later chapters.

\medskip
\medskip
\subsection{Deterministic Sequential Decision Making}
\label{subsec:det-decision-making}
We begin our study of dynamic programming by introducing a simple, yet extremely general, formulation of sequential decision making.
Despite its apparent simplicity, this formulation captures a wide range of problems arising in robotics, control, operations research, and artificial intelligence.
We first consider the deterministic case, where the outcome of each action is fully predictable, before addressing stochasticity in the next section.

In this formulation, time is modeled as a sequence of \emph{discrete} stages at which decisions are made. 
That is, we consider a discrete-time setting\sidenote{The continuous-time counterpart to dynamic programming is the Hamilton-Jacobi-Bellman (HJB) equation. While the HJB equation is beyond the scope of this chapter, we refer interested readers to \citet{Bertsekas2000} for a comprehensive treatment of continuous-time decision-making.}, where the evolution of the system is described by a difference equation of the form:
\begin{equation}
\label{eq:detmodel}
\x_{t+1} = \dynmodel_t(\x_t, \u_t), \quad t = 0, \dots, T-1,
\end{equation}
where $\x_t \in \R^\statedim$ denotes the system state at time step $t$, $\u_t \in \R^\controldim$ is the control applied at that step, and $\dynmodel_t$ specifies how the state evolves.
The integer $T$ defines a finite planning horizon.

At each time step, not all controls may be available.
We therefore associate with each state $\x_t$ a set of admissible controls, denoted by $\controlspace(\x_t)$, and impose the constraint:
\begin{equation}
\label{eq:SDMPconstraints}
\u_t \in \controlspace(\x_t), \quad t = 0, \dots, T-1.
\end{equation}
No particular structure is assumed for $\controlspace(\x_t)$.
Depending on the application, it may be a finite set of discrete actions, a continuous region of allowable inputs, or a state-dependent subset encoding physical, logical, or resource limitations.

The objective of the decision making problem is specified through an additive cost function defined over the planning horizon:
\begin{equation}
\label{eq:detcost}
J(\x_0, \u_0, \dots, \u_{T-1}) =
g_T(\x_T) + \sum_{t=0}^{T-1} g_t(\x_t, \u_t),
\end{equation}
where $g_t$ represents the stage cost incurred at time $t$ and $g_T$ is a terminal cost applied to the final state.
The additivity of the cost over time is a central structural assumption: it is this property that enables the decomposition of the problem into simpler subproblems, which lies at the heart of dynamic programming.
No assumptions are made regarding smoothness, convexity, or time invariance of the cost functions.

The deterministic sequential decision making problem can be formally defined as follows:
\begin{definition}[Deterministic Sequential Decision Making Problem]
\label{def:det-decision-making-problem}
Given the discrete-time system \eqref{eq:detmodel}, the control constraints \eqref{eq:SDMPconstraints}, and the additive cost function \eqref{eq:detcost}, the deterministic sequential decision making problem can be stated as the following optimization problem:
\begin{equation} 
	\label{eq:dproblem}
	\begin{split}
	J^*(\x_0) = \underset{\u_t, \:\:t=0,\dots,T-1}{\text{minimize}} & \quad g_T(\x_T) + \sum_{t=0}^{T-1} g_t(\x_t, \u_t), \\
	\text{subject to} & \quad \x_{t+1} = \dynmodel_t(\x_t, \u_t), \quad t = 0, \dots, T-1, \\
	& \quad \u_t \in \controlspace(\x_t), \quad t = 0, \dots, T-1.
	\end{split}
\end{equation}
\end{definition}

As we will see throughout the remainder of this chapter, the central goal of dynamic programming is to compute a solution to Problem~\eqref{eq:dproblem} in the form of an optimal \emph{closed-loop} control policy:
\begin{equation}
\begin{split}
\pi = & \{\pi_0, \ldots, \pi_{T-1}\}, \\
\u_t^* = & \pi_t^*(\x_t), \quad t = 0, \dots, T-1,
\end{split}
\end{equation}
where the policy $\pi$ denotes a sequence of functions $\pi_t$ mapping any state $\x_t$ into a control $\u_t$\sidenote{In the deterministic setting considered here, the system evolution is fully predictable, and closed-loop and open-loop optimal solutions therefore coincide. The true advantages of closed-loop policies emerge mainly in the stochastic setting, which we address in the next section.}.

In \cref{ch:closedloop}, we were able to derive closed-loop control laws primarily by exploiting strong structural assumptions, such as linear system dynamics, quadratic cost functions, or special system properties like differential flatness. 
While these assumptions yield elegant and computationally efficient solutions, they substantially limit the range of problems that can be addressed.
Dynamic programming offers a fundamentally different approach. 
Rather than relying on restrictive structural properties, it exploits only the sequential nature of the decision-making process and the additive structure of the cost function, thereby providing a systematic framework for computing optimal closed-loop policies for a much broader class of systems, including those with nonlinear dynamics and non-quadratic costs.

In what follows, we will explore how dynamic programming can be employed to solve the deterministic decision making problem defined above through the so-called \emph{principle of optimality}.

\subsubsection{The Principle of Optimality}
The dynamic programming approach to sequential decision making rests on a very simple, yet powerful, idea, known as the \emph{principle of optimality}\sidenote{Often referred to as \emph{Bellman’s principle of optimality}.}.
Despite its simplicity, this principle is the key that transforms an otherwise intractable optimization problem into one that can be solved efficiently through recursive decomposition.

At a high level, the principle of optimality expresses the following insight.
Suppose that a policy is optimal for a given decision making problem.
Then, if we consider any intermediate time and the state reached at that time while following this policy, the remaining decisions prescribed by the policy must themselves be optimal for the subproblem that starts from that state.
In other words, any \emph{tail segment} of an optimal policy must itself be optimal for the corresponding \emph{tail subproblem}.

The intuitive justification for this principle is straightforward.
If the remaining decisions after some time were not optimal for the corresponding tail subproblem, then we could replace them with a better alternative and thereby reduce the total cost.
This would contradict the assumption that the original policy was optimal.
Thus, optimal policies must be composed of optimal solutions to all of their tail subproblems.

\begin{figure}[ht]
    \centering
    \includegraphics[width=0.5\textwidth]{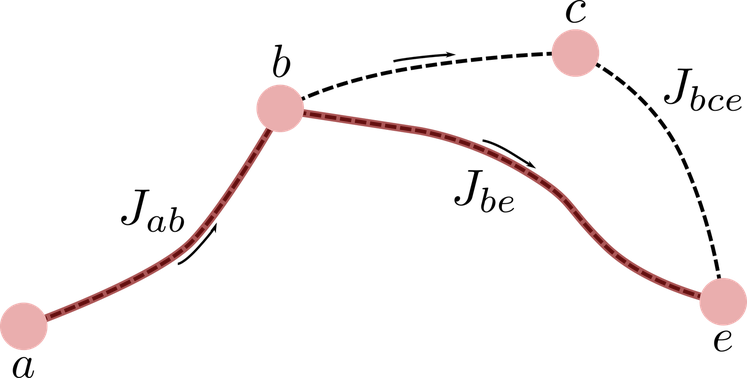}
    \caption{Illustration of the principle of optimality for a shortest-path problem. If the path $a \rightarrow b \rightarrow e$ is optimal from $a$ to $e$, then the subpath $b \rightarrow e$ must be optimal when starting from $b$. Adapted from \citet[Ch.~3]{Kirk2004}.}
    \label{fig:princopt1}
\end{figure}

This idea is most easily visualized in shortest-path problems, as illustrated in \cref{fig:princopt1}.
If the optimal path from a starting point $a$ to a destination $e$ passes through an intermediate point $b$, then the portion of the path from $b$ to $e$ must itself be optimal among all paths that start at $b$.
Otherwise, a shorter path from $b$ to $e$ could be substituted, yielding a shorter overall path from $a$ to $e$.

We now state the principle of optimality formally for deterministic decision making problems.
\begin{theorem}[Principle of Optimality (Deterministic Case)]
\label{thm:principle-optimality-det}
Let $\{\u_0^*, \u_1^*, \dots, \u_{T-1}^*\}$ be an optimal control sequence to the deterministic decision making problem defined by Problem~\ref{eq:dproblem} with initial condition $\x^*_0$, and let $\{\x_0^*, \x_1^*, \dots, \x_T^*\}$ denote the corresponding optimal state trajectory. 
Then, for any time $t \in \{0, \dots, T-1\}$, the truncated control sequence $\{\u_t^*, \dots, \u_{T-1}^*\}$ is optimal for the subproblem that starts from state $\x_t^*$ at time $t$ and minimizes the tail cost:
\begin{equation*}
J_{\text{tail}}(\x_t, \u_t, \dots, \u_{T-1}) = g_T(\x_T) + \sum_{i=t}^{T-1} g_i(\x_i, \u_i),
\end{equation*}
subject to the same system dynamics and control constraints, over the horizon $t$ to $T$.
\end{theorem}
The principle of optimality suggests that an optimal policy can be constructed by solving a sequence of smaller subproblems.
One may first solve the tail subproblem involving only the final stage, then extend this solution to the tail subproblem involving the last two stages, and continue in this manner until an optimal policy for the entire horizon is obtained.
The dynamic programming algorithm is based precisely on this idea: it proceeds backward in time, solving tail subproblems of increasing length by reusing solutions to shorter tail subproblems.

\begin{example}[Shortest-Path Problem]
\label{ex:principle-optimality}
Consider the deterministic shortest-path problem shown in \cref{fig:princopt2}, where the objective is to find an optimal path from point $b$ to point $f$.
Suppose that the optimal costs from points $c$, $d$, and $e$ to $f$ are already known.

\begin{figure}[ht]
    \centering
    \includegraphics[width=0.65\textwidth]{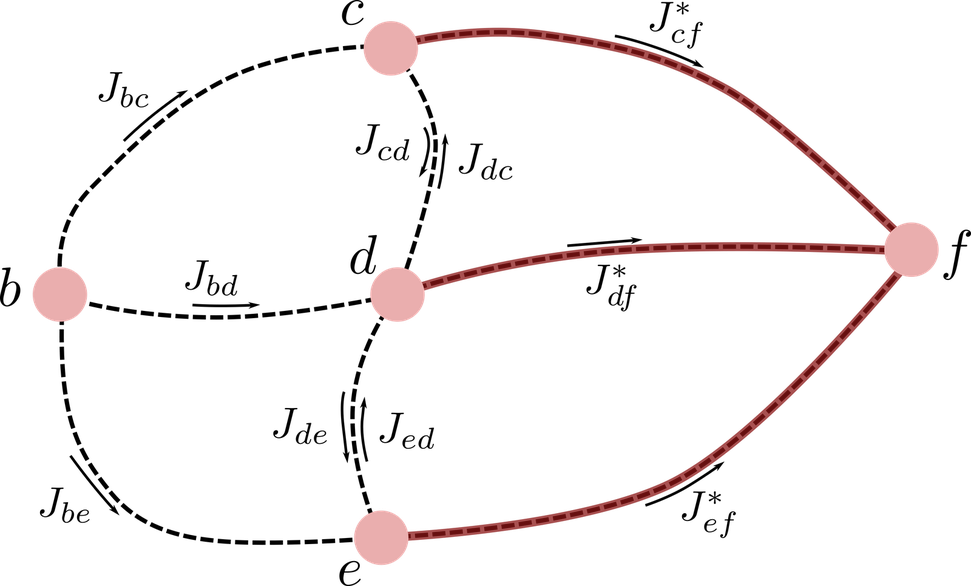}
    \caption{By leveraging optimal tail costs, the number of candidate paths that must be evaluated when searching from $b$ to $f$ is dramatically reduced. Adapted from \citet[Ch.~3]{Kirk2004}.}
    \label{fig:princopt2}
\end{figure}
\end{example}
A brute-force approach would require evaluating all possible paths from $b$ to $f$, including:
\begin{equation*}
\begin{split}
\{b\rightarrow c\rightarrow f,\;
b\rightarrow c\rightarrow d\rightarrow f,\;
b\rightarrow c\rightarrow d\rightarrow e\rightarrow f,\;
b\rightarrow d\rightarrow c\rightarrow f,\;
b\rightarrow d\rightarrow f, \\
b\rightarrow d\rightarrow e\rightarrow f,\;
b\rightarrow e\rightarrow d\rightarrow c\rightarrow f,\;
b\rightarrow e\rightarrow d\rightarrow f,\;
b\rightarrow e\rightarrow f\}.
\end{split}
\end{equation*}
By exploiting the principle of optimality, we know that any optimal path from $b$ to $f$ must consist of an immediate step from $b$ to one of its neighboring points ($c$, $d$, or $e$), followed by an optimal path from that neighboring point to $f$.
Thus, only three candidate paths need to be evaluated:
\begin{equation*}
	b\rightarrow c\rightarrow f,\quad
	b\rightarrow d\rightarrow f,\quad
	b\rightarrow e\rightarrow f,
\end{equation*}
where the optimal path can be found by selecting the minimum-cost path among:
$$
\{J_{bc} + J^*_{cf},\; J_{bd} + J^*_{df},\; J_{be} + J^*_{ef}\}.
$$
This simple example illustrates the central idea underlying dynamic programming, where optimal tail costs are reused to efficiently construct optimal solutions to larger problems. 
We now formalize this idea into a systematic algorithm for solving sequential decision making problems.

\subsubsection{The Dynamic Programming Algorithm}
The principle of optimality provides a powerful structural property of optimal solutions, but by itself it does not specify how such solutions should be computed. 
Dynamic programming turns this structural insight into a concrete computational procedure to find optimal control policies.

Rather than attempting to find an optimal control sequence over the entire horizon at once, dynamic programming proceeds by solving a sequence of smaller subproblems backward in time.
This motivates the introduction of the so-called \emph{cost-to-go} function. 
For each time $t \in \{0,\dots,T\}$ and each state $\x_t$, define the optimal cost-to-go $J_t^*(\x_t)$ as the minimum achievable cost when the system starts from state $\x_t$ at time $t$ and evolves optimally until the terminal time $T$. 
By definition, the cost-to-go at time $T$ coincides with the terminal cost:
$$
J_T^*(\x_T) = g_T(\x_T), \quad \forall \x_T \in \statespace,
$$
since no further decisions remain to be made after time $T$.

The principle of optimality implies that the optimal cost-to-go functions satisfy a recursive relationship, where the cost-to-go at time $t$ can be expressed in terms of the stage cost at time $t$ and the cost-to-go at time $t+1$:
\begin{equation}
	J_t^*(\x_t) = \min_{\u_t \in \controlspace(\x_t)} \Bigl[g_t(\x_t,\u_t) + J_{t+1}^*\bigl(\dynmodel_t(\x_t,\u_t)\bigr)\Bigr], \qquad t = 0,\dots,T-1.
	\label{eq:bellman}
\end{equation}
This equation, often referred to as the Bellman equation, expresses the \emph{global} optimization problem in terms of a \emph{local} minimization combined with the optimal solution of a shorter-horizon problem.

In practice, the dynamic programming algorithm leverages the Bellman equation to compute the optimal cost-to-go functions via a backward-in-time recursion, as summarized in \cref{alg:dDP}.
Starting from the known terminal cost $J_T^*$, one computes $J_{T-1}^*$, then $J_{T-2}^*$, and so on, until $J_0^*$ is obtained. 
At each stage of this backward recursion, the optimal tail cost is computed for every state in the state space. 
The result is a collection of functions $\{J_t^*(\cdot)\}_{t=0}^T$ that completely characterize the optimal performance of the system from any state and time.

\begin{algorithm}[ht]
 $J^*_T(\x_T) = g_T(\x_T),$ for all $\x_T \in \statespace$\\
 \For{$t=T-1$ \KwTo $0$}{
  $J^*_t(\x_t) = \min_{\u_t \in \controlspace(\x_t)} \Bigl[g_t(\x_t,\u_t) + J_{t+1}^*\bigl(\dynmodel_t(\x_t,\u_t)\bigr)\Bigr], \qquad$ for all $\x_t \in \statespace$\\
 }
 \Return $J^*_0(\cdot),\dots,J^*_T(\cdot)$
 \caption{Dynamic Programming (Deterministic)}
 \label{alg:dDP}
\end{algorithm}
Once the cost-to-go functions have been computed, the optimal control at each time step is obtained by minimizing the sum of the immediate cost and the optimal cost-to-go of the resulting next state:
\begin{equation*}
\u_t^* = \argmin_{\u_t \in \controlspace(\x_t^*)} \Bigl[g_t(\x_t^*,\u_t) + J_{t+1}^*\bigl(\dynmodel_t(\x_t^*,\u_t)\bigr)\Bigr].
\end{equation*}
The system is then propagated to the next state $\x_{t+1}^* = \dynmodel_t(\x_t^*,\u_t^*)$, and the process is repeated until the terminal time $T$ is reached.
Conceptually, the backward pass computes the optimal cost-to-go functions of every tail subproblem, while the forward pass uses these functions to compute the actions that realize these optimal costs.

Although dynamic programming yields an exact solution to the deterministic decision making problem, its direct application is often limited by computational considerations. 
The backward recursion requires evaluating the Bellman equation for every possible state at every time step, which may be infeasible when the state space is continuous or very large. 
Discretization of the state space may render the algorithm implementable, but even then the computational burden can grow rapidly with the dimension of the state. 
These challenges motivate the development of approximate methods that retain the conceptual framework of dynamic programming while relaxing its computational demands.

Despite these practical limitations, dynamic programming occupies a central role in sequential decision making. 
It provides the canonical solution method for finite-horizon problems, offers a precise interpretation of optimality through cost-to-go functions, and serves as the conceptual foundation for a wide range of learning-based control algorithms.

\begin{example}[Grid Navigation] 
\label{ex:detDP}
\theoremstyle{definition}
Consider the environment shown in \cref{fig:detDPprob}, where the objective is to move from point $a$ to point $h$ while incurring the minimum possible cost. 
The state $\x = \{a,b,c,d,e,f,g,h\}$ corresponds to the agent's current location on the grid, and the available control actions at each state are encoded by the arrows indicating allowable directions of travel. 
For example, at point $c$ the agent may move either right or up, but not left or down. 
Each directed edge is associated with a nonnegative traversal cost, as shown in the figure.
\begin{figure}[ht]
    \centering
    \includegraphics[width=0.8\textwidth]{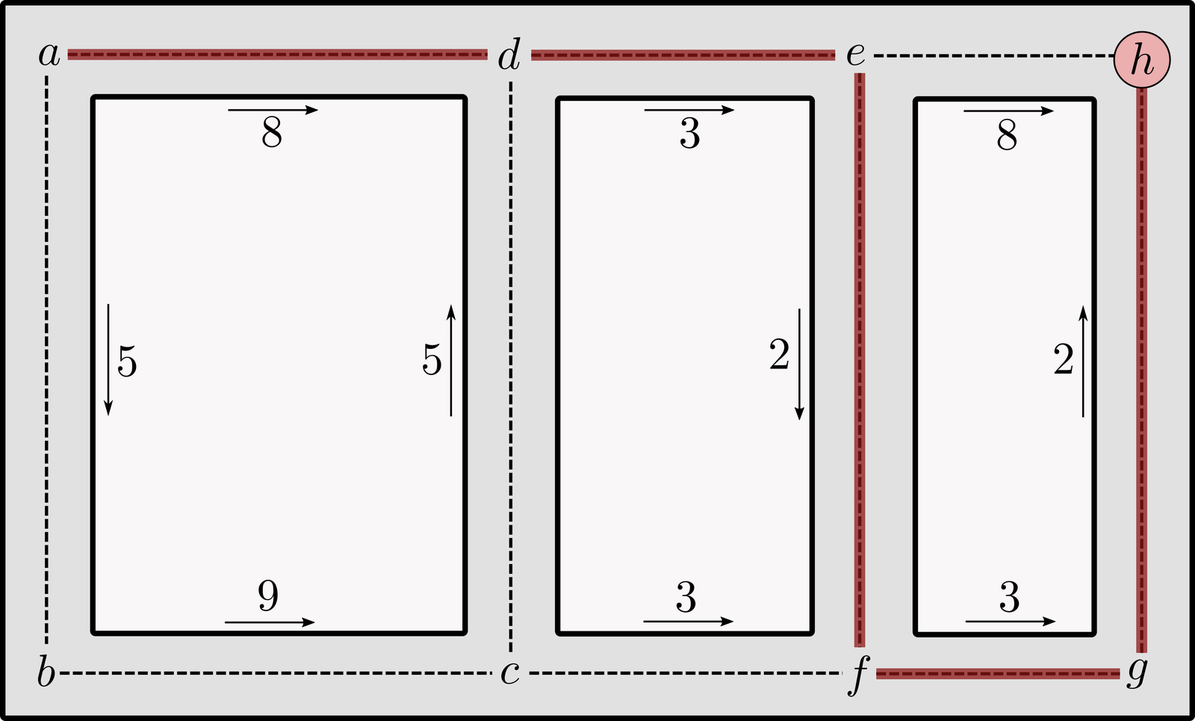}
    \caption{A deterministic decision making problem where the goal is to move from point $a$ to point $h$ while incurring the minimal amount of cost. 
    The path $a\rightarrow d \rightarrow e \rightarrow f \rightarrow g \rightarrow h$ is the optimal path. 
    We solve this problem by dynamic programming in \cref{ex:detDP}. Adapted from \citet[Ch.~3]{Kirk2004}.}
    \label{fig:detDPprob}
\end{figure}

We treat point $h$ as a terminal state with zero terminal cost, so that:
\begin{equation*}
    J^*_T(h) = 0,
\end{equation*}
and we allow the agent to remain at $h$ at zero cost once it is reached.
The dynamic programming recursion is initialized at this terminal condition and proceeds backward in time, successively computing the optimal cost-to-go for states that can reach $h$ within an increasing number of steps.

At the first backward step, corresponding to one step from the terminal time, only states that can transition directly to $h$ are relevant. 
These are the points $e$ and $g$, together with $h$ itself. 
The optimal costs-to-go are obtained by adding the immediate transition cost to the terminal cost:
\begin{equation*}
\begin{split}
J^*_{T-1}(h) &= 0 + J^*_T(h) = 0, \quad u^*_{T-1}(h) = \text{stay}, \\
J^*_{T-1}(e) &= 8 + J^*_T(h) = 8, \quad u^*_{T-1}(e) = \text{right}, \\
J^*_{T-1}(g) &= 2 + J^*_T(h) = 2, \quad u^*_{T-1}(g) = \text{up}. \\
\end{split}
\end{equation*}
At the next step of the recursion, states such as $d$ and $f$ become relevant, since they can reach $h$ in two steps.
Their optimal costs-to-go are computed by adding the immediate cost of moving to $e$ or $g$ and then using the previously computed one-step tail costs:
\begin{equation*}
\begin{split}
J^*_{T-2}(d) = 3 + J^*_{T-1}(e) = 11, \quad &u^*_{T-2}(d) = \text{right}, \\
J^*_{T-2}(f) = 3 + J^*_{T-1}(g) = 5, \quad &u^*_{T-2}(f) = \text{right}. \\
\end{split}
\end{equation*}
At this point, the cost-to-go values represent the optimal cost of reaching $h$ in at most two steps from each of the states $\{d,e,f,g\}$.

As the recursion continues, additional states enter the computation, and some states acquire multiple feasible paths to the terminal state. 
For example, at the next backward step, state $e$ may either move directly to $h$ or move downward to $f$ and then proceed optimally from there. 
The Bellman equation automatically selects the cheaper of these alternatives:
$$
J^*_{T-3}(e) = \min\{8 + J^*_{T-2}(h),\, 2 + J^*_{T-2}(f)\} = 7, \quad u^*_{T-3}(e) = \text{down}.
$$
Similarly, the other relevant states at this step are updated as follows:
\begin{equation*}
\begin{split}
J^*_{T-3}(g) = 2, \quad &u^*_{T-3}(g) = \text{up},  \\
J^*_{T-3}(d) = 3 + J^*_{T-2}(e) = 11, \quad &u^*_{T-3}(d) = \text{right}, \\
J^*_{T-3}(f) = 5, \quad &u^*_{T-3}(f) = \text{right}, \\
J^*_{T-3}(a) = 8 + J^*_{T-2}(d) = 19, \quad &u^*_{T-3}(a) = \text{right}, \\
J^*_{T-3}(c) = \min \{5 + J^*_{T-2}(d), \:3 + J^*_{T-2}(f)\} = 8, \quad &u^*_{T-3}(c) = \text{right}.
\end{split}
\end{equation*}
At this stage, we see that the goal $h$ is reachable from $a$ in three time steps on path $a\rightarrow d \rightarrow e \rightarrow h$, and that we would incur a cost of 19. 

Extending the horizon further allows the algorithm to discover lower-cost paths that take advantage of additional intermediate states. 
In particular, the recursion continues one more step to yield:
\begin{equation*}
	\begin{split}
	J^*_{T-4}(e) = 7, \quad &u^*_{T-4}(e) = \text{down}, \\
	J^*_{T-4}(g) = 2, \quad &u^*_{T-4}(g) = \text{up}, \\
	J^*_{T-4}(d) = 3 + J^*_{T-3}(e) = 10, \quad &u^*_{T-4}(d) = \text{right}, \\
	J^*_{T-4}(f) = 5,  \quad &u^*_{T-4}(f) = \text{right},\\
	J^*_{T-4}(a) = 8 + J^*_{T-3}(d) = 19, \quad &u^*_{T-4}(a) = \text{right}, \\
	J^*_{T-4}(c) = \min \{5 + J^*_{T-3}(d), \:3 + J^*_{T-3}(f)\} = 8, \quad &u^*_{T-4}(c) = \text{right}, \\
	J^*_{T-4}(b) = 9 + J^*_{T-3}(c) = 17, \quad &u^*_{T-4}(b) = \text{right}.\\
	\end{split}
	\end{equation*}
	Finally, extending the horizon one last time allows the algorithm to find the optimal path from the initial state $a$ to the goal state $h$, corresponding to the path $a\rightarrow d \rightarrow e \rightarrow f \rightarrow g \rightarrow h$, with total cost 18:
	\begin{equation*}
	\begin{split}
	J^*_{T-5}(e) = 7, \quad &u^*_{T-5}(e) = \text{down}, \\
	J^*_{T-5}(g) = 2, \quad &u^*_{T-5}(g) = \text{up}, \\
	J^*_{T-5}(d) = 10, \quad &u^*_{T-5}(d) = \text{right}, \\
	J^*_{T-5}(f) = 5, \quad &u^*_{T-5}(f) = \text{right}, \\
	J^*_{T-5}(a) = \min \{8 + J^*_{T-4}(d), \:5 + J^*_{T-4}(b)\} = 18, \quad &u^*_{T-5}(a) = \text{right}, \\
	J^*_{T-5}(c) = \min \{5 + J^*_{T-4}(d), \:3 + J^*_{T-4}(f)\} = 8, \quad &u^*_{T-5}(c) = \text{right}, \\
	J^*_{T-5}(b) = 9 + J^*_{T-4}(c) = 17, \quad &u^*_{T-5}(b) = \text{right}. \\
	\end{split}
\end{equation*}
Several important features of dynamic programming are illustrated by this example. 
First, the algorithm does not search over complete paths from $a$ to $h$. 
Instead, it incrementally builds optimal solutions by reusing previously computed tail costs. 
Second, the algorithm computes optimal costs and controls for \emph{all} states, not just the initial state of interest. 
As a result, once the cost-to-go functions have been computed, optimal paths can be generated immediately from any starting point and for any horizon length. 
For example, starting from point $c$ with a horizon of three steps, the optimal path $c \rightarrow f \rightarrow g \rightarrow h$ and its associated cost of $8$ can be read off directly, without any additional computation.
\end{example} 

\subsection{Decision Making Under Uncertainty: Markov Decision Processes}
\label{subsec:stoch-decision-making}
The deterministic decision making framework developed in \cref{subsec:det-decision-making} provides a clean and powerful lens through which sequential decision making problems can be understood and solved. 
In realistic robotic systems, however, the environment is never perfectly known, and uncertainty is an intrinsic feature of physical interaction and perception.
Sensor measurements are noisy, actuation is imperfect, and the environment may evolve in ways that cannot be modeled exactly or anticipated in advance.
As a result, the evolution of the system state cannot be described deterministically, and decision quality must be evaluated in a statistical sense.

The goal of this section is to extend the deterministic sequential decision making problem in Definition \ref{def:det-decision-making-problem} to explicitly account for such uncertainty.
We begin by deriving a stochastic formulation of the sequential decision making problem that incorporates uncertainty into both the system dynamics and the cost structure, thereby introducing the framework of \emph{Markov decision processes} (MDPs).
This framework is a cornerstone of modern decision making under uncertainty, and will serve as a bridge to the learning-based methods developed in \cref{ch:reinforcement-learning} and \cref{ch:imitation-learning}.
We then adapt the principle of optimality and the dynamic programming algorithm to this stochastic setting.

\subsubsection{Problem Formulation}
\label{subsubsec:stoch-problem-formulation}
We begin by modifying the state transition model in \eqref{eq:detmodel} to include a stochastic disturbance:
\begin{equation} 
\label{eq:smodel}
\x_{t+1}= \dynmodel_t(\x_t, \u_t, \w_t), \quad t = 0, \dots, T-1,
\end{equation}
where $\w_t$ denotes a stochastic disturbance at time $t$.
The disturbance $\w_t$ is assumed to be drawn from a known conditional probability distribution:
\begin{equation}
\w_t \sim p_t(\w_t \given \x_t, \u_t).
\end{equation}
This assumption implies that the distribution of the next state depends only on the current state and action, and not on the full history of the system.
This conditional independence assumption is an instance of the \emph{Markov property}\sidenote{Which we previously encountered in \cref{ch:intro-to-localization} in the context of Bayesian filtering.}, which states that, given the present state and action, the future evolution of the system is independent of the past.
Accordingly, the state $\x_t$ can be interpreted as a sufficient summary of all past information relevant for future decision making.

The admissible control constraints remain unchanged. 
At each time step, the control must satisfy:
\begin{equation}
\label{eq:stoch-control-constraints}
\u_t \in \controlspace(\x_t), \quad t = 0, \dots, T-1,
\end{equation}
where $\controlspace(\x_t)$ may encode physical limitations, logical constraints, or discrete action choices.

We also allow the instantaneous cost to depend explicitly on the disturbance:
\begin{equation}
\label{eq:stoch-stage-cost}
g_t: \statespace \times \controlspace \times \mathcal{W} \to \reals, \quad t = 0, \dots, T-1,
\end{equation}
where $\mathcal{W}$ denotes the space of possible disturbance values.

Because the system evolution is now stochastic, performance can no longer be evaluated along a single trajectory.
Instead, we measure performance in expectation.
Specifically, given a policy $\pi = \{\pi_0, \dots, \pi_{T-1}\}$, we define the associated expected cost starting from an initial state $\x_0$ as:
\begin{equation}
\label{eq:scost}
J_\pi(\x_0)
=
\expected{\w}{\,
g_T(\x_T) + \sum_{t=0}^{T-1} g_t(\x_t, \pi_t(\x_t), \w_t)
\,},
\end{equation}
where the expectation is taken with respect to the joint distribution of the disturbance sequence $\{\w_0,\dots,\w_{T-1}\}$ induced by the policy $\pi$ and the stochastic dynamics \eqref{eq:smodel}.
This formulation corresponds to a \emph{risk-neutral} objective, where policies are compared based on their average performance\sidenote{While the expected cost formulation is the most common in the literature, alternative risk measures---such as worst-case performance or risk-sensitive criteria---can also be considered, but are beyond the scope of this chapter.}.

We can now formally state the stochastic sequential decision making problem.
\begin{definition}[Stochastic Sequential Decision Making Problem]
\label{def:stoch-decision-making-problem}
Given the stochastic dynamics \eqref{eq:smodel}, the control constraints \eqref{eq:stoch-control-constraints}, and the expected cost \eqref{eq:scost}, the stochastic sequential decision making problem consists of computing an optimal policy:
$$
\pi^* = \{\pi_0^*, \dots, \pi_{T-1}^*\},
$$
that solves:
\begin{equation} 
\label{eq:sproblem}
\begin{split}
	J^*(\x_0) = \minimize[\pi] &\expected{\w}{\,
g_T(\x_T) + \sum_{t=0}^{T-1} g_t(\x_t, \pi_t(\x_t), \w_t)
\,} ,\\
	\subjectto & \x_{t+1}= \dynmodel_t(\x_t, \u_t, \w_t), \quad t = 0, \dots, T-1, \\
& \u_t = \pi_t(\x_t) \in \controlspace(\x_t), \:\:t=0,\dots,T-1.
\end{split}
\end{equation}
\end{definition}
This formulation is also known as a finite-horizon \emph{Markov decision process}.
It mirrors the deterministic problem in structure, differing only in the introduction of random disturbances and the use of expected cost as the performance criterion.
As we will see next, this similarity allows us to extend the principle of optimality and dynamic programming methods to this stochastic setting\sidenote{This problem formulation lies at the core of several disciplines, including optimal control, operations research, robotics, economics, and machine learning. As a result, it is common to encounter substantially different notation and terminology across communities, even when describing essentially the same underlying problem. In this chapter, we adopt notation that is standard in optimal control. In the next chapter, we will reintroduce the same problem using notation that is more common in the machine learning and reinforcement learning literature. For a broader discussion of the connections, overlaps, and distinctions among these perspectives, we refer the reader to~\citet{Powell2012}.}.

\subsubsection{Stochastic Decision Making with Dynamic Programming}
\label{subsubsec:stoch-principle-dp}
As in the deterministic case, the key property that enables efficient solution methods for the stochastic sequential decision making problem is the principle of optimality.
In this subsection, we first state the principle of optimality for the stochastic setting, and then show how it leads to a dynamic programming recursion for computing optimal policies.

\paragraph{Principle of Optimality.}
In the stochastic setting, the core intuition behind the principle of optimality—that optimal policies can be constructed by composing optimal solutions to tail subproblems—remains valid. 
However, it must be formulated in terms of \emph{expected} future cost.
Because state transitions are random, it is no longer meaningful to reason in terms of a single ``optimal trajectory''. 
Instead, policies are evaluated by the expected cumulative cost they induce under the stochastic elements in the system.

The crucial assumption that enables a recursive decomposition is the Markov property.
Specifically, as introduced in \cref{ch:intro-to-localization}, the Markov property ensures that the future evolution of the system depends only on the current state and action, and not on the full history leading up to that state.
Consequently, the expected cost incurred from time $t$ onward depends only on $\x_t$ and the future actions selected by the policy.

Under this assumption, if a policy is optimal from the initial condition, then after reaching any intermediate state $\x_t$, the remaining portion of that policy must still be optimal for the tail problem that starts at $\x_t$.
If this were not the case, the tail could be replaced by an alternative policy with strictly lower expected cost, thereby reducing the overall expected cost and contradicting the optimality of the original policy.

Formally, we can state the principle of optimality for the stochastic decision making problem as follows:
\begin{theorem}[Principle of Optimality (Stochastic Case)]
Let $\pi^* = \{\pi_0^*, \pi_1^*, \dots, \pi_{T-1}^*\}$ be an optimal policy for the stochastic decision making problem defined in Problem~\ref{eq:sproblem}. 
For any time $t$ and any state $\x_t$ that is reachable under $\pi^*$, the tail policy $\{\pi_t^*, \dots, \pi_{T-1}^*\}$ is an optimal policy for the tail subproblem that starts at time $t$ from state $\x_t$ and minimizes the expected cost:
$$
J_{\pi}( \x_t ) = \expected{\w}{\, g_T(\x_T) + \sum_{i=t}^{T-1} g_i(\x_i, \pi_i(\x_i), \w_i) \,}.
$$
\end{theorem}
As in the deterministic case, this result implies that optimal policies can be constructed by solving a sequence of nested tail subproblems, with dynamic programming providing a systematic procedure for carrying out this backward construction, as we describe next.

\paragraph{Dynamic Programming.}
The dynamic programming algorithm for the stochastic case closely mirrors its deterministic counterpart.
It proceeds backward in time, starting from the terminal cost:
$$
J_T^*(\x_T) = g_T(\x_T), \quad \forall \x_T \in \statespace,
$$
and successively computing the optimal cost-to-go for earlier stages.
At each step, the algorithm evaluates, for every state, the expected cost associated with each admissible control and selects the minimizing one.
Similarly, the cost-to-go can be expressed recursively, yielding the (stochastic) Bellman equation:
\begin{equation}
	\label{eq:stoch-bellman}
	J_t^*(\x_t)
	=
	\min_{\u_t \in \controlspace(\x_t)}
	\expected{\w}{
	g_t(\x_t,\u_t,\w_t)
	+
	J_{t+1}^*\!\bigl(\dynmodel_t(\x_t,\u_t,\w_t)\bigr)
	},
	\qquad t = 0,\dots,T-1.
\end{equation}
The resulting dynamic programming algorithm is summarized in \cref{alg:sDP}.

\begin{algorithm}[ht!]
	$J_T^*(\x) = g_T(\x),$ for all $\x \in \statespace$\\
	\For{$t=T-1$ \KwTo $0$}{
	 $J_t^*(\x) = \underset{\u \in \controlspace(\x)}{\min} \expected{\w}{g_t(\x,\u,\w) + J_{t+1}^*(\dynmodel_t(\x, \u, \w))},$ for all $\x \in \statespace $\\
	}
	\Return $J_0^*(\cdot),\dots,J_T^*(\cdot)$
	\caption{Dynamic Programming (Stochastic Case)}
	\label{alg:sDP}
\end{algorithm}
Once the cost-to-go functions have been computed, an optimal policy is obtained by selecting:
\begin{equation}
\label{eq:stoch-opt-policy}
\pi_t^*(\x_t)
=
\argmin_{\u_t \in \controlspace(\x_t)}
\expected{\w}{
g_t(\x_t,\u_t,\w_t)
+
J_{t+1}^*\!\bigl(\dynmodel_t(\x_t,\u_t,\w_t)\bigr)
}.
\end{equation}

\begin{example}[Inventory Control]
\label{ex:stoDP}
Consider a simple inventory control problem in which the state $x_t \in \mathbb{N}$ denotes the available stock of an item at time $t$.
At each step, the decision maker chooses how many items to order, $u_t \in \mathbb{N}$, before facing a random demand $w_t \in \mathbb{N}$.

The system dynamics are given by:
$$
x_{t+1} = \max\{0,\, x_t + u_t - w_t\},
$$
which captures the fact that demand reduces inventory, restocking increases it, and inventory cannot go below zero.
We impose the constraint:
$$
x_t + u_t \leq 2,
$$
so that the inventory capacity is limited to at most two units.

Demand is modeled as a discrete random variable with the following probability distribution:
$$
p(w_t=0)=0.1, \quad p(w_t=1)=0.7, \quad p(w_t=2)=0.2.
$$

We consider a finite horizon of $T=3$ steps and define the expected cost:
$$
\expected{w}{\sum_{t=0}^{2} \bigl(u_t + (x_t + u_t - w_t)^2\bigr)},
$$
which penalizes both ordering costs and the squared deviation between inventory and demand.

To solve this problem using dynamic programming, we first identify the state and control spaces:
$$
\statespace = \{0,1,2\}, \quad \controlspace(x_t) = \{0,1,2 - x_t\}.
$$
Following \cref{alg:sDP}, we initialize the terminal cost-to-go function at time $t=3$ as:
$$
J_3^*(x_3) = 0, \quad \forall x_3 \in \{0,1,2\},
$$
since no costs are incurred after the final stage.
Working backward, we compute the cost-to-go at time $t=2$ by minimizing the expected one-step cost:
\begin{equation*}
	\begin{split}
	J^*_2(0) &= \underset{u_2 \in \{0,1,2\}}{\text{minimize}} \:\: \expected{w}{u_2 + (u_2 - w_2)^2}, \\
	&= \underset{u_2 \in \{0,1,2\}}{\text{minimize}} \:\: u_2 + 0.1u_2^2 + 0.7(u_2-1)^2 + 0.2(u_2 - 2)^2 = 1.3, \\
	J^*_2(1) &= \underset{u_2 \in \{0,1\}}{\text{minimize}} \:\: \expected{w}{u_2 + (1 + u_2 - w_2)^2}, \\
	&= \underset{u_2 \in \{0,1\}}{\text{minimize}} \:\: u_2 + 0.1(1 + u_2)^2 + 0.7(u_2)^2 + 0.2(u_2 - 1)^2 = 0.3, \\
	J^*_2(2) &= \expected{w}{(2 - w_2)^2} = 0.1(2)^2 + 0.7(1)^2 + 0.2(0)^2 = 1.1.
	\end{split}
\end{equation*}
Note that for $x_2=2$, the only admissible action is $u_2 = 0$ due to the inventory capacity constraint, and hence the minimum is achieved at $u_2=0$.

From these computations, we directly obtain the optimal actions at time $t=2$:
\begin{equation*}
	\begin{split}
	\pi^*_2(0) &= 1, \\ 
	\pi^*_2(1) &= 0, \\ 
	\pi^*_2(2) &= 0.
	\end{split}
\end{equation*}
Continuing this process, we compute the cost-to-go at time $t=1$:
\begin{equation*}
\begin{split}
J^*_1(0) &= \underset{u_1 \in \{0,1,2\}}{\text{minimize}} \:\: \expected{w}{u_1 + (u_1 - w_1)^2 + J^*_2(\max\{0, u_1 - w_1\})} = 2.5, \\
J^*_1(1) &= \underset{u_1 \in \{0,1\}}{\text{minimize}} \:\: \expected{w}{u_1 + (1 + u_1 - w_1)^2 + J^*_2(\max\{0, 1 + u_1 - w_1\})} = 1.5,\\
J^*_1(2) &= \expected{w}{(2 - w_1)^2  + J^*_2(\max\{0, 2 - w_1\})} = 1.68,
\end{split}
\end{equation*}
with optimal stage actions:
\begin{equation*}
	\begin{split}
	\pi^*_1(0) &= 1, \\ 
	\pi^*_1(1) &= 0, \\ 
	\pi^*_1(2) &= 0. \\ 
	\end{split}
\end{equation*}
Finally, in the last step:
\begin{equation*}
\begin{split}
J^*_0(0) &= \underset{u_0 \in \{0,1,2\}}{\text{minimize}} \:\: \expected{w}{ u_0 + (u_0 - w_0)^2 + J^*_1(\max\{0, u_0 - w_0\}) } = 3.7, \\
J^*_0(1) &= \underset{u_0 \in \{0,1\}}{\text{minimize}} \:\: \expected{w}{ u_0 + (1 + u_0 - w_0)^2 + J^*_1(\max\{0, 1 + u_0 - w_0\})} = 2.7,\\
J^*_0(2) &= \expected{w}{(2 - w_0)^2  + J^*_1(\max\{0, 2 - w_0\})} = 2.818,
\end{split}
\end{equation*}
with:
\begin{equation*}
\begin{split}
\pi^*_0(0) &= 1, \\ 
\pi^*_0(1) &= 0, \\ 
\pi^*_0(2) &= 0.
\end{split}
\end{equation*}
For this example, the optimal policy is time-invariant: \emph{order one unit when the inventory is empty, and order nothing otherwise}.
This policy balances the risk of unmet demand against the cost of carrying inventory, and it emerges naturally from the dynamic programming recursion.
\end{example}

\subsection{Limitations of Dynamic Programming}
\label{subsec:limitations_dp}
Dynamic programming is a powerful algorithmic framework that underlies a wide range of methods for solving sequential decision-making problems. 
However, despite its generality and conceptual elegance, it suffers from several important practical limitations.

First, in its standard form, dynamic programming requires \emph{perfect knowledge} of the environment.
This includes an accurate model of the system dynamics---whether deterministic or stochastic---as well as a known cost or reward function.
In many real-world applications, particularly in robotics, this requirement can be highly restrictive, as physical interactions involving friction, contact dynamics, or complex nonlinear effects are often difficult to model accurately.
Moreover, the algorithms presented in this chapter assume that the full system state is known and directly observable, which is often not the case in practice.

A second major limitation of dynamic programming is the so-called \emph{curse of dimensionality}. 
The computational and storage requirements of dynamic programming grow exponentially with the dimension of the state space. 
Concretely, if the state is $n$-dimensional and each state variable can take on $M$ discrete values, then the Bellman equation must be evaluated $M^n$ times at each stage of the algorithm. 
While this may be tractable for low-dimensional problems, it quickly becomes infeasible as the dimensionality increases.
This issue is especially relevant in robotics, where the state and action spaces can be very high-dimensional due to the presence of multiple degrees of freedom, sensors, and actuators.

Together, these limitations motivate the development of alternative approaches that relax some of the assumptions underlying classical dynamic programming. 
In particular, they have led to the study of learning-based methods that trade exact optimality for computational tractability.

In the next chapter, we first introduce \emph{reinforcement learning}, which can be viewed as a natural extension of the ideas developed here.
At a high level, reinforcement learning may be viewed as a form of \emph{approximate dynamic programming}, as it retains the central concepts of value functions, Bellman recursions, and policy improvement, while addressing the key limitations of classical dynamic programming by allowing policies to be learned from interaction rather than computed from a perfectly known model over an explicitly enumerated state space. 
For this reason, reinforcement learning provides the most direct next step after the present chapter.

We then turn to \emph{imitation learning}, which offers a broader and highly practical framework for robot learning from demonstrations. 
Rather than learning solely through trial-and-error interaction with the environment, imitation learning leverages expert behavior to acquire control policies, and has become one of the most widely used paradigms for training robotic systems in practice. 
Taken together, reinforcement learning and imitation learning provide two complementary responses to the limitations of classical dynamic programming, and they will be the focus of \cref{ch:reinforcement-learning} and \cref{ch:imitation-learning}, respectively.

\section{Summary}
In this chapter, we introduced a general, optimization-based framework for sequential decision-making that allows us to compute optimal policies for complex tasks, even under uncertainty. 
We began in \cref{subsec:det-decision-making} by framing deterministic sequential decision-making as a discrete-time optimal control problem, introducing dynamic programming as the primary computational tool for solving such problems.
Grounded in the principle of optimality, dynamic programming allows long-horizon problems to be decomposed into a sequence of simpler, recursively defined subproblems.
In \cref{subsec:stoch-decision-making}, we extended this framework to the stochastic setting, where uncertainty in system dynamics and outcomes must be explicitly accounted for. This led to the formulation of Markov decision processes and the use of dynamic programming to compute policies that optimize expected performance under known sources of randomness.
Finally, in \cref{subsec:limitations_dp}, we discussed the practical limitations of dynamic programming, most notably the curse of dimensionality and the reliance on accurate models of the environment. 
These challenges motivate the development of learning-based approaches, which are the focus of the next chapter.

\paragraph{To learn more.}
For a foundational and comprehensive treatment of dynamic programming and sequential decision-making, \emph{Dynamic Programming and Optimal Control} by \citet{Bertsekas2000} is an indispensable resource, providing rigorous derivations for both deterministic and stochastic problems. 
For an introduction to Markov decision processes with a focus on their role in modern artificial intelligence and robotics, \emph{Reinforcement Learning: An Introduction} by \citet{SuttonBarto2018} is the standard reference. 
A more in-depth mathematical treatment of MDPs can be found in \emph{Markov Decision Processes: Discrete Stochastic Dynamic Programming} by \citet{Puterman2014}.

\section{Exercises}
The starter code for the exercises provided below is available online through GitHub. 
To get started, download the code by running in a terminal window:

\begin{tcolorbox}[colback=gray!10]
\begin{minted}{bash}
    git clone https://github.com/StanfordASL/pora-exercises.git
\end{minted}
\end{tcolorbox}

We denote Problems requiring hand-written solutions and coding in Python with \adjustbox{height=2ex, valign=c}{\includegraphics{figs/write.png}} and \adjustbox{height=2ex, valign=c}{\includegraphics{figs/code.png}}, respectively.

\subsection*{\adjustbox{height=2ex, valign=c}{\includegraphics{figs/write.png}}\ Problem 1: Shortest Path Through A Grid}
\begin{figure}[ht!]
		\begin{center}
			\begin{tikzpicture}[node distance=0cm, ->]
				\tikzstyle{node} = [draw=black, circle, rounded corners, minimum height=2.0em, minimum width=2.0em, fill=red!30]
				\node[node](s1){\small A};
				\node[node, right of=s1, xshift=1cm, yshift=1cm](s2){\tiny};
				\node[node, right of=s1, xshift=1cm, yshift=-1cm](s3){\tiny};
				\node[node, right of=s2, xshift=1cm, yshift=1cm](s4){\tiny};
				\node[node, right of=s2, xshift=1cm, yshift=-1cm](s5){\tiny};
				\node[node, right of=s3, xshift=1cm, yshift=-1cm](s6){\tiny};
				\node[node, right of=s4, xshift=1cm, yshift=1cm](s7){\tiny};
				\node[node, right of=s4, xshift=1cm, yshift=-1cm](s8){\tiny};
				\node[node, right of=s6, xshift=1cm, yshift=1cm](s9){\tiny};
				\node[node, right of=s6, xshift=1cm, yshift=-1cm](s10){\tiny};
				\node[node, right of=s8, xshift=1cm, yshift=1cm](s11){\tiny};
				\node[node, right of=s8, xshift=1cm, yshift=-1cm](s12){\tiny};
				\node[node, right of=s9, xshift=1cm, yshift=-1cm](s13){\tiny};
				\node[node, right of=s12, xshift=1cm, yshift=1cm](s14){\tiny};
				\node[node, right of=s12, xshift=1cm, yshift=-1cm](s15){\tiny};
				\node[node, right of=s14, xshift=1cm, yshift=-1cm](s16){\small B};
				\draw[->] (s1) to[] node[midway,above left,inner sep=2pt] {\small $5$} (s2);
				\draw[->] (s1) to[] node[midway,below left,inner sep=2pt] {\small $7$} (s3);
				\draw[->] (s2) to[] node[midway,above left,inner sep=2pt] {\small $6$} (s4);
				\draw[->] (s2) to[] node[midway,below left,inner sep=2pt] {\small $8$} (s5);
				\draw[->] (s3) to[] node[midway,above left,inner sep=2pt] {\small $7$} (s5);
				\draw[->] (s3) to[] node[midway,below left,inner sep=2pt] {\small $9$} (s6);
				\draw[->] (s4) to[] node[midway,above left,inner sep=2pt] {\small $10$} (s7);
				\draw[->] (s4) to[] node[midway,below left,inner sep=2pt] {\small $6$} (s8);
				\draw[->] (s5) to[] node[midway,above left,inner sep=2pt] {\small $10$} (s8);
				\draw[->] (s5) to[] node[midway,below left,inner sep=2pt] {\small $8$} (s9);
				\draw[->] (s6) to[] node[midway,above left,inner sep=2pt] {\small $10$} (s9);
				\draw[->] (s6) to[] node[midway,below left,inner sep=2pt] {\small $6$} (s10);
				\draw[->] (s7) to[] node[midway,below left,inner sep=2pt] {\small $7$} (s11);
				\draw[->] (s8) to[] node[midway,above left,inner sep=2pt] {\small $5$} (s11);
				\draw[->] (s8) to[] node[midway,below left,inner sep=2pt] {\small $9$} (s12);
				\draw[->] (s9) to[] node[midway,above left,inner sep=2pt] {\small $5$} (s12);
				\draw[->] (s9) to[] node[midway,below left,inner sep=2pt] {\small $12$} (s13);
				\draw[->] (s10) to[] node[midway,above left,inner sep=2pt] {\small $7$} (s13);
				\draw[->] (s11) to[] node[midway,below left,inner sep=2pt] {\small $8$} (s14);
				\draw[->] (s12) to[] node[midway,above left,inner sep=2pt] {\small $6$} (s14);
				\draw[->] (s12) to[] node[midway,below left,inner sep=2pt] {\small $7$} (s15);
				\draw[->] (s13) to[] node[midway,above left,inner sep=2pt] {\small $9$} (s15);
				\draw[->] (s14) to[] node[midway,below left,inner sep=2pt] {\small $10$} (s16);
				\draw[->] (s15) to[] node[midway,above left,inner sep=2pt] {\small $11$} (s16);
			\end{tikzpicture}
		\end{center}
\caption{Simple grid for shortest-path problem.}
\label{fig:shortest-path-grid}
\end{figure}
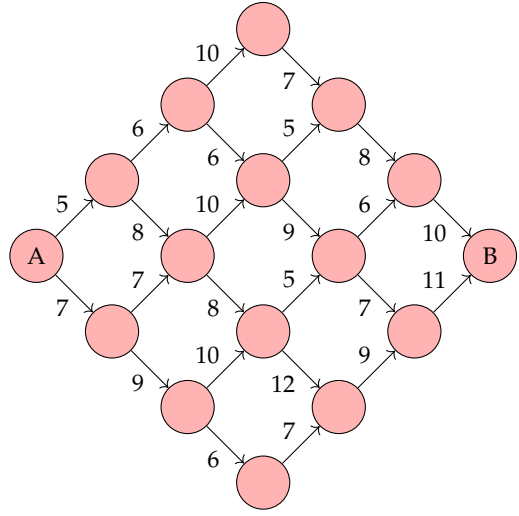

Consider the graph shown in \cref{fig:shortest-path-grid}, where it is only possible to move to the right and the numbers associated with each edge represent a cost to traverse that edge.
The decision to be made at each node is whether to go ``up'' or ``down'', and we can assume the state transitions are deterministic.

For this exercise:
\begin{enumerate}
\item Use dynamic programming to find the shortest path from A to B.
\item Consider a generalized version of the shortest path problem in \cref{fig:shortest-path-grid} where the grid has $n$ segments along each side. 
Find the number of computations required by an exhaustive search algorithm (i.e., the number of routes that such an algorithm would need to evaluate) and the number of computations required by a DP algorithm (i.e., the number of DP evaluations). 
For example, for the $n=3$ case shown in \cref{fig:shortest-path-grid}, an exhaustive search algorithm requires~20 computations, while the DP algorithm requires only~15.
\end{enumerate}

\subsection*{\adjustbox{height=2ex, valign=c}{\includegraphics{figs/write.png}}\ Problem 2: Machine Maintenance}
Suppose we have a machine that is either running or is broken down. 
If it runs throughout one week, it makes a gross profit of $\$100$. 
If it fails during the week, gross profit is zero. 
If it is running at the start of the week and we perform preventive maintenance, the probability that it will fail during the week is $0.4$. 
If we do not perform maintenance, the probability of failure is $0.7$. 
However, preventative maintenance will cost $\$20$. 
When the machine is broken down at the start of the week, it may either be repaired at a cost of $\$40$, in which case it will fail during the week with a probability of $0.4$, or it may be replaced at a cost of $\$150$ by a new machine that is guaranteed to run through its first week of operation. 
Find the optimal repair, replacement, and maintenance policy that maximizes total profit over four weeks, assuming a new machine at the start of the first week (that is guaranteed to run during the first week of operation).

\subsection*{\adjustbox{height=2ex, valign=c}{\includegraphics{figs/code.png}}\ Problem 3: Markovian Drone}
In this problem, we will model the task of flying a drone to its destination through a storm as a Markov Decision Process (MDP), and solve for the optimal policy using dynamic programming.
The world is represented as an $n \times n$ grid, so the state space is:
\begin{equation*}
    \mathcal{X}
    \defn \{ (x_1,x_2) \in \R^2 \mid x_1,x_2 \in \{0,1,\dots,n-1\} \}.
\end{equation*}
In these coordinates, $(0,0)$ represents the bottom left corner of the map and $(n-1,n-1)$ represents the top right corner of the map. 
From any location $x = (x_1,x_2) \in \mathcal{X}$, the drone has five possible controls it can apply:
\begin{equation*}
    \mathcal{U}
    \defn \{\texttt{up}, \texttt{down}, \texttt{left}, \texttt{right}, \texttt{land}\}.
\end{equation*}
The corresponding state changes for each control input are:
\begin{itemize}
    \item \texttt{up}: $(x_1, x_2) \mapsto (x_1,x_2+1)$
    \item \texttt{down}: $(x_1, x_2) \mapsto (x_1,x_2-1)$
    \item \texttt{left}: $(x_1, x_2) \mapsto (x_1-1,x_2)$
    \item \texttt{right}: $(x_1, x_2) \mapsto (x_1+1,x_2)$
    \item \texttt{land}: $(x_1, x_2) \mapsto (x_1,x_2)$
\end{itemize}
Additionally, there is a storm centered at $x_\mathrm{eye} \in \mathcal{X}$. 
The storm's influence is strongest at its center and decays farther from the center according to the equation $\omega(x) = \exp \left( - \frac{\norm{x - x_\mathrm{eye}}_2^2}{2 \sigma^2} \right)$. 
Given its current state $x$ and control input $u$, the drone's next state is determined as follows:
\begin{itemize}
	\item The control to land is deterministic.
    \item With probability $\omega(x)$, the storm will cause the drone to move in a uniformly random direction (for non-landing controls).
    \item With probability $1-\omega(x)$, the drone will move in the direction specified by the control (for non-landing controls).
    \item If the resulting movement would cause the drone to leave $\mathcal{X}$, then it will not move at all. For example, if the drone is on the right boundary of the map, then moving right will do nothing.
\end{itemize}
The quadrotor's objective is to reach $x_\mathrm{goal} \in \mathcal{X}$ as quickly as possible, so the cost function is the indicator function $g_t(x_t) = 1 - I_{x_\mathrm{goal}}(x_t)$. 
In other words, the drone will receive a cost of $1$ whenever it is not at $x_\mathrm{goal} \in \mathcal{X}$, and a cost of $0$ at the goal.
The drone has limited fuel capacity, so it must reach the goal in at most $T$ timesteps.
If the drone is not at the goal at the end of the horizon, it will crash and we will incur a cost of $100$ to replace the drone, therefore $g_T(x_T) = 100(1 - I_{x_\mathrm{goal}}(x_T))$.

In the notebook \colorcode{ch17/exercises/markovian\_drone.ipynb}, complete the following exercises:
\begin{enumerate}
    \item Implement the functions of \colorcode{DroneMDP} to compute the optimal cost-to-go values via dynamic programming (\cref{alg:sDP}) and extract the optimal policy using \cref{eq:stoch-opt-policy}. 
    \item Given $T=100$, $n = 20$, $\sigma = 10$, $x_\mathrm{eye} = (15,15)$, and $x_\mathrm{goal} = (19,9)$, compute and plot a heatmap of the optimal cost-to-go for $t=0$ over the grid $\mathcal{X}$.
    Then, using the code provided, simulate the MDP with the optimal policy for $T = 100$ time steps with the state initialized at $x = (0,19)$.
\end{enumerate}

\newpage
\printbibliography[segment=\therefsegment,heading=subbibliography,title={References}]
\chapter{Reinforcement Learning}
\label{ch:reinforcement-learning}
\newrefsegment
In \cref{ch:sequential-decision-making}, we introduced the deterministic and stochastic sequential decision-making problems, and showed how they can be solved using dynamic programming. 
These approaches, however, suffer from two major limitations: (i) they typically assume full knowledge of the system dynamics, and (ii) their computational complexity grows exponentially with the dimensionality of the state space.
In this chapter, we provide an introduction and overview of the field of \emph{Reinforcement Learning}\cite{Bertsekas2019}\cite{SuttonBarto2018} (RL).
At a high level, reinforcement learning can be described as the problem of \emph{learning what to do}---that is, learning a mapping from states to control actions---with the goal of maximizing a cumulative numerical reward.
More specifically, reinforcement learning formalizes learning through interaction.
Rather than being explicitly instructed on the correct actions to take, an agent must discover effective strategies through trial and error, guided only by feedback from the environment. 
Crucially, unlike the methods discussed in the previous chapter, reinforcement learning does not require prior knowledge of the system dynamics or even full observability of the environment. 
This makes it a highly general and practical framework for autonomous decision-making in complex and uncertain environments.

In this chapter, we begin in \cref{subsec:rl_problem} by introducing key concepts and theoretical foundations of the reinforcement learning problem. 
Next, in \cref{subsec:dyn_prog_methods}, we introduce reinforcement learning algorithms based on exact dynamic programming that leverage ideas from \cref{ch:sequential-decision-making}. 
Motivated by practical limitations of dynamic programming, we introduce two foundational model-free learning paradigms known as Monte Carlo methods and temporal-difference learning in \cref{subsec:model_free_control}. 
Finally, after introducing a taxonomy of reinforcement learning algorithms in \cref{subsec:taxonomy_rl}, we discuss widely used model-free and model-based algorithms in \cref{subsec:model_free_rl} and \cref{subsec:model_based_rl}, respectively.

\subsection{The Reinforcement Learning Problem}
\label{subsec:rl_problem}
At its core, reinforcement learning represents a mathematical formalism for learning-based decision making.
While the mathematical formulation---which entails solving an optimal control problem within an incompletely-known Markov decision process (MDP)---will be detailed in the remainder of this chapter, the basic idea is simple: capture the essential aspects of a learning agent interacting over time with its environment to achieve a goal or objective.

Within the landscape of machine learning paradigms, there are a number of key distinctions and specific challenges that are unique to reinforcement learning.

\paragraph{Learning Without a Teacher.}
Supervised learning, arguably the most extensively studied paradigm in machine learning, assumes access to a dataset of labeled examples provided by a knowledgeable supervisor, with the objective of learning to imitate the supervisor’s behavior. 
While highly effective in many settings, this paradigm is ill-suited for interactive decision-making problems, where labeled data is unavailable or prohibitively expensive to obtain. 
Reinforcement learning, by contrast, is fundamentally concerned with learning from interaction, where the agent is not told which actions to take, but must instead learn through trial and error, guided solely by evaluative feedback from the environment in the form of rewards or penalties. 
This absence of an explicit teacher introduces a distinct set of challenges.

\paragraph{Exploration vs.\ Exploitation.}
In reinforcement learning, the agent must resolve the intrinsic trade-off between exploration and exploitation. 
On the one hand, it must \textit{explore} the environment to discover potentially rewarding actions; on the other hand, it must \textit{exploit} its current knowledge to maximize cumulative reward. 
Focusing exclusively on exploration prevents the agent from capitalizing on what it has learned, while focusing solely on exploitation risks converging to a suboptimal policy. 
Effective learning therefore requires the agent to judiciously balance both behaviors---trying a diverse set of actions while progressively concentrating on those that yield the highest returns. 
This trade-off is largely absent in other learning paradigms and remains a defining challenge of reinforcement learning.

\paragraph{Delayed Rewards.}
In many real-world problems, feedback from the environment is delayed or sparse. 
Rewards may only be observed after a sequence of actions has been executed---for example, in chess, where the outcome is revealed only at the end of the game. 
Such delayed feedback introduces the challenge of \textit{credit assignment}, making it difficult for the agent to determine which actions were responsible for the observed outcome.

\paragraph{Data Not I.I.D.}
A core assumption underlying much of statistical learning theory is that data points are independent and identically distributed (i.i.d.). 
Reinforcement learning clearly violates this assumption, where the data collected by the agent is inherently sequential and highly correlated, since each action influences both the future states of the environment and the data observed thereafter. 
This dependence fundamentally changes both the theoretical analysis and practical design of learning algorithms.

\medskip
In the following sections, we introduce the mathematical formalism of reinforcement learning and the key components that define the problem, together with strategies for addressing the challenges outlined above.

\subsubsection{Elements of Reinforcement Learning}
\label{subsubsec:elements_of_rl}
In this and previous chapters, we have encountered and briefly discussed several key elements of a reinforcement learning system: a \emph{policy}, an \emph{environment}, a \emph{reward signal}, a \emph{value function}, and, optionally, a \emph{model} of the environment.
In this section, we provide a more formal definition of these elements and discuss how they interact in the context of the reinforcement learning problem.

\paragraph{Policy.} A \emph{policy}, $\pi(\u \given \x)$, is a mapping from states to actions that defines the agent's behavior.
Depending on the application, the policy can take the form of a simple function, such as a lookup table or a parametric function, or a more complex one, for instance involving an explicit search process.
In general, policies may be either deterministic, mapping each state to a single action, or stochastic, defining a probability distribution over actions.

\paragraph{Environment.} The \emph{environment} is the system the agent interacts with.
We mathematically represent the environment by a transition model, $p(\x' \given \x, \u)$, which defines the probability of transitioning to a new state, $\x'$, given the current state, $\x$, and action, $\u$.
The transition model may also be either deterministic or stochastic, depending on the nature of the environment.

\paragraph{Reward signal.} A \emph{reward signal} defines the goal of the agent. 
At each time step, we assume the agent receives a scalar \emph{reward}, $r \in \R$, from the environment that indicates how well the agent is performing.
Reward signals are deterministic or stochastic functions of the state of the environment and the action taken, and we denote the function that produces the reward as $R(\x, \u)$.

\paragraph{Value function.} While the reward signal represents an immediate measure of performance, the \emph{value function} represents performance in the long run.
Specifically, the value of a state defines how much reward the agent can expect to accumulate from that state onwards.
This is clearly different from immediate reward.
For example, a state might have a low immediate reward but a high value if it usually leads to states with high rewards, and vice versa.
An effective agent chooses actions by considering the value of the action rather than just the immediate reward.
Because of this, numerous reinforcement learning algorithms are typically centered around accurately estimating values.

\paragraph{Model.} Lastly, a \emph{model} of the environment is an optional component of the reinforcement learning problem that represents the agent's understanding of the environment.
The model's goal is to mimic the behavior of the environment, and can be used to make hypotheses about how the environment will evolve\sidenote{For example, we can use models to evaluate different actions before executing them.}.
In this chapter, we explore reinforcement learning algorithms that use models for learning, referred to as \emph{model-based} algorithms, as well as more direct \emph{model-free} algorithms that do not attempt to learn a model of the environment and solely focus on discovering optimal policies by trial-and-error learning.

\medskip
At a high level, most reinforcement learning algorithms follow the same basic learning cycle. 
First, the agent interacts with the environment by observing the state $\x_t$, applying an action $\u_t$ from a chosen \emph{behavior policy}\sidenote{The behavior policy does not necessarily have to match the learned policy $\pi(\u \given \x)$.}, and then observing the next state $\x_{t+1}$ and scalar reward $r_t$.
This procedure, as illustrated in \cref{fig:rl}, may repeat for multiple steps, during which the agent uses the observed transitions $(\x_t, \u_t, r_t, \x_{t+1})$ to update its policy.

\begin{figure}[tbh]
    \begin{center}
    \begin{tikzpicture}[>=stealth, thick]
    
      \tikzstyle{robot}=[draw=black,rounded corners,minimum height=2.2em,
        minimum width=5.2em,thick,fill=red!30]
      \tikzstyle{environment}=[shape=ellipse,draw=black,minimum height=3em,
        minimum width=3em,thick,fill=gray!30]
    
      \node[robot] (agent) {Agent};
      \node[environment, right=3cm of agent] (env) {Environment};
    
      \coordinate (topA) at ($(agent.north)+(0,1.4)$);
      \coordinate (topE) at ($(env.north)+(0,1.3)$);
      \draw[->] (agent.north) -- (topA)
                -- node[midway,above,inner sep=2pt] {\footnotesize Control $\u_t$} (topE)
                -- (env.north);
    
      \coordinate (rout) at ($(env.south west)+(0,-1.0)$);
      \coordinate (sout) at ($(env.south east)+(0,-1.0)$);
    
      \draw (env.south west) -- node[midway,left,inner sep=2pt] {\footnotesize $R(\x_t, \u_t)$} (rout);
      \draw (env.south east) -- node[midway,right,inner sep=2pt] {\footnotesize $\x_{t+1}$} (sout);
    
      \draw[dashed] ($(env.south west)+(-0.4,-0.8)$) -- ($(env.south east)+(0.4,-0.8)$);
    
      \coordinate (rLeft) at ($(rout)+(-3.2,0)$);
      \coordinate (rIn)   at ($(agent.south east)+(-0.5,-1.0)$);
        \draw[->] (rout) -- (rLeft) -- (rIn) -- node[midway,right,inner sep=2pt] {\footnotesize Reward $r_t$} ($(agent.south east)+(-0.5,0)$);
    
      \coordinate (bottom) at ($(sout)+(0,-1)$);
      \coordinate (aBottom) at ($(agent.south west)+(0.5,-2)$);
      \draw (sout) -- (bottom);
      \draw (bottom) -- (aBottom);
      \draw (aBottom) -- ($(agent.south west)+(0.5,-2)$);
      \draw[->] ($(agent.south west)+(0.5,-2)$) -- node[midway,left,yshift=14pt,inner sep=2pt] {\footnotesize State $\x_t$} ($(agent.south west)+(0.5,0)$);

    
    \end{tikzpicture}
    \end{center}
    \caption{The reinforcement learning problem consists of an agent that learns how to make decisions by interacting with the environment. At each time step $t$, the agent observes the current state $\x_t$ and selects an action $\u_t$ to execute. The environment then transitions to a new state $\x_{t+1}$ and provides a reward $R(\x_t, \u_t)$ to the agent. This process continues as the agent learns to optimize its actions based on the received rewards.}
    \label{fig:rl}
    \end{figure}

\subsubsection{Problem Formulation}
\label{subsubsec:problem_formulation}
Let us briefly revisit the MDP framework, which provides a standard mathematical formalization of the reinforcement learning problem. 
Formally, an MDP is defined as a tuple:
$$
\mathcal{M} = \left(\mathcal{X}, \mathcal{U}, p, R, \gamma \right),
$$
where $\mathcal{X}$ denotes the \textit{state space}, which is the set of all possible environment states $\x \in \mathcal{X}$ and can be either discrete or continuous, $\mathcal{U}$ denotes the \textit{action space}, which is the set of admissible actions $\u \in \mathcal{U}$ and can also be discrete or continuous, $p$ characterizes the system dynamics through the transition probability distribution $p(\x_{t+1} \mid \x_t, \u_t)$, $R : \mathcal{X} \times \mathcal{U} \rightarrow \mathbb{R}$ specifies the reward function, and $\gamma \in (0,1]$ is a discount factor that determines the relative importance of future rewards.
From a reinforcement learning perspective, the goal is to learn a policy defined as a probability distribution over actions given states, $\pi(\u \given \x)$.
We will use the term \emph{trajectory} to refer to a sequence of states and actions of length $T$, given by:
\begin{equation*}
\tau \definedas (\x_0, \u_0, \ldots , \x_T),
\end{equation*}
where $T$ may be infinite.
Given a policy $\pi$, the induced \emph{trajectory distribution} $p_{\pi}$ is:
\begin{equation}
p_{\pi}(\tau) = p_0(\x_0)\prod_{t=0}^{T-1} \pi(\u_t \given \x_t) p(\x_{t+1} \given \x_t, \u_t).
\end{equation} 
where $p_0$ is the initial-state distribution.
The reinforcement learning objective is to maximize the expected discounted cumulative reward under this trajectory distribution, namely:
\begin{equation}
	V^\pi_T \definedas \expected{\tau \sim p_{\pi}(\tau)} {\sum_{t=0}^{T-1} \gamma^t r_t}.
\end{equation}

An additional concept required to fully characterize $V^\pi_T$ is that of \emph{discounting}.
In particular, the discount factor $\gamma$ is a scalar value in the range $[0,1]$ that determines the relative importance of future rewards, whereby a smaller $\gamma$ will make the agent focus more on immediate rewards, while a larger $\gamma$ will make the agent give more importance to future rewards.
For example, in the limit case with $\gamma = 0$, the agent will only consider immediate rewards, while in the case with $\gamma = 1$, the agent will consider all future rewards equally.
Mathematically, discounting is also crucial in ensuring that the sum of rewards in $V^\pi$ is finite even in the infinite horizon case with $T = \infty$, such that if $\gamma < 1$ and the rewards $r_t$ are bounded, the sum of the rewards will be finite.
In practice, the choice of $\gamma$ is often problem-dependent, and it is common to use a value close to $1$ to ensure that the agent considers future rewards.

\subsubsection{Value Functions and Bellman Equations}
\label{subsubsec:value_functions_bellman_equations}
Almost all reinforcement learning algorithms involve estimating \emph{value functions}.
At its core, a value function is a function of state or state-action pairs that defines how \emph{good}\sidenote{As defined in the previous section, value functions quantify quality in terms of expected cumulative future rewards.} it is for the agent to be in a given state or to take a given action in a given state.
Since the reward an agent expects to receive in the future depends on the actions it will take, the value function is inherently defined with respect to a particular policy $\pi$.

We define the \emph{state-value function}, $V^{\pi}(\x)$, as the expected sum of future rewards when starting in state $\x$ and following policy $\pi$ thereafter\sidenote{Throughout this chapter, we primarily consider the infinite-horizon case when referring to $V^\pi$, although the same concepts extend to the finite-horizon case.}:
\begin{equation}
    V^{\pi}(\x) \definedas \expected{\tau \sim p_{\pi}(\tau)} {\sum_{k=0}^{\infty} \gamma^{k} R(\x_{t+k}, \pi(\x_{t+k})) \given \x_t = \x}.
\end{equation}

Similarly, the \emph{action-value function}, $Q^{\pi}(\x, \u)$, is the expected return when starting in state $\x$, taking action $\u$, and then following policy $\pi$ thereafter:
\begin{equation}
    Q^{\pi}(\x, \u) \definedas \expected{\tau \sim p_{\pi}(\tau)} {\sum_{k=0}^{\infty} \gamma^{k} R(\x_{t+k}, \pi(\x_{t+k})) \mid \x_t = \x, \pi(\x_{t}) = \u}.
\end{equation}

A key property of value functions used in the context of reinforcement learning and dynamic programming is that they satisfy the \emph{Bellman equations}.
The Bellman equations describe a recursive relationship that decomposes the value of a state or state-action pair into the immediate reward and the value of the next state or state-action pair.
Formally, for any policy $\pi$ and any state $\x$, the Bellman equation defines the following self-consistency condition:
\begin{equation}
    V^{\pi}(\x) = \expected{\u \sim \pi(\cdot \given \x)} {R(\x, \u) + \gamma \expected{\x' \sim p(\cdot \given \x, \u)} {V^{\pi}(\x')}}, \label{eq:bellman_state_value}
\end{equation}
where, to simplify notation, we have omitted the time index $t$ and used $\x'$ to denote the next state.

Similarly, the Bellman equation for the action-value function is:
\begin{equation}
    Q^{\pi}(\x, \u) = R(\x, \u) + \gamma \expected{\x' \sim p(\cdot \given \x, \u), \, \u' \sim \pi(\cdot \given \x')} {Q^{\pi}(\x', \u')}. 
\label{eq:bellman_action_value}
\end{equation}
Crucially, the value functions $V^{\pi}$ and $Q^{\pi}$ are unique solutions to the Bellman equations.

In the remainder of this chapter, we show how we can use the Bellman equations to derive algorithms for estimating and approximating value functions, and how we can use these value functions to derive optimal policies.

\medskip
Central to the solution of reinforcement learning problems are the notions of \textit{optimal policies} and \textit{optimal value functions}. 
The term \emph{optimal} derives from the fact that value functions induce a partial ordering over policies, where a policy $\pi$ is said to be better than or equal to another policy $\pi'$ if its value is no worse in every state. 
Formally, we write $\pi \geq \pi'$ if and only if:
$$
V^{\pi}(\x) \geq V^{\pi'}(\x), \quad \forall \x \in \statespace.
$$
An \emph{optimal policy} $\pi^*$ is a policy that is better than or equal to all other policies, such that $\pi^* \geq \pi$ for all policies $\pi$.
While the optimal policy does not need to be unique, all optimal policies share the same \emph{optimal value function} $V^*$, defined as:
\begin{equation*}
    V^*(\x) \definedas \max_{\pi} V^{\pi}(\x), \quad \forall \x \in \statespace.
\end{equation*}
Similarly, optimal policies also share the same optimal action-value function $Q^*(\x, \u)$, defined as:
\begin{equation*}
    Q^*(\x, \u) \definedas \max_{\pi} Q^{\pi}(\x, \u), \quad \forall \x \in \statespace, \, \u \in \controlspace.
\end{equation*}

As discussed above, $V^*$ and $Q^*$ are value functions for the optimal policy, thus, they must satisfy the Bellman equations with respect to the optimal policy.
However, because $V^*$ and $Q^*$ are the optimal value functions, the Bellman equations can be written in a policy-independent form. 
This is achieved by exploiting the fact that, under an optimal policy, the value of a state is equal to the expected return obtained by selecting the \emph{best available action} in that state.

Formally, the Bellman equations for the optimal state-value function and action-value function---referred to as the \textit{Bellman optimality equations}---can be derived by substituting the expectation over the policy from \cref{eq:bellman_state_value} and \cref{eq:bellman_action_value} with a maximization over actions, and are given by:
\begin{align}
    V^*(\x) &= \max_{\u} \left[ R(\x, \u) + \gamma \expected{\x' \sim p(\cdot \given \x, \u)} {V^*(\x')} \right], \label{eq:bellman_optimality_state_value} \\
    Q^*(\x, \u) &= R(\x, \u) + \gamma \expected{\x' \sim p(\cdot \given \x, \u)} {\max_{\u'} Q^*(\x', \u')}. \label{eq:bellman_optimality_action_value}
\end{align}

Why are $V^*$ and $Q^*$ so central to reinforcement learning? 
The key reason is that, once either of these functions is known, deriving an optimal policy is substantially simpler. 
In particular, an optimal policy can be derived by acting greedily with respect to the optimal value functions.
For example, given the optimal state-value function $V^*$, the optimal policy can be computed as:
\begin{equation}
\label{eq:one-step-lookahead-policy}
    \pi^*(\x) = \argmax_{\u} \left[ R(\x, \u) + \gamma \expected{\x' \sim p(\cdot \given \x, \u)} {V^*(\x')} \right].
\end{equation}
That is, once $V^*$ is known, determining the optimal policy reduces to a one-step lookahead where, at each state, the agent selects the action that leads to successor states with the highest expected value.

Access to the optimal action-value function, $Q^*$, simplifies the process even further. 
For any state $\x$, we can obtain the optimal policy by selecting the action that maximizes $Q^*$:
\begin{equation}
\label{eq:argmax-qstar}
    \pi^*(\x) = \argmax_{\u} Q^*(\x, \u).
\end{equation}
Thus, by representing a function over states (or state-action pairs), the optimal value functions allow for the direct computation of the optimal policy.

\subsection{Dynamic Programming Methods}
\label{subsec:dyn_prog_methods}
As we introduced in \cref{ch:sequential-decision-making}, the key idea of dynamic programming is to decompose a complex problem into simpler subproblems.
This is achieved by using value functions to systematically organize and structure the search for optimal policies.
In this section, we show how we can leverage dynamic programming algorithms in the context of reinforcement learning by turning the Bellman equations into iterative update rules for the estimation of value functions.
In particular, we explore how to use dynamic programming ideas to derive algorithms for two distinct but interconnected tasks: \emph{prediction} and \emph{control}.
\begin{definition}[Prediction]
In the context of reinforcement learning, we often refer to the task of estimating the value function for a given policy as \emph{prediction}.
\end{definition}
\begin{definition}[Control]
In the context of reinforcement learning, we often refer to the task of finding the optimal policy as \emph{control}.
\end{definition}

\subsubsection{Prediction: Policy Evaluation}
\label{subsubsec:policy_evaluation}
We first consider the prediction problem of estimating the value function, $V^{\pi}$, under a given policy $\pi$.
According to the Bellman equation in \cref{eq:bellman_state_value}, the value of a state $\x$ under policy $\pi$ is defined as an expectation with respect to the policy and state transition model. 
For simplicity, we assume that the state transition model and policy describe probability distributions over discrete states and actions, respectively, which allows us to express the expectations in the Bellman equation as sums rather than integrals\sidenote{The extension to continuous states and actions is fundamentally equivalent and requires the replacement of summations with integrals.}:
\begin{equation}
    V^{\pi}(\x) = \sum_{\u \in \controlspace} \pi(\u \given \x) \left[ R(\x, \u) + \gamma \sum_{\x' \in \statespace} p(\x' \given \x, \u) V^{\pi}(\x') \right].
    \label{eq:policy_evaluation}
\end{equation}

\emph{Policy evaluation} is an iterative algorithm to solve the prediction problem.
Consider a sequence of approximations to the value function, denoted as $V_0, V_1, V_2, \ldots, V^{\pi}$, where $V_0$ is an arbitrarily chosen initial guess\sidenote{Under the condition that any \emph{terminal state}, occurring when $t = T$ in the finite-horizon setting or when the episode terminates in the infinite-horizon setting, must be assigned a value of zero.}.
Policy evaluation uses the Bellman equation in \cref{eq:policy_evaluation} as an update rule, such that at iteration $k$, the value function for all states $x \in \statespace$ is updated according to:
\begin{equation}
    V_{k+1}(\x) = \sum_{\u \in \controlspace} \pi(\u \given \x) \left[ R(\x, \u) + \gamma \sum_{\x' \in \statespace} p(\x' \given \x, \u) V_k(\x') \right].
    \label{eq:policy_evaluation_update}
\end{equation}
It is important to note that $V_k = V^{\pi}$ is a fixed point of the update rule in \cref{eq:policy_evaluation_update} since the Bellman equation for $V^{\pi}$ ensures equality in this case.
Moreover, under mild regularity conditions, it can be shown that the sequence of value functions $\{V_k\}$ converges to $V^{\pi}$ as $k \rightarrow \infty$.
As we will see in the remainder of this chapter, the ideas described above are at the core of many reinforcement learning algorithms, including both model-based and model-free methods.

\subsubsection{Policy Improvement}
\label{subsubsec:policy_improvement}
Having introduced an approach for solving the prediction problem, namely, estimating the value function under a fixed policy, we now turn to the problem of control, where the objective is to compute an optimal policy. 
To address the control problem, we rely on the \emph{policy improvement theorem}, which provides a principled mechanism for transforming a given policy into a new policy that is guaranteed to be better than or equal to the original.

Consider two policies, $\pi$ and $\pi'$, such that for all states $x \in \statespace$:
\begin{equation}
    Q^{\pi}(\x, \pi'(\x)) \geq V^{\pi}(\x).
    \label{eq:policy_improvement}
\end{equation}
Then, the policy $\pi'$ is guaranteed to be better than or equal to $\pi$, such that:
\begin{equation*}
    V^{\pi'}(\x) \geq V^{\pi}(\x).
\end{equation*}
Intuitively, if $\pi'$ chooses actions that are at least as good---according to the action-value function of $\pi$ in \cref{eq:policy_improvement}---as those prescribed by $\pi$ in every state (i.e., $V^{\pi}(\x)$), then following $\pi'$ cannot result in worse long-term performance than continuing to follow $\pi$.

Consider the \emph{greedy} policy $\pi'$, which selects, in each state, the action that maximizes the action-value function $Q^{\pi}$ associated with policy $\pi$\sidenote{In other words, the greedy policy performs a one-step lookahead using the current state-value function $V^{\pi}$.}:
\begin{equation}
\begin{split}
    \pi'(\x) &\definedas \argmax_{\u} Q^{\pi}(\x, \u) \\
    & = \argmax_{\u} \left[ R(\x, \u) + \gamma \sum_{\x' \in \statespace} p(\x' \given \x, \u) V^{\pi}(\x') \right].
\label{eq:greedy_policy}
\end{split}
\end{equation}
By construction, this greedy policy satisfies the condition of the policy improvement theorem in \cref{eq:policy_improvement} and is therefore guaranteed to be better than or equal to the original policy.
We refer to the process of constructing a new policy by greedily selecting actions with respect to the current value function as \emph{policy improvement}.

Suppose now that the greedy policy $\pi'$ is as good as the original policy $\pi$, such that $V^{\pi'} = V^{\pi}$.
From the definition of the greedy policy in \cref{eq:greedy_policy}, we have:
\begin{equation*}
    V^{\pi'}(\x) = \max_{\u} \left[ R(\x, \u) + \gamma \sum_{\x' \in \statespace} p(\x' \given \x, \u) V^{\pi}(\x') \right].
\end{equation*}
Since $V^{\pi'} = V^{\pi}$, this expression is equivalent to the Bellman optimality equation in \cref{eq:bellman_optimality_state_value}.
Therefore, the value function $V^{\pi}$ must equal the optimal value function, and the policy $\pi'$ is therefore optimal.

In other words, policy improvement provides a systematic procedure for iteratively improving a policy using its value function, and it converges once the optimal policy is reached.

\subsubsection{Control: Policy Iteration}
\label{subsubsec:policy_iteration}
The policy improvement theorem provides a concrete strategy to improve a policy by greedily selecting actions with respect to the current value function.
In this section, we discuss how we use this strategy, in tandem with policy evaluation, to construct an algorithm for finding the optimal policy.
We refer to this algorithm for finding the optimal policy as \emph{policy iteration}.
At a high level, the key idea of policy iteration is as follows: starting from a given policy $\pi$, we first evaluate it to obtain its value function $V^{\pi}$ and then improve the policy using this value function to produce a new policy $\pi'$. 
The improved policy $\pi'$ is subsequently evaluated to compute $V^{\pi'}$, which is in turn used to derive an improved policy $\pi''$. 
This alternating process of policy evaluation and policy improvement is repeated until the policy converges to the optimal policy.

More formally, policy iteration defines a sequence of monotonically improving policies by alternating between policy evaluation and policy improvement:
\begin{equation*}
    V_0 \xrightarrow{\hspace{2mm} \text{PE} \hspace{2mm} } \pi_0 \xrightarrow{\hspace{2mm} \text{PI} \hspace{2mm} } V_1 \xrightarrow{\hspace{2mm} \text{PE} \hspace{2mm} } \pi_1 \xrightarrow{\hspace{2mm} \text{PI} \hspace{2mm} } V_2 \xrightarrow{\hspace{2mm} \text{PE} \hspace{2mm} } \pi_2 \xrightarrow{\hspace{2mm} \text{PI} \hspace{2mm} } \ldots,
\end{equation*}
where PE denotes policy evaluation and PI denotes policy improvement.
We outline the policy iteration algorithm in \cref{alg:policy_iteration}.
\begin{algorithm}[ht]
\caption{Policy Iteration}
\label{alg:policy_iteration}
\KwData{Initial policy, $\pi$, and value function, $V_0$, arbitrarily initialized for all $\x \in \statespace$.}
\KwResult{Policy, $\pi \approx \pi^*$, and value function, $V^{\pi} \approx V^*$.}
{
    \For{$i = 0, \ldots, \infty$}
    {
        \textbf{Policy Evaluation:} \\
        \For{$k = 0, \ldots, \infty$}
        {
            \For{$x \in \statespace$}
            {
                $V_{k+1}(\x) = \sum_{\u \in \controlspace} \pi(\u \given \x) \left[ R(\x, \u) + \gamma \sum_{\x' \in \statespace} p(\x' \given \x, \u) V_k(\x') \right]$ \\
            }
            \If{$\| V_{k+1} - V_k \| < \epsilon$}
            {
                $V^{\pi} = V_{k+1}$ \\
                \textbf{break}
            }
        }
        \textbf{Policy Improvement:} \\
        \For{$x \in \statespace$}
        {
            $\pi'(\x) = \argmax_{\u} \left[ R(\x, \u) + \gamma \sum_{\x' \in \statespace} p(\x' \given \x, \u) V^{\pi}(\x') \right]$
        }
        \If{policy has converged}
        {
            \Return $\pi \approx \pi^*$ and $V^{\pi} \approx V^*$
        }
        $\pi = \pi'$ \\
    }
}
\end{algorithm}

Policy iteration is guaranteed to converge to the optimal policy and value function, given enough iterations. 
In practice, since the policy evaluation step is an iterative algorithm, we typically initialize the value function to the value function from the previous step of policy iteration.
This can increase the speed of convergence since the value function does not typically change substantially between iterations.

\begin{example}[Grid World Policy Iteration]
Explore an implementation of policy iteration for a simple grid-world environment in the repository \\\noindent \colorcode{github.com/StanfordASL/pora-exercises} in the notebook \\\noindent \colorcode{ch18/policy\_iteration.ipynb}.
The grid-world environment is a simple grid with action space:
\begin{equation*}
    \mathcal{U}
    \defn \{\texttt{up}, \texttt{down}, \texttt{left}, \texttt{right}\},
\end{equation*}
that has a set of absorbing states that get a reward of $0$, and every other state gets a reward of $-1$.
For this world, the state transitions are deterministic, so taking an action is guaranteed to result in moving in the desired direction (i.e. transition occurs with probability $1$), but we will consider a stochastic policy, $\pi(\u \given \x)$.
In the notebook for this example:
\begin{enumerate}
\item Run the provided code to see how the policy evaluation algorithm updates for a simple random policy.
Try playing around with the discount factor to see how it affects the value function.
\item Take a look at the policy iteration algorithm code.
Note that since we are considering a stochastic policy we use the \colorcode{softmax} function:
\begin{equation*}
\pi'(\u \given \x) = \frac{e^{\beta Q^\pi(\x, \u)}}{\sum_{\u' \in \controlspace} e^{\beta Q^\pi(\x, \u')}},
\end{equation*}
to define the policy probability, where $\beta$ is the Boltzmann constant that when increased will decrease the entropy of the distribution for $\x$ (i.e. as $\beta \rightarrow \infty$ the policy will approach being deterministic).
Play around with the value of $\beta$ to see how it affects the optimal value function.
\end{enumerate}
\end{example}

\subsubsection{Control: Value Iteration}
\label{subsubsec:value_iteration}
One drawback of policy iteration is that it requires a full policy evaluation step at each iteration.
As a result, the algorithm can be computationally expensive, since it must wait for the value function to converge before performing a policy improvement step, which only happens in the limit\sidenote{Several variants of policy iteration mitigate this cost by using truncated or approximate policy evaluation steps.}. 
To address this limitation, \emph{value iteration} provides an alternative approach that combines policy evaluation and policy improvement into a single step.
Value iteration defines the following update rule:
\begin{equation}
    V_{k+1}(\x) = \max_{\u} \left[ R(\x, \u) + \gamma \sum_{\x' \in \statespace} p(\x' \given \x, \u) V_k(\x') \right].
\label{eq:value_iteration}
\end{equation}
Starting from an arbitrary initial value function $V_0$, the sequence of value functions $\{V_k\}$ generated by value iteration is guaranteed to converge to the optimal value function $V^*$.

Value iteration can be interpreted through the lens of the Bellman optimality equation, where the update in \cref{eq:value_iteration} corresponds to applying the Bellman optimality operator from \cref{eq:bellman_optimality_state_value} to the current value function $V_k$.
In contrast to policy iteration, which alternates between iteratively solving the Bellman equation for a fixed policy and performing a separate policy improvement step, value iteration directly applies the Bellman optimality equation at every iteration.
We outline the complete value iteration algorithm in \cref{alg:value_iteration}.
\begin{algorithm}[ht]
\caption{Value Iteration}
\label{alg:value_iteration}
\KwData{Initial value function, $V_0$, arbitrarily initialized for all $\x \in \statespace$.}
\KwResult{Policy, $\pi \approx \pi^*$, and value function, $V^{\pi} \approx V^*$.}
\For{$k=0$ \KwTo $\infty$}{
	\For{$x \in \statespace$}{
            $V_{k+1}(\x) = \max_{\u} \left[ R(\x, \u) + \gamma \sum_{\x' \in \statespace} p(\x' \given \x, \u) V_k(\x') \right]$
        }
        \If{$\| V_{k+1} - V_k \| < \epsilon$}
        {
            $V^\pi = V_{k+1}$ \\
            $\pi(\x) = \argmax_{\u} \left[ R(\x, \u) + \gamma \sum_{\x' \in \statespace} p(\x' \given \x, \u) V_k(\x') \right]$, for all $\x \in \statespace$ \\
            \Return $V^{\pi} \approx V^*$ and $\pi \approx \pi^*$
        }
    }
\end{algorithm}

Policy iteration and value iteration are two of the most foundational algorithms in reinforcement learning and are the basis for many modern reinforcement learning algorithms.

\subsubsection{Inheriting the Limitations of Dynamic Programming}
\label{subsubsec:inheriting_limitations}
Dynamic programming methods, including policy iteration and value iteration, are powerful tools for solving MDPs.
However, these methods also inherit the limitations of dynamic programming as they require a complete model of the environment, $p(\x' \given \x, \u)$, to compute the expectations in the Bellman equations, and are computationally expensive for large state and action spaces.
In the next sections, we discuss two key approaches to address these limitations.
We first introduce \emph{sampling methods}, which relax the requirement of having a complete model of the environment, and then we introduce the concept of \emph{function approximation}, which helps address the computational complexity of dynamic programming methods.

\subsection{Learning Paradigms for Model-free Control}
\label{subsec:model_free_control}
In this section, we introduce two classes of learning methods that estimate value functions and compute optimal policies without requiring a complete model of the environment: \emph{Monte Carlo (MC) methods} and \emph{temporal-difference (TD) learning}.
These methods are particularly effective in practice since they are applicable to a wide range of problems where modeling the environment dynamics is either impractical or infeasible.
Similar to our discussion of dynamic programming methods, we first address the prediction problem for both Monte Carlo and temporal-difference methods before extending the analysis to the control problem.

\subsubsection{Monte Carlo Methods}
\label{subsubsec:monte_carlo_methods}
The term \emph{Monte Carlo} broadly refers to a class of algorithms that rely on random sampling to estimate quantities of interest.
In the context of reinforcement learning, Monte Carlo methods represent a class of approaches for solving the reinforcement learning problem based on averaging observed cumulative rewards from experience\sidenote{We often use the terms \emph{samples} or \emph{experience} to refer to sequences of states, actions, and rewards collected by interacting with the environment.}.
We assume that the agent interacts with the environment for a fixed number of time steps $T$, referred to as an \emph{episode}, during which it collects a trajectory consisting of states, actions, and rewards. 
Episodic MDPs naturally describe tasks with well-defined beginnings and endings, such as navigating a maze, playing a game, or completing a robotic assembly. 
At the end of each episode, the agent uses these observed trajectories to estimate quantities of interest, such as value functions or policy updates.

\medskip
Monte Carlo methods for solving the prediction problem aim to learn the value function $V^{\pi}$ (or equivalently, $Q^{\pi}$) given a policy $\pi$.
In this section, we first address the problem of using Monte Carlo methods for learning the state-value function $V^{\pi}$, and then extend the discussion to learning the action-value function $Q^{\pi}$.
Since the value of a state is defined as the expected cumulative reward obtained when starting from that state, it can be estimated by averaging the cumulative rewards observed across episodes that visit the state. 
As the agent visits a state more frequently, the estimate of its value becomes increasingly accurate.

Formally, suppose we wish to learn the value $V^{\pi}(\x)$ of the state $\x$ under the policy $\pi$, using a set of $N$ episodes $\{\tau_1, \tau_2, \ldots, \tau_N\}$ passing through $\x$.
The state-value function can be estimated using the Monte Carlo prediction procedure outlined in \cref{alg:monte_carlo_prediction}.
\begin{algorithm}[ht]
\caption{Monte Carlo Prediction}
\label{alg:monte_carlo_prediction}
	\KwData{Initial value function, $V$, arbitrarily initialized for all $\x \in \statespace$.}
	\KwResult{Value function estimate, $V^\pi$.}
    Initialize the state visit count, $N(\x) = 0$, for all $\x \in \statespace$ \\
    \For{each episode $\tau_i = \{\x_0, \u_0, r_0, \x_1, \u_1, r_1, \ldots, \x_T\}$}
    {
        \For{$t = T-1$ \KwTo $0$}
        {
            Compute the cumulative future reward from state $\x_t$: $G_t = \sum_{k=t}^{T-1} \gamma^{k-t} r_k$ \\
            Increment the visit count: $N(\x_t) \leftarrow N(\x_t) + 1$ \\
            Update the value estimate: $V(\x_t) \leftarrow V(\x_t) + \frac{1}{N(\x_t)} (G_t - V(\x_t))$
        }
    }
    \Return $V \approx V^{\pi}$
\end{algorithm}
In particular, Monte Carlo prediction updates the value estimate $V(x_t)$ using the incremental update rule:
$$
V(\x_t) \leftarrow V(\x_t) + \frac{1}{N(\x_t)} (G_t - V(\x_t)),
$$ 
which moves the current estimate toward the observed return $G_t$, where $N(x_t)$ denotes the number of times state $x_t$ has been visited.
This incremental form provides an efficient way to compute sample averages and is equivalent to the empirical mean:
$$
V(\x_t) = \frac{1}{N(\x_t)} \sum_{i=1}^{N(\x_t)} G_t^i,
$$
where $G_t^i$ is the cumulative future reward observed in the $i$-th visit to state $\x_t$.

We can also use Monte Carlo methods for learning action values\sidenote{Learning action values $Q^{\pi}$ is useful in practice because we can use them to directly derive an optimal policy using \cref{eq:argmax-qstar}.}.
At a high level, Monte Carlo methods for estimating action value functions $Q^{\pi}(\x, \u)$ are essentially equivalent to the method presented above for estimating state values, with the only difference being that we now consider visits to state-action pairs instead of states.
We consider a state-action pair to have been visited in an episode if the agent is in state $\x$ and takes action $\u$ at some point during the episode.
As with state-value estimation, Monte Carlo methods estimate the action-value function $Q^{\pi}(x, u)$ by averaging the cumulative rewards observed from visits to the state–action pair $(x, u)$.
However, estimating action values introduces an additional challenge.

Suppose that experience is generated using a deterministic policy $\pi$. 
Under such a policy, the agent selects the same action whenever it encounters a given state and therefore observes rewards for only one action per state. 
As a result, the agent receives no information about the values of alternative actions in that state, making it impossible to compare actions and improve the policy.
This challenge is known as the problem of maintaining exploration and is central to reinforcement learning.
As we will see throughout the remainder of this chapter, a common strategy for addressing this issue is to employ policies that ensure every state–action pair is visited with nonzero probability. 
One way to achieve this is by using stochastic policies that assign positive probability to all actions in each state, thereby guaranteeing sufficient exploration of the environment.

\subsubsection{Temporal-Difference Learning}
\label{subsubsec:td_learning}
\emph{Temporal-difference (TD) learning} is widely considered one of the most influential concepts in reinforcement learning.
At a high level, temporal-difference learning combines elements of both Monte Carlo methods and dynamic programming. 
Like dynamic programming methods, temporal-difference learning updates value estimates using other learned estimates rather than waiting for complete returns.
This mechanism is known as \emph{bootstrapping} and allows learning from incomplete sequences of experience\sidenote{In other words, without waiting for the end of an episode.}
At the same time, like Monte Carlo methods, temporal-difference learning learns directly from sampled experience and does not require an explicit model of the environment: a property referred to as \emph{sampling}.
By combining bootstrapping with sampling, temporal-difference learning effectively bridges the gap between Monte Carlo and dynamic programming methods, inheriting many of the advantages of both approaches.

\medskip
Similar to Monte Carlo methods, temporal-difference methods address the prediction problem by collecting samples from the environment and using them to update value estimates. 
Recall that in Monte Carlo methods, we must wait until the end of the episode to compute the cumulative reward following time $t$, which we denote as $G_t$, and then use $G_t$ to define the target for the value function update:
\begin{equation}
    V(\x_t) \leftarrow V(\x_t) + \alpha (G_t - V(\x_t)),
\end{equation}
where $\leftarrow$ denotes the assignment operator and $\alpha$ is an externally-specified step-size parameter\sidenote{In \cref{alg:monte_carlo_prediction}, we used $\alpha = 1/N(\x_t)$}.
In contrast, temporal-difference methods update the value function estimate at each time step $t$ based on the observed reward $r_t$ and the estimate of the value function at the next state $V(\x_{t+1})$ by the update:
\begin{equation}
    V(\x_t) \leftarrow V(\x_t) + \alpha (r_t + \gamma V(\x_{t+1}) - V(\x_t)).
\end{equation}
This update rule is known as the \emph{TD($0$)} update, where the subscript $0$ indicates that the update relies on a single step of experience\sidenote{TD($0$) is a special case of the more general TD($\lambda$) family of methods.}. 
By comparing the Monte Carlo and temporal-difference update rules, we observe that computing the Monte Carlo target $G_t$ requires access to an entire episode, whereas the temporal-difference target $r_t + \gamma V(\x_{t+1})$ can be computed immediately at each time step $t$.
We provide a complete algorithm for TD($0$) in \cref{alg:td_prediction}.
\begin{algorithm}[ht]
\caption{Temporal-Difference Learning (TD($0$))}
\label{alg:td_prediction}
\KwData{Initial value function, $V$, arbitrarily initialized for all $\x \in \statespace$.}
\KwResult{Value function estimate, $V^\pi$.}
\For{each episode}{
    Initialize state $\x_0$ \\
    \For{$t = 0, 1, \ldots$ until $\x_t$ is terminal}{
        Take action $\u_t$ according to $\pi$ \\
        Observe $r_t$ and $\x_{t+1}$ \\
        $V(\x_t) \leftarrow V(\x_t) + \alpha (r_t + \gamma V(\x_{t+1}) - V(\x_t))$ \\
    }
}
\Return $V \approx V^\pi$
\end{algorithm}

It is worth noting that the quantity $r_t + \gamma V(\x_{t+1}) - V(\x_t)$ can be interpreted as an error that measures the discrepancy between the current estimate of the value function, $V(\x_t)$, and an improved target estimate, $r_t + \gamma V(\x_{t+1})$\sidenote{This target is more informative because it incorporates the realized reward $r_t$ from the transition $\x_t \rightarrow \x_{t+1}$, together with the current estimate of the value of the successor state.}. 
This quantity, known as the \emph{TD error}, plays a crucial role in the development of many reinforcement learning algorithms.

\subsubsection{Example: Monte Carlo Control}
\label{subsubsec:mc_control}
We have already discussed how we can use the Monte Carlo and temporal-difference learning paradigms to address the problem of learning without a model of the environment.
However, we solely discussed these methods in the context of the prediction problem, where we are trying to estimate the value function of a given policy.
In this section, we introduce our first complete example of a reinforcement learning algorithm for learning optimal policies through Monte Carlo methods.
Later in this chapter, we introduce various reinforcement learning algorithms that, in one way or another, build upon the principles of model-free control that we discuss here.

The central idea behind using Monte Carlo methods for control mirrors the principles underlying policy iteration, as introduced in \cref{subsec:dyn_prog_methods}. 
We use the term \emph{Generalized Policy Iteration} (GPI) to refer to the broad framework encompassing all methods that alternate between policy evaluation and policy improvement.
To illustrate this idea, consider a straightforward Monte Carlo extension of the classical policy iteration algorithm. 
Starting from an arbitrary initial policy $\pi_0$, the algorithm alternates between two phases: policy evaluation, in which the value function of the current policy is estimated from sampled experience, and policy improvement, in which the policy is updated based on these value estimates. 
This process is repeated until convergence.

As discussed in \cref{subsubsec:monte_carlo_methods}, learning action-value functions is often more convenient than learning state-value functions, since action values can be used directly to derive policies. 
Accordingly, in this example we focus on learning the action-value function. 
At a high level, the algorithm proceeds by repeatedly alternating between the following steps:
\begin{equation*}
    \pi_0 \xrightarrow{\hspace{2mm} \text{E} \hspace{2mm} } Q^{\pi_0} \xrightarrow{\hspace{2mm} \text{I} \hspace{2mm} } \pi_1 \xrightarrow{\hspace{2mm} \text{E} \hspace{2mm} } Q^{\pi_1} \xrightarrow{\hspace{2mm} \text{I} \hspace{2mm} } \pi_2 \xrightarrow{\hspace{2mm} \text{E} \hspace{2mm} } \dots \xrightarrow{\hspace{2mm} \text{I} \hspace{2mm} } \pi^* \xrightarrow{\hspace{2mm} \text{E} \hspace{2mm} } Q^{\pi^*},
\end{equation*}
where $\xrightarrow{\hspace{2mm} \text{E} \hspace{2mm} }$ denotes policy evaluation, and $\xrightarrow{\hspace{2mm} \text{I} \hspace{2mm} }$ denotes policy improvement.
In contrast to the Policy Iteration algorithm, we use the Monte Carlo prediction approach from \cref{subsubsec:monte_carlo_methods} in the policy evaluation step rather than using the exact Bellman equation to update the value function, which would require a model of the environment.
The policy improvement step remains the same, where we define the new policy by acting greedily with respect to the current action-value function by choosing $\pi_{i+1}(\x) = \argmax_{\u} Q^{\pi_i}(\x, \u)$.

While this algorithm captures several core principles underlying many reinforcement learning algorithms, it remains relatively simplistic. 
As discussed in earlier sections, learning accurate action-value estimates from experience requires the agent to maintain sufficient exploration. 
More formally, the behavior policy must ensure that every state–action pair is visited with nonzero probability.
In the formulation presented above, however, no explicit exploration mechanism is incorporated to guarantee this condition. 
As a result, the algorithm is only valid under the \emph{exploring starts} assumption, which posits that each episode begins in every possible state–action pair with nonzero probability. 
Although useful for theoretical analysis, this assumption is generally unrealistic in practical applications.
To obtain a practical and broadly applicable algorithm, this assumption must be relaxed. 
In the following sections, we introduce several strategies for ensuring adequate exploration, enabling effective learning from experience in realistic settings.

\subsubsection{A Unifying View of Reinforcement Learning}
\label{subsubsec:unifying}
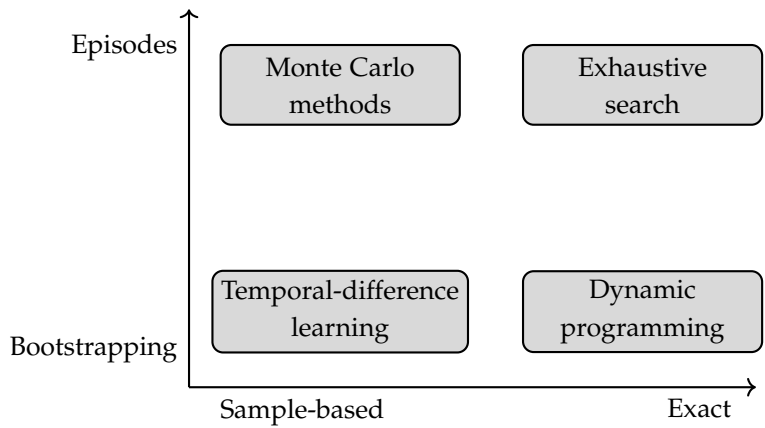
\begin{figure}[ht]
\begin{center}
    \begin{tikzpicture}[node distance=2cm, ->]
        \tikzstyle{node} = [draw=black, rectangle, rounded corners, minimum height=3em, minimum width=9em, thick, fill=white!30]
        \node[node, align=center, fill=gray!30](td){Temporal-difference \\ learning};
        \node[node, align=center, right of=td, xshift=2cm, yshift=0cm, fill=gray!30](dp){Dynamic \\ programming};
        \node[node, align=center, above of=td, xshift=0cm, yshift=1cm, fill=gray!30](mc){Monte Carlo \\ methods};
        \node[node, align=center, above of=dp, xshift=0cm, yshift=1cm, fill=gray!30](es){Exhaustive \\ search};
        \draw[->, thick] (-2,-1) to[] node[pos=.2,below,inner sep=4pt] {Sample-based} node[pos=0.9,below,inner sep=4pt] {Exact} (5.5,-1);
        \draw[->, thick] (-2,-1) to[] node[pos=.1,left,inner sep=4pt] {Bootstrapping} node[pos=0.9,left,inner sep=4pt] {Episodes} (-2,4);
    \end{tikzpicture}
\end{center}
\caption{We can categorize reinforcement learning methods along two-axes based on whether they are sample-based and whether they bootstrap.}
\label{fig:rl-unified-view}
\end{figure}

Monte Carlo, temporal-difference, and dynamic programming methods are often presented as distinct approaches to reinforcement learning.
However, it is worth noting that these methods are extremes of a spectrum.
To appreciate this, we can consider the advantages, disadvantages, and commonalities of each of these paradigms.

Monte Carlo and temporal-difference methods have an advantage over dynamic programming methods in that they do not require an explicit model of the environment and can learn directly from interaction with it—that is, they rely on \emph{sampling}.
This property greatly extends the applicability of these methods to real-world problems, where the environment could be unknown or too complex to model.

Temporal-difference and dynamic programming methods, in turn, offer a key advantage over Monte Carlo methods in that they update value estimates without waiting for an episode to terminate and can therefore learn from incomplete sequences of experience through \emph{bootstrapping}.
This capability is particularly important in practice, as many real-world tasks involve very long or even non-terminating episodes.

On the other hand, Monte Carlo methods also possess an advantage over temporal-difference methods in that they provide unbiased estimates of the \emph{true} value function.
This is because Monte Carlo updates rely on the actual return $G_t$, whereas temporal-difference methods update toward an estimated target, $R(\x_t, \u_t) + \gamma V(\x_{t+1})$.
By doing so, temporal-difference methods deliberately trade some bias---arising from bootstrapped targets---for a reduction in variance, since these targets depend on fewer sources of stochasticity.

These relationships are summarized in \cref{fig:rl-unified-view}, which organizes reinforcement learning methods along two dimensions: the use of samples and the degree of bootstrapping.
At the top right lies exhaustive search, where quantities of interest---such as value functions---are computed exactly through model-based simulation of all possible system evolutions.
At the bottom right are dynamic programming methods, which leverage the principle of optimality and a known model to perform one-step lookahead updates via bootstrapping.
Moving leftward relaxes the requirement of a model or exhaustive computation, replacing it with learning from sampled experience.
At the top left are Monte Carlo methods, which avoid exhaustive search by learning from complete episodes. 
Finally, at the bottom left are temporal-difference methods, which combine the sampling of Monte Carlo methods with the bootstrapping of dynamic programming, enabling efficient learning from incomplete sequences of experience.

\subsection{A Taxonomy of Reinforcement Learning}
\label{subsec:taxonomy_rl}
\begin{figure}[tbh]
\begin{center}
    \begin{tikzpicture}[node distance=2cm, ->]
        \tikzstyle{node} = [draw=black, rectangle, rounded corners, minimum height=3em, minimum width=3em, thick, fill=white!30]
        \tikzstyle{examples} = [draw=black, rectangle, rounded corners, minimum height=3em, minimum width=8em, thick, fill=red!30]
        \node[node, fill=red!30](rl){Reinforcement Learning};
        \node[node, right of=rl, xshift=1cm, yshift=-1.75cm, fill=gray!30](modelfree){Model-free};
        \node[node, left of=rl, xshift=-1cm, yshift=-1.75cm, fill=gray!30](modelbased){Model-based};
        \node[node, right of=modelbased, xshift=-1cm, yshift=-1.75cm](learnmodel){Learned Model};
        \node[node, left of=modelbased, xshift=-1cm, yshift=-1.75cm](knownmodel){Known Model};
        \node[node, right of=modelfree, xshift=1cm, yshift=-1.75cm](policybased){Policy Optimization};
        \node[node, left of=modelfree, xshift=1cm, yshift=-1.75cm](valuebased){Value-based};
		\draw[->, thick] (rl) to[bend left=25] (modelfree);
		\draw[->, thick] (rl) to[bend right=25]  (modelbased);
		\draw[->, thick] (modelbased) to[bend left=25]  (learnmodel);
		\draw[->, thick] (modelbased) to[bend right=25]  (knownmodel);
		\draw[->, thick] (modelfree) to[bend left=25]  (policybased);
		\draw[->, thick] (modelfree) to[bend right=25]  (valuebased);
    \end{tikzpicture}
\end{center}
\caption{A taxonomy of reinforcement learning algorithms based on whether or not they leverage a model of the environment.}
\label{fig:taxonomy_rl}
\end{figure}
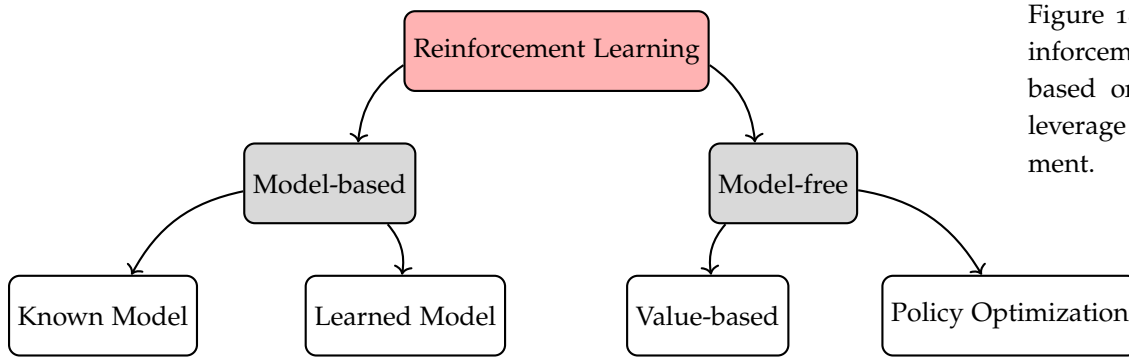

Over the last years, the field of reinforcement learning has seen a rapid growth in the number of algorithms and methods, each with its own strengths and weaknesses.
While an exhaustive treatment of all these methods is beyond the scope of this chapter, we aim to provide an overall picture of the different types of algorithms that exist, a deeper understanding of the core principles that underlie these algorithms, and a number of representative examples from each category.
In this section, we provide a bird's-eye view of the field of reinforcement learning and classify the different algorithms into a taxonomy, shown graphically in \cref{fig:taxonomy_rl}, that can serve as reference through the rest of the chapter.

The first fundamental distinction among reinforcement learning algorithms is between \emph{model-free} and \emph{model-based} methods.
Model-free methods, which are the focus of \cref{subsec:model_free_rl}, attempt to learn the optimal policy directly from experience, without explicitly modeling the environment.
Model-based methods, on the other hand, aim to learn a model of the environment and then use this model to derive optimal policies.
We cover model-based methods in \cref{subsec:model_based_rl}.

Looking at model-free methods, we can further distinguish between \emph{value-based} and \emph{policy optimization} methods.
Value-based methods, similar to the ones described in previous sections, define the policy implicitly through a value function.
By doing so, the main focus of value-based methods lies in accurately estimating the optimal value function, which we can then use to derive the optimal policy.
Policy optimization methods, on the other hand, represent the policy explicitly via a parametric function and optimize the parameters of this function to maximize the reinforcement learning objective.

On the other side of the graph in \cref{fig:taxonomy_rl}, we can see how model-based methods can be further divided into algorithms that either focus on learning a model of the environment from data, or algorithms that use a known model\sidenote{For example, the model could be known from physics or other domain knowledge.} to derive optimal policies.
In this chapter, we focus on methods that learn a model of the environment from data.
As we will see in \cref{subsec:model_based_rl}, learned models can be used in various ways, such as within planning routines or for accelerating model-free algorithms.

\subsubsection{On-policy vs Off-policy Learning}
\label{subsubsec:on_off_policy_learning}
Another crucial distinction among reinforcement learning algorithms is between \emph{on-policy} and \emph{off-policy} learning.
On-policy methods aim to evaluate or improve the policy that is used to interact with the environment.
Off-policy methods, on the other hand, evaluate or improve a policy that is different from the one used to interact with the environment.
For example, the Monte Carlo control algorithm we presented in \cref{subsubsec:mc_control} is an on-policy method, as it attempts to improve the same policy used to make decisions in the environment.
Off-policy methods define two distinct policies: one that must be updated and ideally becomes the optimal policy and one used to interact with the environment.
We refer to the policy being learned as the \emph{target policy}, while we refer to the policy used to generate samples from the environment as the \emph{behavior policy}.

In general, on-policy methods are simpler to implement and tend to be more reliable than off-policy methods. 
Off-policy approaches, on the other hand, are typically more complex and require additional algorithmic considerations, which can make the learning phase less stable and lead to slower convergence in practice.
Off-policy learning enables agents to learn from data generated by arbitrary behavior policies, including data collected by human demonstrators or by conventional, non-learning-based controllers. 
Moreover, off-policy methods make it possible to learn about multiple policies simultaneously, or to learn an optimal policy while continuing to explore the environment.

Throughout the next sections, we introduce various on-policy and off-policy methods for learning and discuss the advantages and disadvantages of each approach.

\subsection{Model-free Reinforcement Learning}
\label{subsec:model_free_rl}
Model-free reinforcement learning methods are commonly considered to be the most popular and widely used class of algorithms in the field.
This popularity is largely due to their recent successes in a wide range of applications, including playing games\cite{SilverEtAl2016}, robot control\cite{LevineEtAl2016}, and the fine-tuning of large-scale AI chatbots\cite{BaiEtAl2022}.
In this section, we discuss representative reinforcement learning algorithms while maintaining our focus on key principles important for understanding and implementing new algorithms.

\subsubsection{Value-based Methods}
\label{subsubsec:value_based_methods}
We begin our discussion on model-free reinforcement learning with value-based methods, which, similarly to the Monte Carlo control algorithm presented in \cref{subsubsec:mc_control}, estimate the value function of a policy and use this estimate to derive the optimal policy.

\paragraph{Q-learning.}
One of the first breakthroughs in reinforcement learning was the introduction of the \emph{Q-learning} algorithm\cite{WatkinsDayan1992}.
Q-learning is an off-policy algorithm that learns the optimal action-value function $Q^*(x, u)$ via TD learning.
Because Q-learning is off-policy, it estimates the optimal action value function $Q^*(\x, \u)$ independently of the behavior policy used to interact with the environment.
The behavior policy therefore does not determine the objective of learning, but instead governs which state–action pairs are explored and, consequently, which action-value estimates are updated.
Provided that the behavior policy is sufficiently exploratory\sidenote{That is, all state–action pairs are visited infinitely often in the limit.}, Q-learning is guaranteed to converge to the optimal action-value function $Q^*(\x, \u)$.

Given a transition sampled from the environment, $(\x, \u, r, \x')$, Q-learning updates the action-value estimate $Q(\x, \u)$ according to:
\begin{equation}
    Q(\x, \u) \leftarrow Q(\x, \u) + \alpha \left( r + \gamma \max_{\u'} Q(\x', \u') - Q(\x, \u) \right),
\label{eq:q_learning_update}
\end{equation}
where $\alpha > 0$ is a tunable learning rate parameter.
Importantly, regardless of the behavior policy used to generate the transition $(\x, \u, r, \x')$, the update in \cref{eq:q_learning_update} adjusts the current estimate $Q(\x, \u)$ toward the action value induced by the greedy policy with respect to the current Q-function, namely $r + \gamma \max_{\u'} Q(\x', \u')$.
In this sense, Q-learning repeatedly enforces the Bellman optimality equation for the action-value function until convergence\sidenote{As discussed in \cref{subsubsec:value_iteration}, the Bellman optimality operator has a unique fixed point, which corresponds to the optimal action-value function $Q^*(\x, \u)$.}.
From this perspective, Q-learning can be viewed as a sample-based approximation of Q-value iteration\sidenote{That is, value iteration applied to action-value functions rather than state-value functions.}, in which the Bellman optimality update is performed using a single transition sample.

To ensure that the behavior policy is sufficiently exploratory, Q-learning typically employs an \emph{$\epsilon$-greedy} action-selection strategy.
Despite its simplicity, the $\epsilon$-greedy policy approach is an effective and widely used exploration mechanism where, with probability $1 - \epsilon$, the agent selects the action that is greedy with respect to the current Q-function, while with probability $\epsilon$ it selects an action uniformly at random.
This strategy ensures that $\pi(\u \given \x) > 0$ for all states $\x$ and actions $\u$, which is a key requirement for the convergence of Q-learning.
We present the complete Q-learning algorithm in \cref{alg:q_learning}.
\begin{algorithm}[ht]
\caption{Q-Learning}
\label{alg:q_learning}
	\KwData{Initial action values, $Q(\x, \u)$, learning rate, $\alpha$, discount factor, $\gamma$, exploration rate, $\epsilon$.}
	\KwResult{Updated action values, $Q(\x, \u)$.}
    \For{each episode}
    {
        Initialize the state, $\x_0$. \\
        \For{each step in the episode}
        {
            Select an action, $\u_t$, using an $\epsilon$-greedy policy with respect to $Q(\x_t, \u)$. \\
            Execute the action $\u_t$ and observe the reward, $r_t$, and the next state, $\x_{t+1}$. \\
            Update the action-value function using the transition $(\x_t, \u_t, r_t, \x_{t+1})$. \\
            $Q(\x_t, \u_t) \leftarrow Q(\x_t, \u_t) + \alpha \left( r_t + \gamma \max_{\u'} Q(\x_{t+1}, \u') - Q(\x_t, \u_t) \right)$
        }
    }
    \Return $Q(\x, \u) \approx Q^*(\x, \u)$
\end{algorithm}

\paragraph{Value Function Approximation.}
So far, we have assumed that the state and action spaces, $\statespace$ and $\controlspace$, are finite and small enough to be easily stored in a look-up table and to allow for meaningful state-action space exploration within a reasonable computation budget.
However, in robotics applications, the state and action spaces are often continuous or extremely high-dimensional, making it impractical to compactly store and efficiently update value functions.
This challenge motivates us to consider methods that rely on \emph{parametrized function approximation}.
Within this class of methods, we represent value functions by a parametric function, $Q_{\btheta}(\x, \u)$, or alternatively $V_{\btheta}(\x)$, with parameters $\btheta$.
The goal of learning is then to find the optimal parameters $\btheta^*$, such that the value function estimator is close to the optimal value function.

Function approximation has two key advantages.
First, it allows us to represent value functions compactly, as the number of parameters $\btheta$ is typically much smaller than the number of states and actions.
Second, it enables the value function estimator to generalize to unseen states and actions, potentially reducing the amount of exploration required to learn a good policy\sidenote{In other words, if the function approximator is able to generalize successfully, the agent does not need to visit every state-action pair.}.
While there are many choices for the function approximator, including linear functions, neural networks, and decision trees, we focus our discussion on differentiable functions.

Given a dataset, $\mathcal{D}$, of transitions $(\x, \u, r, \x')$, policy evaluation via function approximation entails learning the parameters $\btheta$ that minimize the loss function $J(\btheta)$: 
\begin{equation}
    J(\btheta) = \expected{(\x, \u, r, \x') \sim \mathcal{D}} {Q^{\pi}(\x, \u) - Q_{\btheta}(\x, \u)}.
\end{equation}
Intuitively, $J(\btheta)$ measures the discrepancy between the estimated value function, $Q_{\btheta}(\x, \u)$, and the target value function, $Q^{\pi}(\x, \u)$, where $Q^{\pi}(\x, \u)$ is the value function under the policy $\pi$ that generated the transitions in $\mathcal{D}$.
This optimization problem can be solved via stochastic gradient descent, which defines the following update rule:
\begin{equation}
\label{eq:function-approx-param-update}
    \btheta \leftarrow \btheta + \Delta_{\btheta},
\end{equation}
where:
\begin{equation*}
	\Delta_{\btheta} = \alpha \left(Q^{\pi}(\x, \u) - Q_{\btheta}(\x, \u) \right) \nabla_{\btheta} Q_{\btheta}(\x, \u).
\end{equation*}
Here, $\nabla_{\btheta} Q_{\btheta}(\x, \u)$ denotes the gradient of the value function estimator with respect to its parameters $\btheta$, and $\alpha > 0$ is the learning rate.

In practice, however, the update rule in \cref{eq:function-approx-param-update} is not directly applicable since the true target value function $Q^{\pi}(\x, \u)$ is unknown.
Leveraging ideas from model-free control, the unknown target can be replaced by an estimate derived from Monte Carlo, temporal-difference, or dynamic programming methods.
For example, using a Monte Carlo estimate, the parameter update can be written as:
\begin{equation}
\Delta_{\btheta} = \alpha \left( G_t - Q_{\btheta}(\x_t, \u_t) \right) \nabla_{\btheta} Q_{\btheta}(\x_t, \u_t),
\end{equation}
where $G_t = \sum_{k=t}^{T-1} \gamma^{k-t} r_k$ denotes the return observed from time step $t$.
Alternatively, a temporal-difference approach replaces the Monte Carlo return with a one-step bootstrap estimate, yielding the update:
\begin{equation}
\Delta_{\btheta} = \alpha \left( r_t + \gamma Q_{\btheta}(\x_{t+1}, \u_{t+1}) - Q_{\btheta}(\x_t, \u_t) \right) \nabla_{\btheta} Q_{\btheta}(\x_t, \u_t).
\end{equation}
In both cases, the Monte Carlo and temporal-difference targets are computed from trajectories generated by applying the behavior policy in the environment.

\paragraph{Fitted Q-learning.}
A particularly popular algorithm that combines function approximation with temporal-difference learning is the \emph{Fitted Q-Learning} algorithm.
Fitted Q-Learning updates the parameters $\btheta$ of a Q-function estimator by applying the update rule in \cref{eq:function-approx-param-update} with:
\begin{equation}
    \Delta_{\btheta} = \alpha \left( r_t + \gamma \max_{\u'} Q_{\btheta}(\x_{t+1}, \u') - Q_{\btheta}(\x_t, \u_t) \right) \nabla_{\btheta} Q_{\btheta}(\x_t, \u_t).
\end{equation}
It is important to note that the target $r_t + \gamma \max_{\u'} Q_{\btheta}(\x_{t+1}, \u')$ is equivalent to the temporal-difference target used in Q-learning.
Essentially, Fitted Q-Learning mimics the update rule of Q-learning, but rather than directly updating the action values to explicitly enforce the Bellman optimality equation it updates the parameters $\btheta$ of the Q-function estimator to approximately enforce it.
In other words, rather than updating the entries of a look-up table representing the value function, Fitted Q-learning updates the parameters $\btheta$ to minimize the error with respect to the fixed point of the Bellman optimality operator.

\medskip
In this section, we discussed a few foundational examples of value-based reinforcement learning methods.
While these are only a subset of the vast literature in value-based methods, they convey the key ideas and challenges of learning value functions for control.
Specifically, these methods highlight the central idea of approximating value functions from experience and leveraging them to derive optimal policies. 
Key challenges include ensuring sufficient exploration, which we can address through strategies like $\epsilon$-greedy policies, and addressing stability and convergence of the learning process when using value function approximators, particularly in high-dimensional or continuous spaces.
Value-based methods in reinforcement learning are fundamentally built on a concise set of core principles, such as generalized policy iteration and (approximate) value iteration, and differ primarily in their usage of value update targets\sidenote{For instance, Monte Carlo, temporal-difference, or dynamic programming approaches.}, function approximators, or behavior policies. 
Therefore, understanding these foundational concepts provides us a lens through which we can interpret the majority of value-based algorithms.

\subsubsection{Policy Optimization Methods}
\label{subsubsec:policy_optimization_methods}
We now continue our discussion of model-free reinforcement learning by turning to policy optimization (PO) methods. 
In contrast to value-based approaches, policy optimization methods adopt a fundamentally different strategy for solving the reinforcement learning problem by directly optimizing the policy itself.
To better motivate policy optimization, let us recall the reinforcement learning objective from \cref{subsec:rl_problem}:
\begin{equation*}
    V^\pi = \expected{\tau \sim p_\pi(\tau)}{\sum_{t=0}^{T-1} \gamma^t r_t}.
\end{equation*}
Rather than learning a value function and deriving a policy from it, as in value-based methods, policy optimization methods define a parametric policy, $\pi_{\btheta}$, and directly optimize the parameters $\btheta$ to maximize the reinforcement learning objective $V^\pi$.
Formally, the goal of policy optimization is to find the optimal policy parameters:
\begin{equation}
    \btheta^* = \underset{\btheta}{\arg \max } \, \, V(\btheta),
\end{equation}
where, for simplicity, we use $V(\btheta)$ to refer to the reinforcement learning objective, $V^{\pi_{\btheta}}$, under policy $\pi_{\btheta}$ with parameters $\btheta$.

Policy optimization methods typically address this problem in two stages. 
First, they estimate the gradient of the objective with respect to the policy parameters, $\nabla_{\boldsymbol{\theta}} V(\boldsymbol{\theta})$. 
Then, they update the parameters by performing (approximate) gradient ascent:
\begin{equation}
    \btheta \leftarrow \btheta + \alpha \nabla_{\btheta} V(\btheta),
\end{equation}
where $\alpha$ is a user-defined learning rate.

The first challenge that all policy optimization methods face is estimating the gradient of the reinforcement learning objective.
To simplify the notation, we define the cumulative reward as $R(\tau) = \sum_{t=0}^{T-1} \gamma^t r_t$ and assume $\gamma = 1$\sidenote{The extension to the discounted case is equivalent and relatively straightforward.}.
By definition of expectation, we have:
\begin{equation}
    V(\btheta) = \expected{\tau \sim p_{\btheta}(\tau)} {R(\tau)} = \int p_{\btheta}(\tau) R(\tau)  \d\tau,
    \label{eq:expectation_rl_objective}
\end{equation}
where $p_{\btheta}(\tau)$ denotes the trajectory distribution induced by the policy $\pi_{\btheta}$.
\cref{eq:expectation_rl_objective} allows us to write the gradient of the reinforcement learning objective as:
\begin{equation}
    \nabla_{\btheta} V(\btheta) = \nabla_{\btheta} \int R(\tau) p_{\btheta}(\tau) \d\tau = \int \nabla_{\btheta} p_{\btheta}(\tau) R(\tau) \d\tau.
    \label{eq:gradient_rl_objective}
\end{equation}
However, we cannot compute this gradient directly because it depends on unknown dynamics through the trajectory distribution, $p_{\btheta}(\tau)$\sidenote{Recall that we do not assume access to the system dynamics and thus cannot compute $p_{\btheta}(\tau)$ explicitly.}.

To address this issue, we resort to the following useful identity:
\begin{equation}
    p_{\btheta}(\tau)\nabla_{\btheta} \log p_{\btheta}(\tau) = p_{\btheta}(\tau) \frac{\nabla_{\btheta} p_{\btheta}(\tau)}{p_{\btheta}(\tau)} = \nabla_{\btheta} p_{\btheta}(\tau),
\end{equation}
and use it to rewrite the gradient of the reinforcement learning objective in \cref{eq:gradient_rl_objective} as:
\begin{align*}
    \nabla_{\btheta} V(\btheta) & = \int \nabla_{\btheta}p_{\btheta}(\tau) R(\tau) \d\tau \\
    & = \int p_{\btheta}(\tau)\nabla_{\btheta} \log p_{\btheta}(\tau) R(\tau) \d\tau \\
    & = \expected{\tau \sim p_{\btheta}(\tau)} {\nabla_{\btheta} \log p_{\btheta}(\tau) R(\tau)}.
\end{align*}

We can then approximate the expectation using Monte Carlo methods by sampling from the trajectory distribution, $p_{\btheta}(\tau)$, through interaction with the environment. 
However, the gradient of the log-probability, $\nabla_{\btheta} \log p_{\btheta}(\tau)$, remains intractable to compute directly.
To address this, let us recall the definition of the trajectory distribution, $p_{\btheta}(\tau)$:
\begin{equation}
    p_{\btheta}(\tau) \definedas p(\x_0) \prod_{t=0}^{T-1} p(\x_{t+1} \given \x_t, \u_t) \pi_{\btheta}(\u_t \given \x_t),
\end{equation}
where taking the logarithm yields:
\begin{equation}
    \log p_{\btheta}(\tau) = \log p(\x_0) + \sum_{t=0}^{T-1} \log p(\x_{t+1} \given \x_t, \u_t) + \log \pi_{\btheta}(\u_t \given \x_t).
    \label{eq:log_prob_traj} 
\end{equation}
By substituting \cref{eq:log_prob_traj} into the gradient of the reinforcement learning objective, we obtain:
\begin{equation*}
\begin{split}
    \nabla_{\btheta} V(\btheta) & = \expected{\tau \sim p_{\btheta}(\tau)} {\nabla_{\btheta} \log p_{\btheta}(\tau) R(\tau) }\\
    & = \expected{\tau \sim p_{\btheta}(\tau)} {\nabla_{\btheta} \left( \log p(\x_0) + \sum_{t=0}^{T-1} \log p(\x_{t+1} \given \x_t, \u_t) + \log \pi_{\btheta}(\u_t \given \x_t) \right) R(\tau)}.
\end{split}
\end{equation*}
Notably, the terms $\log p(\x_0)$ and $\log p(\x_{t+1} \given \x_t, \u_t)$ do not depend on $\btheta$, and thus can be ignored when computing the gradient of the reinforcement learning objective.
Moreover, evaluating the gradient of the log-probability of the action, $\log \pi_{\btheta}(\u_t \given \x_t)$, is tractable and we can easily compute it, for example by using automatic differentiation tools.

This leads to the following expression for the gradient of the reinforcement learning objective:
\begin{equation}
    \nabla_{\btheta} V(\btheta) = \expected{\tau \sim p_{\btheta}(\tau)}{\sum_{t=0}^{T-1} \nabla_{\btheta} \log \pi_{\btheta}(\u_t \given \x_t) R(\tau)},
    \label{eq:gradient_rl_objective_final}
\end{equation}
which is tractable to compute and which we can estimate using samples from the environment.
For example, given $N$ episodes of interaction with the environment, we can estimate the gradient of the reinforcement learning objective as:
\begin{equation}
    \nabla_{\btheta} V(\btheta) \approx \frac{1}{N} \sum_{i=1}^{N} \sum_{t=0}^{T-1} \nabla_{\btheta} \log \pi_{\btheta}(\u_t^i \given \x_t^i) R(\tau^i).
    \label{eq:gradient_rl_objective_final_estimate}
\end{equation}
This is a crucial result and lays the foundations for almost all policy optimization algorithms in reinforcement learning.

\paragraph{REINFORCE.}
The derivations above directly lead to one of earliest examples of policy optimization methods\sidenote{In the reinforcement learning literature, these are often also referred to as \emph{policy gradient methods}.} known as the \emph{REINFORCE algorithm}\cite{Williams1992}.
At a high level, the REINFORCE algorithm estimates the gradient of the reinforcement learning objective in \cref{eq:gradient_rl_objective_final_estimate} using samples from the environment, and after each episode updates the policy parameters $\btheta$ in the direction of the gradient.
We outline the REINFORCE algorithm in \cref{alg:reinforce}.
\begin{algorithm}
\caption{REINFORCE Algorithm}
\label{alg:reinforce}
\KwData{Initial policy parameters, $\btheta$, learning rate, $\alpha$}
\KwResult{Update policy parameters, $\btheta$}
\For{each episode} {
    Generate a trajectory, $\tau = \{\x_0, \u_0, r_0, \ldots, \x_T\}$, using the policy $\pi_{\btheta}$ in the environment. \\
    Compute the return, $R(\tau) = \sum_{t=0}^{T-1} r_t$. \\
    $\btheta \leftarrow \btheta + \alpha \nabla_{\btheta}V(\btheta)$, where $\nabla_{\btheta}V(\btheta)$ is computed by \cref{eq:gradient_rl_objective_final_estimate}.
}
\Return $\btheta$
\end{algorithm}

From \cref{eq:gradient_rl_objective_final}, we can see that the gradient is computed as the sum of the gradients of the log-probabilities of the actions, weighted by the return of the trajectory.
Although this result follows from a mathematical derivation, it also admits a clear and intuitive interpretation. 
By updating the policy parameters in the direction of the policy gradient, the algorithm increases the log-probability of actions that lead to high returns while decreasing the log-probability of actions that lead to low returns. 
In this way, the policy gradient formalizes the notion of \emph{trial-and-error} learning, where behaviors that prove effective are reinforced, whereas ineffective behaviors are gradually suppressed.

Policy optimization methods represent a popular and intuitive approach to reinforcement learning and have several advantages and disadvantages compared to value-based methods.
A first key advantage of policy optimization methods is that they can naturally handle both discrete and continuous action spaces because we can compute the gradient of the policy using automatic differentiation tools\sidenote{Assuming the policy is parameterized by a differentiable function.}.
For example, in the case of continuous action spaces, we can parameterize the policy as a Gaussian distribution, and all of the derivations developed in this section would still hold.
Additionally, policy optimization methods have the notable advantage of directly optimizing the reinforcement learning objective.
This ensures that improvements to the policy are measured against a well-defined metric, since better values of the reinforcement learning objective imply a better policy. 
In contrast, value-based methods rely on fixed-point iterations of value functions to satisfy the Bellman equation with the goal to eventually converging to the optimal value function. 
While this approach is theoretically sound, it is unclear how suboptimal the policy is during intermediate iterations.

Policy gradient methods also have some disadvantages. 
First, the policy optimization methods presented so far are inherently \emph{on-policy} methods, which can result in high sample inefficiency\sidenote{A number of off-policy policy optimization algorithms have been introduced to allow the policy to be updated using experiences collected from different policies. 
These methods aim to approximate the behavior of classical on-policy algorithms while improving sample efficiency. 
Despite this advantage, they often introduce additional complexities, such as the need for more sophisticated exploration strategies and managing the stability of the off-policy updates.}.
A second disadvantage is that the gradient defined in \cref{eq:gradient_rl_objective_final_estimate} is a high-variance estimator of the true gradient from \cref{eq:gradient_rl_objective_final}.
In practice, this can lead to extremely noisy updates and therefore slow convergence.

As we will see in the remainder of this section, a lot of research in the domain of policy optimization has focused on addressing these limitations to develop sample-efficient and lower-variance policy gradient estimates.

\paragraph{Actor-Critic Methods.}
Actor-critic methods represent an important extension of policy optimization that reduces the high variance associated with policy gradient estimates. 
Let us recall the definition of the policy gradient and slightly rearrange the summation terms to highlight the sum over future rewards\footnote{Where again, for simplicity, we assume $\gamma = 1$.}:
\begin{equation*}
    \nabla_{\btheta} V(\btheta) = \frac{1}{N} \sum_{i=1}^{N} \sum_{t=0}^{T-1} \nabla_{\btheta} \log \pi_{\btheta}(\u_t^i \given \x_t^i) \left( \sum_{t'=t}^{T-1} r_{t'}^i \right).
\end{equation*}
We refer to the term $\sum_{t’=t}^{T-1} r_{t’}^i$ as the \emph{reward-to-go}.
The reward-to-go is a one-sample estimate of the true return, which is defined as the expected cumulative reward under the trajectory distribution, $\expected{\tau \sim p_{\btheta}(\tau)}{\sum_{t’=t}^{T-1} r_{t’}}$.
While conceptually straightforward, this reward-to-go estimate introduces significant variance, leading to noisy policy gradient updates.

Actor-critic methods address this challenge by introducing a \emph{critic}, which is a parametric approximation of the value function. 
Since the value function estimates the expected reward-to-go, the critic enables us to replace the high-variance sample-based estimate with a lower-variance, learned approximation. 
Concretely, the policy gradient becomes:
\begin{equation}
    \nabla_{\btheta} V(\btheta) \approx \frac{1}{N} \sum_{i=1}^{N} \sum_{t=0}^{T-1} \nabla_{\btheta} \log \pi_{\btheta}(\u_t^i \given \x_t^i) Q_{\phi}(\x_t^i, \u_t^i),
\end{equation}
where $Q_{\phi}(\x_t, \u_t)$ is the critic’s estimate of the action-value function. 
We can update the critic using any value estimation method, such as the value-based methods with function approximation discussed in \cref{subsubsec:value_based_methods}.

A particularly popular choice for the definition of the policy gradient in actor-critic methods is through the \emph{advantage function}:
\begin{equation}
    A^{\pi}(\x_t, \u_t) = Q^{\pi}(\x_t, \u_t) - V^{\pi}(\x_t),
\end{equation}
which quantifies the relative merit of taking action $\u_t$ in state $\x_t$, compared to the average value of the state. 
Intuitively, if the policy gradient in the REINFORCE algorithm can be thought of as a way to increase the probability of good actions, the advantage function can be thought of as a way to increase the probability of actions that are better than the average.
This normalization of the policy gradient through the advantage defines the so-called \emph{Advantage Actor Critic (A2C)} \cite{MnihEtAl2016} algorithm, which significantly reduces variance and improves learning stability. 
In practice, we often approximate the advantage function to avoid estimating both $Q^{\pi}$ and $V^{\pi}$, using:
\begin{equation}
    A^{\pi}(\x_t, \u_t) \approx r_t + \gamma V^{\pi}(\x_{t+1}) - V^{\pi}(\x_t).
\end{equation}

Actor-critic methods typically involve iterative updates of both the policy parameters, $\btheta$, and the value function parameters, $\phi$. 
At each step, we improve the policy based on the estimated advantage and we refine the value function to better approximate future rewards. 
The pseudocode for A2C is presented in \cref{alg:a2c} and highlights these alternating updates.
\begin{algorithm}[ht]
\caption{Advantage Actor Critic (A2C)}
\label{alg:a2c}
	\KwData{Initial policy parameters, $\btheta$, and value function parameters, $\phi$, learning rates, $\alpha_{\btheta}$, $\alpha_{\phi}$, discount factor, $\gamma$}
	\KwResult{Updated policy parameters, $\btheta$}
    \For{each episode} {
        Sample trajectories, $\tau = \{\x_0, \u_0, r_0, \ldots, \x_T\}$, using $\pi_{\btheta}$. \\
        \For{$t=0$ \textbf{to} $T-1$} {
            Compute the value target, for example using temporal-difference: $y_t = r_t + \gamma V_{\phi}^{\pi}(\x_{t+1})$. \\
            $\phi \leftarrow \phi + \alpha_{\phi} \nabla_{\phi} ( y_t - V_{\phi}^{\pi}(\x_t))^2$ \\
            $A_{\phi}^{\pi}(\x_t, \u_t) = y_t - V_{\phi}^{\pi}(\x_t)$ \\
            $\btheta \leftarrow \btheta + \alpha_{\btheta} \nabla_{\btheta} \log \pi_{\btheta}(\u_t \given \x_t) A_{\phi}^{\pi}(\x_t, \u_t)$
        }
    }
    \Return $\btheta$
\end{algorithm}

Actor-critic methods blend the strengths of policy optimization and value-based approaches, achieving a balance between expressive policy representations and efficient variance reduction. 
However, the introduction of a critic also adds computational complexity and tuning challenges that we must consider in practical implementations.

\subsubsection{Limitations of Model-free Reinforcement Learning}
Despite their successes, model-free reinforcement learning methods face several important challenges. 
A primary limitation is \emph{sample efficiency}.
Model-free algorithms often require a large number of interactions with the environment to learn an effective policy, which can be prohibitively expensive or impractical in many real-world settings, particularly in robotics.
A second limitation is that, in their standard formulation, model-free methods are inherently \emph{single-task learners}. 
Given a fixed reward function defining a specific task, these methods learn a policy optimized exclusively for that task, making it difficult to reuse knowledge or transfer learned behaviors across tasks.
Finally, many real-world applications provide reward signals that are sparse, delayed, or noisy. 
Such reward structures significantly represent a challenge to the learning process, as useful feedback may be infrequent or difficult to attribute to specific actions. 
As a result, model-free methods can struggle to discover effective policies in these environments.

\subsection{Model-based Reinforcement Learning}
\label{subsec:model_based_rl}
Model-based reinforcement learning methods aim to address the limitations of model-free methods by learning a model of the environment.
In this section, we introduce two broad classes of model-based reinforcement learning methods: \emph{model-based planning} methods that learn a model and use it to plan and \emph{model-based policy optimization} methods that learn a model and use it to accelerate model-free policy learning.

\subsubsection{Model-based Planning}
\label{subsubsec:model_based_planning}
If we had access to a model of the dynamics, $p(\x_{t+1} \given \x_t, \u_t)$, we could directly leverage tools from model-based optimal control to compute an optimal action sequence or policy.
Motivated by this observation, the central idea behind model-based planning methods is to learn an approximate model of the environment dynamics from data, and then use this learned model to plan.

A generic model-based planning procedure can be summarized as follows:
\begin{enumerate}
    \item Run a base policy, $\pi_0$, in the environment and collect a dataset of transitions, $\mathcal{D} = \{(\x_t, \u_t, r_t, \x_{t+1})\}$.
    \item Fit a dynamics model, $p_{\btheta}(\x_t, \u_t)$, to the observed data to minimize the prediction error, for example by minimizing the mean squared error between the predicted and true next state: 
    $$\minimize[\btheta] \sum_{(\x_t, \u_t, r_t, \x_{t+1}) \in \mathcal{D}} \left\| p_{\btheta}(\x_t, \u_t) - \x_{t+1} \right\|^2.$$
    \item Use the learned dynamics model to plan a sequence of actions for the agent to execute.
\end{enumerate}
Despite its simplicity, this scheme works for relatively well-behaved systems, where the dataset $\mathcal{D}$ guarantees sufficient coverage of the state-action space, and where the learned model $p_{\btheta}$ is accurate enough to enable effective planning\sidenote{This scheme is essentially equivalent to a task known as \emph{system identification}.}.
However, in practice, learning an accurate model of the dynamics is often challenging, especially when dealing with high-dimensional, non-linear, and stochastic systems.
Most importantly, inaccuracies in the learned model are particularly problematic when used within an optimization process. 
Optimization algorithms will naturally exploit inaccuracies in the model that have high \emph{predicted} performance, but may not correspond to high \emph{realized} performance.

A popular approach to address this issue is to consider a measure of uncertainty in the model's predictions, and to use this uncertainty to inform the planning process.
While there are many ways to quantify uncertainty, we consider methods that aim to learn a \emph{posterior distribution} over the model parameters.
In these methods, rather than learning a single estimate of the model parameters $\btheta$ through standard maximum likelihood estimation:
\begin{equation}
    \btheta^* = \argmax_{\btheta} \log p_{\btheta}(\mathcal{D} \given \btheta),
\end{equation}
where $p_{\btheta}(\mathcal{D} \given \btheta)$ is the likelihood of the data given the model parameters, we instead aim to learn a posterior distribution, $p(\btheta \given \mathcal{D})$, over the model parameters.
In other words, we aim to learn a full distribution over the model parameters that is consistent with the observed data, potentially capturing multiple plausible models that explain the data, and ultimately enabling us to reason about the uncertainty in the model's predictions.
We can achieve this by applying Bayes' rule to compute the posterior distribution:
\begin{align}
    p(\btheta \given \mathcal{D}) = \frac{p(\mathcal{D} \given \btheta) p(\btheta)}{p(\mathcal{D})},
\end{align}
where $p(\mathcal{D} \given \btheta)$ is the likelihood of the data given the model parameters, $p(\btheta)$ is the prior distribution over the model parameters, and $p(\mathcal{D})$ is the marginal likelihood of the data.
While a complete treatment of Bayesian inference is beyond the scope of this book, we refer the interested reader to standard textbooks on the subject, such as by \citet{Murphy2022}.
For the purpose of this chapter, it will be sufficient to assume that Bayesian inference provides us with computational methods to derive an estimate of the posterior distribution, which we can then use to inform the planning process.

Once we have an estimate of the posterior distribution over the model parameters, we can apply it in the following model-based planning scheme:
\begin{enumerate}
    \item Run a base policy, $\pi_0$, in the environment and collect a dataset of transitions, $\mathcal{D} = \{(\x_t, \u_t, r_t, \x_{t+1})\}$.
    \item Use $\mathcal{D}$ to estimate a posterior distribution, $p(\btheta \given \mathcal{D})$, over the model parameters.
    \item Sample a set of $K$ plausible models, $\{\btheta_1, \ldots, \btheta_K\} \sim p(\btheta \given \mathcal{D})$.
    \item For each sampled model, $\btheta_k$, and given a candidate action plan, $(\u_1, \ldots, \u_{T-1})$, use the model to compute the expected return: 
    \begin{equation*}
    		V(\u_1, \ldots, \u_{T-1}) = \frac{1}{K} \sum_{k=1}^K \sum_{t=1}^{T-1} R(\x_t, \u_t),
    \end{equation*}
    where $\x_{t+1} \sim p_{\btheta_k}(\x_{t+1} \given \x_t, \u_t)$.
    \item Execute the first action from the best plan according to the expected return.
\end{enumerate}
This scheme allows us to leverage the uncertainty in the model's predictions by considering multiple plausible models and to reason about the expected return under each model rather than optimizing under a single model.

Despite its effectiveness, it is important to note that this scheme is just a high-level description of the model-based planning process, and there are many practical considerations that we would need to address to make this approach work in practice.

\subsubsection{Model-based Policy Optimization}
\label{subsubsec:model_based_policy_optimization}
The second class of model-based reinforcement learning methods we consider is \emph{model-based policy optimization}.
In contrast to model-based planning, which uses the learned model to plan a sequence of actions, model-based policy optimization uses the learned model to improve model-free policy learning.

Specifically, having a learned model allows us to consider two sources of experience: real-world data collected by executing the policy in the environment and synthetic data generated by the model.
Given an MDP, $\mathcal{M} = (\mathcal{S}, \mathcal{A}, p, R, \gamma)$, and a learned model, $p_{\btheta}(\x_{t+1}, r_t \given \x_t, \u_t)$\sidenote{Here we consider the general case where we learn both the next state and the reward, but we can extend the discussion to the case where we only have to learn one of the two.}, we can consider two sources of experience:
\begin{align*}
    (\x_t, \u_t, r_t, \x_{t+1}) &\sim \text{Environment data}, \\
    (\x_t, \u_t, \hat r_t, \hat \x_{t+1}) &\sim p_{\btheta}(\x_{t+1}, r_t \given \x_t, \u_t).
\end{align*} 

The basic idea of model-based policy optimization is to use both sources of experience to improve model-free policy learning.
One of the earliest and most popular methods in this category is the \emph{Dyna-Q algorithm}\cite{Sutton1991}.

\paragraph{Dyna-Q.}
The Dyna-Q algorithm is a model-based reinforcement learning algorithm that improves the learning efficiency of Q-learning by using the learned model to generate synthetic data\sidenote{The term Dyna-Q derives from the fact that the algorithm combines Q-learning with model-based acceleration. 
The term \emph{Dyna} more generally refers to the idea of using a learned model to generate synthetic data for model-free learning.}.
The algorithm is based on the idea that in addition to updating the Q-function using real-world data, we can also update the Q-function using synthetic data generated by the learned model.
At a high-level, the algorithm alternates between three main steps.
First, it performs standard Q-learning updates, using real experience collected from interactions with the environment to update the Q-function. 
Second, it carries out a model learning step, in which the dynamics model is updated based on the same real-world data. 
Finally, the algorithm executes a model-based acceleration step, during which the learned model is used to generate synthetic experience that is then leveraged to perform additional Q-function updates.
A detailed description of this algorithm is provided in \cref{alg:dyna_q}.
\begin{algorithm}
\caption{Dyna-Q}
\label{alg:dyna_q}
\KwData{Model, $p_{\btheta}(\x_{t+1}, r_t \given \x_t, \u_t)$, number of model-based acceleration steps, $n$}
\KwResult{Updated model parameters, $\btheta$, action value function $Q(\x, \u) \approx Q^*(\x,\u)$}
\For{each episode}{
    Initialize $\x_0$. \\
    \For{each step $t$}{
        Select action $\u_t = \pi(\x_t)$. \\
        Observe reward $r_t$ and next state $\x_{t+1}$. \\
        $Q(\x_t, \u_t) \leftarrow Q(\x_t, \u_t) + \alpha \left( r_t + \gamma \max_{\u'} Q(\x_{t+1}, \u') - Q(\x_t, \u_t) \right)$ \\
        Update model parameters, $\btheta$, using sample $(\x_t, \u_t, r_t, \x_{t+1})$. \\
        \For{$i = 1, \ldots, n$}{
            Sample $\x_t, \u_t$ from real-world data. \\
            Generate synthetic data: $(\hat \x_{t+1}, \hat r_t) \sim p_{\btheta}(\x_{t+1}, r_t \given \x_t, \u_t)$. \\
            $Q(\x_t, \u_t) \leftarrow Q(\x_t, \u_t) + \alpha \left( \hat r_t + \gamma \max_{\u'} Q(\hat \x_{t+1}, \u') - Q(\x_t, \u_t) \right)$
        }
    }
    \Return $\btheta$, $Q(\x, \u) \approx Q^*(\x,\u)$
}
\end{algorithm}

The Dyna-Q algorithm is a simple yet powerful method that demonstrates the potential of model-based reinforcement learning to improve the learning efficiency of model-free algorithms. 

\subsubsection{Limitations of Model-based Reinforcement Learning}
\label{subsubsec:limitations_of_model_based_rl}
Model-based methods are an extremely active and promising area of research in reinforcement learning, but they are also subject to several limitations.
First, model learning entails optimizing the parameters of the model to minimize prediction error. 
However, this objective does not necessarily align with the objective of the agent, which is to maximize the expected cumulative reward, and this discrepancy can lead to suboptimal policies.
Second, model-based methods are sensitive to model errors, which can cause the agent to learn suboptimal policies or exploit the model errors to achieve high rewards, potentially leading to catastrophic failures.
Finally, learning an accurate model of the environment is a challenging task, especially in complex environments with high-dimensional state and action spaces. 

\subsection{Summary}
\label{subsec:learning_from_interaction}
In this chapter, we provided a comprehensive overview of the field of reinforcement learning. 
Rather than presenting an exhaustive list of algorithms, we focused on a conceptual understanding of the key ideas and principles that underlie reinforcement learning. 
In particular, we discussed how Monte Carlo methods and temporal-difference learning represent two foundational paradigms in reinforcement learning, and how we can use these methods to estimate value functions and learn optimal policies. 
Most importantly, we highlighted how these methods, together with dynamic programming, define a full spectrum of possible approaches to the problem of learning from interaction. 
Finally, we also discussed concrete examples of the main algorithmic families in reinforcement learning, including model-free and model-based methods, and highlighted the key ideas behind some of the most popular algorithms within these categories.

\paragraph{To learn more.}
For readers interested in a deeper exploration of the topics covered in this chapter, several authoritative resources are available. 
The definitive reference in the field is the textbook by \citet{SuttonBarto2018}, which provides a comprehensive and accessible introduction to the core principles of reinforcement learning, including Monte Carlo methods, temporal-difference learning, and function approximation. 
For a more rigorous treatment of the subject that emphasizes the connections between reinforcement learning, dynamic programming, and optimal control, we refer the reader to \citet{Bertsekas2019}.

\section{Exercises}
The starter code for the exercises provided below is available online through GitHub. 
To get started, download the code by running in a terminal window:

\begin{tcolorbox}[colback=gray!10]
\begin{minted}{bash}
    git clone https://github.com/StanfordASL/pora-exercises.git
\end{minted}
\end{tcolorbox}

We denote Problems requiring hand-written solutions and coding in Python with \adjustbox{height=2ex, valign=c}{\includegraphics{figs/write.png}} and \adjustbox{height=2ex, valign=c}{\includegraphics{figs/code.png}}, respectively.

\subsection*{\adjustbox{height=2ex, valign=c}{\includegraphics{figs/code.png}}\ Problem 1: Q-learning Widget Sales}
You are the owner of Widget Co., a shop in the business of buying widgets wholesale and selling them to consumers at a markup. 
The shop is able to store between 0 and 5 widgets at a time, and we denote the number of widgets held in the shop on day $t$ as $s_t$. 
Every day, you choose how many widgets to order from your supplier. 
You can order either zero widgets, a ``half order'' of $2$ widgets, or a ``full order'' of $4$ widgets. 
We write the number of widgets ordered to arrive on day $t$ as $a_t$. 
A random number of customers (following an unknown distribution, though this distribution may be assumed to be consistent across all days) come to Widget Co.~every day; each customer buys a widget if there are any available. 
We write the demand on day $t$ as $d_t$, and assume $d_t \leq 5$. 
At the end of each day, you record a net profit $r_t$ for that day.

In this exercise, we will explore using Q-learning to help model returns and optimize the shop's performance.
In the notebook for this exercise, \\\noindent\colorcode{ch18/exercises/widget\_sales.ipynb}, complete the following:
\begin{enumerate}
    \item We have a dataset $\mathcal{D} = \{ (s_t, a_t, r_t) \}_{t=1}^T$ containing records for each day $t$ of the last three years of the shop's operation. 
    In the provided notebook, fill in the function \colorcode{q\_learning} to implement a $Q$-learning algorithm to learn tabulated $Q$-values from this dataset.
    \item Suppose you find that the dynamics of the number of widgets in the shop each day are described by:
\begin{equation*}
    x_{t+1} = f(x_t, u_t, d_t) \defn \begin{cases}
        0,                  &x_t + u_t - d_t < 0 \\
        5,                  &x_t + u_t - d_t > 5 \\
        x_t + u_t - d_t,    &\text{otherwise}
    \end{cases},
\end{equation*}
and the daily net profit is:
\begin{equation*}
    R(x_t, u_t, d_t)
    = c_\text{sell}\min(x_t + u_t, d_t) - c_\text{rent} - c_\text{storage}x_t - g_\text{order}(u_t),
\end{equation*}
where $c_\text{sell} = 1.2$ is the price you set for each widget, $c_\text{rent} = 1$ is the fixed rent on your shop, $c_\text{storage} = 0.05$ is the cost for storing each widget overnight, and $g_\text{order}(u_t) = \sqrt{u_t}$ is the cost of ordering widgets from your supplier. 
The quantity $\min(x_t + u_t, d_t)$ is the ``satisfied demand'' on day~$t$.

Let's also suppose that after a few weeks of sales, you determine that the daily demand distribution for your widgets seems to be:
\begin{equation*}
    d_t = \begin{cases}
        0,  &\text{with probability}~0.1 \\
        1,  &\text{with probability}~0.3 \\
        2,  &\text{with probability}~0.3 \\
        3,  &\text{with probability}~0.2 \\
        4,  &\text{with probability}~0.1
    \end{cases}.
\end{equation*} 
	In the function \colorcode{action\_value\_iteration}, implement value iteration to learn tabulated $Q$-values from the model described above.
    Specifically, use the following update equation:
\begin{equation*}
    Q_{k+1}(x, u) = \expected{d \sim p(d_t)} {R(x, u, d) + \gamma \max_{u'} Q_k(f(x, u, d), u')},
\end{equation*}
	which is a slight variation of \cref{eq:value_iteration} for learning $Q$-values adapted for this problem's model (see also \cref{eq:bellman_optimality_action_value}).
    Compare the $Q$-values from $Q$-learning compared to those from value iteration. 
    What do you notice about the learned $Q$-values compared to those from value iteration? 
    Why do you think this occurs?

    \item Finally, compute an optimal policy $\pi^*_\text{QL}(x_t)$ based on the $Q$-learning approach from the first part, and another optimal policy $\pi^*_\text{VI}(x_t)$ based on the value iteration part. 
    Run the provided code to simulate each one over five years, and compute the cumulative profit $\sum_{k=0}^t r_k$ for each day $t$ and for each optimal policy. 
    Compare the cumulative profits over time. 
    What do you notice about the difference between the two cumulative profit trends? 
    Why do you think this occurs?
\end{enumerate}

\subsection*{\adjustbox{height=2ex, valign=c}{\includegraphics{figs/code.png}}\ Problem 2: Cart-pole Balancing via Model-free Reinforcement Learning}
In this problem, we will return to the classic ``cart-pole'' balancing control problem where our goal is to design a controller to balance an inverted pendulum upright on a cart.
We have already explored this problem in the context of model-based optimal control (specifically LQR control) in a \cref{ch:closedloop} exercise, but in this exercise we will approach the problem through model-free reinforcement learning.

To summarize the environment setup, the agent observes the state of the environment as $s \defn (x,\dot{x},\theta,\dot{\theta}) \in \R^4$, where $x\in\mathbb{R}$ denotes the horizontal position of the cart and $\theta\in\mathbb{R}$ denotes the angle of the pendulum from the upright position. 
At each instant, the agent chooses an action $u_t \in \{0, 1\}$ indicating whether to push the cart to the left $u_t = 0$ or to the right $u_t = 1$\sidenote{Note this is slightly different from the exercise in \cref{ch:closedloop} where the control was the horizontal force on the cart.}. 

This exercise is split up into several parts, and the starter code can be found in the notebook \colorcode{ch18/exercises/cartpole\_balance.ipynb}.
\begin{enumerate}
\item First, you will implement a deep Q-learning algorithm with experience replay, originally introduced in ``Playing Atari with Deep Reinforcement Learning"\cite{MnihEtAl2013}.
You can find the code for this part in \colorcode{q\_learning.py}:
\begin{enumerate}
    \item Implement the function \colorcode{QLearning.build\_network} to create a model for the Q-function that takes as input a state vector and outputs a vector of Q-values for that state and each action.
   	\item Implement the function \colorcode{QLearning.policy\_train} that will be used to compute actions during the training process.
   	Implement an $\epsilon$-greedy approach that samples a random action with probability $\epsilon$ to ensure exploration.
    \item Implement the functions \colorcode{QLearning.compute\_target} and \colorcode{QLearning.train} to sample episodes and train the model using experience replay, see \cref{alg:dqn} for the training algorithm.
    \item Use the provided code to start training with your choice of hyperparameters.
\end{enumerate}

\begin{algorithm}[ht!]
\caption{Deep Q-learning with Experience Replay}
\label{alg:dqn}
\DontPrintSemicolon
\KwData{Initial action-value model, $Q_\phi$, parameterized by $\phi$, number of episodes $M$}
\KwResult{Improved action-value model $Q_\phi$}
$\mathcal{D} \leftarrow \{\}$ \tcc*[r]{Initialize replay buffer}
\For{$\text{episode} \leftarrow 1$ \KwTo $M$}{
    Initialize $\x_0$. \\
    \For{each step $t$ of episode}{
    		\If{$\mathrm{rand()} < \epsilon$}{
			$\u_t \leftarrow \mathrm{random\_sample()}$
		}
		\Else{
			$\u_t \leftarrow \max_{\u} Q_\phi(\x_t, \u)$ \\
		}
        Execute $\u_t$, observe reward $r_t$ and next state $\x_{t+1}$. \\
        $\mathcal{D}.\mathrm{add}((\x_t, \u_t, r_t, \x_{t+1}))$ \\
        Sample random minibatch of transitions $(\x, \u, r, \x')$ from $\mathcal{D}$ \\
        $L \leftarrow 0$\\
        \For{sample $(\x, \u, r, \x')$ in minibatch}{
        		\If{$\x$ is terminal}{
        			$y \leftarrow r$
        		}
        		\Else{
        			$y \leftarrow r + \gamma\max_{\u'} Q_\phi(\x', \u')$
        		}
        		$L \leftarrow L + (y - Q_\phi(\x, \u))^2$\\
        }
        Update model parameters, $\phi$, using gradient descent on $L$. \\
    }
}
\Return $Q_\phi$
\end{algorithm}

\item Next, you will implement the REINFORCE algorithm introduced in \cref{subsubsec:policy_optimization_methods}.
Specifically, in this exercise you will implement three variations of the REINFORCE algorithm with slightly different definitions of the policy gradient, one of which is the standard version from \cref{alg:reinforce}.
You can find the code for this part in \colorcode{reinforce.py}:
\begin{enumerate}
\item Implement the function \colorcode{Reinforce.build\_network} to create a model for the policy that takes as input a state vector and outputs a vector of action probabilities.
   	\item Implement the function \colorcode{Reinforce.policy\_train} that will be used to compute actions during the training process, as well as the $\log \pi_{\btheta}(\u_t \given \x_t)$ value for the chosen action.
    \item Implement the function \colorcode{Reinforce.train} to sample episodes and train the model using the outline in \cref{alg:reinforce}.
    Here, you will implement options for three different policy gradient definitions (i.e. $\nabla_{\btheta} V(\btheta)$).
    First, implement the standard REINFORCE algorithm where the objective gradient $\nabla_{\btheta} V(\btheta)$ is estimated using:
\begin{equation*}
    \nabla_{\btheta} V(\btheta) \approx \sum_{t=0}^{T-1} \nabla_{\btheta} \log \pi_{\btheta}(\u_t \given \x_t) R(\tau),
\end{equation*}
based on an episode sampled from the environment and $R(\tau) = \sum_{t=0}^{T-1} \gamma^t r_t$ is the total discounted reward of the episode.

Next, implement the policy gradient with a ``causality trick''.
This trick avoids reinforcing actions taken later in the episode based on rewards obtained early in the episode, since those later actions would have no causality in receiving earlier rewards.
Specifically, with this trick you will compute the gradient estimate as:
\begin{equation*}
    \nabla_{\btheta} V(\btheta) \approx \sum_{t=0}^{T-1} \nabla_{\btheta} \log \pi_{\btheta}(\u_t \given \x_t) R_t(\tau),
\end{equation*}
where the term $R_t = \sum_{t’=t}^{T-1} \gamma^{t'-t} r_{t’}$ is the discounted \emph{reward-to-go} from time $t$.

Finally, you will implement a policy gradient that uses the ``causality trick'' with a ``baseline''.
The baseline helps to reduce the variance of the policy gradients by ``centering'' their returns.
Specifically, compute the policy gradient as:
\begin{equation*}
    \nabla_{\btheta} V(\btheta) \approx \sum_{t=0}^{T-1} \nabla_{\btheta} \log \pi_{\btheta}(\u_t \given \x_t) (R_t(\tau) - b(\tau)),
\end{equation*}
where the baseline $b(\tau)$ is the average of the $R_t(\tau)$ reward-to-go values over the episode.
You can additionally scale the baseline by the inverse of the standard deviation of the reward-to-go values.
\item Use the provided code to start training with your choice of hyperparameters.
What differences do you observe between the deep Q-learning algorithm and the REINFORCE algorithm in terms of performance during and after training?
Which method is more sample efficient, and why?
How do the different versions of the policy gradient compare within the REINFORCE method?
\end{enumerate} 
\end{enumerate}

\subsection*{\adjustbox{height=2ex, valign=c}{\includegraphics{figs/code.png}}\ Problem 3: Advantage Actor Critic}
As discussed in \cref{subsubsec:policy_optimization_methods}, actor-critic methods are a popular variance reduction technique for policy optimization.
These methods use a value function as a baseline (the ``critic'') and the learned policy is the ``actor''.
In this problem, you will implement key parts of the Advantage Actor-Critic (A2C) algorithm, which we described in \cref{alg:a2c}.

In the file \colorcode{ch18/exercises/advantage\_actor\_critic.ipynb}, complete the following tasks:
\begin{enumerate}
\item Using the model components already provided, write the code that implements the complete model, consisting of a couple linear layers to transform the input before being passed to heads for the policy and the value function (i.e. for the actor and critic).
For the actor, the model outputs mean and standard deviation parameters for a multivariate normal distribution.
\item For this implementation of A2C, we will use a Monte Carlo estimate to produce value targets for the critic, denoted by $y_t$ in \cref{alg:a2c}.
To compute the Monte Carlo value target, implement the function to compute the discounted returns:
\begin{equation*}
G_t = \sum_{k=t}^{T-1} \gamma^{k-t} r_k,
\end{equation*}
for an episode.
What is the advantage of using Monte Carlo estimates over temporal-difference estimates in terms of the bias-variance tradeoff?
\item Then, implement the function to compute the training loss for an episode:
\begin{equation*}
J = \sum_{t=0}^{T-1} \nabla_{\btheta} \log \pi_{\btheta}(\u_t \given \x_t) A_{\phi}^{\pi}(\x_t, \u_t) + ( A_{\phi}^{\pi}(\x_t, \u_t))^2,
\end{equation*}
where $A_{\phi}^{\pi}(\x_t, \u_t) = G_t - V_{\phi}^{\pi}(\x_t)$ is the advantage and where the first term is the actor loss and the second (quadratic) term is the critic loss.
Note that we are using JAX to compute gradients of the training loss with respect to the model parameters.
When computing the training loss for the actor component, its important to use \colorcode{jax.lax.stop\_gradient} to stop JAX from backpropagating gradients in this term.
Why is this important?
\item Finally, run the provided code to train the model for a toy lunar lander environment.
\end{enumerate}
\newpage
\printbibliography[segment=\therefsegment,heading=subbibliography,title={References}]
\chapter{Imitation Learning}
\label{ch:imitation-learning}
\newrefsegment
In \cref{ch:finite-state-machines}, we introduced a strategy for autonomous robot decision making that requires a very manual and sometimes intractable process of specifying desired actions from every possible state.
Then, in \cref{ch:sequential-decision-making} and \cref{ch:reinforcement-learning}, we formulated the sequential decision making problem as an optimization problem where we must specify a cost or reward function that we want the robot to minimize or maximize.
This optimization-based approach is more general and scalable, but it still requires us to figure out how to appropriately embed our preferences into the form of a mathematical function.
Reward design can be very challenging in practice, and by the nature of optimization-based approaches, the cost or reward function can be inadvertently exploited in undesirable ways.
Additionally, in the reinforcement learning context, we require continuous and exploratory interactions with the environment that could be costly or unsafe\sidenote{For example, some robots operate in close collaboration with humans or in other safety-critical environments where the risk of exploring sub-optimal actions is significant.} as well as sophisticated learning algorithms that are able to learn from experience.

In practice, it can sometimes be easier, more efficient, or safer for human experts to \emph{demonstrate} the desired task or behavior than it is to precisely program it, try to encode it in a cost function, or let the robot freely interact with the environment.
The goal of \emph{imitation learning} in the context of robotics is to leverage a limited set of expert demonstrations to accelerate or completely train a robot to autonomously perform a desired behavior. 
In this chapter, we begin in \cref{sec:il-robotics} by introducing the concept of imitation learning in the context of robotics, provide a formal problem formulation, and discuss key design considerations.
We then present a canonical imitation learning approach known as \emph{Behavior Cloning} in \cref{sec:il-bc}, which aims to directly learn a policy from expert demonstrations.
Lastly, in \cref{sec:il-irl}, we introduce \emph{Inverse Reinforcement Learning}, an alternative approach to imitation learning that learns a reward function from expert demonstrations.

\subsection{Imitation Learning in Robotics}
\label{sec:il-robotics}
Imitation learning is a class of methods that enable skills to be transferred from an expert to a learner. 
In the context of robotics, the expert is typically a human operator or a pre-existing control policy, and the learner is the robot that aims to mimic the expert's behavior. 
While the literature on imitation learning is vast, in this section, we focus on core design decisions and concepts essential for understanding and applying imitation learning to robotic systems.
Specifically, when designing an imitation learning system, several key aspects must be considered:

\paragraph{Is imitation learning the right approach?} 
Imitation learning might not always be the most suitable method for learning a task. 
For example, reinforcement learning might be a more effective approach if it is inexpensive for us to obtain samples from the environment\sidenote{Such as if we have a good \emph{simulator} for the task because it could be very safe and cheap to collect data.}. 
Moreover, if the expert's behavior is suboptimal or inconsistent, imitation learning may not yield the desired performance. 
Therefore, it is crucial to evaluate whether imitation learning is the most appropriate method for the given task.

\paragraph{What should we learn to imitate?} 
Expert demonstrations often contain a substantial amount of information that is irrelevant to the task at hand. 
For example, not all sensor measurements or control signals observed in a demonstration are necessary for successful task execution. 
Consequently, a central challenge is to identify and extract the aspects of the expert's behavior that are truly relevant and should be imitated by the learner.

\paragraph{Who is the expert?} 
The choice of expert can significantly impact the quality of the learned behavior. 
In many cases, the expert is a human operator who demonstrates the task. 
However, the expert could also be a pre-existing control policy, a set of historical data, or a mixture of multiple experts. 
Understanding which expert to learn from is crucial for the success of the imitation learning process.

\paragraph{How should we represent the policy?} 
The choice of policy representation can greatly influence the learning process. 
For instance, expert behavior can equivalently be represented at different levels of abstraction, such as low-level motor commands, high-level symbolic actions, or trajectory-level demonstrations. 
Moreover, different functional forms of the policy, such as whether the policy is defined as a linear function or a neural network, can impact the expressiveness and generalization capabilities of the learned policy.

\paragraph{What learning algorithm is most suitable?} 
The choice of learning algorithm can significantly impact the efficiency and performance of the imitation learning process. 
Many algorithms have been proposed for imitation learning, each with its own strengths and limitations. 
Understanding the characteristics of different algorithms and their suitability for the given task is essential for designing an effective imitation learning system.

\subsubsection{Differences Among Imitation Learning, Supervised Learning, and Reinforcement Learning}
Imitation learning is often compared with supervised learning and reinforcement learning, as all three paradigms involve learning from data.
While these methods share similarities, they also differ in several key aspects.

Supervised learning aims to learn a mapping from input data to output labels\sidenote{For example, from camera image inputs to object category outputs.} based on a dataset of input-output pairs.
While the imitation learning task of deriving a policy from a dataset of expert demonstrations is closely related to supervised learning, there are several key differences.
First, in imitation learning, the solution may have inherent structural properties, such as physical constraints or temporal dependencies\sidenote{For example, in robot planning and control we often have actuation limits.}, that are not present in standard supervised learning tasks.
Second, in a traditional supervised learning setting, we assume that the source domain, which includes the dataset used for training, and the target domain, which includes the test data, are the same.
In imitation learning, we may not be able to directly transfer the expert's behavior to the learner's environment.
For example, the embodiment of the expert may differ from the learner, such as if the expert is a human and the learner is a robot, leading to expert demonstrations of actions that are not directly executable by the robot.
Imitation learning is also typically exposed to the \emph{covariate shift} problem, where the distribution of the expert's data may differ from the distribution of the learner's data.
Specifically, the learner may encounter situations not represented in the expert's demonstrations, requiring it to generalize beyond the expert's behavior\sidenote{Strategies for addressing the issues arising from covariate shift will be discussed in more depth later in this chapter.}.
Lastly, obtaining expert demonstrations can be costly or time-consuming, making data collection a significant concern.

Imitation learning is also closely related to reinforcement learning, as both paradigms involve learning a policy from data that maximizes a reward.
However, reinforcement learning methods typically require a predefined reward function to guide the robot's behavior. 
In contrast, imitation learning assumes that the expert directly provides optimal, or at least good, behavior, bypassing the need for a reward function.

\subsubsection{Problem Formulation}
In imitation learning problems, we typically assume that we have access to a dataset, $\mathcal{D}$, of expert demonstrations.
The dataset generally consists of a set of trajectories and contexts, and we denote it mathematically as $\mathcal{D} = \{(\tau_i, s_i)\}_{i=1}^N$ where $N$ is the number of samples and $\tau_i = \{\x_{i,0}, \u_{i,0}, \ldots, \x_{i,T}\}$ is a trajectory executed by the expert in a given context, $s_i$. 
The context $s_i$ may represent a task description, an environmental configuration, or any other relevant information characterizing the expert's behavior. 
Alternatively, the dataset may consist of state-action pairs, where we would write $\mathcal{D} = \{(\x_i, \u_i)\}_{i=1}^N$.

Given such a dataset, we can broadly identify two main strategies for reproducing the expert's behavior. 
A first approach is to directly learn a mapping from contexts to trajectories, or from states to actions, using supervised learning techniques. 
That is, to learn:
$$
\pi(s) = \tau \quad \text{or} \quad \pi(\x) = \u.
$$
This approach is commonly referred to as \textit{Behavior Cloning} (BC)\cite{OsaEtAl2018}.

Alternatively, we can use the expert demonstrations to learn a reward function, $R(\x, \u)$, that implicitly defines the expert's behavior, and then infer a policy that maximizes this reward:
\begin{equation*}
\pi^*(\x) = \argmax_{\pi} V^\pi = \argmax_{\pi} \expected{\tau \sim p_{\pi}(\tau)}{\sum_{t=0}^{T-1} \gamma^t R(\x_t, \u_t)},
\end{equation*}
where $V^\pi$ is the expected sum of future rewards for policy $\pi$ where the expectation is over possible trajectories $\tau$ that are distributed according to $p(\tau)$.
This approach is known as \emph{Inverse Reinforcement Learning} (IRL)\cite{AroraEtAl2021} or \emph{Inverse Optimal Control} (IOC).

Behavior cloning and inverse reinforcement learning are the two primary approaches to imitation learning, each with distinct strengths and limitations.
In the following sections, we discuss these approaches in more detail and provide insights into when each method is most appropriate.

\subsection{Behavior Cloning}
\label{sec:il-bc}
Behavior cloning is an approach to imitation learning that focuses on directly learning a mapping from states (or contexts) to actions (or trajectories) without explicitly modeling the reward function.
The behavior cloning task can be formulated as a supervised learning problem, where the policy $\pi$ is learned by solving a regression problem.
We outline the general procedure for behavior cloning in \cref{alg:bc}.
\begin{algorithm}[ht]
\caption{Behavior Cloning}
\label{alg:bc}
Collect a dataset, $\mathcal{D}$, of expert demonstrations. \\
Define a model architecture for the policy, $\pi_{\btheta}$. \\
Define a loss function, $\mathcal{L}$. \\
Optimize the loss function, $\mathcal{L}$, with respect to the model parameters, $\btheta$. \\
\Return Trained policy, $\pi_{\btheta}$. 
\end{algorithm}

The first step consists of collecting a dataset $\mathcal{D}$ of expert demonstrations, for example from logged data generated by a human operator.
Next, a model architecture for the policy $\pi_{\btheta}$ is specified, which may take the form of a neural network, a linear model, or another function class described by parameters $\btheta$.
The choice of model architecture depends on the complexity of the task and the amount of available data, since the model must be expressive enough to capture the expert's behavior, yet not excessively complex to avoid overfitting.
Then, a loss function $\mathcal{L}$ is defined to quantify the discrepancy between the actions predicted by the policy and those demonstrated by the expert. 
Common choices include mean squared error, $\ell_1$ loss, hinge loss, and Kullback–Leibler divergence. 
Finally, the policy parameters $\boldsymbol{\theta}$ are optimized by minimizing the loss function $\mathcal{L}$ over the demonstration dataset.

Behavior cloning methods are an attractive approach to learning-based decision making, primarily due to their simplicity, effectiveness, and broad applicability.
However, ensuring the learned policy performs reliably in real-world settings presents significant challenges. 
One of the primary obstacles to trustworthy deployment of policies learned through behavior cloning is the issue of \emph{covariate shift}.

\subsubsection{The Covariate Shift Problem}
Formally, covariate shift refers to a mismatch between the distribution of data encountered during training and the distribution observed at deployment. 
In the context of behavior cloning, this issue arises when the learned policy is executed in the environment, causing small prediction errors to \textit{accumulate} over time. 
As these errors compound, the learner is increasingly likely to visit states that were rarely or never encountered in the expert's demonstrations, thereby drifting into poorly represented regions of the state space. 
As a result, the agent is forced to make decisions in unfamiliar situations, leading to poor performance and potentially catastrophic failures.

While it is impractical to gather data covering all possible states a robot might encounter, several strategies have been developed to mitigate the impact of covariate shift.
These strategies typically follow an iterative process that alternates between updating the robot's policy and targeted data collection based on the robot's current state distribution.
In this section, we outline two primary approaches to address covariate shift: \emph{confidence-based methods} and \emph{data aggregation methods}.

\paragraph{Confidence-Based Methods.}
In the class of confidence-based methods\cite{ChernovaEtAl2009}, the agent is endowed with a mechanism for estimating uncertainty in its predictions. 
This uncertainty estimate is used to identify situations in which the learned policy is likely to make errors, thereby enabling targeted corrective interventions. 
A common strategy exploits the uncertainty measure to detect regions of the state space where the agent's decisions are unreliable, after which additional data is collected in those regions to improve policy performance. 
In some cases, this data collection is also prompted by expert intervention, where the expert temporarily takes control to correct the agent's actions.

At a high level, methods based on this iterative refinement process seek to empirically align the training data distribution with the state distribution induced by the learned policy, thereby mitigating the effects of covariate shift. 
A schematic overview of this approach is presented in Algorithm~\ref{alg:cbm}.

\begin{algorithm}[ht]
\caption{Confidence-based Methods}
\label{alg:cbm}
\KwData{Dataset of expert demonstrations, $\mathcal{D}$, confidence estimation function, $c(\x)$, confidence threshold, $c_0$}
\KwResult{Trained policy, $\pi_{\btheta}$}
Train a policy, $\pi_{\btheta}$, on the dataset, $\mathcal{D}$. \\
\While{true}{
	Observe the state, $\x_t$. \\
	Compute the confidence estimate, $c(\x_t)$. \\
	\If{$c(\x_t) < c_0$ \textbf{or} expert intervention is necessary}{
		Compute additional demonstration data, $(\x_t, \u_t^{\text{expert}})$. \\
		$\mathcal{D} \leftarrow \mathcal{D} \cup \{(\x_t, \u_t^{\text{expert}})\}$. \\
	}
	Train the policy, $\pi_{\btheta}$, on the updated dataset, $\mathcal{D}$. \\
}
\Return Trained policy, $\pi_{\btheta}$.
\end{algorithm}

\paragraph{Data Aggregation Methods.}
Data aggregation methods constitute another major class of approaches within behavior cloning.
A prominent example is $\dagname$\cite{RossEtAl2011}, which mitigates covariate shift by explicitly collecting expert demonstrations under the state distribution induced by the learner's own policy. 

As outlined in \cref{alg:dagger}, $\dagname$ follows an iterative two-step procedure. 
First, the agent is allowed to interact with the environment, thereby generating states according to its current policy and induced state distribution. 
Second, these visited states are relabeled with expert actions, and the resulting data is aggregated into the training set to refine the policy.
\begin{algorithm}[ht]
\caption{$\dagname$ Algorithm}
\label{alg:dagger}
\KwData{Initial dataset of expert demonstrations, $\mathcal{D}$, initial policy, $\pi_{\btheta}^1$, number of iterations, $N$}
\KwResult{Trained policy, $\pi_{\btheta}^N$}
\For{$i = 1, 2, \ldots, N$}{
	Collect trajectories, $\tau = \{(\x_t, \u_t^{\text{robot}})\}$, using the policy $\pi_{\btheta}^i$. \\
	Gather dataset of states visited by the robot and actions given by the expert, $\mathcal{D}^i = \{(\x_t, \u_t^{\text{expert}})\}$. \\
	Aggregate the dataset, $\mathcal{D} \leftarrow \mathcal{D} \cup \mathcal{D}^i$. \\ 
	Train the policy, $\pi_{\btheta}^{i+1}$, on the updated dataset, $\mathcal{D}$. \\
}
\Return Trained policy, $\pi_{\btheta}^N$.
\end{algorithm}

In its simplest form, $\dagname$ begins by initializing the policy, $\pi_{\btheta}^1$, using a set of previously collected expert demonstrations.
The robot then interacts with the environment with policy $\pi_{\btheta}^1$, collecting trajectories $\tau$ that reflect the state distribution under the current policy.
These trajectories are subsequently relabeled using the expert’s actions for the visited states.
The relabeled trajectories are used to train an updated policy $\pi_{\theta}^{2}$, which is then employed to collect additional trajectories under the state distribution induced by the updated policy.
This process is repeated for a fixed number of iterations, resulting in the final trained policy $\pi_{\theta}^N$.

By collecting expert demonstrations under the learner's state distribution, $\dagname$ effectively reduces covariate shift and enhances the performance of the learned policy. 
The method can be viewed as a form of interactive supervised learning, in which the agent actively gathers data to refine its performance.
This iterative process minimizes the amount of expert data required and has proven highly effective across a wide range of tasks.

\medskip
In summary, confidence-based methods and data aggregation techniques both provide solutions for addressing the covariate shift problem in behavior cloning.
While there are many variations of these methods, the core principles we outline in \cref{alg:cbm} and \cref{alg:dagger} provide a foundational understanding of how to mitigate covariate shift through targeted data collection.
However, methods following these principles still suffer from other common limitations of behavior cloning, such as dependence on the quality of expert demonstrations.
In \cref{sec:rl-vs-sl}, we discuss approaches that leverage ideas from behavior cloning to learn from broader, and potentially suboptimal, sets of expert demonstrations.

\subsubsection{Reinforcement Learning via Supervised Learning (RvS)}
\label{sec:rl-vs-sl}
Recent work\cite{EmmonsEtAl2021} has explored the idea of converting the reinforcement learning problem, which we discussed in \cref{ch:reinforcement-learning}, into a \emph{conditional}, \emph{filtered}, or \emph{weighted} imitation learning problem.
These approaches are motivated by a simple but powerful insight: rather than relying exclusively on optimal demonstrations, one can leverage a much broader collection of demonstrations generated by suboptimal policies or gathered across diverse---yet related---tasks.
Methods in this class are often referred to as \textit{reinforcement learning via supervised learning} (RvS).
These approaches typically operate by conditioning the policy on goals or desired reward levels, and may additionally incorporate mechanisms for reweighting or filtering demonstrations

\paragraph{Filtering or Weighting Demonstrations.}
One common approach to RvS is to filter or weight the expert demonstrations based on their quality or relevance to the task.
For example, one might assign higher weights to expert demonstrations that obtain higher rewards.
Revisiting the outline of the behavior cloning algorithm in \cref{alg:bc}, this idea can be instantiated by modifying the dataset to retain only high-quality demonstrations, as measured by reward information.

A simple instantiation of this approach proceeds as follows. 
First, expert demonstrations are ranked according to their \emph{return}\sidenote{The \emph{return} of a trajectory is the sum of rewards accumulated along the trajectory. When ranking trajectories, it is generally desirable to consider long-term performance rather than immediate rewards.}:
\begin{equation*}
	r(\tau) = \sum_{t=0}^{T-1} \gamma^t R(\x_t, \u_t).
\end{equation*}
Then, the original dataset $\mathcal{D}$ is filtered to retain only the top $k\%$ of trajectories based on their return:
\begin{equation*}
	\tilde{\mathcal{D}} = \{ \tau \in \mathcal{D} \given r(\tau) \geq \bar{r} \}.
\end{equation*}
where $\bar{r}$ denotes the return threshold such that $k\%$ of the trajectories in $\mathcal{D}$ achieve a return greater than or equal to this value.
Finally, the policy $\pi_{\btheta}$ is trained using the filtered dataset $\tilde{\mathcal{D}}$.

This process represents a simple instance of RvS, in which expert trajectories are filtered according to their return. 
More sophisticated variants instead operate at the level of individual transitions, rather than filtering entire trajectories. 
In these cases, the quality of individual actions can be assessed using their \emph{advantage}\sidenote{Recall from \cref{ch:reinforcement-learning} that the advantage of an action is defined as the difference between the action-value function and the value function, $A(x_t, u_t) = Q(x_t, u_t) - V(x_t)$. Intuitively, the advantage quantifies how much better an action is relative to the average action in a given state.} or, equivalently, their \emph{Q-value}, instead of relying solely on immediate rewards.

Once action weights have been computed, they can be incorporated into the behavior cloning procedure by modifying the loss function in \cref{alg:bc} as:
\begin{equation*}
	\mathcal{L}(\btheta) = \expected{(\x, \u) \sim \tilde{\mathcal{D}}}{-\log \pi_{\btheta}(\u \given \x) A(\x, \u)}.
\end{equation*}
This objective can be interpreted as a weighted version of the standard behavior cloning loss, where the log-likelihood of each action is scaled by its corresponding advantage. 
This method, commonly referred to as \emph{advantage-weighted behavior cloning}, has been shown to improve the empirical performance of behavior cloning by upweighting higher-quality actions during training.

\paragraph{Goal or Reward Conditioning.}
Another common approach to RvS is to condition the policy on a goal or reward value.
This approach is particularly useful in settings where the expert demonstrations are suboptimal or collected from a different task.
Consider a dataset of previously collected trajectories, $\mathcal{D} = \{\tau_i\}$.
Each trajectory $\tau_i$ might be described using different outcomes\sidenote{In other words, a condition that is verified during or at the end of the trajectory.}, such as the final state of the trajectory, the total reward obtained, or a specific state visited during the trajectory.
Let $\omega$ denote a specific outcome occurring in a trajectory $\tau$.
The goal of conditioning-based RvS is to learn an outcome-conditioned policy, $\pi_{\btheta}(\u \given \x, \omega)$, that optimizes:
\begin{equation*}
	\mathcal{L}(\btheta) = \expected{(\x, \u, \omega) \sim \mathcal{D}}{-\log \pi_{\btheta}(\u \given \x, \omega)}.
\end{equation*}

Among the various conditioning strategies, \emph{goal-conditioned} and \emph{state-conditioned} RvS are particularly relevant. 
In these approaches, the policy is conditioned on a desired outcome, typically specified as a target state or goal $\omega = x \in \statespace$ that the agent is expected to reach. 
For instance, in a robotic manipulation task, the policy may be conditioned on achieving a specific end-effector configuration.
Another widely used conditioning strategy is \emph{reward-conditioned} RvS, in which the policy is conditioned on a target reward value, $\omega = \sum_{t=0}^{T-1} R(x_t, u_t)$. 
In this setting, the conditioning variable encodes the desired level of performance, allowing the policy to adapt its behavior accordingly.

In both cases, conditioning on outcomes enables the robot to extract meaningful information from suboptimal or diverse expert demonstrations, often leading to improved performance in practice.
For example, consider two policies, $\pi_{\btheta_1}(\u \given \x)$ and $\pi_{\btheta_2}(\u \given \x, \omega)$, that are trained on the same dataset $\mathcal{D}$.
Suppose $\pi_{\btheta_1}$ is trained to imitate expert demonstrations that implicitly optimize a specific reward function. 
This reward-centric approach restricts $\pi_{\btheta_1}$ to behaviors that closely follow the expert's trajectory distribution.
On the other hand, $\pi_{\btheta_2}$ is goal-conditioned and trained to achieve any specified goal state, $\omega$, independent of the underlying reward function. 
By explicitly incorporating the goal into its policy, $\pi_{\btheta_2}$ decouples the process of achieving desired outcomes from the reward structure.
As a result, $\pi_{\btheta_2}$ is likely to generalize better to novel tasks or unseen goal states, as it learns a flexible mapping from states and goals to actions. 
In contrast, $\pi_{\btheta_1}$ remains constrained by the expert's reward-aligned demonstrations, making it less adaptable to scenarios with divergent or ambiguous reward structures.

\subsection{Inverse Reinforcement Learning}
\label{sec:il-irl}
In the previous section, we discussed behavior cloning as a form of imitation learning that directly learns a policy from expert demonstrations.
Inverse reinforcement learning\cite{NgRussell2000} takes an orthogonal approach to imitation learning by attempting to recover a reward function from a policy, or from demonstrations of a policy.
In certain cases, identifying the reward function can offer deeper insights into the task's underlying structure, making it potentially more informative than directly learning a policy. 
Additionally, a policy that is optimal for the expert may not be optimal for the agent if they have different dynamics, morphologies, or capabilities\sidenote{Learned reward representations can also potentially generalize across different robot platforms that tackle similar problems.}.

\begin{example}[Inverse Reinforcement Learning vs Behavior Cloning] 
\label{ex:apprentice}
Consider a scenario where the robot's objective is to drive across a city as quickly as possible. 
In the context of imitation learning, we assume the reward function is unknown, but an expert provides example routes to navigate the city. 
Behavior cloning approaches attempt to replicate the expert's actions, such as by learning to turn right at a particular intersection. 
This strategy lacks robustness since it can fail when the robot encounters intersections that the expert never visited.
Inverse reinforcement learning approaches offer a more generalizable alternative by focusing on identifying key features of the expert's trajectories, rather than just mimicking actions. 
For example, instead of merely copying the expert's turns, the robot could learn to recognize useful patterns, such as preferring roads with higher speed limits or fewer stop signs. 
The robot can then develop a policy that takes routes with similar advantageous characteristics, even if they differ from the exact paths the expert took.
\end{example}

Formally, the goal of inverse reinforcement learning is to recover a reward function, $R: \statespace \times \controlspace \rightarrow \R$, from a set of expert demonstrations, $\mathcal{D} = \{\tau_i\}$, where $\tau_i = \{(\x_0, \u_0, \ldots, \x_T)\}$ is an example trajectory.
The recovered reward function can then be used to train a policy that is optimal with respect to this reward. 
In particular, given a parametric representation of the reward function with parameters $w$, inverse reinforcement learning seeks a parameter configuration that best explains the observed expert behavior.

However, the inverse reinforcement learning problem is inherently ill-posed: multiple reward functions may induce the same optimal policy, making the reward unidentifiable from demonstrations alone. 
To mitigate this ambiguity, a variety of alternative objectives have been proposed in the literature, including maximum-margin formulations that separate optimal and suboptimal policies\cite{RatliffBagnellEtAl2006}, as well as maximum-entropy approaches that prefer the least-committal reward function consistent with the demonstrations\cite{ZiebartMaasEtAl2008}.

In practice, most inverse reinforcement learning algorithms follow an iterative optimization procedure involving two coupled steps. 
First, the reward function parameters $w$ are updated according to an algorithm-specific objective. 
Second, the policy parameters $\btheta$ are adjusted to maximize the expected return under the current reward estimate. 
These steps are repeated until convergence. 
Although individual inverse reinforcement learning methods differ in how these updates are performed, Algorithm~\ref{alg:irl} provides a high-level schematic overview of the inverse reinforcement learning process.

\begin{algorithm}[ht]
\caption{High-level IRL Algorithm}
\label{alg:irl}
\KwData{Expert demonstrations, $\mathcal{D}$, initialized reward function parameters, $\w$, initialized policy parameters, $\btheta$}
\KwResult{Learned reward function parameters, $\w$, learned policy parameters, $\btheta$}
\While{not converged}{
	Update the reward function parameters, $\w$. \\
	Update the policy parameters, $\btheta$, to maximize the current estimate of the reward function. \\
}
\Return Optimized reward and policy parameters: $\w$, $\btheta$.
\end{algorithm}

In the following sections, we first introduce the concept of \emph{feature expectation} and then discuss three popular inverse reinforcement learning methods: apprenticeship learning\nocite{AbbeelNg2004}, maximum margin planning\nocite{RatliffBagnellEtAl2006}, and maximum entropy IRL\nocite{ZiebartMaasEtAl2008}.

\subsubsection{Feature Expectation}
\label{sec:il-fe}
We begin by assuming the existence of a true reward function $R^*$ that can be expressed as a linear combination of features:
\begin{equation*}
	R^*(x, u) = {\boldsymbol{w}^*}^\top \phi(x, u),
\end{equation*}
where $\phi : \statespace \times \controlspace \rightarrow [0,1]^d$ denotes a vector of feature functions\sidenote{To ensure that rewards are bounded by 1, we assume $\|\boldsymbol{w}^*\|_2 \leq 1$.}.
In \cref{ch:reinforcement-learning}, we saw that the value function for a policy $\pi$ is defined as the expected cumulative discounted reward as:
\begin{equation}
\label{eq:irl-value-function}
V_T^\pi(\x) = \expected{\tau \sim p_{\pi}(\tau)}{\sum_{t=0}^{T-1} \gamma^tR(\x_t, \pi(\x_t)) \given \x_0 = \x}.
\end{equation}
Substituting the linear reward model $R(x, u) = \boldsymbol{w}^\top \phi(x, u)$ into the expression above yields:
\begin{equation}
\label{eq:irl-value-function-fe}
V_T^\pi(\x) = \w^\top \mu(\pi, \x),
\end{equation}
where:
\begin{equation*}
\mu(\pi, \x) = \expected{\tau \sim p_{\pi}(\tau)}{\sum_{t=0}^{T-1} \gamma^t \phi(\x_t, \pi(\x_t)) \given \x_0 = \x}.
\end{equation*}
The quantity $\mu(\pi, x)$ is referred to as the \emph{feature expectation} of policy $\pi$\sidenote{For brevity, we may also denote $\mu(\pi, x)$ simply as $\mu(\pi)$ when the dependence on the initial state is clear or omitted.}.

An important insight is that, by definition, the optimal expert policy $\pi^*$ will always yield a value function greater than or equal to that of any other policy and therefore:
\begin{equation*}
    V_T^{\pi^*}(\x) \geq V_T^\pi(\x), \quad \forall \x \in \statespace, \quad \forall \pi.
\end{equation*}
Using the feature-expectation formulation of the value function in \cref{eq:irl-value-function-fe}, this condition can be equivalently expressed as:
\begin{equation} 
\label{eq:irlcondition}
 {\w^*}^\top  \mu(\pi^*, \x) \geq  {\w^*}^\top  \mu(\pi, \x), \quad \forall \x \in \statespace, \quad \forall \pi.
\end{equation}
In principle, one could attempt to recover the expert's reward vector $\w^*$ by finding a vector $\w$ that satisfies the inequality in \cref{eq:irlcondition}. 
However, this formulation is inherently ambiguous. 
For example, the trivial choice $\w = \boldsymbol{0}$ satisfies the inequality for all policies without conveying any meaningful preference. 
More generally, multiple reward functions may induce the same optimal policy, a phenomenon known as \emph{reward ambiguity}. 
This issue lies at the core of inverse reinforcement learning\cite{NgRussell2000}, and the algorithms discussed below introduce additional structure or optimization criteria to resolve this ambiguity.

\subsubsection{Apprenticeship Learning}
\label{sec:il-al}
The apprenticeship learning algorithm\cite{AbbeelNg2004} addresses the problem of reward ambiguity by finding a policy $\pi$ such that the feature expectation induced by $\pi$ is close to that of the expert policy $\pi^*$.
Mathematically, the goal of apprenticeship learning is to find a policy such that $\lVert \mu(\pi, \x) - \mu(\pi^*, \x) \rVert_2 \leq \epsilon$ for all $\x \in \statespace$, where $\epsilon$ is a small positive constant.
For such a policy $\pi$, we would have that for any $\w$ with $\lVert \w \rVert_2 \leq 1$:
\begin{equation}
\begin{split}
\lvert V_T^\pi(\x) - V_T^{\pi^*}(\x) \rvert &= \lvert\w^\top  \mu(\pi, \x) - \w^\top  \mu(\pi^*, \x)\rvert, \\
&\leq \lVert \w \rVert_2 \lVert \mu(\pi, \x) - \mu(\pi^*, \x) \rVert_2 \\
&\leq 1 \cdot \epsilon = \epsilon,
\label{eq:al}
\end{split}
\end{equation}
where the first equality follows from the definition of the value function as a function of the feature expectation in \cref{eq:irl-value-function-fe}, the first inequality follows from the fact that $\lvert x^\top y \rvert \leq \lVert x \rVert_2 \lVert y \rVert_2$ for any vectors $x$ and $y$, and the second inequality follows from the assumption that $\lVert \w \rVert_2 \leq 1$.
This result motivates a practical reformulation of the inverse reinforcement learning problem: rather than attempting to recover the true reward parameters $\w^*$, it suffices to learn a policy whose feature expectations match those of the expert within a small tolerance. 
Under this condition, the learned policy is guaranteed to achieve performance comparable to that of the expert, even when the recovered reward function differs from the true underlying reward.

Within this framework, the inverse reinforcement learning problem reduces to finding a policy $\pi$ whose induced feature expectation $\mu(\pi)$ closely matches the one of the expert policy, $\mu(\pi^*)$. 
A schematic overview of the apprenticeship learning algorithm is provided in \cref{alg:apprentice}.
\begin{algorithm}[ht]
\KwData{Expert's feature expectations, $\mu^* = \mu(\pi^*)$, initial policy, $\pi_0$}
\KwResult{Learned parameters, $\w$, and policy, $\hat{\pi}^*$}
$i \leftarrow 1$ \\
\While{true}{
  Compute $\mu^{(i-1)} = \mu(\pi_{i-1})$ (or approximate via Monte Carlo methods). \\
  Compute $t^{(i)} = \max_{w:\lVert w \rVert_2 \leq 1} \min_{j \in \{0, \ldots, (i-1)\}} \w^\top \left(\mu^* - \mu^{(j)}\right)$ by solving:
  \begin{equation} 
  \label{eq:irlcomputew}
    \begin{split}
    (\w_i, t_i) \leftarrow \maximize[\w, t] & t,\\
    \subjectto & \w^\top \mu^* \geq \w^\top \mu^{(j)} + t, \quad \forall j \in  \{0, \dots, (i-1)\},\\
    & \lVert \w \rVert_2 \leq 1.
    \end{split}
  \end{equation}
  \If{$t_i \leq \epsilon$}{
    $\hat{\pi}^* \xleftarrow{}$ best feature matching policy from $\{ \pi_0, \dots, \pi_{i-1} \}$\\
    $\hat \w^* \xleftarrow{} \w_i$ \\
    \Return $\hat{\pi}^*$, $\hat \w^*$
  }
  Compute an optimal policy, $\pi_i$, for the reward function defined by $\hat \w$. \\
  $i \leftarrow i + 1$ \\
 }
\caption{Apprenticeship Learning}
\label{alg:apprentice}
\end{algorithm}

At iteration $i$ of \cref{alg:apprentice}, we have already identified a set of policies $\pi_0, \pi_1, \ldots, \pi_{i-1}$ along with their corresponding feature expectations $\mu^{(0)}, \mu^{(1)}, \ldots, \mu^{(i-1)}$. Within the inner loop, we solve the optimization problem defined in \cref{eq:irlcomputew} to estimate a reward parameter vector $\boldsymbol{w}$ that explains the expert’s behavior.
Specifically, using the feature-expectation formulation of the value function in \cref{eq:irl-value-function-fe}, the constraint:
$$
\w^\top \mu^* \geq \w^\top \mu^{(j)} + t,
$$
can be rewritten as:
$$
V_T^{\pi^*}(\x_0) \geq V_T^{\pi_j}(\x_0) + t,
$$
meaning that, under the recovered reward function, the expert outperforms each previously learned policy by a margin of at least $t$.
Given the resulting reward parameters $\w_i$, we then compute a new policy $\pi_i$\sidenote{For example, using reinforcement learning methods introduced in \cref{ch:reinforcement-learning}.} that is optimal with respect to this reward. 
This iterative procedure continues until the margin $t_i$ falls below a predefined threshold $\epsilon$, indicating that the feature expectations of the learned policy sufficiently match those of the expert.

\subsubsection{Maximum Margin Planning}
Maximum margin planning (MMP)\cite{RatliffBagnellEtAl2006} is a generalization of apprenticeship learning that aims to find a reward function that maximally separates the expert policy from a set of policies.
Specifically, MMP modifies \cref{eq:irlcomputew} from the apprenticeship learning algorithm as follows:
\begin{equation} 
\label{eq:mmp}
\begin{split}
\hat{\w}^* = \underset{\w, \xi}{\arg\min} \quad & \lVert \w \rVert_2^2 + C\xi,\\
\subjectto & \w^\top \mu^* \geq \w^\top \mu^{(j)} + m(\pi^*, \pi^{(j)}) - \xi, \quad \forall j \in  \{0, \dots, (i-1)\},
\end{split}
\end{equation}
where $m(\pi, \pi')$ is a distance function\sidenote{For example, $m(\pi, \pi')$ could measure the number of states in which $\pi$ and $\pi'$ select different actions.} between two policies $\pi$ and $\pi'$, $\xi$ is a slack variable that allows for violations of the margin constraints, and $C$ is a hyperparameter that controls the penalty for such violations.
Intuitively, this formulation enforces larger margins for policies that are more dissimilar from the expert policy $\pi^*$.

An advantage of the MMP formulation over the apprenticeship learning approach in \cref{eq:irlcomputew} arises when the expert is suboptimal. 
In such cases, it may be impossible to find a reward vector $\w$ that makes the expert policy outperform all other policies.
As a result, the apprenticeship learning optimization problem may yield trivial solutions such as $\w_i = \boldsymbol{0}$ and $t_i = 0$.
By contrast, the MMP formulation introduces slack variables that relax the margin constraints, enabling the computation of a nontrivial and informative reward vector even when the expert demonstrations are imperfect. 
This added flexibility allows MMP to remain effective in realistic scenarios where expert behavior may be noisy or suboptimal.

\subsubsection{Maximum Entropy Inverse Reinforcement Learning}
As we described in \cref{sec:il-irl}, the inverse reinforcement learning problem is inherently ill-posed since there are infinitely many reward functions that could explain the expert's behavior.
While maximum margin approaches are highly effective when there is a single reward function that is clearly better than alternatives, in some cases, optimizing for a distribution of reward functions is more appropriate.
Maximum entropy inverse reinforcement learning (MaxEnt IRL)\cite{ZiebartMaasEtAl2008} aims to find a distribution over reward functions that explains the expert's behavior, that matches the feature expectations of the expert\sidenote{Similar to maximum margin methods.}, and has maximum entropy\sidenote{That is, remaining as non-committal or ``random'' as possible beyond the imposed constraints.}.

Let us denote the distribution over trajectories induced by a policy $\pi$ as $p_{\pi}(\tau)$. The feature expectations can be rewritten in terms of this distribution as:
\begin{equation*}
\mu(\pi) = \expected{\pi}{f(\tau)} = \int p_{\pi}(\tau)f(\tau) \d\tau,
\end{equation*}
where $f(\tau) = \sum_{t=0}^{T-1} \gamma^t \phi(\x_t, \pi(\x_t))$.
Within this framework, MaxEnt IRL seeks a trajectory distribution that maximizes entropy:
\begin{equation}
\mathcal{H}(p(\tau)) = \int -p(\tau) \log p(\tau) \d\tau,
\end{equation}
subject to the constraints:
\begin{equation}
\begin{split}
  \int p(\tau)f(\tau) d\tau & = \int p_{\pi^*}(\tau)f(\tau) \d\tau, \\
  \int p(\tau) \d\tau = 1,
\end{split}
\end{equation}
where the first constraint enforces that the feature expectations of the learned behavior match those of the expert policy, while the second ensures that $p(\tau)$ is a valid probability distribution.

Among the distributions that satisfy the constraint:
\begin{equation*}
\int p(\tau)f(\tau) d\tau = \int p_{\pi^*}(\tau)f(\tau) \d\tau,
\end{equation*}
the maximum entropy distribution follows the exponential form:
\begin{equation*}
p(\tau) \propto \exp(\w^\top f(\tau)).
\end{equation*}
Specifically, we can express the trajectory distribution as a function of $\w$ as:
\begin{equation}
p(\tau \given \w) = \frac{1}{Z(\w)}\exp\left(\w^\top  f(\tau)\right),
\label{eq:maxent-traj-dist}
\end{equation}
where $Z(\w)$ is the partition function given by $Z(\w) = \int \exp\left(\w^\top  f(\tau)\right) \d\tau$.

However, \cref{eq:maxent-traj-dist} only holds for deterministic environments where the next state is fully determined by the current state and action. 
In stochastic environments, the trajectory distribution is also influenced by the random environment dynamics, and in this case we express the distribution over trajectories as:
\begin{equation}
p(\tau \given \w) = \frac{1}{Z(\w)}\exp\left(\w^\top  f(\tau)\right) \prod_{t=0}^{T-1} p(\x_{t+1} \given \x_t, \u_t).
\label{eq:maxent-dist}
\end{equation}
We can therefore obtain the parameter vector $\w$ for the reward function by maximizing the likelihood of the observed data under the maximum entropy distribution defined in \cref{eq:maxent-dist} as:
\begin{equation}
  \hat \w^* = \arg\max_{\w} \mathcal{L}_{\mathrm{MLE}} = \arg\max_{\w} \sum_{\tau \in \mathcal{D}} \ln p(\tau \given \w).
\end{equation}

\section{Summary}
In this chapter, we introduced imitation learning, a paradigm for learning control policies by demonstration. 
We began by formally defining the imitation learning problem, discussing key design considerations, and highlighting its relationship with, and key differences from, supervised and reinforcement learning. 
Central to this formulation, we identified two primary strategies: directly learning a policy and inferring an underlying reward function.

The first strategy, behavior cloning, was presented as a straightforward approach that treats imitation as a supervised learning problem, mapping expert states to actions. 
We discussed its primary limitation, the covariate shift problem, where small errors accumulate and lead the robot to unfamiliar states. 
To address this, we explored interactive methods like $\dagname$, which mitigate covariate shift by collecting new demonstrations in states visited by the learner's policy. 
We also introduced reinforcement learning via supervised learning, an extension that leverages suboptimal or diverse demonstration data by filtering, weighting, or conditioning the policy on outcomes like rewards or goals.

The second strategy, inverse reinforcement learning, was introduced as an alternative that seeks to recover the expert's underlying reward function from demonstrations. 
We highlighted that this approach can lead to more generalizable and robust policies, as it captures the expert's intent rather than their exact actions. 
We discussed the core concept of feature expectation matching and explored several prominent inverse reinforcement learning algorithms, including Apprenticeship Learning, Maximum Margin Planning, and Maximum Entropy IRL, each offering a different method to resolve the inherent ambiguity in recovering a reward function.

\paragraph{To learn more.}
For a comprehensive survey of imitation learning in robotics, including behavior cloning and its variants, we refer the reader to \citet{OsaEtAl2018}. 
The seminal paper on the $\dagname$ algorithm is presented by \citet{RossEtAl2011}. 
For an introduction to inverse reinforcement learning, the foundational work by \citet{NgRussell2000} is essential. 
A modern survey on inverse reinforcement learning methods can be found in \citet{AroraEtAl2021}. 
Specific influential algorithms, such as Apprenticeship Learning and Maximum Entropy IRL, are detailed in \citet{AbbeelNg2004} and \citet{ZiebartMaasEtAl2008}, respectively.
\newpage
\printbibliography[segment=\therefsegment,heading=subbibliography,title={References}]

\backmatter

\chapter{Prospects}
\notessection{A Journey through the landscape of Robot Autonomy}
At the outset of this book, we asked a deceptively simple question: \emph{What is robot autonomy?} 
We described a space between two extremes, where on one end we have the fantastical general-purpose android of science fiction, and on the other end we have the simple, pre-programmed machine that is capable of only a single, narrowly defined task.
In that middle ground lives the modern autonomous robot, a system that must perceive, reason, and act in a world that is complex, dynamic, and uncertain.

In this book, we have explored the core principles that underpin the design and operation of modern autonomous robots.
Specifically, we anchored our discussion in the \emph{See-Think-Act} cycle, which provides a structured way to understand the flow of information and decision-making in an autonomous system.
By now, that abstraction should feel less like a theoretical construct and more like a technical blueprint.
Let us briefly revisit each of these stages and reflect on the key insights that emerged from our exploration:

\paragraph{See.}
We emphasized how the \emph{See} stage is not merely about the passive acquisition of data, but about the active process of extracting meaningful information from the environment. 
Much of this discussion appeared in Part II, where we began by examining the fundamental properties and limitations of common sensing modalities, both proprioceptive and exteroceptive, and by establishing the physical and geometric principles that govern their measurements.
Building on this foundation, we explored how raw sensor data can be transformed into structured and informative representations.
We first surveyed classical approaches, including filtering techniques, feature detection, and methods for extracting geometric information about the environment.
We then turned to modern, learning-based techniques, examining how deep neural network architectures enable end-to-end learning of rich representations directly from raw sensory inputs, including images as well as point-based and voxel-based representations of 3D geometry.
Finally, we discussed higher-level visual understanding tasks such as object detection and segmentation, in which algorithms must interpret scenes by identifying object instances and reasoning about their spatial relationships.

Taken together, the \emph{See} phase is not simply about sensing, but about perception, where the robot must transform raw sensory signals into the rich, semantic representations that are necessary for it to think and act intelligently.

\paragraph{Think.}
Once the \emph{See} stage has produced a structured and semantically meaningful representation of the world, the \emph{Think} stage is where that representation is transformed into actionable intent. 
This is the stage in which the robot reasons about where it is with respect to its environment, its goals, and how its actions will shape future outcomes.
In Part III, we discussed how this process typically begins with robot localization and mapping, where local sensor measurements are synthesized into a coherent global estimate of the robot's state and a map of the environment.
Building on this foundation, Part~IV introduced techniques for high-level decision-making.
Specifically, we discussed approaches to \emph{sequential decision-making}, where the robot determines, loosely speaking, \emph{what} to do, ultimately defining the objectives that guide subsequent stages of planning and control.
To address this problem, we explored methods rooted in optimal control, such as dynamic programming, and extended this perspective to modern learning-based approaches, such as reinforcement learning and imitation learning.
Across these frameworks, we emphasized different but complementary strategies for reasoning under uncertainty, balancing model-based formulations with data-driven approximations.

Viewed holistically, the \emph{Think} phase emerges as a structured process that transforms information about the environment and abstract goals into concrete plans and intentions that can be executed by the robot.

\paragraph{Act.}
The final stage of the cycle, \emph{Act}, is where high-level decisions are translated into physical motion. 
While the \emph{Think} stage determines \emph{what} the robot should do, the \emph{Act} stage determines \emph{how} those decisions are realized by a dynamical system subject to physical constraints.
In Part~I, we developed the tools required to bridge this gap.
We began with modeling the robot as a dynamical system, introducing the representations and notation needed to describe its motion and constraints.
Building on this foundation, we discussed how desired behaviors are converted into feasible trajectories through motion planning and trajectory optimization.
Motion planning algorithms produce geometrically feasible, collision-free paths, while trajectory optimization refines these paths into time-parameterized motions that respect dynamics, actuator limits, and task requirements.
Execution, however, is never merely the blind following of precomputed motor commands.
The real world is inherently uncertain, subject to external disturbances, sensor noise, actuator limitations, and inevitable inaccuracies in the robot's internal dynamic models.
For this reason, execution must be grounded in the principle of \emph{feedback control}. 
Rather than assuming that a planned trajectory will unfold exactly as predicted, feedback control continuously compares the robot's measured state against its desired state and corrects deviations in real time.
Our discussion spanned a spectrum of closed-loop control strategies, from classical PID to optimization-based methods, such as Linear Quadratic Regulators (LQR) and Model Predictive Control (MPC), which explicitly reason about system dynamics, performance objectives, and constraints.

Viewed in the context of the full See-Think-Act cycle, the \emph{Act} phase closes the loop between intention and reality. 
It is where plans meet physical embodiment, and where the robot's ability to adapt and respond to the unpredictable nature of the real world is put to the test.

\notessection{From Modular Pipelines to End-to-End Autonomy.}
Throughout this book, we have emphasized that the stages of \emph{See}, \emph{Think}, and \emph{Act} are not isolated modules, but deeply interconnected processes with information flowing asynchronously at different rates, and with multiple feedback loops. 
At the same time, the field is undergoing a noticeable shift from carefully engineered, modular autonomy stacks toward increasingly \emph{end-to-end} frameworks. 
In classical architectures, perception, state estimation, planning, and control are designed and tuned as distinct components, each with explicit interfaces and well-defined responsibilities. 
In contrast, modern data-driven systems often blur or even collapse these boundaries, and can now solve tasks that were once considered beyond the reach of traditional, non-learning-based systems.

This shift toward end-to-end learning has been enabled by the availability of large-scale datasets, increasingly realistic simulation environments, and unprecedented computational resources.
End-to-end learning promises greater adaptability, reduced manual engineering, and the ability to capture complex couplings that are difficult to model analytically. 
Yet, this evolution does not make the material in this book any less relevant, as the principles and techniques of sensing, estimation, planning, and control continue to underpin the design and analysis of these end-to-end systems.
For instance, modular components often serve as strong baselines against which end-to-end approaches are compared. 
Classical algorithms frequently reappear as safety layers, fallback strategies, or constraint-enforcing mechanisms wrapped around learned policies, ensuring that the system maintains a degree of interpretability, reliability, and safety.
Moreover, this data-centric paradigm introduces new system-level challenges. 
The design of datasets, labeling pipelines, simulation environments, evaluation metrics, and benchmarking protocols becomes even more critical, and often draws heavily on traditional techniques from perception, state estimation, and control.
In this sense, the traditional See-Think-Act abstraction continues to provide not only a conceptual framework, but also a powerful lens through which to analyze, debug, and responsibly deploy the increasingly end-to-end systems that are shaping the future of robot autonomy.

\notessection{The Road Ahead}
As we write this book, we find ourselves at a remarkable inflection point in robotics research, where advances in computation, sensing, machine learning, and large-scale data are rapidly reshaping how autonomous systems are designed, developed, and operated.
In this final section, we will briefly reflect on some of the most exciting emerging trends in robotics research, and how they are shaping the future of robot autonomy.

\paragraph{Foundation Models (FMs).}
Over the past few years, artificial intelligence has undergone a dramatic shift driven by large-scale \emph{foundation models}.
Citing the seminal work of \citet{BommasaniEtAl2021FoundationModels}, we define foundation models as:
\begin{displayquote}
    \emph{``[Foundation models are] models that are trained on broad data (generally using self-supervision at scale) that can be adapted (e.g., fine-tuned) to a wide range of downstream tasks.''}
\end{displayquote}
Importantly, the term ``foundation model'' is not meant to refer to a specific architecture.
Rather, it describes a class of models characterized by scale, broad pretraining, and adaptability.
Well-known examples include large language models (LLMs)~\cite{RadfordEtAl2019GPT2}~\cite{BrownEtAl2020GPT3}, which underpin systems such as ChatGPT, as well as vision-language models (VLMs)~\cite{RadfordEtAl2021CLIP}, video generation models~\cite{HoEtAl2022ImagenVideo}, and other large-scale architectures~\cite{RivesEtAl2021ESM}. 
Beyond their impact on purely digital applications, foundation models are increasingly being integrated into physical robotic systems.

Their growing relevance in robotics stems from several key properties.
First, foundation models provide \emph{broad priors about the world}. 
Trained on large and diverse datasets, they encode a wealth of information about the structure of language, visual concepts, and physical interactions, enabling improved generalization in settings with limited task-specific data. 

Second, foundation models are inherently \emph{multimodal}. 
Many modern architectures jointly process text, images, video, audio, and structured data. 
Robotics, by its very nature, is a multimodal domain where robots must integrate information from diverse sensory modalities such as vision, proprioception, tactile sensing, and language-based instructions.
Multimodal foundation models offer a unified representational framework in which these heterogeneous data streams can be fused and reasoned over coherently.

Third, foundation models provide a powerful and flexible \emph{interface for human-robot interaction}. 
Language-conditioned policies enable users to specify goals in natural language rather than low-level commands. 
This dramatically lowers the barrier between human intent and robotic execution, enabling more intuitive forms of supervision, correction, and collaboration.

Within this broader paradigm, several emerging instantiations of foundation models are particularly influential in robotics:

\medskip
\noindent\textbf{Vision–language–action (VLA) and reasoning models.} 
VLA models~\cite{ZitkovichEtAl2023RT2}~\cite{BlackEtAl2024Pi0}~\cite{KimEtAl2025OpenVLA} represent an emerging class of architectures that directly couple vision and language representations to action outputs.
In contrast to traditional pipelines where perception, planning, and control are engineered as separate modules, VLA models attempt to learn end-to-end mappings from multimodal inputs to motor commands or high-level actions, essentially learning the See-Think-Act cycle as a single, integrated process.
Typically pretrained on large-scale vision and language data and subsequently fine-tuned on robotic interaction data, these models aim to combine broad semantic understanding with embodied control.

A closely related development is the emergence of \emph{reasoning models} within embodied systems~\cite{HuangEtAl2023InnerMonologue}~\cite{NVIDIAEtAl2025AlpamayoR1}~\cite{ZawalskiEtAl2024ECOT}.
Beyond mapping observations directly to actions, these architectures allocate explicit computation to intermediate deliberation, generating structured reasoning traces—often in natural language—that attempt to make causal relationships, counterfactuals, and task constraints explicit before committing to specific actions.
Inspired by advances in large language models where inference-time reasoning improves robustness and accuracy, reasoning-enabled robotic systems treat deliberation itself as a tunable resource, where more complex or safety-critical situations can trigger deeper chains of thought, self-reflection, or verification steps.
In embodied contexts, however, reasoning cannot remain purely textual.
It must be grounded in geometry, dynamics, and physical feasibility, and must remain consistent with the actions ultimately executed.
The central challenge, therefore, is not merely to produce articulate explanations, but to ensure alignment between internal rationale and external behavior, so that reasoning becomes a functional component of autonomy rather than a post-hoc narrative. 

\medskip
\noindent\textbf{World models.}
Another especially important instantiation of the foundation model paradigm for robotics is the development of \emph{world models}. 
Broadly speaking, world models aim to capture how the world evolves over time, including its physical, spatial, and visual dynamics~\cite{HaSchmidhuber2018WorldModels}~\cite{NVIDIAEtAl2025Cosmos}. 
Unlike traditional dynamics models, which operate on carefully designed state representations, modern world models are learned directly from rich, high-dimensional inputs such as images, video, depth, and other multimodal signals.
This allows them to model complex aspects of real-world environments—such as object persistence, scene geometry, contact interactions, and temporal visual change—that are difficult to specify analytically but are essential for robust autonomy.

This richer predictive interface is particularly valuable in robotics, where data is a fundamental bottleneck.
Robot interaction data is expensive to collect, embodiment-specific, and limited in scale. 
In contrast, visual data---especially video---is abundant and diverse.
World models provide a pathway for leveraging this disparity by learning general physical and semantic structure from large-scale, largely passive observations. 
In doing so, they enable a partial decoupling between \emph{world understanding} and \emph{control}, allowing robots to acquire broad knowledge about how environments behave without requiring equivalent amounts of task-specific interaction data.

At a high level, world models equip robots with the ability to \emph{predict and imagine}.
Given a current situation and, potentially, a proposed action sequence, a world model can forecast how the environment and the robot within it may evolve.
This enables planning through the evaluation of imagined futures, where a robot may compare alternative outcomes, select actions that lead toward desirable states, or optimize behavior by reasoning directly over predicted trajectories.

A second major advantage is that world models can support the generation of realistic synthetic data. 
Because they model visual, spatial, and physical regularities, world models can be used to synthesize new scenarios, including rare edge cases or safety-critical situations that are difficult to collect in practice.
In this way, world models can augment traditional simulation, providing a data-driven substrate for training and evaluating autonomous systems in a broader range of conditions, while maintaining high visual fidelity.

The precise form that such world models should take remains an open research question. 
Some approaches focus on predicting high-dimensional sensory observations such as future video frames conditioned on actions~\cite{DeepMind2025Genie3}. 
Others learn compact latent representations of the environment and predict future states within this lower-dimensional space rather than at the raw pixel level~\cite{AssranEtAl2025VJEPA2}. 
Additional work explores structured representations that encode objects, physical interactions, or causal relationships in an effort to improve generalization and data efficiency~\cite{LocatelloEtAl2020SlotAttention}. 

\medskip
\noindent Taken together, while the integration of foundation models into robotics remains an active and rapidly evolving area of research, recent developments suggest that they may fundamentally reshape how autonomous systems are designed, developed, and operated. 
Rather than building task-specific autonomy stacks from scratch, future robots may increasingly rely on large pretrained models as adaptable cores, as well as key enablers across the entire autonomy development pipeline.

In this sense, foundation models are not merely new building blocks within the autonomy stack, but are beginning to redefine the abstractions and workflows used to construct it.

\paragraph{Simulation and Closing the Gap to Reality.}
In parallel, progress in simulation has accelerated dramatically, expanding both its fidelity and its role within the autonomy development pipeline.
Advances in physics-based engines, high-fidelity rendering, differentiable simulators, and large-scale synthetic data generation now enable training and evaluation in increasingly realistic environments~\cite{MakoviychukEtAl2021IsaacGym}. 
At the same time, the emergence of generative simulation has begun to complement traditional simulators by producing realistic, diverse, and often open-ended scenarios directly from data. 

Simulation has long been an essential tool in robotics, but it is increasingly becoming a central substrate for large-scale training, evaluation, and red-teaming of autonomous systems. 

\paragraph{Physical AI Safety.}
As autonomy stacks incorporate more data-driven components, the question of safety becomes increasingly critical.
Data-driven models can hallucinate, extrapolate poorly outside their training distribution, or produce internally inconsistent reasoning traces. 
In physical systems, such failures carry real-world consequences.

Physical AI safety is therefore becoming a central topic. 
This includes research on detecting and mitigating failure modes of data-driven models, developing robust training and evaluation protocols, and combining data-driven learning with traditional control-theoretic safety guarantees~\cite{SinhaElhafsiEtAl2024}~\cite{SagawaEtAl2020DRO}~\cite{HoffmanEtAl2018CyCADA}~\cite{TaylorEtAl2020LearningCBF}.
At the systems level, emerging frameworks such as NVIDIA's \emph{Halos} platform aim to provide end-to-end safety validation pipelines for autonomous systems, integrating simulation, scenario generation, and formal evaluation to stress-test models under diverse and safety-critical conditions~\cite{NVIDIAHalos2024}. 
Such efforts highlight the growing recognition that safety must be addressed holistically, spanning model design, data curation, evaluation, and deployment.

\notessection{An Open Frontier}
One thing is clear: it is an exciting time to be working in robotics.
Autonomous systems that were once confined to research laboratories are increasingly deployed in real-world environments and embedded within critical infrastructure. 
From autonomous vehicles and aerial systems to space robotics and medical platforms, the impact of robotics is expanding rapidly.

With this book, our aim has been twofold. 
First, to provide the reader with the mathematical and algorithmic foundations that underpin modern robot autonomy, so that they can understand the core principles and techniques that enable robots to perceive, plan, and act in complex environments.
Second, to cultivate a systems-level perspective, so that the reader may understand not only individual algorithms, but how they interconnect and compose into complete autonomy stacks.
The future of robotics will not be built by isolated techniques, but by thoughtful integration. 
It will be shaped by engineers and researchers who understand both theory and systems, both abstraction and embodiment. 
The tools are evolving rapidly, but the underlying principles will remain enduring guides.
The next chapter of robot autonomy will be written by those who are willing to build, and we hope this book has given the reader the foundation to do just that.





\printbibliography[heading=bibintoc,title={References}]

@book{SiegwartNourbakhshEtAl2011,
    title={Introduction to Autonomous Mobile Robots},
    author={Siegwart, R. and Nourbakhsh, I. R. and Scaramuzza, D.},
    year={2011},
    publisher={MIT Press},
}

@book{Joseph2018,
    Author = {Joseph, L.},
    Title = {Robot Operating System (ROS) for Absolute Beginners: Robotics Programming Made Easy},
    Publisher = {Apress},
    Year = {2018},
}

@book{gelb1974applied,
  title={Applied optimal estimation},
  author={Gelb, A. and others},
  year={1974},
  publisher={MIT press}
}

@book{QuigleyGerkeyEtAl2015,
    Author = {Quigley, M. and Gerkey, B. and Smart, W. D.},
    Title = {Programming Robots with {ROS}: A Practical Introduction to the Robot Operating System},
    Publisher = {O'Reilly Media},
    Year = {2015},
}

@book{Tassa2011,
    title={Theory and Implementation of Biomimetic Motor Controllers. PhD Thesis},
    author={Tassa, Y.},
    year={2011},
    publisher={The Hebrew University of Jerusalem},
}

@inproceedings{CarionEtAl2020,
author="Carion, Nicolas
and Massa, Francisco
and Synnaeve, Gabriel
and Usunier, Nicolas
and Kirillov, Alexander
and Zagoruyko, Sergey",
title="End-to-End Object Detection with Transformers",
booktitle="Computer Vision -- ECCV 2020",
year="2020",
publisher="Springer International Publishing",
pages="213--229",
}

@book{Kirk2004,
    Author = {Kirk, D. E.},
    Title = {Optimal Control Theory: An Introduction},
    Publisher = {Dover Publications},
    Year = {2004},
}

@book{Murray2009,
	title = {Optimization-{Based} {Control}},
	author = {Murray, R. M.},
    year = {2009},
    publisher = {California Institute of Technology},
}

@book{AstromMurray2009,
	title = {Feedback Systems},
	author = {Aström, K. J. and Murray, R. M.},
    year = {2009},
    publisher = {Princeton University Press},
}

@book{Levine2009,
    Author = {Levine, J.},
    Title = {Analysis and Control of Nonlinear Systems: A Flatness-based Approach},
    Publisher = {Springer},
    Year = {2009},
}

@book{LaValle2006,
    author = {LaValle, S. M.},
    title = {Planning Algorithms},
    publisher = {Cambridge University Press},
    address = {Cambridge, U.K.},
    year = {2006}
}

@book{Latombe1991,
    author = {Latombe, J. C.},
    title = {Robot Motion Planning},
    publisher = {Kluwer Academic Publishers},
    address = {USA},
    year = {1991}
}

@article{KavrakiSvestkaEtAl1996,
    author = {{Kavraki}, L. E. and {Svestka}, P. and {Latombe}, J.-C. and {Overmars}, M. H.},
    journal = {IEEE Transactions on Robotics and Automation}, 
    title = {Probabilistic roadmaps for path planning in high-dimensional configuration spaces}, 
    year = {1996},
    volume = {12},
    number = {4},
    pages = {566--580},
}

@misc{LaValle1998,
    author = {LaValle, S. M.},
    title = {Rapidly-Exploring Random Trees: A New Tool for Path Planning},
    institution = {},
    year = {1998},
}

@article{KaramanFrazzoli2011,
    author = {Karaman, S. and Frazzoli, E.},
    title = {Sampling-based Algorithms for Optimal Motion Planning},
    journal = {Int. Journal of Robotics Research},
    volume = {30},
    number = {7},
    pages = {846--894},
    year = {2011},
}

@article{JansonSchmerlingEtAl2015,
    author = {Janson, L. and Schmerling, E. and Clark, A. and Pavone, M.},
    title = {{Fast} {Marching} {Tree:} A Fast Marching Sampling-Based Method for Optimal Motion Planning in Many Dimensions},
    journal = {{Int. Journal of Robotics Research}},
    volume = {34},
    number = {7},
    pages = {883--921},
    year = {2015},
}

@article{JansonIchterEtAl2018,
    author = {Janson, L. and Schmerling, E. and Clark, A. and Pavone, M.},
    title = {Deterministic sampling-based motion planning: Optimality, complexity, and performance},
    journal = {{Int. Journal of Robotics Research}},
    volume = {37},
    number = {1},
    pages = {46-61},
    year = {2018},
}

@article{Sethian1996,
    author = {Sethian, J. A.},
    title = {A fast marching level set method for monotonically advancing fronts},
    journal = {Proceedings of the National Academy of Sciences},
    volume = {93},
    number = {4},
    pages = {1591-1595},
    year = {1996},
}

@inbook{DudekJenkin2008,
    author = {Dudek, G. and Jenkin, M.},
    title = {Inertial Sensors, GPS, and Odometry},
    booktitle = {Springer Handbook of Robotics},
    year = {2008},
    publisher = {Springer},
    pages = {477--490},
}

@incollection{HartleyZisserman2002,
    title = {Camera Models},
    author = {Hartley, R. and Zisserman, A.},
    booktitle = {Multiple View Geometry in Computer Vision},
    year = {2002},
    publisher = {Academic Press}
}

@book{maybeck1982stochastic,
  title={Stochastic models, estimation, and control},
  author={Maybeck, P. S.},
  volume={3},
  year={1982},
  publisher={Academic press}
}

@article{cadena2017past,
  title={Past, present, and future of simultaneous localization and mapping: Toward the robust-perception age},
  author={Cadena, C. and Carlone, L. and Carrillo, H. and Latif, Y. and Scaramuzza, D. and Neira, J. and Reid, I. and Leonard, J. J.},
  journal={IEEE Transactions on robotics},
  volume={32},
  number={6},
  pages={1309--1332},
  year={2017},
  publisher={IEEE}
}

@article{Tsai1987,
    title = {A Versatile Camera Calibration Technique for High-accuracy 3D Machine Vision Metrology Using Off-the-shelf TV Cameras and Lenses},
    author = {Tsai, R.},
    journal = {IEEE Journal on Robotics and Automation},
    volume = {3},
    number = {4},
    pages = {323--344},
    year = {1987},
}

@article{Bradski2000,
    author = {Bradski, G.},
    journal = {Dr. Dobb's Journal of Software Tools},
    title = {{The OpenCV Library}},
    year = {2000}
}

@article{Zhang2000,
    title = {A Flexible New Technique for Camera Calibration},
    author = {Zhang, Z.},
    journal = {IEEE Transactions on Pattern Analysis and Machine Intelligence},
    volume = {22},
    year = {2000}
}

@article{Fusiello2000,
    author = {Fusiello, A. and Trucco, E. and Verri, A.},
    title = {A compact algorithm for rectification of stereo pairs},
    journal = {Machine Vision and Applications},
    year = {2000},
    volume = {12},
    number = {1},
    pages = {16--22},
}

@inproceedings{LoopZhang1999,
    title = {Computing rectifying homographies for stereo vision},
    author = {Loop, C. and Zhang, Z.},
    booktitle = {IEEE Computer Society Conference on Computer Vision and Pattern Recognition},
    volume = {1},
    pages = {125--131},
    year = {1999},
}

@inproceedings{detone2018superpoint,
  title={Superpoint: Self-supervised interest point detection and description},
  author={DeTone, D. and Malisiewicz, T. and Rabinovich, A.},
  booktitle={Proceedings of the IEEE conference on computer vision and pattern recognition workshops},
  pages={224--236},
  year={2018}
}

@inproceedings{Harris1988,
    author = {Harris, C. and  Stephens, M.},
    booktitle = {4th Alvey Vision Conference},
    title = {A combined corner and edge detector},
    year = {1988},
}

@article{Lowe2004,
  title   = {Distinctive Image Features from Scale-Invariant Keypoints},
  author  = {Lowe, D. G.},
  journal = {International Journal of Computer Vision},
  volume  = {60},
  number  = {2},
  pages   = {91--110},
  year    = {2004},
  month   = {11},
  issn    = {1573-1405},
  doi     = {10.1023/B:VISI.0000029664.99615.94}
}

@inproceedings{Moravec1977,
    author = {Moravec, H. P.},
    booktitle = {5th International Joint Conference on Artificial Intelligence},
    title = {Towards automatic visual obstacle avoidance},
    year = {1977},
}

@article{PerveenKumarEtAl2013,
    title = {An overview on template matching methodologies and its applications},
    author = {Perveen, N. and Kumar, D. and Bhardwaj, I.},
    journal = {International Journal of Research in Computer and Communication Technology},
    volume = {2},
    number = {10},
    pages = {988--995},
    year = {2013}
}

@book{barshalom2001estimation,
  title={Estimation with applications to tracking and navigation: theory algorithms and software},
  author={Bar-Shalom, Y. and Li, X. R. and Kirubarajan, T.},
  year={2001},
  publisher={John Wiley \& Sons}
}

@article{olfatisaber2007consensus,
  author  = {Olfati-Saber, R. and Fax, J. A. and Murray, R. M.},
  title   = {Consensus and Cooperation in Networked Multi-Agent Systems},
  journal = {Proceedings of the IEEE},
  volume  = {95},
  number  = {1},
  pages   = {215--233},
  year    = {2007}
}

@article{rauch1965rts,
  title={Maximum likelihood estimates of linear dynamic systems},
  author={Rauch, H. E. and Tung, F. and Striebel, C. T.},
  journal={AIAA journal},
  volume={3},
  number={8},
  pages={1445--1450},
  year={1965}
}

@article{lakshminarayanan2017simple,
  title={Simple and scalable predictive uncertainty estimation using deep ensembles},
  author={Lakshminarayanan, B. and Pritzel, A. and Blundell, C.},
  journal={Advances in neural information processing systems},
  volume={30},
  year={2017}
}

@article{angelopoulos2023conformal,
  title={Conformal prediction: A gentle introduction},
  author={Angelopoulos, A. N. and Bates, S. and others},
  journal={Foundations and trends{\textregistered} in machine learning},
  volume={16},
  number={4},
  pages={494--591},
  year={2023},
  publisher={Now Publishers, Inc.}
}

@article{amini2020deep,
  title={Deep evidential regression},
  author={Amini, A. and Schwarting, W. and Soleimany, A. and Rus, D.},
  journal={Advances in neural information processing systems},
  volume={33},
  pages={14927--14937},
  year={2020}
}

@inproceedings{guo2017calibration,
  title={On calibration of modern neural networks},
  author={Guo, C. and Pleiss, G. and Sun, Y. and Weinberger, K. Q.},
  booktitle={International conference on machine learning},
  pages={1321--1330},
  year={2017},
  organization={PMLR}
}

@inproceedings{julier1997non,
  title={A non-divergent estimation algorithm in the presence of unknown correlations},
  author={Julier, S. J. and Uhlmann, J. K.},
  booktitle={Proceedings of the 1997 American Control Conference (Cat. No. 97CH36041)},
  volume={4},
  pages={2369--2373},
  year={1997},
  organization={IEEE}
}

@misc{alammar2018illustrated,
  author    = {Alammar, J.},
  title     = {The Illustrated Transformer},
  year      = {2018},
  month     = {6},
  howpublished = {\url{https://jalammar.github.io/illustrated-transformer/}},
  note      = {Blog post. Accessed: 2026-04-09}
}

@article{kang2018voxel,
author = {Kang, Z. and Yang, J. and Zhong, R. and Wu, Y. and Shi, Z. and Lindenbergh, R.},
year = {2018},
month = {11},
pages = {4287-4298},
title = {Voxel-Based Extraction and Classification of 3-D Pole-Like Objects From Mobile LiDAR Point Cloud Data},
volume = {11},
journal = {IEEE Journal of Selected Topics in Applied Earth Observations and Remote Sensing},
doi = {10.1109/JSTARS.2018.2869801}
}

@article{uijlings2013selective,
  title     = {Selective Search for Object Recognition},
  author    = {Uijlings, J. R. R. and van de Sande, K. E. A. and Gevers, T. and Smeulders, A. W. M.},
  journal   = {International Journal of Computer Vision},
  year      = {2013}
}

@article{tian2023occ3d,
  title={Occ3D: A Large-Scale 3D Occupancy Prediction Benchmark for Autonomous Driving},
  author={Tian, X. and Jiang, T. and Yun, L. and Wang, Y. and Wang, Y. and Zhao, H.},
  journal={arXiv preprint arXiv:2304.14365},
  year={2023}
}

@inproceedings{zhang2010tracking,
  title={Tracking with multisensor out-of-sequence measurements with residual biases},
  author={Zhang, S. and Bar-Shalom, Y. and Watson, G.},
  booktitle={2010 13th International Conference on Information Fusion},
  pages={1--8},
  year={2010},
  organization={IEEE}
}

@book{Szeliski2010,
    title = {Computer vision: algorithms and applications},
    author = {Szeliski, R.},
    year = {2010},
    publisher = {Springer Science \& Business Media}
}

@book{ThrunBurgardEtAl2005,
    title = {Probabilistic Robotics},
    author = {Thrun, S. and Burgard, W. and Fox, D.},
    year = {2005},
    publisher = {MIT Press}
}

@book{blackman1999tracking,
  author    = {Blackman, S. and Popoli, R.},
  title     = {Design and Analysis of Modern Tracking Systems},
  publisher = {Artech House},
  year      = {1999}
}

@book{Gustafsson2010,
    author = {Gustafsson, F.},
    publisher = {Studentlitteratur},
    pages = {554},
    title = {Statistical Sensor Fusion},
    year = {2013}
}

@inproceedings{LeonardDurrantWhyte1991,
  title={Simultaneous map building and localization for an autonomous mobile robot.},
  author={Leonard, J. J. and Durrant-Whyte, H. F.},
  booktitle={IROS},
  volume={3},
  pages={1442--1447},
  year={1991}
}

@incollection{SmithSelfEtAl1990,
  title={Estimating uncertain spatial relationships in robotics},
  author={Smith, R. and Self, M. and Cheeseman, P.},
  booktitle={Autonomous robot vehicles},
  pages={167--193},
  year={1990},
  publisher={Springer}
}

@book{Simon2006,
    title = {Optimal State Estimation: Kalman, $H_ \infty$, and Nonlinear Approaches},
    author = {Simon, D.},
    year = {2006},
    publisher = {John Wiley \& Sons}
}

@misc{KaelblingWhiteEtAl2011,
    author = {Kaelbling, L. and White, J. and Abelson, H. and Freeman, D. and Lozano-P\'{e}rez, T. and Chuang, I.},
    title = {{6.01SC: Introduction to Electrical Engineering and Computer Science I}},
    year = {2011},
    note = {MIT OpenCourseWare}
}

@book{Bertsekas2019,
    title = {Reinforcement learning and optimal control},
    author = {Bertsekas, D.},
    year = {2019},
    publisher = {Athena Scientific}
}

@article{bertsekas1988auction,
  title={The auction algorithm: A distributed relaxation method for the assignment problem},
  author={Bertsekas, D. P.},
  journal={Annals of operations research},
  volume={14},
  number={1},
  pages={105--123},
  year={1988},
  publisher={Springer}
}

@book{Sipser1996,
    title = {Introduction to the Theory of Computation},
    author = {Sipser, M.},
    year = {1996},
    publisher = {International Thomson Publishing},
}

@book{Bertsekas2000,
    title = {Dynamic Programming and Optimal Control},
    author = {Bertsekas, D.},
    year = {2000},
    publisher = {Athena Scientific}
}

@book{Bertsekas2016,
    title = {Nonlinear Programming},
    author = {Bertsekas, D.},
    year = {2016},
    publisher = {Athena Scientific}
}

@book{How2008,
    title = {Lecture Notes for Principles of Optimal Control},
    author = {How, J. P.},
    year = {2008}
}

@book{SuttonBarto2018,
    title = {Reinforcement learning: An introduction},
    author = {Sutton, R. and Barto, A.},
    year = {2018},
    publisher = {MIT Press}
}

@book{torralba2024foundations,
  author    = {Torralba, A. and Isola, P. and Freeman, W. T.},
  title     = {Foundations of Computer Vision},
  year      = {2024},
  publisher = {The MIT Press},
  address   = {Cambridge, MA},
  isbn      = {978-0-262-04897-2},
  note      = {Available under CC-BY-ND-NC license}
}

@book{Puterman2014,
  title = {Markov Decision Processes: Discrete Stochastic Dynamic Programming},
  author = {Puterman, M.},
  year = {2014},
  publisher = {Wiley}
}

@book{ForsythPonce2011,
    title = {Computer Vision: A Modern Approach},
    author = {Forsyth, D. A. and Ponce, J.},
    year = {2011},
    publisher = {Prentice Hall}
}

@book{GoodfellowBengioCourville2016,
    title={Deep Learning},
    author={Goodfellow, I. and Bengio, Y. and Courville, A.},
    publisher={MIT Press},
    note={\url{http://www.deeplearningbook.org}},
    year={2016}
}

@inproceedings{ZeilerFergus2014,
    author = {Zeiler, M. D. and Fergus, R.},
    title = {Visualizing and Understanding Convolutional Networks},
    booktitle = {European Conference on Computer Vision (ECCV)},
    year = {2014},
    publisher = {Springer},
    pages = {818--833},
}

@inproceedings{zhang2014loam,
  title={LOAM: Lidar odometry and mapping in real-time.},
  author={Zhang, J. and Singh, S. and others},
  booktitle={Robotics: Science and systems},
  volume={2},
  number={9},
  pages={1--9},
  year={2014},
  organization={Berkeley, CA}
}

@article{qin2018vins,
  title={Vins-mono: A robust and versatile monocular visual-inertial state estimator},
  author={Qin, T. and Li, P. and Shen, S.},
  journal={IEEE transactions on robotics},
  volume={34},
  number={4},
  pages={1004--1020},
  year={2018},
  publisher={IEEE}
}

@article{mur2015orb,
  title={ORB-SLAM: A versatile and accurate monocular SLAM system},
  author={Mur-Artal, R. and Montiel, J. M. M. and Tardos, J. D.},
  journal={IEEE transactions on robotics},
  volume={31},
  number={5},
  pages={1147--1163},
  year={2015},
  publisher={IEEE}
}

@inproceedings{klein2007parallel,
  title={Parallel tracking and mapping for small AR workspaces},
  author={Klein, G. and Murray, D.},
  booktitle={2007 6th IEEE and ACM international symposium on mixed and augmented reality},
  pages={225--234},
  year={2007},
  organization={IEEE}
}

@article{vo2006gaussian,
  title={The Gaussian mixture probability hypothesis density filter},
  author={Vo, B.-N. and Ma, W.-K.},
  journal={IEEE Transactions on signal processing},
  volume={54},
  number={11},
  pages={4091--4104},
  year={2006},
  publisher={IEEE}
}

@article{vo2014labeled,
  title={Labeled random finite sets and the Bayes multi-target tracking filter},
  author={Vo, B.-N. and Vo, B.-T. and Phung, D.},
  journal={IEEE Transactions on Signal Processing},
  volume={62},
  number={24},
  pages={6554--6567},
  year={2014},
  publisher={IEEE}
}

@book{mahler2007statistical,
  title={Statistical multisource-multitarget information fusion},
  author={Mahler, R.},
  year={2007},
  publisher={Artech}
}

@inproceedings{AbbeelNg2004,
    title = {Apprenticeship Learning via Inverse Reinforcement Learning},
    author = {Abbeel, P. and Ng, A.},
    booktitle = {Proceedings of the Twenty-First International Conference on Machine Learning},
    year = {2004},
}

@inproceedings{RatliffBagnellEtAl2006,
    title = {Maximum Margin Planning},
    author = {Ratliff, N. and Bagnell, J. A. and Zinkevich, M.},
    booktitle = {Proceedings of the 23rd International Conference on Machine Learning},
    year = {2006},
    pages = {729--736},
}

@inproceedings{NgRussell2000,
    title = {Algorithms for Inverse Reinforcement Learning},
    author = {Ng, A. and Russell, S.},
    booktitle = {Proceedings of the Seventeenth International Conference on Machine Learning},
    year = {2000},
    pages = {663--670},
}

@inproceedings{ZiebartMaasEtAl2008,
    title = {Maximum Entropy Inverse Reinforcement Learning},
    author = {Ziebart, B. D. and Maas, A. and Bagnell, J. A. and Dey, A. K.},
    booktitle = {Proceedings of the Twenty-Third AAAI Conference on Artificial Intelligence},
    year = {2008},
    pages = {1433--1438},
}

@article{kuhn1955hungarian,
  title={The Hungarian method for the assignment problem},
  author={Kuhn, H. W.},
  journal={Naval research logistics quarterly},
  volume={2},
  number={1-2},
  pages={83--97},
  year={1955},
  publisher={Wiley Online Library}
}

@article{Khatib1986,
    author = {Khatib, O.},
    title ={Real-Time Obstacle Avoidance for Manipulators and Mobile Robots},
    journal = {The International Journal of Robotics Research},
    year = {1986},
    volume = {5},
    number = {1},
    pages = {90-98},
}

@article{HarelEtAl1987,
    author = {Harel, D.},
    title ={Statecharts:  A visual formalism for complex systems},
    journal = {Science of Computer Programming},
    year = {1987},
    volume = {8},
    number = {3},
    pages = {231–274},
}

@book{Alur2015,
    title = {Algorithms for Decision Making},
    author = {Alur, R.},
    year = {2015},
    publisher = {MIT Press}
}

@inproceedings{julier1997new,
  title={New extension of the Kalman filter to nonlinear systems},
  author={Julier, S. J. and Uhlmann, J. K.},
  booktitle={Signal processing, sensor fusion, and target recognition VI},
  volume={3068},
  pages={182--193},
  year={1997},
  organization={Spie}
}

@article{dellaert2021factor,
  title={Factor graphs: Exploiting structure in robotics},
  author={Dellaert, F.},
  journal={Annual Review of Control, Robotics, and Autonomous Systems},
  volume={4},
  number={1},
  pages={141--166},
  year={2021},
  publisher={Annual Reviews}
}

@inproceedings{triggs2000bundle,
  title={Bundle adjustment—a modern synthesis},
  author={Triggs, B. and McLauchlan, P. F. and Hartley, R. I. and Fitzgibbon, A. W.},
  booktitle={Vision Algorithms: Theory and Practice: International Workshop on Vision Algorithms Corfu, Greece, September 21--22, 1999 Proceedings},
  pages={298--372},
  year={2000},
  organization={Springer}
}

@inproceedings{lowe1999object,
  title={Object recognition from local scale-invariant features},
  author={Lowe, D. G.},
  booktitle={Proceedings of the seventh IEEE international conference on computer vision},
  volume={2},
  pages={1150--1157},
  year={1999},
  organization={Ieee}
}

@article{SilverEtAl2016,
    author = {Silver, D. and Huang, A. and Maddison, C. J. and Guez, A. and Sifre, L. and van den Driessche, G. and Schrittwieser, J. and Antonoglou, I. and Panneershelvam, V. and Lanctot, M. and Dieleman, S. and Grewe, D. and Nham, J. and Kalchbrenner, N. and Sutskever, I. and Lillicrap, T. and Leach, M. and Kavukcuoglu, K. and Graepel, T. and Hassabis, D.},
    title = {Mastering the game of Go with deep neural networks and tree search},
    journal = {Nature},
    year = {2016},
    volume = {529},
    number = {7587},
    pages = {484--489},
}

@article{LevineEtAl2016,
    author = {Levine, S. and Finn, C. and Darrell, T. and Abbeel, P.},
    title = {End-to-End Training of Deep Visuomotor Policies},
    journal = {Journal of Machine Learning Research},
    year = {2016},
    volume = {17},
    number = {39},
    pages = {1--40},
}

@article{BaiEtAl2022,
    title = {Training a Helpful and Harmless Assistant with Reinforcement Learning from Human Feedback},
    author = {Bai, Y. and Jones, A. and Ndousse, K. and Askell, A. and Chen, A. and DasSarma, N. and Drain, D. and Fort, S. and Ganguli, D. and Henighan, T. and Joseph, N. and Kadavath, S. and Kernion, J. and Conerly, T. and El-Showk, S. and Elhage, N. and Hatfield-Dodds, Z. and Hernandez, D. and Hume, T. and Johnston, S. and Kravec, S. and Lovitt, L. and Nanda, N. and Olsson, C. and Amodei, D. and Brown, T. and Clark, J. and McCandlish, S. and Olah, C. and Mann, B. and Kaplan, J.},
    url = {https://arxiv.org/abs/2204.05862},
    year = {2022},
}

@book{stone2013bayesian,
  title={Bayesian multiple target tracking},
  author={Stone, L. D. and Streit, R. L. and Corwin, T. L. and Bell, K. L.},
  year={2013},
  publisher={Artech House}
}

@article{WatkinsDayan1992,
    author = {Watkins, C. J. C. H. and Dayan, P.},
    title = {Q-learning},
    journal = {Machine Learning},
    year = {1992},
    volume = {8},
    number = {3},
    pages = {279--292},
}

@article{Williams1992,
    author = {Williams, R. J.},
    title = {Simple statistical gradient-following algorithms for connectionist reinforcement learning},
    journal = {Machine Learning},
    year = {1992},
    volume = {8},
    number = {3},
    pages = {229--256},
}

@inproceedings{MnihEtAl2016,
    title = {Asynchronous Methods for Deep Reinforcement Learning},
    author = {Mnih, V. and Badia, A. P. and Mirza, M. and Graves, A. and Lillicrap, T. and Harley, T. and Silver, D. and Kavukcuoglu, K.},
    booktitle = {Proceedings of The 33rd International Conference on Machine Learning},
    year = {2016},
    pages = {1928--1937},
}

@book{Murphy2022,
    author = {Murphy, K. P.},
    title = {Probabilistic Machine Learning: An introduction},
    publisher = {MIT Press},
    year = {2022},
}

@article{Sutton1991,
    author = {Sutton, R. S.},
    title = {Dyna, an integrated architecture for learning, planning, and reacting},
    journal = {SIGART Bull.},
    year = {1991},
    volume = {2},
    number = {4},
    pages = {160--163},
}

@article{OsaEtAl2018,
    title = {An Algorithmic Perspective on Imitation Learning},
    author = {Osa, T. and Pajarinen, J. and Neumann, G. and Bagnell, A. and Abbeel, P. and Peters, J.},
    url = {https://arxiv.org/abs/1811.06711},
    year = {2018},
}

@article{AroraEtAl2021,
title = {A survey of inverse reinforcement learning: Challenges, methods and progress},
journal = {Artificial Intelligence},
volume = {297},
pages = {103500},
year = {2021},
author = {Arora, S. and Doshi, P.},
}

@book{hall2001handbook,
  title={Handbook of multisensor data fusion: theory and practice},
  author={Liggins II, M. and Hall, D. and Llinas, J.},
  year={2017},
  publisher={CRC press}
}

@book{liggins2017handbook,
  title={Handbook of multisensor data fusion: theory and practice},
  author={Liggins II, M. and Hall, D. and Llinas, J.},
  year={2017},
  publisher={CRC press}
}

@article{ChernovaEtAl2009,
author = {Chernova, S. and Veloso, M.},
title = {Interactive policy learning through confidence-based autonomy},
year = {2009},
volume = {34},
number = {1},
issn = {1076-9757},
journal = {Journal of Artificial Intelligence Research},
pages = {1–25},
}

@inproceedings{Powell2012,
  title={AI, OR and control theory: A Rosetta Stone for stochastic optimization},
  author={Powell, W. B.},
  booktitle={Princeton University},
  year={2012},
}

@inproceedings{RossEtAl2011,
  title={A Reduction of Imitation Learning and Structured Prediction to No-Regret Online Learning},
  author={Ross, S. and Gordon, G. and Bagnell, D.},
  booktitle={Proceedings of the Fourteenth International Conference on Artificial Intelligence and Statistics},
  pages={627--635},
  year={2011},
}

@inproceedings{EmmonsEtAl2021,
  title={RvS: What is Essential for Offline RL via Supervised Learning?},
  author={Emmons, S. and Eysenbach, B. and Kostrikov, I. and Levine, S.},
  journal={International Conference on Learning Representations},
  year={2021},
}

@article{FischlerBolles1981,
author = {Fischler, M. A. and Bolles, R. C.},
title = {Random sample consensus: a paradigm for model fitting with applications to image analysis and automated cartography},
year = {1981},
publisher = {Association for Computing Machinery},
address = {New York, NY, USA},
volume = {24},
number = {6},
journal = {Commun. ACM},
pages = {381--395},
numpages = {15},
}

@article{zhang1994iterative,
  title={Iterative point matching for registration of free-form curves and surfaces},
  author={Zhang, Z.},
  journal={International journal of computer vision},
  volume={13},
  number={2},
  pages={119--152},
  year={1994},
  publisher={Springer}
}

@article{MontemerloThrunEtAl2002,
  title={FastSLAM: A factored solution to the simultaneous localization and mapping problem},
  author={Montemerlo, M. and Thrun, S. and Koller, D. and Wegbreit, B. and others},
  journal={Aaai/iaai},
  volume={593598},
  number={2},
  pages={593--598},
  year={2002}
}

@inproceedings{MontemerloThrunEtAl2003,
  author    = {Montemerlo, M. and Thrun, S. and Koller, D. and Wegbreit, B.},
  title     = {FastSLAM 2.0: An Improved Particle Filtering Algorithm for Simultaneous Localization and Mapping that Provably Converges},
  booktitle = {Proceedings of the 18th National Conference on Artificial Intelligence (AAAI)},
  year      = {2003},
  pages     = {1151--1156}
}

@incollection{SchmerlingPavone2019,
  author = {Schmerling, E. and Pavone, M.},
  title = {Kinodynamic Planning},
  booktitle = {Encyclopedia of Robotics},
  publisher = {{Springer}},
  edition = {First},
  year = {2019}
}

@inbook{Hauser2020,
    author = {Hauser, K. },
    title = {Motion and Path Planning},
    booktitle = {Encyclopedia of Robotics},
    year = {2020},
    publisher = {Springer},
    pages = {1--11},
}

@book{Rimon1990,
    title={Exact robot navigation using artificial potential functions. PhD Thesis},
    author={Rimon, E.},
    year={1990},
    publisher={Yale University},
}

@inproceedings{VaswaniEtAl2017,
 author = {Vaswani, A. and Shazeer, N. and Parmar, N. and Uszkoreit, J. and Jones, L. and Gomez, A. N. and Kaiser, \L. and Polosukhin, I.},
 booktitle = {Advances in Neural Information Processing Systems},
 pages = {},
 publisher = {Curran Associates, Inc.},
 title = {Attention is All you Need},
 volume = {30},
 year = {2017}
}

@inproceedings{BrownEtAl2020GPT3,
  title={Language Models are Few-Shot Learners},
  author={Brown, T. B. and Mann, B. and Ryder, N. and Subbiah, M. and Kaplan, J. and Dhariwal, P. and Neelakantan, A. and Shyam, P. and Sastry, G. and Askell, A. and Agarwal, S. and Herbert-Voss, A. and Krueger, G. and Henighan, T. and Child, R. and Ramesh, A. and Ziegler, D. M. and Wu, J. and Winter, C. and Hesse, C. and Chen, M. and Sigler, E. and Litwin, M. and Gray, S. and Chess, B. and Clark, J. and Berner, C. and McCandlish, S. and Radford, A. and Sutskever, I. and Amodei, D.},
  booktitle={Advances in Neural Information Processing Systems},
  pages={1877--1901},
  year={2020},
}

@article{RadfordEtAl2019GPT2,
  title={Language Models are Unsupervised Multitask Learners},
  author={Radford, A. and Wu, J. and Child, R. and Luan, D. and Amodei, D. and Sutskever, I.},
  year={2019}
}

@inproceedings{RadfordEtAl2021CLIP,
  title={Learning Transferable Visual Models From Natural Language Supervision},
  author={Radford, A. and Kim, J. W. and Hallacy, C. and Ramesh, A. and Goh, G. and Agarwal, S. and Sastry, G. and Askell, A. and Mishkin, P. and Clark, J. and Krueger, G. and Sutskever, I.},
  booktitle={Proceedings of the 38th International Conference on Machine Learning},
  pages={8748--8763},
  year={2021},
}

@article{RivesEtAl2021ESM,
  author = {Rives, A. and Meier, J. and Sercu, T. and Goyal, S. and Lin, Z. and Liu, J. and Guo, D. and Ott, M. and Zitnick, C. L. and Ma, J. and Fergus, R.},
  title = {Biological structure and function emerge from scaling unsupervised learning to 250 million protein sequences},
  year = {2021},
  volume = {118},
  number = {15},
  journal = {Proceedings of the National Academy of Sciences},
}

@article{BommasaniEtAl2021FoundationModels,
  title={On the Opportunities and Risks of Foundation Models},
  author={Bommasani, R. and Hudson, D. A. and Adeli, E. and Altman, R. and Arora, S. and others},
  journal={arXiv preprint arXiv:2108.07258},
  year={2021},
}

@article{HoEtAl2022ImagenVideo,
  title={Imagen Video: High Definition Video Generation with Diffusion Models},
  author={Ho, J. and Chan, W. and Saharia, C. and Whang, J. and Gao, R. and Gritsenko, A. and Kingma, D. P. and Poole, B. and Norouzi, M. and Fleet, D. J. and Salimans, T.},
  year={2022},
  journal={arXiv preprint arXiv:2210.02303},
}

@inproceedings{ZitkovichEtAl2023RT2,
  title={RT-2: Vision-Language-Action Models Transfer Web Knowledge to Robotic Control},
  author={Zitkovich, B. and Yu, T. and Xu, S. and Xu, P. and Xiao, T. and Xia, F. and Wu, J. and Wohlhart, P. and Welker, S. and Wahid, A. and others},
  booktitle={Proceedings of The 7th Conference on Robot Learning},
  pages={2165--2183},
  year={2023},
}

@article{BlackEtAl2024Pi0,
  title={$\\pi$0: A Vision-Language-Action Flow Model for General Robot Control},
  author={Black, K. and Brown, N. and Driess, D. and Esmail, A. and Equi, M. and Finn, C. and Fusai, N. and Groom, L. and Hausman, K. and Ichter, B. and Jakubczak, S. and Jones, T. and Ke, L. and Levine, S. and Li-Bell, A. and Mothukuri, M. and Nair, S. and Pertsch, K. and Shi, L. X. and Tanner, J. and Vuong, Q. and Walling, A. and Wang, H. and Zhilinsky, U.},
  booktitle={arXiv preprint arXiv:2410.24164},
  year={2024},
}

@inproceedings{KimEtAl2025OpenVLA,
  title={OpenVLA: An Open-Source Vision-Language-Action Model},
  author={Kim, M. J. and Pertsch, K. and Karamcheti, S. and Xiao, T. and Balakrishna, A. and Nair, S. and Rafailov, R. and Foster, E. P. and Sanketi, P. R. and Vuong, Q. and Kollar, T. and Burchfiel, B. and Tedrake, R. and Sadigh, D. and Levine, S. and Liang, P. and Finn, C.},
  booktitle={Proceedings of The 8th Conference on Robot Learning},
  pages={2679--2713},
  year={2025},
}

@inproceedings{HuangEtAl2023InnerMonologue,
  title={Inner Monologue: Embodied Reasoning through Planning with Language Models},
  author={Huang, W. and Xia, F. and Xiao, T. and Chan, H. and Liang, J. and Florence, P. and Zeng, A. and Tompson, J. and Mordatch, I. and Chebotar, Y. and Sermanet, P. and Jackson, T. and Brown, N. and Luu, L. and Levine, S. and Hausman, K.},
  booktitle={Proceedings of The 6th Conference on Robot Learning},
  pages={1769--1782},
  year={2023},
}

@article{NVIDIAEtAl2025AlpamayoR1,
  author={{NVIDIA}},
  title={Alpamayo-R1: Bridging Reasoning and Action Prediction for Generalizable Autonomous Driving in the Long Tail},
  year={2025},
  journal={arXiv preprint arXiv:2511.00088},
}

@misc{NVIDIAHalos2024,
  author       = {{NVIDIA}},
  title        = {NVIDIA Halos: Autonomous Vehicle Safety},
  year         = {2024},
  url          = {https://www.nvidia.com/en-us/ai-trust-center/halos/autonomous-vehicles/},
}

@article{HaSchmidhuber2018WorldModels,
  title={World Models},
  author={Ha, D. and Schmidhuber, J.},
  year={2018},
  journal={arXiv preprint arXiv:1803.10122},
}

@article{NVIDIAEtAl2025Cosmos,
  author={{NVIDIA}},
  title={Cosmos World Foundation Model Platform for Physical AI},
  year={2025},
  journal={arXiv preprint arXiv:2501.03575},
}

@inproceedings{ZawalskiEtAl2024ECOT,
  title={Robotic Control via Embodied Chain-of-Thought Reasoning},
  author={Zawalski, M. and Chen, W. and Pertsch, K. and Mees, O. and Finn, C. and Levine, S.},
  booktitle={Proceedings of The 8th Conference on Robot Learning},
  year={2024},
}

@misc{DeepMind2025Genie3,
  title={Genie 3: A New Frontier for World Models},
  author={{Google DeepMind}},
  year={2025},
  note={Research blog},
  url={https://deepmind.google/blog/genie-3-a-new-frontier-for-world-models/},
}

@article{AssranEtAl2025VJEPA2,
  author={Assran, M. and Bardes, A. and Fan, D. and Garrido, Q. and Howes, R. and Komeili, M. and Muckley, M. and Rizvi, A. and Roberts, C. and Sinha, K. and others},
  title={V-JEPA 2: Self-Supervised Video Models Enable Understanding, Prediction and Planning},
  year={2025},
  journal={arXiv preprint arXiv:2506.09985},
}

@inproceedings{LocatelloEtAl2020SlotAttention,
  title={Object-Centric Learning with Slot Attention},
  author={Locatello, F. and Weissenborn, D. and Unterthiner, T. and Mahendran, A. and Neumann, G. and R{\"a}tsch, M. and Kohli, P. and Botvinick, M. and Bachem, O.},
  booktitle={Advances in Neural Information Processing Systems 33},
  pages={11525--11538},
  year={2020},
}

@inproceedings{MakoviychukEtAl2021IsaacGym,
  title={Isaac Gym: High Performance GPU Based Physics Simulation For Robot Learning},
  author={Makoviychuk, V. and Wawrzyniak, L. and Guo, Y. and Lu, M. and Storey, K. and Macklin, M. and Hoeller, D. and Rudin, N. and Allshire, A. and Handa, A. and others},
  booktitle={Proceedings of the Neural Information Processing Systems Track on Datasets and Benchmarks},
  year={2021},
}

@inproceedings{TaylorEtAl2020LearningCBF,
  title={Learning for Safety-Critical Control with Control Barrier Functions},
  author={Taylor, A. and Singletary, A. and Yue, Y. and Ames, A.},
  booktitle={Learning for Dynamics and Control},
  pages={708--717},
  year={2020},
}

@inproceedings{SinhaElhafsiEtAl2024,
  title={Real-Time Anomaly Detection and Reactive Planning with Large Language Models},
  author={Sinha, R. and Elhafsi, A. and Agia, C. and Foutter, M. and Schmerling, E. and Pavone, M.},
  booktitle={Proceedings of Robotics: Science and Systems},
  year={2024},
}

@inproceedings{SagawaEtAl2020DRO,
  title={Distributionally Robust Neural Networks for Group Shifts: On the Importance of Regularization for Worst-Case Generalization},
  author={Sagawa, S. and Koh, P. W. and Hashimoto, T. B. and Liang, P.},
  booktitle={Proceedings of the International Conference on Learning Representations},
  year={2020},
}

@inproceedings{HoffmanEtAl2018CyCADA,
  title={CyCADA: Cycle-Consistent Adversarial Domain Adaptation},
  author={Hoffman, J. and Tzeng, E. and Park, T. and Zhu, J.-Y. and Isola, P. and Saenko, K. and Efros, A. A. and Darrell, T.},
  booktitle={Proceedings of the 35th International Conference on Machine Learning},
  pages={1994--2003},
  year={2018},
}

@inproceedings{DosovitskiyEtAl2021,
title={An Image is Worth 16x16 Words: Transformers for Image Recognition at Scale},
author={Dosovitskiy, A. and Beyer, L. and Kolesnikov, A. and Weissenborn, D. and Zhai, X. and Unterthiner, T. and Dehghani, M. and Minderer, M. and Heigold, G. and Gelly, S. and Uszkoreit, J. and Houlsby, N.},
booktitle={International Conference on Learning Representations},
year={2021},
}

@inproceedings{QiEtAl2017a,
  title={Pointnet: Deep learning on point sets for 3d classification and segmentation},
  author={Qi, C. R. and Su, H. and Mo, K. and Guibas, L. J.},
  booktitle={Proceedings of the IEEE conference on computer vision and pattern recognition},
  pages={652--660},
  year={2017}
}

@article{QiEtAl2017b,
  title={Pointnet++: Deep hierarchical feature learning on point sets in a metric space},
  author={Qi, C. R. and Yi, L. and Su, H. and Guibas, L. J.},
  journal={Advances in neural information processing systems},
  volume={30},
  year={2017}
}

@article{WangEtAl2019,
  title={Dynamic graph cnn for learning on point clouds},
  author={Wang, Y. and Sun, Y. and Liu, Z. and Sarma, S. E. and Bronstein, M. M. and Solomon, J. M.},
  journal={ACM Transactions on Graphics (tog)},
  volume={38},
  number={5},
  pages={1--12},
  year={2019},
  publisher={Acm New York, NY, USA}
}

@inproceedings{ZhouTuzel2018,
  title={Voxelnet: End-to-end learning for point cloud based 3d object detection},
  author={Zhou, Y. and Tuzel, O.},
  booktitle={Proceedings of the IEEE conference on computer vision and pattern recognition},
  pages={4490--4499},
  year={2018}
}

@inproceedings{LangEtAl2019,
  title={Pointpillars: Fast encoders for object detection from point clouds},
  author={Lang, A. H. and Vora, S. and Caesar, H. and Zhou, L. and Yang, J. and Beijbom, O.},
  booktitle={Proceedings of the IEEE/CVF conference on computer vision and pattern recognition},
  pages={12697--12705},
  year={2019}
}

@article{YanEtAl2018,
  title={Second: Sparsely embedded convolutional detection},
  author={Yan, Y. and Mao, Y. and Li, B.},
  journal={Sensors},
  volume={18},
  number={10},
  pages={3337},
  year={2018},
  publisher={Multidisciplinary Digital Publishing Institute}
}

@inproceedings{Girshick2015,
  author={Girshick, R.},
  booktitle={2015 IEEE International Conference on Computer Vision (ICCV)}, 
  title={Fast R-CNN}, 
  year={2015},
  volume={},
  number={},
  pages={1440-1448},
  doi={10.1109/ICCV.2015.169}}

@inproceedings{ShiEtAl2019,
    author = {Shi, S. and Wang, X. and Li, H.},
    title = {PointRCNN: 3D Object Proposal Generation and Detection From Point Cloud},
    booktitle = {The IEEE Conference on Computer Vision and Pattern Recognition (CVPR)},
    month = {6},
    year = {2019}
}

@inproceedings{HeEtAl2017,
  title={Mask R-CNN},
  author={He, K. and Gkioxari, G. and Doll{\'a}r, P. and Girshick, R.},
  booktitle={Proceedings of the IEEE International Conference on Computer Vision},
  pages={2961--2969},
  year={2017}
}

@inproceedings{RedmonEtAl2016,
author = {Redmon, J. and Divvala, S. and Girshick, R. and Farhadi, A.},
title = {You Only Look Once: Unified, Real-Time Object Detection},
booktitle = {Proceedings of the IEEE Conference on Computer Vision and Pattern Recognition (CVPR)},
month = {6},
year = {2016}
}

@article{RonnebergerFischerBrox2015,
  author       = {Ronneberger, O. and Fischer, P. and Brox, T.},
  title        = {U-Net: Convolutional Networks for Biomedical Image Segmentation},
  journal      = {CoRR},
  volume       = {abs/1505.04597},
  year         = {2015},
  url          = {http://arxiv.org/abs/1505.04597},
  eprinttype    = {arXiv},
  eprint       = {1505.04597},
  bibsource    = {dblp computer science bibliography, https://dblp.org}
}

@inproceedings{Lozano1990,
author = {Lozano Perez, T.},
title = {Spatial planning: a configuration space approach},
booktitle = {Autonomous Robot Vehicles},
year = {1990}
}

@article{LuMilios1997,
  title={Robot pose estimation in unknown environments by matching 2d range scans},
  author={Lu, F. and Milios, E.},
  journal={Journal of Intelligent and Robotic systems},
  volume={18},
  number={3},
  pages={249--275},
  year={1997},
  publisher={Springer}
}

@inproceedings{KummerleEtAl2011,
  title={g 2 o: A general framework for graph optimization},
  author={K{\"u}mmerle, R. and Grisetti, G. and Strasdat, H. and Konolige, K. and Burgard, W.},
  booktitle={2011 IEEE international conference on robotics and automation},
  pages={3607--3613},
  year={2011},
  organization={IEEE}
}

@inproceedings{Stentz1995,
  title={The focussed D* algorithm for real-time replanning},
  author={Stentz, A.},
  booktitle={14th International Joint Conference on Artificial Intelligence},
  pages={1652–1659},
  year={1995}
}

@article{dellaert2012factor,
  title={Factor graphs and GTSAM: A hands-on introduction},
  author={Dellaert, F.},
  journal={Georgia Institute of Technology, Tech. Rep},
  volume={2},
  number={4},
  year={2012}
}

@article{chaplot2020learning,
  title={Learning to explore using active neural slam},
  author={Chaplot, D. S. and Gandhi, D. and Gupta, S. and Gupta, A. and Salakhutdinov, R.},
  journal={arXiv preprint arXiv:2004.05155},
  year={2020}
}

@article{KaessRanganathanDellaert2008,
  title={iSAM: Incremental smoothing and mapping},
  author={Kaess, M. and Ranganathan, A. and Dellaert, F.},
  journal={IEEE Transactions on Robotics},
  volume={24},
  number={6},
  pages={1365--1378},
  year={2008},
  publisher={IEEE}
}

@inproceedings{dellaert1999monte,
  title={Monte carlo localization for mobile robots},
  author={Dellaert, F. and Fox, D. and Burgard, W. and Thrun, S.},
  booktitle={Proceedings 1999 IEEE international conference on robotics and automation (Cat. No. 99CH36288C)},
  volume={2},
  pages={1322--1328},
  year={1999},
  organization={IEEE}
}

@book{slam-handbook,
  title        = {{SLAM Handbook.} From Localization and Mapping to Spatial Intelligence},
  editor       = {Carlone, L. and Kim, A. and Barfoot, T. and Cremers, D. and Dellaert, F.},
  publisher    = {Cambridge University Press},
  year         = {2026}
}

@article{DellaertKaess2006,
  title={Square root SAM: Simultaneous localization and mapping via square root information smoothing},
  author={Dellaert, F. and Kaess, M.},
  journal={The International Journal of Robotics Research},
  volume={25},
  number={12},
  pages={1181--1203},
  year={2006},
  publisher={Sage Publications Sage CA: Thousand Oaks, CA}
}

@article{forster2016manifold,
  title={On-manifold preintegration for real-time visual--inertial odometry},
  author={Forster, C. and Carlone, L. and Dellaert, F. and Scaramuzza, D.},
  journal={IEEE Transactions on Robotics},
  volume={33},
  number={1},
  pages={1--21},
  year={2016},
  publisher={IEEE}
}

@article{kaess2012isam2,
  title={iSAM2: Incremental smoothing and mapping using the Bayes tree},
  author={Kaess, M. and Johannsson, H. and Roberts, R. and Ila, V. and Leonard, J. J. and Dellaert, F.},
  journal={The International Journal of Robotics Research},
  volume={31},
  number={2},
  pages={216--235},
  year={2012},
  publisher={Sage Publications Sage UK: London, England}
}

@book{SicilianoEtAl2008,
    title={Robotics: Modelling, Planning and Control},
    author={Siciliano, B. and Sciavicco, L. and Villani, L. and Oriolo, G.},
    year={2008},
    publisher={Springer Publishing Company, Incorporated},
}

@book{RawlingsEtAl2017,
    title={Model Predictive Control: Theory, Computation, and Design},
    author={Rawlings, J. and Mayne, D. Q. and Diehl, M.},
    year={2017},
    publisher={Nob Hill Publishing},
}

@book{BorrelliEtAl2017,
    title={Predictive Control for Linear and Hybrid Systems},
    author={Borrelli, F. and Bemporad, A. and Morari, M.},
    year={2017},
    publisher={Cambridge University Press},
}

@book{SicilianoEtAl2008Kinematics,
    title={Robotics: Modelling, Planning and Control},
    author={Siciliano, B. and Sciavicco, L. and Villani, L. and Oriolo, G.},
    year={2008},
    publisher={Springer Publishing Company, Incorporated},
    chapter={2}
}

@book{SicilianoEtAl2008Lagrange,
    title={Robotics: Modelling, Planning and Control},
    author={Siciliano, B. and Sciavicco, L. and Villani, L. and Oriolo, G.},
    year={2008},
    publisher={Springer Publishing Company, Incorporated},
    chapter={7}
}

@book{SicilianoEtAl2007,
    title={Springer Handbook of Robotics},
    author={Siciliano, B. and Khatib, O.},
    year={2007},
    publisher={Springer-Verlag},
}

@book{LynchEtAl2017Lagrange,
    title={Modern Robotics: Mechanics, Planning, and Control},
    author={Lynch, K. M. and Park, K. C.},
    year={2017},
    publisher={Cambridge University Press},
    chapter={8}
}

@article{Shuster1981,
author = {Shuster, M. D.},
title = {Survey of attitude representations},
year = {1993},
volume = {41},
number = {4},
journal = {Journal of the Astronautical Sciences},
pages = {439-517},
}

@article{Rao2010,
author = {Rao, A.},
title = {A Survey of Numerical Methods for Optimal Control},
year = {2010},
volume = {135},
journal = {Advances in the Astronautical Sciences}
}

@article{Kelly2017,
author = {Kelly, M.},
title = {An Introduction to Trajectory Optimization: How to Do Your Own Direct Collocation},
year = {2017},
volume = {59},
number = {4},
pages = {849-904},
journal = {SIAM Review}
}

@article{Hertling1970,
author = {Hertling, J.},
title = {Numerical Methods for Two-Point Boundary Value Problems (Herbert B. Keller)},
year = {1970},
volume = {12},
number = {2},
pages = {313-315},
journal = {SIAM Review}
}

@article{AscherRussell1981,
author  = {Ascher, U. M. and Russell, R. D.},
title   = {Reformulation of boundary value problems into ``standard'' form},
journal = {SIAM Review},
year    = {1981},
volume  = {23},
number  = {2},
pages   = {238--254}
}

@article{MnihEtAl2013,
  title={Playing Atari with Deep Reinforcement Learning},
  author={Mnih, V. and Kavukcuoglu, K. and Silver, D. and Graves, A. and Antonoglou, I. and Wierstra, D. and Riedmiller, M. A.},
  journal={ArXiv},
  year={2013},
  volume={abs/1312.5602},
  url={https://api.semanticscholar.org/CorpusID:15238391}
}

\end{document}